\documentclass[a4paper, 10pt, oneside]{memoir}
\usepackage{style}

\usepackage{subfiles}

\title{
\vspace{4cm}
\normalfont \normalsize 
\horrule{0.5pt} \\[0.4cm]
\huge Конспект по обучению с подкреплением
\horrule{2pt} \\[0.5cm]
}

\author{qbrick@mail.ru}

\date{\normalsize\today}

\begin{document}
\mathversion{bold}

\maketitle
\thispagestyle{empty}

\vspace{-6.3cm}
\begin{adjustwidth}{0.5cm}{}
    \includegraphics[width=0.07\textwidth]{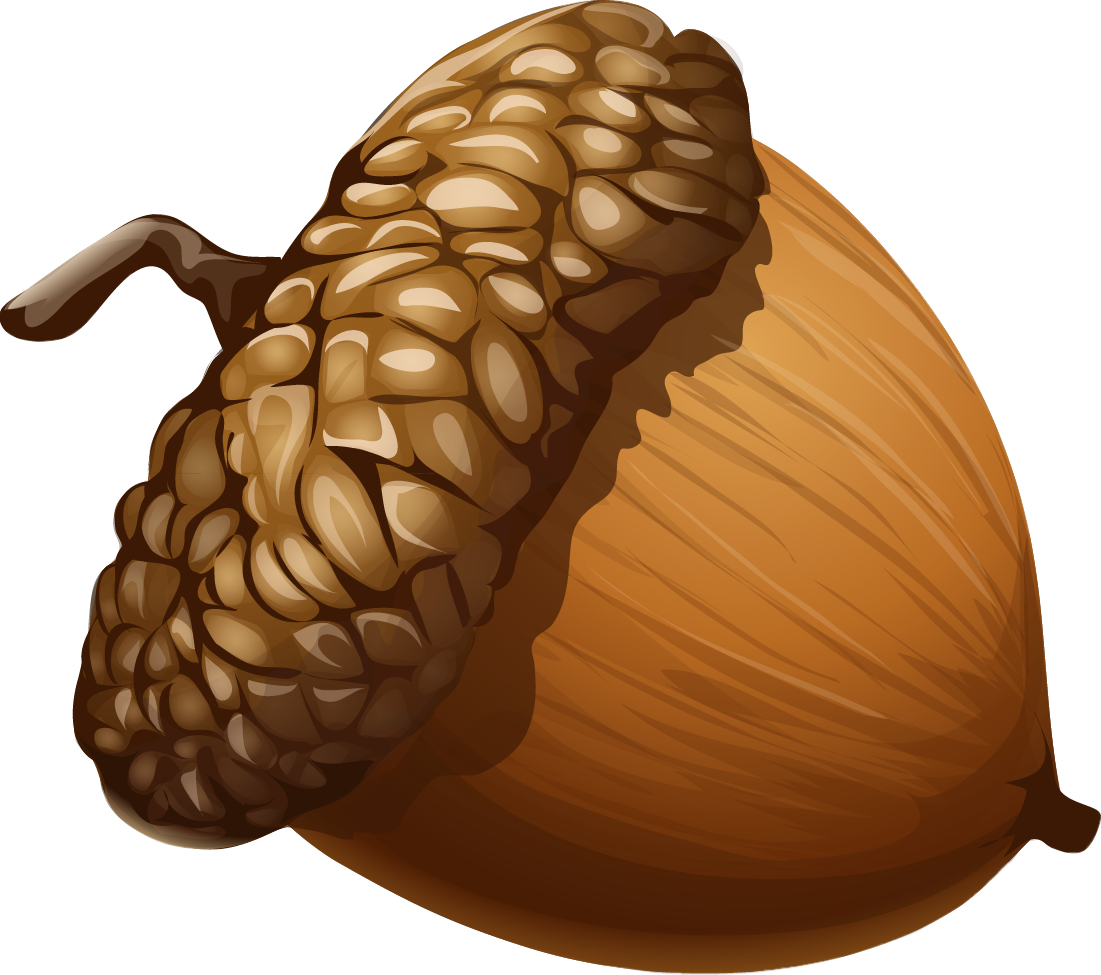}
\end{adjustwidth}
\vspace{4cm}
\begin{center}
    \includegraphics[width=0.3\textwidth]{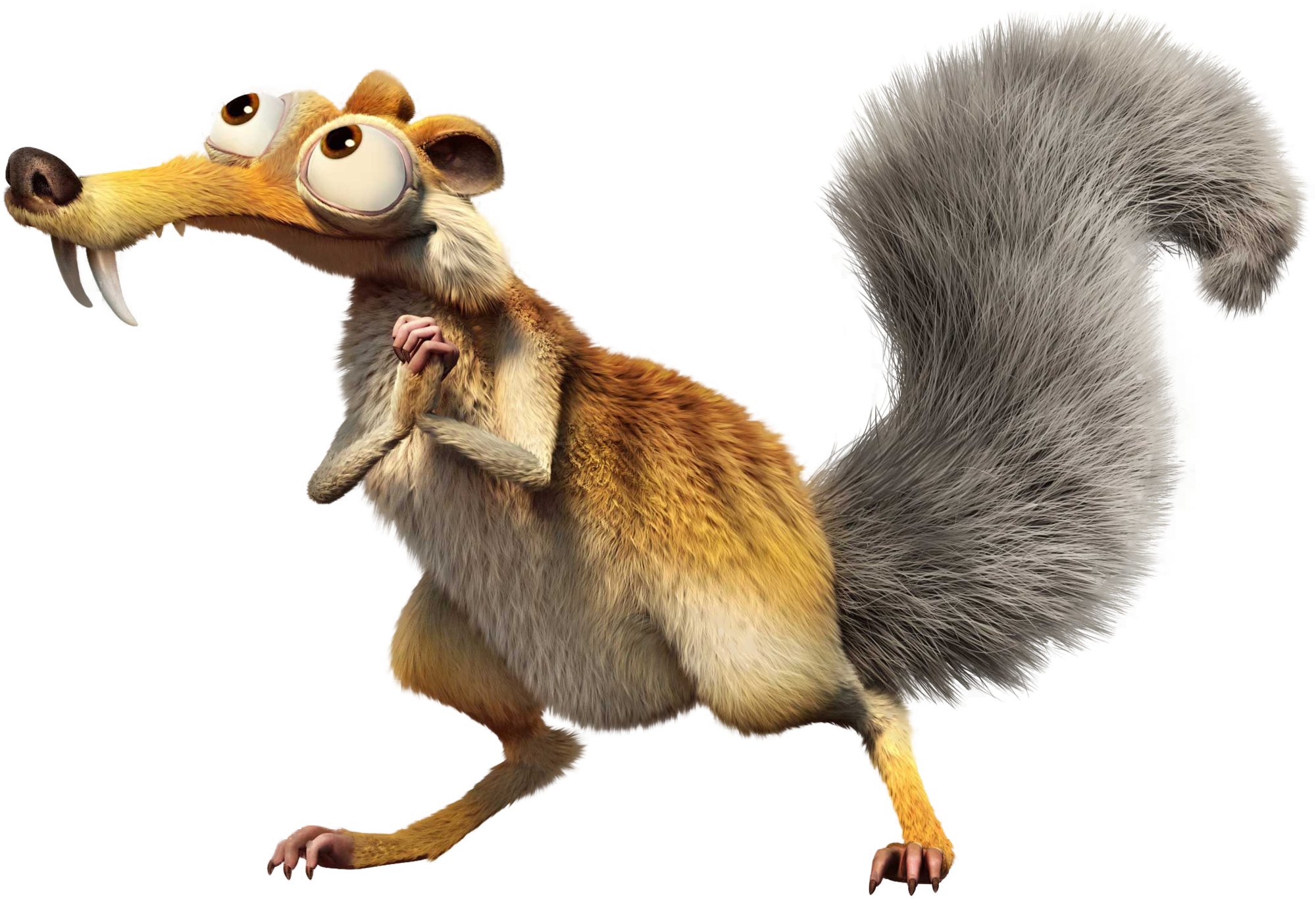}
\end{center}

\vspace{2.2cm}
\begin{center}
\textcolor{ChadBlue}{\underline{\textbf{Аннотация}}}
\end{center}

\vspace{0.5cm}
Современные алгоритмы глубокого обучения с подкреплением способны решать задачи искусственного интеллекта методом проб и ошибок без использования каких-либо априорных знаний о решаемой задаче. В этом конспекте собраны принципы работы основных алгоритмов, достигших прорывных результатов во многих задачах от игрового искусственного интеллекта до робототехники. Вся необходимая теория приводится с доказательствами, использующими единый ход рассуждений, унифицированные обозначения и определения. Основная задача этой работы --- не только собрать информацию из разных источников в одном месте, но понять разницу между алгоритмами различного вида и объяснить, почему они выглядят именно так, а не иначе.

\vspace{0.3cm}
Предполагается знакомство читателя с основами машинного обучения и глубокого обучения. Об ошибках и опечатках в тексте можно сообщать в \href{https://github.com/FortsAndMills/RL-Theory-book}{репозитории проекта}.






\newpage
\tableofcontents*


\chapter{Задача обучения с подкреплением}

Ясно, что обучение с учителем это не та модель <<обучения>>, которая свойственна интеллектуальным сущностям. Термин <<обучение>> здесь подменяет понятие интерполяции или, если уж на то пошло, построение алгоритмов неявным образом (<<смотрите, оно само обучилось, я сам ручками не прописывал, как кошечек от собачек отличать>>). К полноценному обучению, на которое способен не только человек, но и в принципе живые организмы, задачи классического машинного обучения имеют лишь косвенное отношение. Значит, нужна другая формализация понятия <<задачи, требующей интеллектуального решения>>, в которой обучение будет проводится не на опыте, заданном прецедентно, в виде обучающей выборки.

\begin{wrapfigure}{r}{0.4\linewidth}
\centering
\includegraphics[width=0.4\textwidth]{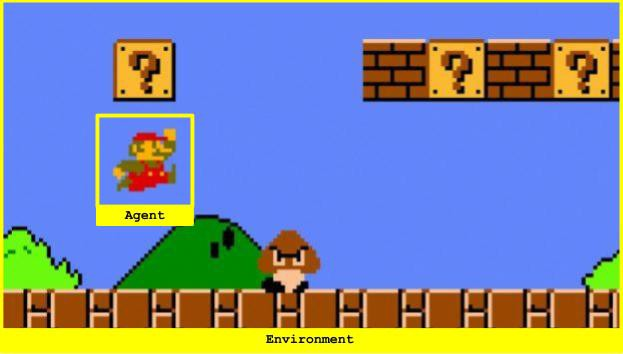}
\end{wrapfigure}

Термин \emph{подкрепление} (reinforcement) пришёл из \href{https://ru.wikipedia.org/wiki/\%D0\%91\%D0\%B8\%D1\%85\%D0\%B5\%D0\%B2\%D0\%B8\%D0\%BE\%D1\%80\%D0\%B8\%D0\%B7\%D0\%BC}{поведенческой психологии} и обозначает награду или наказание за некоторый получившийся результат, зависящий не только от самих принятых решений, но и внешних, не обязательно подконтрольных, факторов. Под обучением здесь понимается поиск способов достичь желаемого результата \emph{методом проб и ошибок} (trial and error), то есть попыток решить задачу и использование накопленного опыта для усовершенствования своей стратегии в будущем.

В данной главе будут введены основные определения и описана формальная постановка задачи. Под желаемым результатом мы далее будем понимать максимизацию некоторой скалярной величины, называемой \emph{наградой} (reward). Интеллектуальную сущность (систему/робота/алгоритм), принимающую решения, будем называть \emph{агентом} (agent). Агент взаимодействует с \emph{миром} (world) или \emph{средой} (environment), которая задаётся зависящим от времени \emph{состоянием} (state). Агенту в каждый момент времени в общем случае доступно только некоторое \emph{наблюдение} (observation) текущего состояния мира. Сам агент задаёт процедуру выбора \emph{действия} (action) по доступным наблюдениям; эту процедуру далее будем называть \emph{стратегией} или \emph{политикой} (policy). Процесс взаимодействия агента и среды задаётся \emph{динамикой среды} (world dynamics), определяющей правила смены состояний среды во времени и генерации награды.

Буквы $s$, $a$, $r$ зарезервируем для состояний, действий и наград соответственно; буквой $t$ будем обозначать время в процессе взаимодействия. 

\section{Модель взаимодействия агента со средой}

\subsection{Связь с оптимальным управлением}

Математика поначалу пришла к очень похожей формализации следующим образом. Для начала, нужно ввести какую-то модель мира; в рамках \href{https://ru.wikipedia.org/wiki/\%D0\%94\%D0\%B5\%D0\%BC\%D0\%BE\%D0\%BD_\%D0\%9B\%D0\%B0\%D0\%BF\%D0\%BB\%D0\%B0\%D1\%81\%D0\%B0}{лапласовского детерминизма} можно предположить, что положение, скорости и ускорения всех атомов вселенной задают её текущее состояние, а изменение состояния происходит согласно некоторым дифференциальным уравнениям:
$$\dot{s} = f(s, t)$$

Коли агент может как-то взаимодействовать со средой или влиять на неё, мы можем промоделировать взаимодействие следующим образом: скажем, что в каждый момент времени $t$ агент в зависимости от текущего состояния среды $s$ выбирает некоторое действие (<<управление>>) $a(s, t)$:
$$\dot{s} = f(s, a(s, t), t)$$

Для моделирования награды положим, что в каждый момент времени агент получает наказание или штраф (cost)\footnote{теория оптимального управления строилась в СССР в формализме минимизации штрафов (потерь, убытков и наказаний), когда построенная в США теория обучения с подкреплением --- в формализме максимизации награды (прибыли, счастья, тортиков). Само собой, формализмы эквивалентны, и любой штраф можно переделать в награду домножением на минус, и наоборот.} в объёме $L(s, a(s, t), t)$. Итоговой наградой агента полагаем суммарную награду, то есть интеграл по времени:
$$
\begin{cases}
-\int\limits L(s, a(s, t), t) \diff t \to \max\limits_{a(s, t)} \\
\dot{s} = f(s, a(s, t), t)
\end{cases}
$$

За исключением того, что для полной постановки требуется задать ещё начальные и конечные условия, поставленная задача является общей формой задачи \emph{оптимального управления} (optimal control). При этом обратим внимание на сделанные при постановке задачи предположения:
\begin{itemize}
    \item время непрерывно;
    \item мир детерминирован;
    \item мир нестационарен (функция $f$ напрямую зависит от времени $t$);
    \item модель мира предполагается известной, то есть функция $f$ задана явно в самой постановке задачи;
\end{itemize}

Принципиально последнее: теория рассматривает способы для заданной системы дифференциальных уравнений поиска оптимальных $a(s, t)$ аналитически. Проводя аналогию с задачей максимизации некоторой, например, дифференцируемой функции $f(x) \to \max\limits_x$, теория оптимального управления ищет необходимые условия решения: например, что в экстремуме функции обязательно $\nabla_x f(x) = 0$. Никакого <<обучения методом проб и ошибок>> здесь не предполагается.

В обучении с подкреплением вместо поиска решения аналитически мы будем строить итеративные методы оптимизации, искать аналоги, например, градиентного спуска. В такой процедуре неизбежно появится цепочка приближений решений: мы начнём с какой-то функции, выбирающей действия, возможно, не очень удачной и не способной набрать большую награду, но с ходом работы алгоритма награда будет оптимизироваться и способ выбора агентом действий будет улучшаться. Так будет выглядеть обучение.

Также RL исходит из других предположений: время дискретно, а среда --- стохастична, но стационарна. В частности, в рамках этой теории можно построить алгоритмы для ситуации, когда модель мира агенту неизвестна, и единственный способ поиска решений --- обучение на собственном опыте взаимодействия со средой.  

\subsection{Марковская цепь}

В обучении с подкреплением мы зададим модель мира следующим образом: будем считать, что существуют некоторые <<законы физики>>, возможно, стохастические, которые определяют следующее состояние среды по предыдущему. При этом предполагается, что в текущем состоянии мира содержится вся необходимая информация для выполнения перехода, или, иначе говоря, выполняется \emph{свойство Марковости} (Markov property): процесс зависит только от текущего состояния и не зависит от всей предыдущей истории.

\begin{definition} 
\emph{Марковской цепью (Markov chain)} называется пара $(\St, \Trans)$, где: 
\begin{itemize}
    \item $\St$ --- множество состояний.
    \item $\Trans$ --- вероятности переходов $\{p(s_{t+1} \HM\mid s_t) \HM\mid t \HM\in \{0, 1, \dots \}, s_t, s_{t+1} \HM\in \St\}$.
\end{itemize}
\end{definition}

Если дополнительно задать стартовое состояние $s_0$, можно рассмотреть процесс, заданный марковской цепью. В момент времени $t \HM= 0$ мир находится в состоянии $s_0$; сэмплируется случайная величина $s_1 \HM\sim p(s_1 \HM\mid s_0)$, и мир переходит в состояние $s_1$; сэмплируется $s_2 \HM\sim p(s_2 \HM\mid s_1)$, и так далее до бесконечности.

Мы далее делаем важное предположение, что законы мира не изменяются с течением времени. В аналогии с окружающей нас действительностью это можно интерпретировать примерно как <<физические константы не изменяются со временем>>.\footnote{вопрос на засыпку для любителей философии: а это вообще правда? Вдруг там через миллион лет гравитационная постоянная или скорость света уже будут другими в сорок втором знаке после запятой?..}

\begin{definition} 
Марковская цепь называется \emph{однородной} (time-homogeneous) или \emph{стационарной} (stationary), если вероятности переходов не зависят от времени:
\begin{equation*}
\forall t \colon p(s_{t+1} \mid s_t) = p(s_1 \mid s_0)
\end{equation*}
\end{definition}

По определению, переходы $\Trans$ стационарных марковских цепей задаются единственным условным распределением $p(s' \HM\mid s)$. Апостроф $'$ канонично используется для обозначения <<следующих>> моментов времени, мы будем активно пользоваться этим обозначением.

\begin{exampleBox}[righthand ratio=0.35, sidebyside, sidebyside align=center, lower separated=false]{Конечные марковские цепи}
Марковские цепи с конечным числом состояний можно задать при помощи ориентированного графа, дугам которого поставлены в соответствие вероятности переходов (отсутствие дуги означает нулевую вероятность перехода). 

На рисунке приведён пример стационарной марковской цепи с 4 состояниями. Такую цепь можно также задать в виде матрицы переходов:
\begin{center}
\begin{tabular}{c|cccc}
 & \colorsquare{black} & \colorsquare{ChadRed} & \colorsquare{ChadPurple} & \colorsquare{ChadBlue} \\
 \hline
\colorsquare{black}      &      & 0.5 & 0.5 &      \\ 
\colorsquare{ChadRed}    & 0.25 &     &     & 0.75 \\ 
\colorsquare{ChadPurple} & 0.1  & 0.2 & 0.1 & 0.6  \\ 
\colorsquare{ChadBlue}   &      & 0.5 &     & 0.5  \\
\end{tabular}
\end{center}

\tcblower
\vspace{-0.3cm}
\animategraphics[controls, width=\linewidth]{1}{Images/MC/MC-}{1}{15}
\end{exampleBox}

\subsection{Среда}

Как и в оптимальном управлении, для моделирования влияния агента на среду в вероятности переходов достаточно добавить зависимость от выбираемых агентом действий. Итак, наша модель среды --- это <<управляемая>> марковская цепь.

\begin{definition} 
\emph{Средой} (environment) называется тройка $(\St, \A, \Trans)$, где: 
\begin{itemize}
    \item $\St$ --- \emph{пространство состояний} (state space), некоторое множество.
    \item $\A$ --- \emph{пространство действий} (action space), некоторое множество.
    \item $\Trans$ --- \emph{функция переходов} (transition function) или \emph{динамика среды} (world dynamics): вероятности $p(s' \HM\mid s, a)$.
\end{itemize}
\end{definition}

В таком определении среды заложена марковость (независимость переходов от истории) и стационарность (независимость от времени). Время при этом дискретно, в частности, нет понятия <<времени принятия решения агентом>>: среда, находясь в состоянии $s$, ожидает от агента действие $a \HM\in \A$, после чего совершает шаг, сэмплируя следующее состояние $s' \HM\sim p(s' \HM\mid s, a)$.

По дефолту, среда также предполагается \emph{полностью наблюдаемой} (fully observable): агенту при выборе $a_t$ доступно всё текущее состояние $s_t$ в качестве входа. Иными словами, в рамках данного предположения понятия состояния и наблюдения для нас будут эквивалентны. Мы будем строить теорию в рамках этого упрощения; если среда не является полностью наблюдаемой, задача существенно усложняется, и необходимо переходить к формализму \emph{частично наблюдаемых MDP} (partially observable MDP, PoMDP). Обобщение алгоритмов для PoMDP будет рассмотрено отдельно в разделе \ref{sec:PoMDP}.

\begin{exampleBox}[righthand ratio=0.5, sidebyside, sidebyside align=center, lower separated=false]{Конечные среды}
Среды с конечным числом состояний и конечным числом действий можно задать при помощи ориентированного графа, где для каждого действия задан свой комплект дуг.

На рисунке приведён пример среды с 2 действиями $\A = \bigl\{ \text{\colorsquare{ChadBlue}}, \text{\colorsquare{ChadRed}} \bigr\} $; дуги для разных действий различаются цветом. Возле каждого состояния выписана стратегия агента.

\tcblower
\animategraphics[controls, width=\linewidth]{1}{Images/ENV/Environment-}{1}{15}
\end{exampleBox}

\begin{exampleBox}[righthand ratio=0.15, sidebyside, sidebyside align=center, lower separated=false]{Кубик-Рубик}
Пространство состояний --- пространство конфигураций Кубика-Рубика. Пространство действий состоит из 12 элементов (нужно выбрать одну из 6 граней, каждую из которых можно повернуть двумя способами). Следующая конфигурация однозначно определяется текущей конфигурацией и действием, соответственно среда Кубика-Рубика \emph{детерминирована}: задаётся вырожденным распределением, или, что тоже самое,  обычной детерминированной функцией $s' = f(s, a)$. 

\tcblower
\includegraphics[width=\textwidth]{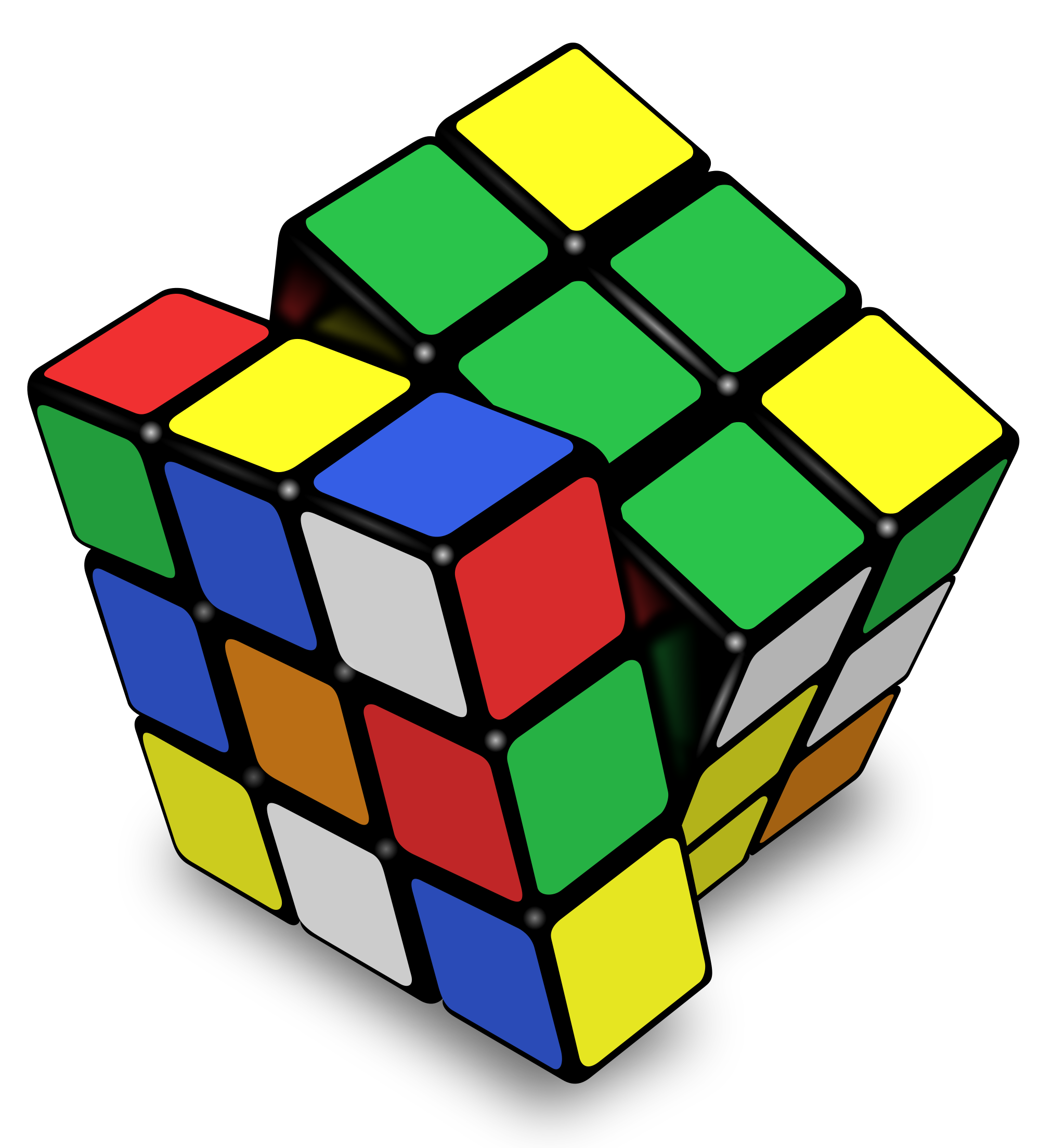}
\end{exampleBox}

\subsection{Действия}

Нас будут интересовать два вида пространства действий $\A$:
\begin{itemize}
    \item[а)] \emph{конечное}, или \emph{дискретное пространство действий} (discrete action space): $|\A| \HM< +\infty$. Мы также будем предполагать, что число действий $|\A|$ достаточно мало.
    \item[б)] непрерывное пространство действий (continuous domain): $\A \HM\subseteq [-1, 1]^m$. Выбор именно отрезков $[-1, 1]$ является не ограничивающем общности распространённым соглашением. Задачи с таким пространством действий также называют задачами \emph{непрерывного управления} (continuous control).
\end{itemize}

\begin{remark}
Заметим, что множество действий не меняется со временем и не зависит от состояния. Если в практической задаче предполагается, что множество допустимых действий разное в различных состояний, в <<законы физики>> прописывается реакция на некорректное действие, например, случайный выбор корректного действия за агента.
\end{remark}

В общем случае процесс выбора агентом действия в текущем состоянии может быть стохастичен. Таким образом, объектом поиска будет являться распределение $\pi(a \mid s), a \HM\in \A, s \HM\in \St$. Заметим, что факт, что нам будет достаточно искать стратегию в классе стационарных (не зависящих от времени), вообще говоря, потребует обоснования.

\subsection{Траектории}

\begin{definition} 
Набор $\Traj \coloneqq \left( s_0, a_0, s_1, a_1, s_2, a_2, s_3, a_3 \dots \right) $ называется \emph{траекторией}.
\end{definition}

\begin{exampleBox}[label=ex:trajectory]{}
Пусть в среде состояния описываются одним вещественным числом, $\St \HM\equiv \R$, у агента есть два действия $\A \HM= \{+1, -1\}$, а следующее состояние определяется как $s' \HM= s + a + \eps$, где $\eps \sim \N(0, 1)$. Начальное состояние полагается равным нулю $s_0 \HM= 0$. Сгенерируем пример траектории для случайной стратегии (вероятность выбора каждого действия равна 0.5):
    \begin{center}
    \includegraphics[width=\textwidth]{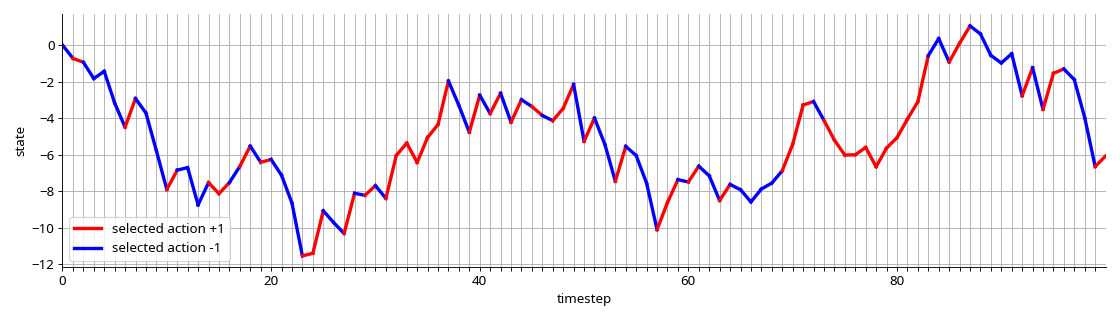}
    \end{center}
\end{exampleBox}

Поскольку траектории --- это случайные величины, которые заданы по постановке задачи конкретным процессом порождения (действия генерируются из некоторой стратегии, состояния --- из функции переходов), можно расписать распределение на множестве траекторий:

\begin{definition} 
Для данной среды, политики $\pi$ и начального состояния $s_0 \HM\in \St$ распределение, из которого приходят траектории $\Traj$, называется \emph{trajectory distribution}:
$$p\left(\Traj \right) = p(a_0, s_1, a_1 \dots) = \prod_{t \ge 0} \pi(a_t \mid s_t) p(s_{t+1} \mid s_t, a_t)$$
\end{definition}

Мы часто будем рассматривать мат.ожидания по траекториям, которые будем обозначать $\E_{\Traj}$. Под этим подразумевается бесконечная цепочка вложенных мат.ожиданий:
\begin{equation}\label{traj_expectation}
\E_{\Traj} \left( \cdot \right) \coloneqq \E_{\pi(a_0 \mid s_0)}\E_{p(s_1 \mid s_0, a_0)}\E_{\pi(a_1 \mid s_1)} \dots \left( \cdot \right)
\end{equation}

Поскольку часто придётся раскладывать эту цепочку, договоримся о следующем сокращении:
$$\E_{\Traj} \left( \cdot \right) = \E_{a_0}\E_{s_1}\E_{a_1} \dots \left( \cdot \right)$$

Однако в такой записи стоит помнить, что действия приходят из некоторой зафиксированной политики $\pi$, которая неявно присутствует в выражении. Для напоминания об этом будет, где уместно, использоваться запись $\E_{\Traj \sim \pi}$.

\subsection{Марковский процесс принятия решений (MDP)}

Для того, чтобы сформулировать задачу, нам необходимо в среде задать агенту цель --- некоторый функционал для оптимизации. По сути, марковский процесс принятия решений --- это среда плюс награда. Мы будем пользоваться следующим определением:

\begin{definition} 
\emph{Марковский процесс принятия решений} (Markov Decision Process, MDP) --- это четвёрка $(\St, \A, \Trans, r)$, где: 
\begin{itemize}
    \item $\St, \A, \Trans$ --- среда.
    \item $r: \St \times \A \to \R$ --- \emph{функция награды} (reward function).
\end{itemize}
\end{definition}

\needspace{7\baselineskip}
\begin{wrapfigure}{r}{0.5\textwidth}
\vspace{-0.4cm}
\centering
\includegraphics[width=0.5\textwidth]{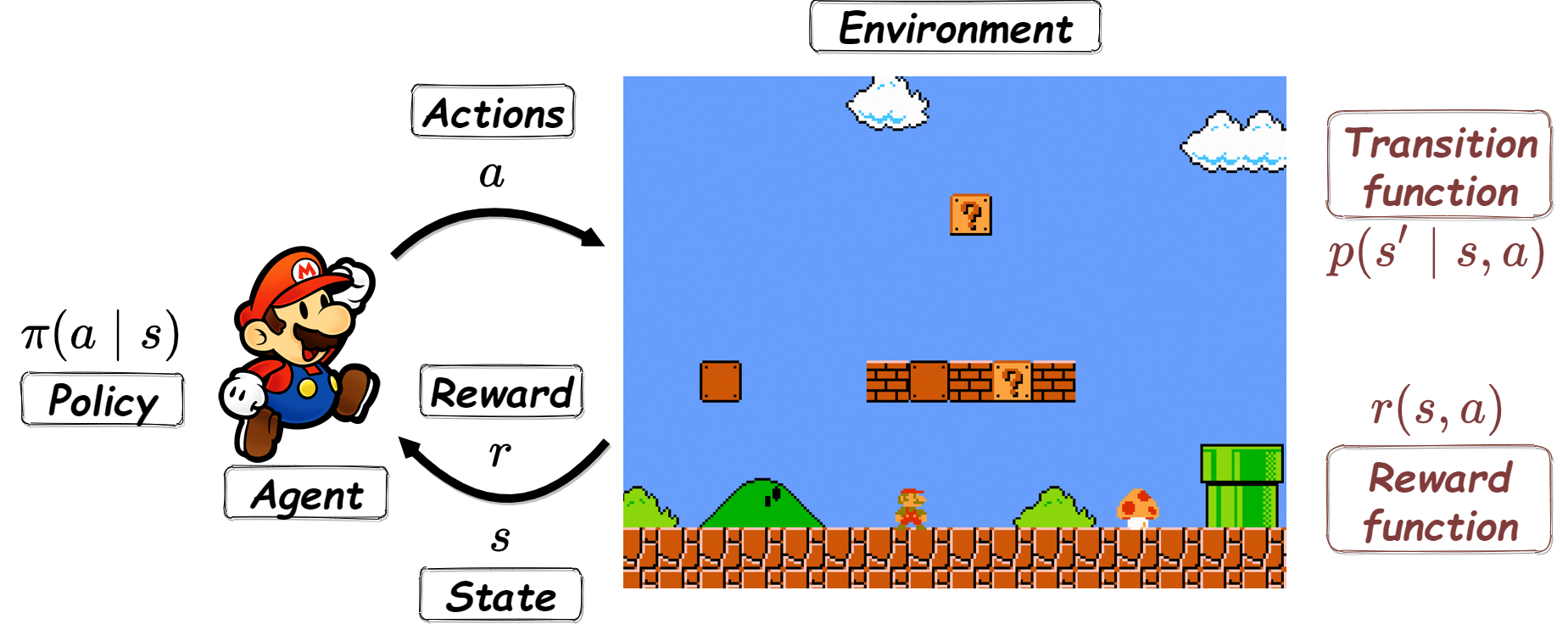}
\vspace{-0.8cm}
\end{wrapfigure}

Сам процесс выглядит следующим образом. Для момента времени $t \HM= 0$ начальное состояние мира полагается $s_0$; будем считать, оно дано дополнительно и фиксировано\footnote{формально его можно рассматривать как часть заданного MDP.}. Агент наблюдает всё состояние целиком и выбирает действие $a_0 \HM\in \A$. Среда отвечает генерацией награды $r(s_0, a_0)$ и сэмплирует следующее состояние $s_1 \HM\sim p(s' \HM\mid s_0, a_0)$. Агент выбирает $a_1 \HM\in \A$, получает награду $r(s_1, a_1)$, состояние $s_2$, и так далее до бесконечности.

\begin{example}
Для среды из примера \ref{ex:trajectory} зададим функцию награды как $r(s, a) = -|s + 4.2|$. Независимость награды от времени является требованием стационарности к рассматриваемым MDP:
\begin{center}
\includegraphics[width=\textwidth]{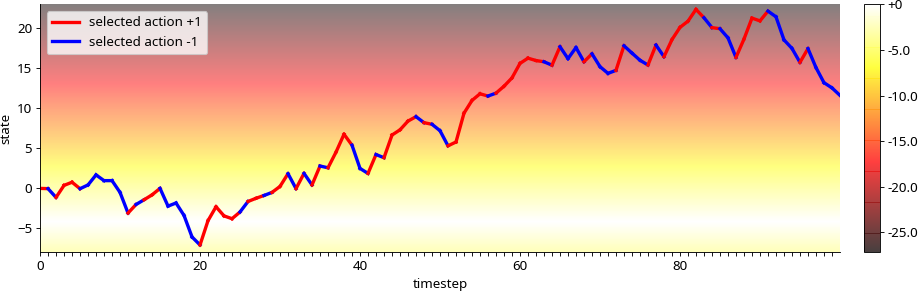}
\end{center}
\end{example}

Формальное введение MDP в разных источниках чуть отличается, и из-за различных договорённостей одни и те же утверждения могут выглядеть совсем непохожим образом в силу разных исходных обозначений. Важно, что суть и все основные теоретические результаты остаются неизменными.

\begin{theorem}[Эквивалентные определения MDP]
Эквивалентно рассматривать MDP, где
\begin{itemize}
    \item функция награды зависит только от текущего состояния;
    \item функция награды является стохастической;
    \item функция награды зависит от тройки ($s$, $a$, $s'$);
    \item переходы и генерация награды задаётся распределением $p(r, s' \HM\mid s, a)$;
    \item начальное состояние стохастично и генерируется из некоторого распределения $s_0 \HM\sim p(s_0)$.
\end{itemize}

\beginproof[Скетч доказательства]{}
Покажем, что всю стохастику можно <<засовывать>>  в $p(s' \HM\mid s, a)$. Например, стохастичность начального состояния можно <<убрать>>, создав отдельное начальное состояние, из которого на первом шаге агент вне зависимости от выбранного действия перейдёт в первое по стохастичному правилу.

\needspace{6\baselineskip}
\begin{wrapfigure}[9]{r}{0.35\textwidth}
\vspace{-0.5cm}
\centering
\includegraphics[width=0.35\textwidth]{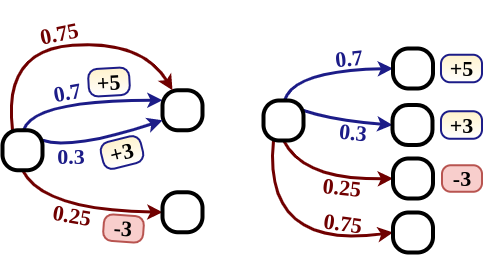}
\vspace{-0.9cm}
\end{wrapfigure}

Покажем, что от самого общего случая (генерации награды и состояний из распределения $p(r, s' \HM\mid s, a)$) можно перейти к детерминированной функции награды только от текущего состояния. Добавим в описание состояний информацию о последнем действии и последней полученной агентом награде, то есть для каждого возможного (имеющего ненулевую вероятность) перехода $(s, a, r, s')$ размножим $s'$ по числу\footnote[*]{которое в худшем случае континуально, так как награда --- вещественный скаляр.} возможных исходов $p(r \HM\mid s, a)$ и по числу действий. Тогда можно считать, что на очередном шаге вместо $r(s, a)$ агенту выдаётся $r(s')$, и вся стохастика процесса формально заложена только в функции переходов. \QED
\end{theorem}

Считать функцию награды детерминированной удобно, поскольку позволяет не городить по ним мат. ожидания (иначе нужно добавлять сэмплы наград в определение траекторий). В любом формализме всегда принято считать, что агент сначала получает награду и только затем наблюдает очередное состояние. 

\begin{definition}
MDP называется \emph{конечным} (finite MDP) или \emph{табличным}, если пространства состояний и действий конечны: $|\St| < \infty, |\A| < \infty$.
\end{definition}

\begin{exampleBox}[label=ex:mdp]{}
Для конечных MDP можно над дугами в графе среды указать награду за переход по ним или выдать награду состояниям (в рамках нашего формализма награда <<за состояние>> будет получена, давайте считать, при выполнении любого действия из данного состояния).
\begin{center}
\animategraphics[controls, width=0.9\linewidth]{1}{Images/MDP/MDP-}{1}{6}
\end{center}
\end{exampleBox}

\subsection{Эпизодичность}

Во многих случаях процесс взаимодействия агента со средой может при определённых условиях <<заканчиваться>>, причём факт завершения доступен агенту.

\begin{definition}
Состояние $s$ называется \emph{терминальным} (terminal) в MDP, если $\forall a \in \A \colon$
$$\Prob (s' = s \mid s, a) = 1 \qquad r(s, a) = 0,$$
то есть с вероятностью 1 агент не сможет покинуть состояние.
\end{definition}

\begin{example}
В примере \ref{ex:mdp} самое правое состояние является терминальным.
\end{example}

Считается, что на каждом шаге взаимодействия агент дополнительно получает для очередного состояния $s$ значение предиката $\done(s) \HM\in \{0, 1\}$, является ли данное состояние терминальным. По сути, после попадания в терминальное состояние дальнейшее взаимодействие бессмысленно (дальнейшие события тривиальны), и, считается, что возможно произвести \emph{reset} среды в $s_0$, то есть начать процесс взаимодействия заново. Введение терминальных состояний именно таким способом позволит всюду в теории писать суммы по времени до бесконечности, не рассматривая отдельно случай завершения за конечное время.

\begin{remark}
Для агента все терминальные состояния в силу постановки задачи неразличимы, и их описания среда обычно не возвращает (вместо этого на практике она обычно проводит ресет и возвращает $s_0$ следующего эпизода).
\end{remark}

\begin{definition}
Один цикл процесса от стартового состояния до терминального называется \emph{эпизодом} (episode).
\end{definition}

Продолжительности эпизодов (количество шагов взаимодействия) при этом, конечно, могут различаться от эпизода к эпизоду.

\begin{definition}
Среда называется \emph{эпизодичной} (episodic), если для любой стратегии процесс взаимодействия гарантированно завершается не более чем за некоторое конечное $T^{\max}$ число шагов.
\end{definition}

\begin{theoremBox}[label=th:episodicmdpistree]{Граф эпизодичных сред есть дерево}
В эпизодичных средах вероятность оказаться в одном и том же состоянии дважды в течение одного эпизода равна нулю.
\begin{proof}
Если для некоторого состояния $s$ при некоторой комбинации действий через $T$ шагов агент с вероятностью $p \HM> 0$ вернётся в $s$, при повторении такой же комбинации действий в силу марковости с вероятностью $p^n \HM> 0$ эпизод будет длиться не менее $nT$ шагов для любого натурального $n$. Иначе говоря, эпизоды могут быть неограниченно долгими. 
\end{proof}
\end{theoremBox}

\begin{exampleBox}[righthand ratio=0.25, sidebyside, sidebyside align=center, lower separated=false]{Cartpole}
К тележке на шарнире крепится стержень с грузиком. Два действия позволяют придать тележке ускорение вправо или влево. Состояние описывается двумя числами: x-координатой тележки и углом, на которой палка отклонилась от вертикального положения.

Состояние считается терминальным, если x-координата стала слишком сильно отличной от нуля (тележка далеко уехала от своего исходного положения), или если палка отклонилась на достаточно большой угол.

Агент получает +1 каждый шаг и должен как можно дольше избегать терминальных состояний. Гарантии завершения эпизодов в этом MDP нет: агент в целом может справляться с задачей бесконечно долго.

\tcblower
\includegraphics[width=\textwidth]{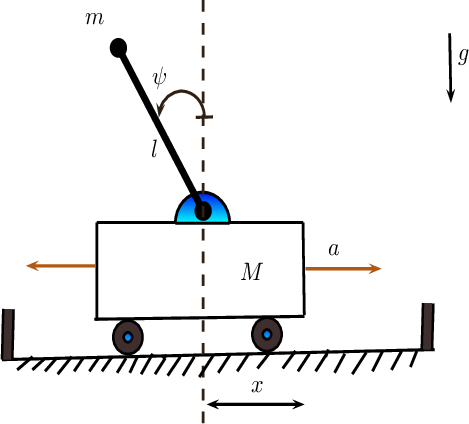}
\end{exampleBox}

\begin{remark}
На практике, в средах обычно существуют терминальные состояния, но нет гарантии завершения эпизодов за ограниченное число шагов. Это лечат при помощи таймера --- жёсткого ограничения, требующего по истечении $T^{\max}$ шагов проводить в среде ресет. Чтобы не нарушить теоретические предположения, необходимо тогда заложить информацию о таймере в описание состояний, чтобы агент знал точное время до прерывания эпизода. Обычно так не делают; во многих алгоритмах RL будет возможно использовать только <<начала>> траекторий, учитывая, что эпизод не был доведён до конца. Формально при этом нельзя полагать $\done \left( s_{T^{\max}} \right) \HM= 1$, но в коде зачастую так всё равно делают. Например, для Cartpole в реализации OpenAI Gym по умолчанию по истечении 200 шагов выдаётся флаг $\done$, что формально нарушает марковское свойство.
\end{remark}

\subsection{Дисконтирование}

Наша задача заключается в том, чтобы найти стратегию $\pi$, максимизирующую среднюю суммарную награду. Формально, нам явно задан функционал для оптимизации:
\begin{equation}\label{sumreward}
\E_{\Traj \sim \pi} \sum_{t \ge 0} r_t \to \max_{\pi},
\end{equation}
где $r_t \HM\coloneqq r(s_t, a_t)$ --- награда на шаге $t$.

Мы хотим исключить из рассмотрения MDP, где данный функционал может улететь в бесконечность\footnote{если награда может быть бесконечной, начинаются всякие парадоксы, рассмотрения которых мы хотим избежать. Допустим, в некотором MDP без терминальных состояний мы знаем, что оптимальная стратегия способна получать +1 на каждом шаге, однако мы смогли найти стратегию, получающую +1 лишь на каждом втором шаге. Формально, средняя суммарная награда равна бесконечности у обоих стратегий, однако понятно, что найденная стратегия <<неоптимальна>>.} или не существовать вообще. Во-первых, введём ограничение на модуль награды за шаг, подразумевая, что среда не может поощрять или наказывать агента бесконечно сильно:
\begin{equation}\label{reward_limit}
\forall s, a \colon |r(s, a)| \le r^{\max}
\end{equation}

Чтобы избежать парадоксов, этого условия нам не хватит\footnote{могут возникнуть ситуации, где суммарной награды просто не существует (например, если агент в бесконечном процессе всегда получает +1 на чётных шагах и -1 на нечётных).}. Введём \emph{дисконтирование} (discounting), коэффициент которого традиционно обозначают $\gamma$:
 
\begin{definition}
\emph{Дисконтированной кумулятивной наградой} (discounted cumulative reward) или \emph{total return} для траектории $\Traj$ с коэффициентом $\gamma \HM\in (0, 1]$ называется
\begin{equation}\label{return}
R(\Traj) \coloneqq \sum_{t \ge 0} \gamma^t r_t
\end{equation}
\end{definition}

У дисконтирования есть важная интерпретация: мы полагаем, что на каждом шаге с вероятностью $1 \HM- \gamma$ взаимодействие обрывается, и итоговым результатом агента является та награда, которую он успел собрать до прерывания. Это даёт приоритет получению награды в ближайшее время перед получением той же награды через некоторое время. Математически смысл дисконтирования, во-первых, в том, чтобы в совокупности с требованием \eqref{reward_limit} гарантировать ограниченность оптимизируемого функционала, а во-вторых, выполнение условий некоторых теоретических результатов, которые явно требуют $\gamma \HM< 1$. В силу последнего, гамму часто рассматривают как часть MDP.

\begin{definition}
\emph{Скором} (score или performance) стратегии $\pi$ в данном MDP называется
\begin{equation}\label{goal}
J(\pi) \coloneqq \E_{\Traj \sim \pi} R(\Traj)
\end{equation}
\end{definition}

Итак, задачей обучения с подкреплением является оптимизация для заданного MDP средней дисконтированной кумулятивной награды:
\begin{equation*}
J(\pi) \to \max_{\pi}
\end{equation*}

\begin{exampleBox}[label=ex:score]{}
Посчитаем $J(\pi)$ для приведённого на рисунке MDP, $\gamma = \frac{10}{11}$ и начального состояния А. Ясно, что итоговая награда зависит только от стратегии агента в состоянии A, поэтому можно рассмотреть все стратегии, обозначив $\pi(a = \text{\colorsquare{ChadRed}} \mid s = A)$ за параметр стратегии $\theta \in [0, 1]$.

\begin{wrapfigure}{r}{0.3\textwidth}
\centering
\includegraphics[width=0.3\textwidth]{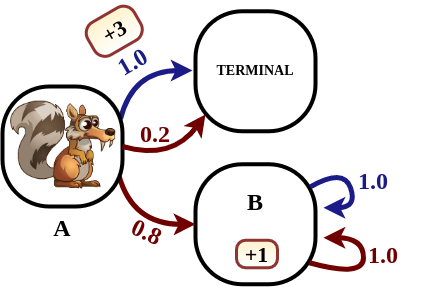}
\vspace{-1cm}
\end{wrapfigure}

C вероятностью $\theta$ агент выберет действие \colorsquare{ChadRed}, после чего попадёт в состояние B с вероятностью 0.8. Там вне зависимости от стратегии он начнёт крутится и получит в пределе
$$\gamma + \gamma^2 + \dots + = \sum\limits_{t \ge 1} \gamma^t = \frac{\gamma}{1 - \gamma} = 10.$$
Ещё с вероятностью $1 - \theta$ агент выберет \colorsquare{ChadBlue}, получит +3 и попадёт в терминальное состояние, после чего эпизод завершится. Итого:
$$J(\pi) = \underbrace{\left( 0.8\theta \right) \cdot 10}_{\text{\colorsquare{ChadRed}}} + \underbrace{\left(1 - \theta \right) \cdot 3}_{\text{\colorsquare{ChadBlue}}} = 3 + 5\theta$$

Видно, что оптимально выбрать $\theta = 1$, то есть всегда выбирать действие \colorsquare{ChadRed}. Заметим, что если бы мы рассмотрели другое $\gamma$, мы бы могли получить другую оптимальную стратегию; в частности, при $\gamma = \frac{15}{19}$ значение $J(\pi)$ было бы константным для любых стратегий, и оптимальными были бы все стратегии. 
\end{exampleBox}

Всюду далее подразумевается\footnote{в качества акта педантичности оговоримся, что также всюду подразумевается измеримость всех функций, необходимая для существования всех рассматриваемых интегралов и мат.ожиданий.} выполнение требования ограниченности награды \eqref{reward_limit}, а также или дисконтирования $\gamma \HM< 1$, или эпизодичности среды.

\begin{proposition}
При сделанных предположениях скор ограничен.
\begin{proof}
Если $\gamma \HM< 1$, то по свойству геометрической прогрессии для любых траекторий $\Traj$:
\begin{equation*}
R(\Traj) = \left| \sum_{t \ge 0} \gamma^t r_t \right| \le \frac{1}{1 - \gamma} r^{\max}
\end{equation*}
Если же $\gamma = 1$, но эпизоды гарантированно заканчиваются не более чем за $T^{\max}$ шагов, то суммарная награда не превосходит по модулю $T^{\max} r^{\max}$. Следовательно, скор как мат.ожидание от ограниченной величины также удовлетворяет этим ограничениям.
\end{proof}
\end{proposition}
\section{Алгоритмы обучения с подкреплением}

\subsection{Условия задачи RL}

Основной постановкой в обучении с подкреплением является задача нахождения оптимальной стратегии на основе собственного опыта взаимодействия. Это означает, что алгоритму обучения изначально доступно только:

\begin{itemize}
    \item вид пространства состояний --- количество состояний в нём в случае конечного числа, или размерность пространства $\R^d$ в случае признакового описания состояний.
    \item вид пространства действий --- непрерывное или дискретное. Некоторые алгоритмы будут принципиально способны работать только с одним из этих двух видов.
    \item взаимодействие со средой, то есть возможность для предоставленной алгоритмом стратегии $\pi$ генерировать траектории $\Traj \sim \pi$; иными словами, принципиально доступны только сэмплы из trajectory distribution \eqref{traj_expectation}.
\end{itemize}

Примерно так выглядит MDP для начинающего обучение агента:

\begin{center}
    \includegraphics[width=0.9\textwidth]{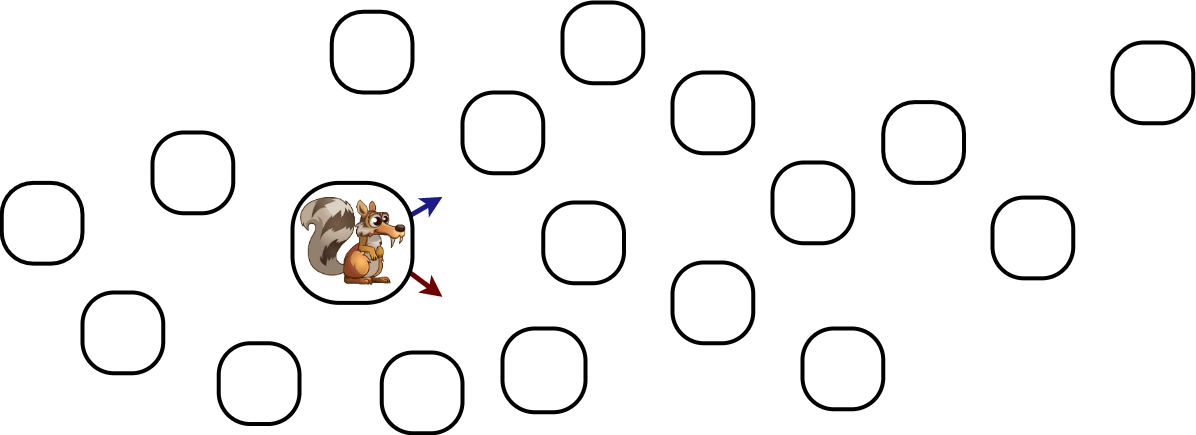}
\end{center}

Итак, в отличие от обучения с учителем, где датасет <<дан алгоритму на вход>>, здесь агент должен сам собрать данные. Находясь в некотором состоянии, обучающийся агент обязан выбрать ровно одно действие, получить ровно один сэмпл $s'$ и продолжить взаимодействие (накопление опыта --- сбор сэмплов) из $s'$. Собираемые в ходе взаимодействия данные и представляют собой всю доступную агенту информацию для улучшения стратегии.

\begin{definition}
Пятёрки $\T \HM\coloneqq \left( s, a, r, s', \done \right)$, где $r \HM\coloneqq r(s, a)$, $s' \HM\sim p(s' \HM\mid s, a)$, $\done \HM\coloneqq \done(s')$, называются \emph{переходами} (transitions). 
\end{definition}

Таким образом, в RL-алгоритме должна быть прописана стратегия взаимодействия со средой во время обучения (\emph{behavior policy}), которая может отличаться от <<итоговой>> стратегии (\emph{target policy}), предназначенной для использования в среде по итогам обучения. 


\subsection{Сравнение с обучением с учителем}

Необходимость собирать данные внутри самого алгоритма --- одно из ключевых отличий задачи RL от обучения с учителем, где выборка подаётся алгоритму на вход. Здесь же <<сигнал>> для обучения предоставляется дизайном функции награды; на практике, чтобы говорить о том, что встретилась задача RL, необходимо задать MDP (предоставить как среду в виде симулятора или интерфейс для взаимодействия настоящего робота в реальном мире, так и функцию награды). Зачастую если какую-то <<репрезентативную>> выборку собрать можно, всегда проще и лучше обратиться к обучению с учителем как менее общей постановке задачи.  

\begin{proposition}
Задача обучения с учителем (с заданной функцией потерь) является частным случаем задачи RL.
\begin{proof}
Пусть дана обучающая выборка в виде пар $(x, y)$, $x$ --- входной объект, $y$ --- целевая переменная. Скажем, что начальное состояние есть описание объекта $x$, случайно сэмплированного из обучающей выборки (начальное состояние будет случайно). Агент выбирает метку класса $\hat{y} \sim \pi(\hat{y} \mid x)$. После этого он получает награду за шаг в в размере $-\Loss(\hat{y}, y)$ и игра завершается. Оптимизация такой функции награды в среднем будет эквивалентно минимизации функции потерь в обучении с учителем. 
\end{proof}
\end{proposition}

Поскольку RL всё-таки предполагает, что в общем случае эпизоды длиннее одного шага, агент будет своими действиями влиять на дальнейшие состояния, которые он встречает. <<Управление>> марковской цепью куда более существенная часть оптимизации RL алгоритмов, нежели максимизация наград за шаг, именно из этого свойства возникает большинство специфических именно для RL особенностей. И зачастую именно в таких задачах появляется такая непреодолимая для обучения с учителем проблема, как то, что правильный ответ никакому человеку, в общем-то, неизвестен. Не знаем мы обычно оптимальный наилучший ход в шахматах или как правильно поворачивать конечности роботу, чтобы начать ходить. 

Да, доступной помощью для алгоритма могут быть \emph{данные от эксперта}, то есть записи взаимодействия со средой некоторой стратегии (или разных стратегий), не обязательно, вообще говоря, оптимальной. Алгоритм RL, возможно, сможет эти данные как-то использовать, или хотя бы как-либо на них предобучиться. В простейшем случае предобучение выглядит так: если в алгоритме присутствует параметрически заданная стратегия $\pi_\theta$, можно \emph{клонировать поведение} (behavior cloning), т.е. восстанавливать по парам $s, a$ из всех собранных экспертом траекторий функцию $\St \HM\to \A$ (это обычная задача обучения с учителем), учиться воспроизводить действия эксперта. Задача обучения по примерам её решения около-оптимальным экспертом называется \emph{имитационным обучением} (imitation learning); её мы обсудим отдельно в разделе \ref{sec:imitationlearning}, однако, если эксперт не оптимален, обученная стратегия вряд ли будет действовать хоть сколько-то лучше. Здесь можно провести прямую аналогию с задачей обучения с учителем, где верхняя граница качества алгоритма определяется качеством разметки; если разметка зашумлена и содержит ошибки, обучение вряд ли удастся. 

В общем же случае мы считаем, что на вход в алгоритм никаких данных эксперта не поступает. Поэтому говорят, что в обучении с подкреплением нам не даны <<правильные действия>>, и, когда мы будем каким-либо образом сводить задачу к задачам регрессии и классификации, нам будет важно обращать внимание на то, как мы <<собираем себе разметку>> из опыта взаимодействия со средой и какое качество мы можем от этой разметки ожидать. В идеале, с каждым шагом алгоритм сможет собирать себе всё более и более <<хорошую>> разметку (получать примеры траекторий с всё большей суммарной наградой), за счёт неё выбирать всё более и более оптимальные действия, и так <<вытягивать сам себя из болота>>.

Здесь важно задать следующий вопрос: а если у нас есть данные из очень плохой стратегии, что можно рассчитывать с них выучить? То есть если мы можем собирать лишь примеры траекторий, в которых агент совершенно случайно себя ведёт (и редко получает набирает положительную награду), можно ли с таких данных выучить, например, оптимальную стратегию? Оказывается, как мы увидим дальше, за счёт структуры задачи RL и формализма MDP ответ положительный. Само собой, такое обучение только будет страшно неэффективным как по времени работы, так и по требуемым объёмам данных.

\subsection{Концепция model-free алгоритмов}

Ещё одним возможным существенным изменением сеттинга задачи является наличие у агента прямого доступа к функции переходов $p(s' \HM\mid s, a)$ и функции награды.

\begin{definition}
Будем говорить, что у агента есть \emph{симулятор} или <<доступ к функции переходов>>, если он знает функцию награды и может в любой момент процесса обучения сэмплировать произвольное число сэмплов из $p(s' \mid s, a)$ для любого набора пар $s, a$.
\end{definition}

Симулятор --- по сути копия среды, которую агент во время обучения может откатывать к произвольному состоянию. Симулятор позволяет агенту строить свою стратегию при помощи \emph{планирования} (planning) --- рассмотрения различных вероятных версий предстоящего будущего и использования их для текущего выбора действия. При использовании планирования в идеальном симуляторе никакого процесса непосредственного обучения в алгоритме может не быть, поскольку если на руках есть симулятор, то данные собирать не нужно: можно из симулятора сколько захотим данных получить.

\begin{exampleBox}[righthand ratio=0.15, sidebyside, sidebyside align=center, lower separated=false]{}
Примером задач, в которых у алгоритма есть симулятор, являются пятнашки или кубик-рубик. То есть, пытаясь создать алгоритм, собирающий кубик-рубик, мы, естественно, можем пользоваться знаниями о том, в какую конфигурацию переводят те или иные действия текущее положение, и таким образом напрашиваются какие-то алгоритмы разумного перебора --- <<планирование>>.

\tcblower
\includegraphics[width=\textwidth]{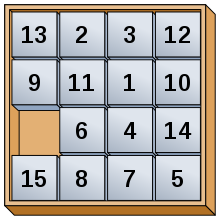}
\end{exampleBox}

\begin{example}
Любые задачи, в которых среда реализована виртуально, можно рассматривать как задачи, где у агента есть симулятор. Например, если мы хотим обучить бота в Марио, мы можем сказать: да у нас есть исходники кода Марио, мы можем взять любую игровую ситуацию (установить симулятор в любое состояние) и для любого действия посмотреть, что будет дальше. В RL по умолчанию считается, что в видеоиграх такого доступа нет: динамика среды изначально агенту неизвестна.
\end{example}

\begin{exampleBox}[righthand ratio=0.3, sidebyside, sidebyside align=center, lower separated=false]{}
Примером задач, в которых у алгоритма принципиально нет симулятора, являются любые задачи реальной робототехники. Важно, что даже если окружение реального робота симулируется виртуально, такая симуляция неточна --- отличается от реального мира. В таких ситуациях можно говорить, что имеется \emph{неидеальный симулятор}. Отдельно стоит уточнить, доступен ли симулятор реальному роботу в момент принятия решения (возможно, симулятор реализован на куда более вычислительно мощной отдельной системе) --- тогда он может использоваться во время обучения, но не может использоваться в итоговой стратегии.

\tcblower
\includegraphics[width=\textwidth]{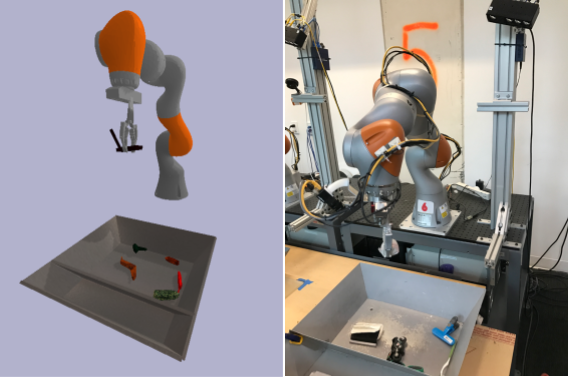}
\end{exampleBox}

По умолчанию всегда считается, что доступа к динамике среды нет, и единственное, что предоставляется алгоритму --- среда, с которой возможно взаимодействовать. Можно ли свести такую задачу к планированию? В принципе, алгоритм может пытаться обучать себе подобный симулятор --- строить генеративную модель, по $s, a$ выдающую $s', r(s, a), \done(s')$ --- и сводить таким образом задачу к планированию. Приближение тогда, естественно, будет неидеальным, да и обучение подобного симулятора сопряжено с рядом других нюансов. Например, в сложных средах в описании состояний может хранится колоссальное количество информации, и построение моделей, предсказывающих будущее, может оказаться вычислительно неподъёмной и неоправданно дорогой задачей. 

Одна из фундаментальных парадигм обучения с подкреплением, вероятно, столь же важная, как парадигма end-to-end обучения для глубокого обучения --- идея model-free обучения. Давайте не будем учить динамику среды и перебирать потенциальные варианты будущего для поиска хороших действий, а выучим напрямую связь между текущим состоянием и оптимальными действиями.

\begin{definition}
Алгоритм RL классифицируется как \emph{model-free}, если он не использует и не пытается выучить модель динамики среды $p(s' \HM\mid s, a)$. 
\end{definition}

\begin{example}
Очень похоже, что когда мы учимся кататься на велосипеде, мы обучаемся как-то в model-free режиме. Мы, находясь в некотором состоянии, не планируем, в какой конфигурации окажется велосипед при разных возможных поворотах наших рук, и вместо этого учимся в точности методом проб и ошибок: мы просто запоминаем, при каком ощущении положения тела как нужно поворачивать руль. 

Здесь очень любопытно пофилософствовать на тему того, почему в таких задачах мы зачастую умеем <<запоминать>> на всю жизнь полученные знания, не разучиваясь ездить на велосипеде после долгих лет без тренировок. Например, нейросетевые модели, которые мы дальше будем обучать для решения задач RL, таким свойством не обладают, и, обучаясь на одном опыте, они старый забывают; если агента RL обучать кататься на велосипеде, а потом долго учить кататься на самокате, стратегия для велосипеда крайне вероятно <<сломается>>.

Оказывается, человек тоже может разучиться ездить на велосипеде при помощи очень хитрой процедуры: нужно придумать такой <<самокат>>, обучение езде на котором требует противоположных навыков, чем велосипед. В качестве такого <<самоката>> можно взять <<обратный велосипед>> (<<The Backwards Bicycle>>): велосипед, в котором поворот руля влево отклоняет колесо вправо, и наоборот. Подробнее про этот эксперимент можно посмотреть в \href{https://www.youtube.com/watch?v=MFzDaBzBlL0}{этом видео}. Интересно, что обе стратегии --- и для езды на велосипеде, и для езды на <<обратном велосипеде>> --- восстанавливаются после некоторой тренировки (причём как-то подозрительно резко, с каким-то <<фазовым переходом>>) и в конечном счёте уживаются вместе.
\end{example}

\subsection{On-policy vs Off-policy}

В model-free алгоритмах сбор данных становится важной составной частью: определяя политику взаимодействия со средой (behavior policy), мы влияем на то, для каких состояний $s, a$ мы получим сэмпл $s'$ из функции переходов. Собираемые данные --- траектории --- алгоритм может запоминать, например, в памяти. Но не каждый алгоритм RL сможет пользоваться такими сохранёнными данными, и поэтому возникает ещё одна важная классификация RL алгоритмов.

\begin{definition}
Алгоритм RL называется \emph{off-policy}, если он может использовать для обучения опыт взаимодействия произвольной стратегии.
\end{definition}

\begin{definition}
Алгоритм RL называется \emph{on-policy}, если для очередной итерации алгоритма ему требуется опыт взаимодействия некоторой конкретной, предоставляемой самим алгоритмом, стратегии.
\end{definition}

Некоторое пояснение названия этих терминов: обычно в нашем алгоритме явно или неявно будет присутствовать некоторая <<текущая>>, <<наилучшая>> найденная стратегия $\pi$: та самая целевая политика (target policy), которую алгоритм выдаст в качестве ответа, если его работу прервать. Если алгоритм умеет улучшать её (проводить очередную итерацию обучения), используя сэмплы любой другой произвольной стратегии $\mu$, то мы проводим <<off-policy>> обучение: обучаем политику <<не по ней же самой>>. On-policy алгоритмам будет необходимо <<отправлять в среду конкретную стратегию>>, поскольку они будут способны улучшать стратегию $\pi$ лишь по сэмплам из неё же самой, будут <<привязаны к ней>>; это существенное ограничение. Другими словами, если обучение с подкреплением --- обучение на основе опыта, то on-policy алгоритмы обучаются на своих ошибках, когда off-policy алгоритмы могут учиться как на своих, так и на чужих ошибках.

Off-policy алгоритм должен уметь проводить очередной шаг обучения на произвольных траекториях, сгенерированных произвольными (возможно, разными, возможно, неоптимальными) стратегиями. Понятие принципиально важно тем, что алгоритм может потенциально переиспользовать траектории, полученные старой версией стратегии со сколь угодно давних итераций. Если алгоритм может переиспользовать опыт, но с ограничениями (например, только с недавних итераций, или только из наилучших траекторий), то мы всё равно будем относить его к on-policy, поскольку для каждой новой итерации алгоритма нужно будет снова собирать сколько-то данных. Это не означает, что для on-policy алгоритма совсем бесполезны данные от (возможно, неоптимального) эксперта; почти всегда можно придумать какую-нибудь эвристику, как воспользоваться ими для хотя бы инициализации (при помощи того же клонирования поведения). Важно, что off-policy алгоритм сможет на данных произвольного эксперта провести <<полное>> обучение, то есть условно сойтись к оптимуму при достаточном объёме и разнообразии экспертной информации, не потребовав вообще никакого дополнительного взаимодействия со средой.

\subsection{Классификация RL-алгоритмов}

При рассмотрении алгоритмов RL мы начнём с рассмотрения именно model-free алгоритмов и большую часть времени посвятим им. Их часто делят на следующие подходы:
\begin{itemize}
    \item \emph{мета-эвристики} (metaheuristic) никак не используют внутреннюю структуру взаимодействия среды и агента, и рассматривают задачу максимизации $J(\pi)$ \eqref{goal} как задачу <<black box оптимизации>>: можно примерно оценивать, чему равно значение функционала для разных стратегий, а структура задачи --- формализм MDP --- не используется; мы рассмотрим мета-эвристики в главе \ref{metaheuristicchapter} как не требующие построения особой теории. 
    \item \emph{value-based} алгоритмы получают оптимальную стратегию неявно через теорию оценочных функций, которую мы рассмотрим в главе \ref{classictheorychapter}. Эта теория позволит нам построить value-based алгоритмы (глава \ref{valuebasedchapter}) и будет использоваться всюду далее.
    \item \emph{policy gradient} алгоритмы максимизируют $J(\pi)$, используя оценки градиента функционала по параметрам стратегии; мы сможем помочь процессу оптимизации, правильно воспользовавшись оценочными функциями (глава \ref{policygradientchapter}).
\end{itemize}

Затем в главе \ref{continuouscontrolchapter} мы отдельно обсудим несколько алгоритмов специально для непрерывных пространств действий, находящихся на стыке value-based и policy gradient подхода, и увидим, что между ними довольно много общего. Наконец, \emph{model-based} алгоритмы, которые учат или используют предоставленную модель среды $p(s' \HM\mid s, a)$, и которые обычно выделяют в отдельную категорию, будут рассмотрены после в главе \ref{modelbasedchapter}.

\subsection{Критерии оценки RL-алгоритмов}

При оценивании алгоритмов принципиально соотношение трёх критериев:
\begin{itemize}
    \item \emph{performance}: Монте-Карло оценка значения $J(\pi)$;
    \item \emph{wall-clock time}: реальное время работы, потребовавшееся алгоритму для достижения такого результата (в полной аналогии с классическими методами оптимизации);
    \item \emph{sample efficiency}: количество сэмплов (или шагов) взаимодействия со средой, потребовавшихся алгоритму. Этот фактор может быть ключевым, если взаимодействие со средой дорого (например, обучается реальный робот); 
\end{itemize}

\begin{remark}Поскольку \href{https://openai.com/projects/five/}{победами над кожаными мешками в дотах} уже никого не удивишь, вторые два фактора начинают играть всю большую роль.
\end{remark}

Конечно, на практике мы хотим получить как можно больший performance при наименьших затратах ресурсов. Время работы алгоритма (wall-clock time) и эффективность по сэмплам связаны между собой, но их следует различать в силу того, что в разных средах сбор данных --- проведение одного шага взаимодействия в среде --- может быть как относительно дешёвым, так и очень дорогим. Например, если мы говорим о реальных роботах, то один шаг взаимодействия со средой --- сверхдорогая операция, по сравнению с которыми любые вычисления на компьютере можно считать очень дешёвыми. Тогда нам выгодно использовать алгоритм, который максимально эффективен по сэмплам. Если же среда задана симулятором, а у нас ещё и есть куча серверов, чтобы параллельно запустить много таких симуляторов, то сбор данных для нас становится дешёвой процедурой. Тогда выгодно не гнаться за sample efficiency и выбирать вычислительно дешёвые алгоритмы.

Очень условно классы RL алгоритмов располагаются по sample efficiency в следующем порядке: самыми неэффективными по требуемым объёмам данных алгоритмами являются мета-эвристики. Зато в них практически не будет никаких вычислений: очень условно, на каждое обновление весов модели будет приходиться сбор нескольких сотен переходов в среде. Их будет иметь смысл применять, если есть возможность параллельно собирать много данных. 

Далее, в Policy Gradient подходе мы будем работать в on-policy режиме: текущая, целевая, политика будет отправляться в среду, собирать сколько-то данных (например, мини-батч переходов, условно проводить порядка 32-64 шагов в среде) и затем делать одно обновление модели. Очень приближённо и с массой условностей говорят, что эффективность по сэмплам будет на порядок выше, чем у эволюционных алгоритмов, за счёт проведения на порядок большего количества вычислений: на одно обновление весов приходится сбор 30-60 переходов.

В value-based подходе за счёт off-policy режима работа можно будет контролировать соотношение числа собираемых переходов к количеству обновлений весов, но по умолчанию обычно полагают, что алгоритм собирает 1 переход и проводит 1 итерацию обучения. Таким образом, вычислений снова на порядок больше, как и (потенциально) sample efficiency.

Наконец, model-based вычислительно страшно тяжеловесные алгоритмы, но за счёт этого можно попытаться добиться максимальной эффективности по сэмплам. 

Отразить четыре класса алгоритмов можно на условной картинке:

\begin{center}
    \includegraphics[width=0.9\textwidth]{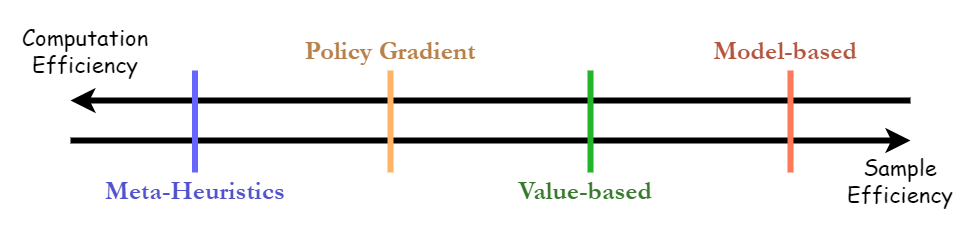}
\end{center}

В зависимости от скорости работы среды, то есть времени, затрачиваемой на сбор данных, а также от особенностей среды (связанных непосредственно со спецификой этих четырёх классов алгоритмов), оптимальное соотношение ресурсы-качество может достигаться на разных классах алгоритмов. Но, к сожалению, на практике редко бывает очевидно, алгоритмы какого класса окажутся наиболее подходящими в конкретной ситуации.

В отличие от классических методов оптимизации, речи о критерии останова идти не будет, поскольку адекватно разумно проверить около-оптимальность текущей стратегии не представляется возможным в силу слишком общей постановки задачи. Считается, что оптимизация (обучение за счёт получения опыта взаимодействия) происходит, пока доступны вычислительные ресурсы; в качестве итога обучения предоставляется или получившаяся стратегия, или наилучшая встречавшаяся в ходе всего процесса.

\subsection{Сложности задачи RL}\label{RLproblems}

Обсудим несколько <<именованных>> проблем задачи обучения с подкреплением, с которыми сталкивается любой алгоритм решения.

Проблема \emph{застревания в локальных оптимумах} приходит напрямую из методов оптимизации. В оптимизируемом функционале \eqref{goal} может существовать огромное количество стратегий $\pi$, для которых его значение далеко не максимально, но все в некотором смысле <<соседние>> стратегии дают в среднем ещё меньшую награду.

\begin{exampleBox}[righthand ratio=0.25, sidebyside, sidebyside align=center, lower separated=false]{}
Часто агент может выучить тривиальное <<пассивное>> поведение, которое не приносит награды, но позволяет избегать каких-то штрафов за неудачи. Например, агент, который хочет научиться перепрыгивать через грабли, чтобы добраться до тортика (+1), может несколько раз попробовать отправиться за призом, но наступить на грабли (-1), и выучить ничего не делать (+0). Ситуация весьма типична: например, Марио может пару раз попробовать отправиться покорять первый уровень, получить по башке и решить не ходить вправо, а тупить в стену и получать <<безопасный>> +0. Для нашей задачи оптимизации это типичнейшие локальные экстремумы: <<похожие стратегии>> набирают меньше текущей, и, чтобы добраться до большей награды, нужно как-то найти совершенно новую область в пространстве стратегий.

\tcblower
\includegraphics[width=\textwidth]{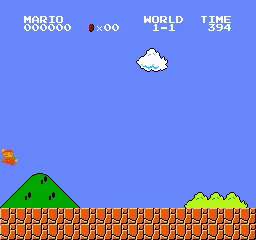}
\end{exampleBox}

Другие проблемы куда более характерны именно для RL. Допустим, агент совершает какое-то действие, которое запускает в среде некоторый процесс. Процесс протекает сам по себе без какого-либо дальнейшего вмешательства агента и завершается через много шагов, приводя к награде. Это проблема \emph{отложенного сигнала} (delayed reward) --- среда даёт фидбэк агенту спустя какое-то (вообще говоря, неограниченно длительное) время.

\begin{exampleBox}[righthand ratio=0.25, sidebyside, sidebyside align=center, lower separated=false]{}
Пример из видеоигр --- в игре Atari Space Invaders очки даются в момент, когда удачный выстрел попал во вражеское НЛО, а не когда агент этот выстрел, собственно, совершает. Между принятием решения и получением сигнала проходит до нескольких секунд, и за это время агент принимает ещё несколько десятков решений.

\tcblower
\includegraphics[width=\textwidth]{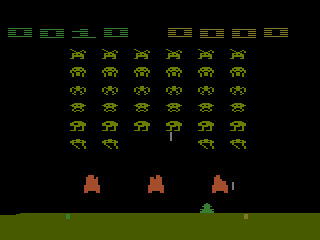}
\end{exampleBox}

Если бы этого эффекта вдруг не было, и агент за каждое своё действие мгновенно получал бы всю награду $r(s, a)$, то решением задачи было бы банальное $\pi(s) = \argmax\limits_a r(s, a)$ --- жадная оптимизация функции награды. Очевидно, это никогда не решение: <<часть>> награды сидит в $s'$, в сэмпле следующего состояния, которого мы добились своим выбором действия, и в описании этого следующего состояния в силу свойства марковости обязана хранится информация о том, сколько времени осталось до получения награды за уже совершённые действия.

Смежная проблема --- какое именно из многих совершённых агентом действий привело к сигналу награды? \emph{Credit assignment problem} --- даже для уже собранного опыта может быть тяжело оценить, какие действия были правильными, а какие нет. 

\begin{example}
Вы поймали скунса, сварили яичницу, завезли банку горчицы в ближайший магазин штор, написали научную статью про шпроты, накормили скунса яичницей, сыграли в боулинг глобусом, вернулись домой после тяжёлого дня и получили +1. Вопрос: чему вы научились? 
\end{example}

Поскольку функция награды может быть произвольная, довольно типично, когда сигнал от среды --- неконстантная награда за шаг --- приходит очень редко. Это проблема \emph{разреженной награды} (sparse reward).

\begin{example}[Mountain Car]
\href{https://gym.openai.com/envs/MountainCar-v0/}{Визуализация задачи в OpenAI Gym}: тележка хочет забраться на горку, но для этого необходимо поехать в противоположном направлении, чтобы набрать разгона. Состояния описываются двумя числами (x-координата тележки и её скорость); действий три (придать ускорения вправо, влево, или ничего не делать). Функция награды равна -1 всюду: задачей агента является как можно скорее завершить эпизод. Однако, терминальное состояние --- это вершина горки, и для того, чтобы достичь его, нужно <<уже уметь>> задачу решать.  
\end{example}

Наконец, проблема, обсуждению которой мы посвятим довольно много времени --- \emph{дилемма исследования-использования} (exploration-exploitation trade-off). Пока ограничимся лишь примером для иллюстрации.

\begin{example}
Вы решили пойти сегодня в ресторан. Следует ли отправится в ваш любимый ресторан, или попробовать новый, в котором вы ещё ни разу не были?
\end{example}

Практическая проблема, отчасти связанная с тем, что алгоритму необходимо постоянно <<пробовать новое>> --- проблема \emph{<<безопасного обучения>>} (Safe RL). Грубо говоря, некоторые взаимодействия агента со средой крайне нежелательны даже во время обучения, в том числе когда агент ещё только учится. Эта проблема возникает в первую очередь для реальных роботов.

\begin{example}
Вы хотите научить реального робота мыть посуду. В начале обучения робот ничего не умеет и рандомно размахивает конечностями, бьёт всю посуду, переворачивает стол, сносит все лампочки и <<в исследовательских целях>> самовыкидывается из окна. В результате, начать второй обучающий эпизод становится довольно проблематично.
\end{example}

\begin{remark}
Здесь надо помнить, что безопасный RL не о том, как <<не сбивать пешеходов>> при обучении автономных автомобилей; он о том, как сбивать меньше пешеходов. RL по определению обучается на собственном опыте и в том числе собственных ошибках, и эти ошибки ему либо нужно совершить, либо получить в форме некоторой априорной информации (экспертных данных), хотя последнее всё равно не защищает от дальнейших исследований любых областей пространства состояний. Это означает, что на практике единственная полноценная защита от нежелательного поведения робота может быть проведена исключительно на этапе построения среды. То есть среда должна быть устроена так, что робот в принципе не может совершить нежелательных действий: опасные ситуации должны детектироваться средой (то есть --- внешне, внешним отдельным алгоритмом), а взаимодействие --- прерываться (с выдачей, например, отрицательной награды для агента).
\end{remark}

\begin{remark}
Одна из причин распространения видеоигр для тестирования RL --- отсутствие проблемы Safe RL: не нужно беспокоиться о том, что робот <<что-то сломает>> в процессе сбора опыта, если среда уже задана программной симуляцией.
\end{remark}

\subsection{Дизайн функции награды}

И есть ещё одна, вероятно, главная проблема\footnote{а точнее, в принципе главная проблема всей нашей жизни (what is your reward function?)}. \emph{Откуда берётся награда?} Алгоритмы RL предполагают, что награда, как и среда, заданы, <<поданы на вход>>, и эту проблему наши алгоритмы обучения, в отличие от предыдущих, решать идеологически не должны. Но понятно, что если для практического применения обучения с учителем боттлнеком часто является необходимость размечивать данные --- <<предоставлять обучающий сигнал>> --- то в RL необходимо аккуратно описать задачу при помощи функции награды.

\begin{definition}
<<\emph{Reward hypothesis}>>: любую интеллектуальную задачу можно задать (определить) при помощи функции награды.
\end{definition}

В обучении с подкреплением принято полагать эту гипотезу истинной. Но так ли это? RL будет оптимизировать ту награду, которую ему предоставят, и дизайн функции награды в ряде практических задач оказывается проблемой.

\begin{example}[<<Взлом>> функции награды]
\href{https://www.youtube.com/watch?v=tlOIHko8ySg}{Классический пример того}, как RL оптимизирует не то, что мы ожидаем, а ту награду, которую ему подсунули. Причина зацикливания агента видна в нижнем левом углу, где отображается счёт игры.
\end{example}

\begin{example}
Попробуйте сформулировать функцию награды для следующих интеллектуальных задач:
\begin{itemize}
    \item очистка мебели от пыли;
    \item соблюдение правил дорожного движения автономным автомобилем;
    \item захват мира;
\end{itemize}
\end{example}

Общее практическое правило звучит так: хорошая функция награды поощряет агента за достигнутые результаты, а не за то, каким способом агент этих результатов добивается. Это логично: если вдруг дизайнеру награды кажется, что он знает, как решать задачу, то вероятно, RL не особо и нужен. К сожалению, такая <<хорошая>> функция награды обычно разреженная.

\begin{exampleBox}[righthand ratio=0.35, sidebyside, sidebyside align=center, lower separated=false]{}
Если вы хотите научиться доезжать до дерева, то хорошая функция награды --- выдать агенту +1 в момент, когда он доехал до дерева. Если попытаться поощрять агента каждый шаг в размере <<минус расстояние до дерева>>, то может возникнуть непредвиденная ситуация: агент поедет прямо в забор, который стоит возле дерева, и это может оказаться с точки зрения такой награды выгоднее, чем делать длинный крюк по району вдали от дерева с целью этот самый забор объехать. 

\tcblower
\includegraphics[width=\textwidth]{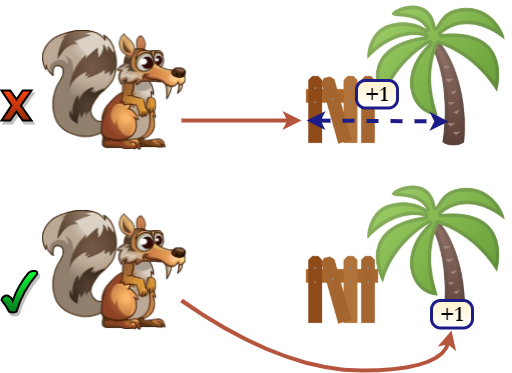}
\end{exampleBox}

Есть, однако, один способ, как можно функцию награды модифицировать, не изменяя решаемую задачу, то есть эквивалентным образом. К сигналу <<из среды>> можно что-то добавлять при условии, что на следующем шаге это что-то обязательно будет вычтено. Другими словами, можно применять трюк, который называется \emph{телескопирующая сумма} (telescoping sums): для любой последовательности $a_t$, т.~ч. $\lim\limits_{t \to \infty} a_t \HM= 0$, верно
\begin{equation}\label{telescopingsum}
    \sum_{t \ge 0}^\infty \left( a_{t+1} - a_t \right) = -a_0
\end{equation}

\begin{definition}
Пусть дана некоторая функция $\Phi(s) \colon \St \to \R$, которую назовём \emph{потенциалом}, и которая удовлетворяет двум требованиям: она ограничена и равна нулю в терминальных состояниях. Будем говорить, что мы проводим \emph{reward shaping} при помощи потенциала $\Phi(s)$, если мы заменяем функцию награды по следующей формуле:
\begin{equation}\label{rewardshaping}
r^{\mathrm{new}}(s, a, s') \coloneqq r(s, a) + \gamma \Phi(s') - \Phi(s)    
\end{equation}
\end{definition}

\begin{theorem}[Reward Shaping]
Проведение любого reward shaping по формуле \eqref{rewardshaping} не меняет задачи.

\begin{proof} Посчитаем суммарную награду для произвольной траектории. До reward shaping суммарная награда равнялась $\sum\limits_{t \ge 0} \gamma^t r_t$. Теперь же суммарная награда равна
\begin{equation*}
\sum_{t \ge 0} \gamma^t r_t + \sum_{t \ge 0} \left[ \gamma^{t+1} \Phi(s_{t+1}) - \gamma^{t} \Phi(s_t) \right]
\end{equation*}

В силу свойства телескопирующей суммы \eqref{telescopingsum}, все слагаемые во второй сумме посокращаются, кроме первого слагаемого $-\Phi(s_0)$, который не зависит от стратегии взаимодействия и поэтому не влияет на итоговую задачу оптимизации, <<последнего слагаемого>>, которое есть ноль: действительно, $\lim\limits_{t \to \infty} \gamma^{t} \Phi(s_t) \HM= 0$ в силу ограниченности функции потенциала. Если же в игре конечное число шагов $T$, не сокращающееся слагаемое $\Phi(s_T)$ есть значение потенциала в терминальном состоянии $s_T$, которая по условию также равно нулю.
\end{proof}
\end{theorem}

Reward shaping позволяет поменять награду, <<не ломая>> задачу. Это может быть способом борьбы с разреженной функцией награды, если удаётся придумать какой-нибудь хороший потенциал. Конечно, для этого нужно что-то знать о самой задаче, и на практике reward shaping --- это инструмент внесения каких-то априорных знаний, дизайна функции награды. Тем не менее, в будущем в главе \ref{PIsection} мы рассмотрим один хороший универсальный потенциал, который будет работать всегда.

\subsection{Бенчмарки}

Для тестирования RL алгоритмов есть несколько распространившихся бенчмарков. Неистощаемым источником тестов для обучения с подкреплением с дискретным пространством действий являются видеоигры, где, помимо прочего, уже задана функция награды --- счёт игры.

\begin{example}[Игры Atari]
Atari --- набор из 57 игр с дискретным пространством действий. Наблюдением является экран видео-игры (изображение), у агента имеется до 18 действий (в некоторых играх действия, соответствующие бездействующим кнопкам джойстика, по умолчанию убраны). Награда --- счёт в игре. \href{https://gym.openai.com/envs/#atari}{Визуализация игр из OpenAI Gym}.

\needspace{12\baselineskip}
\begin{wrapfigure}{r}{0.4\textwidth}
\centering
\vspace{-0.3cm}
\includegraphics[width=0.4\textwidth]{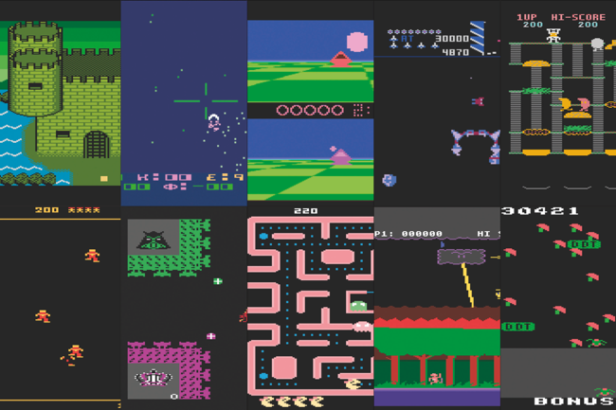}
\vspace{-0.8cm}
\end{wrapfigure}

При запуске алгоритмов на Atari обычно используется препроцессинг. В общем случае MDP, заданный исходной игрой, не является полностью наблюдаемым: например, в одной из самых простых игр \href{https://gym.openai.com/envs/Pong-v0/}{Pong} текущего экрана недостаточно, чтобы понять, в какую сторону летит шарик. Для борьбы с этим состоянием считают последние 4 кадра игры (\emph{frame stack}), что для большинства игр достаточно.

Чтобы агент не имел возможности менять действие сильно чаще человека, применяют \emph{frame skip} --- агент выбирает следующее действие не каждый кадр, а, скажем, раз в 4 кадра. В течение этих 4 кадров агент нажимает одну и ту же комбинацию кнопок. Если агент <<видит>> из-за этого только кратные кадры, могут встретиться неожиданные последствия: например, в Space Invaders каждые 4 кадра <<исчезают>> все выстрелы на экране, и агент может перестать их видеть.

\begin{remark}
Обычно добавляется и препроцессинг самого входного изображения --- в оригинале оно сжимается до размера 84х84 и переводится в чёрно-белое.
\end{remark}

\begin{remark}
В играх часто встречается понятие <<жизней>>. Полезно считать, что потеря жизни означает конец эпизода, и выдавать в этот момент алгоритму флаг $\done$. 
\end{remark}

Функция переходов в Atari --- детерминированная; рандомизировано только начальное состояние. Есть опасения, что это может приводить к <<запоминанию>> хороших траекторий, поэтому распространено использование \emph{sticky actions} --- текущее выбранное действие повторяется для $k$ кадров, где $k$ определяется случайно (например, действие повторяется только 2 раза, затем для очередного кадра подбрасывается монетка, и с вероятностью 0.5 агент повторяет действие снова; монетка подбрасывается снова, и так далее, пока не выпадет останов).

Для оценки алгоритмов используется \emph{Human normalized score}: пусть $\mathrm{agentScore}$ --- полученная оценка $J(\pi)$ для обучившегося агента, $\mathrm{randomScore}$ --- случайной стратегии, $\mathrm{humanScore}$ --- средний результат человека, тогда Human normalized score равен
$$\frac{\mathrm{agentScore} - \mathrm{randomScore}}{\mathrm{humanScore} - \mathrm{randomScore}}.$$
Эта величина усредняется по 57 играм для получения качества алгоритма. При этом по условию бенчмарка алгоритм должен быть запущен на всех 57 играх с одними и теми же настройками и гиперпараметрами.
\end{example}

\begin{example}[Atari RAM]
Игры Atari представлены в ещё одной версии --- <<RAM-версии>>. Состоянием считается не изображение экрана, а 128 байт памяти Atari-консоли, в которой содержится вся информация, необходимая для расчёта игры (координаты игрока и иные параметры). По определению, такое состояние <<полностью наблюдаемое>>, и также может использоваться для тестирования алгоритмов.
\end{example}

Для задач непрерывного управления за тестовыми средами обращаются к физическим движкам и прочим симуляторам. Здесь нужно оговориться, что схожие задачи в разных физических движках могут оказаться довольно разными, в том числе по сложности для RL алгоритмов.

\begin{exampleBox}[label=ex:locomotion]{Locomotion}
Задача научить ходить какое-нибудь существо весьма разнообразна. Под <<существом>> понимается набор сочленений, каждое из которых оно способно <<напрягать>> или <<расслаблять>> (для каждого выдаётся вещественное число в диапазоне $[-1, 1]$). Состояние обычно описано положением и скоростью всех сочленений, то есть является небольшим компактным векторочком. 

\needspace{15\baselineskip}
\begin{wrapfigure}{r}{0.25\textwidth}
\centering
\vspace{-0.3cm}
\includegraphics[width=0.25\textwidth]{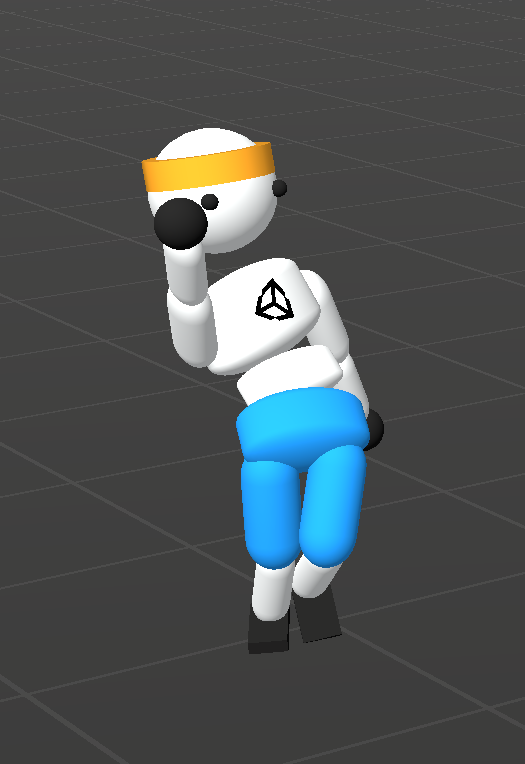}
\vspace{-0.8cm}
\end{wrapfigure}

Типичная задача заключается в том, чтобы в рамках представленной симуляции физики научиться добираться как можно дальше в двумерном или трёхмерном пространстве. Функция награды, описывающая такую задачу, не так проста: обычно она состоит из ряда слагаемых, дающих бонусы за продолжение движения, награду за скоррелированность вектора скорости центра масс с желаемым направлением движения и штрафы за трату энергии.

\begin{remark}
Такая награда, хоть и является довольно плотной, обычно <<плохо отнормирована>>: суммарное значение за эпизод может быть довольно высоким. Нормировать награду <<автоматически>> могут в ходе самого обучения, считая, например, средний разброс встречаемых наград за шаг и деля награду на посчитанное стандартное отклонение. Распространено считать разброс не наград за шаг, а суммарных наград с начала эпизода, и делить награды за шаг на их посчитанное стандартное отклонение.
\end{remark}

Несмотря на малую размерность пространства состояний и действий (случаи, когда нужно выдавать в качестве действия векторы размерности порядка 20, уже считаются достаточно тяжёлыми), а также информативную функцию награды, дающую постоянную обратную связь агенту, подобные задачи непрерывного управления обычно являются довольно сложными.
\end{exampleBox}

\begin{example}
В робототехнике и симуляциях робот может получать информацию об окружающем мире как с камеры, наблюдая картинку перед собой, так и узнавать о располагающихся вокруг объектах с помощью разного рода сенсоров, например, при помощи \emph{ray cast}-ов --- расстояния до препятствия вдоль некоторого направления, возможно, с указанием типа объекта. Преимущество последнего представления перед видеокамерой в компактности входного описания (роботу не нужно учиться обрабатывать входное изображение). В любом случае, входная информация редко когда полностью описывает состояние всего окружающего мира, и в подобных реальных задачах требуется формализм частично наблюдаемых сред. 
\end{example}

\chapter{Мета-эвристики}\label{metaheuristicchapter}

В данной главе мы рассмотрим первый подход к решению задачи, часто называемый <<эволюционным>>. В этом подходе мы никак не будем использовать формализацию процесса принятия решений (а следовательно, не будем использовать какие-либо результаты, связанные с изучением MDP) и будем относиться к задаче как к black-box оптимизации: мы можем отправить в среду поиграть какую-то стратегию и узнать, сколько примерно она набирает, и задача алгоритма оптимизации состоит в том, чтобы на основе лишь этой информации предлагать, какие стратегии следует попробовать следующими. 

\section{Бэйзлайны}

\subsection{Задача безградиентной оптимизации}

\begin{definition}
\emph{Мета-эвристикой} (metaheuristic) называется метод black-box оптимизации
$$J(\theta) \to \max_{\theta \in \Theta}$$
со \emph{стохастическим оракулом нулевого порядка} (stochastic zeroth-order oracle), то есть возможностью для каждой точки $\theta \in \Theta$ получить несмещённую оценку $\hat{J} \HM\approx J(\theta)$.
\end{definition}

Такие методы также называются \emph{безградиентными} (gradient-free), поскольку не используют градиент функции и в принципе не предполагают её дифференцируемости. Понятно, что такие методы --- <<универсальный>> инструмент (читать, <<инструмент последней надежды>>), который можно использовать для любой задачи оптимизации. В первую очередь, этот инструмент полезен, если пространство аргументов $\Theta$ нетривиально (например, графы) или если оптимизируемая функция принципиально недифференцируема, состоит из седловых точек (<<inadequate landscape>>) или есть другие препятствия для градиентной оптимизации.

Заметим, что если $\Theta$ --- конечное множество (о градиентной оптимизации тогда речи идти не может), задача сводится к следующей: надо найти тот аргумент $\theta \in \Theta$, для которого настоящее значение $J(\theta)$ максимально, при этом используя как можно меньше стохастичных оценок $\hat{J}$ оракула. Это задача многорукого бандита, которую мы обсудим отдельно в секции \ref{sec:bandistssection}. В теории мета-эвристик опция <<вызвать оракул в одной и той же точке несколько раз>> в ходе алгоритма обычно не рассматривается; предполагается достаточно богатое пространство $\Theta$, для которого более прагматичной альтернативой кажется запросить значение условно в <<соседней>> точке вместо уточнения значения оракула для одной и той же.

В контексте обучения с подкреплением, чтобы свести задачу к black-box оптимизации, достаточно представить стратегию $\pi_\theta(a \HM\mid s)$ в параметрическом семействе с параметрами $\theta \HM\in \Theta$. В качестве $J$, конечно, выступает наш оптимизируемый функционал \eqref{goal}
$$J(\theta) \coloneqq \E_{\Traj \sim \pi_\theta} R(\Traj) \to \max_{\theta},$$
а в качестве $\hat{J}$ --- его Монте-Карло оценка:
$$\hat{J}(\theta) \coloneqq \frac{1}{B}\sum_{i=1}^{B} R(\Traj_i), \quad \Traj_i \sim \pi_\theta, i \in \{1 \dots B \}$$

\begin{remark}
Если дисперсия оценки достаточно высока (число сэмплов $B$ недостаточно велико), почти все далее рассматриваемые алгоритмы сломаются (будут выживать <<везучие>>, а не <<сильнейшие>>). Поэтому может оказаться крайне существенным использовать $B > 1$, даже если $\pi_\theta$ --- семейство детерминированных стратегий.
\end{remark}

Будем проникаться местной терминологией:
\begin{definition}
Точку $\theta$, в которой алгоритм оптимизации запрашивает значение оракула, будем называть \emph{особью} (individuals, particles), а само значение $\hat{J}(\theta)$ для данной особи --- её \emph{оценкой} или \emph{приспособленностью} (fitness).
\end{definition}

В силу гигантского разнообразия мета-эвристик (от \href{https://en.wikipedia.org/wiki/List_of_metaphor-based_metaheuristics}{метода светлячков до колоний императорских пингвинов}) на полноту дальнейшее изложение, конечно же, не претендует, и стоит воспринимать рассуждение как попытку структурировать мотивации некоторых из основных идей. В частности, нас в первую очередь будут интересовать алгоритмы, в той или иной степени успешно применявшиеся в RL.

\subsection{Случайный поиск}

\emph{Случайный поиск} (random search) --- метод оптимизации и самый простой пример мета-эвристики.

\begin{definition}
Распределение $q(\theta)$ в пространстве $\Theta$ будем называть \emph{стратегией перебора}.
\end{definition}

Случайный поиск сводится к сэмплированию из стратегии перебора особей $\theta_k \HM\sim q(\theta)$ ($k \HM\in \{0, 1, 2 \dots \}$), после чего в качестве результата выдаётся особь с наилучшей оценкой.

\begin{example}[Случайный поиск]
\begin{center}
    \includegraphics[width=\textwidth]{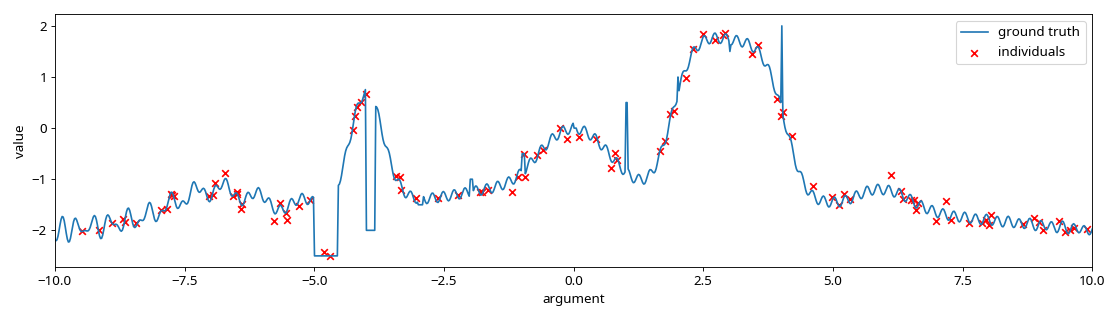}
\end{center}
\end{example}

Забавно, что случайный поиск --- метод глобальной оптимизации: если $\forall \theta \HM\in \Theta \colon q(\theta) \HM> 0$, после достаточного числа итераций метод найдёт сколь угодно близкое к глобальному оптимуму решение\footnote{с оговоркой, что метод сможет понять, что нашёл оптимум, для чего придётся предположить некоторые условия регулярности для $J(\theta)$; понятно, что функцию <<иголка в стоге сена>> (needle in a haystack)
$$J(\theta) = \begin{cases}0 & \theta \neq \theta^* \\ 1 & \theta = \theta^*\end{cases}$$ никакой метод оптимизации в $\Theta \equiv \R$ не прооптимизирует.}. Есть ещё один парадокс грубого перебора: если в наличии есть неограниченное число серверов, то возможно запустить на каждом вычисление приспособленности одной особи, и за время одного вычисления провести <<глобальную>> оптимизацию.

Идея случайного поиска, на самом деле, вводит основные понятия мета-эвристик. Нам придётся так или иначе запросить у оракула приспособленности некоторого набора особей и так или иначе в итоге отобрать лучший. Для имитации умности происходящего введём весёлую нотацию.

\begin{definition}
Набор особей $\Pop \coloneqq \left( \theta_i \mid i \in \{1, 2 \dots N\} \right) $ называется \emph{популяцией} (population) размера $N$.
\end{definition}

\begin{definition}
Запрос оракула для всех особей популяции называется \emph{оцениванием} (evaluation) популяции:
$$\hat{J}(\Pop) \coloneqq \left( \hat{J}(\theta_i) \Bigm| i \in \{1, 2 \dots N\} \right)$$
\end{definition}

\newcommand{\Sel}{\mathrm{select}}
\begin{definition}
Процедурой \emph{отбора} (selection) называется выбор (возможно, случайный, возможно, с повторами) $M$ особей из популяции. Формально, это распределение $\Sel(\Pop^+ \mid \Pop, \hat{J}(\Pop))$, такое что $\forall \theta \in \Pop^+ \colon \theta \in \Pop$ с вероятностью 1.
\end{definition}

\newcommand{\Seltop}{\Sel^{\mathop{top}}}
\begin{definition}
\emph{Жадный} (greedy) отбор $\Seltop_M$ --- выбор топ-$M$ самых приспособленных особей.
\end{definition}

Жадный отбор плох тем, что у нас нет гарантий, что мы на самом деле выбираем наилучшую точку из рассмотренных --- наши оценки $\hat{J}$ могут быть неточны, и наилучшей на самом деле может оказаться особь с не самой высокой приспособленностью. В частности поэтому могут понадобиться альтернативы жадного отбора.

\begin{example}[Пропорциональный отбор]
Зададимся некоторым распределением на особях популяции, которые сэмплирует особь тем чаще, тем выше её приспособленность. Например:
$$p(\theta) \propto \exp{\hat{J}(\theta)}$$
Для отбора $M$ особей засэмплируем (обычно с возвращением) из этого распределения $M$ раз.
\end{example}

\begin{exampleBox}[label=ex:tournir]{Турнирный отбор}
Для отбора $M$ особей $M$ раз повторяется следующая процедура: случайно выбираются $K$ особей популяции (из равномерного распределения) и отбирается та из них, чья приспособленность выше. Число $K$ называется \emph{размером турнира} и регулирует вероятность плохо приспособленной особи быть отобранной: чем больше $K$, тем меньше у слабенькой особи шансов выжить.
\end{exampleBox}

Одна и та же особь в ходе отбора может быть выбрана несколько раз: это можно читать как <<особи повезло оставить больше потомства>>\footnote{С точки зрения эволюционной теории, критерием оптимизации для особи является исключительно преумножение количества своих генов за счёт размножения. Отбор --- не столько про <<выживание сильнейших>>, сколько про количество успешных передач генов потомкам.}.

Итак, случайный поиск можно переформулировать на языке мета-эвристик так: сгенерировать из данной стратегии перебора популяцию заданного размера $N$ и отобрать из неё 1 особь жадно.

\subsection{Hill Climbing}

Процедура отбора позволяет только <<сокращать>> разнообразие популяции. Хочется как-то обусловить процесс генерации новых кандидатов на уже имеющуюся информацию (которая состоит только из особей и их приспособленностей). Мотивация введения мутаций в том, что даже в сложных пространствах $\Theta$ зачастую можно что-то <<поделать>> с точкой $\theta$ так, чтобы она превратилась в другую точку $\hat{\theta}$.

\newcommand{\mut}{\mathfrak{m}}
\newcommand{\Mut}{\mathfrak{M}}
\begin{definition}
\emph{Мутацией} (mutation) называется распределение $\mut(\hat{\theta} \HM\mid \theta)$, где $\theta$ называется \emph{родителем} (parent), $\hat{\theta}$ --- \emph{потомком} (child).
\end{definition}

\begin{exampleBox}[label=ex:graph_mutation]{}
Пусть $\Theta$ --- множество путей обхода вершин некоторого графа. Такое пространство аргументов возникает во многих комбинаторных задачах (таких как \href{https://ru.wikipedia.org/wiki/\%D0\%97\%D0\%B0\%D0\%B4\%D0\%B0\%D1\%87\%D0\%B0_\%D0\%BA\%D0\%BE\%D0\%BC\%D0\%BC\%D0\%B8\%D0\%B2\%D0\%BE\%D1\%8F\%D0\%B6\%D1\%91\%D1\%80\%D0\%B0}{задача коммивояжёра}). Пусть $\theta \in \Theta$ --- некоторый путь обхода, то есть упорядоченное множество вершин графа. Мутацией может выступать выбор случайных двух вершин и смена их местами в порядке обхода. Например, в исходном обходе мы проходили пять вершин графа в порядке (4, 3, 1, 5, 2), а после мутации - в порядке (4, 2, 1, 5, 3). 
\end{exampleBox}

Рассмотрим простейший способ использования мутации. На $k$-ом шаге алгоритма будем генерировать $N$ потомков особи $\theta_k$ при помощи мутации и отбирать из них жадно особь $\theta_{k+1}$. Очень похоже на градиентный подъём: мы сэмплим несколько точек вокруг себя и идём туда, где значение функции максимально. Поэтому отчасти можно считать, что Hill Climbing с большим $N$ --- локальная оптимизация: мы не можем взять наилучшее направление изменения $\theta$ из, например, градиентов, но можем поискать хорошее направление, условно, случайным перебором.

\begin{example}[Hill Climbing]
\begin{center}
\animategraphics[controls, width=\linewidth]{1}{Images/HC/hill_climbing-}{0}{9}
\end{center}
\end{example}

Что получается: если мутация такова, что $\forall \theta, \hat{\theta} \colon \mut(\hat{\theta} \HM\mid \theta) \HM> 0$, остаются гарантии оказаться в любой точке пространства, и алгоритм остаётся методом глобальной оптимизации. При этом, мутация может быть устроена так, что вероятность оказаться <<неподалёку>> от родителя выше, чем в остальной области пространства. 

\begin{example}
В примере \ref{ex:graph_mutation} мутация не удовлетворяла свойству $\forall \theta, \hat{\theta} \colon \mut(\hat{\theta} \HM\mid \theta) \HM> 0$: из текущего пути обхода мы могли получить только очень похожий. Применение мета-эвристик c такой мутацией чревато скорым застреванием в локальном оптимуме: мы можем попасть в точку, применение мутации к которой с вероятностью один даёт ещё менее приспособленные особи. Чтобы полечить это, создадим другую мутацию: засэмплируем натуральное $n$ из \href{https://ru.wikipedia.org/wiki/\%D0\%A0\%D0\%B0\%D1\%81\%D0\%BF\%D1\%80\%D0\%B5\%D0\%B4\%D0\%B5\%D0\%BB\%D0\%B5\%D0\%BD\%D0\%B8\%D0\%B5_\%D0\%9F\%D1\%83\%D0\%B0\%D1\%81\%D1\%81\%D0\%BE\%D0\%BD\%D0\%B0}{распределения Пуассона} или из $p(n) \coloneqq \frac{1}{2^n}$ и применим такое количество мутаций вида <<сменить две вершины местами>>. Так мы гарантируем, что с небольшой вероятностью мы сможем мутировать в произвольную точку пространства аргументов.
\end{example}

Понятно, что если мутация генерирует очень непохожие на родителя особи, алгоритм схлопывается примерно в случайный поиск. И понятно, что если мутация, наоборот, с огромной вероятностью генерирует очень близкие к родителю особи, алгоритм, помимо того, что будет сходиться медленно, будет сильно надолго застревать в локальных оптимумах. Возникает trade-off между \emph{использованием} (exploitation) и \emph{исследованием} (exploration): баланс между выбором уже известных хороших точек и поиском новых <<вдали>>; изучением окрестностей найденных локальных оптимумов и поиском новых. 

\begin{exampleBox}[label=ex:gaussianmutation]{}
Если $\Theta \HM\equiv \R^h$, то типичным выбором мутации является $\mut(\hat{\theta} \HM\mid \theta) \HM\coloneqq \N(\theta, \sigma^2I_{h \times h})$, где $\sigma \HM> 0$ --- гиперпараметр, и, чем ближе $\sigma$ к нулю, тем <<ближе>> к родителю потомки.
\end{exampleBox}

\begin{remark}
Возникает вопрос: а как подбирать гиперпараметры мета-эвристик, тоже мета-эвристикой? Любопытный ответ --- \emph{самоадаптирующиеся параметры} (self-adaptive mutations). Параметры мутации кодируются в пространстве аргументов $\Theta$; то есть одна из координат каждого $\theta \in \Theta$ и отвечает за, например, дисперсию $\sigma$ в добавляемом шуме из нормального распределения. Появляется надежда, что мета-эвристика <<сама подберёт себе хорошие гиперпараметры>>.
\end{remark}

\subsection{Имитация отжига}

\emph{Имитация отжига} (simulated annealing) решает <<проблему исследования>> при помощи более умной процедуры отбора: вероятность выбрать потомка $\theta'_{k+1} \HM \sim \mut(\theta'_{k+1} \HM\mid \theta_k)$, а не остаться в родительской точке $\theta_k$, вводится так:
$$\Sel \left( \theta_{k+1} = \theta'_{k+1} \right) \coloneqq \min \left(1, \exp{\frac{\hat{J}(\theta'_{k+1}) - \hat{J}(\theta_k)}{\tau_k}} \right),$$
где $\tau_k \HM> 0$ --- \emph{температура}, гиперпараметр, зависящий от номера итерации. Иными словами, если новая точка более приспособлена, то мы <<принимаем>> новую точку $\theta'_{k+1}$ с вероятностью 1; если же новая точка менее приспособлена, мы не выкидываем её, а переходим в неё с некоторой вероятностью. Эта вероятность тем ближе к единице, чем <<похожее>> значения оракула, и температура регулирует понятие похожести между скалярами относительно масштаба оптимизируемой функции $J$.

Наша цепочка $\theta_0, \theta_1, \theta_2 \dots$ задаёт марковскую цепь: мы генерируем каждую следующую особь на основе только предыдущей, используя некоторое стохастичное правило перехода. Теория марковских цепей говорит, что может существовать \emph{стационарное распределение} (stationary distribution): распределение, из которого приходят $\theta_k$, при стремлении $k \to \infty$ всё ближе к некоторому $p(\theta)$, которое определяется лишь нашей функцией переходов и не зависит от инициализации $\theta_0$.

\begin{theorem}[Алгоритм Метрополиса-Гастингса]
Пусть в пространстве $\Theta$ задано распределение $p(\theta)$ и распределение $q(\hat{\theta} \HM\mid \theta)$, удовлетворяющее $\forall \hat{\theta}, \theta \colon q(\hat{\theta} \HM\mid \theta) > 0$. Пусть строится цепочка $\theta_0, \theta_1, \theta_2 \dots$ по следующему правилу: генерируется $\theta'_{k+1} \HM\sim q(\theta'_{k+1} \HM\mid \theta_k)$, после чего с вероятностью
$$\min \left( 1, \frac{p(\theta'_{k+1})}{p(\theta_{k})}\frac{q(\theta_k \HM\mid \theta'_{k+1})}{q(\theta'_{k+1} \HM\mid \theta_{k})} \right)$$
$\theta_{k+1}$ полагается равным $\theta'_{k+1}$, а иначе $\theta_{k+1} \coloneqq \theta_k$.
Тогда для любого $\theta_0$:
$$\lim_{k \to \infty} p(\theta_k) = p(\theta)$$
\begin{proof}[Без доказательства]
\end{proof}
\end{theorem}

\begin{proposition}
Пусть оракул точный, то есть $\hat{J}(\theta) \equiv J(\theta)$, а мутация удовлетворяет
$$\forall \theta, \hat{\theta} \colon \mut(\hat{\theta} \HM\mid \theta) = \mut(\theta \HM\mid \hat{\theta}) > 0$$
Тогда, если температура $\tau$ не зависит от итерации, для любой инициализации $\theta_0$ алгоритм имитации отжига строит марковскую цепь со следующим стационарным распределением:
$$\lim_{k \to \infty} p(\theta_k) \propto \exp \frac{J(\theta_k)}{\tau}$$
\end{proposition}

Алгоритм Метрополиса даёт, вообще говоря, <<гарантии сходимости>> для имитации отжига: то, что через достаточно большое количество итераций мы получим сэмпл из распределения $\exp \frac{J(\theta_k)}{\tau}$, нас, в общем-то, устраивает. Если температура достаточно маленькая, это распределение очень похоже на вырожденное в точке максимального значения оптимизируемой функции. Одновременно, правда, маленькая температура означает малую долю исследований в алгоритме, и тогда процедура вырождается в наивный поиск при помощи мутации. Поэтому на практике температуру снижают постепенно; подобный \emph{отжиг} (annealing) часто применяется для увеличения доли исследований в начале работы алгоритма и уменьшения случайных блужданий в конце.

\begin{example}[Имитация отжига]
\begin{center}
\animategraphics[controls, width=\linewidth]{1}{Images/SA/simulated_annealing-}{0}{9}
\end{center}
\end{example}

\begin{remark}
Если в операторе мутации есть параметр, отвечающий за <<силу мутаций>>, связанный с балансом исследования-использования (например, дисперсия $\sigma$ гауссовского шума в примере \ref{ex:gaussianmutation}), его аналогично можно подвергнуть отжигу: в начале алгоритма его значение выставляется достаточно большим, после чего с ходом алгоритма оно постепенно по некоторому расписанию уменьшается.
\end{remark}

\subsection{Эволюционные алгоритмы}

Hill Climbing --- эвристика с <<одним состоянием>>: есть какая-то одна текущая основная особь-кандидат, на основе которой и только которой составляется следующая особь-кандидат. Нам, вообще говоря, на очередном шаге доступна вся история проверенных точек. Первый вариант --- построить суррогат-приближение $\hat{J}(\theta)$, которую легко можно прооптимизировать и найти так следующего кандидата (к нему относятся, например, алгоритмы на основе \href{https://distill.pub/2020/bayesian-optimization/}{гауссовских процессов}), но этот вариант не сработает в сложных пространствах $\Theta$. Второй вариант --- перейти к <<эвристикам с $N$ состояниями>>, то есть использовать последние $N$ проверенных особей для порождения новых кандидатов.

\begin{definition}
Алгоритм называется \emph{эволюционным} (evolutionary), если он строит последовательность популяций $\Pop_1, \Pop_2 \dots$, на $k$-ом шаге строя очередное \emph{поколение} (generation) $\Pop_k$ на основе предыдущего.
\end{definition}

Практически все мета-эвристики сводятся к эволюционным алгоритмам. При этом заметим, что, пока единственным инструментом <<генерации>> новых особей выступает мутация, алгоритмы различаются только процедурой отбора. Таким образом, в алгоритме всегда поддерживается текущая популяция из $N$ особей (первая популяция генерируется из некоторой стратегии перебора $q(\theta)$), из них отбирается $N$ особей (отбор может выбирать одну особь несколько раз, поэтому этот шаг нетривиален), и дальше каждая отобранная особь мутируется. Эвристика \emph{элитизма} (elitism) предлагает некоторые из отобранных особей не мутировать, и оставить для следующей популяции; это позволяет уменьшить шансы популяции <<потерять>> найденную область хороших значений функции, но увеличивает шансы застревания в локальном оптимуме. При элитизме для каждой особи хранится её \emph{возраст} (age) --- число популяций, через которые особь прошла без мутаций; далее этот возраст влияет на отбор, например, выкидывая все особи старше определённого возраста. Далее будем рассматривать алгоритмы без элитизма.

Придумаем простейший эволюционный алгоритм. Любой алгоритм локальной оптимизации (градиентный спуск или Hill Climbing), результат работы которого зависит от начального приближения $\theta_0 \sim q(\theta)$, можно <<заменить>> на метод глобальной оптимизации, запустив на каждом условном сервере по <<\emph{потоку}>> (thread) со своим начальным $\theta_0$. Время работы алгоритма не изменится (считая, конечно, что сервера работают параллельно), а обнаружение неограниченного числа локальных оптимумов с ненулевым шансом найти любой гарантирует нахождение глобального. Набор из текущих состояний всех потоков можно считать текущей популяцией.

При этом, любая процедура отбора позволит потокам <<обмениваться информацией между собой>>. Для примера рассмотрим распространённую схему \emph{$(M, K)$-эволюционной стратегии}, в которой процедура отбора заключается в том, чтобы топ-$M$ особей отобрать по $K$ раз каждую (дать каждой особи из топа породить $K$ детей). Иными словами, мы параллельно ведём $M$ Hill Climbing-ов, в каждом из которых для одного шага генерируется $K$ потомков; если число потоков $M = 1$, алгоритм вырождается в обычный Hill Climbing. Порождённые $N = MK$ особей образуют текущую популяцию алгоритма. Среди них отбирается $M$ лучших особей, которые могут как угодно распределиться по потокам. Получится, что некоторые потоки, которые не нашли хороших областей $\Theta$, будут прерваны, а хорошие получат возможность сгенерировать больше потомков и как бы <<размножаться>> на несколько процессов. 

\begin{algorithm}{$(M, K)$-эволюционная стратегия}
\textbf{Вход:} оракул $\hat{J}(\theta)$ \\
\textbf{Гиперпараметры:} $\mut(\hat{\theta} \mid \theta)$ --- мутация, $q(\theta)$ --- стратегия перебора, $M$ --- число потоков, $K$ --- число сэмплов на поток

\vspace{0.3cm}
Инициализируем $\Pop_0 \coloneqq \left( \theta_i \sim q(\theta) \mid i \in \{1, 2, \dots , MK \} \right)$ \\
\textbf{На $k$-ом шаге:}
\begin{enumerate}
    \item проводим жадный отбор: $\Pop^{+}_k \coloneqq \Seltop_M(\Pop_k, \hat{J}(\Pop_k))$
    \item размножаем: $\Pop_{k+1} \coloneqq \left( \hat{\theta}_i \sim \mut(\hat{\theta} \mid \theta) \mid \theta \in \Pop^{+}_k, i \in \{1, 2, \dots , K\} \right)$
\end{enumerate}
\end{algorithm}

\begin{example}[(4, 5)-эволюционная стратегия]
\begin{center}
\animategraphics[controls, width=\linewidth]{1}{Images/MS_ES/ms_es-}{0}{8}
\end{center}
\end{example}

\subsection{Weight Agnostic Neural Networks (WANN)}

Поскольку мета-эвристики --- универсальный метод оптимизации, они могут применяться к нейронным сетям. Такой подход даёт такие преимущества, как возможность использовать дискретные веса или недопустимые для градиентной оптимизации функции активации вроде пороговой функции Хевисайда ($f(x) \coloneqq \mathbb{I}[x \ge 0]$). 

Особенностью \emph{нейроэволюции} (neuroevolution) является возможность искать топологию сети вместе со значениями самих весов: для этого достаточно предложить некоторый оператор мутации. Рассмотрим распространённый пример: пусть дана некоторая нейросеть с произвольной топологией\footnote{на заре нейросетевого подхода разумность использования полносвязных слоёв была под вопросом. Считаем, что нейрон принимает несколько входов, домножает каждый вход на некоторый вес, складывает и применяет функцию активации, выдавая скалярную величину. Выход нейрона отправляется некоторым из других нейронов на вход; понятие <<слоёв>> обычно не вводится, а единственное ограничение на вычислительный граф --- ацикличность.} и некоторыми весами. Для весов процедуру мутирования можно взять стандартную (добавление шума с заданной дисперсией; дополнительно можно для каждого веса сэмплировать бинарную величину, будет ли данный вес мутировать). Далее случайно выбирается, будет ли проходить мутация архитектуры (топологии) сети, и если да, то какого типа (обычно рассматривается несколько типов мутации).

\begin{exampleBox}[label=ex:wannmutations]{Топологические мутации нейросети}
Распространён набор из трёх видов топологических мутаций: добавление связи, добавление нейрона и смена функции активации. 

\begin{wrapfigure}{r}{0.3\textwidth}
\centering
\vspace{-0.5cm}
\includegraphics[width=0.3\textwidth]{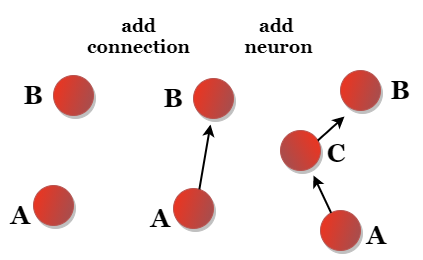}
\vspace{-1cm}
\end{wrapfigure}

Добавление связи означает, что случайно выбираются два ранее не соединённых связью нейрона, и выход одного добавляется ко входу в другой (вес инициализируется, например, случайно). Какой из двух нейронов является входом, а какой --- выходом, однозначно определяется требованием ацикличности к вычислительному графу.

Под добавлением нейрона понимается именно разбиение уже имеющейся связи: имевшаяся связь A--B, где A, B --- нейроны, выключается, и появляются связи A--C и C--B, где C --- новый нейрон. Новые нейроны необходимо добавлять именно так, чтобы они сразу же участвовали в вычислительном процессе. 

Смена функции активации меняет функцию активацию в случайном нейроне на произвольную из предопределённого набора.

\begin{remark}
Заметим, что в таком операторе мутации отсутствуют шансы на уменьшение числа связей. Эволюционный процесс с таким оператором мутации будет рассматривать пространство $\Theta$ нейросетей с различными топологиями от <<более простых>> архитектур к <<более сложным>>. В частности поэтому рекомендуется изначально алгоритм инициализировать <<пустой>> топологией: между входами и выходами связей нет, а появляться они будут в ходе постепенных мутаций. Похоже, что к такой эвристике привёл эмпирический опыт, и введение мутаций, удаляющих часть архитектуры, не приводило к особым успехам; результатом работы нейроэволюционного алгоритма типично является крайне минимальная по современным меркам нейросеть, с довольно малым количеством связей.
\end{remark}
\end{exampleBox}

\emph{Weight Agnostic} сети зашли ещё дальше и, по сути, отказались от настройки весов нейросети в принципе, ограничившись только поиском топологии. Алгоритм WANN основан на $(M, K)$-эволюционной стратегии с оператором мутации из рассмотренного примера \ref{ex:wannmutations}. Единственным изменением является процедура отбора. Для данной топологии оракул запускается шесть раз. В каждом запуске всем весам сети присваивается одно и то же значение (авторы использовали значения $[-2, -1, -0.5, 0.5, 1, 2]$). Рассматривается три критерия приспособленности особи: среднее из шести значение $\hat{J}$, максимальное из шести значение $\hat{J}$ и число связей. Эти три критерия не смешиваются со скалярными гиперпараметрами, вместо этого проводится турнирный отбор (пример \ref{ex:tournir}): из популяции выбираются два кандидата, выбирается случайный критерий, и выживает сильнейший по данному критерию. Так отбирается $M$ особей, которые и порождают $K$ потомков каждый.

\subsection{Видовая специализация}\label{specieisidea}

Возникает простой вопрос к $(M, K)$-эволюционной стратегии: а не случится ли такого, что он схлопнется к Hill Climbing-у, если в очередной популяции среди топ-$M$ останутся только дети одного и того же родителя? Получится, что, хоть мы и исходили из идеи исследовать параллельно много локальных оптимумов, жадный отбор может убить все потоки, кроме одного, и <<область пространства аргументов>>, покрываемая текущей популяцией, <<схлопнется>>.

Для борьбы с этим эффектом в мета-эвристиках рассматривают методы \emph{защиты инноваций} (innovation protection). Если для получения новых хороших свойств необходимо сделать <<несколько>> шагов эволюции, то мы каким-то образом помогаем выживать особям, оказавшихся в не исследуемых местах пространства $\Theta$. Самый простой способ --- использование более <<мягких>> процедур отбора, когда у неприспособленной особи есть небольшой шанс выжить. Более интеллектуально было бы как-то оценить, насколько <<новой>> является область пространства аргументов, в которой оказалась особь, и дать ей больше шансов выжить, если алгоритм эту область ещё не рассматривал: ведь проблема мягких процедур отбора, очевидно, в том, что слабые особи выживают в том числе там, где функция уже в достаточной степени исследована.

Рассмотрим общую идею \emph{видов} (species). Допустим, мы сможем в $\Theta$ придумать метрику (или хоть сколько-то функцию близости) $\rho(\theta_1, \theta_2)$ и на её основе разбить все особи популяции $\Pop$ на непересекающиеся множества --- <<виды>>. Процесс разбиения, что важно, не обязан удовлетворять каким-то особым свойствам и может быть стохастичным, в том числе чтобы быть вычислительно дешёвым. 

\begin{exampleBox}[label=ex:species]{Процедура разделения на виды}
Изначально множество видов пусто. На очередном шаге берём очередную особь из $\Pop$ и в случайном порядке перебираем имеющиеся виды. Для каждого вида сэмплируем одну из ранее отнесённых к нему особей и сравниваем расстояние $\rho$ с порогом-гиперпараметром процесса: меньше порога --- относим рассматриваемую особь к этому виду и переходим к следующей особи, больше порога --- переходим к следующему виду. Если ни для одного вида проверка не прошла, особь относится к новому виду. Пересчёт видов проводится для каждой популяции заново.
\end{exampleBox}

Разделение на виды позволяет делать, например, \emph{explicit fitness sharing}: особи соревнуются только внутри своих видов. Пусть $\widetilde{\Pop} \subseteq \Pop$ --- вид, а $\hat{J}_{\mathop{mean}}(\widetilde{\Pop})$ --- среднее значение приспособленности в данном виде. Тогда на основе этих средних значений между видами проводится (в некоторой <<мягкой>> форме --- слабые виды должны выживать с достаточно высокой вероятностью) некоторый мягкий отбор; например, виду $\widetilde{\Pop}$ позволяется сгенерировать потомков пропорционально $\exp \hat{J}_{\mathop{mean}}(\widetilde{\Pop})$ с учётом того, что в сумме все виды должны породить заданное гиперпараметром число особей. Генерация необходимого числа потомков внутри каждого вида происходит уже, например, стандартным, <<агрессивным>> образом: отбирается некоторая доля топ-особей, к которым и применяется по несколько раз мутация.

Виды защищены мягким отбором, и поэтому заспавнивнишеся вдали особи, образующие новый вид, будут умирать реже; при этом в скоплениях слабых особей в одном месте пройдёт жёсткий внутривидовой отбор, а сам вид получит не так много <<слотов потомства>>, и число точек сократится.

\begin{example}[Эволюция с видовой специализацией]
В данном примере текущая популяция из 20 особей разделяется на виды согласно процедуре из примера \ref{ex:species} с метрикой $\rho(\theta_1, \theta_2) = |\theta_1 - \theta_2|$ и порогом 1. Для видов считается $\hat{J}_{\mathop{mean}}(\widetilde{\Pop})$ (указан как fit в легенде), после чего при помощи сэмплирования из распределения $\propto \exp \hat{J}_{\mathop{mean}}(\widetilde{\Pop})$ 20 раз разыгрываются между видами <<слоты потомства>>. Внутри каждого вида жадно отбирается 1 особь, она и порождает разыгранное число потомков. Разбиение на виды проводится для новой полученной популяции заново.
\begin{center}
\animategraphics[controls, width=\linewidth]{1}{Images/EL_ES/el_es-}{0}{9}
\end{center}
\end{example}

\subsection{Генетические алгоритмы}

До сих пор мы умели создавать новые особи только при помощи мутации. В генетических алгоритмах дополнительно вводится этап \emph{рекомбинации} (recombination), когда новые особи можно строить на основе сразу нескольких особей, как-то <<совмещая>> свойства тех и других в надежде получить <<лучшее от двух миров>>; найти хороший оптимум между двумя локальными оптимумами. В ванильной версии генетических алгоритмов у детей по два родителя, хотя можно рассматривать и скрещивание большего числа особей:

\newcommand{\cross}{\mathfrak{c}}
\begin{definition}
\emph{Кроссинговером} (crossover) называется распределение $\cross(\hat{\theta} \HM\mid \theta_1, \theta_2)$, где $\theta_1, \theta_2$ называются \emph{родителями} (parent), $\hat{\theta}$ --- \emph{потомком} (child).
\end{definition}

\begin{example}
Для $\Theta \equiv \R^d$ или $\Theta \equiv \{0, 1\}^d$ (или их смеси) можно придумать много разных кроссинговеров; генетика подсказывает, что если элементы векторов это <<гены>>, то нужно, например, некоторые гены взять от одного родителя, а другие от другого. Некоторые гены можно \emph{сцеплять} (сцепленные гены должны быть взяты из одного родителя), или вводить порядок на генах (брать от одного родителя <<правую>> часть, от другого <<левую>>).
\end{example}

Чтобы сохранить свойство <<глобальности>> оптимизации, желательно было бы, опять же, чтобы мы могли при помощи такого инструмента порождения оказаться в любой точке пространства, т.е. $\forall \hat{\theta}, \theta_1, \theta_2 \HM\in \Theta \colon \cross(\hat{\theta} \HM\mid \theta_1, \theta_2) > 0$. Однако, для практически любых примеров кроссинговера это не так. Поэтому считается, что это требование НЕ выполняется: процедура рекомбинации, возможно, стохастична, но всегда приводит к точке <<между>> $\theta_1$ и $\theta_2$. Это означает, что, используя только кроссинговер, область, покрываемая потомством, будет уже области, покрываемой родителями: теряется исследование. Чтобы полечить это, в генетических алгоритмах всё равно остаётся этап применения мутации.

\begin{algorithm}[label=geneticsearch]{Генетический поиск}
\textbf{Дано:} оракул $\hat{J}(\theta)$ \\
\textbf{Гиперпараметры:} $\cross(\hat{\theta} \mid \theta_1, \theta_2)$ --- кроссинговер, $\mut(\hat{\theta} \mid \theta)$ --- мутация, $\Sel(\Pop^+ \mid \Pop, \hat{J}(\Pop))$ --- процедура отбора, $q(\theta)$ --- стратегия перебора, $N$ --- размер популяции

\vspace{0.3cm}
Инициализируем $\Pop_0 \coloneqq \left( \theta_i \sim q(\theta) \mid i \in \{1, 2, \dots , N \} \right)$ \\
\textbf{На $k$-ом шаге:}
\begin{enumerate}
    \item проводим отбор: $\Pop^{+}_k \sim \Sel(\Pop^{+}_k \mid \Pop_k, \hat{J}(\Pop_k))$
    \item проводим размножение: $\Pop_{k+1} \coloneqq \left( \theta_i \sim \cross \left( \theta \mid \theta_l, \theta_r \right) \mid \theta_l, \theta_r \in \Pop^{+}_k \right)$
    \item проводим мутирование: $\Pop_{k+1} \leftarrow \left( \hat{\theta} \sim \mut (\hat{\theta} \mid \theta) \mid \theta \in \Pop_{k+1} \right)$
\end{enumerate}
\end{algorithm}

\begin{remark}
При эволюционном обучении нейросетей, в отличие от ряда других задач, кроссинговер во многом неудобен. Нельзя взять <<половинку>> одной хорошей нейросети и присоединить к <<половинке>> другой хорошей нейросети --- каждый нейрон рассчитывает на тот набор входов, для которого он был обучен (неважно, эволюционно или градиентно). Поэтому генетические алгоритмы для нас не представляют особого интереса, по крайней мере на момент написания данного текста. 
\end{remark}

\begin{example}[Neuroevolution of Augmented Topologies (NEAT)]
Рассмотрим в качестве примера набор эвристик нейроэволюционного алгоритма NEAT, который всё-таки был основан на генетическом поиске (алг.~\ref{geneticsearch}), то есть дополнительно вводил оператор кроссинговера для нейросетей (двух произвольных топологий).

\begin{wrapfigure}{r}{0.35\textwidth}
\centering
\vspace{-0.5cm}
\includegraphics[width=0.35\textwidth]{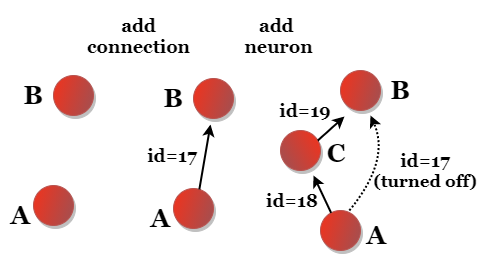}
\vspace{-0.7cm}
\end{wrapfigure}

В NEAT все связи всех особей, помимо веса, имеют статус <<включена-выключена>> и уникальный идентификационный номер (id), или \emph{исторический маркер} (historical marker). Связи могут появляться только в ходе мутаций (используется оператор мутации из примера \ref{ex:wannmutations}); в момент создания связи ей присваивается уникальный (в рамках всего алгоритма) id и статус <<включена>>. Наличие у двух особей связи с одним id будет означать наличие общего предка. У изначальной <<пустой>> топологии связей нет вообще. Связь может попасть в статус <<выключена>> только во время мутации вида <<добавление нейрона>>, когда имевшаяся связь A--B <<исчезает>>: то есть, соответствующий <<ген>> не удаляется из генома особи, а переходит в статус <<выключена>>. Выключенная связь означает, что у особи был предок, у которого связь была включена.

Как введение статусов и id-шников связей позволяет устраивать между разными топологиями кроссинговер? Если связь с данным id имеется у обоих скрещиваемых особей, связь сохраняется и у потомка, с весом и статусом случайного родителя. Связи, имеющиеся только у одного из родителей, будем называть \emph{непарными}; они копируются из того родителя, чья приспособленность выше (или из обоих сразу, если приспособленности одинаковые).

\begin{center}
    \includegraphics[width=\textwidth]{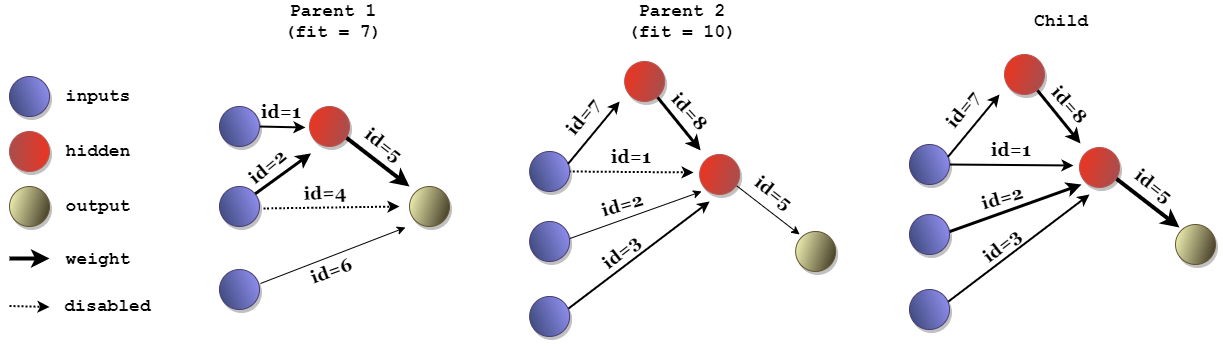}
\end{center}

NEAT также использует видовую специализацию (как описано в разделе \ref{specieisidea}) для процесса отбора, для чего на таких генотипах необходимо задать метрику $\rho$. Пусть $\theta_1, \theta_2$ --- две особи, $G_1, G_2$ --- количество связей у этих особей, $D$ --- число непарных связей, $w_1, w_2$ --- веса особей в парных связях. Понятно, что веса мы можем сравнить только для парных связях, и понятно, что чем больше непарных связей, тем больше должно быть расстояние. В NEAT предлагается просто объединить эти два критерия: 
$$\rho(\theta_1, \theta_2) \coloneqq \alpha_1 \frac{D}{max(G_1, G_2)} + \alpha_2 \|w_1 - w_2\|_1$$
где $\alpha_1, \alpha_2$ --- гиперпараметры. Внутри самих видов отбор жадный.

NEAT --- исторически один из первых алгоритмов, которые можно с каким-то результатом применить для RL задач без подготовленного удобного признакового описания состояний, например, изображений. Можно найти много интересных примеров применения алгоритма к разным задачам, с визуализацией получающейся сети (например, \href{https://www.youtube.com/watch?v=qv6UVOQ0F44}{Mario}). 

\end{example}

\section{Эволюционные стратегии}

\subsection{Идея эволюционных стратегий}

Когда мы пытаемся генерировать $k \HM+ 1$-ое поколение, используя особей $k$-го поколения в качестве исходного материала, единственная зависимость от всей истории заложена в составе $k$-ой популяции. Мы можем рассмотреть распределение, из которого появляются особи очередной популяции: весь смысл нашей процедуры отбора в том, чтобы это распределение для очередной итерации поменялось так, что вероятность появления более хороших особей стала выше. Давайте обобщим эту идею: будем в явном виде хранить распределение для порождения особей новой популяции и по информации со всей популяции аккумулировать всю информацию внутри его параметров --- <<скрещивать все особи>>.

\begin{definition}
Распределение $q(\theta \HM\mid \lambda_k)$, из которого генерируются особи $k$-ой популяции, называется \emph{эволюционной стратегией} (evolutionary strategy, ES):
$$\Pop_{k} \coloneqq \{\theta_i \sim q(\theta \HM\mid \lambda_k) \mid i \in \{1, 2 \dots N\} \}$$
где $\lambda_k$ --- параметры эволюционной стратегии.
\end{definition}

\needspace{7\baselineskip}
\begin{wrapfigure}{r}{0.4\textwidth}
\centering
\vspace{-0.6cm}
\includegraphics[width=0.4\textwidth]{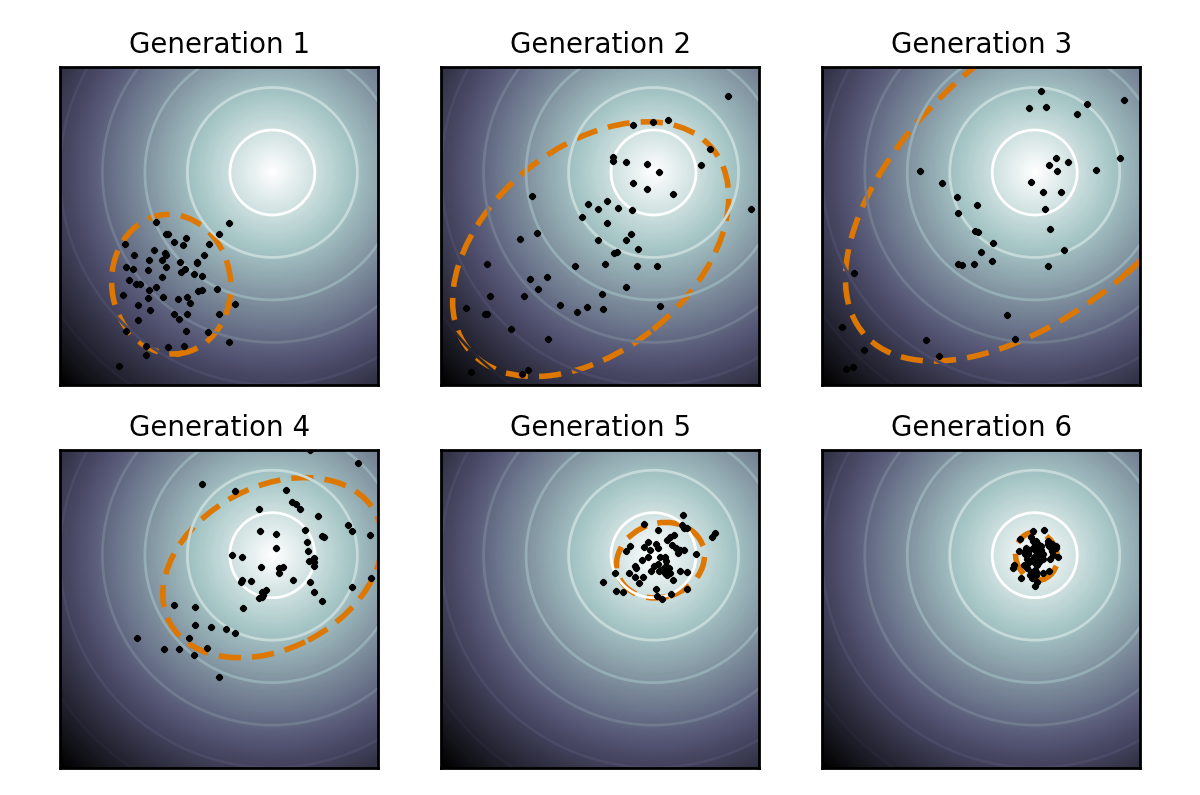}
\vspace{-0.5cm}
\end{wrapfigure}

Из каких соображений подбирать $\lambda_k$ на очередном шаге? В принципе, мы хотели бы найти такой генератор особей, что их оценки как можно больше, то есть нащупать область $\Theta$ с высоким значением $J(\theta)$. Эти соображения можно формализовать довольно по-разному и таким образом оправдывать разные мета-эвристики. В частности, мы можем сказать, что $\lambda$ есть особь или несколько особей (что приведёт нас к примерно ранее рассматривавшимся алгоритмам\footnote{понятие эволюционных стратегий довольно общее и размытое --- любой эволюционный алгоритм <<неявно>> определяет распределение для порождения особей очередной популяции и формально подпадает под эволюционные стратегии. Можно считать, что здесь ключевая идея заключается в том, что мы явно ищем это распределение в некотором параметрическом семействе.}), но мы можем отойти от пространства $\Theta$ и учить модель-генератор особей с какой-то хорошей параметризацией $\lambda$. Мы дальше рассмотрим две основные идеи, как это можно делать.

\subsection{Оценка вероятности редкого события}

Первую идею возьмём немного сбоку. Допустим, стоит задача оценки вероятности редкого события:
\begin{equation}\label{rareeventestimation}
l = \Prob(f(x) \ge \gamma) = \E_{x \sim p(x)} \mathbb{I}\left[ f(x) \ge \gamma \right] \,\text{---}\, ?
\end{equation}
где $p(x)$ --- некоторое распределение, $f \colon X \to \R$ --- функционал, $\gamma$ --- некоторый порог. 

Под словами <<редкое событие>> подразумевается, что условие внутри индикатора $f(x) \HM\ge \gamma$ выполняется с вероятностью, крайне близкой к нулю. Это означает, что Монте-Карло оценка с разумным на практике числом сэмплов $N$ выдаст или ноль или $\frac{1}{N}$, если один раз повезёт; короче, лобовой подход не годится.

Хочется сэмплировать $x$ не из того распределения, которое нам дали --- $p(x)$, --- а из чего-нибудь получше. Для этого применим \emph{importance sampling} с некоторым распределением $q(x)$, которое мы будем выбирать сами:
\begin{equation}\label{is_repe}
l = \E_{x \sim q(x)} \frac{p(x)}{q(x)} \mathbb{I}\left[ f(x) \ge \gamma \right]
\end{equation}

Нам хочется выбрать такое $q(x)$, чтобы дисперсия Монте-Карло оценки такого интеграла была как можно меньше. Желание может быть исполнено:
\begin{proposition}
Дисперсия Монте-Карло оценки \eqref{is_repe} минимальна при 
\begin{equation}\label{optimal_q_is_repe}
q(x) \propto p(x)\mathbb{I}\left[ f(x) \ge \gamma \right]
\end{equation}

\begin{proof}
Искомое значение $l$ \eqref{rareeventestimation} является нормировочной константой такого распределения. Подставим данное $q(x)$ в подынтегральную функцию:
$$
\frac{p(x)}{q(x)} \mathbb{I}\left[ f(x) \ge \gamma \right] = l \frac{p(x)\mathbb{I}\left[ f(x) \ge \gamma \right]}{p(x)\mathbb{I}\left[ f(x) \ge \gamma \right]} = l
$$
Поскольку всё сократилось, для любых сэмплов $x \sim q(x)$ значение Монте-Карло оценки будет равно $l$; то есть, дисперсия равна нулю.
\end{proof}
\end{proposition}

Посчитать такое $q(x)$ мы не можем, однако можем пытаться приблизить в параметрическом семействе $q(x \HM\mid \lambda)$, минимизируя, например, такую KL-дивергенцию\footnote{здесь и всюду далее под обозначением $\const(\lambda)$ мы будем подразумевать функции, константные относительно $\lambda$; такие слагаемые не влияют на оптимизацию по $\lambda$ и могут быть опущены.}:
$$\KL(q(x) \parallel q(x \mid \lambda)) = \const(\lambda) - \E_{q(x)} \log q(x \mid \lambda) \to \min_\lambda$$
Единственное зависящее от параметров $\lambda$ слагаемое называется \emph{кросс-энтропией} (cross entropy) и даёт название методу.

Пока что мы променяли шило на мыло, поскольку для такой оптимизации всё равно нужно уметь сэмплировать из $q(x)$. Однако от задачи оценки числа (которую мы кроме как через Монте-Карло особо решать не умеем) мы перешли к поиску распределения. Поскольку это распределение мы строили так, чтобы оно помогало сэмплировать нам точки из редкого события, можно воспользоваться им же с прошлой итерации, чтобы помочь самим себе решать ту же задачу лучше. А то есть: строим последовательность $q(x \HM\mid \lambda_k)$, $\lambda_0$ --- любое, на очередной итерации:
\begin{align*}
\lambda_{k+1} &= \argmin_{\lambda} -\E_{q(x)} \log q(x \mid \lambda) = \\
\{\text{подставляем вид оптимального $q(x)$ из \eqref{optimal_q_is_repe}}\} &= \argmin_{\lambda} -\E_{p(x)} \mathbb{I}\left[ f(x) \ge \gamma \right] \log q(x \mid \lambda) = \\
\{\text{importance sampling через $q(x \mid \lambda_{k})$}\} &= \argmin_{\lambda} -\E_{q(x \mid \lambda_{k})} \frac{p(x)}{q(x \mid \lambda_{k})}\mathbb{I}\left[ f(x) \ge \gamma \right] \log q(x \mid \lambda)
\end{align*}

Каждая задача нахождения $\lambda_k$ всё ещё тяжела в связи с тем, что подынтегральное выражение всё ещё почти всегда ноль. Ключевая идея: поскольку мы теперь строим целую последовательность, мы можем поначалу решать сильно более простую задачу, разогревая $\gamma$. Будем на $k$-ом шаге брать $\gamma$ не из условия задачи, а поменьше, так, чтобы с итерациями $\gamma$ увеличивалась (и мы решали бы задачу, всё более похожую на ту, что требовалось решить исходно), и одновременно достаточное число сэмплов значения подынтегральной функции были отличны от нуля.

Важно, что мы можем не задавать заранее последовательность $\gamma_k$, а определять очередное значение прямо на ходу, например, исходя из сэмплов $x_1 \dots x_N \sim q(x \mid \lambda_{k})$ и значений $f(x)$ в них.

\begin{algorithm}[label=alg:rareeventestimate]{Метод Кросс-Энтропии для оценки вероятности редкого события}
\textbf{Вход:} распределение $p(x)$, функция $f(x)$, порог $\gamma$ \\
\textbf{Гиперпараметры:} $q(x \mid \lambda)$ --- параметрическое семейство, $N$ --- число сэмплов, $M$ --- порог отбора

\vspace{0.3cm}
Инициализируем $\lambda_0$ произвольно. \\
\textbf{На $k$-ом шаге:}
\begin{enumerate}
    \item сэмплируем $x_1 \dots x_N \sim q(x \mid \lambda_k)$
    \item сортируем значения $f(x_i)$: $f_{(1)} \le f_{(2)} \le \dots \le f_{(N)}$
    \item полагаем $\gamma_k \coloneqq \min(\gamma, f_{(M)})$
    \item решаем задачу оптимизации:
    \begin{equation}\label{CEM_optimization_general}
    \lambda_{k+1} \leftarrow \argmax_{\lambda} \frac{1}{N} \sum_{j=1}^N \mathbb{I}[f(x_j) \ge \gamma_k]\frac{p(x_j)}{q(x_j \mid \lambda_k)} \log q(x_j \mid \lambda)
    \end{equation}
    \item \textbf{критерий останова:} $\gamma_k = \gamma$
\end{enumerate}

\textbf{Получение итоговой оценки:}
\begin{enumerate}
    \item сэмплируем $x_1 \dots x_N \sim q(x \mid \lambda_k)$
    \item возвращаем
    $$l \approx \frac{1}{N} \sum_{j=1}^N \mathbb{I}[f(x_j) \ge \gamma_k]\frac{p(x_j)}{q(x_j \mid \lambda_k)}$$
\end{enumerate}
\end{algorithm}

\subsection{Метод Кросс-Энтропии для стохастической оптимизации}

Ну, в рассуждении было видно, что мы практически учим $q(x \mid \lambda)$ нащупывать область с высоким значением заданной функции без использования какой-либо информации о ней. Поэтому мы можем адаптировать метод, чтобы он стал мета-эвристикой. Для этого вернёмся к нашей задаче безградиентной оптимизации:
$$J(\theta) \to \max_{\theta}$$
и перепишем алгоритм \ref{alg:rareeventestimate} в условиях, когда порог $\gamma$ <<не ограничен>>, ну или что тоже самое, $\gamma \HM\coloneqq \max\limits_{\theta} J(\theta)$. Формально мы также можем выбирать любое $p(x)$; положим $p(x) \coloneqq q(x \mid \lambda_k)$, просто чтобы в задаче \eqref{CEM_optimization_general} сократилась importance sampling коррекция. Мы получим очень простой на вид алгоритм, в котором фактически на очередном шаге минимизируется такое расстояние:
$$\KL(\mathbb{I}[J(x) \ge \gamma_k] q(x \mid \lambda_{k-1}) \parallel q(x \mid \lambda_k)) \to \min_{\lambda_k},$$
где первое распределение задано с точностью до нормировочной константы.

\begin{algorithm}{Метод Кросс-Энтропии для оптимизации с оракулом нулевого порядка}
\textbf{Вход:} оракул $\hat{J}(\theta)$ \\
\textbf{Гиперпараметры:} $q(\theta \mid \lambda)$ --- параметрическое семейство, $N$ --- число сэмплов, $M$ --- порог отбора

\vspace{0.3cm}
Инициализируем $\lambda_0$ произвольно. \\
\textbf{На $k$-ом шаге:}
\begin{enumerate}
    \item сэмплируем $\Pop_k \coloneqq \left( \theta_i \sim q(\theta \mid \lambda_k) \mid i \in \{1, 2, \dots, N\} \right)$
    \item проводим отбор $\Pop^{+}_k \coloneqq \Seltop_M(\Pop_k)$
    \item решаем задачу оптимизации:
    $$\lambda_{k+1} \leftarrow \argmax_{\lambda} \sum_{\theta \in \Pop^{+}_k} \log q(\theta \mid \lambda)$$
\end{enumerate}
\end{algorithm}

Видно, что мы по сути действуем эволюционно: хотим генерировать при помощи распределения $q$ точки из области, где значение функции велико; берём и сэмплируем несколько точек из текущего приближения; из сгенерированных отбираем те, где значение функции было наибольшим и учим методом максимального правдоподобия повторять эти точки. Поскольку некоторая доля плохих точек была выкинута из выборки, распределение, которое учит очередное $q(x \mid \lambda_k)$, лучше предыдущего. Это первый способ обучения эволюционных стратегий.

\begin{example}[Кросс-энтропийный метод для black-box оптимизации]
\begin{center}
\animategraphics[controls, width=\linewidth]{1}{Images/CEM/cem-}{0}{9}
\end{center}
\end{example}

\subsection{Метод Кросс-Энтропии для обучения с подкреплением (CEM)}

В обучении с подкреплением в кросс-энтропийном методе можно сделать ещё один очень интересный шаг. В отличие от всех остальных рассматриваемых в этой главе мета-эвристик, мы можем проводить эволюционный отбор не в пространстве возможных стратегий (в пространстве $\Theta$), а в пространстве траекторий. У нас будет одна текущая стратегия, из которой мы сгенерируем несколько траекторий, и в силу стохастичности некоторые из этих траекторий выдадут лучший результат, чем другие. Мы отберём лучшие и будем методом максимального правдоподобия (по сути, имитационным обучением) учиться повторять действия из лучших траекторий.  

\begin{algorithm}[label=alg:cem]{Cross Entropy Method}
\textbf{Гиперпараметры:} $\pi(a \mid s, \theta)$ --- стратегия с параметрами $\theta$, $N$ --- число сэмплов, $M$ --- порог отбора

\vspace{0.3cm}
Инициализируем $\theta_0$ произвольно. \\
\textbf{На $k$-ом шаге:}
\begin{enumerate}
    \item сэмплируем $N$ траекторий $\Traj_1 \dots \Traj_N$ игр при помощи стратегии $\pi(a \mid s, \theta_k)$
    \item считаем кумулятивные награды $R(\Traj_i)$
    \item сортируем значения: $R_{(1)} \le R_{(2)} \le \dots \le R_{(N)}$
    \item полагаем $\gamma_k \coloneqq R_{(M)}$
    \item решаем задачу оптимизации:
    $$\theta_{k+1} \leftarrow \argmax_{\theta} \frac{1}{N} \sum_{j=1}^N \mathbb{I}[R(\Traj_j) \ge \gamma_k] \sum_{s, a \in \Traj_j} \log \pi(a \mid s, \theta)$$
\end{enumerate}
\end{algorithm}


\subsection{Натуральные эволюционные стратегии (NES)}\label{subsec:nes}

Рассмотрим альтернативный вариант\footnote{смысл названия <<натуральные эволюционные стратегии>> (natural evolution strategies) будет объяснён позже.} подбора параметров эволюционной стратегии $q(\theta \mid \lambda)$. Будем подбирать $\lambda$, исходя из следующего функционала:
\begin{equation}\label{es}
g(\lambda) \coloneqq \E_{\theta \sim q(\theta \mid \lambda)} J(\theta) \to \max_{\lambda}
\end{equation}

Будем оптимизировать этот функционал градиентно по $\lambda$. Давайте подробно разберём, как дифференцировать функции подобного вида, поскольку в дальнейшем мы будем активно пользоваться этой техникой.

\begin{theorem}
\begin{equation}\label{ESgradient}
    \nabla_\lambda g(\lambda) = \E_{\theta \sim q(\theta \mid \lambda)} \nabla_\lambda \log q(\theta \mid \lambda) J(\theta)
\end{equation}
\beginproof
\begin{align*}
\nabla_\lambda g(\lambda) &= \nabla_\lambda \E_{\theta \sim q(\theta \mid \lambda)} J(\theta) = \\
= \{ \text{мат.ожидание --- это интеграл} \} &= 
\nabla_\lambda \int\limits_\Theta q(\theta \mid \lambda)J(\theta) \diff \theta = \\
= \{ \text{проносим градиент внутрь интеграла} \} &=
\int\limits_\Theta \nabla_\lambda q(\theta \mid \lambda)J(\theta) \diff \theta = (*)
\end{align*}

Теперь мы применим стандартный технический трюк, называемый \emph{log-derivative trick}: мы хотим преобразовать данное выражение к виду мат.ожидания по $q(\theta \mid \lambda)$ от чего-то. Как мы сейчас увидим, это <<что-то>> --- градиент логарифма правдодобия. Мы воспользуемся следующим тождеством:
\begin{equation}\label{logderivtrick}
\nabla_\lambda \log q(\theta \mid \lambda) = \frac{\nabla_\lambda q(\theta \mid \lambda)}{q(\theta \mid \lambda)}
\end{equation}

Домножим и поделим наше выражение на $q(\theta \mid \lambda)$, чтобы получить внутри интеграла мат.ожидание:
\begin{align*}
(*) &= \int\limits_\Theta q(\theta \mid \lambda) \frac{\nabla_\lambda q(\theta \mid \lambda)}{q(\theta \mid \lambda)} J(\theta) \diff \theta = \\
= \{ \text{замечаем градиент логарифма \eqref{logderivtrick} } \}
&= \int\limits_\Theta q(\theta \mid \lambda) \nabla_\lambda \log q(\theta \mid \lambda) J(\theta) \diff \theta = \\
= \{ \text{выделяем мат.ожидание} \}
&= \E_{\theta \sim q(\theta \mid \lambda)} \nabla_\lambda \log q(\theta \mid \lambda) J(\theta) \tagqed
\end{align*}

\end{theorem}

Итак, есть следующая идея: сгенерируем популяцию $\Pop_k$ при помощи $q(\theta \mid \lambda_k)$, после чего воспользуемся особями как сэмплами для несмещённой оценки градиента функционала \eqref{es}, чтобы улучшить параметры $\lambda$ и впоследствии сгенерировать следующее поколение $\Pop_{k+1}$ из более хорошего распределения.



\subsection{OpenAI-ES}

Рассмотрим подход на примере OpenAI-ES, где рассматривается обучение нейросети (с фиксированной топологией) с вещественными весами $\Theta \equiv \R^h$, и полагается
\begin{equation}\label{openai_es}
q(\theta \mid \lambda) \coloneqq \N(\lambda, \sigma^2I_{h \times h})
\end{equation}
где $\sigma$ --- гиперпараметр, $\lambda \in \R^h$ --- по сути, кодирует одну особь, $I_{h \times h}$ --- диагональная единичная матрица размера $h \times h$.

\begin{theorem}
Для эволюционной стратегии \eqref{openai_es} оценка градиента \eqref{es} равна
\begin{equation}\label{openai_es_gradient}
\nabla_\lambda g(\lambda) = \frac{1}{N\sigma^2} \sum_{\theta \in \Pop} \hat{J}(\theta) (\theta - \lambda)
\end{equation}
\begin{proof}
$$\nabla_\lambda \log q(\theta \mid \lambda) = -\nabla_\lambda \frac{\left(\theta - \lambda \right)^T \left( \theta - \lambda \right)}{2 \sigma^2} = \frac{\theta - \lambda}{\sigma^2}$$
Достаточно подставить выражение в общую формулу \eqref{ESgradient}.
\end{proof}
\end{theorem}

Полученная формула легко интерпретируема: мы находимся в некотором <<центре>> $\lambda$, который является нашим <<текущим найденным решением>>. Дальше мы сэмплируем несколько векторов $\theta - \lambda$ из стандартного нормального распределения и складываем эти вектора (<<скрещиваем всех детей>>) с весами, пропорциональными приспособленности.


\begin{remark}
Подход работает при большом количестве серверов и ещё одном трюке, позволяющем не обмениваться векторами $\theta$ (имеющими размерность, например, по числу параметров нейросети) между процессами. Для этого достаточно зафиксировать random seed на всех процессорах, в каждом процессе генерировать всё поколение, оценивать только особь с соответствующим процессу номеру и обмениваться с другими процессами исключительно оценками $\hat{J}$ (скалярами!). Цель такой процедуры --- избежать обмена весами нейросетей (пусть даже не очень больших) между серверами. За счёт параллелизации удаётся так <<обучать>> сетки играть в одну игру Атари за 10 минут. Если у вас есть 1440 процессоров. Естественно, ни про какой sample efficiency речь не идёт. 
\end{remark}

\begin{example}[OpenAI-ES]
\begin{center}
\animategraphics[controls, width=\linewidth]{1}{Images/ES/es-}{0}{5}
\end{center}
\end{example}

Рассмотрим альтернативный взгляд на этот алгоритм. Допустим, мы находимся в точке $\theta$ и хотим сдвинуться как бы по градиенту $J(\theta)$, который вычислить мы не можем. Но мы можем приблизить градиент вдоль любого направления $\nu$, например, так:
$$ \left. \nabla \right|_{\nu} J(\theta) \approx \frac{\hat{J}(\theta + \sigma \nu) - \hat{J}(\theta)}{\sigma}$$
для некоторого небольшого скаляра $\sigma$.

Лобовая идея\footnote{в алгоритме ARS (Advanced Random Search, хотя такой подход не совсем <<случайный поиск>>) делается ровно это, за тем небольшим исключением, что используется приближение градиента по направлению за два вызова оракула:
$$ \left. \nabla \right|_{\nu} J(\theta) \approx \frac{\hat{J}(\theta + \sigma \nu) - \hat{J}(\theta - \sigma \nu)}{2\sigma}$$}: давайте возьмём $N$ случайных направлений $\nu_1 \dots \nu_N \sim \N(0, I)$ и сделаем шаг по всем этим направлениям:
\begin{equation}\label{ars}
\theta_{k+1} \coloneqq \theta_k + \alpha \underbrace{\sum_{i=0}^N \frac{\hat{J}(\theta + \sigma \nu_i) - \hat{J}(\theta)}{\sigma}\nu_i}_{\text{приближение градиента}}
\end{equation}
где $\alpha$ --- learning rate.

\begin{proposition}
Формулы \eqref{ars} и \eqref{openai_es_gradient} эквивалентны.
\begin{proof}
Поскольку $\nu$ сэмплируется из стандартной гауссианы, то в среднем:
$$\E_{\nu \sim \N(0, I)} \frac{\hat{J}(\theta)}{\sigma} \nu = \frac{\hat{J}(\theta)}{\sigma} \E_{\nu \sim \N(0, I)} \nu = 0$$
Следовательно, \eqref{ars} оценивает тот же градиент, что и формула
$$\theta_{k+1} \coloneqq \theta_k + \alpha \sum_{i=0}^N \frac{\hat{J}(\theta + \sigma \nu_i)}{\sigma}\nu_i,$$
что совпадает с формулой OpenAI-ES с точностью до замены обозначений: достаточно заметить, что $\theta + \sigma \nu$ есть сэмпл из $\N(\theta, \sigma^2 I)$.
\end{proof}
\end{proposition}

\subsection{Адаптация матрицы ковариации (CMA-ES)}

Более глубокомысленно было бы для $\Theta \equiv \R^{h}$ адаптировать не только среднее, но и матрицу ковариации, которая имеет смысл <<разброса>> очередной популяции. Covariance Matrix Adaptation Evolution Strategy (CMA-ES) --- алгоритм, включающий довольной большой набор эвристик, в основе которого лежит эволюционная стратегия, адаптирующая не только среднее, но и матрицу ковариации:
\begin{equation}\label{covadapt}
q(\theta \mid \lambda) \coloneqq \N(\mu, \Sigma)
\end{equation}
где $\lambda \coloneqq (\mu, \Sigma)$.

\begin{remark}
Нам придётся хранить матрицу ковариации размера $\R^{h \times h}$, где $h$ --- количество параметров нашей стратегии ($\Theta \HM\equiv \R^h$). Если стратегия задана нейронной сетью, то, чтобы такое было возможно, сеть должна быть по современным меркам минимальнейшей. Однако, во многих задачах непрерывного управления это довольно типичная ситуация, когда на вход в качестве состояния подаётся небольшой (размера 100-200) вектор, а на выходе также ожидается вектор размера порядка 10-30. Тогда стратегия может быть задана полносвязной нейросетью всего в несколько слоёв, и хранить $\Sigma$ теоретически становится возможно. 
\end{remark}

Мы рассмотрим только основную часть формул алгоритма, касающихся формулы обновления $\Sigma$. Посмотрим на формулу для обновления среднего \eqref{openai_es_gradient} (с точностью до learning rate):
\begin{equation}\label{cmaes_mu_update}
\mu_{k+1} = \mu_k + \alpha\sum_{\theta \in \Pop_k} \underbrace{\hat{J}(\theta)}_{\text{вес}} \underbrace{\left( \theta - \mu_k \right)}_{\text{\shortstack{<<предлагаемое>> \\ особью изменение}}}
\end{equation}

Для обновления матрицы ковариации будем рассуждать также: каждая особь популяции $\theta$ <<указывает>> на некоторую ковариацию $\left( \theta - \mu_k \right)\left( \theta - \mu_k \right)^T$ и таким образом <<предлагает>> следующее изменение:
$$\left( \theta - \mu_k \right)\left( \theta - \mu_k \right)^T - \Sigma_k,$$
где $\Sigma_k$ --- матрица ковариации на текущей итерации. Усредним эти <<предложения изменения>> по имеющейся популяции, взвесив их на $\hat{J}(\theta)$, и получим <<градиент>> для обновления матрицы:
\begin{equation}\label{cmaes_sigma_update}
\Sigma_{k+1} \coloneqq \Sigma_k + \alpha \sum_{\theta \in \Pop_k} \hat{J}(\theta) \left( \left( \theta - \mu_k \right)\left( \theta - \mu_k \right)^T - \Sigma_k \right)
\end{equation}

Здесь нужно оговориться, что мы используем оценку ковариации как бы <<методом максимального правдоподобия при условии известного среднего $\mu_k$>> (мы знаем, что именно с таким средним генерировались особи прошлой популяции)\footnote{мы бы пришли к немного другим формулам, если бы использовали метод максимального правдоподобия при условии неизвестного среднего ($\mu_k$ заменилось бы на $\mu_{k+1}$):
$$\left( \theta - \mu_{k+1} \right)\left( \theta - \mu_{k+1} \right)^T - \Sigma_k$$}. Исторически к этим формулам пришли эвристически, но позже у формулы появилось теоретическое обоснование.

\begin{theorem}
Формулы \eqref{cmaes_mu_update} и \eqref{cmaes_sigma_update} для обновления $\lambda = (\mu, \Sigma)$ являются формулами натурального градиентного спуска для \eqref{es}.
\begin{proof}[Доказательство требует обсуждения такой большой темы, как натуральный градиентный спуск (см. приложение \ref{appendix:ng}), а вывод формулы потребует небольшого введения в Кронекерову алгебру, поэтому доказательство вынесено в приложение \ref{appendix:cmaes}] 
\end{proof}
\end{theorem}

Именно поэтому данный вид алгоритмов для обучения эволюционных стратегий называется <<\emph{натуральными}>> (natural): считается, что функционал \eqref{es} корректнее оптимизировать именно при помощи натурального градиентного спуска, инвариантного к параметризации $q(\theta \mid \lambda)$, а не обычного.

\begin{example}[CMA-ES (упрощ.)]
\needspace{10\baselineskip}
\begin{center}
\animategraphics[controls, width=\linewidth]{1}{Images/CMA_ES/cma_es-}{0}{9}
\end{center}
\end{example}

\begin{example}
Эволюционные стратегии и особенно CMA-ES весьма успешно применимы в задачах непрерывного управления вроде Locomotion (пример \ref{ex:locomotion}), в которых нужно научить разных существ ходить. Много интересных примеров можно найти в \href{https://www.youtube.com/watch?v=pgaEE27nsQw}{этом видео}.
\end{example}

Если мы попробуем проделать с данным подходом (оптимизацией \eqref{es}) тот же трюк, что и с кросс-энтропийным методом, и попытаемся считать градиент <<в пространстве траекторий>>, а не в пространстве стратегий, то получим методы оптимизации $J(\theta)$ уже первого порядка --- <<policy gradient>> методы. К ним мы перейдём в главе \ref{policygradientchapter}.

\chapter{Классическая теория}\label{classictheorychapter}

В данной главе будут доказаны основные теоретические результаты о MDP и получены важные <<табличные>> алгоритмы, работающие в случае конечных пространств состояний и действий; они лягут в основу всех дальнейших алгоритмов. 

\section{Оценочные функции}\label{valuefunctionssection}

\subsection{Свойства траекторий}

Как это всегда бывает, чем более общую задачу мы пытаемся решать, тем менее эффективный алгоритм мы можем придумать. В RL мы сильно замахиваемся: хотим построить алгоритм, способный обучаться решению <<произвольной>> задачи, заданной средой с описанной функцией награды. Однако в формализме MDP в постановке мы на самом деле внесли некоторые ограничения: марковость и стационарность. Эти предположения практически не ограничивают общность нашей задачи с точки зрения здравого смысла с одной стороны и при этом вносят в нашу задачу некоторую <<структуру>>; мы сможем придумать более эффективные алгоритмы решения за счёт эксплуатации этой структуры. 

Что значит <<структуру>>? Представим, что мы решаем некоторую абстрактную задачу последовательного принятия решения, максимизируя некоторую кумулятивную награду. Вот мы находимся в некотором состоянии и должны выбрать некоторое действие. Интуитивно ясно, что на прошлое --- ту награду, которую мы уже успели собрать --- мы уже повлиять не можем, и нужно максимизировать награду в будущем. Более того, мы можем отбросить всю нашу предыдущую историю и задуматься лишь над тем, как максимизировать награду с учётом сложившейся ситуации --- <<текущего состояния>>. 

\begin{example}[Парадокс обжоры]
Обжора пришёл в ресторан и заказал кучу-кучу еды. В середине трапезы выяснилось, что оставшиеся десять блюд явно лишние и в него уже не помещаются. Обидно: они будут в счёте, да и не пропадать же еде, поэтому надо бы всё равно всё съесть. Однако, с точки зрения функции награды нужно делать противоположный вывод: блюда будут в счёте в любом случае, вне зависимости от того, будут ли они съедены --- это награда за уже совершённое действие, <<прошлое>>, --- а вот за переедание может прилететь ещё отрицательной награды. Обжора понимает, что в прошлом совершил неоптимальное действие, и пытается <<прооптимизировать>> неизбежную награду за прошлое, в результате проигрывая ещё.
\end{example}

Давайте сформулируем эту интуицию формальнее. Как и в обычных Марковских цепях, в средах благодаря марковости действует закон <<независимости прошлого и будущего при известном настоящем>>. Формулируется он так: 

\begin{proposition}[Независимость прошлого и будущего при известном настоящем]\label{pr:futurepastindependent}
Пусть $\Traj_{:t} \HM\coloneqq \{s_0, a_0 \dots s_{t-1}, a_{t-1} \}$ --- <<прошлое>>, $s_t$ --- <<настоящее>>, $\Traj_{t:} \HM\coloneqq \{a_t, s_{t+1}, a_{t+1} \dots \}$ --- <<будущее>>. Тогда:
$$p\left( \Traj_{:t}, \Traj_{t:} \mid s_t \right) = p\left( \Traj_{:t} \mid s_t \right)p\left( \Traj_{t:} \mid s_t \right)$$

\beginproof 
По правилу произведения:
$$p\left( \Traj_{:t}, \Traj_{t:} \mid s_t \right) = p\left( \Traj_{:t} \mid s_t \right)p\left( \Traj_{t:} \mid s_t, \Traj_{:t} \right)$$
Однако в силу марковости будущее зависит от настоящего и прошлого только через настоящее:
\begin{equation*}
p\left( \Traj_{t:} \mid s_t, \Traj_{:t} \right) = p\left( \Traj_{t:} \mid s_t \right) \tagqed
\end{equation*}
\end{proposition}

Для нас утверждение означает следующее: если мы сидим в момент времени $t$ в состоянии $s$ и хотим посчитать награду, которую получим в будущем (то есть величину, зависящую только от $\Traj_{t:}$), то нам совершенно не важна история попадания в $s$. Это следует из свойства мат.~ожиданий по независимым переменным:
$$\E_{\Traj \mid s_t = s} R(\Traj_{t:}) = \{ \text{утв. \ref{pr:futurepastindependent}} \} = \E_{\Traj_{:t} \mid s_t = s} \underbrace{\E_{\Traj_{t:} \mid s_t = s} R(\Traj_{t:})}_{\text{не зависит от $\Traj_{:t}$}} = \E_{\Traj_{t:} \mid s_t = s} R(\Traj_{t:})$$

\begin{definition}
Для траектории $\Traj$ величина
\begin{equation}\label{rewardtogo}
R_t \coloneqq R \left( \Traj_{t:} \right) = \sum_{\hat{t} \ge t} \gamma^{\hat{t} - t} r_{\hat{t}}
\end{equation}
называется \emph{reward-to-go} с момента времени $t$.
\end{definition}

Благодаря второму сделанному предположению, о стационарности (в том числе стационарности стратегии агента), получается, что будущее также не зависит от текущего момента времени $t$: всё определяется исключительно текущим состоянием. Иначе говоря, агенту неважно не только, как он добрался до текущего состояния и сколько награды встретил до настоящего момента, но и сколько шагов в траектории уже прошло. Формально это означает следующее: распределение будущих траекторий имеет в точности тот же вид, что и распределение всей траектории при условии заданного начала.

\begin{proposition}\label{pr:timeindepend} Будущее определено текущим состоянием:
$$p\left( \Traj_{t:} \mid s_t = s \right) \equiv p\left( \Traj \mid s_0 = s \right) $$
\beginproof
По определению:
$$
p\left( \Traj_{t:} \mid s_t = s \right) = \prod_{\hat{t} \ge t} p(s_{\hat{t}+1} \mid s_{\hat{t}}, a_{\hat{t}})\pi(a_{\hat{t}} \mid s_{\hat{t}}) = (*)
$$
Воспользуемся однородностью MDP и однородностью стратегии, а именно:
\begin{gather*}
\pi(a_{\hat{t}} \mid s_{\hat{t}}=s) = \pi(a_0 \mid s_0=s) \\
p(s_{\hat{t}+1} \mid s_{\hat{t}}=s, a_{\hat{t}}) = p(s_1 \mid s_0=s, a_0) \\
\pi(a_{\hat{t}+1} \mid s_{\hat{t}+1}) = \pi(a_1 \mid s_1) \\
p(s_{\hat{t}+2} \mid s_{\hat{t}+1}, a_{\hat{t}+1}) = p(s_2 \mid s_1, a_1)
\end{gather*}
и так далее, получим:
\begin{equation*}
(*) = \prod_{t \ge 0} p(s_{t+1} \mid s_{t}, a_{t})\pi(a_{t} \mid s_{t}) = p\left( \Traj \mid s_0 = s \right) \tagqed
\end{equation*}
\end{proposition}

\begin{proposition}\label{pr:timeindependexpectation}
Для любого $t$ и любой функции $f$ от траекторий:
$$\E_{\Traj \mid s_0 = s} f(\Traj) = \E_{\Traj \mid s_t = s} f(\Traj_{t:})$$
\beginproof
\begin{align*}
\E_{\Traj \mid s_0 = s} f(\Traj) = \{ \text{утв. \ref{pr:timeindepend}} \} = \E_{\Traj_{t:} \mid s_t = s} f(\Traj_{t:}) = \{ \text{утв. \ref{pr:futurepastindependent}} \} = \E_{\Traj \mid s_t = s} f(\Traj_{t:}) \tagqed 
\end{align*}
\end{proposition}

Мы показали, что все свойства reward-to-go определяются исключительно стартовым состоянием.

\subsection{V-функция}

Итак, наша интуиция заключается в том, что, когда агент приходит в состояние $s$, прошлое не имеет значения, и оптимальный агент должен максимизировать в том числе и награду, которую он получит, стартуя из состояния $s$. Поэтому давайте <<обобщим>> наш оптимизируемый функционал, варьируя стартовое состояние:

\begin{definition} 
Для данного MDP \emph{V-функцией} (value function) или оценочной функцией состояний (state value function) для данной стратегии $\pi$ называется величина 
\begin{equation}\label{Vdefinition}
V^\pi(s) \coloneqq \E_{\Traj \sim \pi \mid s_0 = s} R \left( \Traj \right)
\end{equation}
\end{definition}

По определению функция ценности состояния, или V-функция --- это сколько набирает в среднем агент из состояния $s$. Причём в силу марковости и стационарности неважно, случился ли старт на нулевом шаге эпизода или на произвольном $t$-ом:

\begin{proposition}\label{Vtimeindepend}
Для любого $t$ верно:
$$V^\pi(s) = \E_{\Traj \sim \pi \mid s_t = s} R_t$$
\begin{proof}[Пояснение] Применить утверждение \ref{pr:timeindependexpectation} для $R(\Traj)$.
\end{proof}
\end{proposition}

\begin{proposition}
$V^\pi(s)$ ограничено.
\end{proposition}

\begin{proposition}
Для терминальных состояний $V^\pi(s) \HM= 0$.
\end{proposition}

Заметим, что любая политика $\pi$ индуцирует $V^\pi$. То есть для данного MDP и данной стратегии $\pi$ функция $V^\pi$ однозначно задана своим определением; совсем другой вопрос, можем ли мы вычислить эту функцию.

\begin{exampleBox}[label=ex:vfunction]{}
Посчитаем V-функцию для MDP и стратегии $\pi$ с рисунка, $\gamma \HM= 0.8$. Её часто удобно считать <<с конца>>, начиная с состояний, близких к терминальным, и замечая связи между значениями функции для разных состояний.

\begin{wrapfigure}{r}{0.4\textwidth}
\vspace{-0.7cm}
\centering
\includegraphics[width=0.4\textwidth]{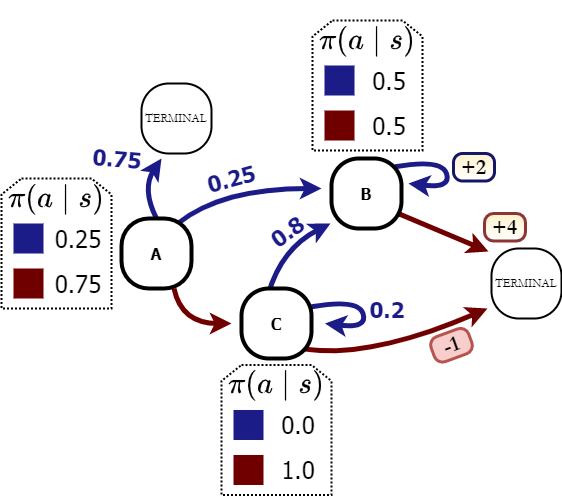}
\vspace{-1.5cm}
\end{wrapfigure}

Начнём с состояния C: там агент всегда выбирает действие \colorsquare{ChadRed}, получает -1, и эпизод заканчивается: $V^{\pi}(s \HM= C) \HM= -1$.

Для состояния B с вероятностью 0.5 агент выбирает действие \colorsquare{ChadRed} и получает +4. Иначе он получает +2 и возвращается снова в состояние B. Вся дальнейшая награда будет дисконтирована на $\gamma \HM= 0.8$ и тоже равна $V^{\pi}(s = B)$ по определению. Итого:
$$V^{\pi}(s = B) = \underbrace{0.5 \cdot 4}_{\text{\colorsquare{ChadBlue}}} + \underbrace{0.5 \cdot \left(2 + \gamma V^{\pi}(s = B)\right)}_{\text{\colorsquare{ChadRed}}}$$
Решая это уравнение относительно $V^{\pi}(s \HM= B)$, получаем ответ 5.

Для состояния A достаточно аналогично рассмотреть все дальнейшие события:
$$V^{\pi}(s = A) = \underbrace{0.25}_{\text{\colorsquare{ChadBlue}}} \cdot \left(\underbrace{0.75 \cdot 0}_{\text{terminal}} + \underbrace{0.25 \gamma V^{\pi}(s = B)}_{B} \right) + \underbrace{0.75}_{\text{\colorsquare{ChadRed}}} \underbrace{\gamma V^{\pi}(s = C)}_{C}$$
Подставляя значения, получаем ответ $V^{\pi}(s \HM= A) \HM= -0.35$.
\end{exampleBox}

\subsection{Уравнения Беллмана}

Если $s_0$ --- стартовое состояние, то $V^\pi(s_0)$ по определению и есть функционал \eqref{goal}, который мы хотим оптимизировать. Формально, это единственная величина, которая нас действительно волнует, так как она нам явно задана в самой постановке задачи, но мы понимаем, что для максимизации $V^\pi(s_0)$ нам нужно промаксимизировать и $V^{\pi}(s)$ (строго мы это пока не показали). Другими словами, у нас в задаче есть \emph{подзадачи эквивалентной структуры}: возможно, они, например, проще, и мы можем сначала их решить, а дальше как-то воспользоваться этими решениями для решения более сложной. Вот если граф MDP есть дерево, например, то очевидно, как считать $V^\pi$: посчитать значение в листьях (листья соответствуют терминальным состояниям --- там ноль), затем в узлах перед листьями, ну и так далее индуктивно добраться до корня.

Мы заметили, что в примере \ref{ex:vfunction} на значения V-функции начали появляться рекурсивные соотношения. В этом и есть смысл введения понятия оценочных функций --- <<дополнительных переменных>>: в том, что эти значения связаны между собой \emph{уравнениями Беллмана} (Bellman equations).

\begin{theorem}[Уравнение Беллмана (Bellman expectation equation) для $V^\pi$]
\,
\begin{equation}\label{VV}
V^\pi(s) = \E_{a} \left[ r(s, a) + \gamma \E_{s'} V^\pi(s') \right]
\end{equation}

\needspace{10\baselineskip}
\begin{wrapfigure}{r}{0.4\textwidth}
\vspace{-0.5cm}
\centering
\includegraphics[width=0.4\textwidth]{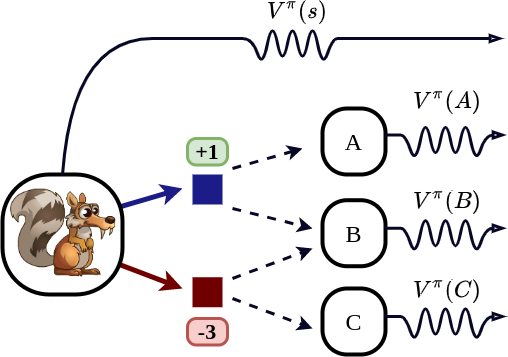}
\end{wrapfigure}
\beginproof
Интуиция: награда за игру равна награде за следующий шаг плюс награда за оставшуюся игру; награда за хвост равна следующей награде плюс награда за хвост. Действительно, для всех траекторий $\Traj$ и для любых $t$ верно:
$$R_t = r_t + \gamma R_{t+1}$$

Соответственно, для формального доказательства раскладываем сумму по времени как первое слагаемое плюс сумма по времени и пользуемся утверждением \ref{Vtimeindepend} о независимости V-функции от времени:
\begin{align*}
V^\pi(s) = \E_{\Traj \mid s_t = s} R_t &= \E_{a_t} \left[ r_t + \gamma \E_{s_{t+1}} \E_{\Traj \sim \pi \mid s_{t+1}} R_{t+1} \right] = \\ &= \E_{a} \left[ r(s, a) + \gamma \E_{s'} V^\pi(s') \right] \tagqed
\end{align*}
\end{theorem}

\begin{example}
Выпишем уравнения Беллмана для MDP и стратегии $\pi$ из примера \ref{ex:vfunction}. Число уравнений совпадает с числом состояний. Разберём подробно уравнение для состояния $A$:

$$
V^{\pi}(A) = \underbrace{0.25}_{\colorsquare{ChadBlue}} (0 + \gamma 0.25 V^{\pi}(B)) + \underbrace{0.75}_{\colorsquare{ChadRed}} (0 + \gamma V^{\pi}(C) )
$$

\begin{wrapfigure}{r}{0.45\textwidth}
\vspace{-0.8cm}
\centering
\includegraphics[width=0.45\textwidth]{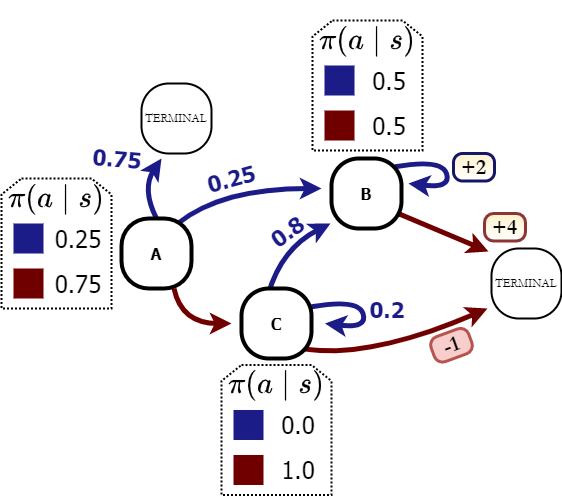}
\vspace{-1cm}
\end{wrapfigure}

С вероятностью 0.25 будет выбрано действие \colorsquare{ChadBlue}, после чего случится дисконтирование на $\gamma$; с вероятностью 0.75 эпизод закончится и будет выдана нулевая награда, с вероятностью 0.25 агент перейдёт в состояние B. Второе слагаемое уравнения будет отвечать выбору действия \colorsquare{ChadRed}; агент тогда перейдёт в состояние C и, начиная со следующего шага, получит в будущем $V^{\pi}(C)$. Аналогично расписываются два оставшихся уравнения.

\vspace{-0.4cm}
\begin{align*}
V^{\pi}(A) &= \frac{1}{16} \gamma V^{\pi}(B) + \frac{3}{4} \gamma V^{\pi}(C) \\
V^{\pi}(B) &= 0.5 \left(2 + \gamma V^{\pi}(B) \right) + 0.5 \cdot 4 \\
V^{\pi}(C) &= -1
\end{align*}

Заметим, что мы получили систему из трёх линейных уравнений с тремя неизвестными.
\end{example}

Позже мы покажем, что $V^\pi$ является единственной функцией $\St \HM\to \R$, удовлетворяющей уравнениям Беллмана для данного MDP и данной стратегии $\pi$, и таким образом однозначно ими задаётся.

\subsection{Оптимальная стратегия}

У нас есть конкретный функционал $J(\pi) \HM= V^{\pi}(s_0)$, который мы хотим оптимизировать. Казалось бы, понятие оптимальной политики очевидно как вводить:
\begin{definition} 
Политика $\pi^*$ \emph{оптимальна}, если $\forall \pi \colon V^{\pi^*}(s_0) \ge V^\pi(s_0)$.
\end{definition}

Введём альтернативное определение:

\begin{definition} 
Политика $\pi^*$ \emph{оптимальна}, если $\forall \pi, s \colon V^{\pi^*}(s) \ge V^\pi(s)$.
\end{definition}

\begin{theorem}
Определения не эквивалентны.

\needspace{10\baselineskip}
\begin{wrapfigure}{r}{0.3\textwidth}
\centering
\vspace{-0.6cm}
\includegraphics[width=0.3\textwidth]{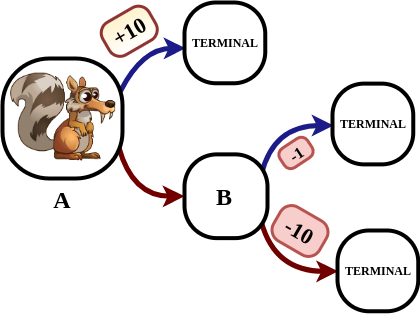}
\end{wrapfigure}
\beginproof{}
Из первого не следует второе (из второго первое, конечно, следует). Контрпример приведён на рисунке. С точки зрения нашего функционала, оптимальной будет стратегия сразу выбрать \colorsquare{ChadBlue} и закончить игру. Поскольку оптимальный агент выберет \colorsquare{ChadRed} с вероятностью 0, ему неважно, какое решение он будет принимать в состоянии $B$, в котором он никогда не окажется. Согласно первому определению, оптимальная политика может действовать в $B$ как угодно. Однако, чтобы быть оптимальной согласно второму определению и в том числе максимизировать $V^\pi(s = B)$, стратегия обязана выбирать в $B$ только действие \colorsquare{ChadBlue}. \QED
\end{theorem}

Интуиция подсказывает, что различие между определениями проявляется только в состояниях, которые оптимальный агент будет избегать с вероятностью 1 (позже мы увидим, что так и есть). Задавая оптимальность вторым определением, мы чуть-чуть усложняем задачу, но упрощаем теоретический анализ: если бы мы оставили первое определение, у оптимальных политик могли бы быть разные V-функции (см. пример из последнего доказательства); согласно второму определению, V-функция всех оптимальных политик совпадает. 
\begin{definition}
Оптимальные стратегии будем обозначать $\pi^*$, а соответствующую им \emph{оптимальную V-функцию} --- $V^*$:
\begin{equation}\label{optimalVdefinition}
V^*(s) = \max_\pi V^{\pi}(s)
\end{equation}
\end{definition}

Пока нет никаких обоснований, что найдётся стратегия, которая максимизирует $V^{\pi}(s)$ сразу для всех состояний. Вдруг в одних $s$ макисмум \eqref{optimalVdefinition} достигается на одной стратегии, а в другом --- на другой? Тогда оптимальных стратегий в сильном смысле вообще не существует, хотя формальная величина \eqref{optimalVdefinition} существует. Пока заметим лишь, что для ситуации, когда MDP --- дерево, существование оптимальной стратегии в смысле второго определения можно опять показать <<от листьев к корню>>. 

\begin{example}
Рассмотрим MDP из примера \ref{ex:score}; $\gamma \HM= \frac{10}{11}$, множество стратегий параметризуется единственным числом $\theta \HM\coloneqq \pi(a \HM= \text{\colorsquare{ChadRed}} \HM\mid s \HM= A)$.

\begin{wrapfigure}{r}{0.25\textwidth}
\centering
\vspace{-0.8cm}
\includegraphics[width=0.25\textwidth]{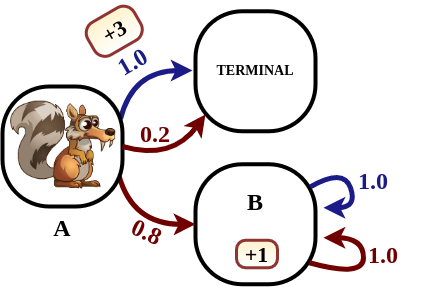}
\vspace{-1cm}
\end{wrapfigure}

По определению оптимальная V-функция для состояния A равна 
$$V^*(s = A) = \max\limits_{\theta \in [0, 1]} J(\pi) = \max\limits_{\theta \in [0, 1]} \left[ 3 + 5\theta \right] = 8.$$

Оценочные функции для состояния B для всех стратегий совпадают и равны $V^*(s \HM= B) \HM= 1 \HM+ \gamma \HM+ \gamma^2 \HM+ \dots \HM= 11$. Для терминальных состояний $V^*(s) \HM= 0$.
\end{example}

\subsection{Q-функция}

V-функции нам не хватит. Если бы мы знали оптимальную value-функцию $V^*(s)$, мы не смогли бы восстановить хоть какую-то оптимальную политику из-за отсутствия в общем случае информации о динамике среды. Допустим, агент находится в некотором состоянии и знает его ценность $V^*(s)$, а также знает ценности всех других состояний; это не даёт понимания того, какие действия в какие состояния приведут --- мы никак не дифференцируем действия между собой. Поэтому мы увеличим количество переменных: введём схожее определение для ценности не состояний, но пар состояние-действие. 

\begin{definition} 
Для данного MDP \emph{Q-функцией} (state-action value function, action quality function) для данной стратегии $\pi$ называется
$$Q^\pi(s, a) \coloneqq \E_{\Traj \sim \pi \mid s_0 = s, a_0 = a} \sum_{t \ge 0} \gamma^t r_t$$
\end{definition}

\begin{theorem}[Связь оценочных функций]
V-функции и Q-функции взаимозависимы, а именно:
\begin{equation}\label{QV}
    Q^\pi(s, a) =  r(s, a) + \gamma \E_{s'} V^\pi(s')
\end{equation}
\begin{equation}\label{VQ}
    V^\pi(s) = \E_{a \sim \pi(a \mid s)} Q^\pi(s, a)
\end{equation}
\begin{proof}
Следует напрямую из определений.
\end{proof}
\end{theorem}

Итак, если V-функция --- это сколько получит агент из некоторого состояния, то Q-функция --- это сколько получит агент после выполнения данного действия из данного состояния. Как и V-функция, Q-функция не зависит от времени, ограничена по модулю при рассматриваемых требованиях к MDP, и, аналогично, для неё существует уравнение Беллмана:

\begin{theorem}[Уравнение Беллмана (Bellman expectation equation) для Q-функции]
\,
\begin{equation}\label{QQ}
    Q^\pi(s, a) =  r(s, a) + \gamma \E_{s'} \E_{a'} Q^\pi(s', a')
\end{equation}
\begin{proof}
Можно воспользоваться \eqref{QV} + \eqref{VQ}, можно расписать как награду на следующем шаге плюс хвостик.
\end{proof}
\end{theorem}

\begin{example}
Q-функция получает на вход пару состояние-действие и ничего не говорится о том, что это действие должно быть как-то связано с оцениваемой стратегией $\pi$. 

\needspace{10\baselineskip}
\begin{wrapfigure}{r}{0.35\textwidth}
\vspace{-0.2cm}
\centering
\includegraphics[width=0.35\textwidth]{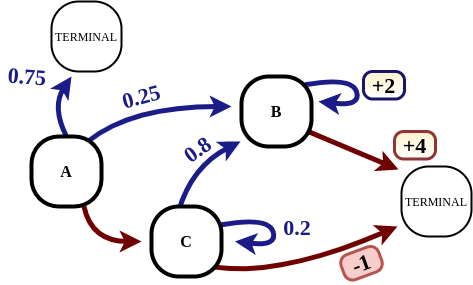}
\vspace{-1.3cm}
\end{wrapfigure}

Давайте в MDP с рисунка рассмотрим стратегию $\pi$, которая всегда детерминировано выбирает действие $\colorsquare{ChadRed}$. Мы тем не менее можем посчитать $Q^\pi(s, \colorsquare{ChadBlue})$ для любых состояний (например, для терминальных это значение формально равно нулю). Сделаем это при помощи QV уравнения:
\begin{align*} 
Q^\pi(s = A, \colorsquare{ChadBlue}) &= 0.25 \gamma V^\pi(s = B) \\ 
Q^\pi(s = B, \colorsquare{ChadBlue}) &= 2 + \gamma V^\pi(s = B) \\
Q^\pi(s = C, \colorsquare{ChadBlue}) &= 0.8 \gamma V^\pi(s = B) + 0.2 \gamma V^\pi(s = C)
\end{align*}
Внутри $V^\pi$ сидит дальнейшее поведение при помощи стратегии $\pi$, то есть выбор исключительно действий $\colorsquare{ChadRed}$: соответственно, $V^\pi(s = B) = 4$, $V^\pi(s = C) = -1$.
\end{example}

Мы получили все уравнения Беллмана для оценочных функций (с условными названиями VV, VQ, QV и, конечно же, QQ). Как видно, они следуют напрямую из определений; теперь посмотрим, что можно сказать об оценочных функциях оптимальных стратегий.

\subsection{Принцип оптимальности Беллмана}

\begin{definition}
Для данного MDP \emph{оптимальной Q-функцией} (optimal Q-function) называется
\begin{equation}\label{optimalQdefinition}
Q^*(s, a) \coloneqq \max_\pi Q^\pi(s, a)
\end{equation}
\end{definition}

Формально очень хочется сказать, что $Q^*$ --- оценочная функция для оптимальных стратегий, но мы пока никак не связали введённую величину с $V^*$ и показать это пока не можем. Нам доступно только такое неравенство пока что:
\begin{proposition}\,
$$Q^*(s, a) \le r + \gamma \E_{s'} V^*(s')$$
\beginproof
\begin{align*}
Q^*(s, a) =
\{ \text{определение $Q^*$ \eqref{optimalQdefinition} } \} &= \max_{\pi} Q^{\pi}(s, a) = \\
= \{ \text{связь QV \eqref{QV}}\} &= \max_{\pi} \left[ r + \gamma \E_{s'} V^{\pi}(s') \right] \le \\
\le \{ \text{ максимум среднего не превосходит среднее максимума } \} &\le r + \gamma \E_{s'} \max_{\pi} V^{\pi}(s') = \\ 
= \{ \text{ определение $V^*$ \eqref{optimalVdefinition} } \} &= r + \gamma \E_{s'} V^*(s') \tagqed
\end{align*}
\end{proposition}

Равенство в месте с неравенством случилось бы, если бы мы доказали следующий факт: что вообще существует такая стратегия $\pi$, которая максимизирует $V^\pi$ сразу для всех состояний $s$ одновременно (и которую мы определили как оптимальную). Другими словами, нужно показать, что максимизация $V^\pi(s)$ для одного состояния <<помогает>> максимизировать награду для других состояний. Для V-функции мы можем построить аналогичную оценку сверху:

\begin{proposition}\,
$$V^*(s) \le \max_{a} Q^*(s, a)$$
\beginproof
\begin{align*}
V^*(s) =
\{ \text{определение $V^*$ \eqref{optimalVdefinition} } \} &= \max_{\pi} V^{\pi}(s) = \\
= \{ \text{связь VQ \eqref{VQ}}\} &= \max_{\pi} \E_{a \sim \pi(a \mid s)} Q^{\pi}(s, a) \le \\
\le \{ \text{ по определению $Q^*$ \eqref{optimalQdefinition} } \} &\le \max_{\pi} \E_{a \sim \pi(a \mid s)} Q^*(s, a) \le \\ 
\le \{ \text{ свойство $\E_x f(x) \le \max\limits_x f(x)$ } \} &\le \max_{a} Q^*(s, a)   \tagqed
\end{align*}
\end{proposition}

\needspace{8\baselineskip}
\begin{wrapfigure}{r}{0.25\textwidth}
\vspace{-0.3cm}
\centering
\includegraphics[width=0.25\textwidth]{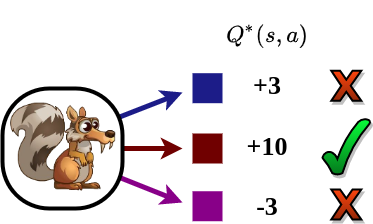}
\vspace{-0.5cm}
\end{wrapfigure}

Можно ли получить $\max\limits_{a} Q^*(s, a)$, то есть достигнуть этой верхней оценки? Проведём нестрогое следующее рассуждение: представим, что мы сидим в состоянии $s$ и знаем величины $Q^*(s, a)$, определённые как \eqref{optimalQdefinition}. Это значит, что если мы сейчас выберем действие $a$, то в дальнейшем сможем при помощи какой-то стратегии $\pi$, на которой достигается максимум\footnote{мы знаем, что Q-функция ограничена, и поэтому точно существует супремум. Для полной корректности рассуждений надо говорить об $\eps$-оптимальности, но для простоты мы это опустим.} для конкретно данной пары $s, a$, получить $Q^*(s, a)$. Следовательно, мы, выбрав сейчас то действие $a$, на которых достигается максимум, в предположении <<дальнейшей оптимальности своего поведения>>, надеемся получить из текущего состояния $\max\limits_a Q^*(s, a)$.

\begin{definition}\label{greedy}
Для данного приближения Q-функции стратегия $\pi(s) \coloneqq \argmax\limits_a Q(s, a)$ называется \emph{жадной} (greedy with respect to Q-function).
\end{definition}

\begin{definition}
\emph{Принцип оптимальности Беллмана}: жадный выбор действия в предположении оптимальности дальнейшего поведения оптимален.
\end{definition}

Догадку несложно доказать для случая, когда MDP является деревом: принятие решения в текущем состоянии $s$ никак не связано с выбором действий в <<поддеревьях>>. Если в поддереве, соответствующему одному действию, можно получить больше, чем в другом поддереве, то понятно, что выбирать нужно его. В общем случае, однако, нужно показать, что жадный выбор в $s$ <<позволит>> в будущем набрать то $Q^*(s, a)$, которое мы выбрали --- вдруг для того, чтобы получить в будущем $Q^*(s, a)$, нужно будет при попадании в то же состояние $s$ выбирать действие как-то по-другому? Если бы это было так, было бы оптимально искать стратегию в классе нестационарных стратегий.

\subsection{Отказ от однородности}

Утверждение, позволяющее, во-первых, получить вид оптимальной стратегии, а как следствие связать оптимальные оценочные функции, будет доказано двумя способами. В этой секции докажем через отказ от однородности (<<классическим>> способом), а затем в секции \ref{PIsection} про Policy Improvement мы поймём, что все желаемые утверждения можно получить и через него.

Отказ от однородности заключается в том, что мы в доказательстве будем искать максимум $\max\limits_\pi V^\pi(s)$ не только среди стационарных, но и нестационарных стратегий. Заодно мы убедимся, что достаточно искать стратегию в классе стационарных стратегий. Ранее стационарность означала, что вне зависимости от момента времени наша стратегия зависит только от текущего состояния. Теперь же, для каждого момента времени $t = 0, 1 \dots$ мы запасёмся своей собственной стратегией $\pi_t(a \mid s)$. Естественно, что теорема \ref{Vtimeindepend} о независимости оценочной функции от времени тут перестаёт быть истинной, и, вообще говоря, оценочные функции теперь зависимы от текущего момента времени $t$. 

\begin{definition}
Для данного MDP и нестационарной стратегии $\pi = \{ \pi_t(a \mid s) \mid t \ge 0 \}$ обозначим её \emph{оценочные функции} как 
$$V^\pi_t(s) \coloneqq \E_{\pi_t(a_t \mid s_t = s)}\E_{p(s_{t+1} \mid s_t=s, a_t)}\E_{\pi_{t+1}(a_{t+1} \mid s_{t+1})} \dots R_t$$
$$Q^\pi_t(s, a) \coloneqq \E_{p(s_{t+1} \mid s_t=s, a_t=a)}\E_{\pi_{t+1}(a_{t+1} \mid s_{t+1})} \dots R_t$$
\end{definition}

\begin{example}
Действительно, мы можем в состоянии $s$ смотреть на часы, если $t=0$ --- кушать тортики, а если $t=7$ --- бросаться в лаву, т.е. $V^{\pi}_{t=0}(s) \ne V^{\pi}_{t=7}(s)$ для неоднородных $\pi$.
\end{example}

\begin{proposition}
Для нестационарных оценочных функций остаются справедливыми уравнения Беллмана:
\begin{equation}\label{VQ_nonstat}
V^\pi_t(s) = \E_{\pi_t(a \mid s)} Q^\pi_t(s, a)
\end{equation}
\begin{equation}\label{QV_nonstat}
Q^\pi_t(s, a) = r(s, a) + \gamma \E_{s'} V^\pi_{t+1}(s')
\end{equation}
\begin{proof}
Всё ещё следует из определений.
\end{proof}
\end{proposition}

\begin{definition}
Для данного MDP \emph{оптимальными оценочными функциями среди нестационарных стратегий} назовём
\begin{equation}\label{V*_nonstat}
V^*_t(s) \coloneqq \max_{\substack{\pi_t \\ \pi_{t+1} \\ \cdots}} V^\pi_{t}(s)
\end{equation}
\begin{equation}\label{Q*_nonstat}
Q^*_t(s, a) \coloneqq \max_{\substack{\pi_{t+1} \\ \pi_{t+2} \\ \cdots}} Q^\pi_{t}(s, a)
\end{equation}
\end{definition}

Заметим, что в определении Q-функции максимум берётся по стратегиям, начиная с $\pi_{t+1}$, поскольку по определению Q-функция не зависит от $\pi_t$ (действие в момент времени $t$ уже <<дано>> в качестве входа).

\begin{proposition}\label{pr:nonstat_optimal_are_stat}
В стационарных MDP (а мы рассматриваем только их) оптимальные оценочные функции не зависят от времени, т.е. $\forall s, a, t_1, t_2$ верно:
$$V^*_{t_1}(s) = V^*_{t_2}(s) \qquad Q^*_{t_1}(s, a) = Q^*_{t_2}(s, a)$$
\beginproof Вообще говоря, по построению, так как зависимость от времени заложена исключительно в стратегиях, по которым мы берём максимум (а его мы берём по одним и тем же симплексам вне зависимости от времени):
\begin{equation*}
V^*_t(s) = \max_{\substack{\pi_t \\ \pi_{t+1} \\ \cdots}} \E_{a_t, s_{t+1} \dots \mid s_t = s} R_t = \max_{\substack{\pi_0 \\ \pi_{1} \\ \cdots}} \E_{a_0, s_{1} \dots \mid s_0 = s} R_0 \tagqed
\end{equation*}
\end{proposition}

Последнее наблюдение само по себе нам ничего не даёт. Вдруг нам в условном MDP с одним состоянием выгодно по очереди выбирать каждое из трёх действий?

\subsection{Вид оптимальной стратегии (доказательство через отказ от однородности)}

Мотивация в отказе от однородности заключается в том, что наше MDP теперь стало деревом: эквивалентно было бы сказать, что мы добавили в описание состояний время $t$. Теперь мы не оказываемся в одном состоянии несколько раз за эпизод; максимизация $Q^*_t(s, a)$ требует оптимальных выборов <<в поддереве>>, то есть настройки $\pi_{t+1}, \pi_{t+2}$ и так далее, а для $\pi_t(a \mid s)$ будет выгодно выбрать действие жадно. Покажем это формально.

\begin{theoremBox}[label=th:nonstatbellmancriterion]{}
Cтратегия $\pi_t(s) \coloneqq \argmax\limits_{a} Q^*_t(s, a)$ оптимальна, то есть для всех состояний $s$ верно $V^\pi_t(s) \HM= V^*_t(s)$, и при этом справедливо:
\begin{equation}\label{V*Q*_nonstat}
V^*_t(s) = \max_{a} Q^*_t(s, a)
\end{equation}
\begin{proof}
В силу VQ уравнения \eqref{VQ_nonstat}, максимизация $V^{\pi}_t(s)$ эквивалентна максимизации
$$V^*_t(s) = \max_{\substack{\pi_t \\ \pi_{t+1} \\ \cdots}} V^\pi_t(s) = \max_{\substack{\pi_t \\ \pi_{t+1} \\ \cdots}} \E_{\pi_t(a \mid s)} Q^\pi_t(s, a)$$

Мы уже замечали, что $Q^\pi_t$ по определению зависит только от $\pi_{t+1}(a \mid s), \pi_{t+2}(a \mid s) \dots$. Максимум $Q^\pi_t(s, a)$ по ним по определению \eqref{Q*_nonstat} есть $Q^*_t(s, a)$. Значит,
$$V^*_t(s) \le \max_{\pi_t} \E_{\pi_t(a \mid s)} \max_{\substack{\pi_{t+1} \\ \pi_{t+2} \\ \cdots}} Q^\pi_t(s, a) = \max_{\pi_t} \E_{\pi_t(a \mid s)} Q^*_t(s, a)$$

Покажем, что эта верхняя оценка достигается. Сначала найдём $\pi_t$ такую, что:
$$
\begin{cases}
\E_{\pi_t(a \mid s)} Q^*_t(s, a) \to \max\limits_{\pi_t} \\
\int\limits_\A \pi_t(a \mid s) \diff a = 1; \qquad \forall a \in \A \colon \pi_t(a \mid s) \ge 0
\end{cases}
$$

Решением такой задачи, в частности\footnote[*]{здесь записана просто задача линейного программирования на симплексе ($\pi_t$ обязано быть распределением); общим решением задачи, соответственно, будет любое распределение, которое размазывает вероятности между элементами множества $\Argmax\limits_{a} Q^*_t(s, a)$.}, будет детерминированная стратегия
$$\pi_t^*(s) \coloneqq \argmax_{a} Q^*_t(s, a)$$
а сам максимум, соответственно, будет равняться $\max\limits_{a} Q^*_t(s, a)$. Соответственно, $\max\limits_{\substack{\pi_t \\ \pi_{t+1} \\ \cdots}} V^\pi_t(s)$ достигает верхней оценки при этой $\pi^*_t$ и том наборе $\pi^*_{t+1}, \pi^*_{t+2} \dots$, на котором достигается значение $Q^*_t(s, \pi_t^*(s) )$.
\end{proof}
\end{theoremBox}

\begin{proposition}
Для нестационарных оценочных функций верно:
\begin{equation}\label{Q*V*_nonstat}
Q^*_t(s, a) = r(s, a) + \gamma \E_{s'} V^*_{t+1}(s')
\end{equation}
\begin{proof}
Получим аналогично оценку сверху на $Q^*_t(s, a)$:
\begin{align*}Q^*_t(s, a) = \max_{\substack{\pi_{t+1} \\ \pi_{t+2} \\ \cdots}} Q^\pi_t(s, a) = \{ \text{связь QV \eqref{QV_nonstat}}\} &= \max_{\substack{\pi_{t+1} \\ \pi_{t+2} \\ \cdots}} \left[ r(s, a) + \gamma \E_{s'} V^\pi_{t+1}(s') \right] \le \\ \le \{ \text{определение $V^*$ \eqref{V*_nonstat} } \} &\le r(s, a) + \gamma \E_{s'} V^*_{t+1}(s')
\end{align*}

Эта верхняя оценка достигается на стратегии $\pi_t(s) = \argmax\limits_a Q^*_t(s, a)$, на которой, как мы доказали в предыдущей теореме \ref{th:nonstatbellmancriterion}, достигается максимум $V^\pi_t(s) = V^*(s)$ сразу для всех $s$ одновременно.
\end{proof}
\end{proposition}

Таким образом мы показали, что в нестационарном случае наши $Q^*$ и $V^*$ являются оценочными функциями оптимальных стратегий, максимизирующих награду из всех состояний. Осталось вернуться к стационарному случаю, то есть показать, что для стационарных стратегий выполняется то же утверждение.

\begin{proposition}
Оптимальные оценочные функции для стационарных и нестационарных случаев совпадают, то есть, например, для V-функции:
$$\max_\pi V^\pi (s) = \max_{\substack{\pi_t \\ \pi_{t+1} \\ \cdots}} V^\pi_t(s),$$
где в левой части максимум берётся по стационарным стратегиям, а в правой --- по нестационарным.
\begin{proof}
По теореме \ref{th:nonstatbellmancriterion} максимум справа достигается на детерминированной $\pi_t(s) \HM= \argmax\limits_{a} Q^*_t(a, s)$. В силу утверждения \ref{pr:nonstat_optimal_are_stat}, для всех моментов времени $Q^*_t$ совпадают, следовательно такая $\pi_t$ тоже совпадает для всех моментов времени и является стационарной стратегией.
\end{proof}
\end{proposition}

Интуитивно: мы показали, что об MDP <<с циклами в графе>> можно думать как о дереве. Итак, в полученных результатах можно смело заменять все нестационарные оптимальные оценочные функции на стационарные. 

\subsection{Уравнения оптимальности Беллмана}

\begin{theorem}[Связь оптимальных оценочных функций]
\begin{equation}\label{V*Q*}
V^*(s) = \max_a Q^*(s, a)
\end{equation}
\begin{equation}\label{Q*V*}
    Q^*(s, a) = r(s, a) + \gamma \E_{s'} V^*(s')
\end{equation}
\end{theorem}

Теперь $V^*$ выражено через $Q^*$ и наоборот. Значит, можно получить выражение для $V^*$ через $V^*$ и $Q^*$ через $Q^*$:

\begin{theorem}[\emph{Уравнения оптимальности Беллмана} (Bellman optimality equation)]
\,
\begin{equation}\label{Q*Q*}
    Q^*(s, a) =  r(s, a) + \gamma \E_{s'} \max_{a'} Q^*(s', a')
\end{equation}
\begin{equation}\label{V*V*}
    V^*(s) =  \max_a \left[ r(s, a) + \gamma \E_{s'} V^*(s') \right]
\end{equation}
\begin{proof}
Подставили \eqref{V*Q*} в \eqref{Q*V*} и наоборот.
\end{proof}
\end{theorem}

Хотя для строгого доказательства нам и пришлось поднапрячься и выписать относительно громоздкое рассуждение, уравнения оптимальности Беллмана крайне интуитивны. Для $Q^*$, например, можно рассудить так: что даст оптимальное поведение из состояния $s$ после совершения действия $a$? Что с нами случится дальше: мы получим награду за этот выбор $r(s, a)$, на что уже повлиять не можем. Остальная награда будет дисконтирована. Затем среда переведёт нас в какое-то следующее состояние $s'$ --- нужно проматожидать по функции переходов. После этого мы, пользуяся принципом Беллмана, просто выберем то действие, которое позволит в будущем набрать наибольшую награду, и тогда сможем получить $\max\limits_{a'} Q^*(s', a')$. 

\begin{exampleBox}[label=ex:optimalvfunction]{}
Сформулируем для MDP с рисунка уравнения оптимальности Беллмана для $V^*$. Мы получим систему из трёх уравнений с трёмя неизвестными.

\begin{wrapfigure}{r}{0.35\textwidth}
\vspace{-0.2cm}
\centering
\includegraphics[width=0.35\textwidth]{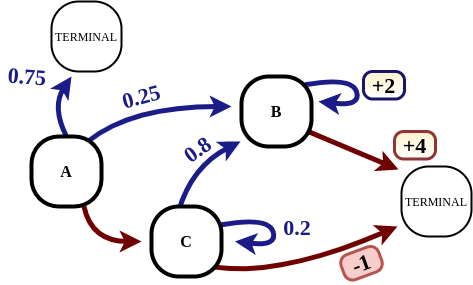}
\vspace{-1.3cm}
\end{wrapfigure}

\begin{align*} 
V^*(s = A) &= \max \bigl( \underbrace{0.25 \gamma V^*(s = B)}_{\colorsquare{ChadBlue}}, \underbrace{\gamma V^*(s = C)}_{\colorsquare{ChadRed}} \bigr) \\ 
V^*(s = B) &= \max \bigl( \underbrace{2 + \gamma V^*(s = B)}_{\colorsquare{ChadBlue}}, \underbrace{4}_{\colorsquare{ChadRed}} \bigr) \\ 
V^*(s = C) &= \max \bigl( \underbrace{0.8 \gamma V^*(s = B) + 0.2 \gamma V^*(s = C)}_{\colorsquare{ChadBlue}}, \underbrace{-1}_{\colorsquare{ChadRed}} \bigr) \\ 
\end{align*}
\end{exampleBox}

Заметим, что в полученных уравнениях не присутствует мат.ожиданий по самим оптимальным стратегиям --- предположение дальнейшей оптимальности поведения по сути <<заменяет>> их на взятие максимума по действиям. Более того, мы позже покажем, что оптимальные оценочные функции --- единственные решения систем уравнений Беллмана. А значит, вместо поиска оптимальной стратегии можно искать оптимальные оценочные функции! Таким образом, мы свели задачу оптимизации нашего функционала к решению системы нелинейных уравнений особого вида. Беллман назвал данный подход <<\emph{динамическое программирование}>> (dynamic programming).






\subsection{Критерий оптимальности Беллмана}

Давайте сформулируем критерий оптимальности стратегий в общей форме, описывающей вид всего множества оптимальных стратегий. Для доказательства нам понадобится факт, который мы технически докажем в рамках повествования чуть позже: для данного MDP $Q^*$ --- единственная функция $\St \times \A \to \R$, удовлетворяющая уравнениям оптимальности Беллмана.

\begin{theoremBox}[label=th:optimalitycriterion]{Критерий оптимальности Беллмана}
$\pi$ оптимальна тогда и только тогда, когда $ \forall s, a \colon \pi(a \mid s) > 0$ верно:
$$a \in \Argmax_a Q^\pi(s, a)$$

\begin{proof}[Необходимость] Пусть $\pi$ --- оптимальна. Тогда её оценочные функции совпадают с $V^*, Q^*$, для которых выполнено уравнение \eqref{V*Q*}:
$$V^{\pi}(s) = V^*(s) = \max_{a} Q^*(s, a) = \max_{a} Q^{\pi}(s, a)$$
С другой стороны из связи VQ \eqref{VQ} верно $V^{\pi}(s) = \E_{\pi(a \mid s)} Q^{\pi}(s, a)$; получаем
$$\E_{\pi(a \mid s)} Q^{\pi}(s, a) = \max_{a} Q^{\pi}(s, a),$$
из чего вытекает доказываемое.
\end{proof}
\begin{proof}[Достаточность] Пусть условие выполнено. Тогда для любой пары $s, a$:
$$Q^{\pi}(s, a) = \{ \text{связь QQ \eqref{QQ}} \} = r(s, a) + \gamma \E_{s'} \E_{\pi(a' \mid s')} Q^{\pi}(s', a') = r(s, a) + \gamma \E_{s'} \max_{a'} Q^{\pi}(s', a')$$
Из единственности решения этого уравнения следует $Q^{\pi}(s, a) \HM= Q^*(s, a)$, и, следовательно, $\pi$ оптимальна.
\end{proof}
\end{theoremBox}

Иначе говоря: теорема говорит, что оптимальны ровно те стратегии, которые пользуются принципом оптимальности Беллмана. Если в одном состоянии два действия позволят в будущем набрать максимальную награду, то между ними можно любым способом размазать вероятности выбора. Давайте при помощи этого критерия окончательно ответим на вопросы о том, существует ли оптимальная стратегия и сколько их вообще может быть.

\begin{proposition}
Если $|\A| < +\infty$, всегда существует оптимальная стратегия.
\begin{proof}
$\Argmax\limits_a Q^*(s, a)$ для конечных множеств $\A$ всегда непуст, следовательно существует детерминированная оптимальная стратегия $\pi(s) \coloneqq \argmax\limits_a Q^*(s, a)$.
\end{proof}
\end{proposition}

\begin{proposition}
Оптимальной стратегии может не существовать.
\begin{proof}[Контрпример]Одно состояние, $\A = [-1, 1]$, после первого выбора эпизод заканчивается; в качестве награды $r(a)$ можем рассмотреть любую не достигающую своего максимума функцию. Просто придумали ситуацию, когда $\Argmax\limits_{a} Q^*(a)$ пуст.
\end{proof}
\end{proposition}

\begin{proposition}
Если существует хотя бы две различные оптимальные стратегии, то существует континуум оптимальных стратегий.
\begin{proof}
Существование двух различных оптимальных стратегий означает, что в каком-то состоянии $s$ множество $\Argmax\limits_a Q^*(s, a)$ содержит по крайней мере два элемента. Между ними можно размазать вероятности выбора любым способом и в любом случае получить максимальную награду.
\end{proof}
\end{proposition}

\begin{proposition}\label{pr:deterministicoptimal}
Если существует хотя бы одна оптимальная стратегия, то существует детерминированная оптимальная стратегия.
\begin{proof}
Пусть $\pi^*$ --- оптимальна. Значит, $\Argmax\limits_a Q^*(s, a)$ не пуст для всех $s$, и существует детерминированная оптимальная стратегия $\pi(s) \coloneqq \argmax\limits_a Q^*(s, a)$.
\end{proof}
\end{proposition}

\begin{example}
Найдём все оптимальные стратегии в MDP из примера \ref{ex:optimalvfunction} для $\gamma \HM= 0.5$.

\begin{wrapfigure}{r}{0.35\textwidth}
\vspace{-0.2cm}
\centering
\includegraphics[width=0.35\textwidth]{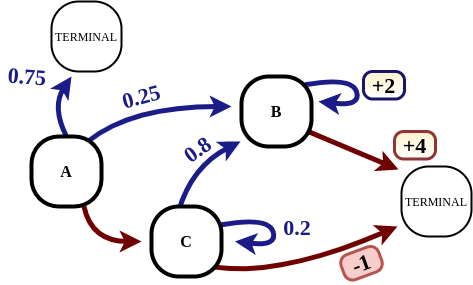}
\vspace{-1cm}
\end{wrapfigure}

Мы могли бы составить уравнения оптимальности Беллмана для $Q^*$ и решать их, но сделаем чуть умнее и воспользуемся критерием оптимальности Беллмана (теорема \ref{th:optimalitycriterion}). Например, в состоянии B оптимально или выбирать какое-то одно из двух действий с вероятностью 1, или действия эквивалентны, и тогда оптимально любое поведение. Допустим, мы будем выбирать всегда \colorsquare{ChadBlue}, тогда мы получим $\frac{2}{1 - \gamma} \HM= 4$; если же будем выбирать \colorsquare{ChadRed}, то получим +4. Значит, действия эквивалентны, оптимально любое поведение, и $V^*(s \HM= B) \HM= 4$.

Проведём аналогичное рассуждение для состояния C. Если оптимально действие \colorsquare{ChadBlue}, то $$Q^*(s \HM= C, \text{\colorsquare{ChadBlue}}) = 0.2 \gamma Q^*(s \HM= C, \text{\colorsquare{ChadBlue}}) + 0.8 \gamma V^*(s = B)$$
Решая это уравнение относительно неизвестного $Q^*(s \HM= C, \text{\colorsquare{ChadBlue}})$, получаем $\frac{16}{9} \HM> Q^*(s \HM= C, \text{\colorsquare{ChadRed}}) \HM= -1$. Значит, в C оптимальная стратегия обязана выбирать \colorsquare{ChadBlue}, и $V^*(C) \HM= \frac{16}{9}.$

Для состояния A достаточно сравнить $Q^*(s \HM= A, \text{\colorsquare{ChadBlue}}) \HM= 0.25 \gamma V^*(s \HM= B) \HM= \frac{1}{4}$ и $Q^*(s \HM= A, \text{\colorsquare{ChadRed}}) \HM= \gamma V^*(s \HM= C) \HM= \frac{8}{9}$, определив, что оптимальная стратегия должна выбирать \colorsquare{ChadRed}.
\end{example}
\section{Улучшение политики}\label{PIsection} 

\subsection{Advantage-функция}\label{subsection:advantage}

Допустим, мы находились в некотором состоянии $s$, и засэмплировали $a \HM\sim \pi(a \HM\mid s)$ такое, что $Q^\pi(s, a) \HM> V^\pi(s)$. Что можно сказать о таком действии? Мы знаем, что вообще в среднем политика $\pi$ набирает из данного состояния $V^\pi(s)$, но какой-то выбор действий даст в итоге награду больше $V^\pi(s)$, а какой-то меньше. Если $Q^\pi(s, a) \HM> V^\pi(s)$, то после того, как мы выбрали действие $a$, <<приняли решение>>, наша средняя будущая награда вдруг увеличилась.

Мы ранее обсуждали в разделе \ref{RLproblems} такую особую проблему обучения с подкреплением, как credit assingment, которая звучит примерно так: допустим, мы засэмплировали траекторию $s, a, s', a', \dots $ до конца эпизода, и в конце в финальном состоянии через $T$ шагов получили сигнал (награду) +1. Мы приняли $T$ решений, но какое из всех этих действий повлекло получение этого +1? <<За что нас наградили?>> Повлияло ли на получение +1 именно то действие $a$, которое мы засэмплировали в стартовом $s$? Вопрос нетривиальный, потому что в RL есть отложенный сигнал: возможно, именно действие $a$ в состоянии $s$ запустило какую-нибудь цепочку действий, которая дальше при любом выборе $a', a'', \cdots$ приводит к награде +1. Возможно, конечно, что первое действие и не имело никакого отношения к этой награде, и это поощрение именно за последний выбор. А ещё может быть такое, что имело место везение, и просто среда в какой-то момент перекинула нас в удачное состояние.

Но мы понимаем, что если какое-то действие <<затриггерило>> получение награды через сто шагов, в промежуточных состояниях будет информация о том, сколько времени осталось до получения этой отложенной награды. Например, если мы выстрелили во вражеский инопланетный корабль, и через 100 шагов выстрел попадает во врага, давая агенту +1, мы будем видеть в состояниях расстояние от летящего выстрела до цели, и знать, что через такое-то время нас ждёт +1. Другими словами, вся необходимая информация лежит в идеальных оценочных функциях $Q^\pi$ и $V^\pi$.

Так, если в некотором состоянии $s$ засэмплировалось такое $a$, что $Q^\pi(s, a) \HM= V^\pi(s)$, то мы можем заключить, что выбор действия на этом шаге не привёл ни к какой <<неожиданной>> награде. Если же $Q^\pi(s, a) \HM> V^\pi(s)$ --- то мы приняли удачное решение, $Q^\pi(s, a) \HM< V^\pi(s)$ --- менее удачное, чем обычно. Если, например, $r(s, a) \HM+ V^\pi(s') \HM> Q^\pi(s, a)$, то мы можем заключить, что имело место везение: среда засэмплировала такое $s'$, что теперь мы получим больше награды, чем ожидали после выбора $a$ в состоянии $s$. И так далее: мы сможем отследить, в какой конкретно момент случилось то событие (сэмплирование действия или ответ среды), за счёт которого получена награда.

Таким образом, идеальный <<кредит>> влияния действия $a$, выбранного в состоянии $s$, на будущую награду равен
$$Q^\pi(s, a) - V^\pi(s),$$
и именно эта величина на самом деле будет для нас ключевой. Поэтому из соображений удобства вводится ещё одно обозначение:

\begin{definition} 
Для данного MDP \emph{Advantage-функцией} политики $\pi$ называется
\begin{equation}\label{advantage}
A^\pi(s, a) \coloneqq Q^\pi(s, a) - V^\pi(s)
\end{equation}
\end{definition}

\begin{proposition}\label{pr:advantageiszero}
Для любой политики $\pi$ и любого состояния $s$:
$$\E_{\pi(a \mid s)} A^\pi(s, a) = 0$$
\beginproof
\begin{align*}
\E_{\pi(a \mid s)} A^\pi(s, a) &= \E_{\pi(a \mid s)} Q^\pi(s, a) - \E_{\pi(a \mid s)} V^\pi(s) = \\
\{ \text{$V^\pi$ не зависит от $a$} \} &= \E_{\pi(a \mid s)} Q^\pi(s, a) - V^\pi(s) = \\
\{ \text{связь $V$ через $Q$ \eqref{VQ}} \} &= V^\pi(s) - V^\pi(s) = 0   \tagqed
\end{align*}
\end{proposition}

\begin{proposition}
\label{adv_is_positive}
Для любой политики $\pi$ и любого состояния $s$:
$$\max_a A^\pi(s, a) \ge 0$$
\end{proposition}

Advantage --- это, если угодно, <<центрированная>> Q-функция. Если $A^\pi(s, a) \HM> 0$ --- действие $a$ <<лучше среднего>> для нашей текущей политики в состоянии $s$, меньше нуля --- хуже.  И интуиция, что процесс обучения нужно строить на той простой идеи, что первые действия надо выбирать чаще, а вторые --- реже, нас не обманывает. 

Естественно, подвох в том, что на практике мы не будем знать точное значение оценочных функций, а значит, и истинное значение Advantage. Решая вопрос оценки значения Advantage для данной пары $s, a$, мы фактически будем проводить credit assingment --- это одна и та же задача.

\subsection{Relative Performance Identity (RPI)}

Мы сейчас докажем одну очень интересную лемму, которая не так часто нам будет нужна в будущем, но которая прям открывает глаза на мир. Для этого вспомним формулу reward shaping-а \eqref{rewardshaping} и заметим, что мы можем выбрать в качестве потенциала V-функцию произвольной стратегии $\pi_2$:
$$\Phi(s) \coloneqq V^{\pi_2}(s)$$
Действительно, требований к потенциалу два: ограниченность (для V-функций это выполняется в силу наших ограничений на рассматриваемые MDP) и равенство нулю в терминальных состояниях (для V-функций это верно по определению). Подставив такой потенциал, мы получим связь между performance-ом $J(\pi) = V^\pi(s_0)$ двух разных стратегий. В общем виде лемма сравнивает V-функции двух стратегий в одном состоянии:

\begin{theoremBox}[label=th:rpi]{Relative Performance Identity}
Для любых двух политик $\textcolor{ChadBlue}{\pi_1}$, $\textcolor{ChadPurple}{\pi_2}$:
\begin{equation}\label{RPI}
\textcolor{ChadPurple}{V^{\pi_2}}(s) - \textcolor{ChadBlue}{V^{\pi_1}}(s) = \E_{\textcolor{ChadPurple}{\Traj \sim \pi_2} \mid s_0 = s} \sum_{t \ge 0} \gamma^t \textcolor{ChadBlue}{A^{\pi_1}}(s_t, a_t)
\end{equation}
\beginproof
\begin{align*}
\textcolor{ChadPurple}{V^{\pi_2}}(s) - \textcolor{ChadBlue}{V^{\pi_1}}(s) &= \E_{\textcolor{ChadPurple}{\Traj \sim \pi_2} \mid s_0 = s} \sum_{t \ge 0} \gamma^t r_t - \textcolor{ChadBlue}{V^{\pi_1}}(s) = \\
&= \E_{\textcolor{ChadPurple}{\Traj \sim \pi_2} \mid s_0 = s} \left[\sum_{t \ge 0} \gamma^t r_t - \textcolor{ChadBlue}{V^{\pi_1}}(s_0) \right] = \\
\{ \text{телескопирующая сумма \eqref{telescopingsum}} \} &= \E_{\textcolor{ChadPurple}{\Traj \sim \pi_2} \mid s_0 = s} \left[\sum_{t \ge 0} \gamma^t r_t + \sum_{t \ge 0} \left[ \gamma^{t+1} \textcolor{ChadBlue}{V^{\pi_1}}(s_{t+1}) - \gamma^t \textcolor{ChadBlue}{V^{\pi_1}}(s_t) \right] \right] = \\
\{ \text{перегруппируем слагаемые} \} &= \E_{\textcolor{ChadPurple}{\Traj \sim \pi_2} \mid s_0 = s} \sum_{t \ge 0} \gamma^t \left( r_t + \gamma \textcolor{ChadBlue}{V^{\pi_1}}(s_{t+1}) - \textcolor{ChadBlue}{V^{\pi_1}}(s_t) \right) = \\
\{ \text{фокус $\E_x f(x) = \E_x \E_x f(x)$} \} &= \E_{\textcolor{ChadPurple}{\Traj \sim \pi_2} \mid s_0 = s} \sum_{t \ge 0} \gamma^t \left( r_t + \gamma \E_{s_{t+1}} \textcolor{ChadBlue}{V^{\pi_1}}(s_{t+1}) - \textcolor{ChadBlue}{V^{\pi_1}}(s_t) \right) = \\
\{ \text{выделяем Q-функцию \eqref{QV}} \} &= \E_{\textcolor{ChadPurple}{\Traj \sim \pi_2} \mid s_0 = s} \sum_{t \ge 0} \gamma^t \left( \textcolor{ChadBlue}{Q^{\pi_1}}(s_t, a_t) - \textcolor{ChadBlue}{V^{\pi_1}}(s_t) \right) \\
\{ \text{по определению \eqref{advantage}} \} &= \E_{\textcolor{ChadPurple}{\Traj \sim \pi_2} \mid s_0 = s} \sum_{t \ge 0} \gamma^t \textcolor{ChadBlue}{A^{\pi_1}}(s_t, a_t)   \tagqed
\end{align*}
\end{theoremBox}

Мы смогли записать наш функционал как мат.ожидание по траекториям, сгенерированным одной политикой, по оценочной функции другой стратегии. Фактически, мы можем награду заменить Advantage-функцией произвольной другой стратегии, и это сдвинет оптимизируемый функционал на константу! Прикольно.

Конечно, это теоретическое утверждение, поскольку на практике узнать точно оценочную функцию какой-то другой стратегии достаточно сложно (хотя ничто не мешает в качестве потенциала использовать произвольную функцию, приближающую $\textcolor{ChadBlue}{V^{\pi_1}}(s)$). Однако в этой <<новой>> награде замешаны сигналы из будущего, награды, которые будут получены через много шагов, и эта <<новая>> награда априори информативнее исходной $r(s, a)$.

Представим, что мы оптимизировали исходный функционал
$$\E_{\textcolor{ChadPurple}{\Traj \sim \pi_2} \mid s_0 = s} \sum_{t \ge 0} \gamma^t r(s_t, a_t) \to \max_{\textcolor{ChadPurple}{\pi_2}}$$
и сказали: слушайте, мы не знаем, как управлять марковской цепью, не очень понимаем, как выбор тех или иных действий в состоянии влияет на структуру траектории $p(\textcolor{ChadPurple}{\Traj \mid \pi_2})$. А давайте мы притворимся, что у нас нет в задаче отложенного сигнала (что очень существенное упрощение), и будем просто во всех состояниях $s$ оптимизировать $r(s, a)$: выбирать <<хорошие>> действия $a$, где функция награды высокая. То есть будем просто выбирать $\textcolor{ChadPurple}{\pi_2}(s) = \argmax\limits_{a} r(s, a)$. Смысла в этом будет мало.

Теперь же мы преобразовали функционал, сменив функцию награды:
$$\E_{\textcolor{ChadPurple}{\Traj \sim \pi_2} \mid s_0 = s} \sum_{t \ge 0} \gamma^t \textcolor{ChadBlue}{A^{\pi_1}}(s_t, a_t) \to \max_{\textcolor{ChadPurple}{\pi_2}}$$

Что, если мы поступим также с новой наградой? Мы, например, знаем, что Advantage --- не произвольная функция, и она обязана в среднем равняться нулю (утв. \ref{pr:advantageiszero}). Значит, если мы выберем
$$\textcolor{ChadPurple}{\pi_2}(s) = \argmax_{a} \textcolor{ChadBlue}{A^{\pi_1}}(s, a),$$
то все встречаемые пары $(s, a)$ в траекториях из $\pi_2$ будут обязательно с неотрицательными наградами за шаг $\textcolor{ChadBlue}{A^{\pi_1}}(s, a) \ge 0$. Значит и вся сумма наград будет положительна для любого стартового состояния:
$$\E_{\textcolor{ChadPurple}{\Traj \sim \pi_2} \mid s_0 = s} \sum_{t \ge 0} \gamma^t \underbrace{\textcolor{ChadBlue}{A^{\pi_1}}(s_t, a_t)}_{\ge 0} \ge 0$$

И тогда из теоремы \ref{th:rpi} об RPI мы можем заключить, что для любого $s$:
$$\textcolor{ChadPurple}{V^{\pi_2}}(s) - \textcolor{ChadBlue}{V^{\pi_1}}(s) \ge 0$$
Это наблюдение - ключ к оптимизации стратегии при известной оценочной функции другой стратегии.

\subsection{Policy Improvement}

\begin{definition}
Будем говорить, что стратегия $\pi_2$ <<\emph{не хуже}>> $\pi_1$ (запись: $\pi_2 \succeq \pi_1$), если $\forall s \colon$
$$V^{\pi_2}(s) \ge V^{\pi_1}(s),$$
и \emph{лучше} (запись: $\pi_2 \succ \pi_1$), если также найдётся $s$, для которого неравенство выполнено строго:
$$V^{\pi_2}(s) > V^{\pi_1}(s)$$
\end{definition}

Мы ввели частичный порядок на множестве стратегий (понятно, что можно придумать две стратегии, которые будут <<не сравнимы>>: когда в одном состоянии одна будет набирать больше второй, в другом состоянии вторая будет набирать больше первой).

Зададимся следующим вопросом. Пусть для стратегии $\pi_1$ мы знаем оценочную функцию $Q^{\pi_1}$; тогда мы знаем и $V^{\pi_1}$ из VQ уравнения \eqref{VQ} и $A^{\pi_1}$ по определению \eqref{advantage}. Давайте попробуем построить $\pi_2 \succ \pi_1$. Для этого покажем более <<классическим>> способом, что стратегии $\pi_2$ достаточно лишь в среднем выбирать действия, дающие неотрицательный Advantage стратегии $\pi_1$, чтобы быть не хуже.

\begin{theoremBox}[label=th:policyimprovement]{Policy Improvement}
Пусть стратегии $\textcolor{ChadBlue}{\pi_1}$ и $\textcolor{ChadPurple}{\pi_2}$ таковы, что для всех состояний $s$ выполняется:
$$\E_{\textcolor{ChadPurple}{\pi_2}(a \mid s)} \textcolor{ChadBlue}{Q^{\pi_1}}(s, a) \ge \textcolor{ChadBlue}{V^{\pi_1}}(s),$$
или, в эквивалентной форме:
$$\E_{\textcolor{ChadPurple}{\pi_2}(a \mid s)} \textcolor{ChadBlue}{A^{\pi_1}}(s, a) \ge 0.$$
Тогда $\textcolor{ChadPurple}{\pi_2} \succeq \textcolor{ChadBlue}{\pi_1}$; если хотя бы для одного $s$ неравенство выполнено строго, то $\textcolor{ChadPurple}{\pi_2} \succ \textcolor{ChadBlue}{\pi_1}$.
\begin{proof}
Покажем, что $\textcolor{ChadPurple}{V^{\pi_2}}(s) \ge \textcolor{ChadBlue}{V^{\pi_1}}(s)$ для любого $s$:
\begin{align*}
\textcolor{ChadBlue}{V^{\pi_1}}(s) = \{ \text{связь VQ \eqref{VQ}} \} &= \E_{\textcolor{ChadBlue}{\pi_1}(a \mid s)} \textcolor{ChadBlue}{Q^{\pi_1}}(s, a) \le \\
= \{ \text{по построению $\textcolor{ChadPurple}{\pi_2}$} \} &= \E_{\textcolor{ChadPurple}{\pi_2}(a \mid s)} \textcolor{ChadBlue}{Q^{\pi_1}}(s, a) = \\
= \{ \text{связь QV \eqref{QV}} \} &= \E_{\textcolor{ChadPurple}{\pi_2}(a \mid s)} \left[ r + \gamma \E_{s'} \textcolor{ChadBlue}{V^{\pi_1}}(s') \right] \le \\
\le \{ \text{применяем это же неравенство рекурсивно} \} &= \E_{\textcolor{ChadPurple}{\pi_2}(a \mid s)} \left[ r + \E_{s'} \E_{\textcolor{ChadPurple}{\pi_2}(a' \mid s')} \left[ \gamma r' + \gamma^2 \E_{s''} \textcolor{ChadBlue}{V^{\pi_1}}(s'') \right] \right] \le \\
\le \{ \text{раскручиваем цепочку далее} \} &\le \dots \le \E_{\textcolor{ChadPurple}{\Traj \sim \pi_2} \mid s_0 = s} \sum_{t \ge 0} \gamma^t r_t = \\
= \{ \text{по определению \eqref{Vdefinition}} \} &= \textcolor{ChadPurple}{V^{\pi_2}}(s)
\end{align*}
Если для какого-то $s$ неравенство из условия теоремы было выполнено строго, то для него первое неравенство в этой цепочке рассуждений выполняется строго, и, значит, $\textcolor{ChadPurple}{V^{\pi_2}}(s) > \textcolor{ChadBlue}{V^{\pi_1}}(s)$.
\end{proof}
\end{theoremBox}

Что означает эта теорема? Знание оценочной функции позволяет улучшить стратегию. Улучшать стратегию можно прямо в отдельных состояниях, например, выбрав некоторое состояние $s$ и сказав: неважно, как это повлияет на частоты посещения состояний, но будем конкретно в этом состоянии $s$ выбирать действия так, что значение
\begin{equation}\label{pi_optimization}
\E_{\textcolor{ChadPurple}{\pi_2}(a \mid s)} \textcolor{ChadBlue}{Q^{\pi_1}}(s, a)    
\end{equation}
как можно больше. Тогда, если в $s$ действие выбирается <<новой>> стратегией $\textcolor{ChadPurple}{\pi_2}$, а в будущем агент будет вести себя \textit{не хуже}, чем $\textcolor{ChadBlue}{\pi_1}$, то и наберёт он в будущем не меньше $\textcolor{ChadBlue}{Q^{\pi_1}}(s, a)$. Доказательство теоремы \ref{th:policyimprovement} показывает, что выражение \eqref{pi_optimization} является нижней оценкой на награду, которую соберёт <<новый>> агент со стратегией $\textcolor{ChadPurple}{\pi_2}$. 

Если эта нижняя оценка поднята выше $\textcolor{ChadBlue}{V^{\pi_1}}(s)$, то стратегию удалось улучшить: и тогда какой бы ни была $\pi_1$, мы точно имеем гарантии $\pi_2 \succeq \pi_1$. Важно, что такой policy improvement работает всегда: и для <<тупых>> стратегий, близких к случайному поведению, и для уже умеющих что-то разумное делать.

В частности, мы можем попробовать нижнюю оценку \eqref{pi_optimization} максимально поднять, то есть провести \emph{жадный} (greedy) policy improvement. Для этого мы формально решаем такую задачу оптимизации:
$$
\E_{\textcolor{ChadPurple}{\pi_2}(a \mid s)} \textcolor{ChadBlue}{Q^{\pi_1}}(s, a) \to \max_{\textcolor{ChadPurple}{\pi_2}},
$$
и понятно, что решение находится в детерминированной $\pi_2$:
$$\textcolor{ChadPurple}{\pi_2}(s) = \argmax_{a} \textcolor{ChadBlue}{Q^{\pi_1}}(s, a)= \argmax_{a} \textcolor{ChadBlue}{A^{\pi_1}}(s, a)$$


Конечно, мы так не получим <<за один ход>> сразу оптимальную стратегию, поскольку выбор $\textcolor{ChadPurple}{\pi_2}(a \mid s)$ сколь угодно хитро может изменить распределение траекторий, но тем не менее.

\needspace{15\baselineskip}
\begin{example}
Попробуем улучшить стратегию $\pi$ из примера \ref{ex:vfunction}, $\gamma = 0.8$. Например, в состоянии C она выбирает \colorsquare{ChadRed} с вероятностью 1 и получает -1; попробуем посчитать $Q^{\pi}(s \HM= C, \colorsquare{ChadBlue})$:
$$
Q^{\pi}(s = C, \colorsquare{ChadBlue}) = 0.2\gamma V^{\pi}(C) + 0.8 \gamma V^{\pi}(B)
$$
\needspace{13\baselineskip}
\begin{wrapfigure}{r}{0.45\textwidth}
\vspace{-0.8cm}
\centering
\includegraphics[width=0.4\textwidth]{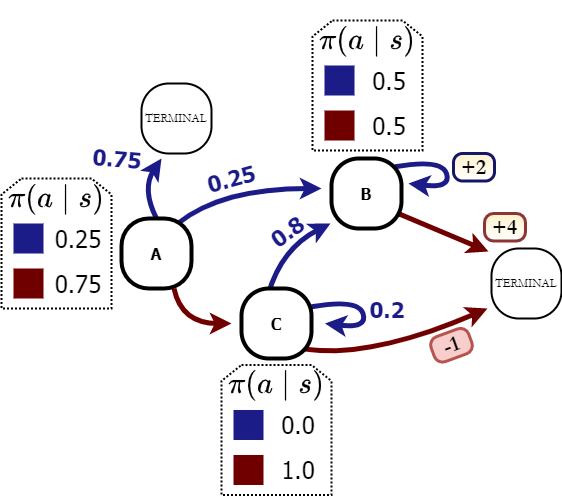}
\end{wrapfigure}
Подставляя ранее подсчитанные $V^{\pi}(C) \HM= -1, V^{\pi}(B) \HM= 5$, видим, что действие \colorsquare{ChadBlue} принесло бы нашей стратегии $\pi$ куда больше -1, а именно $Q^{\pi}(s = C, \colorsquare{ChadBlue}) \HM= 3.04$. Давайте построим $\pi_2$, скопировав $\pi$ в A и B, а в C будем с вероятностью 1 выбирать \colorsquare{ChadBlue}.

Что говорит нам теория? Важно, что она не даёт нам значение $V^{\pi_2}(C)$; в частности, нельзя утверждать, что $Q^{\pi_2}(s = C, \colorsquare{ChadBlue}) \HM= 3.04$, и повторение вычислений подтвердит, что это не так. Однако у нас есть гарантии, что, во-первых, $Q^{\pi_2}(s = C, \colorsquare{ChadBlue}) \ge 3.04$, и, что важнее, из состояния C мы начали набирать больше награды: $V^{\pi_2}(C) \HM> V^{\pi_1}(C)$ строго. Во-вторых, есть гарантии, что мы не <<сломали>> стратегию в других состояниях: во всех остальных состояниях гарантированно $V^{\pi_2}(s) \HM\ge V^{\pi_1}(s)$. Для Q-функции, как можно показать, выполняются аналогичные неравенства.
\end{example}

\subsection{Вид оптимальной стратегии (доказательство через PI)}

Что, если для некоторой $\textcolor{ChadBlue}{\pi_1}$ мы <<не можем>> провести Policy Improvement? Под этим будем понимать, что мы не можем выбрать $\textcolor{ChadPurple}{\pi_{2}}$ так, что $\E_{\textcolor{ChadPurple}{\pi_{2}}(a \mid s)} \textcolor{ChadBlue}{Q^{\pi_{1}}}(s, a) \HM> \textcolor{ChadBlue}{V^{\pi_{1}}}(s)$ строго хотя бы для одного состояния $s$ (ну, равенства в любом состоянии $s$ мы добьёмся всегда, скопировав $\textcolor{ChadBlue}{\pi_1}(\cdot \HM\mid s)$). Такое может случиться, если и только если $\textcolor{ChadBlue}{\pi_1}$ удовлетворяет следующему свойству:
$$\max_a \textcolor{ChadBlue}{Q^{\pi_{1}}}(s, a) = \textcolor{ChadBlue}{V^{\pi_{1}}}(s) \quad \Leftrightarrow \quad \max\limits_a \textcolor{ChadBlue}{A^{\pi_1}}(s, a) = 0$$

Но это в точности критерий оптимальности Беллмана, теорема \ref{th:optimalitycriterion}! Причём мы можем, воспользовавшись теоремами RPI \ref{th:rpi} и о Policy Improvement \ref{th:policyimprovement}, теперь доказать этот критерий альтернативным способом, не прибегая к формализму оптимальных оценочных функций\footnote{в доказательствах RPI и Policy Improvement мы не использовали понятия $Q^*$ и $V^*$ и их свойства; тем не менее, из этих теорем все свойства оптимальных оценочных функций следуют: например, пусть $A^*, Q^*, V^*$ --- оценочные функции оптимальных стратегий, тогда в силу выводимого из RPI критерия оптимальности (теорема \ref{th:optimalitycriterion_pi}) $\forall s \colon \max\limits_a A^*(s, a) \HM= 0$, или, что тоже самое, $\max\limits_a \left[ Q^*(s, a) \HM- V^*(s) \right] \HM= 0$; отсюда $V^*(s) \HM= \max\limits_a Q^*(s, a)$. Аналогично достаточно просто можно получить все остальные утверждения об оптимальных оценочных функциях, не прибегая к рассуждению с отказом от стационарности.} и не требуя рассуждения про отказ от стационарности и обоснования единственности решения уравнений оптимальности Беллмана.

\begin{theoremBox}[label=th:optimalitycriterion_pi]{Критерий оптимальности (альт. доказательство)}
$\pi$ оптимальна тогда и только тогда, когда $\forall s \colon \max\limits_a A^\pi(s, a) = 0$.
\begin{proof}[Достаточность]
Допустим, это не так, и существует $\pi_2, s \colon V^{\pi_2}(s) > V^{\pi}(s)$. Тогда по RPI \eqref{RPI}
$$\E_{\Traj \sim \pi_2 \mid s_0 = s} \sum_{t \ge 0} \gamma^t A^{\pi}(s_t, a_t) > 0,$$
однако все слагаемые в сумме неположительны. Противоречие.
\end{proof}
\begin{proof}[Необходимость]
Допустим, что $\pi$ оптимальна, но для некоторого $\hat{s}$ условие не выполнено, и $\max\limits_{a} A^\pi(\hat{s}, a) > 0$ (меньше нуля он, ещё раз, быть не может в силу утв.~ \ref{adv_is_positive}). Рассмотрим детерминированную $\pi_2$, которая в состоянии $\hat{s}$ выбирает какое-нибудь $\hat{a}$, такое что $A^\pi(\hat{s}, \hat{a}) > 0$ (это можно сделать по условию утверждения --- сам максимум может вдруг оказаться недостижим для сложных пространств действий, но какое-то действие с положительным advantage-ем мы найдём), а в остальных состояниях выбирает какое-нибудь действие, т.ч. advantage-функция неотрицательна. Тогда  
$$V^{\pi_2}(\hat{s}) - V^{\pi}(\hat{s}) = \E_{\Traj \sim \pi_2 | s_0 = \hat{s}} \sum_{t \ge 0} \gamma^t A^{\pi}(s_t, a_t) > 0$$
поскольку все слагаемые неотрицательны, и во всех траекториях с вероятностью\footnote[*]{мы специально стартовали из $\hat{s}$, чтобы пары $\hat{s}, \hat{a}$ <<встретились>> в траекториях, иначе могло бы быть такое, что агент в это состояние $\hat{s}$ <<никогда не попадает>>, и отделиться от нуля не получилось бы.} 1 верно $s_0 \HM= \hat{s}, a_0 \HM= \hat{a}$, то есть первое слагаемое равно $A(\hat{s}, \hat{a}) \HM> 0$. 
\end{proof} 
\end{theoremBox}

Итак, мораль полученных результатов такая: зная $Q^{\pi}$, мы можем придумать стратегию лучше. Не можем --- значит, наша текущая стратегия $\pi$ уже оптимальная.




\section{Динамическое программирование}\label{valueiterationsection}

\subsection{Метод простой итерации}

\begin{wrapfigure}{r}{0.45\textwidth}
\vspace{-1cm}
\centering
\includegraphics[width=0.4\textwidth]{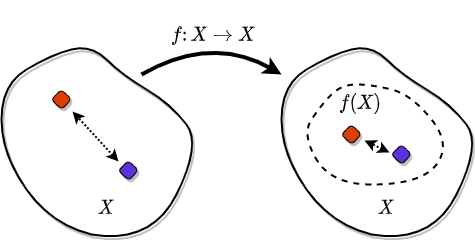}
\vspace{-0.4cm}
\end{wrapfigure}

Мы увидели, что знание оценочных функций открывает путь к улучшению стратегии. Напрямую по определению считать их затруднительно; попробуем научиться решать уравнения Беллмана. И хотя уравнения оптимальности Беллмана нелинейные, они, тем не менее, имеют весьма определённый вид и, как мы сейчас увидим, обладают очень приятными свойствами. Нам понадобится несколько понятий внезапно из функана о том, как решать системы нелинейных уравнений вида $x \HM= f(x)$.

\begin{definition}
Оператор $f \colon X \to X$ называется \emph{сжимающим} (contraction) с коэффициентом сжатия $\gamma < 1$ по некоторой метрике $\rho$, если $\forall x_1, x_2 \colon$
$$\rho(f(x_1), f(x_2)) < \gamma \rho(x_1, x_2)$$
\end{definition}

\begin{definition}
Точка $x \in X$ для оператора $f \colon X \to X$ называется \emph{неподвижной} (fixed point), если
$$x = f(x)$$
\end{definition}

\begin{definition}
Построение последовательности $x_{k+1} = f(x_k)$ для начального приближения $x_0 \in X$ называется методом \emph{простой итерации} (point iteration) решения уравнения $x = f(x)$.
\end{definition}

\begin{theorem}[Теорема Банаха о неподвижной точке]\label{Banach}
В полном\footnote[*]{любая фундаментальная последовательность имеет предел} метрическом пространстве $X$ у сжимающего оператора $f \colon X \to X$ существует и обязательно единственна неподвижная точка $x^*$, причём метод простой итерации сходится к ней из любого начального приближения.
\begin{proof}[Сходимость метода простой итерации] Пусть $x_0$ --- произвольное, $x_{k+1} = f(x_k)$. Тогда для любого $k > 0$:
\begin{equation}
\begin{aligned}\label{onestepcontraction}
\rho(x_k, x_{k+1}) = \{ \text{определение $x_k$} \} &= \rho(f(x_{k-1}), f(x_k)) \le \\ 
\le \{ \text{свойство сжатия} \} &\le \gamma \rho(x_{k-1}, x_k) \le \dots \le \\
\le \{ \text{аналогичным образом} \} &\le \dots \le \gamma^k \rho(x_0, x_1)
\end{aligned}
\end{equation}

Теперь посмотрим, что произойдёт после применения оператора $f$ $n$ раз:
\begin{align*}
\rho(x_k, x_{k + n}) &\le \\
\{ \text{неравенство треугольника} \}
&\le \rho(x_k, x_{k + 1}) + \rho(x_{k+1}, x_{k+2}) + \dots + \rho(x_{k + n - 1}, x_{k + n}) \le \\
\le \{ \text{\eqref{onestepcontraction}} \} 
&\le (\gamma^k + \gamma^{k+1} + \dots \gamma^{k+n-1}) \rho(x_0, x_1) \le \\
\le \{ \text{геом. прогрессия} \} &\le \frac{\gamma^k}{1 - \gamma} \rho(x_0, x_1) \xrightarrow{k \to \infty} 0
\end{align*}

Итак, последовательность $x_k$ --- фундаментальная, и мы специально попросили такое метрическое пространство (<<полное>>), в котором обязательно найдётся предел $x^* \coloneqq \lim\limits_{k \to \infty} x_k$.
\end{proof}
\begin{proof}[Существование неподвижной точки] Покажем, что $x^*$ и есть неподвижная точка $f$, то есть покажем, что наш метод простой итерации конструктивно её построил. Заметим, что для любого $k > 0$:
\begin{align*}
\rho(x^*, f(x^*)) &\le \\
\le \{ \text{неравенство треугольника} \} &\le \rho(x^*, x_k) + \rho(x_k, f(x^*)) = \\
= \{ \text{определение $x_k$} \} &= \rho(x^*, x_k) + \rho(f(x_{k-1}), f(x^*)) \le \\ 
\le \{ \text{свойство сжатия} \} &\le \rho(x^*, x_k) + \gamma \rho(x_{k-1}, x^*)
\end{align*}
Устремим $k \to \infty$; слева стоит константа, не зависящая от $k$. Тогда расстояние между $x_k$ и $x^*$ устремится к нулю, ровно как и между $x_{k-1}, x^*$ поскольку $x^*$ --- предел $x_k$. Значит, константа равна нулю, $\rho(x^*, f(x^*)) = 0$, следовательно, $x^* = f(x^*)$.
\end{proof}

\begin{proof}[Единственность] Пусть $x_1, x_2$ --- две неподвижные точки оператора $f$. Ну тогда:
$$\rho(x_1, x_2) = \rho(f(x_1), f(x_2)) \le \gamma \rho(x_1, x_2)$$
Получаем, что такое возможно только при $\rho(x_1, x_2) = 0$, то есть только если $x_1$ и $x_2$ совпадают.
\end{proof}
\end{theorem}


\subsection{Policy Evaluation}

Вернёмся к RL. Известно, что $V^\pi$ для данного MDP и фиксированной политики $\pi$ удовлетворяет уравнению Беллмана \eqref{VV}. Для нас это система уравнений относительно значений $V^\pi(s)$. $V^\pi(s)$ --- объект (точка) в функциональном пространстве $\St \to \R$.

Будем решать её методом простой итерации\footnote{вообще говоря, это система линейных уравнений относительно значений $V^\pi(s)$, которую в случае табличных MDP можно решать любым методом решения СЛАУ. Однако, дальнейшие рассуждения через метод простой итерации обобщаются, например, на случай непрерывных пространств состояний $\St \HM\subseteq \R^n$.}. Для этого определим оператор $\B$, то есть преобразование из одной функции $\St \HM\to \R$ в другую. На вход этот оператор принимает функцию $V \colon \St \HM\subseteq \R^n$ и выдаёт некоторую другую функцию от состояний $\B V$. Чтобы задать выход оператора, нужно задать значение выходной функции в каждом $s \HM\in \St$; это значение мы будем обозначать $\left[\B V\right] (s)$ (квадратные скобки позволяют не путать применение оператора с вызовом самой функции) и определим его как правую часть решаемого уравнения \eqref{VV}. Итак:

\begin{definition}
Введём \emph{оператор Беллмана} (Bellman operator) для заданного MDP и стратегии $\pi$ как
\begin{equation}\label{bellmanoperatorV}
\left[\B V\right] (s) \coloneqq \E_{a \sim \pi(a \mid s)} \left[ r(s, a) + \gamma \E_{s'} V(s') \right]
\end{equation}
\end{definition}

Также нам нужна метрика на множестве функций $\St \to \R$; возьмём
$$d_\infty(V_1, V_2) \coloneqq \max_s | V_1(s) - V_2(s) |$$

\begin{theorem}
Если $\gamma < 1$, оператор $\B$ --- сжимающий с коэффициентом сжатия $\gamma$.
\beginproof
\begin{align*}
&d_\infty(\B V_1, \B V_2) = \max_s \left| [\B V_1](s) - [\B V_2](s) \right| = \\
&= \{ \text{подставляем значение операторов, т.е. правые части решаемого уравнения} \} = \\
&= \max_s \left| \E_{a} \left[ r(s, a) + \gamma \E_{s'} V_1(s') \right] - \E_{a} \left[ r(s, a) + \gamma \E_{s'} V_2(s') \right] \right| = \\
&= \{ \text{слагаемые $r(s, a)$ сокращаются} \} = \\
&= \gamma \max_s \left| \E_{a} \E_{s'} \left[ V_1(s') - V_2(s') \right] \right| \le \\
&\le \{ \text{используем свойство $\E_x f(x) \le \max\limits_x f(x)$} \} \le \\
&\le \gamma \max_s \max_{s'} \left| V_1(s') - V_2(s') \right| = \gamma d_\infty( V_1, V_2 ) \tagqed
\end{align*}
\end{theorem}

Итак, мы попали в теорему Банаха, и значит, метод простой итерации
$$V_{k+1} \coloneqq \B V_k$$
гарантированно сойдётся к единственной неподвижной точке при любой стартовой инициализации $V_0$. По построению мы знаем, что $V^\pi(s)$ такова, что $\B V^\pi = V^\pi$ (это и есть уравнение Беллмана), поэтому к ней и придём.

Важно помнить, что на каждой итерации такой процедуры текущее приближение не совпадает с истинной оценочной функцией: $V_k(s) \HM\approx V^\pi(s)$, но точного равенства поставить нельзя. Распространено (но, к сожалению, не везде применяется) соглашение обозначать аппроксимации оценочных функций без верхнего индекса: просто $V$ или $Q$. Однако, иногда, чтобы подчеркнуть, что алгоритм учит именно $V^\pi$, верхний индекс оставляют, что может приводить к путанице.

Обсудим, что случится в ситуации, когда $\gamma = 1$; напомним, что в таких ситуациях мы требовали эпизодичность сред, с гарантиями завершения всех эпизодов за $T^{\max}$ шагов. Оператор Беллмана формально сжатием являться уже не будет, и мы не подпадаем под теорему, поэтому этот случай придётся разобрать отдельно.

\begin{theoremBox}[label=th:policyevalepisodic]{}
В эпизодичных средах метод простой итерации сойдётся к единственному решению уравнений Беллмана не более чем за $T^{\max}$ шагов даже при $\gamma = 1$.
\begin{proof}
Мы уже доказывали теорему \ref{th:episodicmdpistree}, что граф таких сред является деревом. Будем говорить, что состояние $s$ находится на ярусе $T$, если при старте из $s$ у любой стратегии есть гарантии завершения за $T$ шагов. Понятно, что для состояния $s$ на ярусе $T$ верно, что $\forall s', a$, для которых $p(s' \mid s, a) > 0$, ярус $s'$ не превосходит $T - 1$. 

Осталось увидеть, что на $k$-ой итерации метода простой итерации вычисляет точные значения $V_k(s) \HM= V^{\pi}(s)$ для всех состояний на ярусах до $k$: действительно, покажем по индукции. Считаем, что терминальные состояния имеют нулевой ярус; а на $k$-ом шаге при обновлении $V_{k+1}(s) \HM\coloneqq \left[\B V_k\right](s)$ для $s$ на $k$-ом ярусе в правой части уравнения Беллмана будет стоять мат.~ожидание по $s'$ с ярусов до $k - 1$-го, для которых значение по предположению индукции уже посчитано точно.

Соответственно, за $T^{\max}$ шагов точные значения распространятся на все состояния, и конструктивно значения определены однозначно.
\end{proof}
\end{theoremBox}

\begin{remark}
Если $\gamma = 1$, а среда неэпизодична (такие MDP мы не допускали к рассмотрению), метод простой итерации может не сойтись, а уравнения Беллмана могут в том числе иметь бесконечно много решений. Пример подобного безобразия. Пусть в MDP без терминальных состояний с нулевой функцией награды (где, очевидно, $V^{\pi}(s) \HM= 0$ для всех $\pi, s$) мы проинициализировали $V_0(s) \HM= 100$ во всех состояниях $s$. Тогда при обновлении наша аппроксимация не будет меняться: мы уже в неподвижной точке уравнений Беллмана. В частности поэтому на практике практически никогда не имеет смысл выставлять $\gamma \HM = 1$, особенно в сложных средах, где, может быть, даже и есть эпизодичность, но, тем не менее, есть <<похожие состояния>>: они начнут работать <<как петли>>, когда мы перейдём к приближённым методам динамического программирования в дальнейшем.
\end{remark}

\begin{proposition}
Если некоторая функция $\tilde{V} \colon \St \to \R$ удовлетворяет уравнению Беллмана \eqref{VV}, то $\tilde{V} \equiv V^\pi$.
\end{proposition}

Мы научились решать задачу \emph{оценивания стратегии} (Policy Evaluation): вычислять значения оценочной функции по данной стратегии $\pi$ в ситуации, когда мы знаем динамику среды. На практике мы можем воспользоваться этим результатом только в \emph{<<табличном>> случае} (tabular RL), когда пространство состояний и пространство действий конечны и достаточно малы, чтобы все пары состояние-действие было возможно хранить в памяти компьютера и перебирать за разумное время. В такой ситуации $V^\pi(s)$ --- конечный векторочек, и мы умеем считать оператор Беллмана и делать обновления $V_{k+1} \HM= \B V_k$.

\begin{algorithm}[label=policyevaluation]{Policy Evaluation}
\textbf{Вход:} $\pi(a \mid s)$ --- стратегия \\
\textbf{Гиперпараметры:} $\eps$ --- критерий останова

\vspace{0.3cm}
Инициализируем $V_0(s)$ произвольно для всех $s \in \St$ \\
\textbf{На $k$-ом шаге:}
\begin{enumerate}
    \item $\forall s \colon V_{k+1}(s) \coloneqq \E_{a} \left[ r(s, a) + \gamma \E_{s'} V_k(s')\right]$
    \item \textbf{критерий останова:} $\max\limits_s |V_k(s) - V_{k+1}(s)| < \eps$
\end{enumerate}

\vspace{0.3cm}
\textbf{Выход:} $V_k(s)$
\end{algorithm}

\begin{exampleBox}[label=ex:pe, righthand ratio=0.25, sidebyside, sidebyside align=center, lower separated=false]{}
Проведём оценивание стратегии, случайно выбирающей, в какую сторону ей пойти, с $\gamma = 0.9$. Угловые клетки с ненулевой наградой терминальны; агент остаётся в той же клетке, если упирается в стенку. На каждой итерации отображается значение текущего приближения $V_k(s) \HM\approx V^{\pi}(s)$.

\tcblower
\animategraphics[controls, width=0.9\linewidth]{1}{Images/PE/policyeval}{0}{9}
\end{exampleBox}

Итак, мы научились считать $V^\pi$ в предположении известной динамики среды. Полностью аналогичное рассуждение верно и для уравнений QQ \eqref{QQ}; то есть, расширив набор переменных, в табличных MDP можно методом простой итерации находить $Q^{\pi}$ и <<напрямую>>. Пока модель динамики среды считается известной, это не принципиально: мы можем посчитать и Q-функцию через V-функцию по формуле QV \eqref{QV}.





\subsection{Value Iteration}

Теорема Банаха позволяет аналогично Policy Evaluation (алг. \ref{policyevaluation}) решать уравнения оптимальности Беллмана \eqref{Q*Q*} через метод простой итерации. Действительно, проведём аналогичные рассуждения (мы сделаем это для $Q^*$, но совершенно аналогично можно было бы сделать это и для $V^*$):

\begin{definition}
Определим \emph{оператор оптимальности Беллмана} (Bellman optimality operator, Bellman control operator) $\B^*$:
$$\left[\B^* Q\right] (s, a) \coloneqq r(s, a) + \gamma \E_{s'} \max_{a'}Q(s', a')$$
\end{definition}

В качестве метрики на множестве функций $\St \times \A \to \R$ аналогично возьмём
$$d_\infty(Q_1, Q_2) \coloneqq \max_{s, a} | Q_1(s, a) - Q_2(s, a) |$$

Нам понадобится следующий факт:
\begin{proposition}\,
\begin{equation}\label{diffmax}
| \max_x f(x) - \max_x g(x) | \le \max_x | f(x) - g(x) |
\end{equation}
\begin{proof}
Рассмотрим случай $\max\limits_x f(x) \HM> \max\limits_x g(x)$. Пусть $x^*$ --- точка максимума $f(x)$. Тогда:
\begin{align*}
    \max_x f(x) - \max_x g(x) = f(x^*) - \max_x g(x) \le f(x^*) - g(x^*) \le \max_x | f(x) - g(x) |
\end{align*}
Второй случай рассматривается симметрично.
\end{proof}
\end{proposition}

\begin{theorem}
Если $\gamma < 1$, оператор $\B^*$ --- сжимающий.
\beginproof
\begin{align*}
&d_\infty(\B^*Q_1, \B^*Q_2) = \max_{s, a} \left| [\B^*Q_1](s, a) - [\B^*Q_2](s, a) \right| = \\
&= \{ \text{подставляем значения операторов, т.е. правые части решаемой системы уравнений} \} = \\
&= \max_{s, a} \left| \left[ r(s, a) + \gamma \E_{s'} \max_{a'} Q_1(s', a') \right] - \left[ r(s, a) + \gamma \E_{s'} \max_{a'} Q_2(s', a') \right] \right| = \\
&= \{ \text{слагаемые $r(s, a)$ сокращаются} \} = \\
&= \gamma \max_{s, a} \left| \E_{s'} \left[ \max_{a'} Q_1(s', a') - \max_{a'} Q_2(s', a') \right] \right| \le \\
&\le \{ \text{используем свойство $\E_x f(x) \le \max\limits_x f(x)$} \} \le \\
&\le \gamma \max_{s, a} \max_{s'} \left| \max_{a'} Q_1(s', a') - \max_{a'} Q_2(s', a') \right| = \\
&\le \{ \text{используем свойство максимумов \eqref{diffmax}} \} \le \\
& \le \gamma \max_{s, a} \max_{s'} \max_{a'} \left| Q_1(s', a') - Q_2(s', a') \right| \le \\
&= \{\text{внутри стоит определение $d_\infty(Q_1, Q_2)$, а от внешнего максимума ничего не зависит}\} = \\
&= \gamma d_\infty( Q_1, Q_2 )   \tagqed
\end{align*}
\end{theorem}

\begin{theorem}
В эпизодичных средах метод простой итерации сойдётся к единственному решению уравнений оптимальности Беллмана не более чем за $T^{\max}$ шагов даже при $\gamma = 1$.
\begin{proof} Полностью аналогично доказательству теоремы \ref{th:policyevalepisodic}.
\end{proof}
\end{theorem}

\begin{proposition}
Если некоторая функция $\tilde{Q} \colon \St \times \A \to \R$ удовлетворяет уравнению оптимальности Беллмана \eqref{Q*Q*}, то $\tilde{Q} \equiv Q^*$.
\end{proposition}

\begin{proposition}
Метод простой итерации сходится к $Q^*$ из любого начального приближения.
\end{proposition}

Вообще, если известна динамика среды, то нам достаточно решить уравнения оптимальности для $V^*$ --- это потребует меньше переменных. Итак, в табличном случае мы можем напрямую методом простой итерации решать уравнения оптимальности Беллмана и в пределе сойдёмся к оптимальной оценочной функции, которая тут же даёт нам оптимальную стратегию.

\begin{algorithm}[label=valueiteration]{Value Iteration}
\textbf{Вход:} $\varepsilon$ --- критерий останова

\vspace{0.3cm}
Инициализируем $V_0(s)$ произвольно для всех $s \in \St$ \\
\textbf{На $k$-ом шаге:}
\begin{enumerate}
    \item для всех $s$: $V_{k+1}(s) \coloneqq \max\limits_a \left[ r(s, a) + \gamma \E_{s'} V_k(s')\right]$
    \item \textbf{критерий останова:} $\max\limits_{s} | V_{k+1}(s) - V_k(s) | < \varepsilon$
\end{enumerate}

\vspace{0.3cm}
\textbf{Выход:} $\pi(s) \coloneqq \argmax\limits_a \left[ r(s, a) + \gamma \E_{s'} V(s')\right]$
\end{algorithm}

\begin{wrapfigure}{r}{0.35\textwidth}
\centering
\includegraphics[width=0.3\textwidth]{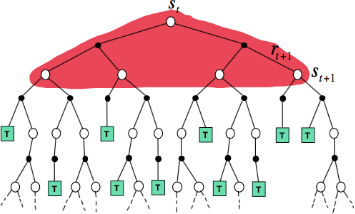}
\end{wrapfigure}

Итак, мы придумали наш первый табличный алгоритм планирования --- алгоритм, решающий задачу RL в условиях известной модели среды. На каждом шаге мы обновляем (<<бэкапим>>) нашу текущую аппроксимацию V-функции на её \emph{одношаговое приближение} (one-step approximation): смотрим на один шаг в будущее ($a, r, s'$) и приближаем всё остальное будущее текущей же аппроксимацией. Такой <<бэкап динамического программирования>> (dynamic programming backup, DP-backup) --- обновление <<бесконечной ширины>>: мы должны перебрать все возможные варианты следующего одного шага, рассмотреть все свои действия (по ним мы возьмём максимум) и перебрать всевозможные ответы среды --- $s'$ (по ним мы должны рассчитать мат.ожидание). Поэтому этот алгоритм в чистом виде напоминает то, что обычно и понимается под словами <<динамическое программирование>>: мы <<раскрываем дерево игры>> полностью на один шаг вперёд.

\begin{exampleBox}[righthand ratio=0.25, sidebyside, sidebyside align=center, lower separated=false]{}
Решим задачу из примера \ref{ex:pe}, $\gamma = 0.9$; на каждой итерации отображается значение текущего приближения $V_k(s) \HM\approx V^{*}(s)$. В конце концов в силу детермнированности среды станет понятно, что можно избежать попадания в терминальное -1 и кратчайшим путём добираться до терминального +1.

\tcblower
\animategraphics[controls, width=0.9\linewidth]{1}{Images/VI/valueiter}{0}{9}
\end{exampleBox}




\subsection{Policy Iteration}

\needspace{7\baselineskip}
\begin{wrapfigure}{r}{0.35\textwidth}
\vspace{-1.3cm}
\centering
\includegraphics[width=0.35\textwidth]{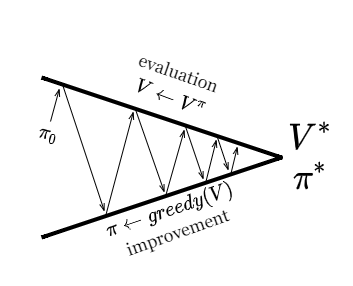}
\vspace{-1.3cm}
\end{wrapfigure}

Мы сейчас в некотором смысле <<обобщим>> Value Iteration и придумаем более общую схему алгоритма планирования для табличного случая.

Для очередной стратегии $\pi_k$ посчитаем её оценочную функцию $Q^{\pi_k}$, а затем воспользуемся теоремой Policy Improvement \ref{th:policyimprovement} и построим стратегию лучше; например, жадно:
$$\pi_{k+1}(s) \coloneqq \argmax\limits_{a} Q^{\pi_k}(s, a)$$

Тогда у нас есть второй алгоритм планирования, который, причём, перебирает детерминированные стратегии, обладающие свойством монотонного возрастания качества: каждая следующая стратегия не хуже предыдущей. Он работает сразу в классе детерминированных стратегий, и состоит из двух этапов:
\begin{itemize}
    \item \textbf{Policy Evaluation}: вычисление $Q^\pi$ для текущей стратегии $\pi$;
    \item \textbf{Policy Improvement}: улучшение стратегии $\pi(s) \leftarrow \argmax\limits_a Q^\pi(s, a)$;
\end{itemize}

При этом у нас есть гарантии, что когда алгоритм <<останавливается>> (не может провести Policy Improvement), то он находит оптимальную стратегию. Будем считать\footnote{считаем, что аргмакс берётся однозначно для любой Q-функции: в случае, если в $\Argmax$ содержится более одного элемента, множество действий как-то фиксированно упорядочено, и берётся действие с наибольшим приоритетом.}, что в такой момент остановки после проведения Policy Improvement наша стратегия не меняется: $\pi_{k+1} \equiv \pi_k$.

\begin{theorem}
В табличном сеттинге Policy Iteration завершает работу за конечное число итераций.
\begin{proof}
Алгоритм перебирает детерминированные стратегии, и, если остановка не происходит, каждая следующая лучше предыдущей:
$$\pi_k \succ \pi_{k-1} \succ \dots \succ \pi_0$$
Это означает, что все стратегии в этом ряду различны. Поскольку в табличном сеттинге число состояний и число действий конечны, детерминированных стратегий конечное число; значит, процесс должен закончится.
\end{proof}
\end{theorem}

\begin{algorithm}[label=policyiteration]{Policy Iteration}
\textbf{Гиперпараметры:} $\eps$ --- критерий останова для процедуры $\operatorname{PolicyEvaluation}$

\vspace{0.3cm}
Инициализируем $\pi_0(s)$ произвольно для всех $s \in \St$ \\
\textbf{На $k$-ом шаге:}
\begin{enumerate}
    \item $V^{\pi_k} \coloneqq \operatorname{PolicyEvaluation}(\pi_k, \eps)$
    \item $Q^{\pi_k}(s, a) \coloneqq r(s, a) + \gamma \E_{s'} V^{\pi_k}(s')$
    \item $\pi_{k+1}(s) \coloneqq \argmax\limits_a Q^{\pi_k}(s, a)$
    \item \textbf{критерий останова:} $\pi_k \equiv \pi_{k+1}$
\end{enumerate}
\end{algorithm}

\begin{exampleBox}[righthand ratio=0.5, sidebyside, sidebyside align=center, lower separated=false]{}
Решим задачу из примера \ref{ex:pe}, $\gamma = 0.9$; на каждой итерации слева отображается $V^{\pi_k}(s)$; справа улучшенная $\pi_{k+1}$. За 4 шага алгоритм сходится к оптимальной стратегии.

\tcblower
\animategraphics[controls, width=\linewidth]{1}{Images/PI/policyiter}{0}{3}
\end{exampleBox}

\subsection{Generalized Policy Iteration}

\needspace{9\baselineskip}
\begin{wrapfigure}{r}{0.35\textwidth}
\vspace{-1.3cm}
\centering
\includegraphics[width=0.3\textwidth]{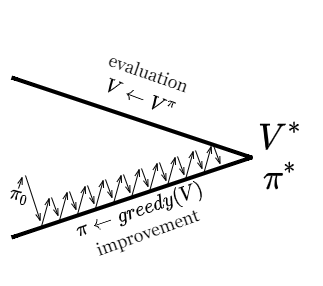}
\vspace{-0.7cm}
\end{wrapfigure}

Policy Iteration --- идеализированный алгоритм: на этапе оценивания в табличном сеттинге можно попробовать решить систему уравнений Беллмана $V^\pi$ с достаточно высокой точностью за счёт линейности этой системы уравнений, или можно считать, что проводится достаточно большое количество итераций метода простой итерации. Тогда, вообще говоря, процедура предполагает бесконечное число шагов, и на практике нам нужно когда-то остановиться; теоретически мы считаем, что доводим вычисления до некоторого критерия останова, когда значения вектора не меняются более чем на некоторую погрешность $\eps \HM> 0$. 

Но рассмотрим такую, пока что, эвристику: давайте останавливать Policy Evaluation после ровно $N$ шагов, а после обновления стратегии не начинать оценивать $\pi_{k+1}$ с нуля, а использовать последнее $V(s) \HM\approx V^{\pi_k}$ в качестве инициализации. Тогда наш алгоритм примет следующий вид: 

\begin{algorithm}[label=generalizedpolicyiteration]{Generalized Policy Iteration}
\textbf{Гиперпараметры:} $N$ --- количество шагов

\vspace{0.3cm}
Инициализируем $\pi(s)$ произвольно для всех $s \in \St$ \\
Инициализируем $V(s)$ произвольно для всех $s \in \St$ \\ 
\textbf{На $k$-ом шаге:}
\begin{enumerate}
    \item \textbf{Повторить $N$ раз:}
    \begin{itemize}
        \item $\forall s \colon V(s) \leftarrow \E_{a} \left[ r(s, a) + \gamma \E_{s'} V(s')\right]$
    \end{itemize}
    \item $Q(s, a) \leftarrow r(s, a) + \gamma \E_{s'} V(s')$
    \item $\pi(s) \leftarrow \argmax\limits_a Q(s, a)$
\end{enumerate}
\end{algorithm}

Мы формально теряем гарантии улучшения стратегии на этапе Policy Improvement, поэтому останавливать алгоритм после того, как стратегия не изменилась, уже нельзя: возможно, после следующих $N$ шагов обновления оценочной функции, аргмакс поменяется, и стратегия всё-таки сменится. Но такая схема в некотором смысле является наиболее общей, и вот почему:

\begin{proposition}
Generalized Policy Iteration (алг. \ref{generalizedpolicyiteration}) совпадает с Value Iteration (алг. \ref{valueiteration}) при $N \HM= 1$ и с Policy Iteration (алг. \ref{policyiteration}) при $N = \infty$.
\begin{proof}
Второе очевидно; увидим первое. При $N = 1$ наше обновление V-функции имеет следующий вид:
$$V(s) \leftarrow \E_{a} \left[ r(s, a) + \gamma \E_{s'} V(s') \right]$$
Вспомним, по какому распределению берётся мат.~ожидание $\E_a$: по $\pi$, которая имеет вид 
$$\pi(s) = \argmax\limits_a Q(s, a) = \argmax\limits_a \left[ r(s, a) + \gamma \E_{s'} V(s') \right]$$
Внутри аргмакса как раз стоит содержимое нашего мат.ожидания в обновлении V, поэтому это обновление выродится в
$$V(s) \leftarrow \max\limits_{a} \left[ r(s, a) + \gamma \E_{s'} V(s') \right]$$
Это в точности обновление из алгоритма Value Iteration.
\end{proof}
\end{proposition}

Итак, Generalized Policy Iteration при $N \HM= 1$ и при $N \HM= \infty$ --- это ранее разобранные алгоритмы, физический смысл которых нам ясен. В частности, теперь понятно, что в Value Iteration очередное приближение $Q_k(s, a) \HM\approx Q^*(s,a)$ можно также рассматривать как приближение $Q_k(s, a) \HM \approx Q^{\pi}(s, a)$ для $\pi(s) \HM \coloneqq \argmax\limits_{a} Q_k(s, a)$; то есть в алгоритме хоть и не потребовалось в явном виде хранить <<текущую>> стратегию, она всё равно неявно в нём присутствует. 

Давайте попробуем понять, что происходит в Generalized Policy Iteration при промежуточных $N$. Заметим, что повторение $N$ раз шага метода простой итерации для решения уравнения $\B V^\pi \HM= V^\pi$ эквивалентно одной итерации метода простой итерации для решения уравнения $\B^N V^\pi \HM= V^\pi$ (где запись $\B^N$ означает повторное применение оператора $\B$ $N$ раз), для которого, очевидно, искомая $V^{\pi}$ также будет неподвижной точкой. Что это за оператор $\B^N$? 

В уравнениях Беллмана мы <<раскручивали>> наше будущее на один шаг вперёд и дальше заменяли оставшийся <<хвост>> на определение V-функции. Понятно, что мы могли бы раскрутить не на один шаг, а на $N$ шагов вперёд.

\begin{theorem}[$N$-шаговое уравнение Беллмана]Для любого состояния $s_0$:
\begin{equation}\label{NstepBellman}
 V^{\pi}(s_0) = \E_{\Traj_{:N} \sim \pi \mid s_0} \left[ \sum_{t=0}^{N-1} \gamma^{t}r_t +  \gamma^N \E_{s_N} V^{\pi}(s_N) \right]   
\end{equation}
\begin{proof}[Доказательство по индукции]
Для получения уравнения на $N$ шагов берём $N-1$-шаговое и подставляем в правую часть раскрутку на один шаг из уравнения \eqref{VV}. Это в точности соответствует применению оператора Беллмана $N$ раз.
\end{proof}
\begin{proof}[Доказательство без индукции]
Для любых траекторий $\Traj$ верно, что
$$R(\Traj) = \sum_{t=0}^{N-1} \gamma^{t}r_t + \gamma^N R_N$$
Возьмём мат.ожидание $\E_{\Traj \sim \pi \mid s_0}$ слева и справа:
$$\E_{\Traj \sim \pi \mid s_0}R(\Traj) = \E_{\Traj \sim \pi \mid s_0} \left[ \sum_{t=0}^{N-1} \gamma^{t}r_t + \gamma^N R_N \right]$$
Слева видно определение V-функции. Справа достаточно разделить мат.ожидание на мат.ожидание по первым $N$ шагам и хвост:
$$V^{\pi}(s_0) = \E_{\Traj_{:N} \sim \pi \mid s_0} \left[ \sum_{t=0}^{N-1} \gamma^{t}r_t + \gamma^N \E_{s_N} \E_{\Traj_{N:} \sim \pi \mid s_N} R_N \right]$$
Осталось выделить справа во втором слагаемом определение V-функции.
\end{proof}
\end{theorem}

\begin{proposition}
$\B^N$ --- оператор с коэффициентом сжатия $\gamma^N$.
\beginproof
\begin{align*}
\rho(\B^N V_1, \B^N V_2) \le \gamma \rho (\B^{N-1} V_1, \B^{N-1} V_2) \le \dots \le \gamma^N \rho (V_1, V_2)   \tagqed
\end{align*}
\end{proposition}

Означает ли это, что метод простой итерации решения $N$-шаговых уравнений сойдётся быстрее? Мы по сути просто <<за один шаг>> делаем $N$ итераций метода простой итерации для решения обычного одношагового уравнения; в этом смысле, мы ничего не выигрываем. В частности, если мы устремим $N$ к бесконечности, то мы получим просто определение V-функции; формально, в правой части будет стоять выражение, вообще не зависящее от поданной на вход оператору $V(s)$, коэффициент сжатия будет ноль, и метод простой итерации как бы сходится тут же за один шаг. Но для проведения этого шага нужно выинтегрировать все траектории --- <<раскрыть дерево полностью>>.

Но теперь у нас есть другой взгляд на Generalized Policy Iteration: мы чередуем одну итерацию решения $N$-шагового уравнения Беллмана с Policy Improvement-ом.

\begin{theorem}
Алгоритм Generalized Policy Iteration \ref{generalizedpolicyiteration} при любом $N$ сходится к оптимальной стратегии и оптимальной оценочной функции.
\begin{proof}[Без доказательства]
\end{proof}
\end{theorem}

\needspace{7\baselineskip}
\begin{wrapfigure}{r}{0.35\textwidth}
\vspace{-0.7cm}
\centering
\includegraphics[width=0.3\textwidth]{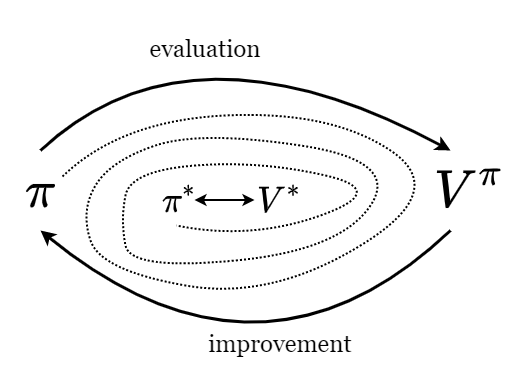}
\vspace{-0.5cm}
\end{wrapfigure}

Интуитивно, такой алгоритм <<стабилизируется>>, если оценочная функция будет удовлетворять уравнению Беллмана для текущей $\pi$ (иначе оператор $\B^N$ изменит значение функции), и если $\pi$ выбирает $\argmax\limits_{a} Q(s, a)$ из неё; а если аппроксимация V-функции удовлетворяет уравнению Беллмана, то она совпадает с $V^\pi$, и значит $Q(s, a) \HM= Q^\pi(s, a)$. То есть, при сходимости стратегия $\pi$ будет выбирать действие жадно по отношению к своей же $Q^\pi$, а мы помним, что это в точности критерий оптимальности.

Все алгоритмы, которые мы будем обсуждать далее, так или иначе подпадают под обобщённую парадигму <<оценивание-улучшение>>. У нас будет два процесса оптимизации: обучение \emph{актёра} (actor), политики $\pi$, и \emph{критика} (critic), оценочной функции (Q или V). Критик обучается оценивать текущую стратегию, текущего актёра: сдвигаться в сторону решения какого-нибудь уравнения, для которого единственной неподвижной точкой является $V^{\pi}$ или $Q^{\pi}$. Актёр же будет учиться при помощи policy improvement-а: вовсе не обязательно делать это жадно, возможно учиться выбирать те действия, где оценка критика <<побольше>>, оптимизируя в каких-то состояниях (в каких - пока открытый вопрос) функционал \eqref{pi_optimization}:
$$\E_{\pi(a \mid s)} Q(s, a) \to \max_{\pi}$$

Причём, возможно, в этом функционале нам не понадобится аппроксимация (модель) Q-функции в явном виде, и тогда мы можем обойтись лишь какими-то оценками $Q^{\pi}$; в таких ситуациях нам достаточно будет на этапе оценивания политики обучать лишь модель $V^{\pi}$ для текущей стратегии. А, например, в эволюционных методах мы обошлись вообще без обучения критика именно потому, что смогли обойтись лишь Монте-Карло оценками будущих наград. Этот самый простой способ решать задачу RL --- погенерировать несколько случайных стратегий и выбрать среди них лучшую --- тоже условно подпадает под эту парадигму: мы считаем Монте-Карло оценки значения $J(\pi)$ для нескольких разных стратегий (evaluation) и выбираем наилучшую стратегию (improvement). Поэтому Policy Improvement, как мы увидим, тоже может выступать в разных формах: например, возможно, как в Value Iteration, у нас будет приближение Q-функции, и мы будем просто всегда полагать, что policy improvement проводится жадно, и текущей стратегией неявно будет $\argmax\limits_a Q(s, a)$.

Но главное, что эти два процесса, оценивание политики (обучение критика) и улучшение (обучение актёра) можно будет проводить стохастической оптимизацией. Достаточно, чтобы лишь \textit{в среднем} модель оценочной функции сдвигалась в сторону $V^{\pi}$ или $Q^{\pi}$, а актёр лишь \textit{в среднем} двигался в сторону $\argmax\limits_a Q(s, a)$. И такой <<рецепт>> алгоритма всегда будет работать: пока оба этих процесса проводятся корректно, итоговый алгоритм запустится на практике. Это в целом фундаментальная идея всего RL. В зависимости от выбора того, как конкретно проводить эти процессы, получатся разные по свойствам алгоритмы, и, в частности, отдельно интересными будут алгоритмы, <<схлопывающие>> схему Generalized Policy Iteration в её предельную форму, в Value Iteration. 

Мы далее начнём строить model-free алгоритмы, взяв наши алгоритмы планирования --- Policy Iteration и Value Iteration, --- и попробовав превратить их в табличные алгоритмы решения задачи.

\section{Табличные алгоритмы}\label{sec:tabularrlsection}

\subsection{Монте-Карло алгоритм}

Value iteration и Policy iteration имели два ограничения: 1) необходимо уметь хранить табличку размером $| \St |$ в памяти и перебирать все состояния и действия за разумное время 2) должна быть известна динамика среды $p(s' \mid s, a)$. Первое полечим нейронками, а сейчас будем лечить второе: в сложных средах проблема даже не столько в том, чтобы приблизить динамику среды, а в том, что интегралы $\E_{s' \sim p(s' \mid s, a)}$ мы не возьмём в силу огромного числа состояний и сможем только оценивать по Монте-Карло. 
Итак, мы хотим придумать табличный model-free RL-алгоритм: мы можем отправить в среду пособирать траектории какую-то стратегию, и дальше должны проводить итерации алгоритма, используя лишь эти сэмплы траекторий. Иначе говоря, для данного $s$ мы можем выбрать $a$ и получить ровно один сэмпл из очередного $p(s' \mid s, a)$, причём на следующем шаге нам придётся проводить сбор сэмплов именно из $s'$. Как в таких условиях <<решать>> уравнения Беллмана --- неясно.

Рассмотрим самый простой способ превратить Policy Iteration в model-free метод. Давайте очередную стратегию $\pi_k$ отправим в среду, сыграем несколько эпизодов, и будем оценивать $Q^{\pi_k}$ по Монте-Карло:
$$Q^{\pi_k}(s, a) \approx \frac{1}{N} \sum_{i=0}^N R(\Traj_i), \quad \Traj_i \sim \pi_k \mid s_0 = s, a_0 = a$$

\needspace{7\baselineskip}
\begin{wrapfigure}{r}{0.35\textwidth}
\vspace{-0.3cm}
\centering
\includegraphics[width=0.3\textwidth]{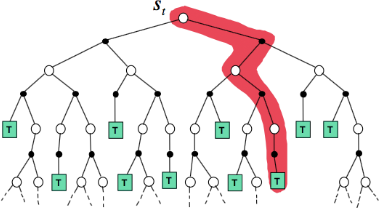}
\vspace{-0.3cm}
\end{wrapfigure}
Теперь, доиграв эпизод до конца, мы для каждой встретившейся пары $s, a$ в полученной траектории можем посчитать reward-to-go и использовать этот сэмпл для обновления нашей аппроксимации Q-функции --- проведения \emph{Монте-Карло бэкапа} (MC-backup). Такое обновление полностью противоположно по свойствам бэкапу динамического программирования: это <<бэкап ширины один>> бесконечной длины --- мы использовали лишь один сэмпл будущего и при этом заглянули в него на бесконечное число шагов вперёд.

\begin{remark}
Формально, из-за петлей сэмплы являются скоррелированными. Если мы крутимся в петле, а потом в какой-то момент эпизода вышли и получили +1, то сэмплы будут выглядеть примерно так: $\gamma^5, \gamma^4, \gamma^3 \dots$. Причина в том, что мы взяли по несколько сэмплов для одной и той же Монте-Карло оценки (для одной и той же пары $s, a$) из одного и того же эпизода (<<\emph{every-visit}>>); для теоретической корректности следует гарантировать независимость сэмплов, например, взяв из каждого эпизода для каждой пары $s, a$ сэмпл только для первого посещения (<<\emph{first-visit}>>).
\end{remark}

\needspace{7\baselineskip}
\begin{wrapfigure}{l}{0.35\textwidth}
\vspace{-0.3cm}
\centering
\includegraphics[width=0.3\textwidth]{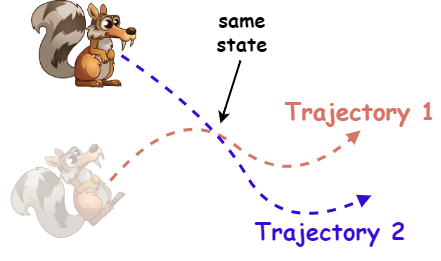}
\vspace{-0.3cm}
\end{wrapfigure}

Монте-Карло алгоритм, на первый взгляд, плох примерно всем. Нужно доигрывать игры до конца, то есть алгоритм неприменим в неэпизодичных средах; Монте-Карло оценки также обладают огромной дисперсией, поскольку мы заменили на сэмплы все мат.ожидания, стоящие в мат.ожидании по траекториям; наконец, мы практически перестали использовать структуру задачи. Если мы получили сэмпл +100 в качестве очередного сэмпла для $Q^{\pi}(s, a)$, то мы <<забыли>>, какую часть из этой +100 мы получили сразу же после действия $a$, а какая была получена в далёком будущем --- не использовали разложение награды за эпизод в сумму наград за шаг. Также мы посеяли <<информацию о соединениях состояниях>>: пусть у нас было две траектории (см. рисунок), имевших пересечение в общем состоянии. Тогда для начал этих траекторий мы всё равно считаем, что собрали лишь один сэмпл reward-to-go, хотя в силу марковости у нас есть намного больше информации.

Ещё одна проблема алгоритма: если для некоторых $s, a \colon \pi(a \mid s) \HM= 0$, то мы ничего не узнали об $Q^\pi(s, a)$. А, как было видно в алгоритмах динамического программирования, мы существенно опираемся в том числе и на значения Q-функции для тех действий, которые $\pi$ никогда не выбирает; только за счёт этого мы умеем проводить policy improvement для детерминированных стратегий.

А ещё в таком Монте-Карло алгоритме встаёт вопрос: когда заканчивать оценивание Q-функции и делать шаг Policy Improvement-а? Точное значение $Q^{\pi_k}$ за конечное время мы Монте-Карло оценкой всё равно не получим, и в какой-то момент improvement проводить придётся, с потерей теоретических гарантий. Возникает вопрос: насколько разумно в таких условиях после очередного обновления стратегии начинать расчёт оценочной функции $Q^{\pi_{k+1}}$ <<с нуля>>? Может, имеет смысл проинициализировать Q-функцию для новой стратегии $\pi_{k+1}$ как-то при помощи текущего приближения $Q(s, a) \HM\approx Q^{\pi_k}(s, a)$? Да, хоть оно и считалось для <<предыдущей>> стратегии и формально содержит сэмплы не из того распределения, но всё-таки этих сэмплов там было аккумулировано много, да и стратегия потенциально поменялась не сильно; и всяко лучше какой-нибудь нулевой инициализации. Возникает желание усреднять сэмплы с приоритетом более свежих, приходящих из <<правильной>> стратегии; а <<неправильные>> сэмплы, из старой стратегии, всё-таки использовать, но с каким-то маленьким весом. Всё это хочется делать как-то онлайн, не храня всю историю Монте-Карло оценок.

\subsection{Экспоненциальное сглаживание}

Рассмотрим такую задачу. Нам приходят сэмплы $x_1, x_2 \dots x_n \sim p(x)$. Хотим по ходу получения сэмплов оценивать мат.ожидание случайной величины $x$. Давайте хранить Монте-Карло оценку, усредняя все имеющиеся сэмплы; для этого достаточно пользоваться следующим рекурсивным соотношением:
$$m_{k} \coloneqq \frac{1}{k} \sum_{i=1}^{k} x_i = \frac{k - 1}{k}m_{k-1} + \frac{1}{k}x_{k}$$
Обозначим за $\alpha_k \coloneqq \frac{1}{k}$. Тогда формулу можно переписать так:
$$m_{k} \coloneqq (1 - \alpha_k)m_{k-1} + \alpha_k x_{k}$$

\begin{definition}
\emph{Экспоненциальным сглаживанием} (exponential smoothing) для последовательности $x_1, x_2, x_3 \dots$ будем называть следующую оценку:
$$m_{k} \coloneqq (1 - \alpha_k) m_{k - 1} + \alpha_k x_k,$$
где $m_0$ --- некоторое начальное приближение, последовательность $\alpha_k \in [0, 1]$ --- гиперпараметр, называемый \emph{learning rate}. 
\end{definition}

Можно ли оценивать мат.ожидание как-то по-другому? В принципе, любая выпуклая комбинация имеющихся сэмплов будет несмещённой оценкой. В частности, если $\alpha_k > \frac{1}{k}$, то мы <<выдаём>> более свежим сэмплам больший вес. Зададимся таким техническим вопросом: при каких других последовательностях $\alpha_k$ формула позволит оценивать среднее?

\begin{definition}
Будем говорить, что learning rate $\alpha_k \in [0, 1]$ удовлетворяет \emph{условиям Роббинса-Монро} (Robbins-Monro conditions), если:
\begin{equation}\label{RobbinsMonro}
\sum_{k \ge 0}^{\infty} \alpha_k = +\infty, \quad \sum_{k \ge 0}^{\infty} \alpha_k^2 < +\infty
\end{equation}
\end{definition}

\begin{theoremBox}[label=th:expsmoothingconvergence]{}
Пусть $x_1, x_2 \dots$ --- независимые случайные величины, $\E x_k = m, \mathbb{D} x_k \le C < +\infty$, где $C$ --- некоторая конечная константа. Пусть $m_0$ --- произвольно, а последовательность чисел $\alpha_k \in [0, 1]$ удовлетворяет условиям Роббинса-Монро \eqref{RobbinsMonro}. Тогда экспоненциальное сглаживание 
\begin{equation}\label{expsmoothingtheoremexpr}
m_k \coloneqq (1 - \alpha_k) m_{k-1} + \alpha_k x_k
\end{equation}
сходится к $m$ с вероятностью 1.
\begin{proof}
Без ограничения общности будем доказывать утверждение для $m = 0$, поскольку для сведения к этому случаю достаточно вычесть $m$ из правой и левой части \eqref{expsmoothingtheoremexpr} и перейти к обозначениям $\hat{m_k} \HM\coloneqq m_k - m, \hat{x_k} \HM\coloneqq x_k \HM- m$.

Итак, пусть $\E x_k \HM= 0$. Будем доказывать, что $v_k \HM\coloneqq \E m_k^2 \xrightarrow{k \to \infty} 0$. Для начала возведём обе стороны уравнения \eqref{expsmoothingtheoremexpr} в квадрат:
$$m_{k}^2 = (1 - \alpha_k)^2 m^2_{k-1} + \alpha^2_k x^2_k + 2\alpha_k(1 - \alpha_k)x_km_{k-1}$$
Возьмём справа и слева мат.ожидание:
$$\E m_{k}^2 = (1 - \alpha_k)^2 \E m^2_{k-1} + \alpha^2_k \E x^2_k + 2\alpha_k(1 - \alpha_k)\E (x_k m_{k-1})$$
Последнее слагаемое зануляется, поскольку в силу независимости $\E (x_k m_{k-1}) \HM= \E x_k \E m_{k-1}$, а $\E x_k$ равно нулю по условию. Используя введённое обозначение, получаем такой результат:
\begin{equation}\label{recurrentdispersion}
v_{k} = (1 - \alpha_k)^2 v_{k-1} + \alpha_k^2 \E x_k^2
\end{equation}

Сейчас мы уже можем доказать, что $v_k \HM\le C$. Действительно, сделаем это по индукции. База: $v_0 \HM= 0 \HM\le C$ по определению. Шаг: пусть $v_{k-1} \le C$, тогда
$$v_{k} = (1 - \alpha_k)^2 v_{k-1} + \alpha_k^2 \E x_k^2 \le (1 - \alpha_k)^2 C + \alpha_k^2 C \le C$$
где последнее неравенство верно при любых $\alpha_k \in [0, 1]$.

Особенность дальнейшего доказательства в том, что последовательность $v_k$ вовсе не обязана быть монотонной. Поэтому применим пару фокусов в стиле матана. Сначала раскроем скобки в рекурсивном выражении \eqref{recurrentdispersion}:
$$v_{k} - v_{k-1} = - 2\alpha_k v_{k-1} + \alpha_k^2 (v_{k-1} + \E x_k^2)$$

Мы получили счётное число равенств, проиндексированных $k$. Просуммируем первые $n$ из них:
\begin{equation}\label{expsmoothdisperrec}
v_{n} - v_0 = - 2 \sum_{k = 0}^{n-1} \alpha_k v_k + \sum_{k = 0}^{n-1} \alpha_k^2 (v_k + \E x_k^2)
\end{equation}

Заметим, что $v_0 \HM= 0$, а $v_{n} \HM\ge 0$ по определению как дисперсия. Значит: 
$$2 \sum_{k = 0}^{n-1} \alpha_k v_k \le \sum_{k = 0}^{n-1} \alpha_k^2 (v_k + \E x_k^2)$$
Применяем ограниченность $v_k$ и $\E x_k^2$:
$$2 \sum_{k = 0}^{n-1} \alpha_k v_k \le \sum_{k = 0}^{n-1} \alpha_k^2 (C + C) = 2C \sum_{k = 0}^{n-1} \alpha_k^2$$
Ряд справа сходится при $n \to +\infty$. Значит, сходится и ряд слева. Коли так, имеет предел правая часть \eqref{expsmoothdisperrec}. Значит, имеет предел и левая.

Было доказано, что последовательность $v_k$ имеет предел. Понятно, что он неотрицателен. Допустим, он положителен и отделён от 0, равен некоторому $b \HM> 0$. Возьмём какое-нибудь небольшое $\eps \HM> 0$, так что $b \HM- \eps \HM> 0$. Тогда, начиная с некоторого номера $i$ все элементы последовательности $v_k \HM> b \HM- \eps$ при $k \HM\ge i$. Получим:
$$\sum_{k = 0}^{n-1} \alpha_k v_k \ge \sum_{k = i}^{n-1} \alpha_k v_k \ge (b - \eps) \sum_{k = i}^{n-1} \alpha_k$$
Правая часть неравенства расходится, поскольку мы просили расходимость ряда из $\alpha_k$; но ранее мы доказали сходимость левой части. Значит, предел равен нулю.
\end{proof}
\end{theoremBox}


\subsection{Стохастическая аппроксимация}

Мы научились, можно считать, решать уравнения такого вида:
$$x = \E_{\eps} f(\eps),$$
где справа стоит мат.ожидание по неизвестному распределению $\eps \sim p(\eps)$ от какой-то функции $f$, которую для данного значения сэмпла $\eps$ мы умеем считать. Это просто стандартная задача оценки среднего, для которой мы даже доказали теоретические гарантии сходимости следующего итеративного алгоритма:
$$x_k \coloneqq (1 - \alpha_k)x_{k-1} + \alpha_k f(\eps), \quad \eps \sim p(\eps)$$

Аналогично у нас есть итеративный алгоритм для решения систем нелинейных уравнений
$$x = f(x),$$
где справа стоит, если угодно, <<хорошая>> функция --- сжатие. Формула обновления в методе простой итерации выглядела вот так:
$$x_k \coloneqq f(x_{k-1})$$

\begin{definition}
\emph{Стохастическая аппроксимация} (Stochastic approximation) --- задача решения уравнения вида
\begin{equation}\label{stochapproxeq}
x = \E_{\eps \sim p(\eps)} f(x, \eps),
\end{equation}
где справа стоит мат.ожидание по неизвестному распределению $p(\eps)$, из которого доступны лишь сэмплы, от функции, которую при данном сэмпле $\eps$ и некотором значении неизвестной переменной $x$ мы умеем считать. 
\end{definition}

Можно ли объединить идеи метода простой итерации и экспоненциального сглаживания (<<перевзвешанной>> Монте-Карло оценки)? Давайте запустим аналогичный итеративный алгоритм: на $k$-ой итерации подставим текущее приближение неизвестной переменной $x_k$ в правую часть $f(x_k, \eps)$ для $\eps \HM\sim p(\eps)$, но не будем <<жёстко>> заменять $x_{k+1}$ на полученное значение, так как оно является лишь несмещённой оценкой правой части; вместо этого сгладим старое значение $x_k$ и полученный новый <<сэмпл>>:
\begin{equation}\label{stochapprox}
x_k = (1 - \alpha_k)x_{k-1} + \alpha_k f(x_{k-1}, \eps), \quad \eps \sim p(\eps)
\end{equation}

Есть хорошая надежда, что, если функция $f$ <<хорошая>>, распределение $p(\eps)$ не сильно страшное (например, имеет конечную дисперсию, как в теореме о сходимости экспоненциального сглаживания), а learning rate $\alpha_k$ удовлетворяют условиям \eqref{RobbinsMonro}, то такая процедура будет в пределе сходиться.

Поймём, почему задача стохастической аппроксимации тесно связана со стохастической оптимизацией. Для этого перепишем формулу \eqref{stochapprox} в альтернативной очень интересной форме:
\begin{equation}\label{stochapproxgrad}
x_k = x_{k-1} + \alpha_k \left( f(x_{k-1}, \eps) - x_{k-1} \right), \quad \eps \sim p(\eps)
\end{equation}
Да это же формула \emph{стохастического градиентного спуска} (stochastic gradient descent, SGD)! Действительно, в нём мы, оптимизируя некоторую функцию $f(x)$, прибавляем к очередному приближению $x_{k-1}$ с некоторым learning rate $\alpha_k$ несмещённую оценку градиента $\nabla f(x_{k-1})$, которую можно обозначить как $\nabla f(x_{k-1}, \eps)$:
$$x_k = x_{k-1} + \alpha_k \nabla f(x_{k-1}, \eps), \quad \eps \sim p(\eps)$$
О стохастической оптимизации можно думать не как об оптимизации, а как о поиске решения уравнения $\nabla f(x) \HM= 0$. Действительно: SGD, как и любая локальная оптимизация, может идти как к локальному оптимуму, так и к седловым точкам, но в них $\E_{\eps} \nabla f(x, \eps) \HM= 0$.

И, глядя на формулу \eqref{stochapproxgrad}, мы понимаем, что, видимо, выражение $f(x_{k-1}, \eps) \HM- x_{k-1}$ есть какой-то <<стохастический градиент>>. Стохастический он потому, что это выражение случайно: мы сэмплируем $\eps \sim p(\eps)$; при этом это несмещённая оценка <<градиента>>, поскольку она обладает следующим свойством: в точке решения, то есть в точке $x^*$, являющейся решением уравнения \eqref{stochapproxeq}, в среднем его значение равно нулю: 
$$\E_{\eps} \left( f(x^*, \eps) - x^* \right) = 0$$
И по аналогии можно предположить, что если стохастический градиентный спуск ищет решение $\E_{\eps} \nabla f(x, \eps) \HM= 0$, то процесс \eqref{stochapproxgrad} ищет решение уравнения $\E_{\eps} f(x, \eps) \HM- x \HM= 0$.

Это наблюдение будет иметь для нас ключевое значение. В model-free режиме как при оценивании стратегии, когда мы решаем уравнение Беллмана
$$Q^{\pi}(s, a) = \E_{s'} \left[ r(s, a) + \gamma \E_{a'} Q^{\pi}(s', a') \right],$$
так и когда мы пытаемся реализовать Value Iteration и напрямую решать уравнения оптимальности Беллмана
$$Q^*(s, a) = \E_{s'} \left[ r(s, a) + \gamma \max_{a'} Q^*(s', a') \right],$$
мы сталкиваемся в точности с задачей стохастической аппроксимации \eqref{stochapproxeq}! Справа стоит мат.ожидание по недоступному нам распределению, содержимое которого мы, тем не менее, при данном сэмпле $s' \HM\sim p(s' \mid s, a)$ и текущем приближении Q-функции, умеем считать. Возникает идея проводить стохастическую аппроксимацию: заменять правые части решаемых уравнений на несмещённые оценки и сглаживать текущее приближение с получаемыми сэмплами.

Отметим интересный факт: уравнение V*V* \eqref{V*V*}, имеющее вид <<максимум от мат.ожидания по неизвестному распределению>>
$$V^*(s) = \max_{a} \left[ r(s, a) + \gamma \E_{s'}V^*(s') \right],$$
под данную форму не попадает. И с такими уравнениями мы ничего сделать не сможем. К счастью, нам и не понадобится.

\subsection{Temporal Difference}


Итак, попробуем применить идею стохастической аппроксимации для решения уравнений Беллмана. На очередном шаге после совершения действия $a$ из состояния $s$ мы получаем значение награды $r \HM\coloneqq r(s, a)$, сэмпл следующего состояния $s'$ и генерируем $a' \HM\sim \pi$, после чего сдвигаем с некоторым learning rate наше текущее приближение $Q(s, a)$ в сторону сэмпла
$$y \coloneqq r + \gamma Q(s', a'),$$
который также будем называть \emph{таргетом}\footnote{в англоязычной литературе встречается слово guess --- <<догадка>>; этот таргет имеет смысл наших собственных предположений о том, какое будущее нас ждёт.} (Bellman target). Получаем следующую формулу обновления:

\begin{equation}\label{TDupdate}
Q_{k+1}(s, a) \leftarrow Q_k(s, a) + \alpha_k \underbrace{\left( \overbrace{r + \gamma Q_k(s', a')}^{\text{\emph{таргет}}} - 
Q_k(s, a) \right)}_\text{\emph{временная разность}} 
\end{equation}

Выражение $r(s, a) + \gamma Q_k(s', a') - Q_k(s, a)$ называется \emph{временной разностью} (temporal difference): это отличие сэмпла, который нам пришёл, от текущей оценки среднего.

\needspace{7\baselineskip}
\begin{wrapfigure}{r}{0.35\textwidth}
\centering
\includegraphics[width=0.3\textwidth]{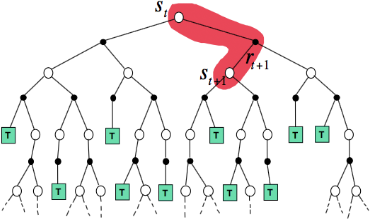}
\end{wrapfigure}

Мы таким образом придумали \emph{TD-backup}: обновление, имеющее как ширину, так и длину один. Мы рассматриваем лишь одну версию будущего (один сэмпл) и заглядываем на один шаг вперёд, приближая всё дальнейшее будущее своей собственной текущей аппроксимацией. Этот ход позволяет нам учиться, не доигрывая эпизоды до конца: мы можем обновить одну ячейку нашей Q-функции сразу же после одного шага в среде, после сбора одного перехода $(s, a, r, s', a')$.

Формула \eqref{TDupdate} отличается от Монте-Карло обновлений Q-функции лишь способом построения таргета: в Монте-Карло мы бы взяли в качестве $y$ reward-to-go, когда здесь же используем одношаговое приближение. Поэтому смотреть на эту формулу также можно через интуицию \emph{бутстрапирования} (bootstrapping)\footnote{бутстрэп --- <<ремешки на ботинках>>, происходит от выражения <<потянуть самого себя за ремешки на ботинках и так перелезть через ограду>>. Русскоязычный аналог --- <<тащить самого себя за волосы из болота>>. Наши таргеты построены на принципе такого бутстрапирования, поскольку мы начинаем <<из воздуха>> делать себе сэмплы из уже имеющихся сэмплов.}. Мы хотим получить сэмплы для оценки нашей текущей стратегии $\pi$, поэтому делаем один шаг в среде и приближаем всю оставшуюся награду нашей текущей аппроксимацией среднего. Такой <<псевдосэмпл>> уже не будет являться корректным сэмплом для $Q^{\pi}(s, a)$, но в некотором смысле является <<более хорошим>>, чем то, что у нас есть сейчас, за счёт раскрытия дерева на один шаг и получения информации об $r, s'$. Такое движение нашей аппроксимации в сторону чего-то хоть чуть-чуть более хорошего и позволяет нам чему-то учиться.

\begin{exampleBox}[label=ex:cafe]{}
Вы сидите в кафе ($s$) и хотите вернуться домой. Вы прикидываете, что в среднем этой займёт $-Q(s, a) = 30$ минут. Вы тратите одну минуту ($-r$) на выход из кафе и обнаруживаете пробку ($s'$). За счёт этой новой информации вы можете дать более точную оценку времени до возвращения до дома: $-Q(s', a') = 40$ минут. Как можно в будущем откорректировать наши прогнозы? Можно засечь время, сколько в итоге заняла поездка --- доиграть эпизод до конца и посчитать Монте-Карло оценку --- а можно уже сделать вывод о том, что случилась некоторая <<временная разность>>, ошибка за один шаг, равная $41 - 30 \HM= 11$ минут. Заменять исходное приближение расстояния от кафе до дома $-Q(s, a)$ на 41 минуту, конечно же, слишком грубо: но временная разность говорит, что 30 минут было заниженной оценкой и её надо чуть-чуть увеличить.
\end{exampleBox}

Обсудим следующий важный технический момент. Какие есть ограничения на переходы $(s, a, r, s', a')$, которые мы можем использовать для обновлений по формуле \eqref{TDupdate}? Пока мы говорили, что мы хотим заниматься оцениванием стратегии $\pi$, и поэтому предлагается, видимо, ею и взаимодействовать со средой, собирая переходы. С точки же зрения стохастической аппроксимации, для корректности обновлений достаточно выполнения всего лишь двух требований:
\begin{itemize}
    \item $s' \sim p(s' \mid s, a)$; если $s'$ приходит из какой-то другой функции переходов, то мы учим Q-функцию для какого-то другого MDP.
    \item $a' \sim \pi(a' \mid s')$; если $a'$ приходит из какой-то другой стратегии, то мы учим Q-функцию для вот этой другой стратегии.
\end{itemize}
Оба этих утверждения вытекают из того, что обновление \eqref{TDupdate} неявно ищет решение уравнения
$$Q(s, a) = \E_{s'} \E_{a'} y$$
как схема стохастической аппроксимации. Поэтому мы будем уделять много внимания тому, из правильных ли распределений приходят данные, которые мы <<скармливаем>> этой формуле обновления. Но и с другой стороны: схема не требует, например, чтобы после ячейки $Q(s, a)$ мы обязательно на следующем шаге обновили ячейку $Q(s', a')$, ровно как и не говорит ничего о требованиях на распределение, из которого приходят пары $s, a$. Как мы увидим позже, это наблюдение означает, что нам вовсе не обязательно обучаться с онлайн-опыта.

\subsection{Q-learning}

Поскольку довести $Q(s, a)$ до точного значения $Q^\pi(s, a)$ с гарантиями мы всё равно не сможем, однажды в алгоритме нам всё равно придётся сделать policy improvement. Что, если мы будем обновлять нашу стратегию $\pi_k(s) \HM\coloneqq \argmax\limits_a Q_k(s, a)$ после каждого шага в среде и каждого обновления Q-функции? Наше приближение Policy Iteration схемы, аналогично ситуации в динамическом программировании, превратится в приближение Value Iteration схемы:
\begin{align*}
y &= r + \gamma Q_k(s', a') = \\
&= r + \gamma Q_k(s', \pi_k(s')) = \\
&= r + \gamma Q_k(s', \argmax_{a'} Q_k(s', a')) = \\
&= r + \gamma \max_{a'} Q_k(s', a')
\end{align*}

Мы получили в точности таргет для решения методом стохастической аппроксимации уравнения оптимальности Беллмана; поэтому для такого случая будем обозначать нашу Q-функцию как аппроксимацию $Q^*$, и тогда наше обновление принимает следующий вид: для перехода $s, a, r, s'$ обновляется только одна ячейка нашего приближения Q-функции:
\begin{equation}\label{Qlearningupdate}
Q_{k+1}(s, a) \coloneqq Q_k(s, a) + \alpha_k \left( r + \gamma \max_{a'} Q_k(s', a') - 
Q_k(s, a) \right) 
\end{equation}

\begin{theoremBox}[label=th:TDconvergencetheorem]{Сходимость Q-learning}
Пусть пространства состояний и действий конечны, $Q_0(s, a)$ --- начальное приближение, на $k$-ой итерации $Q_k(s, a)$ для всех пар $s, a$ строится по правилу
$$Q_{k+1}(s, a) \coloneqq Q_k(s, a) + \alpha_k(s, a) \left( r(s, a) + \gamma \max_{a'} Q_k(s'_k(s, a), a') - Q_k(s, a)\right)$$
где $s'_k(s, a) \sim p(s' \mid s, a)$, а $\alpha_k(s, a) \in [0, 1]$ --- случайные величины, с вероятностью один удовлетворяющие для каждой пары $s, a$ условиям Роббинса-Монро:
\begin{equation}\label{TDconvergence}
\sum_{k \ge 0}^{\infty} \alpha_k(s, a) = +\infty \qquad \sum_{k \ge 0}^{\infty} \alpha_k(s, a)^2 < +\infty
\end{equation}
Тогда $Q_k$ сходится к $Q^*$ с вероятностью один.
\begin{proof}[Доказательство вынесено в приложение \ref{appendix:qlearning}]
\end{proof}
\end{theoremBox}

В частности, если агент некоторым образом взаимодействует со средой и на $k$-ом шаге обновляет своё текущее приближение $Q \HM \approx Q^*$ только для одной пары $s, a$, то можно считать, что $\alpha_k(s, a) \HM \ne 0$ только для неё. При этом ограничения \eqref{TDconvergence} всё ещё могут оказаться выполненными, поэтому эта интересующая нас ситуация есть просто частный случай сформулированной теоремы.

Заметим, что для выполнения условий сходимости необходимо для каждой пары $s, a$ проводить бесконечное число обновлений $Q(s, a)$: в противном случае, в ряду $\sum\limits_{k \ge 1} \alpha_k(s, a)$ будет конечное число ненулевых членов, и мы не попадём под теорему. Значит, наш сбор опыта должен удовлетворять условию \emph{infinite visitation} --- все пары $s, a$ должны гарантированно встречаться бесконечно много раз. По сути теорема говорит, что это требование является достаточным условием на процесс порождения переходов $s, a, r, s'$:

\begin{proposition}\label{infinitepairsisenough}
Пусть сбор опыта проводится так, что пары $s, a$ встречаются бесконечное число раз. Пусть $n(s, a)$ --- счётчик количества выполнений действия $a$ в состоянии $s$ во время взаимодействия агента со средой. Тогда можно положить $\alpha_k(s, a) \HM\coloneqq \frac{1}{n(s, a)}$, чтобы гарантировать сходимость алгоритма к $Q^*$.
\end{proposition}

\subsection{Exploration-exploitation дилемма}

Так какой же стратегией играть со средой, чтобы получать траектории? Коли мы учим $Q^*$, у нас на очередном шаге есть текущее приближение $\pi_k(s) \HM\coloneqq \argmax\limits_a Q_k(s, a)$, и мы можем \emph{использовать} (exploit) эти знания.

\needspace{11\baselineskip}
\begin{theorem}
Собирая траектории при помощи жадной стратегии, есть риск не сойтись к оптимальной.

\begin{wrapfigure}{r}{0.3\textwidth}
\vspace{-0.9cm}
\centering
\includegraphics[width=0.25\textwidth]{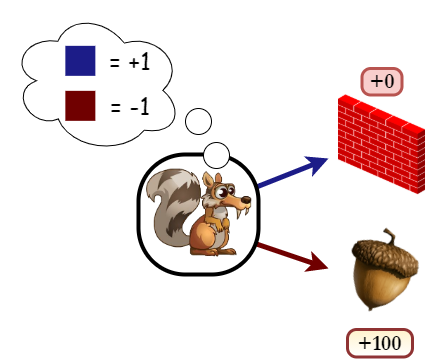}
\vspace{-0.5cm}
\end{wrapfigure}
\beginproof
Приведём простой пример. Есть два действия: получить приз (+100) и тупить в стену (+0), после выполнения любого игра заканчивается. Заинициализировали $Q_0(\text{приз}) \HM= -1$, $Q_0(\text{стена}) \HM= +1$. В первой игре, пользуясь текущей аппроксимацией $\pi_0(s) \HM\coloneqq \argmax\limits_a Q_0(s, a)$, выбираем тупить в стену. Получая 0, сглаживаем 0 и текущее приближение +1, получая новое значение $Q_{k=1}(\text{стена}) \HM\ge 0$. Оно превосходит $Q_{k=1}(\text{приз}) \HM= -1$. Очевидно, и в дальнейших играх агент никогда не выберет приз, и оптимальная стратегия не будет найдена. \qed
\end{theorem}

Действительно, детерминированные стратегии не позволяют получить свойство infinite visitation: многие пары $s, a$ просто приниципиально не встречаются в порождаемых ими траекториях. В частности, из-за неё нельзя ограничиваться рассмотрением только класса детерминированных стратегий, хоть и была доказана теорема \ref{pr:deterministicoptimal} о существовании в нём оптимальной стратегии: мы показали, что для сбора данных --- взаимодействия со средой --- детерминированные стратегии не подходят. Какие подходят?

\begin{theorem}
Любая стратегия, для которой $\forall s, a \colon \mu(a \mid s) > 0$, удовлетворяет условию infinite visitation, то есть с отличной от нуля вероятностью в траектории, порождённой $\pi$, встретится любая пара $s, a$.
\begin{proof}
Для любого $s$ существует набор действий $a_0, a_1 \dots a_N$, которые позволяют с некоторой ненулевой вероятностью добраться до него из $s_0$: $p(s \mid s_0, a_0 \dots a_N) > 0$; если это не так, то ни одна стратегия никогда не попадёт в $s$, и мы без ограничения общности можем считать, что таких <<параллельных вселенных>> в нашей среде не существует. Значит, $\pi$ с некоторой ненулевой вероятностью выберет эту цепочку действий $a_0 \dots a_N$, после чего с ненулевой вероятностью окажется в $s$ и с ненулевой вероятностью выберет заданное $a$.
\end{proof}
\end{theorem}

Если мы воспользуемся любой такой стохастической стратегией, мы попадём под действие теоремы о сходимости \ref{th:TDconvergencetheorem}. Совершая случайные действия, мы рискуем творить ерунду, но занимаемся \emph{исследованием} (exploration) --- поиском новых возможностей в среде. Означает ли это, что нам достаточно взять полностью случайную стратегию, которая равновероятно выбирает из множества действий, отправить её в среду порождать переходики $s, a, r, s'$, обучать на них Q-функцию по формуле \eqref{Qlearningupdate}, и на выходе в пределе мы получим $Q^*$? В пределе --- да.

\begin{example}
Представьте, что вы отправили случайную стратегию играть в Марио. Через экспоненциально много попыток она случайно пройдёт первый уровень и попадёт на второй; для состояний из второго уровня будет проведено первое обновление Q-функции. Через условно бесконечное число таких удач Q-функция для состояний из второго уровня действительно будет выучена...
\end{example}

Иначе говоря, исследование --- корректный, но неэффективный способ генерации данных. Использование --- куда более эффективный метод, позволяющий быстро добираться до <<труднодоступных областей>> в среде --- такими областями обычно являются области с высокой наградой, иначе задача RL скорее всего не является особо интересной --- и быстрее набирать сэмплы для обновлений пока ещё редко обновлённых ячеек $Q(s, a)$. Но это <<некорректный>> способ: детерминированная стратегия может застрять в локальном оптимуме и никогда не увидеть, что в каком-то месте другое действие даёт ещё большую награду.

Обсудим базовые варианты, как можно решать этот exploration-exploitation trade-off в контексте вычисления оптимальной Q-функции методом временных разностей. Нам нужно взять нашу стратегию использования $\pi(s) \HM\coloneqq \argmax\limits_{a} Q(s, a)$ и что-нибудь с ней поделать так, чтобы она стала формально стохастичной.

\begin{definition}
\emph{$\eps$-жадной} ($\eps$-greedy) называется стратегия
\begin{equation}\label{egreedy}
\mu(s) \coloneqq \begin{cases}
a \sim \Uniform(\A) & \text{с вероятностью $\eps$;} \\
\argmax\limits_a Q(s, a) & \text{иначе.}
\end{cases}
\end{equation}
\end{definition}

\begin{definition}
\emph{Больцмановской} с температурой $\tau$ называется стратегия
\begin{equation}\label{boltzman}
\mu(a \mid s) \coloneqq \softmax\limits_a \left( \frac{Q(s, a)}{\tau} \right)
\end{equation}
\end{definition}

Первый вариант никак не учитывает, насколько кажутся хорошими другие действия помимо жадного, и, если решает <<исследовать>>, выбирает среди них случайно. Второй же вариант будет редко выбирать очень плохие действия (это может быть крайне полезным свойством), но чувствителен к масштабу приближения Q-функции, что обычно менее удобно и требует настройки температуры $\tau$. Для Больцмановской стратегии мы увидим интересную интерпретацию в контексте обсуждения Maximum Entropy RL (раздел \ref{SACsection}).

\subsection{Реплей буфер}

Итак, мы теперь можем собрать классический алгоритм табличного RL под названием Q-learning. Это метод временных разностей для вычисления оптимальной Q-функции с $\eps$-жадной стратегией исследования.

\begin{algorithm}[label=alg:qlearning]{Q-learning}
\textbf{Гиперпараметры:} $\alpha$ --- параметр экспоненциального сглаживания, $\eps$ --- параметр исследований

\vspace{0.3cm}
Инициализируем $Q(s, a)$ произвольно для всех $s \in \St, a \in \A$ \\
Наблюдаем $s_0$ \\ 
\textbf{На $k$-ом шаге:}
\begin{enumerate}
    \item с вероятностью $\eps$ играем $a_k \sim \Uniform(\A)$, иначе $a_k = \argmax\limits_{a_k} Q(s_k, a_k)$
    \item наблюдаем $r_k, s_{k+1}$
    \item обновляем $Q(s_k, a_k) \leftarrow Q(s_k, a_k) + \alpha \left( r_k + \gamma \max\limits_{a_{k+1}} Q(s_{k+1}, a_{k+1}) - Q(s_k, a_k) \right)$
\end{enumerate}
\end{algorithm}

\begin{remark}
Естественно, в эпизодичных средах нужно всегда следить за флагом $\done$ и учитывать, что для терминальных состояний оценочные функции равны нулю. Мнемонически можно запомнить, что этот флаг является частью дисконтирования: домножение на $1 \HM- \done$ происходит всюду, где мы домножаем приближение будущей награды на $\gamma$.
\end{remark}

Q-learning является типичным представителем off-policy алгоритмов: нам нигде не требовались сэмплы взаимодействия со средой конкретной стратегии. Наша стратегия сбора данных могла быть, вообще говоря, произвольной. Это крайне существенное свойство, потому что в таких алгоритмах возможно обучение с буфера: допустим, некоторый <<эксперт>> $\pi^{\expert}$ провёл много-много сессий взаимодействия со средой и собрал для нас кучу траекторий. Рассмотрим их как набор переходов. Тогда мы можем, вообще не взаимодействуя больше со средой, провести обучение $Q^*$ с буфера: сэмплируем равномерно переход $(s, a, r, s')$ и делаем обновление ячейки $Q(s, a)$ по формуле \eqref{Qlearningupdate}. Что мы тогда выучим?

\begin{definition}
Для данного буфера --- набора переходов $(s, a, r, s')$ --- будем называть \emph{эмпирическим MDP} (empirical MDP) MDP с тем же пространством состояний, действий и функций награды, где функция переходов задана следующим образом:
$$\hat{p}(s' \mid s, a) \coloneqq \frac{N(s, a, s')}{N(s, a)},$$
где $N(s, a, s')$ --- число троек $s, a, s'$, входящих в буфер, $N(s, a)$ --- число пар $s, a$, входящих в буфер.
\end{definition}

\begin{proposition}\label{prop:qlearningempiricmdp}
При выполнении условий на learning rate, Q-learning, запущенный с фиксированного буфера, выучит $Q^*$ для эмпирического MDP.
\begin{proof}
Именно из эмпирического распределения $\hat{p}(s' \mid s, a)$ приходит $s'$ в формуле обновления (при равномерном сэмплировании переходов из буфера). Следовательно, для такого MDP мы и выучим оптимальную Q-функцию в силу теоремы \ref{th:TDconvergencetheorem}.
\end{proof}
\end{proposition}

Естественно, если буфер достаточно большой, то мы выучим очень близкую $Q^*$ к настоящей. Ещё более интересно, что Q-learning как off-policy алгоритм может обучаться со своего собственного опыта --- со своего же собственного буфера, составленного из порождённых очень разными стратегиями переходов. 

\begin{definition}
\emph{Реплей буфер} (replay buffer, experience replay) --- это память со всеми собранными агентом переходами $(s, a, r, s', \done)$.
\end{definition}

Если взаимодействие со средой продолжается, буфер расширяется, и распределение, из которого приходит $s'$, становится всё больше похожим на настоящее $p(s' \mid s, a)$; на факт сходимости это не влияет.

\begin{algorithm}{Q-learning with experience replay}
\textbf{Гиперпараметры:} $\alpha$ --- параметр экспоненциального сглаживания, $\eps$ --- параметр исследований

\vspace{0.3cm}
Инициализируем $Q(s, a)$ произвольно для всех $s \in \St, a \in \A$ \\
Наблюдаем $s_0$ \\ 
\textbf{На $k$-ом шаге:}
\begin{enumerate}
    \item с вероятностью $\eps$ играем $a_k \sim \Uniform(\A)$, иначе $a_k \coloneqq \argmax\limits_{a_k} Q(s_k, a_k)$
    \item наблюдаем $r_k, s_{k+1}$
    \item кладём $s_k, a_k, r_k, s_{k+1}$ в буфер
    \item сэмплируем случайный переход $s, a, r, s'$ из буфера
    \item обновляем $Q(s, a) \leftarrow Q(s, a) + \alpha \left( r + \gamma \max\limits_{a'} Q(s', a') - Q(s', a') \right)$
\end{enumerate}
\end{algorithm}

Реплей буфер --- ключевое преимущество off-policy алгоритмов: мы сможем с каждого перехода потенциально обучаться бесконечное количество раз. Это развязывает нам руки в плане соотношения числа собираемых данных и количества итераций обновления нашей модели, поскольку здесь мы можем, в общем-то, сами решать, сколько переходов на один шаг обновления мы будем проводить. Проявляется это в том, что в Q-learning, как и в любом другом off-policy алгоритме, есть два независимых этапа: сбор данных (взаимодействие со средой при помощи $\eps$-жадной стратегии и сохранение опыта в памяти --- первые три шага), и непосредственно обучение (сэмплирование перехода из буфера и обновление ячейки --- шаги 4-5). Можно проводить несколько шагов сбора данных на одно обновление, или наоборот: несколько обновлений на один шаг сбора данных, или даже проводить эти процессы независимо параллельно.

\subsection{SARSA}\label{subsec:sarsa}

Мы придумали off-policy алгоритм: мы умеем оценивать нашу текущую стратегию ($\argmax\limits_a Q(s, a)$, неявно сидящий внутри формулы обновления), используя сэмплы другой стратегии. Иными словами, у нас в алгоритме различаются понятия \emph{целевой политики} (target policy) --- стратегии, которую алгоритм выдаст по итогам обучения, она же оцениваемая политика, то есть та политика, для которой мы хотим посчитать оценочную функцию --- и \emph{политики взаимодействия} (behavior policy) --- стратегии взаимодействия со средой, стратегии с подмешанным эксплорейшном. Это различие было для нас принципиально: оптимальны детерминированные стратегии, а взаимодействовать со средой мы готовы лишь стохастичными стратегиями. У этого <<несовпадения>> есть следующий эффект.

\begin{exampleBox}[label=ex:cliffworld]{Cliff World}
Рассмотрим MDP с рисунка с детерминированной функцией переходов, действиями вверх-вниз-вправо-влево и $\gamma < 1$; за попадание в лаву начисляется огромный штраф, а эпизод прерывается. За попадание в целевое состояние агент получает +1, и эпизод также завершается; соответственно, задача агента --- как можно быстрее добраться до цели, не угодив в лаву.

\needspace{7\baselineskip}
\begin{wrapfigure}{r}{0.35\textwidth}
\centering
\includegraphics[width=0.3\textwidth]{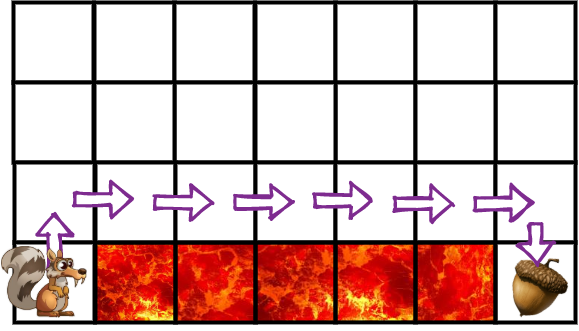}
\end{wrapfigure}
Q-learning, тем не менее, постепенно сойдётся к оптимальной стратегии: кратчайшим маршрутом агент может добраться до терминального состояния с положительной наградой. Однако даже после того, как оптимальная стратегия уже выучилась, Q-learning продолжает прыгать в лаву! Почему? Проходя прямо возле лавы, агент каждый шаг подбрасывает монетку и с вероятностью $\eps$ совершает случайное действие, которое при невезении может отправить его гореть! Если речь не идёт о симуляции, подобное поведение даже во время обучения может быть крайне нежелательно.
\end{exampleBox}

Возможны ситуации, когда небезопасное поведение во время обучения не является проблемой, но для, например, реальных роботов сбивать пешеходов из-за случайных действий --- не самая лучшая идея. Что, если мы попробуем как-то <<учесть>> тот факт, что мы обязаны всегда заниматься исследованиями? То есть внутри оценочной функции должно закладываться, что <<оптимальное>> поведение в будущем невозможно, а возможно только около-оптимальное поведение с подмешиванием исследования. Для примера будем рассматривать $\eps$-жадную стратегию, пока что с константным $\eps$.

Рассмотрим очень похожий на Q-learning алгоритм. Будем использовать пятёрки $s, a, r, s', a'$ (hence the name) прямо из траекторий нашего взаимодействия со средой и апдейтить текущее приближение Q-функции по формуле
\begin{equation}\label{sarsa}
Q(s, a) \leftarrow Q(s, a) + \alpha \left( r(s, a) + \gamma Q(s', a') - Q(s, a) \right)
\end{equation}

Какую Q-функцию такой алгоритм будет учить? Поскольку $a' \sim \mu$, где $\mu$ --- стратегия взаимодействия со средой, то мы стохастически аппроксимируем $Q^{\mu}$ для этой самой $\mu$, сдвигаясь в сторону решения обычного уравнения Беллмана. Формула обновления не будет эквивалентна формуле Q-learning-а \eqref{Qlearningupdate}, где мы сдвигались в сторону $Q^\pi$, где $\pi$ было жадной стратегией по отношению к этой же Q-функции: да, в большинстве случаев (с вероятностью $1 \HM- \eps$) $a'$ будет аргмаксимумом, и обновление будет совпадать с Q-learning, но иногда $a'$ будет оказываться тем самым <<случайным>> действием, случившимся из-за исследований, и наша Q-функция будет сдвигаться в сторону ценности случайных действий. Так в оценочную функцию будет попадать знание о том, что в будущем мы на каждом шаге с вероятностью $\eps$ будем обязаны выбрать случайное действие, и подобное <<дёрганье>> будет мешать нам проходить по краю вдоль обрыва с лавой.

\begin{algorithm}{SARSA}
\textbf{Гиперпараметры:} $\alpha$ --- параметр экспоненциального сглаживания, $\eps$ --- параметр исследований

\vspace{0.3cm}
Инициализируем $Q(s, a)$ произвольно для всех $s \in \St, a \in \A$ \\
Наблюдаем $s_0$, сэмплируем $a_0 \sim \Uniform(\A)$ \\
\textbf{На $k$-ом шаге:}
\begin{enumerate}
    \item наблюдаем $r_k, s_{k+1}$
    \item с вероятностью $\eps$ играем $a_{k+1} \sim \Uniform(\A)$, иначе $a_{k+1} = \argmax\limits_{a_{k+1}} Q(s_{k+1}, a_{k+1})$
    \item обновляем $Q(s_k, a_k) \leftarrow Q(s_k, a_k) + \alpha \left( r_k + \gamma Q(s_{k+1}, a_{k+1}) - Q(s_k, a_k) \right)$
\end{enumerate}
\end{algorithm}

Попробуем понять формальнее, что происходит в таком алгоритме. Можно считать, что он чередует два шага: обновление \eqref{sarsa}, которое учит $Q^{\pi}$ для текущей $\pi$, и, неявно, некий аналог policy improvement-а: замены $\pi$ на $\eps\operatorname{-greedy}(Q)$. Именно при помощи обновлённой стратегии мы будем взаимодействовать со средой на следующем шаге, то есть полагаем $\mu \equiv \pi$ (<<переходим в on-policy режим>>). Является ли такое обновление policy improvement-ом (допустим, для идеально посчитанной $Q^\pi$)? Вообще говоря, нет, но наши стратегии $\pi$, которые мы рассматриваем --- не произвольные. Они все $\eps$-жадные. Введём на минутку такое определение.

\begin{definition}
Будем говорить, что стратегия $\pi$ --- \emph{$\eps$-мягкая} ($\eps$-soft), если $\forall s, a \colon \pi(a \mid s) \HM\ge \frac{\eps}{|\A|}$.
\end{definition}

\begin{proposition}
Если $\pi_1$ --- $\eps$-мягкая, то $\pi_2 \HM\coloneqq \eps\operatorname{-greedy}(Q^{\pi_1})$ не хуже, чем $\pi_1$.
\begin{proof}
Проверим выполнение теоремы \ref{th:policyimprovement}; выглядит немного страшновато, но суть этих выкладок довольно лобовая: если мы переложим всю вероятностную массу в самое <<хорошее>> с точки зрения $Q^{\pi_1}(s, a)$, оставив у остальных, не самых лучших, действий ровно $\frac{\eps}{|\A|}$, то среднее значение увеличится. 
\begin{align*}
V^{\pi_1}(s) &= \{ \text{уравнение VQ \eqref{VQ} } \} = \sum_a \pi_1(a \mid s) Q^{\pi_1}(s, a) = \\
&= \sum_a \left(\pi_1(a \mid s) - \frac{\eps}{|\A|} \right) Q^{\pi_1}(s, a) + \frac{\eps}{|\A|} \sum_a Q^{\pi_1}(s, a) = \\
&= (1 - \eps) \sum_a \frac{\pi_1(a \mid s) - \frac{\eps}{|\A|}}{1 - \eps} Q^{\pi_1}(s, a) + \frac{\eps}{|\A|} \sum_a Q^{\pi_1}(s, a) \le \\
&\le (1 - \eps) \max_a Q^{\pi_1}(s, a) \sum_a \frac{\pi_1(a \mid s) - \frac{\eps}{|\A|}}{1 - \eps} + \frac{\eps}{|\A|} \sum_a Q^{\pi_1}(s, a)
\end{align*}
Заметим, что последний переход был возможен только потому, что $\pi_1(a \mid s) - \frac{\eps}{|\A|} \HM> 0$ по условию, так как $\pi_1$ --- $\eps$-мягкая по условию. Осталось заметить, что 
$$\sum_a \frac{\pi_1(a \mid s) - \frac{\eps}{|\A|}}{1 - \eps} = \frac{\sum\limits_a \pi_1(a \mid s) - \eps}{1 - \eps} = 1,$$
и мы показали, что $V^{\pi_1}(s) \le \E_{\pi_2(a \mid s)}Q^{\pi_1}(s, a)$.
\end{proof}
\end{proposition}

Давайте изменим постановку задачи: скажем, что мы запрещаем к рассмотрению стратегии, <<похожие на детерминированные>>. Наложим ограничение в нашу задачу оптимизации: скажем, что стратегия обязательно должна быть $\eps$-мягкой. В такой задаче будут свои оптимальные оценочные функции.

\begin{definition}
Для данного MDP оптимальными $\eps$-мягкими оценочными функциями назовём:
$$V^*_{\eps\operatorname{-soft}}(s) \coloneqq \max_{\pi \in \eps\operatorname{-soft}} V^{\pi}(s)$$
$$Q^*_{\eps\operatorname{-soft}}(s, a) \coloneqq \max_{\pi \in \eps\operatorname{-soft}} Q^{\pi}(s, a)$$
\end{definition}

\needspace{5\baselineskip}
\begin{wrapfigure}{r}{0.35\textwidth}
\vspace{-0.4cm}
\centering
\includegraphics[width=0.35\textwidth]{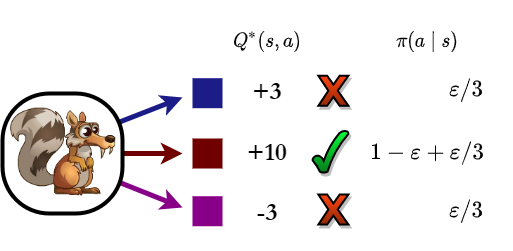}
\vspace{-0.8cm}
\end{wrapfigure}

Как тогда будет выглядеть принцип оптимальности? Раньше нужно было выбирать самое хорошее действие, но теперь так делать нельзя. Проводя аналогичные рассуждения, можно показать, что теперь оптимально выбирать самое хорошее действие с вероятностью $1 \HM- \eps \HM+ \frac{\eps}{|\A|}$, а всем остальным действиям выдавать минимально разрешённую вероятность $\frac{\eps}{|\A|}$. Как это понять? Мы уже поняли, что <<взятие $\eps$-жадной>> стратегии есть местный Policy Improvement. Если мы не можем его провести ни в одном состоянии, а то есть $\pi \equiv \eps\operatorname{-greedy}(Q^{\pi})$, то, видимо, придумать стратегию лучше в принципе невозможно:

\begin{proposition}
Стратегия $\pi$ оптимальна в классе $\eps$-мягких стратегий тогда и только тогда, когда $\forall s, a$ таких, что $a \not\in \Argmax Q^{\pi}(s, a)$ верно $\pi(a \mid s) \HM= \frac{\eps}{|\A|}$.
\begin{proof}[Скетч доказательства]
Притворимся, что в будущем мы сможем выбирать действия как угодно $\eps$-мягко, то есть для данного состояния $s$ для каждого действия $a$ сможем в будущем набирать $Q^*_{\eps\operatorname{-soft}}(s, a)$. Как нужно выбрать действия сейчас? Нужно решить такую задачу оптимизации:
$$
\begin{cases}
\E_{\pi(a \mid s)} Q^*_{\eps\operatorname{-soft}}(s, a) \to \max\limits_{\pi} \\
\int\limits_\A \pi(a \mid s) \diff a = 1; \qquad \forall a \in \A \colon \pi(a \mid s) \ge \frac{\eps}{|\A|}
\end{cases}
$$
Формально решая эту задачу условной оптимизации, получаем доказываемое.
\end{proof}
\end{proposition}

\begin{proposition}
Уравнения оптимальности, соответственно, теперь выглядят так:
$$Q^*_{\eps\operatorname{-soft}}(s, a) = r + \gamma \E_{s'} \left[ (1 - \eps) \max_{a'} Q^*_{\eps\operatorname{-soft}}(s', a') + \frac{\eps}{|\A|} \sum_{a'} Q^*_{\eps\operatorname{-soft}}(s', a') \right]$$
\begin{proof}
Их можно получить, например, взяв обычное уравнение QQ \eqref{QQ} и подставив вид оптимальной стратегии.
\end{proof}
\end{proposition}

Мы теперь понимаем, что наша формула обновления \eqref{sarsa} --- просто метод решения такого уравнения оптимальности: сэмплируя $a'$ из $\eps$-жадной стратегии, мы просто стохастически аппроксимируем по мат.ожиданию из $\eps$-жадной стратегии. Мы могли бы, вообще говоря, взять это мат.ожидание полностью явно, сбив таким образом дисперсию:
\begin{align}\label{expectedsarsa}
Q(s, a) &\leftarrow Q(s, a) + \alpha \left( r(s, a) + \gamma \E_{a' \sim \pi} Q(s', a') - Q(s, a) \right),
\end{align}
где мат.ожидание по $a'$ по определению $\pi$ равно
$$\E_{a' \sim \pi} Q(s', a') = (1 - \eps) \max_{a'} Q(s', a') + \frac{\eps}{|\A|} \sum_{a'} Q(s', a')$$

Такая схема называется Expected SARSA --- <<SARSA, в которой взяли мат.ожидание>>. Мы всё ещё учим Q-функцию текущей политики, но не используем сэмпл из текущей траектории, а вместо этого берём мат.ожидание по действиям из уравнения Беллмана \eqref{QQ} честно. Вообще говоря, в такой схеме мы работаем не с пятёрками $s, a, r, s', a'$, а с четвёрками $s, a, r, s'$ (несмотря на название).

\begin{algorithm}{Expected-SARSA}
\textbf{Гиперпараметры:} $\alpha$ --- параметр экспоненциального сглаживания, $\eps$ --- параметр исследований

\vspace{0.3cm}
Инициализируем $Q(s, a)$ произвольно для всех $s \in \St, a \in \A$ \\
Инициализируем $\pi_0$ произвольно \\
Наблюдаем $s_0$ \\
\textbf{На $k$-ом шаге:}
\begin{enumerate}
    \item играем $a_k \sim \pi_k(a_k \mid s_k)$
    \item наблюдаем $r_k, s_{k+1}$
    \item $\pi_{k+1}$ есть с вероятностью $\eps$ выбрать $a_{k+1} \sim \Uniform(\A)$, иначе $a_{k+1} \coloneqq \argmax\limits_{a_{k+1}} Q(s_{k+1}, a_{k+1})$
    \item обновляем $Q(s_k, a_k) \leftarrow Q(s_k, a_k) + \alpha \left( r_k + \gamma \E_{a_{k+1} \sim \pi_{k+1}}Q(s_{k+1}, a_{k+1}) - Q(s_k, a_k) \right)$
\end{enumerate}
\end{algorithm}

Соответственно, и SARSA, и Expected SARSA будут сходиться уже не к обычной оптимальной оценочной функции, а к $Q^*_{\eps\operatorname{-soft}}(s, a)$ --- оптимальной оценочной функции в семействе $\eps$-мягких стратегий.

\begin{exampleBox}[righthand ratio=0.35, sidebyside, sidebyside align=center, lower separated=false]{Cliff World}
Попробуем запустить в MDP из примера \ref{ex:cliffworld} SARSA. Мы сойдёмся вовсе не к оптимальной стратегии --- а <<к безопасной>> оптимальной стратегии. Внутри нашей оценочной функции сидит вероятность сорваться в лаву при движении вдоль неё, и поэтому кратчайший маршрут перестанет давать нам наибольшую награду просто потому, что стратегии, для которых такой маршрут был оптимален, мы перестали допускать к рассмотрению.

\tcblower
\includegraphics[width=\textwidth]{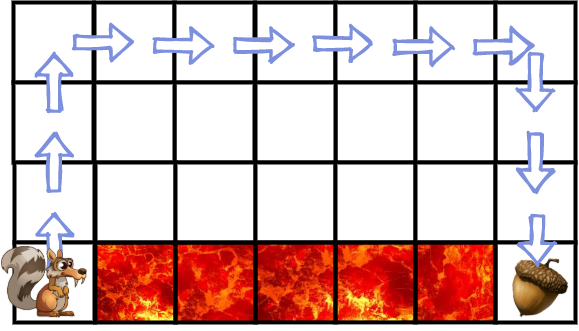}
\end{exampleBox}

Принципиальное отличие схемы SARSA от Q-learning в том, что мы теперь учимся ровно на тех же сэмплах действий $a'$, которые отправляем в среду. Наши behavior и target policy теперь совпадают: для очередного шага алгоритма нужно сделать шаг в среде при помощи текущей $\pi$, и поэтому мы должны учиться онлайн, <<в on-policy режиме>>.

Чтобы понять, к чему это приводит, рассмотрим, что случится, если взять буфер некоторого эксперта $\pi_{\expert}$, генерировать пятёрки $s, a, r, s', a'$ из него и проводить обновления \eqref{sarsa} по ним. Что мы выучим? Применяем наш стандартный ход рассуждений: $a' \HM\sim \pi_{\expert}(a' \mid s')$, и, значит, мы учим $Q^{\pi_{\expert}}$! Если же мы попробуем запустить SARSA с experience replay, то есть обучаться на собственной же истории, то мы вообще творим полную ахинею: на каждом шаге мы движемся в сторону Q-функции для той стратегии $\pi$, которая породила засэмплированный переход (например, если переход был засэмплирован откуда-то из начала обучения --- скорее всего случайной или хуже). Такой алгоритм не просто будет расходиться, но и не будет иметь никакого смысла. Поэтому SARSA нельзя (в таком виде, по крайней мере) запустить с реплей буфера.




\section{Bias-Variance Trade-Off}\label{sec:biasvar}

\subsection{Дилемма смещения-разброса}

Мы обсудили два вида бэкапов, доступных в model-free обучении: Монте-Карло бэкап и Temporal-Difference бэкап. На самом деле, они очень похожи, поскольку делают обновление вида
\begin{equation*}
Q(s, a) \leftarrow Q(s, a) + \alpha \left( y_Q - Q(s, a) \right),
\end{equation*}
и отличаются лишь выбором $y_Q$: Монте-Карло берёт reward-to-go, а TD-backup --- одношаговую бутстрапированную оценку с использованием уже имеющейся аппроксимации Q-функции:
$$y_Q \coloneqq r(s, a) + \gamma Q(s', a')$$

Какой из этих двух вариантов лучше? Мы уже обсуждали недостатки Монте-Карло оценок: высокая дисперсия, необходимость играть до конца эпизодов, игнорирование структуры получаемой награды и потеря информации о соединениях состояний. Но не то, чтобы одношаговые оценки сильно лучше: на самом деле, они обладают полностью противоположными свойствами и проблемами. 

Да, одношаговые оценки аппроксимируют решение одношаговых уравнений Беллмана и приближают алгоритм динамического программирования: поэтому они не теряют информации о том, на каком шаге какой сигнал от среды был получен, и сохраняют информацию о сэмплах $s'$ из функции переходов; в том числе, как мы видели, одношаговые алгоритмы могут использовать реплей буфер, по сути и хранящий собранную выборку таких сэмплов. Взамен в одношаговых алгоритмах возникает \emph{проблема распространения сигнала}.

\begin{example}
Представьте, что за 100 шагов вы можете добраться до сыра (+1). Пока вы не добьётесь успеха, сигнала нет, и ваша аппроксимация Q-функции остаётся всюду нулём. Допустим, вы учитесь с одношаговых оценок с онлайн опыта. После первого успеха +1 распространится лишь в пару $s, a$, непосредственно предшествующей получению сыра; после второго успеха +1 распространится из пары $s, a$ в предыдущую и так далее. Итого, чтобы распространить сигнал на 100 шагов, понадобится сделать 100 обновлений. С другой стороны, если бы использовалась Монте-Карло оценка, после первого же успеха +1 распространился бы во все пары $s, a$ из успешной траектории.
\end{example}

Вместо высокой дисперсии Монте-Карло оценок в одношаговых оценках нас ждёт большое \emph{смещение} (bias): если $y_Q$ оценено через нашу же текущую аппроксимацию через бутстрапирование, то оно не является несмещённой оценкой искомой $Q^{\pi}(s, a)$ и может быть сколь угодно <<неправильным>>. Как мы увидели, гарантии сходимости остаются, но естественно, что методы стохастической аппроксимации из-за смещения будут сходиться сильно дольше экспоненциального сглаживания, которому на вход поступают несмещённые оценки искомой величины. Но дисперсия $y_Q$ в temporal difference обновлениях, например, в алгоритме Q-learning \ref{alg:qlearning}, конечно, сильно меньше дисперсии Монте-Карло оценок: внутри нашей аппроксимации Q-функции уже усреднены все будущие награды, то есть <<взяты>> все интегралы, относящиеся к будущим после первого шага наградам. Итого одношаговая оценка $y_Q$ --- случайная величина только от $s'$, а не от всего хвоста траектории $s', a', s'' \dots$. Выбор между оценками с высокой дисперсией и отсутствием смещения (Монте-Карло) и оценками с низкой дисперсией и большим смещением (одношаговые оценки) --- особая задача в обучении с подкреплением, называемая \emph{bias-variance trade-off}.

\subsection{N-step Temporal Difference}

Какие есть промежуточные варианты между одношаговыми оценками и Монте-Карло оценками? Давайте заглядывать в будущее не на один шаг и не до самого конца, а на $N$ шагов. Итак, пусть у нас есть целый фрагмент траектории:

\begin{definition}
Фрагмент траектории $s, a, r, s', a', r', s'', a'', r'', \dots s^{(N)}, a^{(N)}$, где $s^{(N)}, a^{(N)}$ --- состояние и действие, которое агент встретил через $N$ шагов, будем называть \emph{роллаутом} (rollout) длины $N$.
\end{definition}

\begin{definition}
\emph{$N$-шаговой оценкой} (N-step estimation) для $Q^{\pi}(s, a)$ назовём следующий таргет:
$$y_Q \coloneqq r + \gamma r' + \gamma^2 r'' + \dots + \gamma^{N-1} r^{(N-1)} + \gamma^N Q(s^{(N)}, a^{(N)}),$$
\end{definition}

Такой таргет является стохастической аппроксимацией правой части $N$-шагового уравнения Беллмана \eqref{NstepBellman}, и формула \eqref{generalTD} с таким таргетом позволяет эту систему уравнений решать в model-free режиме. По каким переменным мы заменили интегралы на Монте-Карло приближения в такой оценке? По переменным $s', a', s'', a'' \dots s^{(N)}, a^{(N)}$, которые, пусть и не все присутствуют явно в формуле, но неявно задают то уравнение, которое мы решаем. Соответственно, чтобы выучить $Q^{\pi}(s, a)$, нужно, чтобы состояния приходили из функции переходов, а действия --- из оцениваемой стратегии $\pi$. Другими словами, роллаут, использованный для построения таргета, должен быть порождён оцениваемой стратегией.

Почему гиперпараметр $N$ отвечает за bias-variance trade-off? Понятно, что при $N \HM\to \infty$ оценка переходит в Монте-Карло оценку. С увеличением $N$ всё больше интегралов заменяется на Монте-Карло оценки, и растёт дисперсия; наше же смещённое приближение будущих наград $Q^{\pi}(s^{(N)}, a^{(N)})$, которое может быть сколь угодно неверным, домножается на $\gamma^N$, и во столько же раз сбивается потенциальное смещение; с ростом $N$ оно уходит в ноль. Замешивая в оценку слагаемое $r + \gamma r' + \gamma^2 r'' + \dots + \gamma^{N-1} r^{(N-1)}$ мы теряем информацию о том, что из этого в какой момент было получено, но и начинаем распространять сигнал в $N$ раз <<быстрее>> одношаговой оценки.

Сразу заметим, что при $N \HM> 1$ нам необходимо иметь в $N$-шаговой оценке сэмплы $a' \HM\sim \pi(a \HM\mid s), s'' \HM\sim p(s'' \mid s', a')$. Это означает, что мы не можем получить такую оценку с буфера: действительно, в буфере для данной четвёрки $s, a, r, s'$ не лежит сэмпл $a' \HM\sim \pi(a \HM\mid s)$, ведь в произвольном буфере $a'$ генерируется стратегией сбора данных (старой версией стратегией или <<экспертом>>). Само по себе это, вообще говоря, не беда: мы могли бы взять из буфера четвёрку, прогнать стратегию $\pi$, которую хотим оценить, на $s'$ и сгенерировать себе сэмпл $a'$; но для него мы не сможем получить сэмпл $s''$ из функции переходов! Поэтому обучаться на многошаговые оценки с буфера не выйдет; по крайней мере, без какой-либо коррекции.

Также отметим, что все рассуждения одинаковы применимы как для обучения $Q^\pi$, так и $V^\pi$. Для V-функции общая формула обновления выглядит так:
\begin{equation}\label{generalTD}
V(s) \leftarrow V(s) + \alpha \left( y_V - V(s) \right),
\end{equation}
где $y_V$ --- reward-to-go при использовании Монте-Карло оценки, и $y_V \HM \coloneqq r(s, a) \HM+ \gamma V(s')$ для одношагового метода временных разностей, где $a \HM \sim \pi(a \HM \mid s)$ (сэмплирован из текущей оцениваемой стратегии), $s' \HM \sim p(s' \HM\mid s, a)$. Соответственно, для V-функции обучаться с буфера (без каких-либо коррекций) невозможно даже при $N \HM= 1$, поскольку лежащий в буфере $a$ сэмплирован из стратегии сбора данных, а не оцениваемой $\pi$.

Для простоты и наглядности будем обсуждать обучение $V^\pi$. Также введём следующие обозначения:

\begin{definition}
Введём такое обозначение \emph{$N$-шаговой временной разности} (N-step temporal difference) для пары $s, a$:
\begin{equation}\label{Nstepadvantage}
\Psi_{(N)} (s, a) \coloneqq \sum_{t=0}^{N-1} \gamma^{t} r^{(t)} + \gamma^N V(s^{(N)}) - V(s)
\end{equation}
где $V$ --- текущая аппроксимация V-функции, $s, a, \dots s^{(N)}$ --- роллаут, порождённый $\pi$.
\end{definition}

Ранее мы обсуждали temporal difference, в котором мы сдвигали нашу аппроксимацию на $\Psi_{(1)}(s, a)$:
\begin{align*}
V(s) &\leftarrow V(s) + \alpha \left( r + \gamma V(s') - V(s) \right) = \\ &= V(s) + \alpha \Psi_{(1)}(s, a)
\end{align*}
Теперь же мы можем обобщить наш метод, заменив оценку V-функции на многошаговую оценку:
$$V(s) \leftarrow V(s) + \alpha \Psi_{(N)}(s, a)$$

Но какое $N$ выбрать?

\subsection{Интерпретация через Credit Assingment}

Вопрос, обучаться ли со смещённых оценок, или с тех, которые имеют большую дисперсию, имеет прямое отношение к одной из центральных проблем RL --- credit assingment. На самом деле, это ровно та же самая проблема. 

Рассмотрим проблему credit assingment-а: за какие будущие награды <<в ответе>> то действие, которое было выполнено в некотором состоянии $s$? Как мы обсуждали в разделе \ref{subsection:advantage}, <<идеальное>> решение задачи --- значение $A^\pi(s, a)$, но на практике у нас нет точных значений оценочных функций, а есть лишь аппроксимация, допустим, $V \HM \approx V^\pi$. Положим, мы знаем сэмпл траектории $a, s', r, s'', a'', \dots$ до конца эпизода. 


С одной стороны, мы можем выдавать кредит, полностью опираясь на аппроксимацию:
$$Q^\pi(s, a) = r(s, a) + \gamma \E_{s'} V^\pi(s') \approx r(s, a) + \gamma V(s')$$
\begin{equation}\label{onestepcredit}
A^\pi(s, a) = Q^\pi(s, a) - V^\pi(s) \approx r(s, a) + \gamma V(s') - V(s)
\end{equation}
Проблема в том, что наша аппроксимация может быть сколь угодно неверна, и выдавать полную ерунду. Более <<безопасный>> с этой точки зрения способ --- в приближении Q-функции не опираться на аппроксимацию и использовать reward-to-go:
\begin{equation}\label{infinitestepcredit}
A^\pi(s, a) = Q^\pi(s, a) - V^\pi(s) \approx \sum_{t \ge 0} \gamma^t r^{(t)} - V(s)
\end{equation}
У этого второго способа выдавать кредит есть важное свойство, которого нет у первого: в среднем такой кредит будет больше у тех действий, которые действительно приводят к более высокой награде. То есть, если $a_1, a_2$ таковы, что $Q^\pi(s, a_1) \HM> Q^\pi(s, a_2)$, то при любой аппроксимации $V(s)$ среднее значение кредита для $s, a_1$ будет больше $s, a_2$. Это и есть свойство несмещённости Монте-Карло оценок в контексте проблемы выдачи кредита.

Беда Монте-Карло в том, что в этот кредит закладывается награда не только за выбор $a$ в $s$, но и за будущие решения. То есть: представьте, что через десять шагов агент выбрал действие, которое привело к +100. Эта награда +100 попадёт и в кредит всех предыдущих действий, хотя те не имели к нему отношения.

\begin{example}
Допустим, мы ведём машину, и в состоянии $s$, где аппроксимация V-функции прогнозирует будущую награду ноль, решили выехать на встречку. Среда не сообщает нам никакого сигнала (пока не произошло никакой аварии), но аппроксимация V-функции резко упала; если мы проводим credit assingment одношаговым способом \eqref{onestepcredit}, то мы получаем сильно отрицательный кредит, который сообщает, что это действие было явно плохим.

Дальше агент следующими действиями исправил ситуацию, вернулся на правильную полосу и после решил поехать в магазин тортиков, где оказался юбилейным покупателем и получил бесплатный тортик +10. Если бы мы проводили credit assingment методом Монте-Карло \eqref{infinitestepcredit}, кредит получился бы сильно положительным: $+10 - 0 \HM= +10$: мы положили, что выезд на встречку привёл к тортику.
\end{example}

Видно, что эти два крайних способа выдачи кредита есть в точности <<градиент>> $y_V - V(s)$, по которому учится V-функция в формуле \eqref{generalTD}. Фактически, занимаясь обучением V-функции и используя формулу
\begin{equation*}
V(s) \leftarrow V(s) + \alpha \Psi(s, a),
\end{equation*}
мы выбором функции $\Psi(s, a)$ по-разному проводим credit assingment. 

Таким образом видна новая интерпретации bias-variance trade-off-а: и <<дисперсия>> с этой точки зрения имеет смысл возложения на действие ответственности за те будущие награды, к которым это действие отношения на самом деле не имеет. Одношаговая оценка $\Psi_{(1)}$ говорит: действие влияет только на награду, которую агент получит тут же, и весь остальной сигнал будет учтён в аппроксимации V-функции. Монте-Карло оценка $\Psi_{(\infty)}$ говорит, что действие влияет на все будущие награды. А $N$-шаговая оценка $\Psi_{(N)}$ говорит странную вещь: действие влияет на события, который происходят с агентом в течение следующих $N$ шагов.

Как и в любом trade-off, истина лежит где-то по середине. Однако подбирать на практике <<хорошее>> $N$, чтобы получить оценки с промежуточным смещением и дисперсией, затруднительно. Но что ещё важнее, все $N$-шаговые оценки на самом деле неудачные. Во-первых, они плохи тем, что не используют всю доступную информацию: если мы знаем будущее на $M$ шагов вперёд, то странно не использовать награды за шаг за все эти $M$ шагов, и странно не учесть прогноз нашей аппроксимации оценочной функции $V(s^{(t)})$ для всех доступных $t$. Во-вторых, странно, что в стационарной задаче, где всё инвариантно относительно времени, у нас появляется гиперпараметр, имеющий смысл количества шагов --- времени. Поэтому нас будет интересовать далее альтернативный способ <<интерполировать>> между Монте-Карло и одношаговыми оценками. Мы всё равно оттолкнёмся именно от $N$-шаговых оценок, поскольку понятно, что эти оценки <<корректны>>: они направляют нас к решению уравнений Беллмана, для которых искомая $V^\pi$ является единственной неподвижной точкой.

\needspace{5\baselineskip}
\begin{wrapfigure}{r}{0.35\textwidth}
\centering
\includegraphics[width=0.35\textwidth]{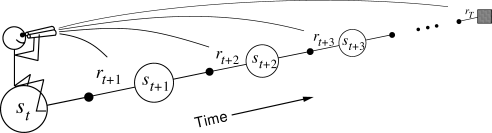}
\vspace{-0.5cm}
\end{wrapfigure}
Мы придумали эти $N$-шаговые оценки, посмотрев на задачу под следующим углом: мы знаем сэмпл будущего (хвост траектории) и хотим выдать кредит <<настоящему>>: самому первому действию. Такой взгляд на оценки называется <<\emph{forward view}>>: мы после выполнения $a$ из $s$ знаем <<вперёд>> своё будущее и можем обновить оценочную функцию для этой пары.

\subsection{Backward View}

Оказывается, мы можем посмотреть на задачу немного по-другому: можно, используя настоящее, обновлять кредит для прошлого. Рассмотрим эту идею, развив пример с кафе \ref{ex:cafe}. 

\begin{example}
Ещё раз сядем в кафе ($s$) и захотим вернуться домой. Текущая аппроксимация даёт $-V(s) \HM= 30$ минут. Делаем один шаг в среде: тратим одну минуту ($-r$) и обнаруживаем пробку ($s'$). Новая оценка времени возвращения даёт: $-V(s') \HM= 40$ минут, соответственно с нами случилась одношаговая временная разность $\Psi_{(1)}(s, a) \HM= 41 - 30 \HM= 11$ минут, которая позволяет нам корректировать $V(s)$.

\needspace{7\baselineskip}
\begin{wrapfigure}{r}{0.3\textwidth}
\vspace{-0.3cm}
\centering
\includegraphics[width=0.3\textwidth]{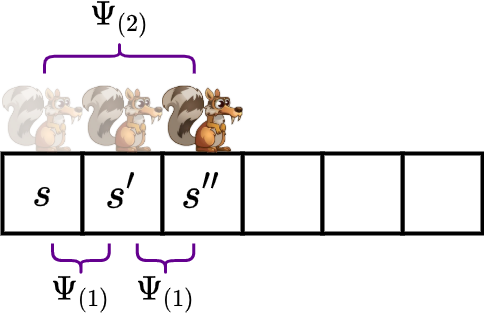}
\vspace{-0.3cm}
\end{wrapfigure}

Давайте сделаем ещё один шаг в среде: тратим одну минуту $-r'$, видим пожар $s''$ и получаем новую оценку $-V(s'') \HM= 60$ минут. Тогда мы можем посчитать как одношаговую временную разность для пары $s', a'$, равную $\Psi_{(1)}(s', a') \HM= 61 \HM- 40 \HM= 21$ минуте, и уточнить свою аппроксимацию V-функции для состояния с пробкой; а можем посчитать и двухшаговую временную разность для кафе: мы потратили две минуты $r + r'$ на два шага и наша нынешнее приближение равно 60 минутам. Итого двухшаговая временная разность равна $\Psi_{(2)}(s, a) \HM= 62 - 30 \HM= 32$ минуты. Forward view говорит следующее: если мы хотим учиться на двухшаговые оценки вместо одношаговых, то нам не следовало на первом шаге использовать 11 минут для обновления $V(s)$, а нужно было дождаться второго шага, узнать двухшаговую ошибку в 32 минуты и воспользоваться ей.

Но понятно, что двухшаговая ошибка это сумма двух одношаговых ошибок! Наши 32 минуты ошибки --- это 11 минут ошибки после выхода из кафе в пробку плюс 21 минута ошибки от выхода из пробки в пожар. Давайте после первого шага используем уже известную часть ошибки в 11 минут, а на втором шаге, если мы готовы обучаться с двухшаговых ошибок, возьмём и добавим недостающие 21 минуту.
\end{example}

Формализуем эту идею. Пусть мы взаимодействуем в среде при помощи стратегии $\pi$, которую хотим оценить; также будем считать learning rate $\alpha$ константным. После совершения первого шага <<из кафе>> мы можем, зная $s, a, r, s'$ сразу же обновить нашу V-функцию:
\begin{equation}\label{twostepsupdatefirstpart}
V(s) \leftarrow V(s) + \alpha \Psi_{(1)}(s, a)    
\end{equation}
Затем мы делаем ещё один шаг в среде, узнавая $a', r', s''$ и временную разность за этот случившийся шаг $\Psi_{(1)}(s', a')$. Тогда мы можем просто ещё в нашу оценку V-функции добавить слагаемое:
\begin{equation}\label{twostepsupdatesecondpart}
V(s) \leftarrow V(s) + \alpha \gamma \Psi_{(1)}(s', a')
\end{equation}
Непосредственной проверкой легко убедиться, что суммарное обновление V-функции получится эквивалентным двухшаговому обновлению:
\begin{proposition}
Последовательное применение обновлений \eqref{twostepsupdatefirstpart} и \eqref{twostepsupdatesecondpart} эквивалентно двухшаговому обновлению
$$V(s) \leftarrow V(s) + \alpha \Psi_{(2)}(s, a)$$
\beginproof
\begin{align*}
V(s) &\leftarrow V(s) + \alpha \left( \Psi_{(1)}(s, a) + \gamma \Psi_{(1)}(s', a') \right) = \\ 
&= V(s) + \alpha \left( r + \gamma V(s') - V(s) + \gamma r' + \gamma^2 V(s'') - \gamma V(s') \right) = \\
&= V(s) + \alpha \left( r + \gamma r' + \gamma^2 V(s'') - V(s) \right) = \\
&= V(s) + \alpha \Psi_{(2)}(s, a)   \tagqed
\end{align*}
\end{proposition}

Понятно, что можно обобщить эту идею с двухшаговых ошибок на $N$-шаговые: действительно, ошибка за $N$ шагов равна сумме одношаговых ошибок за эти шаги.

\begin{theorem}\,
\begin{equation}\label{NstepAdvSimplified}
\Psi_{(N)}(s, a) = \sum_{t = 0}^{N - 1} \gamma^t \Psi_{(1)}(s^{(t)}, a^{(t)})
\end{equation}

\needspace{7\baselineskip}
\begin{wrapfigure}{r}{0.3\textwidth}
\vspace{-0.3cm}
\centering
\includegraphics[width=0.3\textwidth]{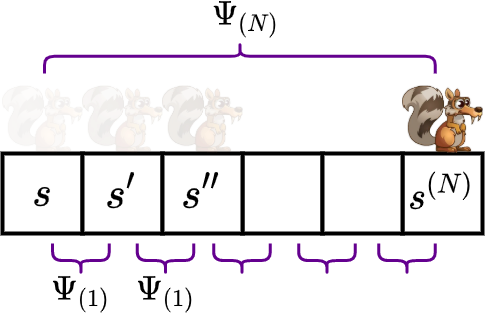}
\vspace{-0.3cm}
\end{wrapfigure}
\beginproof
Докажем по индукции. Для $N = 1$ справа стоит только одно слагаемое, $\Psi_{(1)}(s, a)$, то есть одношаговая оценка.

Пусть утверждение верно для $N$, докажем для $N + 1$. В правой части при увеличении $N$ на единицу появляется одно слагаемое, то есть для доказательства достаточно показать, что
\begin{equation}\label{NtrasformtoNplusone}
\Psi_{(N + 1)}(s, a) = \Psi_{(N)}(s, a) + \gamma^N \Psi_{(1)}(s^{(N)}, a^{(N)})    
\end{equation}
Убедимся в этом, подставив определения:
\begin{align*}
\Psi_{(N + 1)}(s, a) &= \sum_{t = 0}^{N}\gamma^t r^{(t)} + \gamma^{N+1}V(s^{(N+1)}) - V(s) = \\
 &= \sum_{t = 0}^{N - 1}\gamma^t r^{(t)} + \gamma^N V(s^{(N)}) - V(s) + \gamma^N \left( r^{(N)} + \gamma V(s^{(N+1)}) - V(s^{(N)}) \right) = \\
&= \Psi_{(N)}(s, a) + \gamma^N \Psi_{(1)}(s^{(N)}, a^{(N)})   \tagqed
\end{align*}
\end{theorem}

Это наблюдение открывает, что все наши формулы обновления выражаются через одношаговые ошибки --- $\Psi_{(1)}(s, a)$. Это интересный факт, поскольку одношаговая временная разность
$$\Psi_{(1)}(s, a) = r + \gamma V(s') - V(s)$$
очень похожа на reward shaping \eqref{rewardshaping}, где в качестве потенциала выбрана наша текущая аппроксимация $V(s)$. Поэтому эти \emph{дельты}, как их ещё иногда называют, можно интерпретировать как некие <<новые награды>>, центрированные --- которые в среднем должны быть равны нулю, если аппроксимация V-функции точная.

\needspace{7\baselineskip}
\begin{wrapfigure}[11]{l}{0.35\textwidth}
\centering
\includegraphics[width=0.35\textwidth]{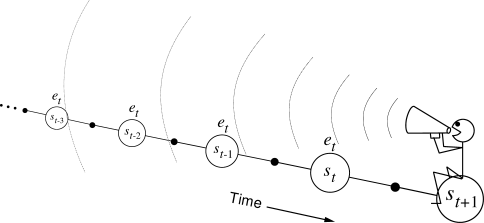}
\end{wrapfigure}
Итого, оказывается, мы можем на каждом шаге добавлять к оценкам V-функции ранее встречавшихся в эпизоде пар $s, a$ только что случившуюся одношаговую ошибку $\Psi_{(1)}(s, a)$ и таким образом получать $N$-шаговые обновления: достаточно пару $s, a$, посещённую $K$ шагов назад, обновить с весом $\gamma^K$. То есть мы начинаем действовать по-другому: зная, что было в прошлом, мы правильным образом обновляем оценочную функцию посещённых состояний из прошлого, используя временную разность за один последний шаг (<<настоящее>>). Такой подход к обновлению оценочной функции называется, соответственно, <<\emph{backward view}>>, и он позволяет взглянуть на обучение оценочных функций под другим углом.

\subsection{Eligibility Trace}

Рассмотрим случай $N = +\infty$, то есть допустим, что мы готовы обновлять V-функцию с reward-to-go. Наше рассуждение, можно сказать, позволяет теперь это делать, не доигрывая эпизоды до конца: мы сразу же в ходе эпизода можем уже <<начинать>> сдвигать V-функцию в правильном направлении. Это позволяет запустить <<как бы Монте-Карло>> алгоритм даже в неэпизодичных средах. Однако, на каждом шаге нам нужно перебирать все встретившиеся ранее в эпизоде состояния, чтобы обновить их, и, если эпизоды длинные (или среда неэпизодична), это хранение истории на самом деле становится избыточным. 

Допустим, мы сделали шаг в среде и получили на этом одном шаге какую-то одношаговую ошибку $\Psi_{(1)}$. Рассмотрим какое-нибудь состояние $s$. С каким весом, помимо learning rate, нужно добавить эту ошибку к нашей текущей аппроксимации? Это состояние в течение прошлого эпизода было, возможно, посещено несколько раз, и за каждое посещение вес увеличивается на $\gamma^K$, где $K$ --- число шагов с момента посещения. Такой счётчик можно сохранить в памяти: заведём вектор $e(s)$ размером с число состояний, проинициализируем его нулём и далее на $t$-ом шаге будем обновлять следующим образом:
\begin{equation}\label{montecarloeligibility}
e_t(s) \coloneqq \begin{cases}
\gamma e_{t - 1}(s) + 1 \quad & \text{если } s = s_t \\
\gamma e_{t - 1}(s) \quad & \text{иначе}
\end{cases}
\end{equation}
После этого на каждом шаге мы будем добавлять текущую одношаговую ошибку, временную разность $\Psi_{(1)}(s_t, a_t) \HM= r_t \HM+ \gamma V^(s_{t+1}) \HM- V(s_t)$, ко \textit{всем} состояниям с коэффициентом $e_t(s)$.

\begin{definition}
Будем называть $e_t(s)$ \emph{следом} (eligibility trace) для состояния $s$ в момент времени $t$ эпизода коэффициент, с которым алгоритм обновляет оценочную функцию $V(s)$ на $t$-ом шаге при помощи текущей одношаговой ошибки $\Psi_{(1)}(s_t, a_t)$:
\begin{equation}\label{eligibilitytraceupdate}
    V(s) \leftarrow V(s) + \alpha e_t(s) \Psi_{(1)}(s_t, a_t)
\end{equation}
\end{definition}

\needspace{7\baselineskip}
\begin{wrapfigure}{r}{0.3\textwidth}
\vspace{-0.8cm}
\centering
\includegraphics[width=0.25\textwidth]{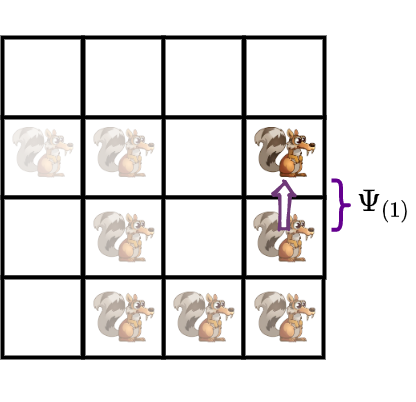}
\vspace{-0.6cm}
\end{wrapfigure}

Допустим, эпизод доигран до конца, и мы в алгоритме используем формулу \eqref{eligibilitytraceupdate} со следом \eqref{montecarloeligibility}. Одношаговое обновление будет превращено в двухшаговое, двухшаговое --- в трёхшаговое и так далее до $N$-шагового, где $N$ --- количество шагов до конца эпизода. Таким образом (если в ходе таких обновлений learning rate и оцениваемая стратегия не меняется) наши обновления в точности соответствуют Монте-Карло.

Важно, что eligibility trace имеет физический смысл <<кредита>>, который выдан решениям, принятым в состоянии $s$: это та степень ответственности, с которой выбранное в этом состоянии действия влияют на события настоящего, на получаемые сейчас награды (временные разности). Действительно, обновление \eqref{eligibilitytraceupdate} говорит следующее: прогноз будущей награды в состоянии $s$ нужно увеличить с некоторым learning rate-ом на получаемую награду (<<$\Psi_{(1)}(s_t, a_t)$>>), домноженную на степень ответственности решений в $s$ за события настоящего (<<$e_t(s)$>>). И пока мы используем Монте-Карло обновления, кредит ведёт себя так: как только мы принимаем в $s$ решение, он увеличивается на единичку и дальше не затухает. Домножение на $\gamma$ можно не интерпретировать как затухание, поскольку это вызвано дисконтированием награды в нашей задаче, <<затуханием>> самой награды со временем. То есть мы рисуем мелом на стене <<я здесь был>>, и выдаём постоянный кредит этому состоянию.

А что, если мы не хотим использовать бесконечношаговые (Монте-Карло) обновления? Мы помним, что это одна крайность, в которой обновления имеют большую дисперсию. Можно броситься в другую крайность: если мы хотим ограничиться лишь, допустим, одношаговыми, то мы можем использовать просто <<другое>> определение следа:
$$e_t(s) \coloneqq \begin{cases}
1 \quad & \text{если } s = s_t \\
0 \quad & \text{иначе}
\end{cases}$$

То есть для одношаговых оценок нам нужно домножать след не на $\gamma$, а на ноль. Тогда вектор $e(s)$ на каждой итерации будет нулевым за исключением одного лишь только что встреченного состояния $s_t$, для которого он будет равен единице, и обновление \ref{eligibilitytraceupdate} будет эквивалентно обычному одношаговому temporal difference. В таком кредите решение, принятое в $s$, влияет только на то, что произойдёт на непосредственно том же шаге; <<мел на стене испаряется мгновенно>>.

\subsection{TD($\lambda$)}

Как можно интерполировать между Монте-Карло обновлениями и одношаговыми с точки зрения backward view? Раз eligibility trace может затухать в $\gamma$ раз, а может в 0, то, вероятно, можно тушить его и любым другим промежуточным способом. Так мы теперь можем придумать другой вид <<промежуточных оценок>> между Монте-Карло и одношаговыми. Пусть на очередном моменте времени след для состояния $s$ затухает сильнее, чем в $\gamma$ раз, но больше, чем в ноль; что тогда произойдёт?

\needspace{15\baselineskip}
\begin{wrapfigure}{r}{0.3\textwidth}
\centering
\includegraphics[width=0.3\textwidth]{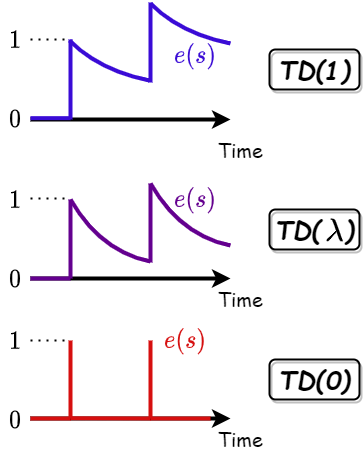}
\vspace{0.2cm}
\end{wrapfigure}

После момента посещения состояния $s$ след для него вырастет на единичку и оценочная функция обновится следующим образом:
$$V(s) \leftarrow V(s) + \alpha \Psi_{(1)}(s, a)$$
Допустим, на следующем шаге мы выбрали $\lambda_1 \HM \in [0, 1]$ --- <<коэффициент затухания>> --- и потушили след не с коэффициентом $\gamma$, а с коэффициентом $\gamma \lambda_1$. Тогда дальше мы добавим новую текущую временную разность $\Psi_{(1)}(s', a')$ с коэффициентом $\lambda \gamma$, получая суммарно следующее обновление:
$$V(s) \leftarrow V(s) + \alpha \left( \Psi_{(1)}(s, a) + \lambda_1 \gamma \Psi_{(1)}(s', a') \right),$$
которое в силу \eqref{NstepAdvSimplified} для $N \HM= 2$ преобразуется в:
$$V(s) \leftarrow V(s) + \alpha \left( (1 - \lambda_1) \Psi_{(1)}(s, a) + \lambda_1 \Psi_{(2)}(s, a) \right)$$

Таким образом, мы \emph{заансамблировали} одношаговое и двухшаговое обновление. Из этого видно, что такая процедура корректна: $V^\pi$ является неподвижной точкой как одношагового, так и двухшагового уравнения Беллмана, и значит неподвижной точкой любой их выпуклой комбинации.

С точки зрения кредита, мы сказали, что решение, принятое в $s$, влияет на то, что случится через 2 шага, но не так сильно, как на то, что случится через 1 шаг: степень ответственности за один шаг затухла в $\lambda_1$ раз. Мы пользуемся здесь следующим прайором из реальной жизни: решения более вероятно влияют на ближайшее будущее, нежели чем на далёкое.

Для понятности проведём ещё один шаг рассуждений. Допустим, мы сделали ещё один шаг в среде и увидели $\Psi_{(1)}(s'', a'')$; потушили eligibility trace $e(s)$, на этот раз, в $\gamma \lambda_2$ раз, где $\lambda_2 \HM \in [0, 1]$. Тогда след стал равен $e(s) \HM= \gamma^2 \lambda_1 \lambda_2$, и мы получим следующее обновление:
$$V(s) \leftarrow V(s) + \alpha \left( (1 - \lambda_1) \Psi_{(1)}(s, a) + \lambda_1 \Psi_{(2)}(s, a) + \gamma^2 \lambda_1 \lambda_2 \Psi_{(1)}(s'', a'') \right)$$
Вспоминая формулу \eqref{NtrasformtoNplusone}, последние два слагаемых преобразуются:
$$\Psi_{(2)}(s, a) + \gamma^2 \lambda_2 \Psi_{(1)}(s'', a'') = (1 - \lambda_2) \Psi_{(2)}(s, a) + \lambda_2 \Psi_{(3)}(s, a)$$
То есть долю $\lambda_2$ двухшаговой ошибки мы превращаем в трёхшаговое, а долю $1 \HM- \lambda_2$ --- нет. Суммарное обновление становится таким ансамблем:
$$V(s) \leftarrow V(s) + \alpha \left( (1 - \lambda_1) \Psi_{(1)}(s, a) + \lambda_1 (1 - \lambda_2) \Psi_{(2)}(s, a) + \lambda_1 \lambda_2 \Psi_{(3)}(s, a) \right)$$

И так далее. Интерпретировать это можно так. Если мы тушим след в $\gamma$ раз, то на очередном шаге обновления мы <<превращаем>> $N$-шаговое обновление в $N \HM+ 1$-шаговое. Если мы тушим след полностью, зануляя его, то мы <<отказываемся превращать>> $N$-шаговое обновление в $N \HM+ 1$-шаговое. Оба варианта корректны, и поэтому мы, выбирая $\lambda_t \in [0, 1]$, решаем заансамблировать их: мы возьмём и долю $\lambda_t$ $N$-шагового обновления <<превратим>> в $N \HM+ 1$-шаговое, а долю $1 \HM- \lambda_t$ трогать не будем и оставим без изменения. Поэтому $\lambda_t$ ещё называют <<коэффициентом смешивания>>.

Мы уже обсуждали, что странно проводить credit assingment нестационарно, то есть чтобы в процедуре была какая-то зависимость от времени, прошедшего с момента посещения состояния, поэтому коэффициент затухания $\lambda \HM\in [0, 1]$ обычно полагают не зависящем от момента времени, каким-то константным гиперпараметром, отвечающим за bias-variance trade-off. Естественно, $\lambda \HM= 1$ соответствует Монте-Карло обновлениям, $\lambda \HM= 0$ --- одношаговым.

\begin{definition}
Будем называть \emph{temporal difference обновлением} с параметром $\lambda \HM\in [0, 1]$ (<<обновление TD($\lambda$)) обновление \eqref{eligibilitytraceupdate} со следом $e_t(s)$, проинициализированным нулём и далее определённым следующим образом:
$$e_t(s) \coloneqq \begin{cases}
\gamma \lambda e_{t - 1}(s) + 1 \quad & \text{если } s = s_t \\
\gamma \lambda e_{t - 1}(s) \quad & \text{иначе}
\end{cases}$$
\end{definition}

Формула следа задаёт алгоритм в парадигме backward view. Естественно, что любая оценка, придуманная в терминах backward view, то есть записанная в терминах следа (в явном виде хранящего <<кредит>> ответственности решений для каждого состояния), переделывается в парадигме forward view (как и наоборот), когда мы, используя сэмплы будущего, строим некоторую оценку Advantage $\Psi(s, a) \HM \approx A^\pi(s, a)$ и просто сдвигаем по нему значение V-функции:
\begin{equation}\label{generalforwardviewupdate}
V(s) \leftarrow V(s) + \alpha \Psi(s, a)
\end{equation}
Какому обновлению в парадигме forward view соответствует обновление TD($\lambda$)?

\begin{theoremBox}[label=th:tdlambda]{Эквивалентные формы TD($\lambda$)}
Обновление TD($\lambda$) эквивалентно \eqref{generalforwardviewupdate} с оценкой
\begin{equation}\label{TDlambda} 
\Psi(s, a) \coloneqq \sum_{t \ge 0} \gamma^t \lambda^t \Psi_{(1)}(s^{(t)}, a^{(t)}) = (1 - \lambda) \sum_{t > 0} \lambda^{t-1} \Psi_{(t)}(s, a)
\end{equation}
\begin{proof}
Составим такую табличку: какое суммарное обновление у нас получается в TD($\lambda$) после $t$ шагов в среде. Справа запишем, с какими весами входят разные $N$-шаговые оценки в получающийся ансамбль.
\begin{center}
\begin{tabular}{|c|c||c|c|c|c|c|}
\hline
\small \textbf{Step} & \small \textbf{Update} & \small \textbf{$\Psi_{(1)}(s, a)$} & \small \textbf{$\Psi_{(2)}(s, a)$} &
\textbf{$\Psi_{(3)}(s, a)$} & 
\textbf{ \dots } & \small \textbf{$\Psi_{(N)}(s, a)$}\\
\hline
1 & $\Psi_{(1)}(s, a)$ &
1 & 0 & 0 & & 0 \\
\hline
2 & $\Psi_{(1)}(s, a) + \gamma \lambda \Psi_{(1)}(s', a')$ &
$1 - \lambda$ & $\lambda$ & 0 & & \\
\hline
\multirow{2}{*}{3} & $\Psi_{(1)}(s, a) + \gamma \lambda \Psi_{(1)}(s', a') +$ &
\multirow{2}{*}{$1 - \lambda$} & \multirow{2}{*}{$(1 - \lambda)\lambda$} & \multirow{2}{*}{$\lambda^2$} & & \multirow{2}{*}{0} \\
& $+ (\gamma \lambda)^2 \Psi_{(1)}(s'', a'')$ &
 & & & &  \\
\hline
$\vdots$ & & & & &  $\ddots$ & \\
\hline
$N$ & $\sum^N_{t \ge 0} (\gamma \lambda)^t \Psi_{(1)}(s^{(t)}, a^{(t)})$ &
$1 - \lambda$ & $(1 - \lambda)\lambda$ & $(1 - \lambda)\lambda^2$ & & $\lambda^N$ \\
\hline
\end{tabular}
\end{center}

Продолжая строить такую табличку, можно по индукции увидеть, что после окончания эпизода суммарное обновление V-функции получится следующим:
$$V(s) \leftarrow V(s) + \alpha \sum_{t \ge 0} (\gamma \lambda)^t \Psi_{(1)}(s^{(t)}, a^{(t)})$$
Это и есть суммарное обновление V-функции по завершении эпизода.
\end{proof}



\end{theoremBox}

Итак, полученная формула обновления имеет две интерпретации. Получается, что подобный ансамбль многошаговых оценок эквивалентен дисконтированной сумме будущих одношаговых ошибок (<<модифицированных наград>> с потенциалом $V(s)$), где коэффициент дисконтирования равен $\gamma \lambda$; исходный коэффициент $\gamma$ отвечает за затухание ценности наград со временем из исходной постановки задачи, а $\lambda$ соответствует затуханию кредита ответственности действия за полученные в будущем награды.

\needspace{5\baselineskip}
\begin{wrapfigure}{r}{0.4\textwidth}
\vspace{-0.4cm}
\centering
\includegraphics[width=0.4\textwidth]{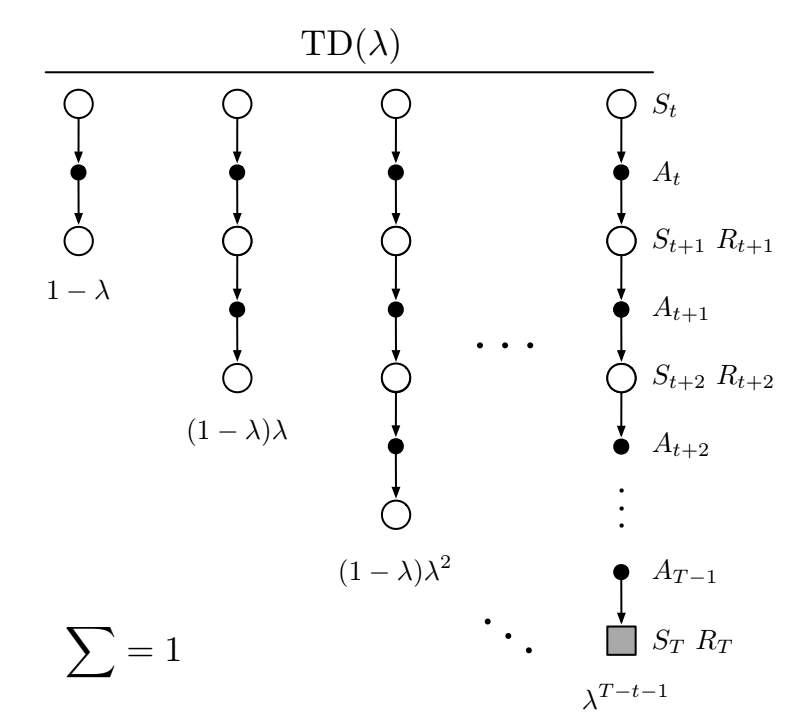}
\vspace{-0.7cm}
\end{wrapfigure}

А с другой стороны можно интерпретировать TD($\lambda$) как ансамбль многошаговых оценок разной длины. Мы взяли одношаговую оценку с весом $\lambda$, двухшаговую с весом $\lambda^2$, $N$-шаговую с весом $\lambda^N$ и так далее. Сумма весов в ансамбле, как водится, должна равняться единице, отсюда в формуле домножение на $1 - \lambda$.

Если через $N$ шагов после оцениваемого состояния $s$ эпизод закончился, то все многошаговые оценки длины больше $N$, понятно, совпадают с $N$-шаговой и равны reward-to-go. Следовательно, в такой ситуации reward-to-go по определению замешивается в оценку с весом $(1 - \lambda)(\lambda^{N-1} + \lambda^{N} + \dots) \HM= \lambda^{N-1}$.

Мы привыкли, что любая формула обновления для нас --- стохастическая аппроксимация решения какого-то уравнения. TD($\lambda$) не исключение. Если $N$-шаговая оценка направляет нас в сторону решения $N$-шагового уравнения Беллмана $V^{\pi} \HM= \B^N V^{\pi}$, то ансамбль из оценок направляет нас в сторону решения ансамбля $N$-шаговых уравнений Беллмана:
\begin{equation}\label{bellman_ensemble}
V^{\pi} = (1 - \lambda)(\B V^{\pi} + \lambda \B^2 V^{\pi} + \lambda^2 \B^3 V^{\pi} + \dots ) = (1 - \lambda) \sum_{N > 0} \lambda^{N-1} \B^N V^{\pi}
\end{equation}

Поскольку $V^{\pi}$ является неподвижной точкой для операторов $\B^N$ для всех $N$, то и для их выпуклой комбинации, <<ансамбля>>, он тоже будет неподвижной точкой. Итого TD($\lambda$) дало нам прикольную идею: мы не могли выбрать одну из многошаговых оценок, и поэтому взяли их все сразу.

Итак, мы получили алгоритм TD($\lambda$) оценивания стратегии, или temporal difference с параметром $\lambda$, который при $\lambda \HM= 1$ эквивалентен Монте-Карло алгоритму (с постоянным обновлением V-функции <<по ходу эпизода>>), а при $\lambda \HM = 0$ эквивалентен одношаговому temporal difference методу, который мы, в частности, применяли в Q-learning и SARSA для оценивания стратегии.

\begin{algorithm}[label=alg:tdlambda]{TD($\lambda$)}
\textbf{Вход:} $\pi$ --- стратегия \\
\textbf{Гиперпараметры:} $\alpha$ --- параметр экспоненциального сглаживания, $\lambda \in [0, 1]$ --- степень затухания следа

\vspace{0.3cm}
Инициализируем $V(s)$ произвольно для всех $s \in \St$ \\
Инициализируем $e(s)$ нулём для всех $s \in \St$ \\
Наблюдаем $s_0$ \\
\textbf{На $k$-ом шаге:}
\begin{enumerate}
    \item выбираем $a_{k} \sim \pi(a_{k} \mid s_{k})$
    \item играем $a_k$ и наблюдаем $r_k, s_{k+1}$
    \item обновляем след $e(s_k) \leftarrow e(s_k) + 1$
    \item считаем одношаговую ошибку $\Psi_{(1)} \coloneqq r_k + \gamma V(s_{k+1}) - V(s_k)$
    \item для всех $s$ обновляем $V(s) \leftarrow V(s) + \alpha e(s) \Psi_{(1)}$
    \item для всех $s$ обновляем $e(s) \leftarrow \gamma \lambda e(s)$
\end{enumerate}

\vspace{0.3cm}
\textbf{Выход:} $V(s)$
\end{algorithm}

Мы обсуждали и выписали этот алгоритм для V-функции для задачи именно оценивания стратегии; естественно, мы могли бы сделать это для Q-функции или добавить policy improvement после, например, каждого шага в среде, получив табличный алгоритм обучения стратегии. Позже в разделе \ref{subsec:retrace} мы рассмотрим формулировку теоремы о сходимости таких алгоритмов для ещё более общей ситуации.

Очевидно, TD($\lambda$) обновление не эквивалентно никаким $N$-шаговым temporal difference формулам: в нём замешана как Монте-Карло оценка, то есть замешана вся дальнейшая награда (весь будущий сигнал), так и приближения V-функции во всех промежуточных состояний (при любом $\lambda \in (0, 1)$). Гиперпараметр $\lambda$ также не имеет смысла времени, и поэтому на практике его легче подбирать.

\begin{remark}
Полезность TD($\lambda$) в том, что $\lambda$ непрерывно и позволяет более гладкую настройку <<длины следа>>. На практике алгоритмы будут чувстительны к выбору $\lambda$ в намного меньшей степени, чем к выбору $N$. При этом даже если $\lambda < 1$, в оценку <<поступает>> информация о далёкой награде, и использование TD($\lambda$) позволит бороться с проблемой распространения сигнала.
\end{remark}

Всюду далее в ситуациях, когда нам понадобится разрешать bias-variance trade-off, мы будем обращаться к формуле forward view TD($\lambda$) обновления \eqref{TDlambda}. Однако как отмечалось ранее, для работы с многошаговыми оценками, а следовательно и с обновлением \eqref{TDlambda} TD($\lambda$), необходимо работать в on-policy режиме, то есть иметь сэмплы взаимодействия со средой именно оцениваемой стратегии $\pi(a \mid s)$, и поэтому возможность разрешать bias-variance trade-off у нас будет только в on-policy алгоритмах.

\subsection{Retrace($\lambda$)}\label{subsec:retrace}

Что же тогда делать в off-policy режиме? Итак, пусть дан роллаут $s, a, r, s', a', r', s'', \dots$, где действия сэмплированы из стратегии $\mu$. Мы хотим при этом провести credit assingment для стратегии $\pi$, то есть понять, как обучать $V \HM\approx V^\pi$ или $Q \HM\approx Q^\pi$. 

Проблема в том, что на самом деле это не всегда даже в принципе возможно. Представим, что $a$ таково, что $\pi(a \HM\mid s) \HM= 0$. Тогда в роллауте хранится информация о событиях, которые произойдут с агентом, использующем стратегию $\pi$, с вероятностью 0. Понятно, что никакой полезной информации о том, как менять аппроксимацию V-функции, мы из таких данных не получим. Поэтому для детерминированных стратегий задача обучения с многошаговых оценок в off-policy режиме может запросто оказаться бессмысленной.

\begin{example}
Допустим, стратегия $\mu$ с вероятностью один в первом же состоянии прыгает в лаву. Мы же хотим посчитать $V^\pi(s)$, будущую награду, для стратегии $\pi$, которая с вероятностью один не прыгает в лаву, а кушает тортики. Поскольку в задаче RL функция переходов $p(s' \HM \mid s, a)$ для разных $a$ может быть произвольно разной, мы ничего не можем сказать о ценности одного действия по информации о другом действии.
\end{example}

Возможность в принципе обучаться off-policy в Q-learning обеспечивалась тем, что, когда мы учим Q-функцию с одношаговых оценок, нам для любых $s, a$, которые можно взять из буфера, достаточно лишь сэмпла $s'$. При этом сэмпл $a' \HM \sim \pi(a' \HM\mid s')$ мы всегда сможем сгенерировать <<онлайн>>, прогнав на взятом из буфера $s'$ оцениваемую стратегию. Обычно в off-policy режиме учат именно Q-функцию, что важно в том числе тем, что по крайней мере одношаговая оценка будет доступна всегда. Если же $a'$ из буфера такого, что $\pi(a' \HM \mid s') \HM= 0$, то любые наши коррекции схлопнутся в одношаговую оценку (или же будут теоретически некорректны), но по крайней мере хоть что-то мы сможем сделать.

Итак, попробуем разрешить bias-variance trade-off-а для Q-функции. Для этого снова вернёмся к идеи следа. Допустим, для взятых из буфера $s, a, r, s'$ мы составили одношаговое обновление:
$$Q(s, a) \leftarrow Q(s, a) + \alpha (r + \gamma Q(s', a'_\pi) - Q(s, a)),$$
где $a'_\pi \HM \sim \pi(\cdot \HM \mid s')$, положив неявно значение следа $e(s, a) \HM= 1$ (след при обучении Q-функции зависит от пары $s, a$). Также лучше, если есть такая возможность, использовать не сэмпл $a'_\pi$, а усреднить по нему (аналогично тому, как было проделано в формуле \eqref{expectedsarsa} Expected SARSA):
\begin{equation}\label{offpolicytwostepupdatefirstpart}
Q(s, a) \leftarrow Q(s, a) + \alpha (r + \gamma \E_{a'_\pi \sim \pi} Q(s', a'_\pi) - Q(s, a))
\end{equation}

Затем возьмём из буфера $a', r', s''$. Какое значение следа мы можем выбрать? Можно $e(s, a) \HM= 0$, оставив одношаговое обновление и <<не превращая>> его в двухшаговое, это одна крайность. А как превратить обновление в двухшаговое целиком? Хотелось бы прибавить
$$\gamma (r' + \gamma \E_{a''_\pi \sim \pi} Q(s'', a''_\pi) - \E_{a'_\pi \sim \pi} Q(s', a'_\pi)),$$
где $s'' \HM\sim p(s'' \mid s', a'_\pi)$. Однако в буфере у нас нет такого сэмпла, а вместо него есть сэмпл $s'' \HM\sim p(s'' \mid s', a')$. Технически, ошибка за второй шаг является случайной величиной от $a'$, который должен приходить из $\pi$, когда у нас есть сэмпл лишь из $\mu$. Поэтому здесь необходимо применить importance sampling коррекцию, после которой ошибка за второй шаг принимает следующий вид\footnote{видно, что в отношении последнего слагаемого возможны вариации, например, ошибку за второй шаг можно оценить как
$$\gamma \frac{\pi(a' \mid s')}{\mu(a' \mid s')} \left( r' + \gamma \E_{a''_\pi \sim \pi} Q(s'', a''_\pi) - \E_{a'_\pi \sim \pi} Q(s', a'_\pi) \right),$$
или даже не корректировать последнее слагаемое importance sampling коррекцией, так как в нём не требуется брать $a'$ обязательно из буфера:
$$\gamma \frac{\pi(a' \mid s')}{\mu(a' \mid s')} \left( r' + \gamma \E_{a''_\pi \sim \pi} Q(s'', a''_\pi)\right) - \gamma \E_{a'_\pi \sim \pi} Q(s', a'_\pi).$$
Далее в формулах предполагается вариант, рассматриваемый в статье про Retrace($\lambda$). 
}:
\begin{equation}\label{offpolicytwostepupdatesecondpart}
Q(s, a) \leftarrow Q(s, a) + \alpha \gamma \frac{\pi(a' \mid s')}{\mu(a' \mid s')} \left( r' + \gamma \E_{a''_\pi \sim \pi} Q(s'', a''_\pi) - Q(s', a') \right),
\end{equation}
где $a', r', s''$ --- взяты из буфера. 

\begin{proposition}
Последовательное применение обновлений \eqref{offpolicytwostepupdatefirstpart} и \eqref{offpolicytwostepupdatesecondpart} является корректным двухшаговым обновлением, то есть эквивалентно
$$Q(s, a) \leftarrow Q(s, a) + \alpha (y(Q) - Q(s, a)),$$
где среднее значение $y(Q)$ по всей заложенной в ней стохастике является правой частью двухшагового уравнения Беллмана для Q-функции:
$$\E y(Q) = \E_{s'}\E_{a' \sim \pi} \E_{s''}\E_{a'' \sim \pi} \left[ r(s, a) + \gamma r(s', a') + \gamma^2 Q(s'', a'') \right]$$
\begin{proof}
После двух рассматриваемых обновлений получается, что <<целевая переменная>>, таргет, равен:
$$y(Q) = r + \gamma \E_{a'_\pi \sim \pi} Q(s', a'_\pi) + \gamma \frac{\pi(a' \mid s')}{\mu(a' \mid s')} \left( r' + \gamma \E_{a''_\pi \sim \pi} Q(s'', a''_\pi) - Q(s', a') \right),$$
где случайными величинами являются $s', a', s''$, и $a' \sim \mu(a' \mid s')$. Возьмём среднее по этим величинам:
$$\E y(Q) = \E_{s'} \left[ r + \gamma \E_{a'_\pi \sim \pi} Q(s', a'_\pi) + \gamma \E_{a' \sim \mu} \frac{\pi(a' \mid s')}{\mu(a' \mid s')} \left( r' + \gamma \E_{s''} \E_{a''_\pi \sim \pi} Q(s'', a''_\pi) - Q(s', a') \right) \right]$$
Раскроем importance sampling коррекцию, воспользовавшись
$$\E_{a' \sim \mu} \frac{\pi(a' \mid s')}{\mu(a' \mid s')} f(a') = \E_{a' \sim \pi} f(a'),$$
получим:
$$\E y(Q) = \E_{s'} \E_{a' \sim \pi} \E_{s''} \E_{a''_\pi \sim \pi} \left[ r + \gamma Q(s', a') + \gamma \left( r' + \gamma  Q(s'', a''_\pi) - Q(s', a') \right) \right]$$
Осталось заметить, что слагаемое $\gamma Q(s', a')$ по аналогии с on-policy режимом сокращается.
\end{proof}
\end{proposition}

Таким образом, одношаговую ошибку за второй шаг для получения полного превращения одношагового обновления в двухшаговое необходимо добавить не с весом $\gamma$, как в on-policy режиме, а с весом $\gamma \frac{\pi(a' \mid s')}{\mu(a' \mid s')}$.

Продолжая рассуждение дальше, можно получить, что одношаговая ошибка через $t$ шагов после выбора оцениваемого действия $a$ в состоянии $s$ в off-policy режиме зависит от случайных величин $s', a', s'', \dots s^{(t + 1)}$, и поэтому importance sampling коррекция для неё будет равна:
$$\prod_{i = 1}^{\hat{t} = t} 
\frac{\pi(a^{(i)} \mid s^{(i)}) p(s^{(i + 1)} \mid s^{(i)}, a^{(i)})}{\mu(a^{(i)} \mid s^{(i)}) p(s^{(i + 1)} \mid s^{(i)}, a^{(i)})} = \prod_{i = 1}^{i = t} \frac{\pi(a^{(i)} \mid s^{(i)}) }{\mu(a^{(i)} \mid s^{(i)})}$$
Здесь и далее считается, что при $t \HM= 0$ подобные произведения равны единице.

Получается, что для того, чтобы строить оценку максимальной длины, нужно на $t$-ом шаге домножать след на
$\gamma \frac{\pi(a^{(t)} \mid s^{(t)})}{\mu(a^{(t)} \mid s^{(t)})}$. Итоговую формулу обновления часто записывают в следующем виде:

\begin{equation}\label{importancesamplingoffpolicycredit}
Q(s, a) \leftarrow Q(s, a) + \alpha 
\sum_{t \ge 0} \gamma^t  \left( \prod_{i = 1}^{i = t} \frac{\pi(a^{(i)} \mid s^{(i)}) }{\mu(a^{(i)} \mid s^{(i)})} \right) \Psi_{(1)}(s^{(t)}, a^{(t)}),
\end{equation}
где
\begin{equation}\label{onestepoffpolicyQdelta}
\Psi_{(1)}(s^{(t)}, a^{(t)}) = r^{(t)} + \gamma \E_{a^{(t+1)}_\pi \sim \pi} Q(s^{(t+1)}, a^{(t+1)}_\pi) - Q(s^{(t)}, a^{(t)})    
\end{equation}

Мы получили <<off-policy>> Монте-Карло оценку в терминах следа. Теперь по аналогии с TD($\lambda$) проинтерполируем между одношаговыми (где след зануляется после первого шага) и бесконечношаговыми обновлениями (где след есть importance sampling дробь): оказывается, оценка будет корректна при любом промежуточном значении следа. Чтобы записать это формально, перепишем формулу \eqref{importancesamplingoffpolicycredit} в следующем виде:
\begin{equation}\label{offpolicyestimators}
Q(s, a) \leftarrow Q(s, a) + \alpha 
\sum_{t \ge 0} \gamma^t  \left( \prod_{i = 1}^{i = t}  c_{i} \right) \Psi_{(1)}(s^{(t)}, a^{(t)}),
\end{equation}
где $c_i$ --- коэффициенты затухания следа: в on-policy они могли быть в диапазоне $[0, 1]$ (и мы выбирали его равным гиперпараметру $\lambda$), а здесь, в off-policy режиме, он может быть в диапазоне
$$c_i \in \left[ 0, \frac{\pi(a^{(i)} \mid s^{(i)}) }{\mu(a^{(i)} \mid s^{(i)})} \right]$$

То, что при любом выборе способа затухания следа табличные алгоритмы, использующие оценку \eqref{offpolicyestimators}, будут сходиться --- один из ключевых и самых общих результатов табличного RL. Приведём несколько нестрогую формулировку этой теоремы:

\begin{theorem}[Retrace]
Пусть число состояний и действий конечно, таблица $Q_0(s, a)$ проинициализирована произвольно. Пусть на $k$-ом шаге алгоритма для каждой пары $s, a$ ячейка таблицы обновляется по формуле
$$Q_{k+1}(s, a) \leftarrow Q_k(s, a) + \alpha_k(s, a)
\sum_{t \ge 0} \gamma^t \left( \prod_{i = 1}^{i = t}  c_{i} \right) \Psi_{(1)}(s^{(t)}, a^{(t)}),$$
$$\Psi_{(1)}(s^{(t)}, a^{(t)}) = r^{(t)} + \gamma \E_{a^{(t+1)}_\pi \sim \pi_k} Q_k(s^{(t+1)}, a^{(t+1)}_\pi) - Q(s^{(t)}, a^{(t)})$$
где $\Traj \HM\sim \mu_k \HM\mid s_0 \HM= s, a_0 \HM= a$ сгенерирована произвольной стратегией сбора данных $\mu_k$ (причём не обязательно стационарной), learning rate $\alpha_k(s, a)$ --- случайно, $\pi_k$ произвольно, и коэффициенты следа --- любые в диапазоне
$$c_i \in \left[ 0, \frac{\pi_k(a^{(i)} \mid s^{(i)}) }{\mu_k(a^{(i)} \mid s^{(i)})} \right].$$
Тогда, если с вероятностью 1 learning rate удовлетворяет условиям Роббинса-Монро \eqref{RobbinsMonro}, а стратегия $\pi_k$ c вероятностью 1 становится жадной по отношению к $Q_k$ в пределе $k \HM\to \infty$, то при некоторых технических ограничениях с вероятностью 1 $Q_k$ сходится к оптимальной Q-функции $Q^*$, а $\pi_k$, соответственно, к оптимальной стратегии.
\begin{proof}[Без доказательства; интересующиеся могут обратиться к оригинальной \href{https://arxiv.org/abs/1606.02647}{статье по Retrace}]
\end{proof}
\end{theorem}

Пользуясь этой теоремой, мы можем в полной аналогии с TD($\lambda$) выбрать гиперпараметр $\lambda \in [0, 1]$, и считать след по формуле
$$
    c_i = \lambda \frac{\pi(a^{(i)} \mid s^{(i)}) }{\mu(a^{(i)} \mid s^{(i)})}
$$
В частности, в on-policy режиме $\pi \HM \equiv \mu$ мы получим коэффициент затухания следа, равный просто $\lambda$. Это очень удобно, но на практике неприменимо.

Такая оценка страдает сразу от двух проблем. Первая проблема --- типичная: \emph{затухающий след} (vanishing trace), когда $\mu(a^{(t)} \mid s^{(t)}) \HM\gg \pi(a^{(t)} \mid s^{(t)})$ для какого-то $t$, и соответствующий множитель близок к нулю. Такое случится, если, например, $\mu$ выбирает какие-то действия, которые $\pi$ выбирает редко, что типично. В предельном случае для детерминированных стратегий может быть такое, что числитель коэффициента равен нулю ($\pi$ не выбирает такого действия никогда), и коррекция скажет, что вся информация, начиная с этого шага, полностью неактуальна. Эта проблема неизбежна.

Вторая проблема --- \emph{взрывающийся след} (exploding trace): $\mu(a^{(t)} \mid s^{(t)}) \HM\ll \pi(a^{(t)} \mid s^{(t)})$. Такое может случиться, если в роллауте попалось редкое для $\mu$ действие, которое тем не менее часто выполняется оцениваемой стратегией $\pi$. С одной стороны, кажется, что как раз такие роллауты наиболее ценны для обучения $Q^\pi$, ведь в них описывается шаг взаимодействия со средой, который для $\pi$ как раз достаточно вероятен. Но на практике взрывающийся importance sampling коэффициент --- источник большой дисперсии рассматриваемой оценки.

\begin{center}
\begin{tabular}{|c|c|c|}
\hline
\textbf{Название} & \textbf{Коффициенты $c_i$} & \textbf{Проблема}\\
\hline
TD($\lambda$) & $\lambda$ &
только on-policy режим \\
\hline
Одношаговые & 0 &
сильное смещение \\
\hline
Importance Sampling & $\lambda \frac{\pi(a^{(i)} \mid s^{(i)}) }{\mu(a^{(i)} \mid s^{(i)})}$ &
легко взрываются \\
\hline
\end{tabular}
\end{center}

Идея борьбы со взрывающимся коэффициентом, предложенной в оценке Retrace($\lambda$), заключается в следующем: если importance sampling коррекция для $t$-го шага взорвалась, давайте воспользуемся тем, что мы теоретически обоснованно можем выбрать любой коэффициент меньше (<<быстрее>> потушить след), и выберем единицу:
\begin{equation}\label{retracecoeff}
c_i \coloneqq \lambda \min \left( 1, \frac{\pi(a^{(i)} \mid s^{(i)}) }{\mu(a^{(i)} \mid s^{(i)})} \right) 
\end{equation}

Коррекция с такими коэффициентами корректна, стабильна, но, конечно, никак не помогает с затуханием следа. Важно помнить, что если $\pi$ и $\mu$ сильно отличаются, то велика вероятность, что коэффициенты затухания будут очень близки к нулю, и мы получим что-то, парктически всегда похожее на одношаговое обновление. С этим мы фундаментально не можем ничего сделать, хотя из-за этого итоговая формула обновления получается сильно смещённой. По этой причине смысла выбирать $\lambda \HM< 1$ в off-policy режиме обычно мало, и его почти всегда полагают равным единицей.

Также обсудим, что говорит формула Retrace в ситуации, когда оцениваемая политика $\pi$ детерминирована. Если на шаге $t$ политика сбора данных $\mu$ выбрала ровно то действие, которое выбирает политика $\pi$, то коэффициент $c_i \HM= \lambda$, то есть ситуация совпадает с on-policy режимом (действительно, в дискретных пространствах действий в числителе единица, а в знаменателе что-то меньшее единицы, когда в непрерывных пространствах действий числитель технически равен бесконечности, поэтому мы обрежем дробь до единицы). В противном же случае дробь $\frac{\pi(a^{(i)} \mid s^{(i)}) }{\mu(a^{(i)} \mid s^{(i)})}$ имеет ноль в числителе, и след занулится. Таким образом, пока в буфере в цепочке $s', a', s'', a'', \cdots$ записанные действия совпадают с теми, которые выбирает детерминированная $\pi$, мы <<пользуемся TD($\lambda$)>>, но вынуждены оборвать след, как только очередное действие разойдётся. Есть некоторая польза в том, чтобы в алгоритме $\pi$ и $\mu$ были стохастичны: тогда след по крайней мере полностью никогда не затухнет.

Мы дальше в off-policy будем обсуждать в основном одношаговые оценки, в том числе для простоты. Из одношаговости будут вытекать все ключевые недостатки таких алгоритмов, связанные со смещённостью подобных оценок и невозможностью полноценно разрешать bias-variance trade-off. Во всех этих алгоритмах с этой проблемой можно будет частично побороться при помощи идей Retrace($\lambda$).

\chapter{Value-based подход}\label{valuebasedchapter}

В данной главе будет рассмотрен второй, Value-based подход к решению задачи, в котором алгоритм ищет не саму стратегию, а оптимальную Q-функцию. Для этого табличный алгоритм Value Iteration будет обобщён на более сложные пространства состояний; требование конечности пространства действий $|\A|$ останется ограничением подхода.

\section{Deep Q-learning}

\subsection{Q-сетка}

В сложных средах пространство состояний может быть непрерывно или конечно, но велико (например, пространство всех экранов видеоигры). В таких средах моделировать функции от состояний, будь то стратегии или оценочные функции, мы можем только приближённо при помощи параметрических семейств.

\needspace{7\baselineskip}
\begin{wrapfigure}{r}{0.35\textwidth}
\vspace{-0.3cm}
\centering
\includegraphics[width=0.3\textwidth]{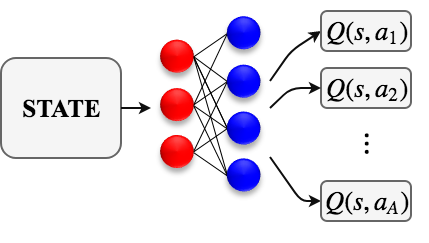}
\vspace{-0.3cm}
\end{wrapfigure}

Попробуем промоделировать в сложных средах алгоритм \ref{alg:qlearning} Q-learning. Для этого будем приближать оптимальную Q-функцию $Q^*(s, a)$ при помощи нейронной сети $Q_{\theta}(s, a)$ с параметрами $\theta$. Заметим, что для дискретных пространств действий сетка может как принимать действия на входе, так и принимать на вход только состояние $s$, а выдавать $|\A|$ чисел $Q_{\theta}(s, a_1) \dots Q_{\theta}(s, a_{|\A|})$. В последнем случае мы можем за константу находить жадное действие $\pi(s) \HM= \argmax\limits_a Q_{\theta}(s, a)$.

В случае, если пространство действий непрерывно, выдать по числу для каждого варианта уже не получится. При этом, если непрерывное действие подаётся на вход вместе с состоянием, то оптимизировать по нему для поиска максимума или аргмаксимума придётся при помощи серии прямых и обратных проходов (для дискретного пространства --- за $|\A|$ прямых проходов), что вычислительно ни в какие ворота. Поэтому такой вариант на практике не встречается, а алгоритм пригоден в таком виде только для дискретных пространств состояний (позже мы пофиксим это при обсуждении алгоритма DDPG в главе \ref{DDPGsection}).

\begin{remark}
Использование нейросеток позволяет обучаться для сред, в которых состояния $s$ заданы, например, пиксельным представлением экранов видеоигр. Стандартным вариантом архитектуры является несколько (не очень много) свёрточных слоёв, обычно без использования макспулингов (важно не убить информацию о расположении распознанных объектов на экране). Использование батч-нормализаций и дроп-аутов сопряжено с вопросами о том, нужно ли их включать-выключать на этапах генерации таргетов (который, как мы увидим позже, тоже распадается на два этапа), сборе опыта с возможностью исследования и так далее. Чаще их не используют, чем используют, так как можно нарваться на неожиданные эффекты. Важно помнить, что все эти блоки были придуманы для решения немного других задач, и стоит осторожно переносить их в контекст обучения с подкреплением.
\end{remark}

\subsection{Переход к параметрической Q-функции}\label{toregression}

Как обучать параметры $\theta$ нейронной сети так, чтобы $Q_{\theta}(s, a) \HM \approx Q^*(s, a)$? Вообще говоря, мы помним, что мы хотим решать уравнения оптимальности Беллмана \eqref{Q*Q*}, и можно было бы, например, оптимизировать невязку:
$$\left( Q_{\theta}(s, a) - r(s, a) - \gamma \E_{s'} \max_{a'} Q_{\theta}(s', a') \right)^2 \to \min_{\theta}$$
однако мат.ожидание $\E_{s'}$ в формуле берётся по неизвестному нам распределению (а даже если известному, то почти наверняка в сложных средах аналитически ничего не возьмётся) и никак не выносится (несмещённые оценки градиента нас бы устроили).

В Q-learning-е мы смогли с сохранением теоретических гарантий побороться с этим при помощи стохастической аппроксимации, получив формулу \eqref{Qlearningupdate}. Хочется сделать какой-то аналогичный трюк:
$$Q_{\theta}(s, a) \leftarrow Q_{\theta}(s, a) + \alpha\left( r(s, a) + \gamma \max_{a'} Q_{\theta}(s', a') - Q_{\theta}(s, a) \right)$$

Мы уже отмечали ключевое наблюдение о том, что формула стохастической аппроксимации очень напоминает градиентный спуск, а $\alpha$ играет роль learning rate. Да даже условия сходимости \ref{TDconvergence}, собственно, те же, что в стохастическом градиентном спуске\footnote{это не случайность: корни теории одни и те же.}! И, действительно, табличный Q-learning является градиентным спуском для решения некоторой задачи регрессии.

Поскольку это очень принципиальный момент, остановимся в этом месте подробнее. Пусть у нас есть текущая версия $Q_{\theta_k}(s, a)$, и мы хотим проделать шаг метода простой итерации для решения уравнения Q*Q* \eqref{Q*Q*}. Зададим следующую задачу регрессии:
\begin{itemize}
\item входом является пара $s, a$
\item искомым (!) значением на паре $s, a$ --- правая часть уравнения оптимальности Беллмана \eqref{Q*Q*}, т.е.
$$f(s, a) \coloneqq r(s, a) + \gamma \E_{s'} \max_{a'} Q_{\theta_k}(s', a') $$
\item наблюдаемым (<<зашумлённым>>) значением целевой переменной или \emph{таргетом} (target)
\begin{equation}\label{guess}
y(s, a) \coloneqq r(s, a) + \gamma \max_{a'} Q_{\theta_k}(s', a')
\end{equation}
где $s' \sim p(s' \mid s, a)$
\item функцией потерь MSE:
$\Loss(y, \hat{y}) = \frac{1}{2}(y - \hat{y})^2$
\end{itemize}

Заметим, что, как и в классической постановке задачи машинного обучения, значение целевой переменной --- это её <<зашумлённое>> значение: по входу $s, a$ генерируется $s'$, затем от $s'$ считается детерминированная функция, и результат $y$ является наблюдаемым значением. Мы можем для данной пары $s, a$ засэмплировать себе таргет, взяв переход $(s, a, r, s')$ и воспользовавшись сэмплом $s'$, но существенно, что такой таргет --- случайная функция от входа. Принципиально: согласно уравнениям Беллмана, мы хотим выучить мат.ожидание такого таргета, а значит, нам нужна квадратичная функция потерь.

\begin{theorem}
Пусть $Q_{\theta_{k+1}}(s, a)$ --- достаточно ёмкая модель (model with enough capacity), выборка неограниченно большая, а оптимизатор идеальный. Тогда решением вышеописанной задачи регрессии будет шаг метода простой итерации для поиска оптимальной Q-функции:
$$Q_{\theta_{k+1}}(s, a) = r(s, a) + \gamma \E_{s'} \max_{a'} Q_{\theta_k}(s', a')$$
\beginproof
Под <<достаточно ёмкой моделью>> подразумевается, что модель может имитировать любую функцию, например для каждой пары $s, a$ из выборки выдавать любое значение. Найдём оптимальное значение для пары $s, a$: для неё $Q_{\theta_{k+1}}(s, a)$ выдаёт некоторое число, которое должно минимизировать MSE:
$$\frac{1}{2}\E_{s'} \left( y(s, a) - Q_{\theta_{k+1}}(s, a) \right)^2 \to \min_{\theta_{k+1}}$$
Оптимальным константным решением регрессии с MSE как раз является среднее, получаем
\begin{align*}
    Q_{\theta_{k+1}}(s, a) = \E_{s'} y(s, a) = r(s, a) + \gamma \E_{s'} \max_{a'} Q_{\theta_k}(s', a') \tagqed
\end{align*}
\end{theorem}

Другими словами, когда мы решаем задачу регрессии с целевой переменной $y(s, a)$ по формуле \eqref{guess}, мы в среднем сдвигаем нашу аппроксимацию в сторону правой части уравнения оптимальности Беллмана. Полезно наглядно увидеть это в формуле самого градиента. Представим на секунду, что мы решаем задачу регрессии с <<идеальной>>, незашумлённой целевой переменной $f(s, a) \coloneqq r(s, a) + \gamma \E_{s'} \max\limits_{a'} Q_{\theta_k}(s', a')$: для одного примера $s, a$ градиент MSE тогда будет равен:

$$\nabla_\theta \frac{1}{2}(f(s, a) - Q_{\theta}(s, a))^2 = \overbrace{(f(s, a) - Q_{\theta}(s, a))}^{\text{скаляр}}\underbrace{\nabla_\theta Q_{\theta}(s, a)}_{\mathclap{\text{\shortstack{направление увеличения \\ значения $Q_{\theta}(s, a)$}}}}$$

И наша <<зашумлённая>> целевая переменная $y(s, a)$ есть несмещённая оценка $f(s, a)$, поскольку специально построена так, что $\E_{s'} y(s, a) = f(s, a)$. Значит, мы можем несмещённо оценить этот градиент как 
\begin{equation}\label{DQNgradient}
(y(s, a) - Q_{\theta}(s, a))\nabla_\theta Q_{\theta}(s, a) = \nabla_\theta \frac{1}{2} \left( y(s, a) - Q_{\theta}(s, a) \right)^2
\end{equation}
Мы всегда оптимизируем нейронные сети стохастической градиентной оптимизации, поэтому несмещённая оценка градиента нам подойдёт.

В частности, формула Q-learning \eqref{Qlearningupdate} --- это частный случай решения указанной задачи регрессии стохастическим градиентным спуском для специального вида параметрических распределений:

\begin{theorem}
Пусть Q-функция задана <<табличным параметрическим семейством>>, то есть табличкой размера $|\St|$ на $|\A|$, где в каждой ячейке записан параметр $\theta_{s, a}$:
$$Q_{\theta}(s, a) = \theta_{s, a}$$
Тогда формула \eqref{Qlearningupdate} представляет собой градиентный спуск для решения обсуждаемой задачи регрессии.

\begin{proof}
Посчитаем градиент функции потерь на одном объекте $s, a$. Предсказанием модели будет $Q_{\theta}(s, a) \HM= \theta_{s, a}$, то есть градиент предсказания по параметрам модели равен
$$\nabla_\theta Q_{\theta}(s, a) = e_{s, a}$$
где $e_{s, a}$ --- вектор из $\R^{|\St| |\A|}$ из всех нулей с единственной единичкой в позиции, соответствующей паре $s, a$. Теперь посчитаем градиент функции потерь:
$$\nabla_\theta \frac{1}{2} \left( y - Q_{\theta}(s, a) \right)^2 = \left( Q_{\theta}(s, a) - y(s, a) \right) \nabla_\theta Q_{\theta}(s, a) = \left( Q_{\theta}(s, a) - y(s, a) \right) e_{s, a}$$
Итак, градиентный спуск делает следующие апдейты параметров:
$$\theta_{k+1} = \theta_k - \alpha \nabla_\theta \Loss \left( y, Q_{\theta}(s, a) \right) = \theta_k + \alpha \left( y(s, a) - Q_{\theta_k}(s, a) \right) e_{s, a}$$
В этой формуле на каждом шаге обновляется ровно одна ячейка таблички $\theta_{s, a}$, и это в точности совпадает с \eqref{Qlearningupdate}.
\end{proof}
\end{theorem}

Само собой, аналогичный приём мы в будущем будем применять не только для обучения модели $Q^*$, но и любых других оценочных функций. Для этого мы будем составлять задачу регрессии, выбирая в качестве целевой переменной несмещённую оценку правой части какого-нибудь уравнения Беллмана, неподвижной точкой которого является искомая оценочная функция. Этот приём является полным аналогом методов временной разности, поэтому и свойства получаемых алгоритмов будут следовать из классической теории табличных алгоритмов. 

\subsection{Таргет-сеть}

Рассмотренное теоретическое объяснение перехода от табличных методов к нейросетевым, конечно, предполагает, что мы решаем задачу регрессии <<полностью>>, обучая $\theta$ при фиксированных $\theta_k$. <<Замороженные>> $\theta_k$ соответствуют фиксированию формулы целевой переменной $y(s, a)$ \eqref{guess}, то есть фиксированию задачи регрессии. Так мы моделируем один шаг метода простой итерации, и только после этого объявляем выученные параметры модели $\theta_{k+1} \HM\coloneqq \theta$. В этот момент задача регрессии изменится (поменяется целевая переменная), и мы перейдём к следующему шагу метода простой итерации. 

Однако в Q-learning же, как и во всех табличных методах, теория стохастической аппроксимации позволяла <<сменять>> задачу регрессии каждый шаг, используя свежие параметры модели при построении целевой переменной. Конечно, как только мы от <<табличных параметрических функций>> переходим к произвольным параметрическим семействам, все теоретические гарантии сходимости Q-learning-а \ref{TDconvergence} теряются (теоремы сходимости использовали <<табличность>> нашего представления Q-функции). Как только мы используем ограниченные параметрические семейства вроде нейросеток, неидеальные оптимизаторы вроде Адама или не доводим каждый этап метода простой итерации до сходимости, гарантий нет.

Но возникает вопрос: сколько шагов градиентного спуска тратить на решение фиксированной задачи регрессии? Возникает естественное желание по аналогии с табличными методами использовать для построения таргета свежую модель, то есть менять целевую переменную в задаче регрессии каждый шаг после каждого градиентного шага:
$$y(s, a) \coloneqq r + \gamma \max_{a'} Q_{\theta}(s', a')$$
Принципиально важно, что зависимость целевой переменной $y(s, a)$ \eqref{guess} от параметров текущей модели $\theta$ игнорировалось. Если вдруг в неё протекут градиенты, мы будем не только подстраивать прошлое под будущее, но и будущее под прошлое, что не будет являться корректной процедурой. 

Эмпирически легко убедиться, что такой подход нестабилен примерно от слова совсем. Стохастическая оптимизация чревата тем, что после очередного шага модель может стать немножко <<сломанной>> и некоторое время выдавать неудачные значения на ряде примеров. В обычном обучении с учителем этот эффект сглаживается большим количеством итераций: обучение на последующих мини-батчах <<исправляют>> предыдущие ошибки, движение идёт в среднем в правильную сторону, но нет гарантий удачности каждого конкретного шага (на то это и стохастическая оптимизация). Здесь же при неудачном шаге сломанная модель может начать портить целевую переменную, на которую она же и обучается. Это приводит к цепной реакции: плохая целевая переменная начнёт портить модель, которая начнёт портить целевую переменную... Этот эффект особенно ярко проявляется ещё и потому, что мы используем одношаговые целевые переменные, которые, как мы обсуждали в главе \ref{sec:biasvar}, сильно смещены (слишком сильно опираются на текущую же аппроксимацию).

\needspace{7\baselineskip}
\begin{wrapfigure}{r}{0.4\textwidth}
\vspace{-0.3cm}
\centering
\includegraphics[width=0.4\textwidth]{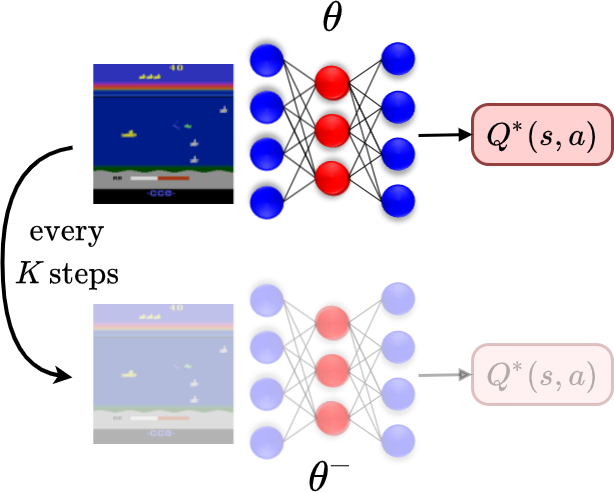}
\vspace{-1cm}
\end{wrapfigure}

Для стабилизации процесса одну задачу регрессии нужно решать более одной итерации градиентного спуска; необходимо сделать хотя бы условно 100-200 шагов. Проблема в том, что если таргет строится по формуле $r + \gamma \max\limits_{a'} Q_{\theta}(s, a)$, то после первого же градиентного шага $\theta$ поменяется.

Поэтому хранится копия $Q$-сетки, называемая \emph{таргет-сетью} (target network), единственная цель которой --- генерировать таргеты текущей задачи регрессии для транзишнов из засэмплированных мини-батчей. Традиционно её параметры обозначаются $\theta^{-}$. Итак, целевая переменная в таких обозначениях генерится по формуле
$$y( \T ) \coloneqq r + \gamma \max\limits_{a'} Q_{\theta^{-}}(s', a')$$
а раз в $K$ шагов веса $\theta^{-}$ просто копируются из текущей модели с весами $\theta$ для <<задания>> новой задачи регрессии.

\begin{remark}
При обновлении задачи регрессии график функции потерь типично подскакивает. Поэтому распространённой, но чуть более вычислительно дорогой альтернативой является на каждом шаге устраивать экспоненциальное сглаживание весов таргет сети. Тогда на каждом шаге:
$$\theta^{-} \leftarrow (1 - \alpha) \theta^{-} + \alpha \theta,$$
где $\alpha$ --- гиперпараметр. Такой вариант тоже увеличивает стабильность алгоритма хотя бы потому, что решаемая задача регрессии меняется <<постепенно>>.
\end{remark}

\subsection{Декорреляция сэмплов}

Q-learning после одного шага в среде делал апдейт одного значения таблицы $Q(s, a) \HM \approx Q^*(s, a)$. Сейчас нас такой вариант, очевидно, не устраивает, потому что обучать нейросетки на мини-батчах размера 1 (особенно с уровнем доверия к нашим таргетам) --- это явно плохая идея, но главное, сэмплы $s, a, y$ сильно скоррелированы: в сложных средах последовательности состояний обычно очень сильно похожи. Нейросетки на скоррелированных данных обучаются очень плохо (чаще --- не обучаются вовсе). Поэтому сбор мини-батча подряд с одной траектории здесь не сгодится.

\begin{center}
    \includegraphics[width=\textwidth]{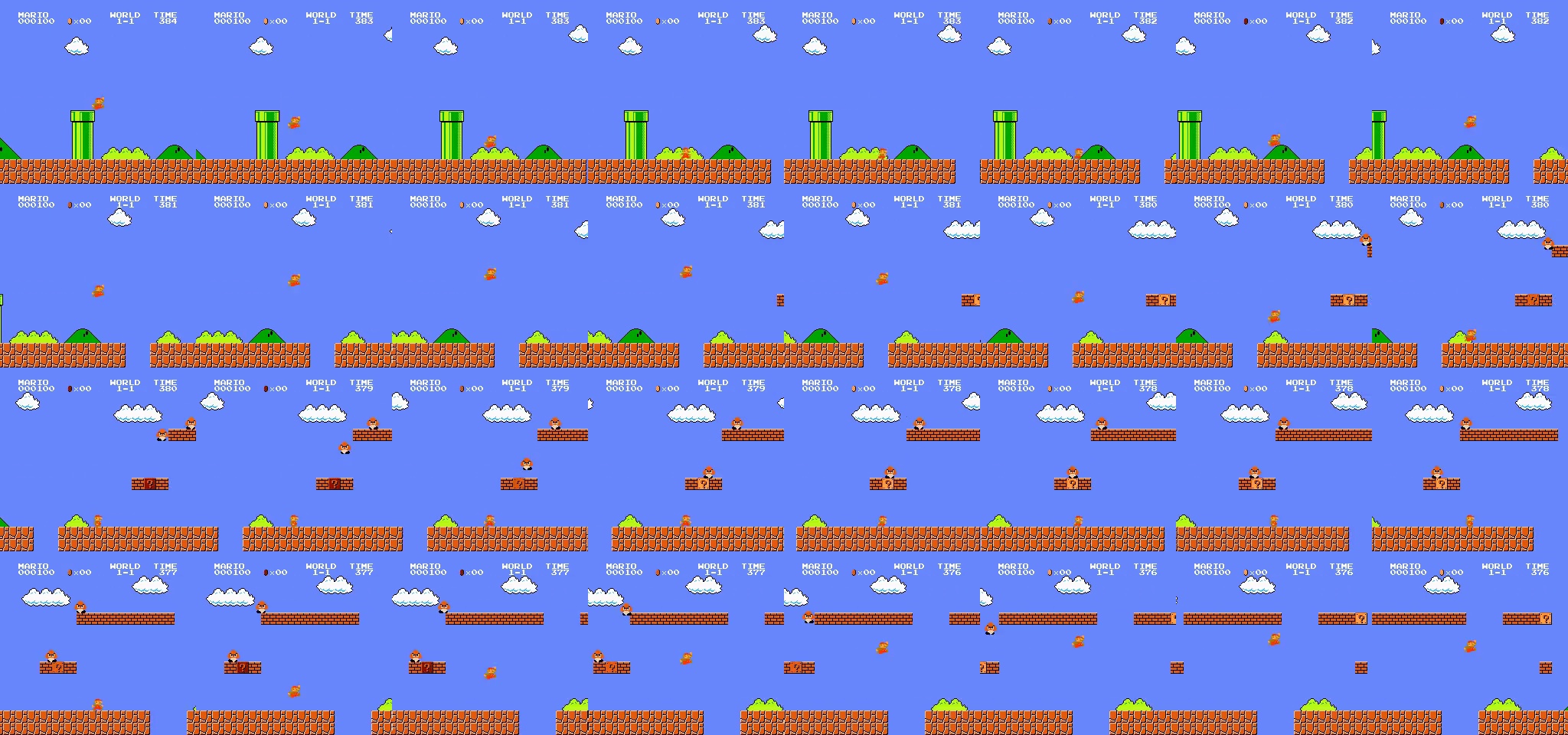}
\end{center}

Есть два доступных на практике варианта \emph{декорреляции сэмплов} (sample decorrelation). Первый --- запуск параллельных агентов, то есть сбор данных сразу в нескольких параллельных средах. Этот вариант доступен всегда, по крайней мере, если среда виртуальна; иначе эта опция может быть дороговатой... Второй вариант --- реплей буфер, который, как мы помним, является прерогативой исключительно off-policy алгоритмов.

При наличии реплей буфера агент может решать задачи регрессии, сэмплируя мини-батчи переходов $\T \HM= (s, a, r', s')$ из буфера, затем делая для каждого перехода расчёт таргета\footnote{наличие реплей буфера не избавляет от необходимости использовать таргет-сеть, поскольку вычисление таргета для всего реплей буфера, конечно же, непрактично: реплей буфер обычно огромен (порядка $10^6$ переходов), а одна задача регрессии будет решаться суммарно 100-200 шагов на мини-батчах размера, там, 32 (итого таргет понадобится считать всего для порядка 3000 переходов).} $y(\T) \coloneqq r \HM+ \gamma \max\limits_{a'} Q_{\theta^{-}}(s', a')$, игнорируя зависимость таргета от параметров, и проводя шаг оптимизации по такому мини-батчу. Такой батч уже будет декоррелирован.

\begin{center}
    \includegraphics[width=\textwidth]{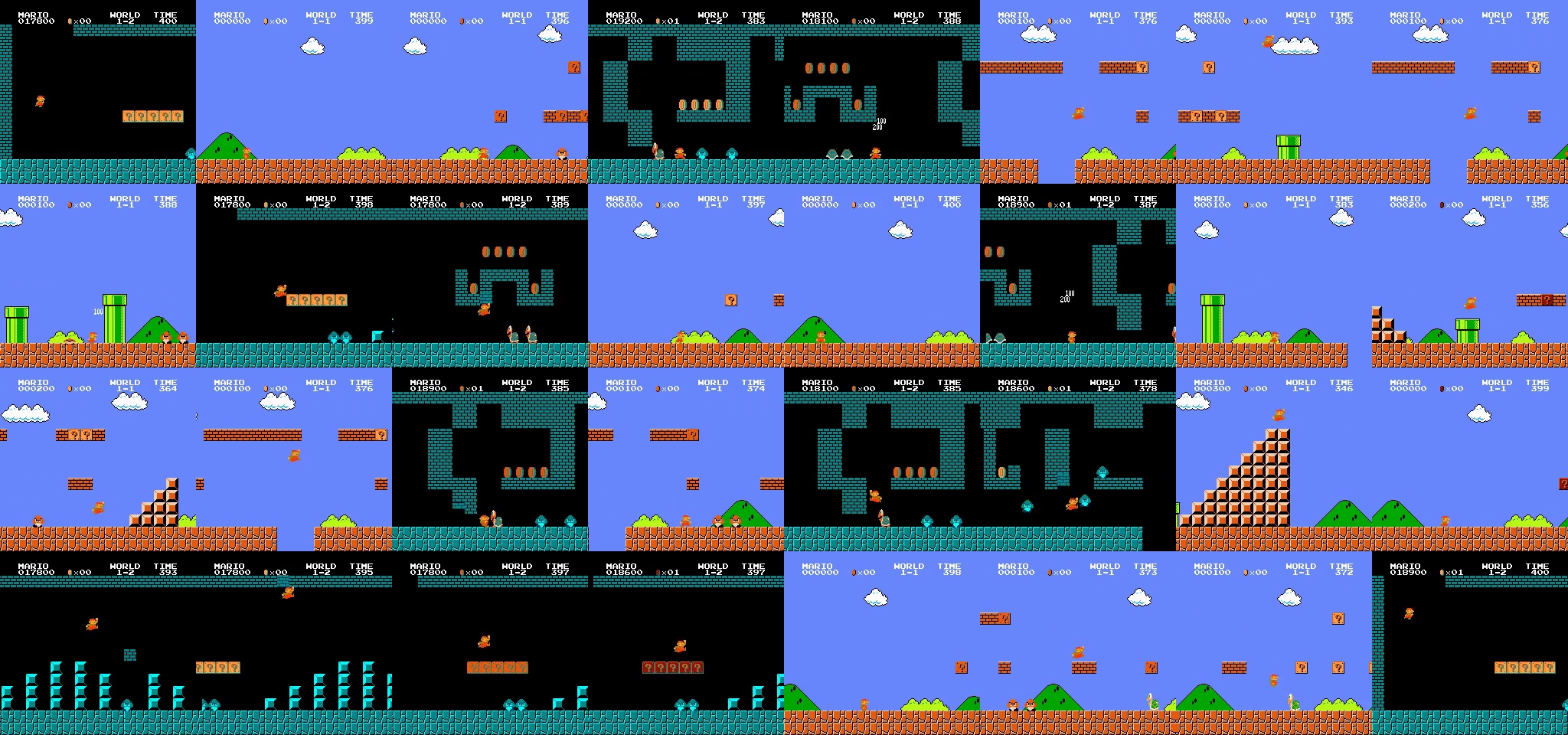}
\end{center}

Почему мы можем использовать здесь реплей буфер? Мы хотим решать уравнения оптимальности Беллмана для всех пар $s, a$. Поэтому в поставленной задаче регрессии мы можем брать тройки $s, a, y$ для обучения из условно произвольного распределения до тех пор, пока оно достаточно разнообразно и нескоррелированно. Единственное ограничение --- внутри $y$ сидит $s'$, который должен приходить из и только из $p(s' \mid s, a)$. Но поскольку среда однородна, любые тройки $s, a, s'$ из любых траекторий, сгенерированных любой стратегией, таковы, что $s' \HM\sim p(s' \HM\mid s, a)$, а значит, $s'$ может быть использован для генерации таргета. Иными словами, в рамках текущего подхода мы, как и в табличном Q-learning-е, находимся в off-policy режиме, и наши условия на процесс порождения переходов $s, a, r, s'$ точно такие же.

\begin{remark}
На практике картинки в какой-то момент начинают переполнять оперативку и начинаются проблемы. Простое решение заключается в том, чтобы удалять самые старые переходы, то есть оставлять самый новый опыт, однако есть альтернативные варианты (см. приоритизированный реплей, раздел \ref{subsec:prioritizedreplay})
\end{remark}

\subsection{DQN}

Собираем алгоритм целиком. Нам придётся оставить $\eps$-жадную стратегию исследования --- с проблемой исследования-использования мы ничего пока не делали, и при стартовой инициализации есть риск отправиться тупить в ближайшую стену. 

Также лишний раз вспомним про то, что в терминальных состояниях обязательно нужно домножаться на $(1 - \done)$, поскольку шансов у приближённого динамического программирования сойтись куда-то без <<отправной точки>> не очень много. 

\begin{algorithm}[label = DQNalgorithm]{Deep Q-learning (DQN)}
\textbf{Гиперпараметры:} $B$ --- размер мини-батчей, $K$ --- периодичность апдейта таргет-сети, $\eps(t)$ --- стратегия исследования, $Q$ --- нейросетка с параметрами $\theta$, SGD-оптимизатор

\vspace{0.3cm}
Инициализировать $\theta$ произвольно \\
Положить $\theta^- \coloneqq \theta$ \\
Пронаблюдать $s_0$ \\
\textbf{На очередном шаге $t$:}
\begin{enumerate}
    \item выбрать $a_t$ случайно с вероятностью $\eps(t)$, иначе $a_t \coloneqq \argmax\limits_{a_t} Q_{\theta}(s_t, a_t)$
    \item пронаблюдать $r_t$,  $s_{t+1}$, $\done_{t+1}$
    \item добавить пятёрку $(s_t, a_t, r_t, s_{t+1}, \done_{t+1})$ в реплей буфер
    \item засэмплировать мини-батч размера $B$ из буфера
    \item для каждого перехода $\T = (s, a, r, s', \done)$ посчитать таргет:
    $$y(\T ) \coloneqq r + \gamma (1 - \done) \max\limits_{a'} Q_{\theta^{-}}(s', a')$$
    \item посчитать лосс:
    $$\Loss(\theta) \coloneqq \frac{1}{B}\sum_{\T} \left( Q_{\theta}(s, a) - y(\T ) \right) ^2$$
    \item сделать шаг градиентного спуска по $\theta$, используя $\nabla_\theta \Loss(\theta)$
    \item если $t \operatorname{mod} K = 0$: $\theta^- \gets \theta$
\end{enumerate}
\end{algorithm}

Все алгоритмы, которые относят к Value-based подходу в RL, будут основаны на DQN: само название <<value-based>> обозначает, что мы учим только оценочную функцию, а такое возможно только если мы учим модель Q-функции и полагаем, что policy improvement проводится на каждом шаге жадно. Неявно в DQN, конечно же, присутствует текущая политика, <<целевая политика>>, которую мы оцениваем --- $\argmax\limits_a Q_{\theta}(s, a)$. Тем не менее ключевое свойство алгоритма, которое стоит помнить --- это то, что он работает в off-policy режиме, и потому потенциально является достаточно sample efficient.

Есть, однако, много причин, почему алгоритм может не раскрыть этот потенциал и <<плохо>> заработать на той или иной среде: либо совсем не обучиться, либо обучаться очень медленно. Со многими из этих недостатков можно пытаться вполне успешно бороться, что и пытаются делать модификации этого алгоритма, которые мы обсудим далее в главе \ref{sec:dqnmods}. 

Выделим отдельно одну особую фундаментальную причину, почему алгоритмы на основе DQN могут не справиться с оптимизацией награды и застрять на какой-то асимптоте, с которой мало что можно сделать, и которая вытекает непосредственно из off-policy режима работы DQN. DQN, как и любые алгоритмы, основанные на одношаговых целевых переменных, страдает от проблемы \emph{накапливающейся ошибки} (compound error). Условно говоря, чтобы распространить награду, полученную в некоторый момент времени, на 100 шагов в прошлое, понадобится провести 100 этапов метода простой итерации. Каждый этап мы решаем задачу регрессии в сильно неидеальных условиях, и ошибка аппроксимации накапливается. На это накладывается необходимость обучать именно Q-функцию, которая должна дифференцировать между действиями.

\begin{example}
Рассмотрим типичную задачу: вы можете перемещаться в пространстве в разные стороны. Если вы отправитесь вправо, через 100 шагов вы получите +1. $Q^*(s, \text{вправо}) \HM= \gamma^{100}$. 

\begin{wrapfigure}{r}{0.3\textwidth}
\vspace{-0.5cm}
\centering
\includegraphics[width=0.3\textwidth]{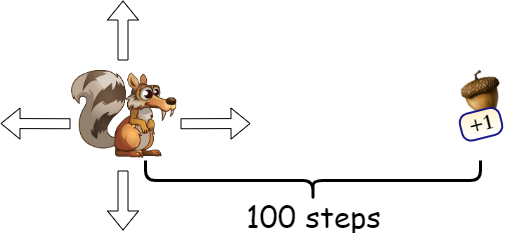}
\vspace{-0.9cm}
\end{wrapfigure}

Посмотрим на значение функции в других действиях: например, $Q^*(s, \text{влево}) \HM= \gamma^{102}$. Вот с такой точностью наша $Q^*$ должна обучиться в этом состоянии, чтобы жадная стратегия выбирала правильные действия. Именно поэтому в подобных ситуациях DQN-подобные алгоритмы не срабатывают: из-за проблемы накапливающейся ошибки сигнал на 100 шагов просто не распространяется с такой точностью.
\end{example}

Если же этих проблем <<с точностью>> не возникает, если в среде нет сильно отложенного сигнала или награда за каждый шаг очень информативна, то value-based подход может раскрыть свой потенциал и оказаться эффективнее альтернатив за счёт использования реплей буфера.

\section{Модификации DQN}\label{sec:dqnmods}

\subsection{Overestimation Bias}\label{subsec:overestimation}

Отмечалось, что без таргет-сетки (при обновлении задачи регрессии каждый шаг) можно наблюдать, как $Q$ начинает неограниченно расти. Хотя таргет-сетка более-менее справляется с тем, чтобы стабилизровать процесс, предотвратить этот эффект полностью у неё не получается: сравнение обучающейся $Q$ с Монте-Карло оценками и зачастую просто со здравым смыслом выдаёт присутствие в алгоритме заметного \emph{смещения в сторону переоценки} (overestimation bias). Почему так происходит?

Очевидно, что источник проблемы --- оператор максимума в формуле построения таргета: 
$$y(\T ) = r + \gamma \max\limits_{a'} Q_{\theta^{-}}(s', a')$$
При построении таргета есть два источника ошибок: 1) ошибка аппроксимации 2) внешняя стохастика. Максимум здесь <<выбирает>> то действие, для которого из-за ошибки нейросети или из-за везения в прошлых играх $Q(s', a')$ больше правильного значения.

\begin{example}
Вы выиграли в лотерею +100 в силу везения. В вашем опыте нет примеров того, как вы купили билет и проиграли, и поэтому алгоритм будет учить завышенное значение $Q(s, a)$ для действия <<купить билет в лотерею>> (напомним, в утверждении \ref{prop:qlearningempiricmdp} мы отмечали, что запущенный с реплей буфера Q-learning, как и DQN, учит оптимальную Q-функцию для эмпирического MDP). Это завышение из-за везения.

Из-за того, что вы моделируете Q-сетку нейросетью, при увеличении $Q(s, a)$ случайно увеличилось значение ценности действия <<играть в казино>>, поскольку оно опиралось на примерно те же признаки описания состояний. Это завышение из-за ошибки аппроксимации.

А дальше начинается <<цепная реакция>>: при построении таргета для состояния <<казино>> $s'$ в задачу регрессии поступают завышенные значения, предвещающие кучу награды. Завышенное значение начинает распространятся дальше на другие состояния: вы начинаете верить, что вы всегда можете пойти в казино и получить кучу награды.
\end{example}

\begin{proposition}
Рассмотрим одно $s'$. Пусть $Q^*(s', a')$ --- истинные значения, и для каждого действия $a'$ есть модель $Q(s', a') \HM \approx Q^*(s', a')$, которая оценивает это действие с некоторой погрешностью $\eps(a')$:
$$Q(s', a') = Q^*(s', a') + \eps(a')$$
Пусть эти погрешности $\eps(a')$ независимы по действиям, а завышение и занижение оценки равеновероятны:
$$\Prob( \eps(a') > 0) = \Prob( \eps(a') < 0) = 0.5$$
Тогда оценка максимума скорее завышена, чем занижена:
$$\Prob \left( \max_{a'} Q(s', a') > \max_{a'} Q^*(s', a') \right) > 0.5$$

\begin{proof}
С вероятностью 0.5 модель переоценит самое хорошее действие, на котором достигается максимум истинной $Q^*(s', a')$; и ещё к тому же с некоторой ненулевой вероятностью возникнет настолько завышенная оценка какого-нибудь из других действий, что $Q(s', a') \HM> \max\limits_{a'} Q^*(s', a')$.
\end{proof}
\end{proposition}

\begin{example}
Предположим, для $s'$ у нас есть три действия, и на самом деле $Q^*(s', a') \HM= 0$ для всех трёх возможных действий. Мы оцениваем каждое действие с некоторой погрешностью. Допустим даже, эта погрешность в среднем равна нулю и распределена по Гауссу: $Q(s', a') \HM \sim \N(0, \sigma^2)$. Тогда случайная величина $\max\limits_{a'} Q(s', a')$ --- взятие максимума из трёх сэмплов из гауссианы --- очевидно такова, что
$$\E \max_{a'} Q(s', a') > 0$$
Хотя истинный максимум $\max\limits_{a'} Q^*(s', a') \HM= 0$. Получается, что, несмотря на то, что каждый элемент оценён несмещённо, максимум по аппроксимациям будет смещён в сторону завышения.
\end{example}

Одно из хороших решений проблемы заключается в разделении (decoupling) двух этапов подсчёта максимума: \emph{выбор действия} (action selection) и \emph{оценка действия} (action evaluation):
$$\max\limits_{a'} Q(s', a') = \overbrace{Q(s', \underbrace{\argmax\limits_{a'} Q(s', a')}_{\text{выбор действия}})}^{\text{оценка действия}}$$

Действительно: мы словно дважды используем одну и ту же погрешность при выборе действия и оценке действия. Из-за скоррелированности ошибки на этих двух этапах и возникает эффект завышения.

Основная идея борьбы с этой проблемой заключается в следующем: предлагается\footnote{когда эту идею предлагали в 2010-ом году в рамках классического RL (тогда нейронки ещё не вставили), то назвали её Double Q-learning, но сейчас под Double DQN подразумевается алгоритм 2015-го года (см.~раздел \ref{subsec:doubledqn}), а этот трюк иногда именуется <<близнецами>> (Twin). Правда, в последнее время из-за алгоритма Twin Delayed DDPG (см. раздел \ref{subsec:td3}) под словом Twin понимается снова не эта формула, и с названиями есть небольшая путаница...} обучать два приближения Q-функции параллельно и использовать аппроксимацию Q-функции <<независимого близнеца>> для этапа оценивания действия:
$$\textcolor{ChadBlue}{y_1}(\T ) \coloneqq r + \gamma \textcolor{ChadPurple}{Q_{\theta_2}}(s', \argmax\limits_{a'} \textcolor{ChadBlue}{Q_{\theta_1}}(s', a'))$$
$$\textcolor{ChadPurple}{y_2}(\T ) \coloneqq r + \gamma \textcolor{ChadBlue}{Q_{\theta_1}}(s', \argmax\limits_{a'} \textcolor{ChadPurple}{Q_{\theta_2}}(s', a'))$$

Если обе аппроксимации Q-функции идеальны, то, понятное дело, мы всё равно получим честный максимум. Однако, если оба DQN честно запущены параллельно и даже собирают каждый свой опыт (что в реальности, конечно, дороговато), их ошибки аппроксимации и везения будут в <<разных местах>>. В итоге, Q-функция близнеца выступает в роли более пессимистичного критика действия, выбираемого текущей Q-функцией. 

\begin{remark}
Если сбор уникального опыта для каждого из близнецов не организуется, Q-функции всё равно получаются скоррелированными. Как минимум, можно сэмплировать для обучения сеток разные мини-батчи из реплей буфера, если он общий.
\end{remark}

Интуиция, почему <<независимая>> Q-функция близнеца используется именно для оценки действия, а не наоборот для выбора: если из-за неудачного градиентного шага наша сетка $ \textcolor{ChadBlue}{Q_{\theta_1}}$ пошла куда-то не туда, мы не хотим, чтобы плохие значения попадали в её же таргет $\textcolor{ChadBlue}{y_1}$. Плохие действия в таргет пусть попадают: пессимистичная оценка здесь предпочтительнее.

\begin{remark}
На практике таргет-сети в такой модели всё равно используют, и в формулах целевой переменной всюду используются именно <<замороженные>> $\textcolor{ChadBlue}{Q_{\theta^-_1}}$ и $\textcolor{ChadPurple}{Q_{\theta^-_2}}$. Это ещё больше противодействует цепным реакциям. Без них всё равно может быть такое, что сломанная $\textcolor{ChadBlue}{Q_{\theta_1}}$ поломает $\textcolor{ChadPurple}{y_2}$, та поломает $\textcolor{ChadPurple}{Q_{\theta_2}}$, та поломает $\textcolor{ChadBlue}{y_1}$, а та в свою очередь продолжить ломать $\textcolor{ChadBlue}{Q_{\theta_1}}$, хотя, конечно, такая <<цепь>> менее вероятна, чем в обычном DQN.
\end{remark}

\begin{example}
На картинке для каждого из четырёх действий указаны значения идеальной $Q^*(s', a')$ и две аппроксимации $Q_1, Q_2$. Каждая аппроксимация условно выдаёт значение с погрешностью, которая в среднем равна нулю (вероятности завышенной оценки или заниженной оценки равны).

\begin{center}
    \includegraphics[width=0.6\textwidth]{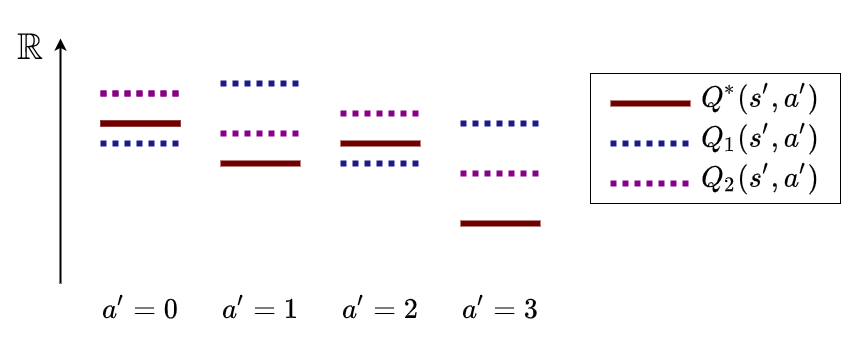}
\end{center}

Видно, что обе аппроксимации оценивают максимум по действиям завышенно. Но если одна из аппроксимаций выберет действие (то $a'$, на котором достигается её максимум), а другая аппроксимация выдаст значение в выбранном индексе, то получится более близкое к адекватной оценки истинного максимума значение.
\end{example}

\subsection{Twin DQN}\label{subsec:clippedtwin}

Разовьём интуицию дальше. Вот мы строим таргет для $Q_{\theta_1}$. Мы готовы выбрать при помощи неё же действия, но не готовы оценить их ею же самой в силу потенциальной переоценки. Для этого мы и берём <<независимого близнеца>> $Q_{\theta_2}$. Но что, если он выдаёт ещё больше? Что, если его оценка выбранного действия, так случилась, потенциально ещё более завышенная? Давайте уж в таких ситуациях всё-таки брать то значение, которое поменьше! Получаем следующую интересную формулу, которую называют twin-оценкой (или ещё clipped double оценкой):
$$\textcolor{ChadBlue}{y_1}(\T ) \coloneqq r + \gamma \min_{i = \textcolor{ChadBlue}{1}, \textcolor{ChadPurple}{2}}Q_{\theta_i}(s', \argmax\limits_{a'} \textcolor{ChadPurple}{Q_{\theta_2}}(s', a'))$$
$$\textcolor{ChadPurple}{y_2}(\T ) \coloneqq r + \gamma \min_{i = \textcolor{ChadBlue}{1}, \textcolor{ChadPurple}{2}}Q_{\theta_i}(s', \argmax\limits_{a'} \textcolor{ChadBlue}{Q_{\theta_1}}(s', a'))$$

Это очень забавная формула, поскольку она говорит бороться с проблемой <<клин клином>>: зная, что взятие максимума по аппроксимациям приводит к завышению, мы вводим в оценку искусственное занижение, добавляя в формулу минимум по аппроксимациям! По сути, это формула <<ансамблирования>> Q-функций: только вместо интуитивного среднего берём минимум, <<для борьбы с максимумом>>. 

\begin{remark}
Конечно, в отличие от обычного ансамблирования в машинном обучении, такой подход плохо масштабируется с увеличением числа обучаемых Q-функций (обычно учат всё-таки только две). В случае ансамбля имеет смысл брать не среднее, и не минимум, а, например, какой-то квантиль, более близкий к минимуму, чем к медиане --- но возникает неудобный гиперпараметр, что же именно брать.
\end{remark}

\subsection{Double DQN}\label{subsec:doubledqn}

Запускать параллельно обучение двух сеток дороговато, а при общем реплей буфере корреляция между ними всё равно будет. Поэтому предлагается простая идея: запускать лишь один DQN, а в формуле таргета для оценивания вместо <<близнеца>> использовать таргет-сеть. То есть: пусть $\theta$ --- текущие веса, $\theta^{-}$ --- веса таргет-сети, раз в $K$ шагов копирующиеся из $\theta$. Тогда таргет вычисляется по формуле:
$$y(\T ) \coloneqq r + \gamma Q_{\theta^{-}}(s', \argmax\limits_{a'} Q_{\theta}(s', a'))$$

Хотя понятно, что таргет-сеть и текущая сетка очень похожи, такое изменение формулы целевой переменной всё равно <<избавляет>> нас от взятия оператора максимума; эмпирически оказывается, что такая декорреляция действительно помогает стабилизации процесса. При этом, в отличие от предыдущих вариантов, такое изменение бесплатно: не требует обучения второй нейросети.

\begin{remark}
Если обучение второй Q-сети тяжеловесно, то рекомендуется использовать формулу Double DQN. Если же речь идёт об обучении маленьких полносвязных нейросеток, то вероятно, обучение <<ансамбля>> хотя бы из двух Q-функций по крайней мере с общего буфера не должно быть особо дорогим, и тогда имеет смысл пользоваться Twin-оценкой из раздела \ref{subsec:clippedtwin}.
\end{remark}

\subsection{Dueling DQN}\label{subsec:duelingdqn}

Рассмотрим ещё одну очевидную беду Q-обучения на примере. 

\begin{example}
Вы в целях исследования попробовали кинуться в яму $s$. Сидя в яме, вы попробовали $a$ --- поднять правую руку. В яме холодно и грустно, поэтому вы получили -100. Какой вывод делает агент? Правильно: для этого состояния $s$ оценку этого действия $a$ нужно понизить. Остальные не трогать; оценка самого состояния (по формуле \eqref{V*Q*} --- максимум Q-функции по действиям) скорее всего не изменится, и надо бы вернуться в это состояние и перепроверить ещё и все остальные действия: попробовать поднять левую руку, свернуться калачиком...
\end{example}

В сложных MDP ситуация зачастую такова, что получение негативной награды в некоторой области пространства состояний означает в целом, что попадание в эту область нежелательно. Верно, что с точки зрения теории остальные действия должны быть <<исследованы>>, но неудачный опыт должен учитываться внутри оценки самого состояния; иначе агент будет возвращаться в плохие состояния с целью перепробовать все действия без понимания, что удача здесь маловероятна, вместо того, чтобы исследовать те ветки, где негативного опыта <<не было>>. Понятно, что проблема тем серьёзнее, чем больше размерность пространства действий $|\A|$.

\needspace{7\baselineskip}
\begin{wrapfigure}{r}{0.4\textwidth}
\vspace{-0.5cm}
\centering
\includegraphics[width=0.4\textwidth]{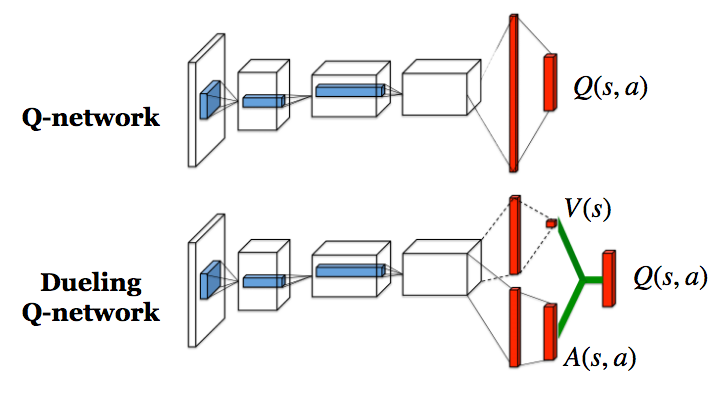}
\vspace{-0.5cm}
\end{wrapfigure}

Формализуя идею, мы хотели бы в модели учить не $Q^*(s, a)$ напрямую, а получать их с учётом $V^*(s)$. Иными словами, модель должна знать ценность самих состояний и с её учётом выдавать ценности действий. При этом при получении, скажем, негативной информации о ценности одного из действий, должна понижаться в том числе и оценка V-функции. \emph{Дуэльная} (dueling) архитектура --- это модификация вычислительного графа нашей модели с параметрами $\theta$, в которой на выходе предлагается иметь две головы, V-функцию $V(s) \HM \approx V^*(s)$ и Advantage-функцию $A(s, a) \HM \approx A^*(s, a)$: 
\begin{equation}\label{naivedueling}
Q_{\theta}(s, a) \coloneqq V_{\theta}(s) + A_{\theta}(s, a)
\end{equation}

Это интересная идея решить проблему просто сменой архитектуры вычислительного графа: при апдейте значения для одной пары $s, a$ неизбежно поменяется значение $V_{\theta}(s)$ ценности всего состояния, которое общее для всех действий. Мы таким образом вводим следующий <<прайор>>: ценности действий в одном состоянии всё-таки связаны между собой, и если одно действие в состоянии плохое, то вероятно, что и остальные тоже плохие (<<само состояние плохое>>).  

Здесь есть подвох: если $V^*(s)$ --- это произвольный скаляр, то $A^*(s, a)$ --- не произвольный. Действительно: мы моделируем $|\A|$ чисел, выдавая почему-то $|\A| \HM+ 1$ число: то есть почему-то вводим <<лишнюю степень свободы>>. Вообще-то, Advantage-функция не является произвольной функцией и обязана подчиняться \eqref{pr:advantageiszero}. Для оптимальных стратегий в предположении жадности нашей стратегии это утверждение вырождается в следующее свойство:

\begin{proposition}
$$\forall s \colon \max_a A^*(s, a) = 0$$
\beginproof
\begin{align*}
    \max_a A^*(s, a) = \max_a Q^*(s, a) - V^*(s) = \{ \text{связь V*Q* \eqref{V*Q*}} \} = 0 \tagqed
\end{align*}
\end{proposition}

Это условие можно легко учесть, вычтя максимум в формуле \eqref{naivedueling}:
\begin{equation}\label{dueling}
Q_{\theta}(s, a) \coloneqq V_{\theta}(s) + A_{\theta}(s, a) - \max_{\hat{a}} A_{\theta}(s, \hat{a})    
\end{equation}
Таким образом мы гарантируем, что максимум по действиям последних двух слагаемых равен нулю, и они корректно моделируют Advantage-функцию.

Заметим, что $V$ и $A$ не учатся по отдельности (для $V^*$ уравнение оптимальности Беллмана не сводится к регрессии, для $A^*$ уравнения Беллмана не существует вообще); вместо этого минимизируется лосс для Q-функции точно так же, как и в обычном DQN.

\begin{remark}
В нейросетях формулу \eqref{dueling} реализовать очень просто при помощи двух голов: одна выдаёт скаляр, другая $|\A|$ чисел, из которых вычитается максимум. Дальше к каждой компоненте второй головы добавляется скаляр, выданный первой головой, и результат считается выданным моделью $Q_{\theta}(s, a)$.
\end{remark}

Нюанс: авторы статьи эмпирически обнаружили, что замена максимума на среднее даёт чуть лучшие результаты. Вероятно, в реплей буфере очень часто встречаются пары $s, a$ такие, что $a$ --- наилучшее действие в этом состоянии по мнению текущей модели (а то есть $a \HM = \argmax\limits_{a} A_{\theta}(s, a)$), и поэтому градиент $A_{\theta}(s, a) \HM- \max\limits_{\hat{a}} A_{\theta}(s, \hat{a})$ часто зануляется. В результате, на текущий день под дуэльной архитектурой понимают альтернативную формулу:
\begin{equation}\label{kludgedueling}
Q_{\theta}(s, a) \coloneqq V_{\theta}(s) + A_{\theta}(s, a) - \frac{1}{|\A|}\sum_{\hat{a}} A_{\theta}(s, \hat{a})    
\end{equation}

\subsection{Шумные сети (Noisy Nets)}\label{subsec:noisynets}

По дефолту, алгоритмы на основе DQN решают дилемму исследования-использования при помощи примитивной $\eps$-жадной стратегии взаимодействия со средой. Этот бэйзлайн-подход плох примерно всем, в первую очередь тем, что крайне чувствителен к гиперпараметрам: для $\eps$ обязательно нужно составлять какое-нибудь расписание, чтобы в начале обучения он был побольше, а потом постепенно затухал, и откуда брать это расписание --- непонятно. При этом слишком большие значения шума существенно замедляют обучение, заставляя агента вести себя случайно, а раннее затухание приведёт к застреванию алгоритма в каком-нибудь локальном оптимуме (агент будет биться головой об стенку, не пробуя её обойти). 
\begin{remark}
Чтобы понять, что случилось именно это, можно посмотреть игры агента: если, например, Марио всё время доходит до середины первого уровня и прыгает в одну и ту же яму, то у него просто нет положительного опыта перепрыгивания этой ямы, и поэтому он не знает, что можно набрать больше награды.
\end{remark}

\needspace{7\baselineskip}
\begin{wrapfigure}{r}{0.4\textwidth}
\vspace{-0.3cm}
\centering
\includegraphics[width=0.4\textwidth]{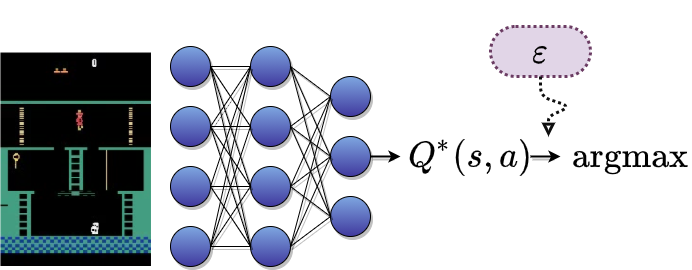}
\vspace{-0.3cm}
\end{wrapfigure}

Ключевая причина, почему $\eps$-жадная стратегия примитивна, заключается в независимости добавляемого шума от текущего состояния. Мы выдаём оценки Q-функции и в зависимости только от времени принимаем решение, использовать ли эти знания или эксплорить. Интуитивно, правильнее было бы принимать это решение в зависимости от текущего состояния: если состояние исследовано, чаще принимать решение в пользу использования знаний, если ново --- в пользу исследования. Открытие новой области пространства состояний скорее всего означает, что в ней стоит поделать разные действия, когда двигаться к ней нужно за счёт использования уже накопленных знаний.

\needspace{7\baselineskip}
\begin{wrapfigure}{l}{0.4\textwidth}
\vspace{-0.3cm}
\centering
\includegraphics[width=0.4\textwidth]{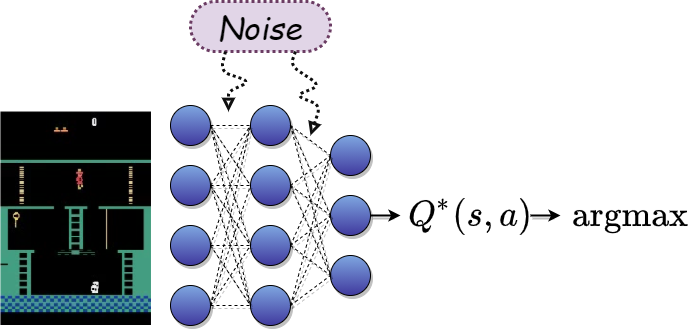}
\vspace{-0.3cm}
\end{wrapfigure}

\emph{Шумные сети} (noisy nets) --- добавление шума с обучаемой и, главное, зависимой от состояния (входа в модель) дисперсией. Хак чисто инженерный: давайте каждый параметр в модели заменим на
$$\theta_i \coloneqq w_i + \sigma_i \eps_i, \qquad \eps_i \sim \N(0, 1),$$
или, другими словами, заменим веса сети на сэмплы из $\N(w_i, \sigma^2_i)$, где $w, \sigma \in \R^h$ --- параметры модели, обучаемые градиентным спуском. Очевидно, что выход сети становится случайной величиной, и, в зависимости от шума $\eps$, будет меняться выбор действия $a \HM= \argmax\limits_a Q_{\theta}(s, a, \eps)$. При этом влияние шума на принятое решение зависит от поданного в модель входного состояния.

\begin{remark}
Если состояния --- изображения, шум в свёрточные слои обычно не добавляется (зашумлять выделение объектов из изображения кажется бессмысленным).
\end{remark}

Формально, коли наша модель стала стохастичной, мы поменяли оптимизируемый функционал: мы хотим минимизировать функцию потерь в среднем по шуму:
$$\E_{\eps} \Loss(\theta, \eps) \to \min_\theta$$
Видно, что градиент такого функционала можно несмещённо оценивать по Монте-Карло:
$$\nabla_\theta \E_{\eps} \Loss(\theta, \eps) = \E_{\eps} \nabla_\theta \Loss(\theta, \varepsilon) \approx \nabla_\theta \Loss(\theta, \eps), \qquad \eps \sim \N(0, I)$$

Гарантий, что магнитуда шума в среднем будет падать для исследованных состояний, вообще говоря, нет. Надежда этой идеи в том, что магнитуда будет подстраиваться в зависимости от текущих в модель градиентов: если модель часто видит какое-то $s$ и функция потерь говорит, что на этом состоянии нужно выдавать, скажем, низкое значение, модель будет учиться при любых сэмплах $\eps$ выдавать указанное низкое значение. Для этого модели будет удобно уменьшать дисперсию впрыскиваемого шума и больше опираться на те нейронные связи, которые мало зашумлены. Если же в сеть поступают противоречивые сигналы о паре $s, a$, или это какое-то новое $s$, которого модель ещё не видела, выходное значение модели будет, интуитивно, сильно зашумлено, и часто аргмаксимум будет достигаться именно на нём. 

Ещё одно ключевое преимущество идеи в том, что в этом подходе отсутствуют гиперпараметры.

\begin{remark}
Кроме инициализации. Неудачная инициализация $\sigma$ всё равно может замедлить процесс обучения; обычно дисперсию шума инициализируют какой-нибудь константой, и эта константа становится в некотором смысле важным гиперпараметром алгоритма.
\end{remark}

Заметим, что таргет $y(\T)$, который мы генерируем для каждого перехода $\T$ из батча, формально теперь тоже должен вычисляться как мат.ожидание по шуму:
$$y(\T) \coloneqq \E_\eps \left[ r + \gamma \max_{a'} Q_{\theta}(s', a', \eps) \right]$$
Опять же, мат.ожидание несмещённо оценивается по Монте-Карло, однако с целью декорреляции полезно использовать в качестве $\eps$ другие сэмплы, нежели используемые при вычислении лосса. Считается, что подобное <<зашумление>> целевой переменной в DQN может даже пойти на пользу.

\begin{remark}
Генерация сэмплов шума по числу параметров нейросети на видеокарте может сильно замедлить время прохода через сеть. Для оптимизации авторы предлагают для матриц полносвязных слоёв генерировать шум следующим образом. Пусть $n$ --- число входов, $m$ --- число выходов в слое. Сэмплируются $\eps_1 \sim \N(0, I_{m \times m}), \eps_2 \sim \N(0, I_{n \times n})$, после чего полагается шум для матрицы равным
$$\eps \coloneqq f(\eps_1) f(\eps_2)^T$$
где $f$ --- масштабирующая функция, например $f(x) = \operatorname{sign}(x)\sqrt{|x|}$ (чтобы каждый сэмпл в среднем всё ещё имел дисперсию 1). Процедура требует всего $m + n$ сэмплов вместо $mn$, но жертвует независимостью сэмплов внутри слоя. К сожалению, даже при таком хаке время работы алгоритма заметно увеличивается, поскольку проходов через нейросеть в DQN нужно делать очень много.
\end{remark}

\begin{remark}
Альтернативно, можно зашумлять выходы слоёв (тогда сэмплов понадобится на порядок меньше) или просто добавлять шум на вход. В обоих случаях, <<зашумлённость>> выхода будет обучаемой, а степень влияния шума на выход сети будет зависеть от состояния.
\end{remark}

\subsection{Приоритизированный реплей (Prioritized DQN)}\label{subsec:prioritizedreplay}


\needspace{7\baselineskip}
\begin{wrapfigure}{r}{0.3\textwidth}
\vspace{-0.3cm}
\centering
\includegraphics[width=0.3\textwidth]{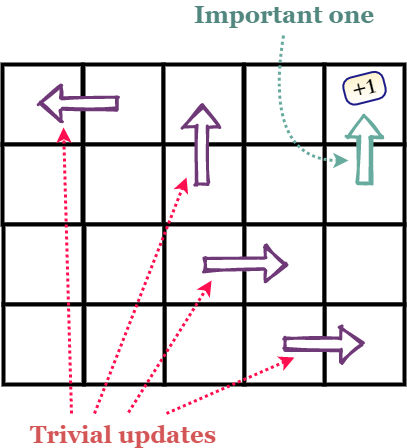}
\vspace{-0.3cm}
\end{wrapfigure}

Off-policy алгоритмы позволяют хранить и переиспользовать весь накопленный опыт. Однако, интуитивно ясно, что встречавшиеся переходы существенно различаются по важности. Зачастую большая часть буфера, особенно поначалу обучения, состоит из записей изучения агентом ближайшей стенки, а переходы, включавшие, например, получение ещё не выученной внутри аппроксимации Q-функции награды, встречаются в буфере сильно реже и при равномерном сэмплировании редко оказываются в мини-батчах.

Важно, что при обучении оценочных функций информация о награде распространяется от последних состояний к первым. Например, на первых итерациях довольно бессмысленно обновлять те состояний, где сигнала награды не было ($r(s, a) \HM= 0$), а Q-функция для следующего состояния примерно случайна (а именно такие переходы чаще всего и попадаются алгоритму). Такие обновления лишь схлопывают выход аппроксимации к константе (которая ещё и имеет тенденцию к росту из-за оператора максимума). Ценной информацией поначалу являются терминальные состояния, где целевая переменная по определению равна $y(\T ) = r(s, a)$ и является абсолютно точным значением $Q^*(s, a)$. Типично, что на таких переходах значения временной разницы (лосса DQN) довольно высоко. Аналогичная ситуация в принципе справедлива для любых наград, которые для агента новы и ещё не распространились в аппроксимацию через уравнение Беллмана.

Очень хочется сэмплировать переходы из буфера не равномерно, а приоритизировано. Приоритет установим, например, следующим образом:
\begin{equation}\label{priority}
    \rho(\T ) \coloneqq \left( y(\T ) - Q_{\theta}(s, a) \right)^2 = \Loss(y(\T ), Q_{\theta}(s, a))
\end{equation}
Сэмплирование переходов из буфера происходит по следующему правилу:
$$\Prob ( \T ) \propto \rho( \T )^\alpha$$
где гиперпараметр $\alpha > 0$ контролирует масштаб приоритетов (в частности, $\alpha = 0$ соответствует равномерному сэмплированию, когда $\alpha \to +\infty$ соответствовало бы жадному сэмплированию самых <<важных>> переходов).

\begin{remark}
Добиться эффективного сэмплирования с приоритетами можно благодаря структуре данных SumTree: бинарному дереву, у которого в каждом узле хранится сумма значений в двух детях. Сам массив приоритетов для буфера хранится на нижнем уровне дерева; в корне, соответственно, лежит сумма всех приоритетов, нормировочная константа. Для сэмплирования достаточно взять случайную равномерную величину и спуститься по дереву. Таким образом, процедура сэмплирования имеет сложность $O(\log M)$, где $M$ --- размер буфера. За ту же сложность проходит обновление приоритета одного элемента, для чего достаточно обновить значения во всех его предках для поддержания структуры.
\end{remark}

Техническая проблема идеи \eqref{priority} заключается в том, что после каждого обновления весов сети $\theta$ приоритеты переходов меняются для всего буфера (состоящего обычно из миллионов переходов). Пересчитывать все приоритеты, конечно же, непрактично, и необходимо ввести некоторые упрощения. Например, можно обновлять приоритеты только у переходов из текущего батча, для которых значение лосса так и так считается. Это, вообще говоря, означает, что если у перехода был низкий приоритет, и до него дошла, условно, <<волна распространения>> награды, алгоритм не узнает об этом, пока не засэмплирует переход с тем приоритетом, который у него был.

\begin{remark}
Новые переходы добавляются в буфер с наивысшим приоритетом $\max\limits_{\T } \rho(\T )$, который можно поддерживать за константу, или вычислять текущий приоритет, дополнительно рассчитывая \eqref{priority} для онлайн-сэмплов.
\end{remark}

В чём у такого подхода есть фундаментальная проблема? После неконтролируемой замены равномерного сэмплирования на какое-то другое, могло случиться так, что для наших переходов $s' \not\sim p(s' \mid s, a)$. Почему это так, проще понять на примере.

\begin{example}
Пусть для данной пары $s, a$ с вероятностью 0.9 мы попадаем в $s' \HM= A$, а с вероятностью 0.1 мы попадаем в $s' \HM= B$. В условно бесконечном буфере для этой пары $s, a$ среди каждых 10 сэмплов будет 1 сэмпл с $s' \HM= B$ и 9 сэмплов с $s' \HM= A$, и равномерное сэмплирование давало бы переходы, удовлетворяющие $s' \sim p(s' \mid s, a)$. 

\needspace{7\baselineskip}
\begin{wrapfigure}{r}{0.15\textwidth}
\vspace{-0.3cm}
\centering
\includegraphics[width=0.15\textwidth]{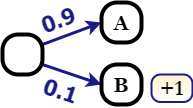}
\end{wrapfigure}

Для приоритизированного реплея, веса у переходов с $s' \HM= A$ могут отличаться от весов для переходов с $s' \HM= B$. Например, если мы оцениваем $V(A) \HM= 0, V(B) \HM= 1$, и уже даже правильно выучили среднее значение $Q(s, a) \HM= 0.1$, то $\Loss(s, a, s'=A)$ будет равен $0.1^2$, а для $\Loss(s, a, s'=B) = 0.9^2$. Значит, $s' = B$ будет появляться в засэмплированных переходах чаще, чем с вероятностью 0.1, и это выбьет $Q(s, a)$ с её правильного значения.
\end{example}

Иными словами, приоритизированное сэмплирование приводит к \emph{смещению} (bias). Этот эффект не так страшен поначалу обучения, когда распределение, из которого приходят состояния, всё равно скорее всего не сильно разнообразно. Более существенно нивелировать этот эффект по ходу обучения, в противном случае процесс обучения может полностью дестабилизироваться или где-нибудь застрять.

Заметим, что равномерное сэмплирование не является единственным <<корректным>> способом, но основным доступным. Мы не очень хотим <<возвращаться>> к нему постепенно с ходом обучения, но можем сделать похожую вещь: раз мы хотим подменить распределение, то можем при помощи importance sampling сохранить тот же оптимизируемый функционал:

\begin{theorem}
При сэмплировании с приоритетами $\Prob( \T )$ использование весов $w(\T) \coloneqq \frac{1}{\Prob (\T )}$ позволит избежать эффекта смещения.
\begin{proof} Пусть $M$ --- размер буфера.
\begin{align*}
\mathbb{E}_{\T \sim \mathop{Uniform}} \Loss(\T ) &= \sum_{i=1}^M \frac{1}{M} \Loss(\T_i) = \\ 
&= \sum_{i=1}^M \Prob (\T_i) \frac{1}{M\Prob (\T_i)} \Loss(\T_i) = \\
&= \mathbb{E}_{\T \sim \Prob (\T)} \frac{1}{M\Prob (\T )} \Loss(\T),
\end{align*}
что с точностью до константы $\frac{1}{M}$ и есть перевзвешивание функции потерь.
\end{proof}
\end{theorem}

Importance sampling подразумевает, что мы берём <<интересные>> переходы, но делаем по ним меньшие шаги (вес меньше именно для <<приоритетных>> переходов). Цена за такую корректировку, конечно, в том, что полезность приоритизированного сэмплирования понижается. Раз поначалу смещение нас не так беспокоит, предлагается вводить веса постепенно: а именно, использовать веса
$$w(\T) \coloneqq \frac{1}{\Prob (\T )^{\beta(t)}}$$
где $\beta(t)$ --- гиперпараметр, зависящий от итерации алгоритма $t$. Изначально $\beta(t=0) = 0$, что делает веса равномерными (корректировки не производится), но постепенно $\beta(t)$ растёт к 1 и полностью избавляет алгоритм от эффекта смещения.

\begin{remark}
На практике веса, посчитанные по такой формуле, могут оказаться очень маленькими или большими, и их следует нормировать. Вариации, как нормировать, различаются в реализациях: можно делить веса на $\max\limits_\T w(\T)$, где максимум берётся, например, только по текущему мини-батчу, чтобы гарантировать максимальный вес 1.
\end{remark}

\subsection{Multi-step DQN}\label{subsec:multistepdqn}

Уже упоминалось, что DQN из-за одношаговых целевых переменных страдает от проблемы отложенного сигнала и сопряжённой с ней в контексте нейросетевой аппроксимации проблемы накапливающейся ошибки. Эта проблема фундаментальна для off-policy подхода: в разделе \ref{sec:biasvar} про bias-variance trade-off упоминалось, что разрешать дилемму смещения-разброса (то есть <<проводить умный credit assingment>>) мы можем только в on-policy режиме. 

\emph{Многошаговый} (multi-step) DQN --- теоретически некорректная эвристика для занижения этого эффекта. Грубо говоря, нам очень хочется распространять за одну итерацию награду сразу на несколько шагов вперёд, то есть решать многошаговые уравнения Беллмана \eqref{NstepBellman}. Мы как бы и можем уравнение оптимальности многошаговое выписать...

\begin{proposition}[$N$-шаговое уравнение оптимальности Беллмана]
\begin{equation}\label{NstepOptimBellman}
 Q^*(s_0, a_0) = \E_{\Traj_{:N} \sim \pi^* \mid s_0, a_0} \left[ \sum_{t=0}^{N-1} \gamma^{t}r_t + \gamma^N \E_{s_N} \max_{a^*_N} Q^*(s_N, a^*_N) \right]   
\end{equation}
\end{proposition}

Что мы можем сделать? Мы можем прикинуться, будто решаем многошаговые уравнения Беллмана, задав целевую переменную следующим образом:
\begin{equation}\label{Nsteptarget}
y(s_0, a_0) \coloneqq \sum_{t=0}^{N-1} \gamma^{t}r_t + \gamma^N \max_{a_N} Q_{\theta}(s_N, a_N)
\end{equation}
где $s_1, a_1 \dots a_{N-1}, s_N$ взяты из буфера. Для этого в буфере вместо одношаговых переходов $\T \HM\coloneqq (s, a, r, s', \done)$ достаточно просто хранить другую пятёрку:
$$\T \coloneqq \left( s, \, a, \, \sum_{n=0}^{N-1} \gamma^{n}r^{(n)}, \, s^{(N)}, \, \done \right)$$
где $r^{(n)}$ --- награда, полученная через $n$ шагов после посещения рассматриваемого состояния $s$, $s^{(N)}$ --- состояние, посещённое через $N$ шагов, и, что важно, флаг $\done$ указывает на то, завершился ли эпизод в течение этого $N$-шагового роллаута\footnote{естественно, алгоритм должен рассматривать все $N$-шаговые роллауты, включая те, которые привели к завершению эпизода за $k \HM< N$ шагов. Для них, естественно, $r^{(k')} \HM= 0$ для $k' \HM> k$, и $Q^*(s^{(N)}, a_N) \equiv 0$ для всех $a_N$.}. Все остальные элементы алгоритма не изменяются, в частности, можно видеть, что случай $N \HM= 1$ соответствует обычному DQN.

Видно, что теперь награда, полученная за один шаг, распространяется на $N$ состояний в прошлое, и мы таким образом не только ускоряем обучение оценочной функции стартовых состояний, но и нивелируем проблему накапливающейся ошибки. 

Почему теоретически это некорректно? Беря $s_1, a_1 \dots a_{N-1}, s_N$ из буфера, мы получаем состояния из функции переходов, которая стационарна и соответствует тому мат.ожиданию, которое стоит в уравнении \eqref{NstepOptimBellman}. Но вот действия в этом мат.ожидании должны приходить из оптимальной стратегии! А в буфере $a_1, a_2 \dots a_{N-1}$ --- действия нашей стратегии произвольной давности (то есть сколь угодно неоптимальные). Вместо того, чтобы оценивать оптимальное поведение за хвост траектории (по определению $Q^*$), мы $N \HM- 1$ шагов ведём себя сколько угодно неоптимально, а затем в $s^{(N)}$ подставляем оценку оптимального поведения за хвост. Иными словами, мы недооцениваем истинное значение правой части $N$-шагового уравнения Беллмана при $N \HM> 1$. Вместо уравнения оптимальности мы решаем такое уравнение: что, если я следующие $N$ шагов веду себя как стратегия $\mu$, когда-то породившая данный роллаут, и только потом соберусь вести себя оптимально? Причём из-за нашего желания делать так в off-policy режиме $\mu$ для каждого перехода своё, то есть схеме Generalized Policy Iteration (алг. \ref{generalizedpolicyiteration}) это не соответствует: в ней мы всегда должны оценивать именно текущую стратегию $\pi$, а текущей стратегией в DQN является $\pi(s) \HM= \argmax\limits_{a} Q_{\theta}(s, a)$.

\begin{exampleBox}[righthand ratio=0.2, sidebyside, sidebyside align=center, lower separated=false]{}
Пример, когда многошаговая оценка приводит к некорректным апдейтам. На втором шаге игры в $s_1$ агент может скушать тортик или прыгнуть в лаву, и в первом эпизоде обучения агент совершил ошибку и получил огромную негативную награду. В буфер при $N \HM> 1$ запишется пример со стартовым состоянием $s_0$ и этой большой отрицательной наградой (в качестве $s^{(N)}$ будет записана лава). Пока этот пример живёт в реплей буфере, каждый раз, когда он сэмплируется в мини-батче, оценочная функция для $s_0$ обновляется этой отрицательной наградой, даже если агент уже научился больше не совершать эту ошибку и в $s_1$ наслаждается тортиками.

\tcblower
\vspace{-0.3cm}
\begin{adjustwidth}{-0.6cm}{}
\includegraphics[width=1.15\textwidth]{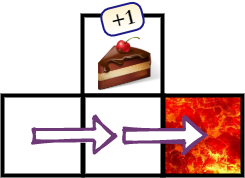} \hspace*{-1cm}
\end{adjustwidth}
\end{exampleBox}

\begin{remark}
Эмпирически большое значение $N$ действительно может полностью дестабилизировать процесс, как и подсказывает теория, поэтому рекомендуется выставлять небольшие значение 2-3, от силы 5. 
\end{remark}

Большие значения могут быть работоспособны в средах, где сколь угодно неоптимальное поведение в течение $N$ шагов не приводит к существенному изменению награды по сравнению с оптимальным поведением, то есть в средах, где нет моментов с <<критическими решениями>> (когда $\max\limits_a Q^*(s, a) \HM- \min\limits_a Q^*(s, a)$ мало, то есть неоптимальное поведение в течение одного шага не приводит к сильно меньшей награде, чем оптимальное).

\begin{example}
Пример среды без <<критических решений>>: вы робот, который хочет добраться до соседней комнаты. Действия вверх-вниз-вправо-влево чуть-чуть сдвигают робота в пространстве. Тогда <<вести себя как угодно>> в течение $N \HM- 1$ шагов и потом отправиться кратчайшим маршрутом до соседней комнаты приносит практически столько же награды, сколько и сразу отправиться кратчайшим маршрутом до соседней комнаты. Поэтому в таких ситуациях использование относительно большого $N$ (5-10) может помочь, хоть алгоритм и может полностью дестабилизироваться (процедура некорректна).
\end{example}

\subsection{Retrace}

Как мы обсуждали в разделе \ref{subsec:retrace}, теоретически корректным способом обучаться в off-policy с многошаговых оценок является использование Retrace оценки. Конечно, она может на практике схлопываться в одношаговые обновления, но по крайней мере гарантирует, что алгоритм не ломается; и важно, что если записанные в засэмплированном из буфера роллауте действия достаточно вероятны для оцениваемой политики, то оценка получается достаточно длинной.

Конечно, сложно говорить про <<достаточно вероятны>>, когда оцениваемая политика детерминирована. Поэтому в практическом алгоритме Retrace предлагается перейти от моделирования Q-learning к моделированию SARSA (см. раздел \ref{subsec:sarsa}): то есть, считать целевой политикой $\pi(a \HM \mid s)$ $\eps$-жадную стратегию по отношению к текущей модели Q-функции. Преимущество в том, что это делает стратегию стохастичной, и любые действия в буфере не приведут к занулению следа и полному схлопыванию в одношаговую оценку.

В буфере также нужно сохранять вероятности выбора сохранённых действий $\mu(a \HM\mid s)$ в момент сбора данных (для $\eps$-жадных стратегий эти значения всё время будут или $\frac{\eps}{|\A|}$, или $1 \HM- \eps \HM+ \frac{\eps}{|\A|}$, где $\eps$ --- параметр эксплорейшна на момент сбора перехода). Вместо отдельных переходов теперь хранятся роллауты --- фрагменты траекторий некоторой длины $N$.

\begin{remark}
Если в DQN обучение проводилось на мини-батчах из, скажем, 64 переходов, то в Retrace (при том же масштабе задачи) нужно засэмплировать для одного мини-батча 4 роллаута длины $N\HM=16$. Такой выбор позволит надеяться на получение оценок длины вплоть до 16-шаговой (что в Retrace будет достигнуто, если политика сбора данных совпадёт с оцениваемой политикой на оцениваемом роллауте). Важно помнить про проблему декорреляции: нужно, чтобы в мини-батче должны оказаться разнообразные примеры, поэтому нельзя, например, взять только один роллаут длины 64.  
\end{remark}

Далее для каждого засэмплированного из буфера роллаута $s_0, a_0, r_0, s_1, a_1, r_1, \dots s_N$ мы сначала для каждой пары $s_t, a_t$ считаем, используя формулы из теории Retrace, следующие вспомогательные величины: значение коэффициента затухания следа $c_t$ по формуле \eqref{retracecoeff} (коэффициент $\lambda$ обычно полагают равным единице) и значение одношаговой ошибки $\Psi_{(1)}(s_t, a_t)$ по формуле \eqref{onestepoffpolicyQdelta}:
$$c_t \coloneqq \min \left( 1, \frac{\pi(a_t \mid s_t) }{\mu(a_t \mid s_t)} \right) $$
$$\Psi_{(1)}(s_t, a_t) = r_t + \gamma \E_{\hat{a}_{t+1} \sim \pi} Q_{\theta^{-}}(s_{t+1}, \hat{a}_{t+1}) - Q_{\theta^{-}}(s_t, a_t)$$

Всюду, где используется $\pi$, используется $\eps$-жадная стратегии по отношению к таргет-сети (хотя, если использовать идею Double DQN из раздела \ref{subsec:doubledqn}, то как раз во всех местах, где используется $\pi$ --- оцениваемая стратегия --- имеет смысл использовать свежую версию Q-функции). В частности, мат.ожидание по $\pi$ можно посчитать явно:
$$\E_{\hat{a} \sim \pi} Q_{\theta^{-}}(s, \hat{a}) = (1 - \eps) \max_{\hat{a}} Q_{\theta^{-}}(s, \hat{a}) + \frac{\eps}{|\A|} \sum_{\hat{a}} Q_{\theta^{-}}(s, \hat{a}) $$

После этого в Retrace для одной пары $s_t, a_t$ все будущие одношаговые ошибки нужно просуммировать (воспользуемся индексом $\hat{t}$ для обозначения этого перебора), но заглядывание на каждый следующий $i$-ый шаг в будущее обязывает нас потушить след в $c_i$ раз: 
$$\Psi^{\mathrm{retrace}}(s_t, a_t) \coloneqq \sum_{\hat{t} \ge t}^N \gamma^{\hat{t} - t}  \left( \prod_{i = t+1}^{i = \hat{t}} c_{i} \right) \Psi_{(1)}(s_{\hat{t}}, a_{\hat{t}})$$

Заметим, что в этой формуле внешняя сумма по $\hat{t}$ идёт не до бесконечности, как в теории Retrace, а до $N$, до конца роллаута. После этого считаем, что след зануляется: это корректно, хотя иногда можно и потерять возможность получить более длинную оценку.

Далее в табличном методе мы бы провели обновление по формуле
$$Q(s, a) \leftarrow Q(s, a) + \alpha \Psi^{\mathrm{retrace}}(s, a),$$
то есть воспользовались бы $\Psi^{\mathrm{retrace}}(s, a)$ как градиентом. Другими словами, оценка указывает, нужно ли увеличивать выход модели для рассматриваемой пары $s, a$ или уменьшать. Чтобы получить задачу регрессии, целевая переменная строится по формуле
$$y(s_t, a_t) \coloneqq \Psi^{\mathrm{retrace}}(s_t, a_t) + Q_{\theta}(s_t, a_t)$$
и дальше оптимизируется MSE, игнорируя зависимость $y$ от $\theta$:
$$\left( y(s_t, a_t) - Q_{\theta}(s_t, a_t) \right)^2 \to \min_{\theta}$$

Тогда градиент функции потерь по $\theta$ для одного примера равен:
$$\nabla_\theta \frac{1}{2}\left( y(s, a) - Q_{\theta}(s, a) \right)^2 = \left( y(s, a) - Q_{\theta}(s, a) \right) \nabla_\theta Q_{\theta}(s, a) = \Psi^{\mathrm{retrace}}(s, a) \nabla_\theta Q_{\theta}(s, a)$$

Это полностью аналогично градиенту обычного DQN \eqref{DQNgradient}, только там оценка $\Psi(s, a) \HM= r + \gamma \max\limits_{a'} Q_{\theta^{-}}(s, a) - Q_{\theta}(s, a)$ была одношаговой, а здесь мы заглядываем настолько максимально далеко вперёд, насколько возможно (в силу использования $\lambda \HM= 1$). 

\begin{remark}
В большинстве последних алгоритмов на основе DQN, таких как Agent57, используются формулы Retrace. Они позволяют максимально возможным образом побороться с ключевыми фундаментальными проблемами off-policy подхода, вытекающими из одношаговых целевых переменных, когда из недостатков можно выделить, пожалуй, лишь громоздкость формул.
\end{remark}

\section{Distributional RL}

\subsection{Идея Distributional подхода}

\needspace{7\baselineskip}
\begin{wrapfigure}{r}{0.25\textwidth}
\vspace{-0.3cm}
\centering
\includegraphics[width=0.25\textwidth]{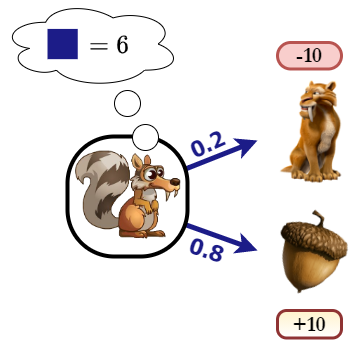}
\vspace{-0.6cm}
\end{wrapfigure}

Задача RL такова, что в среде содержится в том числе неподконтрольная агенту стохастика: \emph{алеаторическая неопределённость} (aleatoric uncertainty)\footnote{которую не стоит путать с \emph{эпистемической неопределённостью} (epistemic uncertainty), или байесовской неопределённостью, связанной с нашим субъективным незнанием о том, как устроена, например, функция переходов и награды на самом деле, связанной с нашим осознанием неточности прогнозов.}. Агент, предсказывающий, что он получит в будущем в среднем суммарную награду 6, на самом деле может получить, например, только -10 или 10, просто последний исход случится с вероятностью 0.8, а первый --- 0.2. Помимо прочего, это означает, что часто агенту приходится рисковать: например, теоретически возможна ситуация, когда агент с малой вероятностью получает гигантскую награду, и тогда оптимальный агент на практике будет постоянно получать, например, какой-то штраф, компенсирующийся редко выпадающими мегаудачами. Вся эта информация заложена в распределении награды $R(\Traj)$ \eqref{return} как случайной величины.

В Distributional-подходе предлагается учить не среднее будущей награды, а всё распределение будущей награды как случайной величины. Складывается эта неопределённость как из подконтрольной агенту стохастики --- его собственных будущих выборов действий --- так и неподконтрольной, переходов (и награды, если рассматривается формализм со случайной функцией награды). Среднее есть лишь одна из статистик этого распределения. 

\needspace{7\baselineskip}
\begin{wrapfigure}{l}{0.3\textwidth}
\vspace{-0.3cm}
\centering
\includegraphics[width=0.3\textwidth]{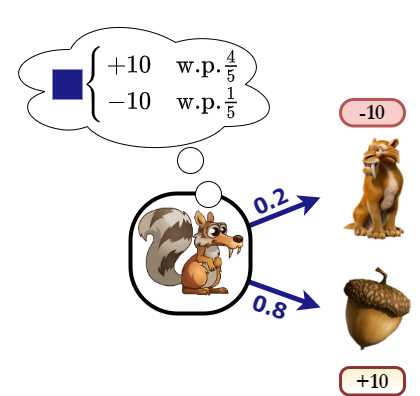}
\vspace{-0.3cm}
\end{wrapfigure}

Здесь стоит заранее оговориться о противоречиях, связанных с этой идеей. Обсуждение этой темы в первую очередь мотивировано эмпирическим превосходством Distributional-подхода по сравнению с алгоритмами, учащими только среднее, однако с теоретической точки зрения ясного обоснования этого эффекта нет. Даже наоборот: мы далее встретим теоретические результаты, показывающие эквивалентность Distributional-алгоритмов с обычными в рамках табличного сеттинга. 

Одно гипотетическое объяснение преимущества distributional-подхода в нейросетевом сеттинге, когда будущая награда предсказывается сложной параметрической моделью, может быть следующая: обучая модель предсказывать не только среднее награды, но и другие величины (другие статистики), сильно связанные по смыслу со средней наградой, в модель отправляются более информативные градиенты. Например, если с вероятностью 0.01 во входном изображении появляется грибочек, намекающий на приближение вкусного +100, обновление среднего будет, абстрактно говоря, проходить с учётом малой вероятности явления; когда обучение 1\%-го квантиля наоборот крайне интересуется именно хвостом распределения и поможет быстрее выучить, например, фильтр для детекции грибочка в первых свёрточных слоях, который, в свою очередь, ускорит и более точное вычисление среднего, для которого распознавание грибочка на самом деле было существенным. Вот ещё один пример, почему это может быть хорошо:

\begin{example}
Допустим, состояние описывается картинкой. Если вы видите на картинке орешек, то получите +1. Если тигра --- то -1. Если и тигра, и орешек, то вам может повезти, а может не повезти, и вы с вероятностью 0.5 получите +1, а с вероятностью 0.5 получите -1. В такой ситуации модель, предсказывающая среднюю будущую награду, должна выдавать 0 --- тоже самое значение, что и для пустой картинки, где никаких объектов нет.

\begin{center}
\vspace{0.2cm}
    \includegraphics[width=0.7\textwidth]{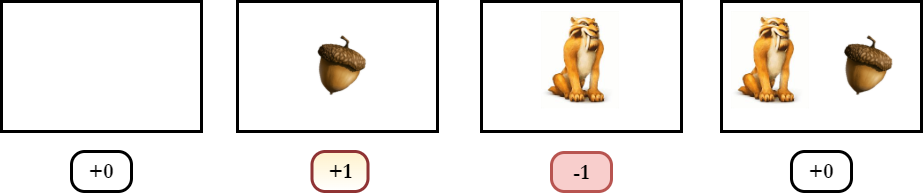}
\vspace{-0.2cm}
\end{center}

Было бы здорово в таких ситуациях в последнем слое модели увидеть что-то вроде признаков, отвечающих на вопрос <<есть ли на картинке орешек?>> или <<есть ли на картинке тигр?>>. Конечно, такой разметки нам никто не предоставит, и обучать такой слой напрямую не получится. Так вот, забавное наблюдение заключается в том, что распределение награды в таких ситуациях содержит информацию о том, <<какие объекты>> есть на картинке: если целевая переменная для пустой картинке будет <<0 с вероятностью 1>>, а для картинки с тигром и орешком <<$\pm 1$ с вероятностями 0.5>>, то модель научится различать такие ситуации, и скорее всего будет быстрее обучаться распознавать тигров и орешки. 
\end{example}

\begin{remark}
Ещё одним способом придумать для модели целевую переменную, предсказание которой сильно связано со средней наградой, является использование нескольких дискаунт-факторов $0 \HM< \gamma_1 \HM< \gamma_2 \HM< \dots \HM< \gamma_G \HM< \gamma$, где $\gamma$ --- коэффициент, для которого решается задача. Иначе говоря, нейросетевая модель выдаёт для каждой пары состояние-действие не одно, а $G+1$ число, соответствующее $Q$-функции для соответствующего коэффициента дисконтирования:
$$Q^\pi(s, a, g) \coloneqq \E_{\Traj \mid s_0 = s, a_0 = a} \sum_{t \ge 0} \gamma_g^t r_t, \quad g \in \{1 \dots G \}$$
Иными словами, мы дополнительно учим, какую награду мы получим в самое ближайшее время, более близкое, чем реальный рассматриваемый горизонт. Эти дополнительные величины самой стратегией использоваться не будут (взаимодействие со средой использует только значения $Q$ для настоящего коэффициента $\gamma$), однако их обучение поможет <<ускорить>> обучение для настоящей $Q$ с самым большим горизонтом. Мы вернёмся к обобщению этой идеи, когда будем обсуждать multi-task RL в разделе \ref{sec:multitask}.
\end{remark}

\subsection{Z-функция}

\begin{definition} 
Для данного MDP \emph{оценочной функцией в distributional форме} (distributional state-action value function) для стратегии $\pi$ называется случайная величина, обусловленная на пару состояние-действие $s, a$ и определяющаяся как reward-to-go для такого старта:
$$\Z^\pi(s, a) \cdfcoloneqq R(\Traj), \qquad \Traj \sim \pi \mathrel{\Big|} \, s_0 = s, a_0 = a $$
\end{definition}

Здесь читателю предлагается заварить себе кофе на том факте, что введённая так называемая Z-функция\footnote{слово <<функция>> здесь, конечно, не очень удачно, однако автор данного текста не справился с нахождением альтернатив. <<Величина>> ещё можно было бы.} является для каждой пары $s, a$ случайной величиной. Во-первых, это скалярная случайная величина, соответственно, она задаётся некоторым распределением на $\R$, во-вторых, как и для любой случайной величины, существенно, на что она обуславливается. Запись $\Z^\pi(s, a)$ предполагает, что мы сидим в состоянии $s$ и выполнили действие $a$, после чего <<бросаем кости>> для сэмплирования случайной величины; нас, вообще говоря, будет интересовать её \emph{функция распределения} (cumulative distribution function, c.~d.~f.):
$$F_{\Z^\pi(s, a)}(x) \coloneqq \Prob (\Z^\pi(s, a) \le x)$$
Нам будет удобнее работать с ними, а не с плотностями, поскольку зачастую распределение $\Z^\pi(s, a)$ --- дискретное или вообще вырожденное (принимающее с вероятностью 1 только какое-то одно значение). Таким образом, пространство всевозможных Z-функций имеет такой вид:
\begin{equation}\label{Zfunctionspace}
\Z^\pi(s, a) \in \St \times \A \to \Prob (\R),
\end{equation}
где $\Prob (\R)$ --- пространство скалярных случайных величин. 

Надпись c.d.f. над равенством здесь и далее означает, что слева и справа стоят случайные величины. Очень важно, что случайные величины справа и слева в подобных равенствах обусловлены на одну и ту же информацию: справа, как и слева, стоит случайная величина, обусловленная на $s, a$. Случайная величина здесь задана процессом генерации: сначала генерируется случайная траектория $\Traj$ при заданных $s_0 \HM= s, a_0 \HM= a$ (это по определению MDP эквивалентно последовательному сэмплированию $s_1, a_1, s_2 \dots$), затем от этой случайной величины считается детерминированная функция $R(\Traj)$. Запись $\cdfcoloneqq$ означает, что $\Z^\pi(s, a)$ имеет в точности то же распределение, что и случайная величина, генерируемая процессом, описанном справа.

По определению:

\begin{proposition}
\begin{equation}\label{QZ}
Q^\pi(s, a) = \E \Z^\pi(s, a)
\end{equation}
\end{proposition}

\begin{proposition}
В терминальных состояниях для всех действий $\Z^{\pi}(s, a)$ есть вырожденная случайная величина, с вероятностью 1 принимающая значение ноль.
\end{proposition}

\begin{example}
Допустим, мы сидим в состоянии и выполнили действие \colorsquare{ChadBlue}. Как будет выглядеть $\Z^{\pi}(s, \colorsquare{ChadBlue})$ для MDP и стратегии $\pi$ с рисунка, $\gamma \HM= 0.5$?

\begin{wrapfigure}{r}{0.4\textwidth}
\centering
\includegraphics[width=0.4\textwidth]{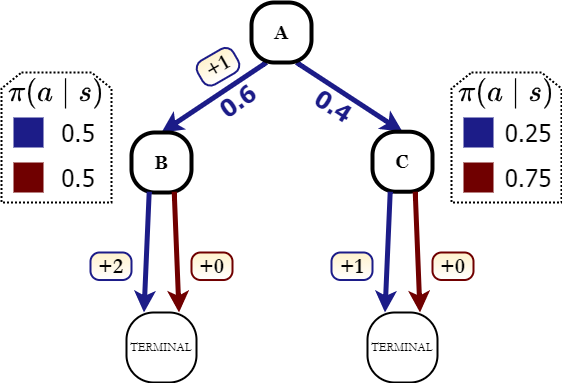}
\end{wrapfigure}

Нас ждёт два источника случайности: сначала среда кинет нас или в состояние B, или в состояние C, затем мы случайно будем определять своё следующее действие. Всего нас ждёт 4 возможных исхода. Для каждого мы можем посчитать его вероятность и получаемый reward-to-go. Итого $\Z^{\pi}(s, \colorsquare{ChadBlue})$ --- дискретное распределение с 4 исходами:

\vspace{0.2cm}
\begin{adjustwidth}{0.7cm}{0cm}
\begin{tabular}{cccc}
\toprule
    Исход $s'$ & Исход $a'$ & Вероятность & reward-to-go \\
\midrule
    B & \colorsquare{ChadBlue} & 0.3 & $1 + 2\gamma$ \\
    B & \colorsquare{ChadRed} & 0.3 & $1$ \\
    C & \colorsquare{ChadBlue} & 0.1 & $\gamma$ \\
    C & \colorsquare{ChadRed} & 0.3 & $0$ \\
\bottomrule
\end{tabular}
\end{adjustwidth}
\vspace{0.2cm}
\end{example}

\needspace{16\baselineskip}
\begin{exampleBox}[label=ex:zfunction]{}
Посчитаем Z-функцию для MDP и стратегии $\pi$ с рисунка, $\gamma \HM= 0.8$.

\begin{wrapfigure}{r}{0.45\textwidth}
\vspace{-0.4cm}
\centering
\includegraphics[width=0.45\textwidth]{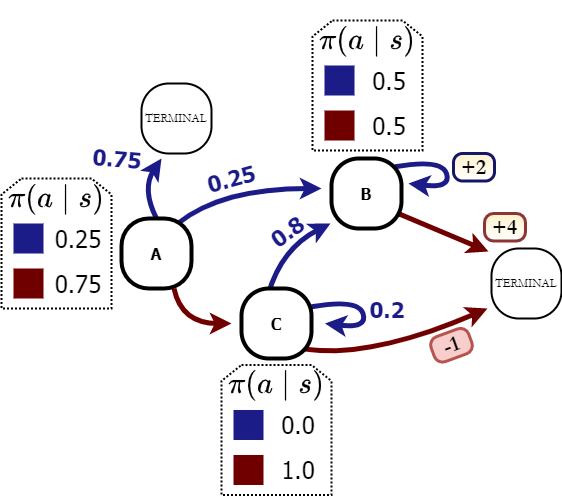}
\vspace{-0.4cm}
\end{wrapfigure}

Начнём с состояния B: если агент выбирает действие \colorsquare{ChadRed}, то он получает +4, и эпизод заканчивается. Значит, $\Z^{\pi}(s \HM= B, \colorsquare{ChadRed})$ всегда принимает значение 4.

Что произойдёт, если он выберет \colorsquare{ChadBlue}? Агент точно получит +2 и вернётся в состояние B. Вся дальнейшая награда будет дисконтирована на $\gamma \HM= 0.8$. После этого начинается первая стохастика: агент снова будет выбирать действие! С вероятностью 0.5 он выберет \colorsquare{ChadRed} и получит итоговый reward-to-go $2 \HM+ \gamma \cdot 4 \HM= 5.2$ за эпизод. Вот мы нашли часть нашей Z-функции для $s \HM= B$, $a \HM= \colorsquare{ChadBlue}$: с вероятностью 0.5 исход будет 5.2. Ищем, что соответствует оставшейся вероятностной массе 0.5: мы выберем снова \colorsquare{ChadBlue}, получим уже суммарно $2 \HM+ \gamma \cdot 2 \HM= 3.6$, снова вернёмся в B, снова случится дисконтирование и снова за нами будет выбор. Мы можем найти, что соответствует ещё 0.25 нашей вероятностной массы: $2 \HM+ \gamma \cdot 2 + \gamma^2 \cdot 4 \HM= 6.16$. Дальше в этом распределении будет ещё исход с вероятностью $\frac{1}{8}$, затем $\frac{1}{16}$ и так далее: $\Z^{\pi}(s \HM= B, \colorsquare{ChadBlue})$ есть распределение со счётным множеством исходов!

Очевидно, что $\Z^{\pi}(s \HM= A, \text{\colorsquare{ChadRed}})$ и $\Z^{\pi}(s \HM= C, \text{\colorsquare{ChadRed}})$ тоже будут вырожденными: reward-to-go для таких стартов однозначно определён и равен -0.8 и -1 соответственно. 

$\Z^{\pi}(s \HM= A, \text{\colorsquare{ChadBlue}})$ содержит компоненту <<с вероятностью 0.75 мы попадём в терминальное состояние и получим 0>>. Оставшиеся 0.25 соответствуют попаданию в B и распределяются пополам между выбором следующего действия \colorsquare{ChadRed} (соответствует награде 3.2) и \colorsquare{ChadBlue} (тогда начнётся та же цепочка исходов, домноженных на $\gamma \HM= 0.8$, что и для $\Z^{\pi}(s \HM= B, \colorsquare{ChadBlue})$). Аналогично можно расписать $\Z^{\pi}(s \HM= C, \text{\colorsquare{ChadBlue}})$; как видно, такая <<оценочная функция>> содержит в себе намного больше информации о будущих наградах.
\end{exampleBox}

\subsection{Distributional-форма уравнения Беллмана}

Заметим, что в доказательстве уравнений Беллмана, например, \eqref{VV}, мы ссылаемся на то, что для reward-to-go любых траекторий верно рекурсивное соотношение. После этого мы берём по траекториям мат.ожидание слева и справа, получая традиционное уравнение Беллмана. Ясно, что мы могли бы вместо среднего взять любую другую статистику от случайной величины (дисперсию, медианы, другие квантили...), а, вообще говоря, верно совпадение левой и правой части по распределению. Иначе говоря, можно зачеркнуть символ мат.ожидания из уравнения Беллмана для получения более общего утверждения.

\begin{theorem}[Уравнение Беллмана в Distributional-форме]\,
\begin{equation}\label{ZZ}
 \Z^\pi(s, a) \cdfeq r(s, a) + \gamma \Z^\pi(s', a'), \quad s' \sim p(s' \mid s, a), a' \sim \pi(a' \mid s') 
\end{equation}
\begin{proof}
Следует из тождества $R(\Traj) = r + \gamma R(\Traj_{1:})$ для любых траекторий $\Traj$.
\end{proof}
\end{theorem}

Читатель подозревается в недоумении от происходящего; остановимся на этом уравнении и обсудим, что тут написано. Во-первых, необходимо пояснить, что данное уравнение есть переформулировка (другая нотация) используемых определений. Reward-to-go $R(\Traj)$ --- детерминированная функция от заданной траектории $\Traj$, $\Z^\pi$ --- по сути тоже самое, только траектория рассматривается как случайная величина (а параметры $s, a$ указывают на начальные условия генерации траекторий). И слева, и справа в уравнении \eqref{ZZ} стоят случайные величины, зависящие от $s, a$; равенство означает, что они имеют одинаковые распределения. Иными словами, слева и справа записаны два процесса генерации одной и той же случайной величины. Мы можем бросить кость $\Z^\pi(s, a)$ (случайная величина слева), а можем --- сначала $s'$, потом $a'$, затем $\Z^\pi(s', a')$ и выдать исход\footnote{если в формализме функция награды также случайна, вместо сдвига на константу здесь было бы сложение двух случайных величин; для расчёта распределения тогда теоретически рассматривалась бы операция свёртки.} $r(s, a) \HM+ \gamma \Z^\pi(s', a')$ (случайная величина справа), и эти две процедуры порождения эквивалентны. 

\begin{exampleBox}[righthand ratio=0.55, sidebyside, sidebyside align=center, lower separated=false]{}
Уравнение Беллмана всё ещё связывает <<содержимое>> Z-функции через неё же саму, раскрывая дерево на один шаг. Эти уравнения теперь затруднительно выписать аналитически, поскольку теперь <<компоненты>> распределения $\Z(s, a)$ есть перевзвешанные на вероятности переходов и выборов действий (и подправленные по значению на дискаунт фактор и смещённые на награду за шаг) $\Z(s', a')$ для всевозможных $s', a'$.

\tcblower
\begin{adjustwidth}{-0.85cm}{}
\includegraphics[width=1.1\textwidth]{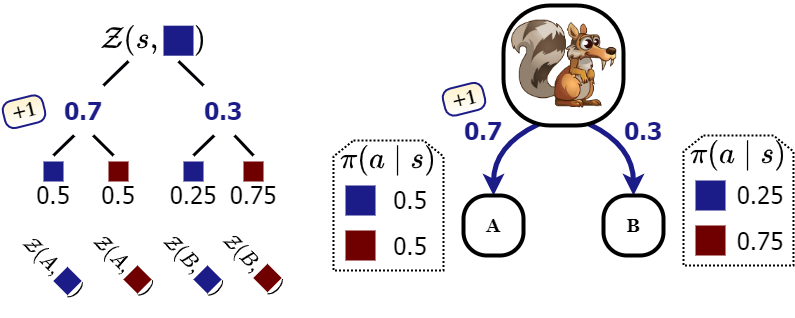}
\end{adjustwidth}
\end{exampleBox}

Подобные уравнения называются \emph{recursive distributional equations} и рассматриваются математикой в одном из разделов теории вероятности. Нам далее не понадобится какая-то особая теория оттуда, однако для вдохновения рассмотрим каноничный местный пример. 
\begin{example}
Пусть
$$X_1 \cdfeq \frac{X_2}{\sqrt{2}} + \frac{X_3}{\sqrt{2}}$$
где $X_1, X_2, X_3$ --- независимые случайные величины из одного распределения $p(x)$. То есть заданы две процедуры порождения: мы можем взять сэмпл из распределения, а может взять два сэмпла из распределения (независимо), отмасштабировать на корень из двух и сложить. Уравнение заявляет, что эти две процедуры эквивалентны. Вопрос для математики такой: каким могло бы быть $p(x)$? Несколько ответов можно угадать: например, ответом является, детерминированный ноль или $\N(0, \sigma^2)$.
\end{example}

\subsection{Distributional Policy Evaluation}

Будем строить аналог Policy Evaluation для distributional-формы оценочной функции. Иными словами, мы хотим чисто теоретический алгоритм, позволяющий для данного MDP и данной стратегии $\pi$ посчитать распределение $\Z^\pi(s, a)$. MDP пока считаем полностью известным (распределение $p(s' \mid s, a)$ считаем данным). Действуем в полной аналогии с обычными уравнениями: начнём с ввода оператора Беллмана.

\begin{definition}
Для данного MDP и стратегии $\pi$ будем называть \emph{оператором Беллмана в distributional форме} оператор $\B_D$, действующий из пространства Z-функций \eqref{Zfunctionspace} в пространство Z-функций, задающий случайную величину для $s, a$ на выходе оператора как правую часть distributional уравнения Беллмана \eqref{ZZ}:
$$\left[\B_D \Z\right](s, a) \cdfcoloneqq r(s, a) + \gamma \Z(s', a'), \quad s' \sim p(s' \mid s, a), a' \sim \pi(a' \mid s')$$
\end{definition}

\begin{wrapfigure}{r}{0.3\textwidth}
\vspace{-0.5cm}
\centering
\includegraphics[width=0.3\textwidth]{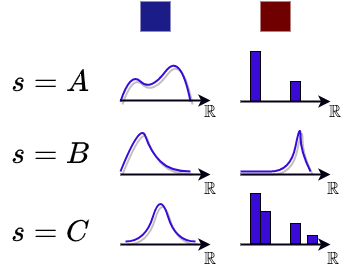}
\vspace{-0.4cm}
\end{wrapfigure}

По определению, истинное $\Z^\pi$ будет неподвижной точкой такого оператора:
$$\Z^\pi = \B_D \Z^\pi$$

Нас интересует вопрос о сходимости метода простой итерации. Что это означает? Если на $k$-ой итерации мы храним большую табличку, где для каждой пары $s, a$ хранится целиком и в точности всё распределение $\Z_k(s, a)$, то на очередном шаге для всех пар $s, a$ происходит обновление
\begin{equation}\label{distrpolicyevalupdate}
    \Z_{k+1}(s, a) \cdfcoloneqq r(s, a) + \gamma \Z_k(s', a'),
\end{equation}
где вероятности случайных величин $s' \sim p(s' \mid s, a), a' \sim \pi(a' \mid s')$ мы знаем и потому можем полностью посчитать свёртку распределений $\Z_k(s', a')$ для всевозможных пар следующих $s', a'$. 

Чтобы показать сходимость такой процедуры, хочется в аналогии с традиционным случаем доказать сжимаемость оператора $\B_D$. Однако, обсуждение сжимаемости имеет смысл только при заданной метрике, а в данном случае даже для конечных пространств состояний и пространств действий пространство Z-функций бесконечномерно, поскольку бесконечномерно $\Prob (\R)$. Нам нужна метрика в таком пространстве, и, внезапно, от её выбора будет зависеть ответ на вопрос о сжимаемости.

\begin{definition}
Пусть $\D$ --- метрика в пространстве $\Prob (\R)$. Тогда её \emph{максимальной формой} (maximal form) будем называть следующую метрику в пространстве Z-функций:
$$\D^{\max}(\Z_1, \Z_2) \coloneqq \sup\limits_{s \in \St, a \in \A} \D \left( \Z_1(s, a), \Z_2(s, a) \right)$$
\end{definition}

\begin{theoremBox}[label=constructmetric]{}
Для любой метрики $\D$ в пространстве $\Prob (\R)$ её максимальная форма $\D^{\max}$ есть метрика в пространстве Z-функций.
\begin{proof}
Проверим неравенство треугольника. Для любых трёх Z-функций $\Z_1, \Z_2, \Z_3$:
\begin{align*}
\D^{\max} (\Z_1, \Z_2) &= \sup\limits_{s, a} \D \left( \Z_1(s, a), \Z_2(s, a) \right) \le \\
\le \{ \text{неравенство треугольника для $\D$} \} &\le \sup\limits_{s, a} \left[ \D \left( \Z_1(s, a), \Z_3(s, a) \right) + \D \left( \Z_3(s, a), \Z_2(s, a) \right) \right] \le \\
\le \{ \text{свойство максимума} \} &\le \sup\limits_{s, a} \D \left( \Z_1(s, a), \Z_3(s, a) \right) + \sup\limits_{s, a} \D \left( \Z_3(s, a), \Z_2(s, a) \right) = \\
&= \D^{\max} (\Z_1, \Z_3) + \D^{\max} (\Z_3, \Z_2)
\end{align*}
Симметричность, неотрицательность и равенство нулю только при совпадении аргументов проверяется непосредственно.
\end{proof}
\end{theoremBox}

\begin{wrapfigure}{r}{0.3\textwidth}
\vspace{-0.5cm}
\centering
\includegraphics[width=0.3\textwidth]{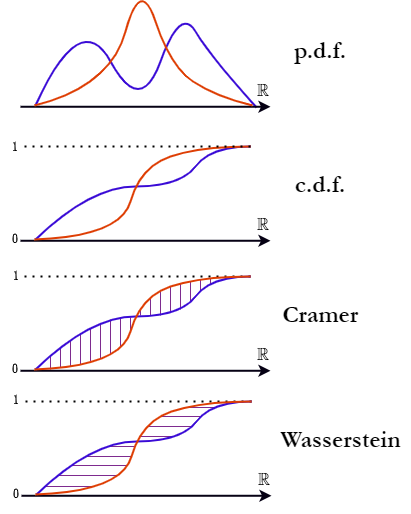}
\vspace{-5.3cm}
\end{wrapfigure}

Соответственно, вопрос о выборе метрики в пространстве Z-функций сводится к вопросу о метрике в пространстве скалярных случайных величин. Вопрос, вообще, довольно богатый. Можно пытаться посчитать расстояние между функциями распределения (такова, например, \href{https://en.wikipedia.org/wiki/Energy_distance}{метрика Крамера}), а можно --- между их обратными функциями:

\begin{definition}
Для скалярной случайной величины $X$ с функцией распределения $F_X(x) \colon \R \to [0, 1]$ \emph{квантильной функцией} (inverse distribution function (quantile function)) называется\footnote[*]{инфинум берётся для однозначности определения в ситуациях, когда в $F_X(x)$ есть плато.}
$$F^{-1}_X(\omega) \coloneqq \inf\{ x \in \R \mid F_X(x) \ge \omega \}$$
Значение этой функции $F^{-1}_X(\omega)$ в точке $\tau \in (0, 1)$ будем называть \emph{$\tau$-квантилем}. 
\end{definition}

\begin{definition}
Для $1 \le p \le +\infty$ для двух случайных скалярных величин\footnote[*]{с ограниченными $p$-ми моментами, что в нашем контексте гарантируется ограниченностью награды \eqref{reward_limit} и ограниченностью домена.} $X, Y$ с функциями распределения $F_X$ and $F_Y$ соответственно \emph{расстоянием Вассерштайна} (Wasserstein distance) называется
\begin{equation}\label{wasserstein}
    \W_p(X, Y) \coloneqq \left( \int\limits_0^1 \left| F_X^{-1}(\omega) - F_Y^{-1}(\omega) \right| ^p \diff \omega \right)^{\frac{1}{p}}
\end{equation}
$$\W_\infty(X, Y) \coloneqq \sup_{\omega \in [0,1]} \left| F_X^{-1}(\omega) - F_Y^{-1}(\omega) \right| $$
\end{definition}

Расстояние Вассерштайна --- это довольно глубокая тема в математике; особый интерес представляет случай $p \HM= 1$ (см., например, \href{https://en.wikipedia.org/wiki/Wasserstein_metric}{википедию}). 

\begin{theorem}[Эквивалентная форма $\W_1$]
\begin{equation}\label{W1equiv}
\W_1(X, Y) = \int\limits_{\R} \left| F_X(x) - F_Y(x) \right| \diff x
\end{equation}
\beginproof
При $p \HM= 1$ расстояние Вассерштайна $\W_1$ есть просто площадь между графиками c.d.f. $F_X, F_Y$; тоже самое записано и в этой форме, только интегрирование (<<суммирование>>) проводится по оси значений, а не оси квантилей, но как график не поверни, площадь остаётся той же.

Формальное обоснование даёт теорема из матана о том, что в двойном интеграле можно менять местами интегралы:
$$ \int\limits_{\R} \left| F_X(x) - F_Y(x) \right| \diff x = \int\limits_{\R} \diff x \int\nolimits_{\min(F_X(x), F_Y(x))}^{\max(F_X(x), F_Y(x))} \diff \omega$$
Это площадь, если просуммировать вдоль оси $x$; давайте попробуем поменять местами интегралы. 

Рассмотрим какое-нибудь $\omega$; есть два симметричных случая. В первом $F_X(x) \le \omega \le F_Y(x)$. Ну тогда из второго неравенства $F_Y^{-1}(\omega) \le x$. Теперь заметим, что $F^{-1}_X(\omega) \ge F^{-1}_X(\hat{\omega})$ для любого $\hat{\omega} \le \omega$ в силу неубывания $F^{-1}_X$; поэтому возьмём в качестве $\hat{\omega} \coloneqq F_X(x)$, который по первому неравенству не больше $\omega$, и получим 
$$F^{-1}_X(\omega) \ge F^{-1}_X(\hat{\omega}) = F^{-1}_X(F_X(x)) = x$$
Мы получили, что $F^{-1}_X(\omega) \le x \le F^{-1}_Y(\omega)$. В симметричном случае, когда $F_Y(x) \le \omega \le F_X(x)$, мы получили бы аналогично $F^{-1}_Y(\omega) \le x \le F^{-1}_X(\omega)$. Таким образом, при данном $\omega$ для подсчёта площади переменной $x$ нужно пробежать от $\min(F^{-1}_X(x), F^{-1}_Y(x))$ до $\max(F^{-1}_X(x), F^{-1}_Y(x))$; мы получаем равенство
$$ \left( \int\limits_{\R} \diff x \int\nolimits_{\min(F_X(x), F_Y(x))}^{\max(F_X(x), F_Y(y))} \diff \omega \right) = \left( \int\limits_0^1 \diff \omega \int\nolimits_{\min(F^{-1}_X(\omega), F^{-1}_Y(\omega))}^{\max(F^{-1}_X(\omega), F^{-1}_Y(\omega))} \diff x \right)$$
Ну а это в точности равно $\W_1$, то есть выинтегрированию модуля разности между $F^{-1}_X$ и $F^{-1}_Y$.\qed

\beginproof[Замечание] Утверждение верно только для $p \HM= 1$: иначе существенно, вдоль какой оси мы растягиваем разность функций возведением в $p$-ую степень (физический смысл <<площади между графиками>> пропадает). Поэтому метрика Крамера, которая есть L2-расстояние между c.d.f., не равна $\W_2$, которая есть L2-расстояние между квантильными функциями.
\end{theorem}

\begin{exampleBox}[label=ex:emd]{}
Расстояние $\W_1$ между двумя распределениями неспроста имеет второе название \emph{Earth Moving Distance}. Аналогия такая: нам даны две кучи песка. Объём песка в кучах одинаков, но у них разные конфигурации, они <<насыпаны>> по-разному. Чтобы перенести каждую песчинку массы $m$ на расстояние $x$, нам нужно затратить <<работы>> объёмом $mx$. Расстояние Вассерштайна $\W_1$ замеряет, какое минимальное количество работы нужно совершить, чтобы перевести конфигурацию первой кучи песка во вторую кучу; объём песка в каждой кучи одинаков. Для дискретных распределений, когда функции распределения (и, соответственно, квантильные функции) --- <<ступеньки>>, минимальная работа полностью соответствует площади между функциями распределений.

\needspace{7\baselineskip}
\begin{wrapfigure}{r}{0.35\textwidth}
\vspace{-0.3cm}
\centering
\includegraphics[width=0.35\textwidth]{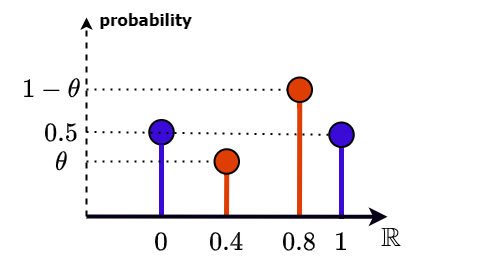}
\vspace{-0.6cm}
\end{wrapfigure}
Посчитаем $\W_1$ между двумя следующими распределениями. Первое распределение (синие на картинке) --- честная монетка с исходами 0 и 1. Вторая случайная величина (красная на картинке) принимает значение 0.4 с вероятностью $\theta \HM< 0.5$ и 0.8 с вероятностью $1 \HM- \theta$. Можно нарисовать функции распределения и посчитать площадь между ними. А можно рассуждать так: давайте <<превратим>> вторую кучу песка в первую. Посмотрим на песок объёма $\theta$ в точке 0.4. Куда его переносить? Наверное, в точку 0, куда его тащить ближе. Перенесли; совершили работы объёмом $0.4\theta$. Посмотрим на песок объёма $1 \HM- \theta$ в точке 0.8. Его удобно тащить в точку 1, но там для получения первой конфигурации нужно только 0.5 песка. Поэтому 0.5 песка из точки 0.8 мы можем перевести в точку 1, совершив работу $0.2 \cdot 0.5$, а оставшийся объём $1 \HM- \theta \HM- 0.5$ придётся переводить в точку 0, совершая работу $0.8 (0.5 \HM- \theta)$. Итого расстояние Вассерштайна равно:
$$\W_1 = 0.4\theta + 0.8(0.5 - \theta) + 0.1$$
\end{exampleBox}

\begin{proposition}
Максимальная форма метрики Вассерштайна $\W^{\max}_p$ есть метрика в пространстве Z-функций.
\end{proposition}

\begin{theorem}
По метрике $\W^{\max}_p(\Z_1, \Z_2)$ оператор $\B_D$ является сжимающим.
\beginproof[Доказательство для $\W_1$] Воспользуемся формой расстояния через c.d.f. \eqref{W1equiv}, тогда
$$\W^{\max}_1(\B_D\Z_1, \B_D\Z_2) = \sup_{s, a} \int\limits_{\R} \left| \Prob(r + \gamma \Z_1(s', a') \le x) - \Prob(r + \gamma \Z_2(s', a') \le x) \right| \diff x = (*)$$
где внутри вероятностей $s', a'$ --- тоже случайные величины! Ну, сначала заметим, что добавление награды не изменяет значение интеграла. Давайте сделаем такую замену: $\hat{x} = \frac{x - r}{\gamma}$. Получаем:
$$(*) = \sup_{s, a} \gamma \int\limits_{\R} \left| \Prob(\Z_1(s', a') \le \hat{x}) - \Prob(\Z_2(s', a') \le \hat{x}) \right| \diff \hat{x} = (**)$$
Осталось справиться со случайностью $s', a'$; к счастью, для функций распределений это несложно. Пусть $s, a$ --- фиксированы, тогда просто по формуле полной вероятности:
$$\Prob(\Z(s', a') \le \hat{x}) = \int\limits_{\St} \int\limits_{\A} p(s' \mid s, a)\pi(a' \mid s') F_{\Z(s', a')}(\hat{x}) \diff s' \diff a',$$
где $F_{\Z(s', a')}(\hat{x})$ --- вероятность, что $\Z(s', a')$ не превзойдёт $\hat{x}$ при фиксированных $s', a'$. Подставляем:
\begin{align*}
(**) &= \gamma \sup_{s, a} \int\limits_{\R} \left| \int\limits_{\St} \int\limits_{\A} p(s' \mid s, a)\pi(a' \mid s') \left(F_{\Z_1(s', a')} - F_{\Z_2(s', a')} \right) \diff s' \diff a' \right| \diff \hat{x} \le \\
&\le \{ \text{наше любимое $\E_{x} f(x) \le \max\limits_x f(x)$} \} \le \gamma \sup_{s, a} \W^{\max}_1(\Z_1, \Z_2)   \tagqed
\end{align*}
\end{theorem}

Это значит, что для систем уравнений \eqref{ZZ} выполняется теорема Банаха \ref{Banach}!

\begin{proposition}
Существует единственная функция $\St \times \A \to \Prob (\R)$, являющаяся решением уравнений \eqref{ZZ}, и метод простой итерации сходится к ней из любого начального приближения по метрике Вассерштайна.
\end{proposition}

\needspace{9\baselineskip}
\begin{example}
Попробуем найти $\Z^{\pi}$ для случайной $\pi$ (выбирающей из двух действий всегда равновероятно) для MDP с рисунка и $\gamma \HM = 0.5$.

\begin{wrapfigure}{r}{0.15\textwidth}
\vspace{-0.5cm}
\centering
\includegraphics[width=0.12\textwidth]{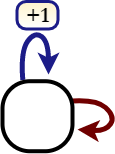}
\vspace{-0.3cm}
\end{wrapfigure}

Сначала попробуем понять, в каких границах может лежать наша награда за весь эпизод. Если, например, мы всё время выбираем \colorsquare{ChadRed}, то получим в итоге ноль; меньше, понятно, получить нельзя. Если же мы всё время выбираем \colorsquare{ChadBlue}, то получим в итоге $1 \HM+ \gamma \HM+ \gamma^2 \HM+ \dots \HM+ \HM= \frac{1}{1 - \gamma} \HM= 2$. Значит, вероятные исходы размазаны на отрезке $[0, 2]$. 

Попробуем посмотреть на $\Z^{\pi}(\text{\colorsquare{ChadRed}})$. По определению, мы предполагаем, что на первом шаге выбирается действие $\text{\colorsquare{ChadRed}}$, и значит, на первом шаге мы гарантированно получим +0. Тогда, проводя аналогичные рассуждения, можно заключить, что дальнейшая возможная награда лежит в отрезке $[0, 1]$. Но что именно это за распределение? Можно рассмотреть распределение случайной величины $\sum\limits_{t = 0}^T \gamma^t r_t \HM\mid a_0 \HM= \colorsquare{ChadRed}$ не при $T \HM= +\infty$, а при меньших $T$. Например, при $T \HM= 1$ мы получим +0, затем в качестве $r_1$ с равными вероятностями получим $+0$ или $+\frac{1}{2}$. Получится равновероятное распределение с исходами ${0, +\frac{1}{2}}$. При $T \HM= 2$ мы получим уже равновероятное распределение с исходами ${0, +\frac{1}{4}, +\frac{1}{2}, +\frac{3}{4}}$. Продолжая рассуждение дальше, можно увидеть, что при $T \to +\infty$ распределение продолжает равномерно размазывать вероятности по $[0, 1]$. Видимо, в пределе получится просто равномерное распределение на $[0, 1]$. Как можно строго доказать, что это правильный ответ?

Попробуем подставить в уравнения Беллмана \eqref{ZZ} в качестве $\Z^{\pi}(\text{\colorsquare{ChadRed}})$ равномерное распределение на отрезке $[0, 1]$, а в качестве $\Z^{\pi}(\text{\colorsquare{ChadBlue}})$ равномерное распределение на отрезке $[1, 2]$ (так как тут мы гарантированно получим +1 на первом же шаге). Что мы получим? Рассмотрим первое уравнение:
$$\Z^{\pi}(\text{\colorsquare{ChadBlue}}) \cdfeq \underbrace{+1}_{r} + \underbrace{0.5}_{\gamma} \Z^{\pi}(a'),$$
где $a'$ --- случайная величина, с равной вероятностью принимающая оба возможных значения. 

Вот мы выбрали $\colorsquare{ChadBlue}$: с одной стороны левая часть уравнения говорит, что мы получим равномерное распределение на $[1, 2]$. С другой стороны правая часть уравнения рассматривает <<одношаговое приближение>>: мы точно получим +1, затем кинем кубик; с вероятностью 0.5 выберем на следующем шаге \colorsquare{ChadBlue} и получим равномерное из $[1, 2]$, а с вероятностью 0.5 выберем \colorsquare{ChadRed} и получим равномерное из $[0, 1]$. Значит, начиная со второго шага мы получим сэмпл из равномерного на $[0, 2]$; он будет дисконтирован на гамму и получится сэмпл из $[0, 1]$; дальше мы его сдвинем на +1, который мы получили на первом шаге, и в итоге как раз получится равномерное из $[1, 2]$! Сошлось; в левой и правой стороне уравнения получается одно и то же распределение! Аналогично проверяется, что сходится второе distributional-уравнение. Из доказанного нами свойства сжатия следует, что это решение --- единственное, и, значит, является истинной $\Z^{\pi}$. 
\end{example}

Итак, мы с каждым шагом алгоритма становимся всё ближе к $\Z^\pi$, однако только если мы понимаем близость в терминах Вассерштайна. Это не единственная метрика в $\Prob ( \R )$, по максимальной форме которой была доказана сжимаемость $\B$; например, ещё она доказана для максимальной формы метрики Крамера. Важно, что есть примеры метрик, для максимальных форм которых свойства сжатия нет (например, \href{https://en.wikipedia.org/wiki/Total_variation_distance_of_probability_measures}{Total Variation}). Для нас важен более практический пример: в реальности нам с Вассерштайном обычно неудобно работать, и мы предпочитаем более удобные \emph{дивергенции}\footnote{снимается требование симметричности и неравенства треугольника.}, например, $\KL$-дивергенцию. Чисто теоретически мы можем задаться вопросом, как ведёт себя расстояние от $\Z_k \HM\coloneqq \B_D^k \Z_0$ до истинной $\Z^{\pi}$ в терминах KL-дивергенции, то есть стремится ли оно хотя бы к нулю? Оказывается, не просто не стремится, а вообще полное безобразие происходит: KL-дивергенция не умеет адекватно мерить расстояние между распределениями с \emph{несовпадающим доменом} (disjoint support).

\begin{theorem}
Расстояние между $\B_D^k \Z_0$ и истинным $\Z^\pi$ по максимальной форме $\KL$-дивергенции может быть равно бесконечности для всех $k$.

\needspace{7\baselineskip}
\begin{wrapfigure}{r}{0.2\textwidth}
\vspace{-1cm}
\centering
\includegraphics[width=0.2\textwidth]{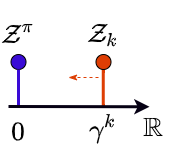}
\vspace{-0.5cm}
\end{wrapfigure}
\beginproof[Пример]
Пусть в MDP с одним состоянием, одним действием и нулевой функцией награды мы проинициализировали $\Z_0$ вырожденной случайной величиной, всегда принимающей значение 1. Тогда на первом шаге метода мы получим случайную величину, с вероятностью 1 равную $\gamma$; на $k$-ом шаге, по индукции, случайную величину, с вероятностью 1 равную $\gamma^k$. При этом $\KL$-дивергенция между ней и истинным распределением $\Z^\pi$ --- вырожденной в нуле --- равна бесконечности! \qed
\end{theorem}

Так, ну ладно: оно же сходится по Вассерштайну, который лишён этой проблемы. Мы показали, что мы чисто теоретически умеем конструктивно находить $\Z^{\pi}$, запустив метод простой итерации из произвольного начального приближения (в том числе, кстати, можем стартовать из вырожденных распределений). От практического алгоритма нас пока отделяет тот факт, что даже в табличном сеттинге мы не умеем в ячейках таблицы хранить <<полностью>> распределения на $\R$; мы займёмся этой проблемой чуть позже.

Пока что ответим на такой вопрос: а вот когда мы учим так $\Z^{\pi}$, что там происходит с их мат.ожиданиями, то есть, по сути, с нашими представлениями о $Q^\pi$? Может, они там как-то быстрее сходятся за счёт того, что мы начали дополнительную информацию о распределении учить? Нет: их поведение в точности совпадает с тем, что получилось бы в методе простой итерации для обучения Q-функции непосредственно.

Пусть $\B$ --- обычный оператор Беллмана из пространства Q-функций в пространство Q-функций, а $\B_D$, как и раньше, оператор Беллмана из пространства Z-функций в пространство Z-функций.

\begin{theorem}
Пусть инициализации $\Z_0$ и $Q_0$ удовлетворяют $\E \Z_0 = Q_0$, и рассматривается два метода простой итерации:
$$Q_k \coloneqq \B^k Q_0$$
$$\Z_k \coloneqq \B_D^k \Z_0$$
Тогда:
$$Q_k = \E \Z_k$$
\beginproof
По индукции. Пусть это верно для $k$-ой итерации, покажем для $k+1$-ой:
$$\E \Z_{k+1} = \E \left[ \B_D \Z_k \right](s, a) = \E r(s, a) + \gamma \Z_k(s', a') = (*)$$
Заметим, что мат.ожидание в последнем выражении берётся по $s', a'$ и случайности в $\Z_k(s', a')$ (случайности по хвосту траектории). По предположению индукции:
$$\E \Z_k(s', a') = Q_k(s', a')$$
Подставляем:
\begin{align*}
(*) =  \E_{s'} \E_{a'} r(s, a) + \gamma \E \Z_k(s', a') = \E_{s'}\E_{a'} \left[ r(s, a) + \gamma Q_k(s', a') \right] = \B Q_k = Q_{k+1}   \tagqed
\end{align*}
\end{theorem}

\subsection{Distributional Value Iteration}

По аналогии с традиционным случаем, очень хочется ввести оптимальную оценочную функцию в distributional-форме как Z-функцию оптимальных стратегий:
\begin{equation}\label{optimalZ}
\Z^*(s, a) \cdfcoloneqq Z^{\pi^*}(s, a)
\end{equation}

Мы начинаем спотыкаться уже на этом моменте, и дальше будет только хуже.

\begin{theorem}
Определение \eqref{optimalZ} неоднозначно.

\needspace{7\baselineskip}
\begin{wrapfigure}{r}{0.25\textwidth}
\vspace{-1cm}
\centering
\includegraphics[width=0.25\textwidth]{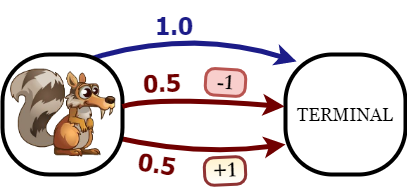}
\end{wrapfigure}
\beginproof
Рассмотрим MDP, где агент может выбрать действие $a \HM= \colorsquare{ChadBlue}$ и получить нулевую награду с вероятностью 1, или $a \HM = \colorsquare{ChadRed}$ и получить +1 или -1 с вероятностями 0.5 (эпизод в обоих случаях заканчивается). Все стратегии будут оптимальными, хотя все Z-функции различны. \qed
\end{theorem}

С уравнением оптимальности Беллмана для $\Z^*$ тоже внезапно есть тонкости. Для любой оптимальной стратегии  $\pi^*$ вследствие \eqref{QZ} верно, что
$$Q^*(s, a) = \E \Z^{\pi^*}(s, a),$$
и мы знаем, что, в частности, среди оптимальных есть стратегия
$$\pi^*(s) = \argmax\limits_a Q^*(s, a) = \argmax\limits_a \E \Z^{\pi^*}(s, a).$$
В принципе, можно взять \eqref{ZZ} для этой $\pi^*(s)$ и использовать её вид.
\begin{equation}\label{Z*Z*}
\Z^*(s, a) \cdfeq r(s, a) + \gamma \Z^*(s', \pi^*(s')), \quad s' \sim p(s' \mid s, a)
\end{equation}
Здесь справа мы для данных $s, a$ описываем следующий процесс генерации случайной величины: генерируем $s'$ из функции переходов, определяем однозначно\footnote{здесь есть нюанс с выбором жадного действия в случае равного среднего для нескольких вариантов: для полной корректности, множество действий должно быть упорядочено, и в <<спорных>> ситуациях следует выбирать действие с наименьшим порядком. Это существенно, поскольку выбор разных оптимальных действий приводит к одному и тому же среднему, но другие статистики могут быть различны.} $a' = \argmax\limits_{a'} \E \Z^*(s', a')$, после чего генерируем сэмпл $\Z^*(s', a')$ и выдаём $r(s, a) + \gamma \Z^*(s', a')$ в качестве результата.

Заметим, что взятие мат.ожидания справа и слева в уравнении \eqref{Z*Z*} приведёт к традиционному уравнению оптимальности Беллмана для Q \eqref{Q*Q*}.

\begin{definition}
Введём \emph{оператор оптимальности Беллмана в distributional-форме} $\B^*_D$:
$$\left[\B^*_D \Z \right] (s, a) \cdfcoloneqq r(s, a) + \gamma \Z(s', \argmax\limits_{a'} \E \Z(s', a')), \quad s' \sim p(s' \mid s, a)$$
\end{definition}

Пусть также $\B^*$ --- обычный оператор оптимальности Беллмана из пространства Q-функций в пространство Q-функций.

\begin{theoremBox}[label=th:distributionalVIiscorrect]{}
Пусть инициализации $\Z_0$ и $Q_0$ удовлетворяют $\E \Z_0 = Q_0$ и рассматривается два метода простой итерации:
$$Q_k \coloneqq \left(\B^*\right)^k Q_0$$
$$\Z_k \coloneqq \left(\B^*_D\right)^k \Z_0$$
Тогда:
$$Q_k = \E \Z_k$$
\beginproof
По индукции. Пусть это верно для $k$-ой итерации, покажем для $k+1$-ой:
$$\E \Z_{k+1} = \E \left[ \B^*_D \Z_k \right](s, a) = \E \left[ r(s, a) + \gamma \Z_k(s', \argmax\limits_{a'} \E \Z_k(s', a')) \right] = (*)$$
Заметим, что внутренний аргмакс эквивалентен аргмаксу по $Q$-функции, так что здесь мы тоже можем воспользоваться предположением индукции:
$$\E \Z_k(s', a') = Q_k(s', a')$$
Подставляем:
\begin{align*}
(*) &= \E_{s'}\E_{a'} \left[ r(s, a) + \gamma \E \Z_k(s', \argmax\limits_{a'} Q_k(s', a')) \right] = \\ 
&= r(s, a) + \gamma Q_k(s', \argmax\limits_{a'} Q_k(s', a')) = \\
&= r(s, a) + \gamma \max\limits_{a'} Q_k(s', a') = \B^* Q_k = Q_{k+1}   \tagqed
\end{align*}
\end{theoremBox}

Итак, мы показали, что в методе простой итерации с оператором $\B^*_D$ средние движутся точно также, как и $Q^*$ в обычном подходе. Однако, хвосты распределений при этом могут вести себя довольно нестабильно, и понятно, почему.

\begin{theorem}
Оператор $\B^*_D$ может не являться непрерывным.

\needspace{13\baselineskip}
\begin{wrapfigure}{r}{0.3\textwidth}
\vspace{-1cm}
\centering
\includegraphics[width=0.28\textwidth]{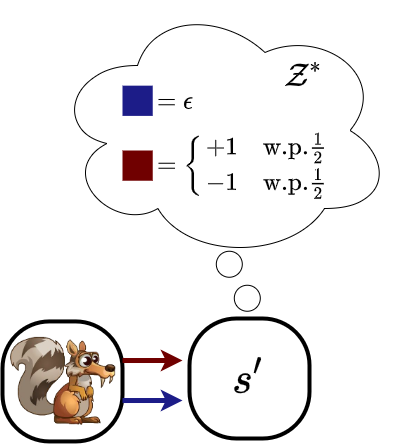}
\vspace{0.2cm}
\end{wrapfigure}
\beginproof
Пусть после выполнения $s, a$ мы точно попадаем в некоторое $s'$, для которого наше приближение $\Z$ указано как на рисунке; варьируя $\epsilon$, мы можем получать близкие (по любой непрерывной метрике) Z-функции. Рассмотрим $\epsilon \HM\to 0$ и поймём, что оператор $\B^*_D$ выдаёт совершенно разные Z-функции в зависимости от того, приближаемся ли мы к нулю с положительной полуоси или отрицательной.

Смотрим на $\left[ \B^*_D \Z \right](s, a) \HM\cdfeq \gamma \Z(s', \pi^*(s'))$, где $\pi^*(s')$ --- жадная. Если $\epsilon \HM> 0$, жадная политика выдаст $a' \HM= \text{\colorsquare{ChadBlue}}$, и результат оператора будет вырожденным: $\left[ \B^*_D \Z \right](s, a)$ скажет, что с вероятностью 1 будет получена награда $\gamma \epsilon$. Если $\epsilon \HM< 0$, то результат оператора будет дискретным распределением с двумя атомами $\gamma$ и $-\gamma$ (с вероятностями $\frac{1}{2}$). Расстояние между этими двумя вариантами по любой метрике не будет нулевым. Иными словами, при переходе $\epsilon$ через ноль при непрерывном изменении аргумента оператора значение оператора может сколько угодно сильно измениться. \qed
\end{theorem}

\begin{proposition}
Оператор $\B^*_D$ может не являться сжимающим.
\begin{proof}[Пояснение] По определению, любой сжимающий оператор \href{https://ru.wikipedia.org/wiki/\%D0\%9B\%D0\%B8\%D0\%BF\%D1\%88\%D0\%B8\%D1\%86\%D0\%B5\%D0\%B2\%D0\%BE_\%D0\%BE\%D1\%82\%D0\%BE\%D0\%B1\%D1\%80\%D0\%B0\%D0\%B6\%D0\%B5\%D0\%BD\%D0\%B8\%D0\%B5}{Липшицев}, и, значит, обязан быть непрерывным.
\end{proof}
\end{proposition}

Мы, тем не менее, показали, что средние сходятся к $Q^*$, поэтому не совсем ясно, насколько страшно, что хвосты распределений могут начать вести как-то нестабильно.

\subsection{Категориальная аппроксимация Z-функций}

Пока что мы не получили даже табличный алгоритм в рамках distributional-подхода: даже если пространства $\St$ и $\A$ конечны, а функция переходов известна, хранить в памяти точное распределение $\Z(s, a)$ мы не умеем. Нам придётся выбрать некоторую параметрическую аппроксимацию. Хорошая новость заключается в том, что награда --- скаляр, и распределения, с которыми мы хотим работать, одномерны. Более того, в силу \eqref{reward_limit} домен распределений ограничен. Мы можем этим воспользоваться и придумать какую-нибудь <<сеточную>> аппроксимацию.

\begin{definition}
Зададимся набором \emph{атомов} (atoms) $r^{\min} \HM= z_0 \HM< z_1 \HM< z_2 \dots < z_A \HM= r^{\max}$, где $A + 1$ --- число атомов. Обозначим \emph{семейство категориальных распределений} $\mathcal{C} \subset \Prob(\R)$ как множество дискретных распределений на домене $\{z_0, z_1 \dots z_A\}$: если $\Z(s, a) \HM\in \mathcal{C}$, то
$$\Prob \left( \Z(s, a) = z_i \right) = p_i,$$
где $p$ --- набор из $A \HM+ 1$ чисел, суммирующихся в единицу.
\end{definition}

\begin{example}[c51]
Типично атомы образуют просто равномерную сетку, для задания которой требуется три гиперпараметра: число атомов, минимальное и максимальное значение награды. Распространённый дефолтный вариант для Atari игр --- 51 атом на отрезке $[-10, 10]$. В честь такой параметризации (categorical with 51 atoms) иногда алгоритм Categorical DQN, к построению которого мы приближаемся, называют c51.
\end{example}

\needspace{7\baselineskip}
\begin{wrapfigure}{r}{0.35\textwidth}
\vspace{-0.3cm}
\centering
\includegraphics[width=0.35\textwidth]{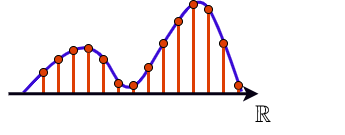}
\vspace{-0.6cm}
\end{wrapfigure}
Итак, для каждой пары $s, a$ мы будем хранить в табличке $A + 1$ неотрицательное число $p_0, p_1 \dots p_A$, суммирующиеся в единицу, и полагать, что $A+1$ узлов нашей сетки $z_0, z_1 \dots z_A$ являются единственно возможными исходами будущей награды. Такова наша аппроксимация.

 Возникает следующая проблема: мы, в принципе, можем посчитать распределения $\B^*_D \Z$, но что, если оно <<не попадёт>> в рассматриваемое семейство аппроксимаций? То есть что, если для какой-то пары $s, a$ $[\B^*_D \Z](s, a) \not\in \mathcal{C}$, то есть что, если оно не является категориальным распределением на домене $\{z_0, z_1 \dots z_A\}$? Нам придётся как-то проецировать полученный результат на нашу сетку...
 
 \begin{proposition}
 В табличном сеттинге если $\Z(s, a) \in \mathcal{C}$ для всех $s, a$, то $\left[\B^*_D \Z\right](s, a)$ --- дискретное распределение с конечным множеством исходов.
 \begin{proof} Распределение является смесью не более чем $|\St| |\A|$ категориальных распределений с $A$ исходами, поэтому у него не может быть более $|\St| |\A| A$ различных исходов.
 \end{proof}
 \end{proposition}
 
 Значит, нам нужно научиться проецировать лишь дискретные распределения.
 
 \begin{definition}
 Введём \emph{оператор проекции} $\Pi$, действующий из пространства произвольных дискретных распределений в $\mathcal{C}$ следующим образом. Пусть $\tilde{\Z}(s, a)$ --- произвольное дискретное распределение с исходами $\tilde{z}_i$ с соответствующими вероятностями $\tilde{p}_i$ (суммирующимися в единицу). Изначально инициализируем все $p_i$ для результата работы оператора нулями.
 
 Дальше перебираем исходы $\tilde{z}_i$; если очередной исход меньше $r^{\min} \HM= z_0$, всю его вероятностную массу отправляем в $p_0$, то есть увеличиваем $p_0$ на $\tilde{z}_i$. Аналогично поступаем если $\tilde{z}_i \HM> r^{\max} = z_A$. В остальных случаях найдётся два соседних атома нашей сетки, такие что $z_j \HM\le \tilde{z}_i \HM\le z_{j+1}$. Распределим вероятностную массу между ними обратно пропорционально расстоянию до них, а то есть:
\begin{equation}\label{projector}
\begin{split} 
p_j \leftarrow p_j + \frac{z_{j+1} - \tilde{z}_i}{z_{j+1} - z_j} \tilde{p}_i  \\
p_{j+1} \leftarrow p_{j+1} + \frac{\tilde{z}_i - z_j}{z_{j+1} - z_j} \tilde{p}_i
\end{split}
\end{equation}
 \end{definition}

Почему именно так мы ввели оператор проекции? Наш метод простой итерации теперь <<подкорректированный>>, после каждого шага мы применяем проекцию:
$$\Z_{k+1} \cdfcoloneqq \Pi \B^*_D \Z_{k},$$
где применение $\Pi$ к Z-функции означает проецирование всех распределений $\Z(s, a)$ для всех $s, a$. Мы очень хотели бы сохранить гарантии теоремы \ref{th:distributionalVIiscorrect} о том, что средние в таком подправленном процессе продолжают вести себя как аппроксимации Q-функции в Value Iteration!

\begin{theorem}
Пусть $\Z(s, a)$ дискретно и выдаёт исходам вне отрезка $[r_{\min}, r_{\max}]$ нулевую вероятность. Тогда оператор проекции \eqref{projector} сохраняет мат.ожидание, $\forall \Z, s, a \colon$
$$\E \Pi \Z (s, a) = \E \Z(s, a)$$
 
\needspace{7\baselineskip}
\begin{wrapfigure}{r}{0.4\textwidth}
\vspace{-1cm}
\centering
\includegraphics[width=0.4\textwidth]{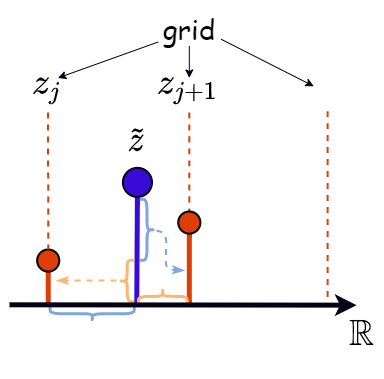}
\vspace{-0.6cm}
\end{wrapfigure}
\beginproof
Рассмотрим один исход $\Z(s, a)$; он вносил в итоговое среднее вклад $\tilde{z}\tilde{p}$, где $\tilde{p}$ --- его вероятность, $\tilde{z}$ --- значение исхода. По условию, вся вероятностная масса размазывалась между двумя соседними узлами $z_j \le \tilde{z} \le z_{j+1}$, и в $\left[\Pi \Z\right](s, a)$ соответственно появляется два слагаемых:
\begin{align*}
    &\overbrace{\frac{z_{j+1} - \tilde{z}}{z_{j+1} - z_j} \tilde{p} z_j}^{\text{левый узел}} + \overbrace{\frac{\tilde{z} - z_j}{z_{j+1} - z_j} \tilde{p} z_{j+1}}^{\text{правый узел}} = \\ = &\frac{z_{j+1}z_j - \tilde{z}z_j + \tilde{z}z_{j+1} - z_{j}z_{j+1}}{z_{j+1} - z_j} \tilde{p} = \\ = &\tilde{z}\tilde{p}   \tagqed
\end{align*}
\end{theorem}

\subsection{Categorical DQN}\label{subsec:c51}

\needspace{7\baselineskip}
\begin{wrapfigure}{r}{0.3\textwidth}
\vspace{-0.5cm}
\centering
\includegraphics[width=0.3\textwidth]{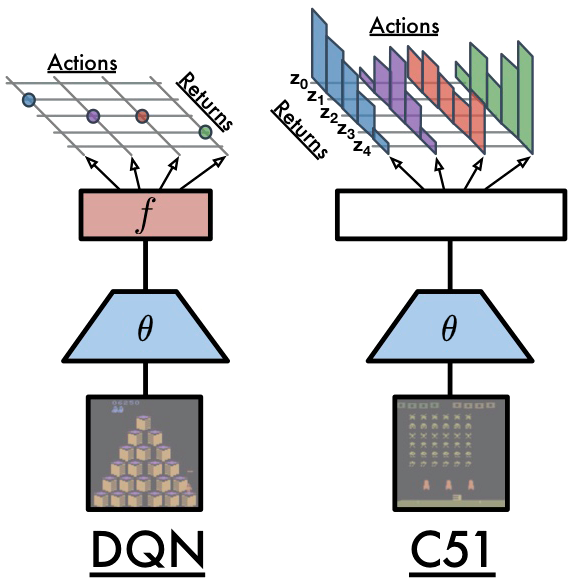}
\vspace{-0.5cm}
\end{wrapfigure}
Попробуем составить уже полностью практический алгоритм. Во-первых, обобщим алгоритм на случай произвольных пространств состояний, моделируя $\Z_\theta \HM\approx \Z^*$ (а точнее --- её распределение) при помощи нейросети с параметрами $\theta$. Для каждой пары $s, a$ такая нейросеть выдаёт $A \HM+ 1$ неотрицательное число $p_0(s, a, \theta), p_1(s, a, \theta) \dots p_A(s, a, \theta)$, суммирующиеся в единицу, и мы предполагаем категориальную аппроксимацию
\begin{equation}\label{neural_categorical}
\Prob \left( \Z_{\theta}(s, a) = z_i \right) \coloneqq p_i(s, a, \theta).
\end{equation}
Как и в DQN, считаем, что у нас есть таргет-сеть с параметрами $\theta^{-}$ --- Z-функция $\Z_{\theta^-}$ с предыдущего (условно, $k$-го) шага метода простой итерации, а мы хотим обучать $\theta$ так, чтобы получить Z-функцию на $k+1$-ом шаге: наша цель --- выучить $\B^*_D \Z_{\theta^-}$. 

В model-free режиме, без доступа к функции переходов, мы не то чтобы посчитать $\B^*_D \Z_{\theta^-}$ не можем, нам даже недоступна большая часть информации о нём. Для данной пары $s, a$ из реплей буфера мы можем получить только один сэмпл $s' \sim p(s' \mid s, a)$, и нам нужен какой-то <<аналог>> метода временных разностей.

Первое соображение: мы умеем сэмплировать из $[\B^*_D \Z_{\theta^-}](s, a)$. Действительно: пусть дано $s, a$; берём сэмпл $s'$ из, например, буфера; смотрим на нашу таргет сеть $\Z_{\theta^-}(s', a')$ для всех действий $a'$, считаем для каждого действия $a'$ мат.ожидание (для ситуации $\Z_{\theta^-}(s', a') \HM\in \mathcal{C}$ это, очевидно, не проблема) и выбираем <<наилучшее>> действие $a' \HM= \argmax\limits_{a'} \E \Z_{\theta^-}(s', a')$. Выбираем такое $a'$, и дальше у нас есть даже не сэмпл, а целая компонента искомого распределения $[\B^*_D \Z_{\theta^-}](s, a)$ в виде распределения $r(s, a) + \gamma \Z_{\theta^-}(s', a')$, которое мы и будем использовать в качестве таргета. 

Итак, пусть $\T \coloneqq (s, a, r, s')$ --- четвёрка из буфера. Введём целевую переменную (таргет) следующим образом:
\begin{equation}\label{distributionaltarget}
y(\T) \cdfcoloneqq r + \gamma \Z_{\theta^{-}}(s', \argmax_{a'} \E \Z_{\theta^{-}}(s', a'))
\end{equation}
где $s'$ в формуле берётся из $\T$, то есть взято из буфера. Такой таргет является дискретным распределением с, очевидно, $A$ атомами, но из-за того, что мы взяли лишь один сэмпл $s'$, он является лишь компонентой из $[\B^*_D \Z_{\theta^-}](s, a)$.

Второе соображение: допустим, для данной пары $s, a$ мы сможем оптимизировать следующий функционал для некоторой дивергенции $\D$, используя лишь сэмплы из первого распределения:
\begin{equation*}
\D ([\B^*_D \Z_{\theta^-} ] (s, a) \parallel \Z_{\theta}(s, a)) \to \min_{\theta}
\end{equation*}

Если бы мы могли моделировать произвольные Z-функции, минимум достигался бы в нуле на совпадающих распределениях, и наша цель была бы достигнута. Однако мы ограничены нашим аппроксимирующим категориальным семейством $\mathcal{C}$, и при оптимизации такого функционала даже чисто теоретически мы получим лишь проекцию на $\mathcal{C}$; здесь возникает вопрос, а не потеряем ли мы свойство сохранения мат.ожидания. Мы знаем, что наш оператор проекции \eqref{projector} обладает этим свойством: мы могли бы приближать наше распределение сразу к <<хорошей>> проекции:
\begin{equation}\label{distributionalopt}
\D ([\Pi \B^*_D \Z_{\theta^-}](s, a) \parallel \Z_{\theta}(s, a)) \to \min_{\theta}
\end{equation}
Тогда мы будем учить категориальное распределение с сохранённым мат.ожиданием. 

Вопрос: не потеряли ли мы возможность сэмплировать из целевого распределения? То есть можем ли мы получить сэмпл из $[\Pi \B^*_D \Z_{\theta^-}](s, a)$?

\begin{theorem}
\begin{equation}\label{projectioncomponents}
[\Pi \B^*_D \Z_{\theta^-}](s, a) \cdfeq \Pi y(\T), \quad s' \sim p(s' \mid s, a)
\end{equation}
\beginproof[Пояснение]
Сначала разберёмся, что здесь написано. Мы можем (теоретически) посчитать полностью одношаговую аппроксимацию $\B^*_D \Z_{\theta^-}$ и спроецировать полученное распредление (случ. величина слева); а можем взять случайный $s'$, посмотреть на распределение $r + \gamma\Z_{\theta^-}(s', a')$ для жадного $a'$ и спроецировать его (случ. величина справа). Утверждается эквивалентность этих процедур: мы можем проецировать лишь компоненты $[\B^*_D \Z_{\theta^-}](s, a)$. Таким образом, сэмплы из $\Pi y(\T)$ при случайных $s' \sim p(s' \mid s, a)$ есть сэмплы $[\Pi \B^*_D \Z_{\theta^-}](s, a)$.

\begin{proof} Следует, в общем-то, из определения нашего оператора проекции \eqref{projector}, который с каждым возможным исходом работает <<независимо>> от всех остальных. Пусть для некоторого $s'$ $p$ --- вероятность исхода $z$, тогда в правой части эта вероятность будет размазана между соседними узлами $z_i \HM\le z \HM\le z_{i+1}$ с некоторыми весами $w_i, w_{i+1}$. Тогда в силу того, что $s'$ случайно, эта вероятностная масса в итоговом распределении в правой части уравнения будет домножена на $p(s' \mid s, a)$. В распределении в левой части вероятностная масса будет сначала домножена на $p(s' \mid s, a)$, а только потом размазана между теми же $z_i, z_{i+1}$ с теми же весами $w_i, w_{i+1}$ (которые по определению зависят только от значения исхода $z$); очевидно, эти процедуры эквивалентны.
\end{proof}
\end{theorem}

Итак, $\Pi y(\T)$ есть компонента $[\Pi \B^*_D \Z_{\theta^-}](s, a)$, то есть у нас, условно, есть сэмплы из этого распределения. Для каких метрик $\D$ мы умеем получать несмещённую оценку градиента \eqref{distributionalopt} по сэмплам? Теория подсказывает, что в пространстве Z-функций осмысленной метрикой является Вассерштайн \eqref{wasserstein}. И тут случается облом.

\begin{theoremBox}[label=th:wassersteingradientsproblem]{}
Градиенты расстояния Вассерштайна до сэмплов не являются несмещёнными оценками градиента расстояния Вассерштайна до полного распределения.

\needspace{7\baselineskip}
\begin{wrapfigure}{r}{0.35\textwidth}
\vspace{-0.3cm}
\centering
\includegraphics[width=0.35\textwidth]{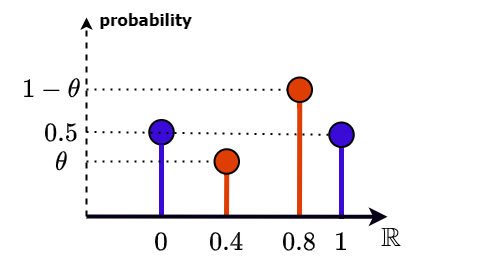}
\vspace{-0.6cm}
\end{wrapfigure}
\beginproof[Контрпример]
Контрпримером будет являться практически любая ситуация, где $s'$ недетерминировано, а наше параметрическое семейство $\Z_{\theta}$ может моделировать невырожденные случайные величины; так в принципе устроено расстояние Вассерштайна.

Разберём какой-нибудь пример, для простоты для $\W_1$. Пусть для данных $s, a$ с вероятностью 0.5 $s'$ такого, что $y_1(\T)$ --- вырожденная в нуле, а с вероятностью 0.5 $s'$ такого, что $y_2(\T)$ --- вырожденная в единице. Тогда понятно, что целиком распределение $[\B^*_D \Z_{\theta^-}](s, a)$ есть распределение с двумя равновероятными исходами, 0 и 1. 

Пусть $\Z_{\theta}$ равна 0.4 с вероятностью $\theta$ и 0.8 с вероятностью $1 \HM- \theta$, других значений не принимает. Будем смотреть на точку $\theta < 0.5$. Мы как раз считали подобные расстояния Вассерштайна в примере \ref{ex:emd}. Тогда:
$$\W_1 \left( \left[ \B^*_D \Z_{\theta^-} \right] (s, a) \parallel \Z_{\theta} \right) = 0.4\theta + 0.8(0.5 - \theta) + 0.1$$
$$\W_1 \left( y_1(\T) \parallel \Z_{\theta} \right) = 0.4 \theta + 0.8 (1 - \theta)$$
$$\W_1 \left( y_2(\T) \parallel \Z_{\theta} \right) = 0.6 \theta + 0.2 (1 - \theta)$$
Получаем:
$$\nabla_\theta \W_1 \left( [\B^*_D \Z_{\theta^-})] \parallel \Z_{\theta} \right) = -0.4$$
$$\nabla_\theta \E_{s'} \W_p \left( y(\T) \parallel \Z_{\theta} \right) = \frac{1}{2} (-0.4) + \frac{1}{2} (+0.4) = 0$$
Не сходится. \qed
\end{theoremBox}

Как мы сейчас покажем, псевдометрикой, которую можно оптимизировать по сэмплам, является наша любимая KL-дивергенция. Мы понимаем, что, с одной стороны, теория подсказывает нам, что в пространстве Z-функций $\KL$-дивергенция потенциально не приближает нас к истинной оптимальной Z-функции, но зато мы сможем оптимизировать её в model-free режиме.

Итак, рассмотрим в \eqref{distributionalopt} в качестве $\D$ $\KL$-дивергенцию (значит, будет важен порядок аргументов). Для неё вылезает ещё одна проблема: домен сравниваемых распределений должен совпадать, иначе KL-дивергенция по определению бесконечность и не оптимизируется. К счастью, мы уже решили, что мы будем в качестве целевого распределения использовать $\Pi y(\T)$, которое имеет тот же домен --- сетку $z_0 \HM< z_1 \HM< \dots \HM< z_A$. 

\begin{theorem}
Градиент $\KL$-дивергенции до целевой переменной $\Pi y(\T)$ есть несмещённая оценка градиента \eqref{distributionalopt}:
\begin{equation}\label{c51loss}
\nabla_\theta \KL ( \left[ \Pi \B^*_D \Z_{\theta^-} \right](s, a) \parallel \Z_{\theta}(s, a)) = \E_{s'} \nabla_\theta \KL (\Pi y(\T) \parallel \Z_{\theta}(s, a))
\end{equation}
\beginproof
\begin{align*}\nabla_\theta \KL ( \left[ \Pi \B^*_D \Z_{\theta^-} \right] (s, a) \parallel \Z_{\theta}(s, a)) &= \nabla_\theta \const(\theta) - \\
&- \nabla_\theta \E_{z \sim \left[ \Pi \B^*_D \Z_{\theta^-} \right] (s, a)} \log \Prob \left( \Z_{\theta}(s, a) = z \right) = \\
= \{ \text{\eqref{projectioncomponents}} \} &= -\nabla_\theta \E_{s'} \E_{z \sim \Pi y(\T)} \log \Prob \left( \Z_{\theta}(s, a) = z \right) = \\
&=\E_{s'} \nabla_\theta \KL (\Pi y(\T) \parallel \Z_{\theta}(s, a))     \tagqed
\end{align*}
\end{theorem}

Итак, градиент KL-дивергенции --- мат.ожидание по целевому распределению, и значит, мы можем вместо мат.ожидания по $\left[ \Pi \B^*_D \Z_{\theta^-} \right](s, a)$ использовать Монте-Карло оценку по сэмплам. При этом поскольку у нас есть даже не просто сэмплы, а целая компонента $\Pi y(\T)$ целевого распределения, то по ней интеграл мы можем взять просто целиком (он состоит всего из $A$ слагаемых, как видно, поскольку $\Pi y(\T) \in \mathcal{C}$).

\needspace{14\baselineskip}
\begin{wrapfigure}{r}{0.4\textwidth}
\centering
\includegraphics[width=0.4\textwidth]{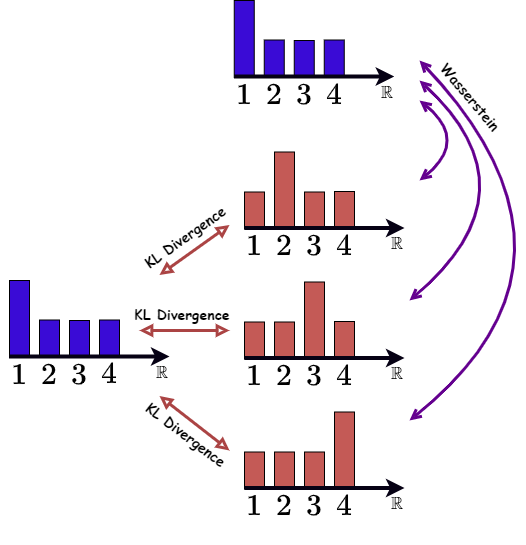}
\vspace{-0.8cm}
\end{wrapfigure}

Получается следующее: для данного перехода мы в качестве функции потерь возьмём $\KL (\Pi y(\T) \HM\parallel \Z_{\theta}(s, a))$, где таргет $y(\T)$ вычисляется по формуле \eqref{distributionaltarget}. Раз мы используем категориальную аппроксимацию \eqref{neural_categorical}, и $\Pi y(\T)$ --- категориальное распределение на той же сетке, то эта KL-дивергенция считается явно и (с точностью до константы, не зависящей от $\theta$) равна
$$-\sum_{i=0}^{A} \Prob \left( \Pi y(\T) = z_i \right) \log p_i(s_t, a_t, \theta).$$

Как видно из этой формулы, мы по сути начинаем решать задачу классификации, где у нас есть для данного входа $s, a$ сразу целая компонента <<целевого>> распределения. Минимизация KL-дивергенции, хоть и является стандартной функций потерь в таких ситуациях, сейчас имеет для нас побочный эффект: мы отчасти потеряли <<физический смысл>> наших <<классов>>. KL-дивергенция смотрит на каждый узел $z_i$ нашей сетки отдельно и сравнивает вероятность, которую мы выдаём сейчас, с вероятностью $z_i$ в таргете. Она не учитывает, находится ли разница в вероятностной массе на соседнем узле, например, $z_{i+1}$ (в <<соседнем>> исходе) или на противоположном конце сетки в условном $z_0$; в обоих случаях KL-дивергенция будет выдавать одно и то же значение. Адекватные метрики в пространстве распределений, например, Вассерштайн, продифференцировали бы эти случаи. Причём заметим, что мы, вообще говоря, могли бы посчитать того же Вассерштайна между $y(\T)$ и $\Z_{\theta}(s, a)$ (эти распределения дискретны, и мы разбирали, как считать расстояние Вассерштайна в таком случае в примере \ref{ex:emd}), но градиенты такой функции потерь в силу теоремы \ref{th:wassersteingradientsproblem} не были бы несмещёнными оценками градиента для минимизации \eqref{distributionalopt}, и такой алгоритм был бы некорректен.

Тем не менее, мы получили первый полноценный Distributional алгоритм. Соберём c51, он же Categorical DQN, целиком.

\begin{algorithm}[label = c51algorithm]{Categorical DQN (c51)}
\textbf{Гиперпараметры:} $B$ --- размер мини-батчей, $V_{\max}, V_{\min}, A$ --- параметры категориальной аппроксимации, $K$ --- периодичность обновления таргет-сети, $\eps(t)$ --- стратегия исследования, $p_i(s, a, \theta)$ --- нейросетка с параметрами $\theta$, SGD-оптимизатор

\vspace{0.3cm}
Предпосчитать узлы сетки $z_i \coloneqq V_{\min} + \frac{i}{A}(V_{\max} - V_{\min})$ \\
Инициализировать $\theta$ произвольно \\
Положить $\theta^- \coloneqq \theta$ \\
Пронаблюдать $s_0$ \\
\textbf{На очередном шаге $t$:}
\begin{enumerate}
    \item выбрать $a_t$ случайно с вероятностью $\eps(t)$, иначе $a_t \coloneqq \argmax\limits_{a_t} \sum\limits_{i=0}^A z_i p_i(s_t, a_t, \theta)$
    \item пронаблюдать $r_t$,  $s_{t+1}$, $\done_{t+1}$
    \item добавить пятёрку $(s_t, a_t, r_t, s_{t+1}, \done_{t+1})$ в реплей буфер
    \item засэмплировать мини-батч размера $B$ из буфера
    \item для каждого перехода $\T \coloneqq (s, a, r, s', \done)$ посчитать таргет:
    $$\Prob \left( y(\T) = r + \gamma (1 - \done) z_i \right) \coloneqq p_i\left( s', \argmax_{a'} \sum_{i=0}^A z_i p_i(s', a', \theta^-), \theta^- \right) $$
    \item спроецировать таргет на сетку $\{ z_0, z_1 \dots z_{A} \}$: $y(\T) \leftarrow \Pi y(\T)$ 
    \item посчитать лосс:
    $$\Loss(\theta) \coloneqq -\frac{1}{B}\sum_{\T} \sum_{i=0}^{A} \Prob \left( y(\T) = z_i \right) \log p_i(s_t, a_t, \theta) $$
    \item сделать шаг градиентного спуска по $\theta$, используя $\nabla_\theta \Loss(\theta)$
    \item если $t \operatorname{mod} K = 0$: $\theta^- \gets \theta$
\end{enumerate}
\end{algorithm}

\subsection{Квантильная аппроксимация Z-функций}

В c51 мы воспользовались тем, что $\KL$-дивергенция --- это мат.ожидание по одному из сравниваемых распределений. Только это позволило нам несмещённо оценивать градиенты, используя лишь один сэмпл $s'$. Иначе говоря, у нас не ложатся карты: по адекватным метрикам такой фокус не прокатывает --- их нельзя так просто <<оптимизировать по сэмплам>> --- и к тому же у нас есть сложности с доменом распределения, нам необходим оператор проекции и аккуратный подбор неудобных гиперпараметров $V_{\max}, V_{\min}$, которые критично подобрать более-менее правильно.

\needspace{7\baselineskip}
\begin{wrapfigure}{r}{0.35\textwidth}
\vspace{-0.3cm}
\centering
\includegraphics[width=0.35\textwidth]{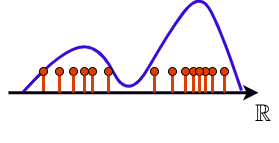}
\vspace{-1.4cm}
\end{wrapfigure}

Оказывается, карты сложатся, если мы выберем другую аппроксимацию распределений в $\Prob (\R)$. Если раньше мы зафиксировали домен (узлы сетки) и подбирали вероятности, то теперь мы зафиксируем вероятности и будем подбирать узлы сетки. На первый взгляд это может показаться странно (как можно отказываться от предсказания вероятностей?), однако на самом деле это весьма гибкое семейство распределений с интересными свойствами. Итак:

\begin{definition}
Обозначим \emph{семейство квантильных распределений} $\mathcal{Q} \subset \Prob(\R)$ с $A$ атомами как множество равномерных дискретных распределений с $A$ произвольными исходами: если $\Z(s, a) \in \mathcal{Q}$, то для некоторых $A$ чисел $z_0, z_1 \dots z_{A-1} \colon$
$$\Prob \left( \Z(s, a) = z_i \right) = \frac{1}{A}$$
\end{definition}

Сразу хорошо то, что нам понадобится всего один гиперпараметр --- число атомов $A$ --- и не понадобится подбирать верхнюю-нижнюю границу ручками. Также заметим, что вырожденное распределение принадлежит $\mathcal{Q}$: просто все $z_i$ в этом случае совпадают.

$\B^*_D \Z_{\theta^-}$, тем не менее, может снова выпадать из такого семейства представлений, и нам всё равно понадобится какая-то проекция. Но на этот раз мы сможем сделать куда более естественную проекцию. На очередном шаге для заданной пары $s, a$ мы будем оптимизировать расстояние Вассерштайна $\W_1$ между правой частью уравнения Беллмана и тем, что мы выдаём:
\begin{equation}\label{qprojection}
\W_1 ([\B^*_D \Z_{\theta^-}](s, a) \parallel \Z) \to \min_{\Z \in \mathcal{Q}}
\end{equation}

Если бы умели выдавать произвольные распределения, мы бы искали $\Z$, условно, среди всех распределений и тогда выдали бы $[\B^*_D \Z_{\theta^-}](s, a)$, получив точный шаг метода простой итерации. Но поскольку мы ограничены семейством квантильных распределений, то лучшее, что мы можем сделать, это спроецировать шаг метода простой итерации в него. 

Ключевой момент: оказывается, задача \eqref{qprojection} имеет аналитическое решение. Введём следующее обозначение (<<середины отрезков сетки>>):
$$\tau_i \coloneqq \frac{\frac{i}{A} + \frac{i + 1}{A}}{2}$$

\begin{theorem}
Пусть $F$ --- функция распределения $[\B^*_D \Z_{\theta^-}](s, a)$. Тогда решение $\Z \in \mathcal{Q}$ задачи \eqref{qprojection} имеет домен $z_0, z_1 \dots z_{A-1}$:
$$z_i = F^{-1}(\tau_i)$$
\begin{proof}
Задача минимизации выглядит так:
\begin{equation}\label{Wassersteinminimization}
\int_0^1 \left| F^{-1}(\omega) - U^{-1}_{z_0, z_1 \dots z_{A-1}}(\omega) \right| \diff \omega \to \min_{z_0, z_1 \dots z_{A-1}}
\end{equation}
где $ U^{-1}_{z_0, z_1 \dots z_{A-1}}$ --- функция распределения равномерного дискретного распределения на домене $\{z_0, z_1 \dots z_{A-1}\}$. Чему она равна? Ну, понятно, что это такая <<лесенка>>:

\needspace{14\baselineskip}
\begin{wrapfigure}{r}{0.4\textwidth}
\vspace{-0.4cm}
\centering
\includegraphics[width=0.4\textwidth]{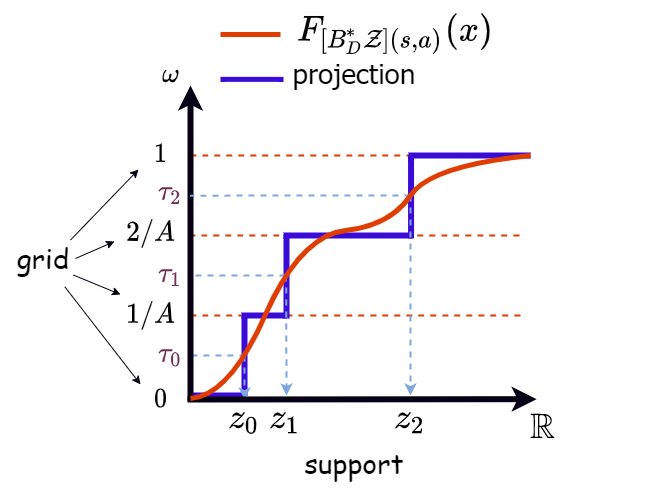}
\vspace{-1.8cm}
\end{wrapfigure}

$$U^{-1}_{z_0, z_1 \dots z_{A-1}}(\omega) = \begin{cases}
z_0 \quad &0 \le \omega < \frac{1}{A} \\
z_1 \quad &\frac{1}{A} \le \omega < \frac{2}{A} \\
\vdots \\
z_{A - 1} \quad &\frac{A - 1}{A} \le \omega < 1
\end{cases}$$

Подставляем это в \eqref{Wassersteinminimization}:
\begin{equation*}
\sum_{i = 0}^{A - 1} \int_{\frac{i}{A}}^{\frac{i + 1}{A}} \left| F^{-1}(\omega) - z_i \right| d\omega \to \min_{z_0, z_1 \dots z_{A-1}}
\end{equation*}

Видим, что задача распадается на $A$ отдельных задач оптимизации:
\begin{equation}\label{w1minim}
\int_{\frac{i}{A}}^{\frac{i + 1}{A}} \left| F^{-1}(\omega) - z_i \right| d\omega \to \min_{z_i}
\end{equation}

Продифференцируем по $z_i$ и приравняем к нулю. Функция $F$ монотонна, поэтому сначала будет кусок интеграла, где градиент равен -1, затем кусок, где +1:
\begin{equation*}
\int_{\frac{i}{A}}^{F(z_i)} -1 \diff \omega + \int_{F(z_i)}^{\frac{i+1}{A}} 1 \diff \omega = 0
\end{equation*}
Берём интегралы от константы:
\begin{equation*}
-\left( F(z_i) - \frac{i}{A} \right) + \frac{i+1}{A} - F(z_i) = 0
\end{equation*}
Откуда видим $F(z_i) = \tau_i$ и получаем доказываемое.
\end{proof}
\end{theorem}

\subsection{Quantile Regression DQN}\label{subsec:qrdqn}

\needspace{7\baselineskip}
\begin{wrapfigure}{r}{0.2\textwidth}
\vspace{-0.5cm}
\centering
\includegraphics[width=0.2\textwidth]{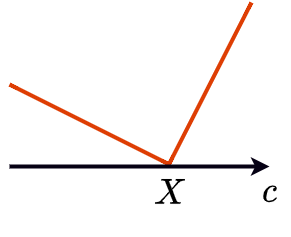}
\vspace{-1.2cm}
\end{wrapfigure}

Мы получили, что нам достаточно уметь искать лишь $A$ определённых квантилей распределения $\left[ \B^*_D \Z_{\theta^-} \right](s, a)$ для вычисления аппроксимации правой части уравнения Беллмана. Можем ли мы это сделать, используя только сэмплы? Конечно. 

\emph{Квантильная регрессия} (quantile regression) --- способ получить $\tau$-ый квантиль некоторого распределения, из которого доступна только лишь выборка. В частном случае, мы получим известный факт о том, что для получения медианы ($\frac{1}{2}$-го квантиля) нужно минимизировать MAE между константным прогнозом и сэмплами из распределения.

\begin{definition}
Для заданного $\tau \in (0, 1)$ \emph{квантильной функцией потерь} (quantile loss) называется:
\begin{equation}\label{quantileloss}
\Loss_{\tau}(c, X) \coloneqq \begin{cases}
\tau (c - X) \quad &c \ge X \\
(1 - \tau) (X - c) \quad &c < X \\
\end{cases}
\end{equation}
\end{definition}

\begin{theorem}[Квантильная регрессия] Решением задачи
\begin{equation}\label{QR}
\E_X \Loss_{\tau}(c, X) \to \min_{c \in \R}
\end{equation}
будет $\tau$-ый квантиль распределения случайной величины $X$.

\beginproof
Пусть $F$ --- функция распределения $X$. Распишем \eqref{QR}: для заданной точки $c$ интеграл будет состоять из двух слагаемых, где в первом $c < X$, а во втором $c \ge X$:
\begin{equation*}
\int_0^{F^{-1}(c)} (1 - \tau)(X - c) \diff \omega + \int_{F^{-1}(c)}^1 \tau (c - X) \diff \omega = 0 
\end{equation*}
Дифференцируем по $c$ и приравниваем к нулю:
\begin{equation*}
\int_0^{F^{-1}(c)} (\tau - 1) \diff \omega + \int_{F^{-1}(c)}^1 \tau \diff \omega = 0 
\end{equation*}
Берём константные интегралы:
\begin{equation*}
F^{-1}(c)(\tau - 1) + (1 - F^{-1}(c)) \tau = -F^{-1}(c) + \tau = 0
\end{equation*}
Отсюда получаем, что $F^{-1}(c) = \tau$, то есть $c$ --- $\tau$-ый квантиль. \qed
\end{theorem}

\begin{proposition}
Формулу \eqref{quantileloss} можно переписать <<в одну строчку>> в следующем виде:
$$\Loss_{\tau}(c, X) = \left( \tau - \mathbb{I}[c < X] \right) \left( c - X \right)$$
\end{proposition}

\needspace{7\baselineskip}
\begin{wrapfigure}{r}{0.2\textwidth}
\centering
\includegraphics[width=0.2\textwidth]{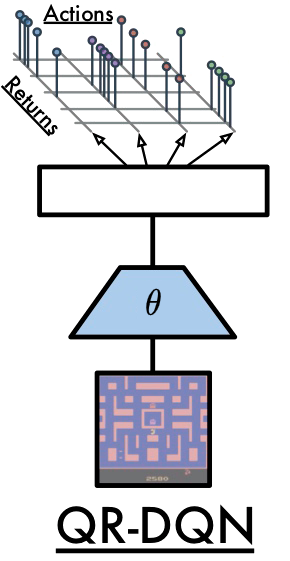}
\vspace{-0.5cm}
\end{wrapfigure}
Итак, соберём всё вместе. У нас есть нейросеть $z_i(s, a, \theta)$ с параметрами $\theta$, которая для данного состояния-действия выплёвывает $A$ произвольных вещественных чисел, которые мы интерпретируем как $A$ равновероятных возможных исходов случайной величины $\Z_{\theta}(s, a)$. Обозначим за $\theta^-$ веса таргет-сети, как обычно. Для очередного перехода $\T \coloneqq (s, a, r, s')$ из буфера мы хотим провести оптимизацию
\begin{equation}\label{wassersteinminim}
\W_1(\left[\B^*_D \Z_{\theta^-} \right](s, a) \parallel \Z_{\theta}(s, a)) \to \min_\theta
\end{equation}
и мы поняли, что это эквивалентно поиску квантилей распределения $\left[\B^*_D \Z_{\theta^-} \right](s, a)$, поэтому для оптимизации $i$-го выхода нейросетки будем оптимизировать квантильный лосс \eqref{quantileloss} (по $i$ просто просуммируем --- хотим учить все $A$ интересующих нас квантилей):
$$\sum_{i=0}^{A-1} \E_{x \sim \left[\B^*_D \Z_{\theta^-} \right](s, a)} \Loss_{\tau_i}(z_i(s, a, \theta), x) \to \min_{\theta}$$

Конечно же, мы не будем доводить этот процесс оптимизации до конца, а сделаем всего один шаг обновления весов $\theta$ по градиенту, а затем сразу же возьмём из буфера другой мини-батч. Мы лишь получили способ получать несмещённые оценки градиентов, указывающих в сторону минимизации \eqref{wassersteinminim}.

Опять же заметим, что $\E_{x \sim \left[\B^*_D \Z_{\theta^-} \right](s, a)}$ распадается в сэмплирование $s'$ и интегрирование по возможным исходам $\Z_{\theta^-}(s', \pi^*(s'))$, где $\pi^*(s')$ выбирает действие жадно. Мат.ожидание по $\Z_{\theta^-}(s', a')$ при данных $s', a'$ есть просто усреднение по $A$ равновероятным исходам, поэтому его мы посчитаем явно. Итого:
$$\underbrace{\sum_{i=0}^{A-1}}_{\mathclap{\text{\shortstack{учим $A$ \\ квантилей}}}} \overbrace{\E_{s'} \frac{1}{A}}^{\mathclap{\text{\shortstack{вероятности \\ сэмплов}}}}\sum_{j=0}^{A-1} \Loss_{\tau_i}(\underbrace{z_i(s, a, \theta)}_{\text{прогноз}}, \overbrace{r + \gamma z_j(s', a', \theta^{-})}^{\text{сэмпл}}) \to \min_{\theta}$$
Занося внешнюю сумму под мат.ожидание по $s'$, получаем функцию потерь, градиент которой можно оценивать по Монте-Карло, используя лишь сэмплы $s'$ из функции переходов.

\begin{algorithm}[label = QRDQNalgorithm]{Quantile Regression DQN (QR-DQN)}
\textbf{Гиперпараметры:} $B$ --- размер мини-батчей, $A$ --- число атомов, $K$ --- периодичность обновления таргет-сети, $\eps(t)$ --- стратегия исследования, $z_i(s, a, \theta)$ --- нейросетка с параметрами $\theta$, SGD-оптимизатор

\vspace{0.3cm}
Предпосчитать середины отрезков квантильной сетки $\tau_i \coloneqq \frac{\frac{i}{A} + \frac{i+1}{A}}{2}$ \\
Инициализировать $\theta$ произвольно \\
Положить $\theta^- \coloneqq \theta$ \\
Пронаблюдать $s_0$ \\
\textbf{На очередном шаге $t$:}
\begin{enumerate}
    \item выбрать $a_t$ случайно с вероятностью $\eps(t)$, иначе $a_t \coloneqq \argmax\limits_{a_t} \sum_{i=0}^{A-1} z_i(s_t, a_t, \theta)$
    \item пронаблюдать $r_t$,  $s_{t+1}$, $\done_{t+1}$
    \item добавить пятёрку $(s_t, a_t, r_t, s_{t+1}, \done_{t+1})$ в реплей буфер
    \item засэмплировать мини-батч размера $B$ из буфера
    \item для каждого перехода $\T \coloneqq (s, a, r, s', \done)$ посчитать таргет:
    $$y(\T)_j \coloneqq r + (1 - \done)\gamma z_j\left( s', \argmax\limits_{a'} \sum_{i} z_i(s', a', \theta^-), \theta^- \right) $$
    \item посчитать лосс:
    $$\Loss(\theta) \coloneqq \frac{1}{BA}\sum_{\T} \sum_{i=0}^{A-1} \sum_{j=0}^{A-1} \left( \tau_i - \mathbb{I}[z_i(s, a, \theta) < y(\T)_j] \right) \left( z_i(s, a, \theta) - y(\T)_j \right)$$
    \item сделать шаг градиентного спуска по $\theta$, используя $\nabla_\theta \Loss(\theta)$
    \item если $t \operatorname{mod} K = 0$: $\theta^- \gets \theta$
\end{enumerate}
\end{algorithm}

\subsection{Implicit Quantile Networks}\label{subsec:iqn}

В QR-DQN мы фиксировали <<равномерную сетку>> на оси квантилей: говорили, что наше аппроксимирующее распределение есть равномерное на домене из $A$ атомов. Идея: давайте будем уметь в нашей нейросети выдавать произвольные квантили, каким-то образом задавая $\tau \in (0, 1)$ дополнительно на вход. Тогда наша модель $z(s, a, \tau, \theta)$ будет неявно (implicit) задавать, вообще говоря, произвольное распределение на $\R$. По сути, мы моделируем квантильную функцию <<целиком>>; очень удобно:
$$F^{-1}_{\Z_{\theta}(s, a)}(\tau) \coloneqq z(s, a, \tau, \theta)$$

Поймём, как тогда считать мат.ожидание (или Q-функцию) в такой модели.

\begin{theoremBox}[label=th:samplingimplicit]{}
Пусть $F$ --- функция распределения случайной величины $X$. Тогда, если $\tau \sim U[0, 1]$, случайная величина $F^{-1}(\tau)$ имеет то же распределение, что и $X$.
\beginproof
Заметим, что функция распределения равномерной случайной величины при $x \HM\in [0, 1]$ равна $\Prob(\tau \HM< x) \HM= x$. Теперь посмотрим на функцию распределения случайной величины $F^{-1}(\tau)$:
\begin{equation*}
\Prob \left( F^{-1}(\tau) < x \right) = \Prob \left( \tau < F(x) \right) = F(x) \tagqed   
\end{equation*}
\end{theoremBox}

\needspace{7\baselineskip}
\begin{wrapfigure}{r}{0.25\textwidth}
\vspace{-0.3cm}
\centering
\includegraphics[width=0.25\textwidth]{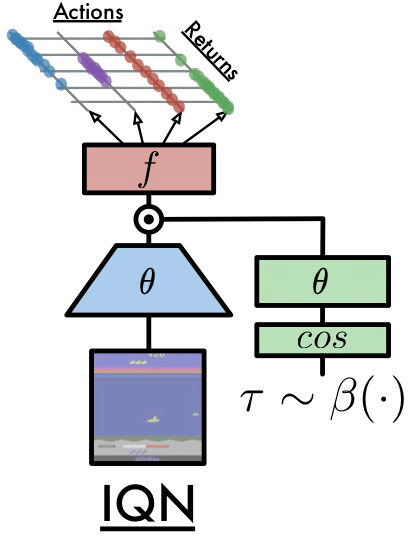}
\vspace{-0.5cm}
\end{wrapfigure}

Итак, мы можем аппроксимировать жадную стратегию примерно так:
$$\pi^*(s) \coloneqq \argmax\limits_{a} \sum_{i=0}^{N} z(s, a, \tau_i, \theta), \qquad \tau_i \sim U[0, 1]$$

В качестве функции потерь предлагается использовать тот же квантильный лосс, что и в QR-DQN, только если в QR-DQN нам были нужны определённые $A$ квантилей, то теперь предлагается засэмплировать $N'$ каких-то квантилей и посчитать лосс для них. Для подсчёта лосса нам было нужно брать мат.ожидание по $\Z_{\theta^-}(s', a')$, для чего в формуле мы пользовались тем, что это распределение в нашей модели равномерно. Теперь же этот интеграл мы заменяем на Монте-Карло оценку с произвольным числом сэмплов $N''$, а для сэмплирования опять же используем теорему \ref{th:samplingimplicit}:
$$\Loss(\T, \theta) \coloneqq \sum_{i=0}^{N'} \frac{1}{N''} \sum_{j=0}^{N''} \Loss_{\tau_i}(z(s, a, \tau_i, \theta), r + \gamma z(s', \pi^*(s'), \tau_j, \theta^{-})),$$
где $\tau_i, \tau_j \sim U[0, 1]$.

\begin{remark}
Возможно так обучать стратегию, которая в меньшей или большей степени предпочитает рисковать. Если при расчёте $\pi$ сэмплировать квантили $\tau$ не из равномерного распределения, а чаще брать квантили, близкие к 1, агент будет в большей степени смотреть на награду, которую он может получить при везении. Если же чаще сэмплировать квантили, близкие к 0, агент будет избегать рискованных ситуаций, когда есть вероятность получить низкую награду, и это может быть полезно в Safe RL. Поэтому в общем случае можно считать, что квантили генерируются в этой схеме из некоторого распределения $\beta(\tau)$. 
\end{remark}

\begin{remark}
Архитектура сети предлагается такая. Входное состояние сжимается в некоторый эмбеддинг при помощи основной части сети $f_{\theta}(s)$. Параллельно строится некоторый эмбеддинг $g_{\theta}(\tau)$, описывающий квантиль $\tau \in (0, 1)$: тут могут быть разные варианты, но авторы остановились на следующем: число $\tau$ <<описывается>> несколькими признаками, где $i$-ый признак равен $\cos(\pi i \tau)$, и преобразуется одним полносвязным слоем (с обучаемыми параметрами) для получения $g_{\theta}(\tau)$. Дальше финальное преобразование $h$ должно взять $f_{\theta}(s)$ и $g_{\theta}(\tau)$ и выдать по одному числу для каждого действия $a$; чтобы не городить ещё слои, взаимодействие этих двух эмбеддингов предлагается организовать при помощи поэлементного перемножения:
$$z(s, a, \tau, \theta) \coloneqq h_{\theta}(f_{\theta}(s) \odot g_{\theta}(\tau))$$
\end{remark}

\subsection{Rainbow DQN}\label{subsec:rainbow}

В разделе про модификации \ref{sec:dqnmods} были рассмотрены весьма разные улучшения DQN, нацеленные на решения очень разных проблем. Хорошо видно, что все эти модификации <<ортогональны>> и могут включаться-выключаться, так сказать, независимо в алгоритм. Distributional-подход, вообще говоря, не решает какую-то проблему внутри DQN, но может рассматриваться как ещё одна модификация базового алгоритма DQN.

Rainbow DQN совмещает 6 модификаций алгоритма DQN:
\begin{itemize}
    \item Double DQN (раздел \ref{subsec:doubledqn})
    \item Dueling DQN (раздел \ref{subsec:duelingdqn})
    \item Noisy DQN  (раздел \ref{subsec:noisynets})
    \item Prioritized Experience Replay (раздел \ref{subsec:prioritizedreplay})
    \item Multi-step DQN (раздел \ref{subsec:multistepdqn})
    \item Distributional RL
\end{itemize}

В последнем пункте исторически в Rainbow используется Categorical DQN (раздел \ref{subsec:c51}), однако понятно, что можно использовать любой другой алгоритм; в частности, QR-DQN (раздел \ref{subsec:qrdqn}) или IQN (раздел \ref{subsec:iqn}). Обсудим только пару нюансов, которые возникают при совмещении некоторых пар из этих идей, а дальше приведём полный-полный алгоритм.

Совмещение приоритизированного реплея и Distributional-подхода требует введения приоритета перехода $\T$: в его качестве, аналогично обычному случаю \eqref{priority}, берётся значение функции потерь \eqref{c51loss}: 
$$\rho(\T) \coloneqq \KL(y(\T) \parallel \Z_{\theta}(s, a))$$

При совмещении шумных сетей с эвристикой Double DQN, шум сэмплируется заново на каждом проходе через сеть или таргет-сеть (то есть генерится отдельный сэмпл шума для выбора действия $a'$, отдельный для оценивания при построении таргета и отдельный для прохода через сеть для подсчёта градиентов).

Забивать костыльми приходится совмещение Categorical DQN с Dueling DQN. Здесь остаётся только идея о том, что при обновлении модели для пары $s, a$ мы должны <<легче обобщаться>> на все остальные действия $\hat{a} \in \A$, для чего вычисление $p_i(s, a, \theta)$ проходит в два потока: <<типа>> V-поток $V_{\theta}(s)$, который выдаёт $A$ атомов как бы общих для всех действий, и <<типа>> A-поток $A_{\theta}(s, a)$, который выдаёт $A$ атомов для каждого из $|\A|$ действий. Дальнейшая формула <<взята>> из обычного Dueling DQN \eqref{dueling}; софтмакс, необходимый, чтобы получить на выходе валидное категориальное распределение, применяется в самом конце:
\begin{equation}\label{rainbowdueling}
p_i(s, a, \theta) \coloneqq \softmax\limits_{i} \left( V_{\theta}(s)_i + A_{\theta}(s, a)_i - \frac{1}{|\A|}\sum_a A_{\theta}(s, a)_i \right)
\end{equation}
Последнее слагаемое --- <<типа>> централизация A-потока к нулю, снова со средним вместо максимума (хотя эта логика к вероятностям исходов $\Z$ уже плохо применима).

\begin{remark}
Ablation study там показывал, что убирание Dueling DQN, в отличие от остальных 5 модификаций, к особому падению качества итогового алгоритма не приводит. Вероятно, это связано с тем, что <<семантика>> потоков ещё больше теряется. Стоит отметить, что если использовать QR-DQN вместо c51, то софтмакс становится не нужен, и формула становится <<более логичной>>.
\end{remark}

\begin{algorithm}[label = rainbowalg]{Rainbow DQN}
\textbf{Гиперпараметры:} $B$ --- размер мини-батчей, $V_{\max}, V_{\min}, A$ --- параметры категориальной аппроксимации, $K$ --- периодичность обновления таргет-сети, $N$ --- количество шагов в оценке, $\alpha$ --- степень приоритизации, $\beta(t)$ --- параметр importance sampling коррекции для приоритизированного реплея, $p_i(s, a, \theta, \eps)$ --- нейросетка с параметрами $\theta$, SGD-оптимизатор

\vspace{0.3cm}
Предпосчитать узлы сетки $z_i \coloneqq V_{\min} + \frac{i}{A}(V_{\max} - V_{\min})$ \\
Инициализировать $\theta$ произвольно \\
Положить $\theta^- \coloneqq \theta$ \\
Пронаблюдать $s_0$ \\
\textbf{На очередном шаге $t$:}
\begin{enumerate}
    \item выбрать $a_t \coloneqq \argmax\limits_{a_t} \sum_{i=0}^A z_i p_i(s_t, a_t, \theta, \eps), \quad \eps \sim \N(0, I)$
    \item пронаблюдать $r_t$, $s_{t+1}$, $\done_{t+1}$
    \item построить $N$-шаговый переход $\T \coloneqq \left( s, a, \sum_{n=0}^N \gamma^n r^{(n)}, s^{(N)}, \done \right)$, используя последние $N$ наблюдений, и добавить его в реплей буфер с максимальным приоритетом 
    \item засэмплировать мини-батч размера $B$ из буфера, используя вероятности $\Prob (\T) \propto \rho(\T)^\alpha$
    \item посчитать веса для каждого перехода:
    $$w(\T) \coloneqq \frac{1}{\rho(\T)^{\beta(t)}}$$
    \item для каждого перехода $\T \coloneqq (s, a, \bar{r}, \bar{s}, \done)$ посчитать таргет:
    $$\eps_1, \eps_2 \sim \N(0, I)$$
    $$\Prob \left( y(\T) = \bar{r} + \gamma^N (1 - \done) z_i \right) \coloneqq p_i\left( \bar{s}, \argmax_{\bar{a}} \sum_{i=0}^A z_i p_i(\bar{s}, \bar{a}, \theta, \eps_1), \theta^-, \eps_2 \right)$$
    \item спроецировать таргет на сетку $\{ z_0, z_1 \dots z_{A} \}$: $y(\T) \leftarrow \Pi y(\T)$ 
    \item посчитать для каждого перехода лосс:
    $$L(\T, \theta) \coloneqq -\sum_{i=0}^{A} \Prob \left( y(\T) = z_i \right) \log p_i(s_t, a_t, \theta, \eps) \quad \eps \sim \N(0, I)$$
    \item обновить приоритеты всех переходов из буфера: $\rho(\T) \gets L(\T, \theta)$
    \item посчитать суммарный лосс:
    $$\Loss(\theta) \coloneqq \frac{1}{B}\sum_{\T} w(\T)L(\T, \theta)$$
    \item сделать шаг градиентного спуска по $\theta$, используя $\nabla_\theta \Loss(\theta)$
    \item если $t \operatorname{mod} K = 0$: $\theta^- \gets \theta$
\end{enumerate}
\end{algorithm}

\begin{remark}
При решении задач дискретного управления в off-policy режиме имеет смысл выбирать Rainbow DQN, но задумываться о том, какие модули нужны, а какие можно отключить. Дело в том, что каждая модификация DQN несколько увеличивает вычислительную сложность каждой итерации. Использование тех модулей, которые несущественны для решаемой задачи, может ускорить работу алгоритма; однако, на практике часто сложно сказать, какие модули важны в том или ином случае. Полезно помнить, какую проблему решает каждая модификация, и пытаться отслеживать, возникает ли она.
\end{remark}

\begin{remark}
Несмотря на увесистую теорию, на практике код Distributional RL алгоритмов отличается от кода DQN буквально несколькими строчками: нужно лишь поменять размерность выхода нейронной сети и поменять функцию потерь. Поэтому их имеет смысл обязательно попробовать при экспериментировании с value-based подходом; обучение всего распределения будущей награды вместо среднего может значительно повысить sample efficiency алгоритма.
\end{remark}

\chapter{Policy Gradient подход}\label{policygradientchapter}

В данной главе будет рассмотрен третий, Policy Gradient подход к решению задачи, в котором целевой функционал будет оптимизироваться градиентными методами. Для этого мы выведем формулу градиента средней награды по параметрам стратегии и обсудим различные способы получения его стохастических оценок. В итоге мы сможем получить общие алгоритмы, основным ограничением которых будет жёсткий on-policy режим.

\section{Policy Gradient Theorem}\label{PGTsection}

\subsection{Вывод первым способом}

Часто говорят, что функционал в задаче обучения с подкреплением не дифференцируем. Имеется в виду, что функция награды $r(s, a)$ не дифференцируема по действиям $a$; например, просто потому что пространство действий дискретно, например, в состоянии $s$ агент выбрал действие $a \HM = 0$ и значение полученной награды можно лишь сравнивать со значениями для других действий. Однако, мы уже видели в примере \ref{ex:score}, что это не мешает дифференцируемости по параметрам стратегии в ситуации, когда стратегия ищется в семейсте стохастичных стратегий. Фактически, оптимизация в пространстве стохастичных стратегий является этакой <<релаксацией>> нашей задачи.

Пусть политика $\pi_{\theta}(a \mid s)$ параметризована $\theta$ и дифференцируема по параметрам. Тогда наш оптимизируемый функционал $J(\pi_\theta) \HM= V^{\pi_\theta}(s_0)$ тоже дифференцируем по $\theta$, и далее мы будем выводить формулу этого градиента. Для этого нам понадобится стандартная техника вычисления градиента мат.ожидания по распределению, зависящего от параметров; мы уже встречались с ней при обсуждении эволюционных стратегий в главе \ref{subsec:nes}. Сейчас в оптимизируемом функционале у нас стоит целая цепочка вложенных мат.ожиданий, и наш вывод будет заключаться просто в последовательном применении той же техники к каждому стоящему там мат.ожиданию $\E_{a \sim \pi(a \mid s)} ( \cdot )$.

Заранее оговоримся, что при минимальных технических условиях регулярности\footnote{это следует из наших условий регулярности на MDP и предположения интегрируемости всех функций; тогда все оценочные функции и награды ограничены, следовательно, все интегралы и ряды в рассуждении сходятся абсолютно и равномерно по параметрам $\theta$; мы просто не <<связываемся>> в контексте нашей задачи с бесконечностью, на которой в теории математического анализа и возникают ситуации, когда так делать нельзя.} мы имеем право менять местами знаки градиента, мат.ожиданий, сумм и интегралов.

\begin{theorem}
\begin{equation}\label{Vgrad}
\nabla_{\theta} V^{\pi_\theta}(s) = \E_{a} \left[ \nabla_{\theta} \log \pi_\theta (a \mid s) Q^{\pi}(s, a) + \nabla_{\theta} Q^{\pi_\theta}(s, a) \right]
\end{equation}
\beginproof
\begin{align*}
\nabla_{\theta} V^{\pi_\theta}(s) &= \{ \text{\eqref{VQ}} \} = \nabla_{\theta} \E_{a \sim \pi_\theta(a \mid s)} Q^{\pi_\theta}(s, a) = \\
&= \{ \text{мат.ожидание --- это интеграл} \} = \\
&= \nabla_{\theta} \int\limits_{\A} \pi_\theta(a \mid s) Q^{\pi_\theta}(s, a) \diff a = \\
&= \{ \text{проносим градиент внутрь интеграла} \} = \\
&= \int\limits_{\A} \nabla_{\theta} \left[ \pi_\theta(a \mid s) Q^{\pi_\theta}(s, a) \right] \diff a = \\
&= \{ \text{правило градиента произведения} \} = \\
&= \int\limits_{\A} \nabla_{\theta} \pi_\theta(a \mid s) Q^{\pi}(s, a) \diff a + \int\limits_{\A} \pi(a \mid s) \nabla_{\theta} Q^{\pi_\theta}(s, a) \diff a = \\
&= \{ \text{второе слагаемое --- это мат.ожидание} \} = \\
&= \int\limits_{\A} \nabla_{\theta} \pi_\theta(a \mid s) Q^{\pi}(s, a) \diff a + \E_{a} \nabla_{\theta} Q^{\pi_\theta}(s, a) = \\
&= \{ \text{log-derivative trick \eqref{logderivtrick}} \} = \\
&= \int\limits_{\A} \pi_\theta(a \mid s) \nabla_{\theta} \log \pi_\theta(a \mid s) Q^{\pi}(s, a) \diff a + \E_{a} \nabla_{\theta} Q^{\pi_\theta}(s, a) = \\
&= \{ \text{первое слагаемое тоже стало мат.ожиданием} \} = \\
&= \E_{a} \left[ \nabla_{\theta} \log \pi_\theta(a \mid s) Q^{\pi}(s, a) + \nabla_{\theta} Q^{\pi_\theta}(s, a) \right] \tagqed
\end{align*}
\end{theorem}

Эта техника вычисления градиента через <<стохастичный узел нашего вычислительного графа>>, когда мы сэмплируем $a \HM \sim \pi(a \HM\mid s)$, носит название REINFORCE. Как видно, эта техника универсальна: применима всегда для любых пространств действий, а также в ситуации, когда функция $Q^{\pi}(s, a)$ не дифференцируема по действиям. Заметим, что в глубоком обучении при некоторых дополнительных условиях может быть также применим другой способ; его мы обсудим позже в главе \ref{continuouscontrolchapter}, когда пространство действий будет непрерывно, а $Q^{\pi}(s, a)$ --- дифференцируема по действиям, и альтернативный метод будет применим. 

Мы смогли выразить градиент V-функции через градиент Q-функции, попробуем сделать наоборот. Для этого нам нужно посчитать градиент от мат.ожидания по функции переходов, не зависящей от параметров нашей стратегии, поэтому здесь всё тривиально. 

\begin{proposition}
\begin{equation}\label{Qgrad}
\nabla_{\theta} Q^{\pi_\theta}(s, a) = \gamma \E_{s'} \nabla_{\theta} V^{\pi_\theta}(s')
\end{equation}
\beginproof
\begin{align*}
\nabla_{\theta} Q^{\pi_\theta}(s, a) &= \{ \text{\eqref{QV}} \} = \nabla_{\theta} \left[ r(s, a) + \gamma \E_{s'} V^{\pi_\theta}(s') \right] = \\
&= \{ \text{$r(s, a)$ не зависит от $\theta$, $p(s' \mid s, a)$ тоже} \} = \\
&= \gamma \E_{s'} \nabla_{\theta} V^{\pi_\theta}(s')  \tagqed
\end{align*}
\end{proposition}

Подставляя \eqref{Qgrad} в \eqref{Vgrad}, получаем:

\begin{proposition}
\begin{equation}
\nabla_{\theta} V^{\pi_\theta}(s) = \E_{a} \E_{s'} \left[ \nabla_{\theta} \log \pi_\theta (a \mid s) Q^{\pi}(s, a) + \gamma \nabla_{\theta} V^{\pi_\theta}(s') \right]
\end{equation}
\end{proposition}

Следовательно, мы получили рекурсивное выражение градиента $V^{\pi_\theta}(s)$ через него же само. Очень похоже на уравнение Беллмана, кстати: в правой части стоит мат.ожидание по выбранному в $s$ действию $a$ и следующему состоянию.

Осталось раскрутить эту рекурсивную цепочку, продолжая раскрывать $\nabla_{\theta} V^{\pi_\theta}(s')$ в будущее до бесконечности. Аккуратно собирая слагаемые, а также собирая из мат.ожиданий мат.ожидание по траектории, получаем следующее:

\begin{proposition}[Policy Gradient Theorem] Выражение для градиента оптимизируемого функционала можно записать следующим образом:
\begin{equation}\label{pgt_firstproof}
\nabla_{\theta} J(\pi_\theta) = \E_{\Traj \sim \pi} \sum_{t \ge 0} \gamma^t \nabla_{\theta} \log \pi_\theta (a_t \mid s_t) Q^{\pi}(s_t, a_t)
\end{equation}
\end{proposition}

\subsection{Вывод вторым способом}

Прежде, чем мы обсудим физический смысл полученной формулы, выведем её альтернативным способом. Применим REINFORCE не к мат.ожиданиям по отдельным действиям, а напрямую к распределению всей траектории следующим образом:

\begin{theorem}
\begin{equation*}
\nabla_{\theta} V^{\pi_\theta}(s) = \E_{\Traj \sim \pi \mid s_0 = s} \sum_{t \ge 0} \nabla_{\theta} \log \pi_\theta (a_t \mid s_t) R(\Traj)
\end{equation*}
\begin{proof}
\begin{align*}
\nabla_{\theta} V^{\pi_\theta}(s) &= \nabla_{\theta} \E_{\Traj \sim \pi \mid s_0 = s} R(\Traj) = \\
&= \{ \text{рассмотрим мат.ожидание как интеграл по всевозможным траекториям} \} = \\
&= \nabla_{\theta} \iint\limits_{\Traj} p\left( \Traj \mid \pi, s_0 = s \right) R(\Traj) \diff \Traj = \\
&= \{ \text{проносим градиент внутрь интеграла} \} = \\
&= \iint\limits_{\Traj} \nabla_{\theta} p\left( \Traj \mid \pi, s_0 = s \right) R(\Traj) \diff \Traj = \\
&= \{ \text{log-derivative trick \eqref{logderivtrick}} \} = \\
&= \E_{\Traj \sim \pi \mid s_0 = s} \nabla_{\theta} \log p\left( \Traj \mid \pi, s_0 = s \right) R(\Traj) = \\
&= \{ \text{вспоминаем, что вероятность траектории есть произведение вероятностей} \} = \\
&= \E_{\Traj \sim \pi \mid s_0 = s} \sum_{t \ge 0} \nabla_{\theta} \log \pi_\theta (a_t \mid s_t) R(\Traj)
\end{align*}
В последнем переходе логарифмы вероятностей переходов $\sum\limits_{t \ge 0} \nabla_{\theta} \log p(s_{t+1} \mid s_t, a_t)$ сократились как не зависящие от параметров стратегии.
\end{proof}
\end{theorem}

Видим, что мы более простым способом получили очень похожую формулу, но с суммарной наградой за игры вместо Q-функции из первого доказательства \eqref{pgt_firstproof}. В силу корректности всех вычислений, уже можно утверждать равенство между этими формулами, что наводит на мысль, что градиент можно записывать в нескольких математически эквивалентных формах. Математически эти формы будут эквивалентны, то есть равны, как интегралы, но их Монте-Карло оценки могут начать вести себя совершенно по-разному. Может быть, мы можем как-то <<хорошую форму>> выбрать для наших алгоритмов.

Рассмотрим, как можно этим вторым способом рассуждений дойти формально до формы градиента из первого способа. Раскрывая $R(\Traj)$ по определению, мы сейчас имеем произведение двух сумм под интегралом:
$$\nabla_{\theta} V^{\pi_\theta}(s) = \E_{\Traj \sim \pi \mid s_0 = s} \left(\sum_{t \ge 0} \nabla_{\theta} \log \pi_\theta (a_t \mid s_t) \right) \left( \sum_{\hat{t} \ge 0} \gamma^{\hat{t}} r_{\hat{t}} \right)$$

Давайте перемножим эти два ряда:
\begin{equation}\label{fulltwosums}
\nabla_{\theta} V^{\pi_\theta}(s) = \E_{\Traj \sim \pi \mid s_0 = s} \sum_{t \ge 0} \sum_{\hat{t} \ge 0} \nabla_{\theta} \log \pi_\theta (a_t \mid s_t) \gamma^{\hat{t}} r_{\hat{t}}
\end{equation}

Видим странную вещь: на градиент по параметрам за решение выбрать $a_t$ в момент времени $t$ влияет награда, собранная при $\hat{t} \HM< t$, то есть величина, на которую наше решение точно не могло повлиять, поскольку принималось после её получения. Но почему формально это так?

\begin{theorem}
Для произвольного распределения $\pi_{\theta}(a)$ с параметрами $\theta$, верно:
\begin{equation}\label{baseline}
\,
\E_{a \sim \pi_{\theta}(a)} \nabla_\theta \log \pi_{\theta}(a) = 0 
\end{equation}
\beginproof
\begin{align*}
\E_{a \sim \pi_{\theta}(a)} \nabla_\theta \log \pi_{\theta}(a) &= \{ \text{производная логарифма} \} 
= \E_{a \sim \pi_{\theta}(a)} \frac{\nabla_\theta \pi_{\theta}(a)}{\pi_{\theta}(a)} = \\
&= \int\limits_\A \nabla_\theta \pi_{\theta}(a) \diff a = \nabla_\theta \int\limits_\A \pi_{\theta}(a) \diff a
= \nabla_\theta 1 = 0 \tagqed
\end{align*}
\end{theorem}

Следующее утверждение формализует этот тезис о том, что <<будущее не влияет на прошлое>>: выбор действий в некоторый момент времени никак не влияет на те слагаемые из награды, которые были получены в прошлом.

\begin{theorem}[Принцип причинности (causality)] 
При $t > \hat{t}$:
\begin{equation*}
\E_{\Traj \sim \pi} \nabla_{\theta} \log \pi_\theta (a_t \mid s_t) \gamma^{\hat{t}} r_{\hat{t}} = 0
\end{equation*}
\beginproof
\begin{align*}
&\E_{\Traj \sim \pi} \nabla_{\theta} \log \pi_\theta (a_t \mid s_t) \gamma^{\hat{t}} r_{\hat{t}} = \\
&= \{ \text{представляем мат.ожидание по траектории как вложенные мат.ожидания} \} = \\
&= \E_{a_1, s_1 \dots s_{\hat{t}}, a_{\hat{t}}} \E_{s_{\hat{t}+1}, a_{\hat{t}+1} \dots s_t, a_t \dots} \nabla_{\theta} \log \pi_\theta (a_t \mid s_t) \gamma^{\hat{t}} r_{\hat{t}} = \\
&= \{ \text{выносим константу из мат.ожидания} \} = \\
&= \E_{a_1, s_1 \dots s_{\hat{t}}, a_{\hat{t}}} \gamma^{\hat{t}} r_{\hat{t}} \E_{s_{\hat{t}+1}, a_{\hat{t}+1} \dots s_t, a_t \dots} \nabla_{\theta} \log \pi_\theta (a_t \mid s_t) = \\
&= \{ \text{мат.ожидание градиента логарифма вероятности есть ноль \eqref{baseline}} \} = \\
&= \E_{a_1, s_1 \dots s_{\hat{t}}, a_{\hat{t}}} \gamma^{\hat{t}} r_{\hat{t}} \cdot 0 = 0   \tagqed
\end{align*}
\end{theorem}

Значит, вместо полной награды за игру можно в весах оставить только reward-to-go \eqref{rewardtogo}, поскольку из всех слагаемых в \eqref{fulltwosums} слагаемые для $\hat{t} \HM< t$ занулятся. Понятно, что дисперсия Монте-Карло оценки такого интеграла будет меньше: мы убрали некоторую зашумляющую часть нашей стохастической оценки, которая, как мы теперь поняли, в среднем равна нулю.

При этом дисконтирование в сумме наград шло с самого начала игры, поэтому для того, чтобы записать формулу в терминах reward-to-go, нужно вынести $\gamma^t$:
$$\sum_{\hat{t} \ge t} \gamma^{\hat{t}} r_{\hat{t}} = \gamma^t \sum_{\hat{t} \ge t} \gamma^{\hat{t} - t} r_{\hat{t}} = \gamma^t R_t$$

\begin{proposition}
\begin{equation}\label{reinforce_grad}
\nabla_{\theta} V^{\pi_\theta}(s) = \E_{\Traj \sim \pi \mid s_0 = s} \sum_{t \ge 0} \gamma^t \nabla_{\theta} \log \pi_\theta (a_t \mid s_t) R_t
\end{equation}
\end{proposition}

Reward-to-go очень похож на Q-функцию, так как является Монте-Карло оценкой Q-функции, а мат.ожидание по распределениям всё равно берётся. Формальное обоснование эквивалентности выглядит так:

\begin{proposition} Формула \eqref{reinforce_grad} эквивалентна
\begin{equation*}
\nabla_{\theta} V^{\pi_\theta}(s) = \E_{\Traj \sim \pi \mid s_0 = s} \sum_{t \ge 0} \gamma^t \nabla_{\theta} \log \pi_\theta (a_t \mid s_t) Q^{\pi}(s_t, a_t)
\end{equation*}
\beginproof
\begin{align*}
\nabla_{\theta} V^{\pi_\theta}(s) &= \E_{\Traj \sim \pi \mid s_0 = s} \sum_{t \ge 0} \gamma^t \nabla_{\theta} \log \pi_\theta (a_t \mid s_t) R_t = \\
&= \{ \text{меняем местами сумму по $t$ и мат.ожидание по траекториям} \} = \\
&= \sum_{t \ge 0} \E_{\Traj \sim \pi \mid s_0 = s} \gamma^t \nabla_{\theta} \log \pi_\theta (a_t \mid s_t) R_t = \\
&= \{ \text{представляем мат.ожидание по траектории как вложенные мат.ожидания} \} = \\
&= \sum_{t \ge 0} \E_{a_0, s_1 \dots s_t, a_t} \E_{s_{t+1}, a_{t+1} \dots} \gamma^t \nabla_{\theta} \log \pi_\theta (a_t \mid s_t) R_t = \\
&= \{ \text{выносим константу из мат.ожидания} \} = \\
&= \sum_{t \ge 0} \E_{a_0, s_1 \dots s_t, a_t} \gamma^t \nabla_{\theta} \log \pi_\theta (a_t \mid s_t)  \E_{s_{t+1}, a_{t+1} \dots} R_t = \\
&= \{ \text{видим определение Q-функции} \} = \\
&= \sum_{t \ge 0} \E_{a_0, s_1 \dots s_t, a_t} \gamma^t \nabla_{\theta} \log \pi_\theta (a_t \mid s_t) Q^\pi(s_t, a_t) = \\
&= \{ \text{формально, берём фиктивное мат.ожидание по $s_{t+1}, a_{t+1}, \cdots$} \} = \\
&= \sum_{t \ge 0} \E_{\Traj \sim \pi \mid s_0 = s} \gamma^t \nabla_{\theta} \log \pi_\theta (a_t \mid s_t) Q^\pi(s_t, a_t) = \\
&= \{ \text{снова меняем местами сумму и мат.ожидание} \} = \\
&= \E_{\Traj \sim \pi \mid s_0 = s} \sum_{t \ge 0} \gamma^t \nabla_{\theta} \log \pi_\theta (a_t \mid s_t) Q^\pi(s_t, a_t)  \tagqed
\end{align*}
\end{proposition}

Итак, мы получили формулу \eqref{pgt_firstproof} вторым способом.

\subsection{Физический смысл}

Обсудим, а в каком направлении, собственно, указывает полученная формула градиента \eqref{pgt_firstproof}. Оказывается, градиент нашего функционала имеет вид градиента взвешенных логарифмов правдоподобий. Чтобы ещё лучше увидеть это, рассмотрим \emph{суррогатную функцию} (surrogate objective) --- другой функционал, который будет иметь в точке текущих значений параметров стратегии $\pi$ такой же градиент, как и $J(\theta)$:
\begin{equation}\label{surrogateobj}
\mathcal{L}_{\textcolor{ChadPurple}{\tilde{\pi}}}(\textcolor{ChadBlue}{\theta}) \coloneqq \E_{\Traj \sim \textcolor{ChadPurple}{\tilde{\pi}(s)}} \sum_{t \ge 0} \gamma^t \log \textcolor{ChadBlue}{\pi_{\theta}}(a \mid s) Q^{\textcolor{ChadPurple}{\tilde{\pi}}}(s, a)
\end{equation}

Это суррогатная функция от двух стратегий: стратегии $\textcolor{ChadBlue}{\pi_{\theta}}$, которую мы оптимизируем, и ещё одной стратегии $\textcolor{ChadPurple}{\tilde{\pi}}$. Давайте рассмотрим эту суррогатную функцию в точке $\theta$ такой, что эти две стратегии совпадают: $\textcolor{ChadPurple}{\tilde{\pi}} \HM\equiv \textcolor{ChadBlue}{\pi_{\theta}}$, и посмотрим на градиент при изменении $\theta$, только одной из них. То есть что мы сказали: давайте <<заморозим>> оценочную Q-функцию, и <<заморозим>> распределение, из которого приходят пары $s, a$. Тогда:

\begin{proposition}
$$\left. \textcolor{ChadBlue}{\nabla_{\theta}} \mathcal{L}_{\textcolor{ChadPurple}{\tilde{\pi}}}(\textcolor{ChadBlue}{\theta}) \right|_{\textcolor{ChadPurple}{\tilde{\pi}} = \textcolor{ChadBlue}{\pi_{\theta}}} = \nabla_{\theta} J(\theta)$$
\begin{proof}
Поскольку мат.ожидание по траекториям не зависит в этой суррогатной функции от $\theta$, то градиент просто можно пронести внутрь:
$$\left. \textcolor{ChadBlue}{\nabla_{\theta}} \mathcal{L}_{\textcolor{ChadPurple}{\tilde{\pi}}}(\textcolor{ChadBlue}{\theta}) \right|_{\textcolor{ChadPurple}{\tilde{\pi}} = \textcolor{ChadBlue}{\pi_{\theta}}} = \E_{\Traj \sim \textcolor{ChadPurple}{\tilde{\pi}(s)}} \sum_{t \ge 0} \gamma^t \left. \textcolor{ChadBlue}{\nabla_{\theta}} \log \textcolor{ChadBlue}{\pi_{\theta}}(a \mid s) \right|_{\textcolor{ChadPurple}{\tilde{\pi}} = \textcolor{ChadBlue}{\pi_{\theta}}} Q^{\textcolor{ChadPurple}{\tilde{\pi}}}(s, a)$$

В точке $\theta \colon \textcolor{ChadPurple}{\tilde{\pi}} \HM= \textcolor{ChadBlue}{\pi_{\theta}}$ верно, что $p(\Traj \HM\mid \textcolor{ChadPurple}{\tilde{\pi}}) \HM\equiv p(\Traj \HM\mid \textcolor{ChadBlue}{\pi})$ и $Q^{\textcolor{ChadPurple}{\tilde{\pi}}}(s, a) \HM = Q^{\textcolor{ChadBlue}{\pi}}(s, a)$; следовательно, значение градиента в этой точке совпадает со значением формулы \eqref{pgt_firstproof}.
\end{proof}
\end{proposition}

Значит, направление максимизации $J(\theta)$ в текущей точке $\theta$ просто совпадает с направлением максимизации этой суррогатной функции! Это принципиально единственное свойство введённой суррогатной функции. Таким образом, можно считать, что в текущей точке мы на самом деле <<как бы>> максимизируем \eqref{surrogateobj}, а это уже в чистом виде логарифм правдоподобия каких-то пар $(s, a)$, для каждой из которых дополнительно выдан <<вес>> в виде значения $Q^{\pi}(s, a)$.

\begin{remark}
Эта суррогатная функция очень удобна для подсчёта градиента $\nabla_\theta J(\pi)$, поскольку она представляет его <<в терминах лосса>>:
$$\nabla_\theta J(\pi) = \nabla_\theta \Loss^{\actor}(\theta)$$
Это полезно в средствах автоматического дифференцирования, где нужно задать некоторый вычислительный граф для получения градиентов; также её можно считать некой <<функцией потерь>> для актёра, хотя это название очень условно хотя бы потому, что значение этой функции вовсе не должно убывать и может вести себя довольно хаотично.
\end{remark}

Проведём такую аналогию с задачей обучения с учителем: если в машинном обучении в задачах регрессии и классификации мы для данной выборки $(x, y)$ максимизировали правдоподобие
$$\sum_{(x, y)} \log p(y \mid x, \theta) \to \max_{\theta},$$
то теперь в RL, когда выборки нет, мы действуем по-другому: мы сэмплируем сами себе входные данные $s$ и примеры выходных данных $a$, выдаём каждой паре какой-то <<кредит доверия>> (credit), некую скалярную оценку хорошести, выраженную в виде $Q^{\pi}(s, a)$, и идём в направлении максимизации
$$\sum_{(s, a)} \log \pi(a \mid s, \theta)Q^{\pi}(s, a) \to \max_{\theta}.$$

\subsection{REINFORCE}

Попробуем сразу построить какой-нибудь практический RL алгоритм при помощи формулы \eqref{pgt_firstproof}. Нам достаточно лишь несмещённой оценки на градиент, чтобы воспользоваться методами стохастической градиентной оптимизации, и поэтому мы просто попробуем всё неизвестное в формуле заменить на Монте-Карло оценки. 

Во-первых, для оценки мат.ожидания по траекториям просто сыграем несколько полных игр при помощи текущей стратегии $\pi$. Сразу заметим, что мы тогда требуем эпизодичности среды и сразу ограничиваем себя on-policy режимом: для каждого следующего шага нам требуется сыграть эпизоды при помощи именно текущей стратегии. Во-вторых, воспользуемся Монте-Карло оценкой для приближения $Q^\pi(s_t, a_t)$, заменив его просто на reward-to-go:
$$Q^\pi(s, a) \approx R(\Traj), \qquad \Traj \sim \pi \mid s_t = s, a_t = a$$

Можно сказать, что мы воспользовались формулой градиента в форме \eqref{reinforce_grad}.

\begin{algorithm}[label = REINFORCE]{REINFORCE}
\textbf{Гиперпараметры:} $N$ --- количество игр, $\pi(a \mid s, \theta)$ --- стратегия с параметрами $\theta$, SGD-оптимизатор.

\vspace{0.3cm}
Инициализировать $\theta$ произвольно \\
\textbf{На очередном шаге $t$:}
\begin{enumerate}
    \item играем $N$ игр $\Traj_1, \Traj_2 \dots \Traj_N \sim \pi$
    \item для каждого $t$ в каждой игре $\Traj$ считаем reward-to-go: $R_t(\Traj ) \coloneqq \sum_{\hat{t} = t} \gamma^{\hat{t} - t} r_{\hat{t}}$
    \item считаем оценку градиента:
    $$\nabla_\theta J(\pi) \coloneqq \frac{1}{N}\sum_{\Traj} \sum_{t \ge 0} \gamma^t \nabla_\theta \log \pi(a_t \mid s_t, \theta) R_t(\Traj ) $$
    \item делаем шаг градиентного подъёма по $\theta$, используя $\nabla_\theta J(\pi)$
\end{enumerate}
\end{algorithm}

Первая беда такого алгоритма очевидна: для одного шага градиентного подъёма нам необходимо играть несколько игр целиком (!) при помощи текущей стратегии. Такой алгоритм просто неэффективен в плане сэмплов, и это негативная сторона on-policy режима. Чтобы как-то снизить этот эффект, хотелось бы научиться как-то делать шаги обучения, не доигрывая эпизоды до конца.

Вторая проблема алгоритма --- колоссальная дисперсия нашей оценки градиента. На одном шаге направление оптимизации указывает в одну сторону, на следующем --- совсем в другую. В силу корректности нашей оценки все гарантии стохастичной оптимизации лежат у нас в кармане, но на практике дождаться каких-то результатов от такого алгоритма в сколько-то сложных задачах не получится.

Чтобы разобраться с этими двумя проблемами, нам понадобится чуть подробнее познакомиться с конструкцией \eqref{pgt_firstproof}.

\subsection{State visitation frequency}

Мы получили, что градиент оптимизируемого функционала по параметрам стратегии \eqref{pgt_firstproof} имеет вид мат.ожиданий по траекториям стратегии $\pi$. Казалось бы, для его оценки нам придётся играть полные эпизоды. Однако видно, что внутри интеграла по траекториям и суммы по времени стоит нечто, зависящее только от пар состояние-действие:
$$f(s, a) \coloneqq \log \pi_\theta (a \mid s) Q^\pi(s, a)$$
Нельзя ли как-то сказать, что если мы делаем стратегией $\pi$ лишь несколько шагов в среде, то собранные пары $s, a$ приходят из того самого распределения, которое мы хотим оценить?

Допустим, мы взаимодействуем со средой при помощи стратегии $\pi$. Из какого распределения нам приходят состояния, которые мы встречаем?
\begin{proposition}
Состояния, которые встречает агент со стратегией $\pi$, приходят из некоторой стационарной марковской цепи.
\begin{proof}
Выпишем вероятность оказаться на очередном шаге в состоянии $s'$, если мы используем стратегию $\pi$:
$$p(s' \mid s) = \int\limits_{\A} \pi(a \mid s)p(s' \mid s, a) \diff a$$
Эта вероятность не зависит от времени и от истории, следовательно, цепочка состояний образует марковскую цепь.
\end{proof}
\end{proposition}

Допустим, начальное состояние $s_0$ фиксировано. Обозначим вероятность оказаться в состоянии $s$ в момент времени $t$ при использовании стратегии $\pi$ как $p(s_t \HM= s \mid \pi)$. Мы могли бы попробовать посчитать, сколько раз мы в среднем оказываемся в некотором состоянии $s$, просто просуммировав по времени:
$$\sum_{t \ge 0} p(s_t \HM= s \mid \pi),$$
однако, как легко видеть, такой ряд может оказаться равен бесконечности (например, если в MDP всего одно состояние). Поскольку мы хотели получить распределение, мы можем попробовать отнормировать этот <<счётчик>>:

\begin{definition} 
Для данного MDP и политики $\pi$ \emph{state visitation frequency} называется
\begin{equation}\label{svf}
\mu_\pi(s) := \lim_{T \to \infty} \frac{1}{T} \sum_{t \ge 0}^T p(s_t = s \mid \pi)
\end{equation}
\end{definition}

Вообще говоря, мы мало что можем сказать про это распределение. При некоторых технических условиях у марковской цепи встречаемых состояний может существовать некоторое стационарное распределение $\lim\limits_{t \to \infty} p(s_t \HM= s \mid \pi)$, из которого будут приходить встречающиеся состояния, условно, через бесконечное количество шагов взаимодействия; если так, то \eqref{svf} совпадает с ним, поскольку, интуитивно, начиная с некоторого достаточно большого момента времени $t$ все слагаемые в ряде будут очень похожи на стационарное распределение.

Можно примерно считать, что во время обучения при взаимодействии со средой состояния приходят из $\mu_{\pi}(s)$, где $\pi$ --- стратегия взаимодействия. Конечно, это верно лишь в предположении, что марковская цепь уже <<разогрелась>> и распределение действительно похоже на стационарное; то есть, в предположении, что обучение продолжается достаточно долго (обычно это так), и предыдущие стратегии, использовавшиеся для взаимодействия, менялись достаточно плавно. 

Естественно, надо помнить, что сэмплы из марковской цепи скоррелированы, и соседние состояния будут очень похожи --- то есть независимости в цепочке встречаемых состояний, конечно, нет. При необходимости набрать мини-батч независимых сэмплов из $\mu_{\pi}(s)$ для декорреляции необходимо воспользоваться параллельными средами: запустить много сред параллельно и для очередного мини-батча собирать состояния из разных симуляций взаимодействия. Вообще, если в батч попадает целая цепочка состояний из одной и той же среды, то это нарушает условие независимости сэмплов и мешает обучению. 

\subsection{Расцепление внешней и внутренней стохастики}

Итак, давайте попробуем формально понять, из какого распределения приходят состояния в формуле градиента \eqref{pgt_firstproof}, и отличается ли оно от $\mu_{\pi}(s)$. Для этого мы сейчас придумаем, как можно записывать функционалы вида
$$\E_{\Traj \sim \pi} \sum_{t \ge 0} \gamma^t f(s_t, a_t),$$
где $f$ --- какая-то функция от пар состояние-действие, немного по-другому.

В MDP есть два вида стохастики:
\begin{itemize}
    \item \emph{внешняя} (extrinsic), связанная со случайностью в самой среде и неподконтрольная агенту; она заложена в функции переходов $p(s' \HM\mid s, a)$.
    \item \emph{внутренняя} (intrinsic), связанная со случайностью в стратегии самого агента; она заложена в $\pi(a \mid s)$. Это стохастика нам подконтрольна при обучении.
\end{itemize}

Мат. ожидание по траектории \eqref{traj_expectation} плохо тем, что мат.ожидания по внешней и внутренней стохастике чередуются. При этом во время обучения из внешней стохастики мы можем только получать сэмплы, поэтому было бы здорово переписать наш функционал как-то так, чтобы он имел вид мат.ожидания по всей внешней стохастике.

Введём ещё один, <<дисконтированный счётчик посещения состояний>> для стратегии взаимодействия $\pi$. При дисконтировании отпадают проблемы с нормировкой.

\begin{definition} 
Для данного MDP и политики $\pi$ \emph{discounted state visitation distribution} называется
\begin{equation}\label{svd}
d_\pi(s) := (1 - \gamma ) \sum_{t \ge 0} \gamma^t p(s_t = s \mid \pi)
\end{equation}
\end{definition}

\begin{proposition} State visitation distribution есть распределение на множестве состояний, то есть:
$$
\int\limits_{\St} d_\pi(s) \diff s = 1
$$
\beginproof
\begin{equation*}
\int\limits_{\St} d_\pi(s) \diff s = \int\limits_{\St} (1 - \gamma )  \sum_{t \ge 0} \gamma^t p(s_t = s) \diff s  = (1 - \gamma ) \sum_{t \ge 0} \gamma^t \int\limits_{\St} p(s_t = s) \diff s = (1 - \gamma ) \sum_{t \ge 0} \gamma^t = 1   \tagqed
\end{equation*}
\end{proposition}

State visitation distribution \eqref{svd} является важным понятием ввиду следующей теоремы, благодаря которой мы можем чисто теоретически (!) расцепить (decouple) внешнюю и внутреннюю стохастику:

\begin{theoremBox}[label=th:decoupling_stoch]{} Для произвольной функции $f(s, a)$:
\begin{equation}\label{decoupling_stoch}
\E_{\Traj \sim \pi} \sum_{t \ge 0} \gamma^t f(s_t, a_t) = \frac{1}{1 - \gamma}\E_{s \sim d_\pi(s)} \E_{a \sim \pi(a \mid s)} f(s, a)
\end{equation}
\beginproof
\begin{align*}
\E_{\Traj \sim \pi} \sum_{t \ge 0} \gamma^t f(s_t, a_t) &= \\
\{ \text{меняем местами сумму и интеграл} \} &= \sum_{t \ge 0} \gamma^t \E_{\Traj \sim \pi} f(s_t, a_t) = \\ 
\{ \text{расписываем мат.ожидание по траектории} \} &= \sum_{t \ge 0} \gamma^t \int\limits_{\St} \int\limits_{\A } p(s_t = s, a_t = a \mid \pi) f(s, a) \diff a \diff s = \\ 
\{ \text{по определению процесса} \} &= \sum_{t \ge 0} \gamma^t \int\limits_{\St} \int\limits_{\A } p(s_t = s \mid \pi) \pi(a \mid s) f(s, a) \diff a \diff s = \\ 
\{ \text{выделяем мат.ожидание по стратегии} \} &= \sum_{t \ge 0} \gamma^t \int\limits_{\St} p(s_t = s \mid \pi) \E_{\pi(a \mid s)} f(s, a) \diff s = \\
\{ \text{заносим сумму по времени обратно} \} &= \int\limits_{\St} \sum_{t \ge 0} \gamma^t p(s_t = s \mid \pi) \E_{\pi(a \mid s)} f(s, a) \diff s = \\
\{ \text{выделяем state visitation distribution \eqref{svd}} \} &= \int\limits_{\St} \frac{d_\pi(s)}{1 - \gamma} \E_{\pi(a \mid s)} f(s, a) \diff s = \\
\{ \text{выделяем мат.ожидание по состояниям} \} &= \frac{1}{1 - \gamma}\E_{s \sim d_\pi(s)} \E_{\pi(a \mid s)} f(s, a) \tagqed
\end{align*}
\end{theoremBox}

Итак, мы научились переписывать мат.ожидание по траекториям в другом виде. В будущем мы будем постоянно пользоваться формулой \eqref{decoupling_stoch} для разных $f(s, a)$, поэтому полезно запомнить эту альтернативную форму записи. Например, мы можем получить применить этот результат к нашему функционалу:

\begin{proposition}
Оптимизируемый функционал \eqref{goal} можно записать в таком виде:
$$J(\pi) = \frac{1}{1 - \gamma}\E_{s \sim d_\pi(s)} \E_{\pi(a \mid s)} r(s, a)$$
\end{proposition}

Интерпретация у полученного результата может быть такая: нам не столько существенна последовательность принимаемых решений, сколько частоты посещений <<хороших>> состояний с высокой наградой. 

Теорема \ref{th:decoupling_stoch} позволяет переписать и формулу градиента:

\begin{proposition} Выражение для градиента оптимизируемого функционала можно записать следующим образом:
\begin{equation}\label{gradient}
\nabla_{\theta} J(\pi) = \frac{1}{1 - \gamma}\E_{d_\pi(s)} \E_{\pi(a \mid s)} \nabla_{\theta} \log \pi_\theta (a \mid s) Q^{\pi}(s, a)
\end{equation}
\begin{proof}
Применить теорему \ref{th:decoupling_stoch} для $f(s, a) \coloneqq \nabla_{\theta} \log \pi_\theta (a \mid s) Q^{\pi}(s, a)$.
\end{proof}
\end{proposition}

Итак, множитель $\gamma^t$ в формуле \eqref{pgt_firstproof} имеет смысл <<дисконтирования частот посещения состояний>>. Для нас это представляет собой, мягко говоря, очень странную проблему. Если мы рассмотрим формулу градиента $J(\theta)$, то есть зафиксируем начальное состояние в наших траекториях, то все слагаемые, соответствующие состояниям, встречающиеся в эпизодах только после, условно, 100-го шага, будут домножены на $\gamma^{100}$. То есть одни слагаемые в нашем оптимизируемом функционале имеют один масштаб, а другие --- домножаются на близкий к нулю $\gamma^{100}$, совершенно иной. Градиентная оптимизация с такими функционалами просто не справится: в градиентах будет доминировать информация об оптимизации наших решений в состояниях около начального, имеющих большой вес.

Откуда это лезет? Давайте посмотрим на то, что мы оптимизируем: $J(\theta)$ \eqref{goal}. Ну действительно: награда, которую мы получим после сотого шага, дисконтируется на $\gamma^{100}$. Мы уже поняли, что для того, чтобы промаксимизировать её, нам всё равно придётся промаксимизировать $V^{\pi}(s)$ для всех состояний, то есть эти слагаемые <<в микро-масштабе>> для достижения глобального оптимума тоже должны оптимизироваться, и задача оптимизации поставлена <<корректно>>. Но для градиентной оптимизации такая <<форма>> просто не подходит.

Сейчас случится неожиданный поворот событий. На практике во всех Policy Gradient методах от дисконтирования частот посещения отказываются. Это означает, что мы заменяем $d_{\pi}(s)$ на $\mu_{\pi}(s)$ \eqref{svf}:

\begin{equation*}
\nabla_{\theta} J(\pi) \approx \frac{1}{1 - \gamma}\E_{\mu_\pi(s)} \E_{\pi(a \mid s)} \nabla_{\theta} \log \pi_\theta (a \mid s) Q^{\pi}(s, a)
\end{equation*}

В формуле \eqref{pgt_firstproof} это эквивалентно удалению множителя $\gamma^t$:

\begin{equation*}
\nabla_{\theta} J(\pi) \approx \E_{\Traj \sim \pi} \sum_{t \ge 0} \nabla_{\theta} \log \pi_\theta (a_t \mid s_t) Q^{\pi}(s_t, a_t)
\end{equation*}

Мы вовсе не отказались от дисконтирования вовсе: оно всё ещё сидит внутри оценочной функции и даёт приоритет ближайшей награде перед получением той же награды в будущем, как мы и задумывали изначально.

Итак, при таком соглашении мы можем получить как бы оценку Монте-Карло на градиент:
$$\nabla_{\theta} J(\pi) \approx \frac{1}{1 - \gamma}\E_{a \sim \pi(a \mid s)} \nabla_{\theta} \log \pi_\theta (a \mid s) Q^{\pi}(s, a), \qquad s \sim \mu_\pi(s)$$

Такую оценку можно считать условно несмещённой (с учётом нашего забивания на множитель $\gamma^t$ и приближённого сэмплирования из $\mu_{\pi}(s)$), если у нас на руках есть точная (или хотя бы несмещённое) оценка <<кредита>> $Q^\pi(s, a)$.

\subsection{Связь с policy improvement}\label{subsec:pg_is_pi}

Формула градиента в форме \eqref{gradient} даёт ещё одну интересную интерпретацию того, в каком направлении указывает полученная формула градиента. Давайте введём ещё одну \emph{суррогатную функцию} (surrogate objective). Как и суррогатная функция \eqref{surrogateobj}, это будет ещё один функционал, который имеет в точке текущих значений параметров стратегии $\pi$ такой же градиент, как и $J(\theta)$:
\begin{equation}\label{PGisPI}
\mathcal{L}_{\textcolor{ChadPurple}{\tilde{\pi}}}(\textcolor{ChadBlue}{\theta}) \coloneqq \frac{1}{1 - \gamma}\E_{\textcolor{ChadPurple}{d_{\tilde{\pi}}(s)}} \E_{a \sim \textcolor{ChadBlue}{\pi_{\theta}}(a \mid s)} Q^{\textcolor{ChadPurple}{\tilde{\pi}}}(s, a)
\end{equation}

Эта суррогатная функция, опять же, от двух стратегий: стратегии $\textcolor{ChadBlue}{\pi_{\theta}}$, которую мы оптимизируем, и ещё одной стратегии $\textcolor{ChadPurple}{\tilde{\pi}}$. Снова смотрим на эту суррогатную функцию в такой точке $\theta$, что две стратегии совпадают: $\textcolor{ChadPurple}{\tilde{\pi}} \HM\equiv \textcolor{ChadBlue}{\pi_{\theta}}$; будем <<шевелить>> $\theta$ и смотреть на градиент. То есть теперь мы <<заморозили>> распределение частот посещений состояний и, как и в прошлый раз, оценочную функцию. Тогда:

\begin{proposition}
$$\left. \textcolor{ChadBlue}{\nabla_{\theta}} \mathcal{L}_{\textcolor{ChadPurple}{\tilde{\pi}}}(\textcolor{ChadBlue}{\theta}) \right|_{\textcolor{ChadPurple}{\tilde{\pi}} = \textcolor{ChadBlue}{\pi_{\theta}}} = \nabla_{\theta} J(\theta)$$
\begin{proof}
Поскольку $\textcolor{ChadPurple}{d_{\tilde{\pi}}(s)}$ и $Q^{\textcolor{ChadPurple}{\tilde{\pi}}}(s, a)$ не зависят от $\theta$, то для дифференцирования суррогатной функции достаточно лишь пронести градиент через мат.ожидание по выбору стратегии при помощи REINFORCE:
\begin{align*}
\textcolor{ChadBlue}{\nabla_{\theta}} \frac{1}{1 - \gamma}\E_{\textcolor{ChadPurple}{d_{\tilde{\pi}}(s)}} \E_{a \sim \textcolor{ChadBlue}{\pi_{\theta}}(a \mid s)} Q^{\textcolor{ChadPurple}{\tilde{\pi}}}(s, a) &= 
\frac{1}{1 - \gamma}\E_{\textcolor{ChadPurple}{d_{\tilde{\pi}}(s)}} \int\limits_{\A} \textcolor{ChadBlue}{\nabla_{\theta}} \textcolor{ChadBlue}{\pi_{\theta}}(a \mid s) Q^{\textcolor{ChadPurple}{\tilde{\pi}}}(s, a) \diff a = \\
= \{ \text{log-derivative trick \eqref{logderivtrick}} \}
&= \frac{1}{1 - \gamma}\E_{\textcolor{ChadPurple}{d_{\tilde{\pi}}(s)}} \E_{a \sim \textcolor{ChadBlue}{\pi_{\theta}}(a \mid s)} \textcolor{ChadBlue}{\nabla_{\theta}} \log \textcolor{ChadBlue}{\pi_{\theta}}(a \mid s) Q^{\textcolor{ChadPurple}{\tilde{\pi}}}(s, a)
\end{align*}

В точке $\theta \colon \textcolor{ChadPurple}{\tilde{\pi}} \HM= \textcolor{ChadBlue}{\pi_{\theta}}$ верно, что $\textcolor{ChadPurple}{d_{\tilde{\pi}}}(s) \HM= \textcolor{ChadBlue}{d_{\pi}}(s)$ и $Q^{\textcolor{ChadPurple}{\tilde{\pi}}}(s, a) \HM = Q^{\textcolor{ChadBlue}{\pi}}(s, a)$; следовательно, значение градиента в этой точке совпадает со значением формулы \eqref{gradient}.
\end{proof}
\end{proposition}

\begin{wrapfigure}{r}{0.35\textwidth}
\centering
\includegraphics[width=0.35\textwidth]{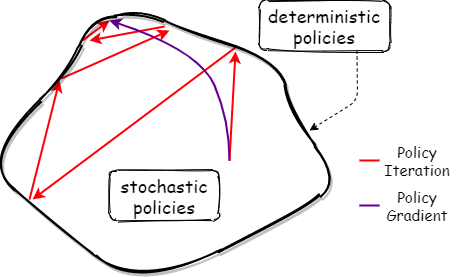}
\vspace{-0.5cm}
\end{wrapfigure}

Это значит, что в текущей точке градиенты указывают туда же, куда и при максимизации $\mathcal{L}_{\textcolor{ChadPurple}{\tilde{\pi}}}(\textcolor{ChadBlue}{\theta})$ по $\textcolor{ChadBlue}{\theta}$. А куда указывает направление градиентов для неё? Выражение $\E_{a \sim \textcolor{ChadBlue}{\pi_{\theta}}(a \mid s)} Q^{\textcolor{ChadPurple}{\tilde{\pi}}}(s, a)$ говорит максимизировать Q-функцию по выбираемым нами действиям: это в чистом виде направление policy improvement-а из теоремы \ref{th:policyimprovement}! 

Но если в табличном сеттинге policy improvement мы делали, так сказать, <<жёстко>>, заменяя целиком стратегию на жадную по действиям, то формула градиента теперь говорит нам, что это лишь направление максимального увеличения $J(\theta)$; после любого шага в этом направлении наша Q-функция тут же, формально, меняется, и в новой точке направление уже должно быть скорректировано. 

Второе уточнение, которое дарит нам эта формула, это распределение, из которого должны приходить состояния. В табличном случае мы не знали, в каких состояниях проводить improvement <<важнее>>, теперь же мы видим, что $s$ должно приходить из $d_{\pi}(s)$. Если какое-то состояние посещается текущей стратегией часто, то улучшать стратегию в нём важнее, чем в состояниях, которые мы посещаем редко.

Но ещё это наблюдение, что градиент будущей награды и policy improvement связаны, в частности, даёт одно из оправданий тому, что мы отказались от дисконтирования частот посещения состояний. Мы используем формулу градиента для максимизации $V^\pi(s_0)$. Но, в общем-то, мы хотим максимизировать $V^\pi(s)$ сразу для всех $s$, и мы можем делать это с некоторыми весами (и эти веса, в целом, наш произвол). Тогда, выходит, частоты появления $s$ в оптимизируемом функционале определяются в том числе этими весами. Другими словами, $d_{\pi}(s)$ указывает на <<самое правильное>> распределение, из которого должны приходить состояния, которые дадут направление именно максимального увеличения функционала, но теория policy improvement-а подсказывает нам, что теоретически корректно выбрать и любое другое. Даже если мы совсем другое какое-то распределение выберем для сэмплирования состояний $s$, то функционал
$$\E_{s} \E_{a \sim \textcolor{ChadBlue}{\pi_{\theta}}(a \mid s)} Q^{\textcolor{ChadPurple}{\tilde{\pi}}}(s, a) \to \max_{\textcolor{ChadBlue}{\theta}}$$
будет давать какое-то направление подъёма, какое-то годное направление оптимизации. 

Получается, исходя из этого рассуждения, что нам для использования формулы градиента даже не нужны on-policy сэмплы, и мы можем брать состояния просто из буфера! Можем ли мы так делать? Допустим, мы готовы заменить распределение состояний на произвольное. Всё равно остаётся существенным сэмплировать действия $a$ именно из текущей версии $\pi(a \HM\mid s)$; значит, брать действие $a$ из буфера нельзя. Следовательно, мы не можем получить $s' \HM\sim p(s' \HM\mid s, a)$; но если у нас как-то есть на руках для любой пары $s, a$ какое-то значение <<кредита>> $Q^\pi(s, a)$, то для градиента актёра нам, согласно формуле, больше ничего и не нужно.

Итак, если у нас есть модель Q-функции, мы даже с оговорками можем обучать стратегию с буфера, если будем брать из него только состояния, а мат.ожидание по $a$ (или его Монте-Карло оценку) считать, используя текущую стратегию $\pi$. Однако, коли уж мы хотим off-policy алгоритм, то и эту Q-функцию тогда нужно учить с буфера. В итоге, мы получим алгоритм, очень похожий на DQN, со схожими недостатками и преимуществами. Мы обсудим такой подход позже в главе \ref{continuouscontrolchapter}, а пока что мы не будем гоняться за возможностью обучаться с буфера и за счёт этого сможем получить некоторые другие полезные преимущества, которые открываются в on-policy режиме.

В частности, раз уж формула градиента подсказывает сэмплировать $s$ из частот посещения состояний, то мы будем стараться так делать, стараться моделировать именно оценку градиента: на практике мы возьмём состояния из недисконтированных частот $\mu_\pi(s)$, но тем не менее.

\subsection{Бэйзлайн}

При стохастической оптимизации ключевым фактором является дисперсия оценки градиента. Когда мы заменяем мат.ожидания на Монте-Карло оценки, дисперсия увеличивается. Понятно, что замена Q-функции --- выинтегрированных будущих наград --- на её Монте-Карло оценку в REINFORCE \eqref{reinforce_grad} повышало дисперсию. Однако, в текущем виде основной источник дисперсии заключается в другом.

\begin{example}
Допустим, у нас два действия $\A = \{0, 1\}$. Мы породили траекторию, в котором выбрали $a \HM= 0$. Выданное значение <<кредита доверия>> --- пусть это точное значение $Q^{\pi}(s, a \HM= 0)$ --- допустим, равно 100. Это значит, что градиент для данной пары $s, a$ указывает на направление изменение параметров, которые увеличат вероятность $\pi(a \HM= 0 \HM\mid s)$, и это направление войдёт в итоговую оценку градиентов с весом 100. Но 100 --- это много или мало?

Рассмотрим один вес $\theta_i \in \R$ в нашей параметризации стратегии. Без ограничения общности будем считать, что для повышения $\pi(a \HM= 0 \HM\mid s)$ вес $\theta_i$ нужно повышать. Тогда если на следующем шаге оптимизации мы в том же $s$ засэмплировали $a \HM= 1$, то для повышения $\pi(a \HM= 1 \HM\mid s)$ вес $\theta_i$, очевидно, нужно уменьшать (так как сумма $\pi(a \HM= 0 \HM\mid s) \HM+ \pi(a \HM= 1 \HM\mid s)$ обязана равняться 1). С каким весом мы будем уменьшать $\theta_i$? С весом $Q^{\pi}(s, a \HM= 1)$. Что, если оно равно 1000? На прошлом шаге мы шли в одну сторону с весом 100, на текущем --- в другую с весом 1000. За счёт разницы весов мы в среднем движемся в правильную сторону, но нас постоянно дёргает то в одном направлении, то в другом.
\end{example}

\needspace{8\baselineskip}
\begin{wrapfigure}{r}{0.3\textwidth}
\vspace{-0.2cm}
\centering
\includegraphics[width=0.3\textwidth]{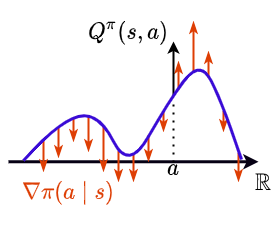}
\vspace{-1cm}
\end{wrapfigure}

Обобщим описанную в примере ситуацию. Для этого вспомним утверждение \eqref{baseline}: градиент логарифма правдоподобия в среднем равен нулю. Это значит, что если для данного $s$ мы выдаём некоторое распределение $\pi(a \HM\mid s)$, для увеличения вероятностей в одной области $\A$ нужно данный вес $\theta_i$ параметризации увеличивать, а в другой области --- уменьшать. В среднем <<магнитуда изменения>> равна нулю. Но у нас в Монте-Карло оценке только один сэмпл $a \sim \pi(a \HM\mid s)$, и для него направление изменения домножится на кредит, на нашу оценку $Q^{\pi}(s, a)$. Если эта оценка в одной области 100, а в другой 1000 --- дисперсия получаемых значений $\nabla_\theta \log \pi_\theta(a \HM\mid s) Q^{\pi}(s, a)$ становится колоссальной. Было бы сильно лучше, если бы <<кредит>> --- вес примеров --- был в среднем центрирован, и тоже колебался возле нуля. Тогда для <<плохих действий>> мы правдоподобие этих действий уменьшаем, а для <<хороших действий>> --- увеличиваем, что даже чисто интуитивно логичнее. И для центрирования весов правдоподобия в Policy Gradient методах всегда вводится \emph{бэйзлайн} (baseline), без которого алгоритмы обычно не заведутся.

\begin{proposition}
Для произвольной функции $b(s) \colon \St \to \R$, называемой бэйзлайном, верно:
\begin{equation}\label{pgt_baseline}
\nabla_{\theta} J(\pi) = \frac{1}{1 - \gamma}\E_{d_\pi(s)} \E_{\pi(a \mid s)} \nabla_{\theta} \log \pi_\theta (a \mid s) \left(Q^{\pi}(s, a) - b(s) \right)
\end{equation}
\begin{proof}
Добавленное слагаемое есть ноль в силу формулы \eqref{baseline}. 
\end{proof}
\end{proposition}

Это верно для произвольной функции от состояний и становится неверно, если вдруг бэйзлайн начинает зависеть от $a$. Мы вольны выбрать бэйзлайн произвольно; он не меняет среднего значения оценок градиента, но изменяет дисперсию. 

\begin{theorem}
Бэйзлайном, максимально снижающим дисперсию Монте-Карло оценок формулы градиентов \eqref{pgt_baseline}, является
$$b^*(s) \coloneqq \frac{\E_a \|\nabla_{\theta} \log \pi(a \mid s)\|_2^2 Q^{\pi}(s, a)}{\E_a \|\nabla_{\theta} \log \pi(a \mid s)\|_2^2}$$
\beginproof
Рассмотрим одно состояние $s$ и попробуем вычислить оптимальное значение $b(s)$. Пусть для данного состояния $m$ --- среднее значение оценки градиента, которое, как мы поняли ранее, не зависит от значения $b$:
$$
m \coloneqq \E_{a \sim \pi(a \mid s)} \nabla_{\theta} \log \pi_\theta (a \mid s) Q^{\pi}(s, a)
$$
Для нас будет важно, что $m$ --- просто какой-то фиксированный вектор той же размерности, что и $\theta$. Распишем дисперсию и будем минимизировать её по $b$:
$$\E_a \| \nabla_{\theta} \log \pi_\theta(a \mid s) \left( Q^{\pi}(s, a) - b \right) - m \|^2_2 \to \min_{b} $$
Дифференцируем по $b$ и приравниваем к нулю:
$$2 \E_a \left( \nabla_{\theta} \log \pi_\theta(a \mid s) \left( Q^{\pi}(s, a) - b \right) - m \right)^T \left(-\nabla_{\theta} \log \pi_\theta(a \mid s) \right) = 0 $$
Выделяем норму градиента логарифма правдоподобия:
\begin{equation}\label{optimalbaselineintermediate}
-\E_a \|\nabla_{\theta} \log \pi_\theta(a \mid s)\|^2_2 Q^{\pi}(s, a) + \E_a \|\nabla_{\theta} \log \pi_\theta(a \mid s)\|^2_2 b + \E_a m^T \left(\nabla_{\theta} \log \pi_\theta(a \mid s) \right) = 0
\end{equation}

Осталось заметить, что третье слагаемое есть ноль. Это обобщение нашей теоремы о бэйзлайне (формулы \eqref{baseline}): условно, бэйзлайн может быть свой для каждой компоненты вектора $\theta$, опять же, до тех пор, пока он не зависит от действий. В данном случае $m$ --- некоторый фиксированный вектор, одинаковый для всех $a$; поэтому, если $d$ --- размерность вектора параметров $\theta$, то:
$$\E_a m^T \left(\nabla_{\theta} \log \pi_\theta(a \mid s) \right) = \E_a \sum_{i=0}^{d} m_i \nabla_{\theta_i} \log \pi_\theta(a \mid s) = \sum_{i=0}^{d} m_i \underbrace{\E_a \nabla_{\theta_i} \log \pi_\theta(a \mid s)}_{\text{0 по формуле \eqref{baseline}}} = 0$$

Убирая это нулевое третье слагаемое из \eqref{optimalbaselineintermediate}, получаем равенство между первыми двумя:
\begin{equation*}
b\E_a \|\nabla_{\theta} \log \pi_\theta(a \mid s)\|^2_2 = \E_a \|\nabla_{\theta} \log \pi_\theta(a \mid s)\|^2_2 Q^{\pi}(s, a)
\end{equation*}
Выражая из него $b$, получаем доказываемое.
\end{theorem}

\begin{example}
Представить себе интуицию этой формулы можно следующим образом. Рассмотрим одно $s$. В нём мы сэмплируем одно какое-то действие $a$ с вероятностями $\pi(a \HM \mid s)$. Допустим, действий четыре: засэмплировалось какое-то одно. Для каждого действия есть свой градиент, показывающий, как повысить вероятность выбора этого действия $\nabla_\theta \log \pi_\theta (a \HM \mid s)$. У него есть некоторая норма $\|\nabla_{\theta} \log \pi_\theta(a \mid s)\|^2_2$ --- его длина. Мы в качестве градиента берём такой вектор, отмасштабированный на $Q^\pi(s, a)$ --- произвольное, в общем-то, число. Как снизить дисперсию получающихся градиентов?

\needspace{12\baselineskip}
\begin{wrapfigure}{r}{0.5\textwidth}
\centering
\vspace{-0.8cm}
\includegraphics[width=0.5\textwidth]{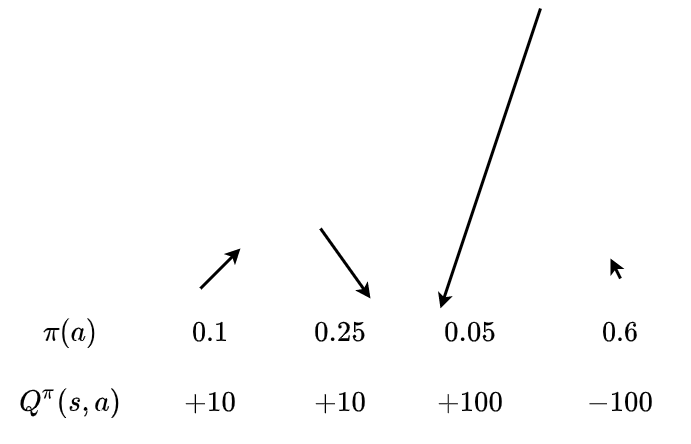}
\end{wrapfigure}

Мы можем вычесть из значения <<кредитов>> $Q^\pi(s, a)$ какое-то общее для всех действий число $b$. И формула говорит: возьмите среднее значение $Q^\pi(s, a)$ с учётом двух вещей: что, во-первых, какие-то вектора мелькают чаще других, и поэтому нужно брать среднее с учётом вероятностей $\pi(a \HM \mid s)$, а во-вторых, какие-то вектора длиннее по норме, чем другие, и поэтому также нужно перевзвесить значения $Q^\pi(s, a)$ пропорционально норме градиента $\|\nabla_{\theta} \log \pi_\theta(a \mid s)\|^2_2$. Второе указывает сделать бэйзлайн таким, чтобы отцентрированный <<кредит>> для самых <<длинных>> векторов был поближе к нулю. 
\end{example}

Практическая ценность результата невысока. Знать норму градиента для всех действий $a$ вычислительно будет труднозатратно даже в дискретных пространствах действий. Поэтому мы воспользуемся небольшим предположением: мы предположим, что норма градиента примерно равна для всех действий. Тогда:
$$b^*(s) = \frac{\E_a \|\nabla_{\theta} \log \pi_\theta(a \mid s)\|_2^2 Q^{\pi}(s, a)}{\E_a \|\nabla_{\theta} \log \pi_\theta(a \mid s)\|_2^2} = \{ \|\nabla_{\theta} \log \pi_\theta(a \mid s)\|_2^2 \approx \const(a) \} \approx \E_a Q^{\pi}(s, a) = V^\pi(s)$$

Эта аппроксимация довольно интуитивная: по логике, для дисперсии хорошо, если значения функции $Q^\pi(s, a) \HM- b(s)$ вертятся вокруг нуля, то есть в среднем дают ноль, и поэтому хорошим (но неоптимальным) бэйзлайном будет
$$\E_{a \sim \pi(a \mid s)} \left( Q^{\pi}(s, a) - b(s) \right) \coloneqq 0 \quad \Rightarrow \quad b(s) \coloneqq \E_{a \sim \pi(a \mid s)} Q^{\pi}(s, a) = V^\pi(s)$$

Итак, всюду далее будем в качестве бэйзлайна использовать $b(s) \HM \coloneqq V^\pi(s)$. Подставляя и вспоминая определение \eqref{advantage} Advantage-функции, получаем:
\begin{proposition}
\begin{equation}\label{advantagepg}
\nabla_{\theta} J(\pi) = \frac{1}{1 - \gamma}\E_{d_\pi(s)} \E_{\pi(a \mid s)} \nabla_{\theta} \log \pi_\theta (a \mid s) A^{\pi}(s, a)
\end{equation}
\end{proposition}

Таким образом, <<кредит>>, который мы выдаём каждой паре $s, a$, будет являться оценкой Advantage, и состоять из двух слагаемых: оценки Q-функции и бэйзлайна. Именно поэтому этот вес и называется <<кредитом>>: задача оценки Advantage и есть в точности тот самый credit assingment, который мы обсуждали в главе \ref{sec:biasvar}!

\section{Схемы <<Актёр-критик>>}\label{ActorCriticSection}

\subsection{Введение критика}

Мы хотели научиться оптимизировать параметры стратегии при помощи формулы градиента, не доигрывая эпизоды до конца. Мы уже поняли, что мат.ожидание по траекториям не представляет для нас проблемы. Тогда осталось лишь придумать, как, не доигрывая эпизоды до конца, проводить credit assignment, то есть определять для каждой пары $s, a$ оценку Advantage-функции.

Раз знание функции $Q^\pi$ позволит обучаться, не доигрывая эпизоды до конца, а функцию в готовом виде нам никто не даст, то возникает логичная идея --- аппроксимировать её. Итак, введём вторую сетку, которая будет <<оценивать>> наши собственные решения --- \emph{критика} (critic). Нейросеть, моделирующую стратегию, соответственно будем называть \emph{актёром} (actor), и такие алгоритмы, в которых обучается как модель критика, так и модель актёра, называются \emph{Actor-Critic}.

Здесь возникает принципиальный момент: мы не умеем обучать модели оценочных функций $Q^\pi$ или $V^\pi$, так чтобы они оценивали будущую награду несмещённо (<<выдавали в среднем правильный ответ>>). Что это влечёт? При смещённой оценке Q-функции любые гарантии на несмещённость градиентов мгновенно теряются. Единственный способ оценивать Q-функцию несмещённо --- Монте-Карло, но она требует полных эпизодов и имеет более высокую дисперсию. Любое замешивание нейросетевой аппроксимации в оценку --- и мы сразу теряем несмещённость оценок на градиент, и вместе с ними --- любые гарантии на сходимость стохастической оптимизации к локальному оптимуму или даже вообще хоть куда-нибудь!

В машинном обучении в любых задачах оптимизации всегда необходимо оценивать градиент оптимизируемого функционала несмещённо, и важной необычной особенностью Policy Gradient методов в RL является тот факт, что в них внезапно используются именно смещённые оценки градиента. Надежда на работоспособность алгоритма после замены значения $Q^\pi(s, a)$ на смещённую оценку связана с тем, что формула градиента говорит проводить policy improvement во встречаемых состояниях (см. физический смысл суррогатной функции \eqref{PGisPI}). Коли градиент есть просто policy improvement, то мы знаем из общей идеи алгоритма \ref{generalizedpolicyiteration} Generalized Policy Iteration, что для улучшения политики вовсе не обязательно использовать идеальную $Q^\pi$, достаточно любой аппроксимации критика $Q(s, a) \HM \approx Q^\pi(s, a)$, которая параллельным процессом движется в сторону идеальной $Q^\pi$ для текущей стратегии $\pi$. Фактически, все Actor-Critic методы моделируют именно эту идею.

Следующий существенный момент заключается в том, что в качестве критика обычно учат именно V-функцию. Во-первых, если бы мы учили модель Q-функции и подставили бы её, то градиент
$$\nabla_\theta \log \pi_\theta(a \mid s) Q(s, a)$$
направил бы нашу стратегию в $\argmax\limits_a Q(s, a)$. Помимо того, что это выродило бы стратегию, это бы привело к тому, что качество обучения актёра выродилось бы в качество обучения критика: пока модель $Q(s, a)$ не похожа на истинную $Q^\pi(s, a)$, у нас нет надежды, что актёр выучит хорошую стратегию. 


\begin{remark}
Можем ли мы в качестве критика напрямую учить сетку, выдающую аппроксимацию $A_{\phi}(s, a) \HM\approx A^\pi(s, a)$, и использовать её выход в качестве нашей оценки $\Psi (s, a)$? Это не очень удобно хотя бы потому, что в отличие от Q-функций и V-функций, advantage --- не абстрактная произвольная функция: она должна подчиняться теореме \ref{pr:advantageiszero}. При этом восстановить по advantage-у Q-функцию без знания V-функции нельзя, а значит, для advantage не получится записать аналога уравнения Беллмана, использующего только advantage-функции.
\end{remark}

Возможность не обучать сложную $Q^*$ является одним из преимуществ подхода прямой оптимизации $J(\theta)$. В DQN было обязательным учить именно Q-функцию, так как, во-первых, мы хотели выводить из неё оптимальную стратегию, во-вторых, для уравнения оптимальности Беллмана для оптимальной V-функции фокус с регрессией не прокатил бы --- там мат.ожидание стоит внутри оператора взятия максимума. Сейчас же мы из соображений эффективности алгоритма (желания не играть полные эпизоды и задачи снижения дисперсии оценок) не сможем обойтись совсем без обучения каких-либо оценочных функций, но нам хватит лишь $V_{\phi}(s) \HM\approx V^{\pi}$, поскольку оценить Q-функцию мы можем хотя бы так:
\begin{equation}\label{Qonlineonestep}
Q^\pi(s, a) = r(s, a) + \gamma \E_{s'} V^\pi(s') \approx r(s, a) + \gamma V_{\phi}(s'), \qquad s' \sim p(s' \mid s, a)
\end{equation}
Иначе говоря, если у нас есть приближение V-функции, то мы можем использовать её как для бэйзлайна, так и для оценки Q-функции. Важно, что V-функция намного проще, чем Q-функция: ей не нужно дифференцировать между действиями, достаточно лишь понимать, какие области пространства состояний --- хорошие, а какие плохие. И поэтому раз можно обойтись ей, то почему бы так не сделать и не упростить критику задачу.

\subsection{Bias-variance trade-off}

Обсудим сначала, как критик $V_{\phi}(s) \HM\approx V^\pi(s)$ с параметрами $\phi$ будет использоваться в формуле градиента по параметрам стратегии \eqref{advantagepg}. Мы собираемся вместо честного advantage подставить некоторую его оценку (advantage estimator) и провести таким образом credit assingment:
\begin{equation}\label{advantageestimation}
\nabla_{\theta} J(\pi) \approx \frac{1}{1 - \gamma}\E_{d_\pi(s)} \E_{a \sim \pi(a \mid s)} \nabla_{\theta} \log \pi_\theta (a \mid s) \underbrace{\Psi (s, a)}_{\approx A^{\pi}(s, a)}
\end{equation}

Мы столкнулись ровно с тем же самым bias-variance trade-off, который мы обсуждали \ref{sec:biasvar}, который как раз и сводится к оцениванию Advantage и определению того, какие действия хорошие или плохие. Только теперь, в контексте policy gradient, речь напрямую идёт о дисперсии и смещении оценок градиента. Мы уже встречались с двумя самыми <<крайними>> вариантами выбора функции $\Psi (s, a)$: Монте-Карло и одношаговая оценка через V-функцию \eqref{Qonlineonestep}:

\vspace{0.2cm}
\begin{center}
\begin{tabular}{ccc}
\toprule
    \textbf{$\Psi (s, a)$} & \textbf{Дисперсия} & \textbf{Смещение}  \\
\midrule
    $R_t - V_{\phi}(s)$ & высокая & нету \\
    \hdashline
    $r(s, a) + \gamma V_{\phi}(s') - V_{\phi}(s)$ & низкая & большое \\
\bottomrule
\end{tabular}
\end{center}
\vspace{0.2cm}

В этой таблице речь идёт именно о дисперсии и смещении оценки градиента $J(\pi)$ при использованной аппроксимации! А то есть, в первом случае оценка Q-функции несмещённая, оценка V-функции смещённая, но поскольку в качестве бэйзлайна в силу \eqref{baseline} может использоваться совершенно произвольная функция от состояний, совершенно несущественно, насколько наша аппроксимация $V^\pi(s)$ вообще похожа на истинную оценочную функцию. Да, если наша аппроксимация V-функции будет неточной, мы, вероятно, не так сильно собьём дисперсию оценок градиента, как могли бы, но ровно на этом недостатки использования смещённой аппроксимации V-функции в качестве бэйзлайна заканчиваются. 

Во втором же случае, аппроксимация $V^\pi(s')$ используется для оценки Q-функции и вызывает смещение в том числе в оценке градиента. Дисперсия же снижается, поскольку в Q-функции аккумулированы мат.ожидания по всему хвосту траектории; она в обоих случаях существенно ниже, чем до введения бэйзлайна, но замена Монте-Карло оценки на бутстрапированную оценку снижает её ещё сильнее.

Вспомним, как можно интерполировать между этими крайними вариантами. Для начала, у нас есть ещё целое семейство промежуточных вариантов --- многошаговых (multi-step) оценок Q-функции, использующих N-шаговое уравнение Беллмана \eqref{NstepBellman}:
$$Q^\pi(s, a) \approx \sum_{t=0}^{N-1} \gamma^{t} r^{(t)} + \gamma^N V_{\phi}(s^{(N)}) $$

Тогда мы пользуемся для credit assingment-а $N$-шаговой оценкой Advantage, или $N$-шаговой временной разностью \eqref{Nstepadvantage}.
$$\Psi_{(N)}(s, a) \coloneqq \sum_{t=0}^{N-1} \gamma^{t} r^{(t)} + \gamma^N V_{\phi}(s^{(N)}) - V_{\phi}(s)$$
С ростом $N$ дисперсия такой оценки увеличивается: всё больший фрагмент траектории мы оцениваем по Монте-Карло, нам становятся нужны сэмплы $a_{t+1} \sim \pi(a_{t+1} \mid s_{t=1})$, $s_{t+2} \sim \pi(s_{t+2} \mid s_{t+1}, a_{t+1})$, \dots , $s_{t+N} \sim \pi(s_{t+N} \mid s_{t+N-1}, a_{t+N-1})$. а при $N \to \infty$ оценка в пределе переходит в полную Монте-Карло оценку оставшейся награды, где дисперсия большая, но зато исчезает смещение в силу отсутствия смещённой аппроксимации награды за хвост траектории. 

Также с ростом $N$ мы начинаем всё меньше опираться на модель критика. Используя сэмплы наград, мы ценой увеличения дисперсии напрямую учим связь между хорошими действиями и высоким откликом от среды, меньше опираясь на промежуточный этап в виде выучивания оценочной функции. Это значит, что нам не нужно будет дожидаться, пока наш критик идеально обучиться: в оценку Advantage попадает сигнал в том числе из далёкого будущего при больших $N$, и актёр поймёт, что удачно совершённое действие надо совершать чаще. Математически это можно объяснить тем, что, увеличивая $N$, мы снижаем смещение наших оценок градиента. Значит, нам в целом даже и не потребуется, чтобы критик оценивал состояния с высокой точностью, поскольку главное, чтобы в итоге получалась более-менее адекватная оценка совершённых нашей стратегией действий.

Trade-off заключается в том, что чем дальше в будущее мы заглядываем, тем выше дисперсия этих оценок; помимо этого, для заглядывания в будущее на $N$ шагов нужно же иметь это самое будущее, то есть из каждой параллельно запущенной среды понадобится собрать для очередного мини-батча не по одному сэмплу, а собрать целый длинный фрагмент траектории. Поэтому, регулируя длину оценок, мы <<размениваем>> смещение на дисперсию, и истина, как всегда, где-то посередине.

Возможность разрешать bias-variance trade-off и выбирать какую-то оценку с промежуточным смещением и дисперсией --- чуть ли не главное преимущество on-policy режима обучения. Напомним, что сэмплы фрагментов траекторий из буфера получить не удастся (действия должны генерироваться из текущей стратегии), и использование многошаговых оценок в off-policy режиме было невозможно.

Пока что в нашем алгоритме роллауты непрактично делать сильно длинными. Дело в том, что пары $s, a$ из длинных роллаутов будут сильно скоррелированы, что потребуется перебивать числом параллельно запущенных сред, а тогда при увеличении длины роллаута начинает раздуваться размер мини-батча. Потом собранные переходы будут использованы всего для одного шага градиентного подъёма, и переиспользовать их будет нельзя; это расточительно и поэтому большой размер мини-батча невыгоден. Поэтому пока можно условно сказать, что самая <<выгодная>> оценка Advantage-а для не очень длинных роллаутов --- оценка максимальной длины: для $s_t$ мы можем построить $N$-шаговую оценку, её и возьмём; для $s_{t+1}$ уже не более чем $N-1$-шаговую; наконец, для $s_{t+N-1}$ нам доступна лишь одношаговая оценка, просто потому что никаких сэмплов после $s_{t+N}$ мы ещё не получали. 

\begin{definition}
Для пар $s_t, a_t$ из роллаута $s_0, a_0, r_0, s_1, a_1, r_1, \dots, s_N$ длины $N$ будем называть \emph{оценкой максимальной длины} (max trace estimation) оценку с максимальным заглядыванием в будущее: для Q-функции
\begin{equation}\label{maxtrace}
y^{\mathrm{MaxTrace}} (s_t, a_t) \coloneqq \sum_{\hat{t}=t}^{N-1} \gamma^{\hat{t}-t} r_{\hat{t}} + \gamma^{N-t} V^\pi(s_N),
\end{equation}
для Advantage функции, соответственно:
\begin{equation*}\label{maxtraceadvantage}
\Psi^{\mathrm{MaxTrace}} (s_t, a_t) \coloneqq y^{\mathrm{MaxTrace}} (s_t, a_t) - V^\pi(s_t)
\end{equation*}
\end{definition}

Заметим, что мы вовсе не обязаны использовать для всех пар $s, a$ оценку одной и той же длины $N$. То есть мы не должны брать для $s_t, a_t$ из $N$-шагового роллаута $N$-шаговую оценку, а остальные пары $s, a$ из роллаута не использовать в обучении лишь потому, что для них $N$-шаговая оценка невозможна; вместо этого для них следует просто использовать ту многошаговую оценку, которая доступна. Это полностью корректно, поскольку любая $N$-шаговая оценка является оценкой Advantage-а для данного слагаемого в нашей формуле градиента. Именно это и говорит оценка максимальной длины: если мы собрали роллаут $s_0, a_0, s_1, a_1 \dots s_5$ длины 5, то одну пару $s_4, a_4$, самую последнюю, нам придётся оценить одношаговой оценкой (поскольку для неё известен сэмпл лишь следующего состояния $s_5$ и только, никаких альтернатив здесь придумать не получится), предпоследнюю пару $s_3, a_3$ --- двухшаговой, и так далее. То есть <<в среднем>> длина оценок будет очень маленькой, такая оценка с точки зрения bias-variance скорее смещена, чем имеет большую дисперсию. 

\subsection{Generalized Advantage Estimation (GAE)}

Пока $N$ не так велико, чтобы дисперсия раздувалась, оценка максимальной длины --- самое разумное решение trade-off-а, и о более умном решении думать не нужно. Однако, впоследствии мы столкнёмся с ситуацией, что on-policy алгоритмы будут собирать достаточно длинные роллауты (порядка тысячи шагов), и тогда брать оценку наибольшей длины уже будет неразумно; с этой же проблемой можно столкнуться и в контексте обсуждаемой Actor-Critic схемы, если длина собираемых роллаутов достаточно большая.

Решение дилеммы bias-variance trade-off подсказывает теория TD($\lambda$) оценки из главы \ref{sec:biasvar}. Нужно применить формулу \eqref{TDlambda} и просто заансамблировать $N$-шаговые оценки разной длины:

\begin{definition}
\emph{GAE-оценкой} Advantage-функции называется ансамбль многошаговых оценок, где оценка длины $N$ \eqref{Nstepadvantage} берётся с весом $\lambda^{N-1}$, где $\lambda \in (0, 1)$ --- гиперпараметр:
\begin{equation}\label{GAEadest}
\Psi_{\mathrm{GAE}}(s, a) \coloneqq (1 - \lambda) \sum_{N > 0} \lambda^{N-1} \Psi_{(N)}(s, a)
\end{equation}
\end{definition}

Как мы помним, при $\lambda \to 0$ такая GAE-оценка соответствует одношаговой оценке; при $\lambda = 1$ GAE-оценка соответствует Монте-Карло оценке Q-функции, которой мы фактически воспользовались, например, в REINFORCE.

\begin{remark}
Как и при дисконтировании, геометрическая прогрессия затухает очень быстро; это означает, что $\lambda \approx 0.9$ больше предпочитает короткие оценки длинным. Поэтому часто $\lambda$ всё-таки близка к 1, типичное значение --- 0.95.
\end{remark}

В текущем виде в формуле суммируются все $N$-шаговые оценки вплоть до конца эпизода. В реальности собранные роллауты могут прерваться в середине эпизода: допустим, для данной пары $s, a$ через $M$ шагов роллаут <<обрывается>>. Тогда на практике используется чуть-чуть другим определением GAE-оценки: если мы знаем $s^{(M)}$, но после этого эпизод ещё не доигран до конца, мы пользуемся формулой \eqref{TDlambda} и оставляем от суммы только <<доступные>> слагаемые:
\begin{equation}\label{truncatedGAE}
\Psi_{\mathrm{GAE}}(s, a) \coloneqq \sum_{t \ge 0}^{M - 1} \gamma^t \lambda^t \Psi_{(1)}(s^{(t)}, a^{(t)})
\end{equation}
Напомним, что это корректно, поскольку соответствует просто занулению следа на $M$-ом шаге, или, что тоже самое, ансамблированию (взятию выпуклой комбинации) первых $M$ многошаговых оценок, где веса <<пропавших>> слишком длинных оценок просто перекладываются в вес самой длинной доступной $M$-шаговой оценки:

\begin{proposition}
Формула \eqref{truncatedGAE} эквивалентна следующему ансамблю $N$-шаговых оценок:
$$\Psi_{\mathrm{GAE}}(s, a) = (1 - \lambda) \sum_{N > 0}^{M-1} \lambda^{N-1} \Psi_{(N)}(s, a) + \lambda^{M-1} \Psi_{(M)}(s, a)$$
\begin{proof}
Следует из доказательства теоремы \ref{th:tdlambda}.
\end{proof}
\end{proposition}

В такой <<обрезанной>> оценке $\lambda \HM= 1$ соответствует оценке максимальной длины \eqref{maxtraceadvantage}, а $\lambda \HM= 0$ всё ещё даст одношаговую оценку.

В коде формула \eqref{truncatedGAE} очень удобна для рекурсивного подсчёта оценки; также для практического алгоритма осталось учесть флаги $\done_t$. Формулы подсчёта GAE-оценки для всех пар $(s, a)$ из роллаута $s_0, a_0, r_0, s_1, a_1, r_1, \cdots s_N$ приобретают такой вид:
\begin{equation*}
\begin{split}
\Psi_{\mathrm{GAE}}(s_{N-1}, a_{N-1}) &\coloneqq \Psi_{(1)}(s_{N-1}, a_{N-1}) \\
\Psi_{\mathrm{GAE}}(s_{N-2}, a_{N-2}) &\coloneqq \Psi_{(1)}(s_{N-2}, a_{N-2}) + \gamma \lambda (1 - \done_{N-2}) \Psi_{\mathrm{GAE}}(s_{N-1}, a_{N-1}) \\
&\vdots \\
\Psi_{\mathrm{GAE}}(s_0, a_0) &\coloneqq \Psi_{(1)}(s_0, a_0) + \gamma \lambda (1 - \done_0) \Psi_{\mathrm{GAE}}(s_1, a_1)
\end{split}
\end{equation*}

Заметим, что эти формулы очень похожи на расчёт кумулятивной награды за эпизод, где <<наградой за шаг>> выступает $\Psi_{(1)}(s, a)$. В среднем наши оценки Advantage должны быть равны нулю, и, если наша аппроксимация этому не удовлетворяет, мы получаем направление корректировки стратегии.

\begin{remark}
Теория говорит, что любые off-policy алгоритмы должны быть более эффективны в плане числа использованных сэмплов, чем on-policy, но на практике может оказаться так, что продвинутые policy gradient алгоритмы, которые мы обсудим позднее, обойдут DQN-подобные алгоритмы из главы \ref{valuebasedchapter} и по sample efficiency. Вероятно, это происходит в ситуациях, когда DQN страдает от проблемы распространения сигнала. При использовании GAE замешивание далёких будущих наград в таргеты позволяет справляться с этой проблемой; если награда разреженная, то имеет смысл выставлять $\lambda$ ближе к единице, если плотная --- ближе к, допустим, 0.9. Распространённый дефолтный вариант 0.95 можно рассматривать как отчасти <<универсальный>>. Если же награда информативная, на каждом шаге поступает какой-то информативный сигнал, то осмыслены и одношаговые обновления; и в таких случаях off-policy алгоритмы на основе value-based подхода скорее всего окажутся более эффективными.
\end{remark}

\subsection{Обучение критика}

Как обучать критика $V_{\phi}(s) \HM\approx V^\pi$? Воспользуемся идеей перехода к регрессии, которую мы обсуждали раньше в контексте DQN (раздел \ref{toregression}). Нам нужно просто решать методом простой итерации уравнение Беллмана \eqref{VV}:
$$V_{\phi_{k+1}}(s) \leftarrow \E_{a} \left[ r + \gamma \E_{s'} V_{\phi_k}(s') \right]$$

Мы можем получить несмещённую оценку правой части $y \HM \coloneqq r \HM+ \gamma V_{\phi_k}(s')$ и с таким таргетом минимизировать MSE. Однако, давайте воспользуемся преимуществами on-policy режима и поймём, что мы можем поступить точно также, как с оценкой Q-функции в формуле градиента: решать многошаговое уравнение Беллмана \eqref{NstepBellman} вместо одношагового. Например, можно выбрать любое $N$-шаговое уравнение и строить целевую переменную как
\begin{equation}\label{criticNsteptarget}
y \coloneqq r + \gamma r' + \gamma^2 r'' + \cdots + \gamma^N V_{\phi_k}(s^{(N)})
\end{equation}
Конечно же, определение того, насколько далеко в будущее заглядывать при построении таргета --- это снова всё тот же самый bias-variance trade-off, и очередное ключевое преимущество on-policy подхода --- разрешать его в том числе при обучении критика.

В чём заключается bias-variance trade-off при обучении критика? С ростом $N$ таргет \eqref{criticNsteptarget} в задаче регрессии становится всё более и более шумным, зато мы быстрее распространяем сигнал и меньше опираемся в таргете на свою же собственную аппроксимацию. Это позволяет бороться с проблемой накапливающейся ошибки, от которой страдают off-policy алгоритмы вроде DQN. В пределе --- при $N \HM= +\infty$ --- мы решаем задачу регрессии, где целевая переменная есть reward-to-go, и начинаем учить V-функцию просто по определению \eqref{Vdefinition}. Такая задача регрессии уже является самой обычной задачей регрессии из машинного обучения, целевая переменная будет являться <<ground truth>>: именно теми значениями, среднее которых мы и хотим выучить. Но такая задача регрессии будет обладать очень шумными целевыми переменными, плюс для сбора таких данных понадобится, опять же, полные эпизоды играть.

Мы можем для оценки Q-функции для обучения политики и для построения целевой переменной для критика использовать разные подходы (скажем, оценки разной длины), но особого смысла в этом немного: хороший вариант для одного будет хорошим вариантом и для другого. Соответственно, можно считать решение trade-off одинаковым для актёра и критика. Тогда если мы оцениваем Advantage как
$$\Psi(s, a) = y - V_{\phi}(s),$$
где $y$ --- некоторая оценка Q-функции, то $y$ же является и таргетом для V-функции, и наоборот. Используя функцию потерь MSE с таким таргетом, мы как раз и учим среднее значение наших оценок Q-функции, то есть бэйзлайн.

Конечно же, мы можем использовать и GAE-оценку \eqref{truncatedGAE} Advantage, достаточно <<убрать бэйзлайн>>:
$$Q^\pi(s, a) = A^\pi(s, a) + V^\pi(s) \approx \Psi_{\mathrm{GAE}}(s, a) + V_{\phi}(s)$$
При этом мы как бы будем решать <<заансамблированные>> N-шаговые уравнения Беллмана для V-функции: 

\begin{proposition}
Таргет $\Psi_{\mathrm{GAE}}(s, a) \HM+ V_{\phi}(s)$ является несмещённой оценкой правой части <<ансамбля>> уравнений Беллмана:
$$V_{\phi}(s) = (1 - \lambda) \sum_{N > 0} \lambda^{N-1} \left[ \B^N V_\phi \right] (s),$$
где $\B$ --- оператор Беллмана для V-функции \eqref{bellmanoperatorV}.
\begin{proof}
По определению, поскольку $\Psi_{(N)}(s, a) \HM+ V(s)$ является несмещённой оценкой правой части $N$-шагового уравнения Беллмана (т.~е. несмещённой оценкой $\left[ \B^N V^\pi \right] (s)$), а
$$(1 - \lambda) \sum_{N > 0} \lambda^{N-1} (\Psi_{(N)}(s, a) + V(s)) = \Psi_{\mathrm{GAE}}(s, a) + V(s)$$
по определению \eqref{GAEadest} GAE-оценки.
\end{proof}
\end{proposition}

Брать для обучения критика набор выходных состояний $s$ мы можем откуда угодно. Поэтому для удобства будем брать $s \HM\sim \mu_\pi(s)$, чтобы можно было использовать для обучения критика и актёра один и тот же мини-батч. Итого получаем следующее: делаем несколько шагов взаимодействия со средой, собирая таким образом роллаут некоторой длины $N$; считаем для каждой пары $s, a$ некоторую оценку Q-функции $y(s, a)$, например, оценку максимальной длины \eqref{maxtrace}; оцениваем Advantage каждой пары как $\Psi(s, a) \HM \coloneqq y(s, a) \HM- V_{\phi}(s)$; далее по Монте-Карло оцениваем градиент по параметрам стратегии
$$\nabla_\theta J(\pi) \approx \frac{1}{N} \sum_{s, a} \nabla_{\theta} \log \pi_\theta (a \mid s) \Psi(s, a)$$
и градиент для оптимизации критика (допустим, критик --- Q-функция):
$$\Loss^{\critic}(\phi) = \frac{1}{N} \sum_{s, a} \left( y(s, a) - V_{\phi}(s) \right)^2$$
Естественно, для декорреляции нужно собрать несколько независимых роллаутов из параллельно запущенных сред.

\begin{remark}
Когда состояния представлены в виде картинок, понятно, что и критику, и актёру нужны примерно одни и те же фичи с изображений (положение одних и тех же распознанных объектов). Поэтому кажется, что если критик и актёр будут двумя разными сетками, их первые слои будут обучаться одному и тому же. Логично для ускорения обучения объединить \emph{экстрактор фич} (feature extractor) для критика и актёра и сделать им просто свои индивидуальные головы. Конечно, тогда нужно обучать всю сеть синхронно, но мы специально для этого считаем градиенты для актёра и критика по одному и тому же мини-батчу. Есть у такого \emph{общего позвоночника} (shared backbone) и свои минусы: лоссы придётся отмасштабировать при помощи скалярного гиперпараметра $\alpha$ так, чтобы одна из голов не забивала градиентами другую:
$$\Loss^{\mathrm{ActorCritic}}(\theta) = \Loss^{\actor}(\theta) + \alpha \Loss^{\critic}(\theta)$$
\end{remark}

\subsection{Advantage Actor-Critic (A2C)}

Мы собрали стандартную схему Advantage Actor Critic (A2C): алгоритма, который работает в чистом виде on-policy режиме. Из-за того, что роллауты в этом алгоритме не очень длинные, используются оценки максимальной длины (или, что тоже самое, GAE с $\lambda \HM= 1$). 

\begin{algorithm}{Advantage Actor-Critic (A2C)}
\textbf{Гиперпараметры:} $M$ --- количество параллельных сред, $N$ --- длина роллаутов, $V_\phi(s)$ --- нейросеть с параметрами $\phi$, $\pi_\theta(a \mid s)$ --- нейросеть для стратегии с параметрами $\theta$, $\alpha$ --- коэф. масштабирования лосса критика, SGD оптимизатор.

\vspace{0.3cm}
Инициализировать $\theta, \phi$ \\
\textbf{На каждом шаге:}
\begin{enumerate}
    \item в каждой параллельной среде собрать роллаут длины $N$, используя стратегию $\pi_{\theta}$:
    $$s_0, a_0, r_0, s_1, \dots , s_N$$
    \item для каждой пары $s_t, a_t$ из каждого роллаута посчитать оценку Q-функции максимальной длины, игнорируя зависимость оценки от $\phi$:
    $$Q(s_t, a_t) \coloneqq \sum_{\hat{t} = t}^{N-1} \gamma^{\hat{t} - t} r_{\hat{t}} + \gamma^{N - t} V_{\phi}(s_N)$$
    \item вычислить лосс критика:
    $$\Loss^{\critic}(\phi) \coloneqq \frac{1}{MN}\sum_{s_t, a_t} \left( Q(s_t, a_t) - V_\phi(s_t) \right) ^2$$
    \item делаем шаг градиентного спуска по $\phi$, используя $\nabla_\phi \Loss^{\critic}(\phi)$
    \item вычислить градиент для актёра:
    $$\nabla^{\actor}_\theta \coloneqq \frac{1}{MN}\sum_{s_t, a_t} \nabla_\theta \log \pi_\theta(a_t \mid s_t) \left( Q(s_t, a_t) - V_\phi(s_t) \right) $$
    \item сделать шаг градиентного подъёма по $\theta$, используя $\nabla^{\actor}_\theta$
\end{enumerate}
\end{algorithm}

\begin{remark}
Для того, чтобы поощрить исследование среды, к суррогатной функции потерь актёра добавляют ещё одно слагаемое, <<регуляризатор>>, поощряющий высокую энтропию. Обычно, энтропию распределения $\pi_{\theta}(\cdot \mid s_t)$ для выбранной параметризации распределения можно посчитать аналитически и дифференцировать по $\theta$. Естественно, этот <<дополнительный лосс>> тоже нужно грамотно взвесить на скалярный коэффициент. Позже в разделе \ref{maximumentropyrlsubsection} мы встретимся с постановкой задачи Maximum Entropy RL, в рамках которой появление этого слагаемого можно объяснить небольшим изменением в самом оптимизируемом функционале.
\end{remark}

\begin{remark}
Известно, что Policy Gradient алгоритмы особенно чувствительны к инициализации нейросетей. Рекомендуют ортогональную инициализацию слоёв. Также крайне существенна обрезка градиентов, которая защищает от взрывов в градиентной оптимизации. При обучении policy gradient алгоритмов важно логгировать норму градиентов; большое значение нормы сигнализирует о неудачном подборе параметров.
\end{remark}

\begin{example}
Перерыв на \href{https://hackernoon.com/intuitive-rl-intro-to-advantage-actor-critic-a2c-4ff545978752}{чтение комиксов}.
\end{example}

Сравним A2C с value-based алгоритмами на основе DQN. Когда мы моделировали Value Iteration, мы целиком и полностью <<полагались>> на этап оценивания стратегии, то есть на обучение критика: модели актёра в явном виде даже не было в алгоритме, поскольку текущая стратегия всё время считалась жадной по отношению к текущему критику. Это означало, что качество стратегии упиралось в качество критика: пока Q-функция не научится адекватно приближать истинную $Q^*$, стратегия хорошей не будет. 

Достаточно интересно, что идея model-free алгоритмов, отказывающихся обучать модель функции переходов в том числе из соображений end-to-end схемы, пришла к тому, что мы иногда сводимся к обучению оценочных функций --- промежуточных величин, вместо того, чтобы напрямую матчить состояния и действия, понимать, какие действия стоит выбирать чаще других. Вообще, интуиция подсказывает, что обучать актёра проще, чем оценочную функцию\footnote{есть, однако, и ряд теоретических наблюдений, которые говорят, что эти задачи скорее эквивалентны по сложности. На это указывают формулы градиентов для обучения критика и актёра: так, для актёра мы идём по градиенту
$$\nabla_\theta \log \pi_\theta(a \mid s) \Psi(s, a),$$
а для критика мы идём по градиенту (продифференцируйте MSE, чтобы убедиться в этом):
$$\nabla_\phi V_\phi(s) \Psi(s, a)$$
Как видно, это как будто бы один и тот же градиент: он просто говорит увеличивать соответствующий выход нейронной сети, а кредит предоставляет скаляр, сообщающий масштаб изменения и, главное, знак.
}: в реальных задачах человек редко когда думает в терминах будущей награды. 

\begin{example}
Представьте, что вы стоите перед кофеваркой и у вас есть два действия: получить кофе и не получить. Вы прекрасно знаете, какое действие лучше другого, но при этом не оцениваете, сколько кофе вы сможете выпить в будущем при условии, например, что сейчас вы выберете первый или второй вариант: то есть не строите в явном виде прогноз значения оценочной функции.
\end{example}

И формула градиента, идея policy gradient методов, как раз предоставляет возможность обучать актёра напрямую, минуя промежуточный этап в виде критика: так, в алгоритме \ref{REINFORCE} REINFORCE мы могли использовать Монте-Карло оценки Q-функции и обходиться без обучения в явном виде модели критика, то есть полностью опираться на этап policy improvement-а. Поэтому policy gradient алгоритмы ещё иногда называют \emph{policy-based}. Таким образом, DQN и REINFORCE --- это два <<крайних случая>>, первый алгоритм полностью опирается на критика, а второй алгоритм --- на актёра. Только у первого недостатки компенсируются возможностью обучаться в off-policy режиме и использовать реплей буфер, а вот недостатки алгоритма REINFORCE --- высокая дисперсия и необходимость играть целые эпизоды --- не имеют аналогичного противовеса. 

Важно, что в Policy Gradient алгоритмах мы за счёт использования метода REINFORCE при расчёте градиента можем заменять значение Q-функции (необходимую для улучшения политики) на какую-то заглядывающую в будущее оценку. И насколько сильно при обучении актёра опираться на критика (<<насколько далеко в будущее заглядывать>>) --- это и есть bias-variance trade-off, который в on-policy алгоритмах возможно разрешать. За счёт этого A2C в отличие от DQN может использовать и в качестве таргета для обучения критика, и в качестве оценки для обучения Q-функции GAE-оценку \eqref{truncatedGAE}. Да, пока что в A2C роллауты (зачастую) получаются слишком короткими, и GAE ансамбль <<бедный>>, приходится $\lambda \HM= 1$ использовать, но главное, что есть такая возможность технически, и даже короткие роллауты уже позволяют существенно справиться с проблемой распространения сигнала: это помогает и критику лучше учиться, и актёр может не полностью на критика опираться. В частности, нам не нужна модель Q-функции, достаточно более простой V-функции. Всё это делает Policy Gradient алгоритмы куда более эффективными в средах с сильно отложенным сигналом.

Наконец, ещё одно небольшое, но важное преимущество Policy Gradient --- обучение стохастичной политики. Хотя мы знаем, что оптимальная политика детерминирована, обучение стохастичной политики позволяет использовать её для сбора данных, и стохастичность --- ненулевая вероятность засэмплировать любое действие --- частично решает проблему исследования. Да, это, конечно, далёкое от идеала решение, но намного лучшее, чем, например, $\eps$-жадная стратегия: можно рассчитывать на то, что если градиент в какой-то области пространства состояний постоянно указывает на то, что одни действия хорошие, а другие плохие, актёр выучит с вероятностью, близкой к единице, выбирать хорошие действия; если же о действиях в каких-то состояниях приходит противоречивая информация, или же такие состояния являются для агента новыми, то можно надеяться, что стратегия будет близка к равномерной, и агент будет пробовать все действия.

Все эти преимущества неразрывно связаны с on-policy режимом. За счёт <<свежести>> данных, можно делать, так сказать, наилучшие возможные шаги обучения стратегия, практически идти по градиенту оптимизируемого функционала. Но из него проистекает и главный недостаток подхода: неэффективность по количеству затрачиваемых сэмплов. Нам всё время нужны роллауты, сгенерированные при помощи текущей политики с параметрами $\theta$, чтобы посчитать оценку градиента в точке $\theta$ и сделать шаг оптимизации. После этого будут нужны уже сэмплы из новой стратегии, поэтому единственное, что мы можем сделать с уже собранными данными --- почистить оперативную память.

Понятно, что это полное безобразие: допустим, агент долго вёл себя псевдослучайно и наконец наткнулся на какой-то хороший исход с существенным откликом среды. Схема сделает по ценному роллауту всего один градиентный шаг, что скорее всего не поможет модели выучить удачное поведение или значение отклика. После этого ценная информация никак не может быть использована и придётся дожидаться следующей удачи, что может случиться нескоро. Это очень sample inefficient и основная причина, почему от любого on-policy обучения в чистом виде нужно пытаться отойти.

\begin{example}
Допустим, вы играете в видео-игру, и в начале обучения мало что умеете, всё время действуя примерно случайно и падая в первую же яму. Среда выдаёт вам всё время ноль, и вы продолжаете вести себя случайно. Вдруг в силу стохастичности стратегии вы перепрыгиваете первую яму и получаете монетку +1. В DQN этот ценнейший опыт будет сохранён в буфере, и постепенно критик выучит, какие действия привели к награде. В A2C же агент сделает один малюююсенький шаг изменения весов моделей и тут же выкинет все собранные данные в мусорку, потому что на следующих итерациях он никак не может переиспользовать их. Агенту придётся ждать ещё много-много сессий в самой игре, пока он не перепрыгнет яму снова, чтобы сделать следующий шаг обучения перепрыгиванию ям.
\end{example}

Частично с этой проблемой нам удастся побороться в следующей главе, что позволит построить алгоритмы лучше A2C. Но важно, что полностью решить эту проблему, полностью перейти в off-policy режим, нам не удастся точно также, как в off-policy мы не смогли адекватно разрешать bias-variance trade-off (и там максимум наших возможностей была Retrace-оценка). Поэтому построить алгоритм, берущий <<лучшее от двух миров>>, нельзя, и поэтому среди model-free алгоритмов выделяют два класса алгоритмов: off-policy, работающие с реплей буфером и потому потенциально более sample efficient, и on-policy алгоритмы, где есть ряд вышеуказанных преимуществ.
\section{Продвинутые Policy Gradient}\label{TRPOPPOsection}

\subsection{Суррогатная функция}

Как можно хотя бы частично побороться с ключевой проблемой Policy Gradient подхода --- с on-policy режимом? Есть два соображения, как это можно попробовать сделать.

Во-первых, можно попробовать вместо градиентного подъёма использовать какой-то более тяжеловесный метод оптимизации, например, методы оптимизации второго порядка. Такие методы обычно требуют меньше обращений к оракулу, делая меньше итераций, за счёт более высокой вычислительной сложности каждого шага. В нашем случае это может быть ровно то, что нам нужно --- вызов оракула для нас это сбор данных, и какие-то дополнительные вычисления <<на компьютере>> можно считать дешёвой процедурой. 

Вторая идея --- можно как-то попробовать посчитать оценку градиента в точке текущих параметров, при этом используя сэмплы, полученные при помощи другой стратегии. Полностью перейти в off-policy режим с сохранением всех преимуществ on-policy у нас не получится, поэтому придётся наложить ограничение на эту другую стратегию сбора данных: она должна быть очень похожей версией оцениваемой стратегии, то есть, можно считать, недавней версии актуальной стратегии. То, что стратегия сбора данных <<похожа>> на текущую, означает, что её траектории тоже в некотором смысле <<похожи>>: понятно, что это важно, ведь много информации о градиенте политики по около-оптимальным параметрам с траекторий случайного агента собрать не получится, просто потому что такие траектории у около-оптимальной стратегии почти не встретятся.

На самом деле эти две идеи примерно об одном и том же, что мы и увидим далее.

Итак, допустим мы хотим оптимизировать стратегию $\textcolor{ChadBlue}{\pi_\theta}$ по параметрам $\textcolor{ChadBlue}{\theta}$, используя только сэмплы из другой стратегии $\textcolor{ChadPurple}{\pi^{\old}}$ (в итоговой схеме это будет недавняя версия стратегии). Что нам тогда мешает оценить значение \eqref{advantagepg}?

$$
\nabla_{\theta} J(\pi) = \frac{1}{1 - \gamma}\E_{\textcolor{ChadBlue}{d_\pi}(s)} \E_{\textcolor{ChadBlue}{\pi}(a \mid s)} \nabla_{\theta} \log \textcolor{ChadBlue}{\pi_\theta} (a \mid s) \textcolor{ChadBlue}{A^{\pi}}(s, a)
$$

Во-первых, у нас нет сэмплов из $\textcolor{ChadBlue}{d_\pi}(s)$, ведь в данных есть только\footnote{с учётом забивания на $\gamma^t$ и наших прочих оговорок; но для корректности выкладок будем в этой главе писать везде дисконтированные частоты посещения состояний $d$} состояния из $\textcolor{ChadPurple}{d_{\pi^{\old}}}(s)$. Мы уже обсуждали, что поскольку формула говорит проводить policy improvement, мы можем подменить это распределение на любое другое, и схема останется рабочей. Однако ключевой является вторая проблема. У нас нет оценки $\textcolor{ChadBlue}{A^{\pi}}(s, a)$. Если мы попробуем оценить Advantage для одной стратегии по данным из другой, то нам придётся вводить все те коррекции, которые мы обсуждали в разделе \ref{subsec:retrace} про Retrace оценку, которая скорее всего схлопнется в смещённую. Очень хотелось бы всё-таки оставить <<чистую>> GAE-оценку, то есть как-то использовать <<несвежего>> критика  $\textcolor{ChadPurple}{A^{\pi^{\old}}}(s, a)$ в формуле градиента.


Как сделать так, чтобы у нас где-то в формулах образовался <<несвежий>> критик? На помощь приходит RPI, формула \eqref{RPI}: мы можем сменить функцию награды на Advantage функцию любой другой стратегии! Давайте запишем это в следующем виде:

\begin{proposition}
\,
\begin{equation}\label{RPIdecoupled}
J(\textcolor{ChadBlue}{\pi_{\theta}}) = J(\textcolor{ChadPurple}{\pi^{\old}}) + \frac{1}{1 - \gamma}\E_{s \sim \textcolor{ChadBlue}{d_{\pi_{\theta}}}(s)} \E_{a \sim \textcolor{ChadBlue}{\pi_{\theta}}(a \mid s)} \textcolor{ChadPurple}{A^{\pi^{\old}}}(s_t, a_t)
\end{equation}
\begin{proof}
Из RPI \eqref{RPI} следует, что для любых двух стратегий верно 
$$J(\textcolor{ChadBlue}{\pi_{\theta}}) = J(\textcolor{ChadPurple}{\pi^{\old}}) + \E_{\Traj \sim \textcolor{ChadBlue}{\pi_{\theta}}} \sum_{t \ge 0} \gamma^t \textcolor{ChadPurple}{A^{\pi^{\old}}}(s_t, a_t)$$
Для получения формулы достаточно переписать мат.ожидание по траекториям из $\textcolor{ChadBlue}{\pi_{\theta}}$ через state visitation distribution, подставив в теореме \ref{th:decoupling_stoch} в качестве $f(s, a) \HM= \textcolor{ChadPurple}{A^{\pi^{\old}}}(s, a)$.
\end{proof}
\end{proposition}

Тогда градиент исходного функционала $\textcolor{ChadBlue}{\nabla_\theta} J(\textcolor{ChadBlue}{\pi_{\theta}})$ есть градиент правой части, и в ней уже как-то <<замешана>> оценка Advantage для стратегии $\textcolor{ChadPurple}{\pi^{\old}}$, которое мы сможем посчитать при помощи GAE-оценки:

$$\textcolor{ChadBlue}{\nabla_\theta} J(\textcolor{ChadBlue}{\pi_{\theta}}) = \textcolor{ChadBlue}{\nabla_\theta} \frac{1}{1 - \gamma}\E_{s \sim \textcolor{ChadBlue}{d_{\pi_{\theta}}}(s)} \E_{a \sim \textcolor{ChadBlue}{\pi_{\theta}}(a \mid s)} \textcolor{ChadPurple}{A^{\pi^{\old}}}(s_t, a_t)$$

Мы также можем справиться с мат.ожиданием $\E_{a \sim \textcolor{ChadBlue}{\pi_{\theta}}(a \mid s)}$ при помощи importance sampling. Да, этот коэффициент может быть ужасен (сильно большим единицы или близким к нулю), но это коррекция всего лишь за один шаг, и такая дробь будет терпимой.

\begin{equation}\label{perfectsurrogate}
\textcolor{ChadBlue}{\nabla_\theta} J(\textcolor{ChadBlue}{\pi_{\theta}}) = \textcolor{ChadBlue}{\nabla_\theta} \frac{1}{1 - \gamma} \E_{s \sim \textcolor{ChadBlue}{d_{\pi_{\theta}}}(s)} \E_{a \sim \textcolor{ChadPurple}{\pi^{\old}}(a \mid s)} \frac{\textcolor{ChadBlue}{\pi_{\theta}}(a \mid s)}{\textcolor{ChadPurple}{\pi^{\old}}(a \mid s)} \textcolor{ChadPurple}{A^{\pi^{\old}}}(s_t, a_t)
\end{equation}

Осталась последняя проблема: $\textcolor{ChadBlue}{d_{\pi_{\theta}}}(s)$. В роллаутах, сгенерированных при помощи $\textcolor{ChadPurple}{\pi^{\old}}$, состояния всё-таки будут приходить из $\textcolor{ChadPurple}{d_{\pi^{\old}}}(s)$, и тут мы importance sampling не сделаем даже при большом желании, так как просто не можем внятно оценить эти величины: из частот посещения состояний мы можем только сэмплировать.

Рассмотрим аппроксимацию: что, если мы в формуле \eqref{perfectsurrogate} заменим $\textcolor{ChadBlue}{d_{\pi_{\theta}}}(s)$ на $\textcolor{ChadPurple}{d_{\pi^{\old}}}(s)$? Это, вообще говоря, будет какая-то другая функция.

\begin{definition}
Введём суррогатную функцию:
\begin{equation}\label{surrogate}
L_{\textcolor{ChadPurple}{\pi^{\old}}}(\textcolor{ChadBlue}{\theta}) \coloneqq \frac{1}{1 - \gamma}\E_{s \sim \textcolor{ChadPurple}{d_{\pi^{\old}}}(s)} \E_{a \sim \textcolor{ChadPurple}{\pi^{\old}}(a \mid s)} \frac{\textcolor{ChadBlue}{\pi_\theta}(a \mid s)}{\textcolor{ChadPurple}{\pi^{\old}}(a \mid s)} \textcolor{ChadPurple}{A^{\pi^{\old}}}(s, a)   
\end{equation}
\end{definition}

\begin{proposition}
Сэмплы из мат.ожиданий в суррогатной функции --- это сэмплы из роллаутов, сгенерированных при помощи $\textcolor{ChadPurple}{\pi^{\old}}$:
\begin{equation*}
L_{\textcolor{ChadPurple}{\pi^{\old}}}(\textcolor{ChadBlue}{\theta}) = \E_{\Traj \sim \textcolor{ChadPurple}{\pi^{\old}}} \sum_{t \ge 0} \gamma^t \frac{\textcolor{ChadBlue}{\pi_\theta}(a \mid s)}{\textcolor{ChadPurple}{\pi^{\old}}(a \mid s)} \textcolor{ChadPurple}{A^{\pi^{\old}}}(s, a)
\end{equation*}
\begin{proof}[Пояснение]
Применить теорему \ref{th:decoupling_stoch} в обратную сторону для $f(s, a) \HM= \frac{\textcolor{ChadBlue}{\pi_\theta}(a \mid s)}{\textcolor{ChadPurple}{\pi^{\old}}(a \mid s)} \textcolor{ChadPurple}{A^{\pi^{\old}}}(s, a)$
\end{proof}
\end{proposition}

Какой у этой суррогатной функции $L_{\textcolor{ChadPurple}{\pi^{\old}}}(\textcolor{ChadBlue}{\theta})$ физический смысл? Мы сказали, что наши настраиваемые параметры $\textcolor{ChadBlue}{\theta}$ не влияют на частоты посещения состояний, то есть выбор действий не влияет на то, какие состояния мы будем посещать в будущем. Это очень сильное допущение, поэтому аппроксимация не самая удачная. Давайте поймём, что будет происходить, если мы будем оптимизировать вместо честного, исходного функционала, такую суррогатную функцию при фиксированной $\textcolor{ChadPurple}{\pi^{\old}}$:
\begin{equation}\label{surrogate_opt}
L_{\textcolor{ChadPurple}{\pi^{\old}}}(\textcolor{ChadBlue}{\theta}) \to \max_{\textcolor{ChadBlue}{\theta}}
\end{equation}

\begin{proposition}\label{prop:surrogateoptispi}
Решением оптимизационной задачи \eqref{surrogate_opt} является жадная стратегия по отношению к $\textcolor{ChadPurple}{Q^{\pi^{\old}}}(s, a)$ (или, что тоже самое, к Advantage-функции стратегии $\textcolor{ChadPurple}{\pi^{\old}}$):
$$\textcolor{ChadBlue}{\pi_\theta}(s) = \argmax_{a} \textcolor{ChadPurple}{A^{\pi^{\old}}}(s, a)$$

\begin{proof}
Эта оптимизационная задача распадается на оптимизационную задачу для каждого $s$, а решением задачи
$$\E_{a \sim \textcolor{ChadPurple}{\pi^{\old}}(a \mid s)} \frac{\textcolor{ChadBlue}{\pi_\theta}(a \mid s)}{\textcolor{ChadPurple}{\pi^{\old}}(a \mid s)} \textcolor{ChadPurple}{A^{\pi^{\old}}}(s, a) = \E_{a \sim \textcolor{ChadBlue}{\pi_\theta}(a \mid s)} \textcolor{ChadPurple}{A^{\pi^{\old}}}(s, a) \to \max_{\textcolor{ChadBlue}{\theta}}$$
является <<жадная>> стратегия.
\end{proof}
\end{proposition}

Другими словами, такая суррогатная функция просто говорит проводить policy improvement стратегии $\textcolor{ChadPurple}{\pi^{\old}}$, но в состояниях, приходящих из частот посещения состояний $\textcolor{ChadPurple}{d_{\pi^{\old}}}(s)$. Такая очередная форма нам сейчас будет удобна, поскольку такая суррогатная функция является локальной аппроксимацией нашего оптимизируемого функционала, причём эта аппроксимация --- не просто, скажем, линейное приближение оптимизируемой функции, а какое-то более умное, учитывающее особенности задачи. 

\subsection{Нижняя оценка}

Итак, у нас есть аппроксимация нашего оптимизируемого функционала через суррогатную функцию, с которой мы можем работать:
$$J(\textcolor{ChadBlue}{\pi}) \approx J(\textcolor{ChadPurple}{\pi^{\old}}) + L_{\textcolor{ChadPurple}{\pi^{\old}}}(\textcolor{ChadBlue}{\theta}) = J(\textcolor{ChadPurple}{\pi^{\old}}) + \frac{1}{1 - \gamma}\E_{s \sim \textcolor{ChadPurple}{d_{\pi^{\old}}}(s)} \E_{a \sim \textcolor{ChadPurple}{\pi^{\old}}(a \mid s)} \frac{\textcolor{ChadBlue}{\pi_\theta}(a \mid s)}{\textcolor{ChadPurple}{\pi^{\old}}(a \mid s)} \textcolor{ChadPurple}{A^{\pi^{\old}}}(s, a)$$

Насколько эта аппроксимация хороша? Наша интуиция была в том, что если $\textcolor{ChadBlue}{\pi}$ <<похожа>> на $\textcolor{ChadPurple}{\pi^{\old}}$, то частоты посещения состояний у них тоже наверняка будут похожи. Формализовать <<похожесть>> стратегий можно, например, так:
\begin{definition}
Введём расстояние между стратегиями $\textcolor{ChadPurple}{\pi^{\old}}, \textcolor{ChadBlue}{\pi_\theta}$ как среднюю KL-дивергенцию между ними по состояниям из частот посещения первой стратегии:
$$\KL(\textcolor{ChadPurple}{\pi^{\old}} \parallel \textcolor{ChadBlue}{\pi_\theta}) \coloneqq \E_{s \sim \textcolor{ChadPurple}{d_{\pi^{\old}}}(s)} \KL(\textcolor{ChadPurple}{\pi^{\old}}(a \mid s) \parallel \textcolor{ChadBlue}{\pi_\theta}(a \mid s))$$
\end{definition}

\begin{theoremBox}[label=th:Lapproximationestimation]{}
\,
\begin{equation*} 
\left| J(\textcolor{ChadBlue}{\pi_\theta}) - J(\textcolor{ChadPurple}{\pi^{\old}}) - L_{\textcolor{ChadPurple}{\pi^{\old}}}(\textcolor{ChadBlue}{\theta}) \right| \le C \sqrt{\KL(\textcolor{ChadPurple}{\pi^{\old}} \parallel \textcolor{ChadBlue}{\pi_\theta}) }
\end{equation*}
где $C$ --- константа, равная $C \HM= \frac{\sqrt{2} \gamma}{(1 - \gamma)^2} \max\limits_{s, a} |\textcolor{ChadPurple}{A^{\pi^{\old}}}(s, a)|$
\begin{proof}[Без доказательства; интересующиеся могут обратиться к статье \href{https://arxiv.org/pdf/1705.10528.pdf}{Constrained Policy Optimization}]
\end{proof}
\end{theoremBox}

Конечно, это очень грубая оценка, хотя бы потому, что она верна для произвольных MDP и произвольных двух стратегий. Но ключевой момент в том, что мы теперь формально можем вывести \emph{нижнюю оценку} (lower bound) на оптимизируемый функционал\footnote{исторически в статьях по TRPO и PPO использовалась чуть более грубая нижняя оценка, в которой ошибка между суррогатной функцией и честным функционалом оценивалась сверху при помощи KL-дивергенции в максимальной форме:
$$\KL^{\max}(\textcolor{ChadPurple}{\pi^{\old}} \parallel \textcolor{ChadBlue}{\pi}) \coloneqq \max_s \KL(\textcolor{ChadPurple}{\pi^{\old}}(a \mid s) \parallel \textcolor{ChadBlue}{\pi_\theta}(a \mid s))$$
Однако такое выражение посчитать в практических алгоритмах нельзя, поэтому далее её приходилось эвристически заменять на среднее по $s \HM\sim \textcolor{ChadPurple}{d_{\pi^{\old}}}(s)$. Уточнённая нижняя оценка обосновывает этот переход и указывает, что из KL-дивергенции также нужно взять корень; в остальном на ход дальнейших рассуждений это не влияет.
}:
\begin{theorem}[Performance Lower Bound]
\begin{equation}\label{lowerbound}
J(\textcolor{ChadBlue}{\pi_\theta}) - J(\textcolor{ChadPurple}{\pi^{\old}}) \ge L_{\textcolor{ChadPurple}{\pi^{\old}}}(\textcolor{ChadBlue}{\theta}) - C \sqrt{\KL(\textcolor{ChadPurple}{\pi^{\old}} \parallel \textcolor{ChadBlue}{\pi_{\theta}})}
\end{equation}
\end{theorem}

Возникает любопытнейшая идея: возможно, мы можем работать не с исходным функционалом, а с нижней оценкой.

\begin{theorem}
Процедура оптимизации
\begin{equation}\label{lowerboundopt}
\textcolor{ChadBlue}{\theta_{k+1}} \coloneqq \argmax_{\textcolor{ChadBlue}{\theta}} \left[ L_{\textcolor{ChadPurple}{\pi_{\theta_k}}}(\textcolor{ChadBlue}{\theta}) - C \sqrt{\KL(\textcolor{ChadPurple}{\pi_{\theta_k}} \parallel \textcolor{ChadBlue}{\pi_\theta})} \right]
\end{equation}
гарантирует монотонное неубывание функционала: $J(\textcolor{ChadBlue}{\pi_{\theta_{k+1}}}) \ge J(\textcolor{ChadPurple}{\pi_{\theta_{k}}})$
\begin{proof}
В точке $\theta \HM= \theta_k$ суррогатная функция $L_{\pi_{\theta_k}}(\theta)$ равна нулю, поскольку
\begin{align*}
L_{\pi_{\theta_k}}(\theta_k) &= \\
= \{\text{по определению \eqref{surrogate}}\} &= \frac{1}{1 - \gamma}\E_{s \sim d_{\pi_{\theta_{k}}}(s)} \E_{a \sim \pi_{\theta_{k}}(a \mid s)} \frac{\pi_{\theta_{k}}(a \mid s)}{\pi_{\theta_{k}}(a \mid s)} A^{\pi_{\theta_{k}}}(s, a) = \\ 
&= \frac{1}{1 - \gamma}\E_{s \sim d_{\pi_{\theta_{k}}}(s)} \E_{a \sim \pi_{\theta_{k}}(a \mid s)} A^{\pi_{\theta_{k}}}(s, a) = \\
\{\text{следствие RPI \eqref{RPIdecoupled}}\} &= J(\pi_{\theta_{k}}) - J(\pi_{\theta_{k}}) = 0
\end{align*}

Понятно, что $\KL(\pi_{\theta_k} \parallel \pi_\theta)$ в точке $\theta \HM= \theta_k$ тоже равна 0, так как во всех состояниях $\KL$ между одинаковыми стратегиями равна нулю. Значит, максимум нижней оценки не меньше нуля, а она есть нижняя оценка на $J(\textcolor{ChadBlue}{\pi_{\theta_{k+1}}}) -\HM J(\textcolor{ChadPurple}{\pi_{\theta_{k}}})$.
\end{proof}
\end{theorem}

\needspace{9\baselineskip}
\begin{wrapfigure}[9]{r}{0.5\textwidth}
\centering
\includegraphics[width=0.5\textwidth]{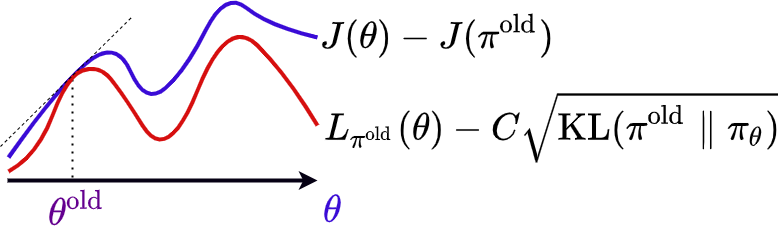}
\end{wrapfigure}

Итак, в воздухе витает идея заняться типичным \emph{minorization-maximization} алгоритмом. Сначала мы подтягиваем нашу нижнюю оценку так, чтобы в текущей точке $\theta \HM= \theta^{\old}$ она в точности совпадала с оптимизируемой функцией (minorization). Затем мы начинаем при фиксированном $\textcolor{ChadPurple}{\theta^{\old}}$ оптимизировать по $\textcolor{ChadBlue}{\theta}$ не наш функционал (который мы не сможем оптимизировать без новых сэмплов из новой стратегии $\textcolor{ChadBlue}{\pi_\theta}$ с текущими значениями параметров), а нижнюю оценку, с которой умеем работать (maximization). 

То есть, что мы получили: мы <<можем>> оптимизировать не $J(\textcolor{ChadBlue}{\pi_\theta}) \HM- J(\textcolor{ChadPurple}{\pi^{\old}})$ по формуле \eqref{perfectsurrogate}, а нашу суррогатную функцию $L_{\textcolor{ChadPurple}{\pi^{\old}}}(\textcolor{ChadBlue}{\theta})$, то есть проводить policy improvement стратегии $\textcolor{ChadPurple}{\pi^{\old}}$, но добавив штраф --- регуляризатор --- за различие между $\textcolor{ChadBlue}{\pi_\theta}$ и стратегией, из которой приходят данные $\textcolor{ChadPurple}{\pi^{\old}}$:

\begin{equation}\label{lowerboundmean}
L_{\textcolor{ChadPurple}{\pi^{\old}}}(\textcolor{ChadBlue}{\theta}) - C \sqrt{\KL(\textcolor{ChadPurple}{\pi^{\old}} \parallel \textcolor{ChadBlue}{\pi_\theta}) } \to \max_{\textcolor{ChadBlue}{\theta}}
\end{equation}

Мы получили процедуру, гарантирующую улучшение стратегии, что звучит подозрительно хорошо. Очень похожие гарантии улучшения стратегии у нас были в policy improvement. Интересно порассуждать, в чём отличие. Мы уже увидели в утверждении \ref{prop:surrogateoptispi}, что оптимизация суррогатной функции без регуляризатора соответствует policy improvement-у стратегии $\textcolor{ChadPurple}{\pi^{\old}}$. Однако как мы помним, жадный policy improvement вовсе не является наилучшим: если мы в некотором состоянии $s$ перекладываем вероятностную массу в какое-то действие, чтобы увеличить среднее значение $\textcolor{ChadPurple}{A^{\pi^{\old}}}(s, a)$, мы попадаем чаще в те состояния, где стратегия $\textcolor{ChadPurple}{\pi^{\old}}$ набирает больше, да, но на самом деле мы таким изменением стратегии можем помешать себе добираться чаще до тех состояний, где $\textcolor{ChadPurple}{A^{\pi^{\old}}}(s, a)$ принимает большие положительные значения!

\begin{exampleBox}[righthand ratio=0.4, sidebyside, sidebyside align=center, lower separated=false]{}
Что говорит policy improvement для MDP с картинки и заданной стратегией $\textcolor{ChadPurple}{\pi^{\old}}$? В начальном состоянии $\textcolor{ChadPurple}{Q^{\pi^{\old}}}(s, \colorsquare{ChadBlue}) \HM= +10$, $\textcolor{ChadPurple}{Q^{\pi^{\old}}}(s, \colorsquare{ChadRed}) \HM= -100$, и поэтому улучшение будет заключаться в том, что новая стратегия должна чаще выбирать именно действие $\colorsquare{ChadBlue}$, выбирать +10. Это связано с тем, что policy improvement игнорирует влияние действий на частоты посещения состояний, и оптимизирует <<новую награду>> $\textcolor{ChadPurple}{A^{\pi^{\old}}}(s, a)$ независимо в каждом состоянии жадным образом. Он не видит, что выбор в начальном состоянии действия $\colorsquare{ChadRed}$ позволит ему попасть в такое состояние, где для одного из действий $\textcolor{ChadPurple}{A^{\pi^{\old}}}(s_2, \colorsquare{ChadBlue}) \HM= +200$!

\tcblower
\vspace{-0.2cm}
\begin{adjustwidth}{-0.35cm}{0cm}
\includegraphics[width=\textwidth]{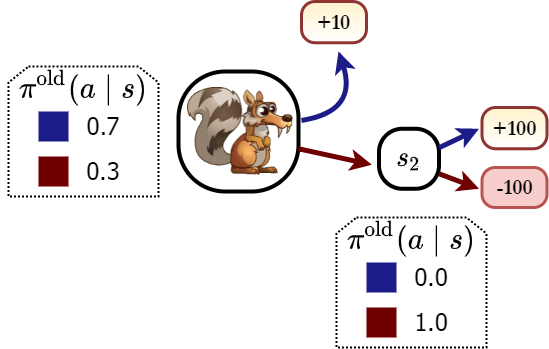}
\end{adjustwidth}
\end{exampleBox}

Поэтому формула policy gradient --- <<наилучшего policy improvement-а>> --- говорит, что как только параметры $\textcolor{ChadBlue}{\theta}$ хоть чуть-чуть изменяются, сразу же стоит улучшать новую стратегию, то есть использовать свежего критика. Мы же свежего критика получить не можем, хотим пользоваться лишь <<несвежим>> $\textcolor{ChadPurple}{A^{\pi^{\old}}}(s, a)$, и нижняя оценка \eqref{lowerboundmean} даёт промежуточную альтернативу, что тогда можно делать: добавить регуляризатор, запрещающий сильное изменение исходной стратегии $\textcolor{ChadPurple}{\pi^{\old}}$.

Однако чтобы оптимизировать \eqref{lowerboundmean}, нужно как-то определить значение константы $C$: мы не умеем считать выражение для неё из теоремы \ref{th:Lapproximationestimation}, поскольку там присутствует максимум Advantage-функции по всем парам состояние-действие\footnote{мы могли бы оценить его сверху как $2R^{\max}$, где $R^{\max}$ --- максимальная награда, но это было бы непрактично грубой оценкой.}. Мы можем заменить его на гиперпараметр, но, чтобы не потерять теоретические гарантии, он должен быть достаточно большим, чтобы превосходить значение из теоремы. Вообще, скорее всего, даже если бы мы знали эту константу, она была бы колоссальной: чего стоит только $(1 \HM- \gamma)^2$ в знаменателе формулы из теоремы \ref{th:Lapproximationestimation}. Поэтому практической пользы от такой нижней оценки всё равно много бы не было: её оптимизация делала бы слишком консервативные шаги обновления политики.



\subsection{Trust Region Policy Optimization (TRPO)}

В TRPO предлагается воспользоваться полученной теорией, чтобы построить на основе идеи суррогатной функции более <<мощный>> метод оптимизации, чем обычный градиентный спуск. Как обычно устроены методы оптимизации? В текущей точке для рассматриваемой функции строится какая-то модель, какая-то локально верная аппроксимация (например, линейная или квадратичная). Далее эта модель либо оптимизируется, и алгоритм сдвигает текущее решение в сторону оптимума модели (такие методы относят к \emph{line search} подходу), либо модель оптимизируется в некотором <<\emph{регионе доверия}>> (trust region), где эта модель, считается, более-менее похожа на исходный функционал. Подход на основе регионов доверия считается более тяжеловесным, поскольку требует решать на каждом шаге работы алгоритма условную задачу оптимизации, зато более эффективным по числу итераций.

\begin{wrapfigure}{r}{0.2\textwidth}
\vspace{-0.5cm}
\centering
\includegraphics[width=0.2\textwidth]{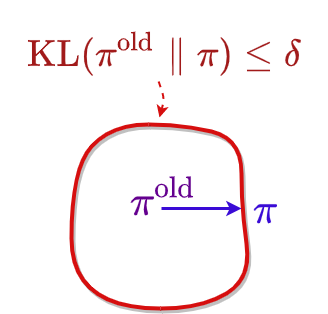}
\vspace{-0.9cm}
\end{wrapfigure}

Основная идея TRPO заключается в переходе от оптимизации \eqref{lowerboundmean} без ограничений к задаче оптимизации с ограничением в trust region форме:
\begin{equation}\label{trustregion}
\begin{cases}
L_{\textcolor{ChadPurple}{\pi^{\old}}}(\textcolor{ChadBlue}{\theta}) \to \max\limits_{\textcolor{ChadBlue}{\theta}} \\
\KL(\textcolor{ChadPurple}{\pi^{\old}} \parallel \textcolor{ChadBlue}{\pi_\theta}) \le \delta
\end{cases}
\end{equation}

То есть, суррогатная функция является локальной аппроксимацией нашего функционала, поэтому на каждом шаге работы алгоритма мы будем работать с ней. При этом второе слагаемое оптимизируемой нижней оценки \eqref{lowerboundmean} подсказывает, что с ростом KL-дивергенции <<нечестный>> функционал всё меньше похож на настоящий, и к нему <<всё меньше доверия>>. Условная задача оптимизации говорит, что оптимизировать суррогатную функцию вместо настоящего функционала можно, но с жёстким ограничением на длину шага.

Вообще говоря, мы получили задачу для оптимизации $L_{\textcolor{ChadPurple}{\pi^{\old}}}(\textcolor{ChadBlue}{\theta})$ методом \emph{натуральных градиентов} (natural gradient): мы оптимизируем функцию, разрешая не шаг некоторой длины по градиенту в пространстве параметров, а шаг некоторой длины в пространстве параметрически заданных распределений. Подробнее о натуральном градиенте можно прочитать в приложении \ref{appendix:ng}. Если раньше градиентный шаг мог сильно поменять стратегию (как распределение в пространстве действий), при том что для других небольших изменений распределения было бы необходимо сильно менять параметры $\textcolor{ChadBlue}{\theta}$, то здесь $\delta$ ограничивает изменение самой стратегии в терминах $\KL$-дивергенции. Ограничение $\delta$ нам при этом всё равно нужно будет выбрать, это аналог learning rate в <<trust region формах>> методов оптимизации.

Как будем задачу \eqref{trustregion} решать? В контексте нашей задачи мы собрали при помощи текущей стратегии некоторое количество данных для Монте-Карло оценок всех мат.ожиданий (в этот момент $\pi^{\old} \HM= \pi_\theta$) и хотим решить задачу \eqref{trustregion}, зафиксировав $\textcolor{ChadPurple}{\pi^{\old}}$ и оптимизируя параметры $\textcolor{ChadBlue}{\theta}$. Обозначим параметры $\textcolor{ChadPurple}{\pi^{\old}}$ как $\textcolor{ChadPurple}{\theta^{\old}}$. Аппроксимируем оптимизируемый функционал $L_{\textcolor{ChadPurple}{\pi^{\old}}}(\textcolor{ChadBlue}{\theta})$ разложением Тейлора до первого порядка с центром в точке $\theta = \theta^{\old}$, а ограничение --- до второго. До второго --- потому что слагаемое первого порядка ноль.

\begin{proposition}
В точке $\theta = \theta^{\old}$ первый член разложения ограничения в ряд Тейлора равен нулю:
\begin{equation*}
\forall s \colon \left. \textcolor{ChadBlue}{\nabla_\theta} \KL(\textcolor{ChadPurple}{\pi^{\old}} \parallel \textcolor{ChadBlue}{\pi_\theta}) \right|_{\theta = \theta^{\old}} = 0
\end{equation*}
\begin{proof}
$\KL$-дивергенция в этой точке равна 0 как среднее по состояниям дивергенций между одинаковыми распределениями, следовательно как функция от $\textcolor{ChadBlue}{\theta}$ она достигает в этой точке глобального минимума $\Rightarrow$ градиент равен нулю.
\end{proof}
\end{proposition}

Итак, введём обозначения для предложенного разложения. Пусть $g$ --- градиент $L_{\textcolor{ChadPurple}{\pi^{\old}}}(\textcolor{ChadBlue}{\theta})$ по параметрам $\textcolor{ChadBlue}{\theta}$ в точке $\theta = \theta^{\old}$, а $F$ --- гессиан ограничения в точке $\theta = \theta^{\old}$:
\begin{equation}\label{trpo_gradient}
    g := \left. \textcolor{ChadBlue}{\nabla_\theta} L_{\textcolor{ChadPurple}{\pi^{\old}}}(\textcolor{ChadBlue}{\theta}) \right|_{\theta = \theta^{\old}}
\end{equation}
\begin{equation}\label{trpo_hessian}
    F := \left. \textcolor{ChadBlue}{\nabla^2_\theta} \KL(\textcolor{ChadPurple}{\pi^{\old}} \parallel \textcolor{ChadBlue}{\pi_\theta}) \right|_{\theta = \theta^{\old}}
\end{equation}

В этих обозначениях аппроксимация задачи получается следующая:
\begin{equation}\label{approximatetrustregion}
\begin{cases}
\langle g , \textcolor{ChadBlue}{\theta} - \textcolor{ChadPurple}{\theta^{\old}} \rangle \to \max\limits_{\textcolor{ChadBlue}{\theta}} \\
\frac{1}{2} (\textcolor{ChadBlue}{\theta} - \textcolor{ChadPurple}{\theta^{\old}})^T F (\textcolor{ChadBlue}{\theta} - \textcolor{ChadPurple}{\theta^{\old}}) \le \delta
\end{cases}
\end{equation}

Градиент $g$, на самом деле, весьма любопытен:
\begin{proposition}
Градиент $g$ совпадает с градиентом из алгоритма Advantage Actor Critic \eqref{advantagepg}.
\begin{proof}
\begin{align*}g = \left. \textcolor{ChadBlue}{\nabla_\theta} L_{\textcolor{ChadPurple}{\pi^{\old}}}(\textcolor{ChadBlue}{\theta}) \right|_{\theta = \theta^{\old}} &= \frac{1}{1 - \gamma}\E_{s \sim \textcolor{ChadPurple}{d_{\pi^{\old}}}(s)} \E_{a \sim \textcolor{ChadPurple}{\pi^{\old}}(a \mid s)} \frac{\left. \textcolor{ChadBlue}{\nabla_\theta} \textcolor{ChadBlue}{\pi_\theta}(a \mid s) \right|_{\theta = \theta^{\old}}}{\textcolor{ChadPurple}{\pi^{\old}}(a \mid s)} \textcolor{ChadPurple}{A^{\pi^{\old}}}(s, a) = \\
= \{ \substack{\text{замечаем определение} \\ \text{производной логарифма} }\} &= \frac{1}{1 - \gamma}\E_{s \sim \textcolor{ChadPurple}{d_{\pi^{\old}}}(s)} \E_{a \sim \textcolor{ChadPurple}{\pi^{\old}}(a \mid s)} \left. \textcolor{ChadBlue}{\nabla_\theta} \log \textcolor{ChadBlue}{\pi_\theta}(a \mid s) \right|_{\theta = \theta^{\old}} \textcolor{ChadPurple}{A^{\pi^{\old}}}(s, a)
\end{align*}
что в точности есть градиент обычного ActorCritic c бэйзлайном в точке $\theta = \theta^{\old}$.
\end{proof}
\end{proposition}

\begin{theorem}
Решение аппроксимированной задачи \eqref{approximatetrustregion} есть
$$\theta - \theta^{\old} = kF^{-1}g$$
где скалярный коэффициент пропорциональности $k$ можно посчитать по формуле $k = \sqrt{\frac{2 \delta}{g^TF^{-1}g}}$
\begin{proof}
Составляем лагранжиан (оптимизируемый функционал входит с минусом, т.к. максимум поменяем на минимум):
$$\mathcal{L}(\lambda, \theta) = -g^T(\theta - \theta^{\old}) + \lambda \left( \frac{1}{2} (\theta - \theta^{\old})^T F (\theta - \theta^{\old}) - \delta \right)$$
Дифференцируем лагранжиан и приравниваем к нулю:
$$\nabla_\theta \mathcal{L}(\lambda, \theta) = -g + \lambda F (\theta - \theta^{\old}) = 0$$
Отсюда получаем:
\begin{equation}\label{solvetrustregion}
\theta - \theta^{\old} = \frac{1}{\lambda} F^{-1}g
\end{equation}
Осталось найти значение $\lambda$. Поскольку решение должно быть допустимой точкой, а оптимизируемый функционал линеен, понятно, что ограничение из задачи превратится в равенство и будет достигнуто на границе. То есть:
$$\frac{1}{2} (\theta - \theta^{\old})^T F (\theta - \theta^{\old}) = \delta$$
Подставляем найденное решение \eqref{solvetrustregion}:
$$\frac{1}{2} \left( \frac{1}{\lambda} F^{-1}g \right)^T F \left( \frac{1}{\lambda} F^{-1}g \right) = \frac{1}{2 \lambda^2} g^TF^{-1}g = \delta$$
Отсюда находим коэффициент\footnote[*]{однозначно в силу положительности коэф. Лагранжа; число $g^TF^{-1}g$ положительно в силу того, что гессиан KL-дивергенции является матрицей Фишера и поэтому является положительно определённой матрицей (см. приложение \ref{appendix:fishermatrix}). Естественно, усреднение по состояниям не нарушает этого факта.}:
$$\lambda = \sqrt{\frac{g^TF^{-1}g}{2 \delta}}$$
Обратная дробь $\frac{1}{\lambda}$, соответственно, является коэффициентом пропорциональности.
\end{proof}
\end{theorem}

Итак, что мы делаем на практике. Во-первых, собираем большой-большой роллаут (порядка 1024 переходов суммарно по средам), чтобы возможно было хоть сколько-то адекватно оценивать гессиан $F$ \eqref{trpo_hessian}. Оцениваем Advantage собранных пар $s, a$ при помощи GAE \eqref{truncatedGAE}. Затем считаем градиент $g$ \eqref{trpo_gradient} аналогично Actor-Critic методу. Далее мы хотим решить систему линейных уравнений:
$$F \left(\theta - \theta^{\old}\right) = g$$
Хранить в памяти гессиан и тем более обращать для нейросеток мы, конечно, не будем и воспользуемся \emph{методом сопряжённых градиентов} (conjugate gradient method), который позволяет решать систему линейных уравнений итеративно, на каждой итерации требуя лишь вычислять $Fh$ для некоторых векторов $h$ (прочитать про него можно, например, \href{https://ru.wikipedia.org/wiki/Метод_сопряжённых_градиентов_(для_решения_СЛАУ)}{в википедии}). Нам придётся для каждой итерации метода сопряжённых градиентов сделать два обратных прохода по вычислительному графу (дважды вычислять все производные: сначала по функции, для которой мы хотим посчитать гессиан, чтобы получить градиент $f \HM \coloneqq \left. \textcolor{ChadBlue}{\nabla_\theta} \KL(\textcolor{ChadPurple}{\pi^{\old}} \parallel \textcolor{ChadBlue}{\pi_\theta}) \right|_{\theta = \theta^{\old}}$, затем для градиента функции $\langle f, h \rangle$), но до сходимости сводить алгоритм не будем и остановим после порядка 10 итераций. Процедура получается дороговатой всё равно, но поскольку в деле замешан гессиан, то что поделать --- мы практически выходим в методы оптимизации второго порядка.

Посчитали приближение $F^{-1}g$, дальше возникает проблема: нам нужно домножить вектор на коэффициент $k$, который напрямую зависит от $\delta$. Мы могли бы поставить $\delta$ гиперпараметром как learning rate в обычном градиентном спуске, но всё-таки мы боремся за то, чтобы делать шаги <<правильного>> размера. Мы уже знаем направление $F^{-1}g$, в котором будем менять параметры политики, но не знаем, насколько. И вот тут становится важно, что когда мы приблизили локально $J(\textcolor{ChadBlue}{\theta})$ суррогатной функцией, мы для суррогатной функции можем посчитать значение в любой точке.

Когда мы делаем шаг градиентного спуска, мы рискуем сделать слишком длинный шаг, даже если он делается в верном направлении. В Policy Gradient методах это особенно критично: если сделать неудачное обновление политики, и $\pi$ сломается, то на следующем шаге собранные данные будут ужасными, поскольку они собираются on-policy, при помощи стратегии $\pi$! Поэтому важно не сломать стратегию ни на одном шаге, и для этого learning rate приходится выбирать маленьким. Естественно, это приводит к тому, что модель будет обучаться очень долго (а у нас тут сэмплы на каждый шаг сжигаются!). При этом, как это обычно бывает при обучении нейронных сетей, запускать какую-нибудь процедуру автоматического подбора learning rate, дороговато --- для этого нужно уметь вычислять значение оптимизируемой функции в разных точках. Тем более в случае с RL-ем, где, чтобы оценить $J(\textcolor{ChadBlue}{\theta})$ и проверить, не сломалось ли чего, нужно отправлять на каждом шаге несколько раз какие-то стратегии в среду и играть целые эпизоды, это совершенно не вариант.

Но здесь, в TRPO, мы применяли аппроксимацию дважды: сначала приблизили функционал на суррогатную функцию, а затем суррогатную функцию приблизили линейной моделью. Можно было бы сказать, что мы просто исходный функционал приблизили линейно (получился бы тот же самый градиент $g$, как мы разобрали), но для промежуточной задачи \eqref{trustregion}, в которой мы работаем с суррогатной функцией, значение которой мы можем посчитать в любой точке, мы на самом деле можем проводить \emph{бэктрэкинг} для адаптивного подсчёта $\delta$ --- аналога learning rate в trust region подходе. Его можно проводить немного по-разному, рассмотрим конкретный вариант из стандартных реализаций TRPO.

Выставляем какое-нибудь большое значение $\delta$ (это начальное значение --- гиперпараметр) и считаем коэффициент пропорциональности $k = \sqrt{\frac{2 \delta}{g^TF^{-1}g}}$ (заметим, что $F^{-1}g$ приближённо мы уже посчитали в методе сопряжённых градиентов). Дальше, во-первых, проверяется, а правда ли мы остались внутри trust region-а, то есть соблюли ли условие $\KL(\textcolor{ChadPurple}{\pi^{\old}} \parallel \textcolor{ChadBlue}{\pi_\theta}) \HM\le \delta$. Мы могли нарушить это условие как из-за перехода к приближённой задаче \eqref{approximatetrustregion}, так и из-за приближённого её решения через метод сопряжённых градиентов; если таки нарушили, то уменьшаем $\delta$ в, допустим, два раза и перепроверяем. Если же условие соблюдено, то проверяем, что значение $L_{\textcolor{ChadPurple}{\pi^{\old}}}(\textcolor{ChadBlue}{\theta})$, оценённое по имеющимся данным, больше нуля (что хотя бы для эмпирических данных удалось увеличить значение суррогатной функции): если нет, то продолжаем уменьшать регион доверия, а если да, то мы можем заканчивать процедуру бэктрэкинга и делать таким образом <<максимально длинный>> шаг.

После такой процедуры мы уже шагнули на границу нашего региона доверия, максимально отступив от $\textcolor{ChadPurple}{\pi^{\old}}$, использовавшихся для сбора данных. Соответственно, смысла <<продолжать>> оптимизировать нижнюю оценку дальше, разложив функционал в $\textcolor{ChadBlue}{\theta} \ne \textcolor{ChadPurple}{\theta^{\old}}$, нет; нужно перестраивать нижнюю оценку, то есть собирать новые данные при помощи обновлённой стратегии. Можно сказать, что переиспользовать данные мы не научились, но мы научились делать тяжёлые и дорогие, но хорошие шаги оптимизации. 

\vspace{0.4cm}
\begin{center}
\includegraphics[width=0.9\textwidth]{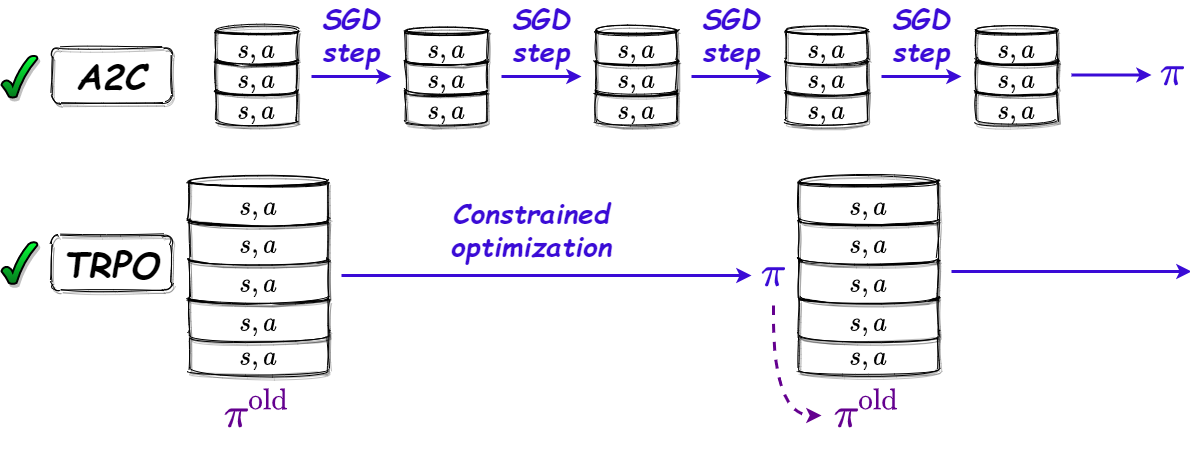}
\end{center}

\begin{remark}
В текущем виде в TRPO критика всё ещё нужно оптимизировать также, как в A2C, то есть собирая маленькие роллауты и делая небольшие шаги. Не очень удобно делать это параллельно со сбором больших роллаутов для актёра, да и оптимизировать актёра и критика в этой схеме, получается, придётся раздельно (иметь для них две раздельные сетки). Поскольку роллауты собирают большие, в имплементациях на критика иногда вообще забивают и используют Монте-Карло оценку как в REINFORCE, просто доигрывая игры до конца (если игры относительно короткие, по 20-100 шагов, то это может быть разумно: всё равно нам длинные роллауты нужны шагов по 100). Конечно, если всё-таки обучать критика, то можно существенно выиграть от использования GAE-оценок.
\end{remark}

\subsection{Proximal Policy Loss}

Считается, что TRPO практически всегда работает лучше A2C, с единственным основным недостатком --- сложностью самого алгоритма. Всё-таки в глубоком обучении привычнее работать с какими-то датасетами, из которых сэмплируются батчи, вычисляется какая-то функция потерь или оптимизируемый функционал, считается градиент и отправляется в условный Adam. Настраивать такой алгоритм тяжело и неприятно.

Proximal Policy Optimization (PPO) --- альтернативный способ рассуждения после вывода нижней оценки \eqref{lowerboundmean}. Давайте не будем прибегать к методам оптимизации, требующим какие-либо гессианы, и будем оптимизировать нижнюю оценку напрямую обычным градиентным спуском, где $C$ --- гиперпараметр:

\begin{equation}\label{lowerboundunconstrained}
\E_{s \sim \textcolor{ChadPurple}{d_{\pi^{\old}}}(s)} \E_{a \sim \textcolor{ChadPurple}{\pi^{\old}}(a \mid s)} \left[ \frac{\textcolor{ChadBlue}{\pi_\theta}(a \mid s)}{\textcolor{ChadPurple}{\pi^{\old}}(a \mid s)} \textcolor{ChadPurple}{A^{\pi^{\old}}}(s, a) - C \sqrt{\KL(\textcolor{ChadPurple}{\pi^{\old}} \parallel \textcolor{ChadBlue}{\pi_\theta})} \right] \to \max_{\textcolor{ChadBlue}{\theta}}
\end{equation}

Сделать так напрямую в лоб не получится; давайте поймём, почему. Какого размера датасет нам понадобится для такого пайплайна? Собрать данных на мини-батч или несколько и <<воспользоваться ими несколько раз>> не получится: Монте-Карло оценки мат.ожиданий в наших функциях потерь просто будут скоррелированы. Чтобы по данным из $\textcolor{ChadPurple}{\pi^{\old}}$ прооптимизировать $\textcolor{ChadBlue}{\theta}$ несколькими шагами градиентного спуска, нужно, чтобы мини-батчи были достаточно разными. Это значит, что при помощи $\textcolor{ChadPurple}{\pi^{\old}}$ придётся собрать датасет достаточно большого размера.

\vspace{0.4cm}
\begin{center}
\includegraphics[width=0.9\textwidth]{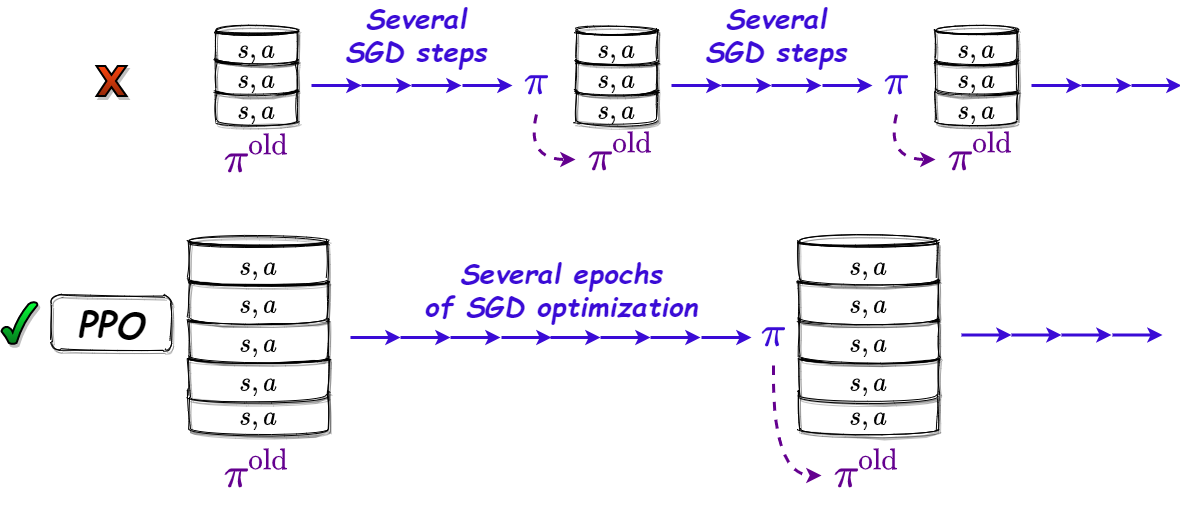}
\end{center}

\begin{wrapfigure}{r}{0.3\textwidth}
\vspace{-0.5cm}
\centering
\includegraphics[width=0.3\textwidth]{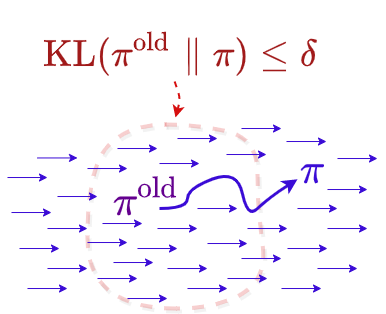}
\vspace{-0.5cm}
\end{wrapfigure}

Чтобы что-то выиграть от такого процесса, нужно пройтись по датасету несколькими эпохами (иначе тот же A2C был бы выгоднее за счёт свежести данных). Это значит, что шагов градиентной оптимизации понадобится сделать достаточно много. Следовательно, стратегия потенциально обновляется за время оптимизации на одном датасете относительно сильно: вместе с тем, что небольшое изменение в пространстве параметров может приводить к большому изменению стратегии $\pi$ (в пространстве распределений) без введения жёсткого региона доверия стратегия может начать сколь угодно сильно подстраиваться под те оценки критика, которые мы выдадим парам в датасете.

Действительно, на практике наш критик $\textcolor{ChadPurple}{A^{\pi^{\old}}}$ неидеален и его оценка $\Psi(s, a) \HM\approx \textcolor{ChadPurple}{A^{\pi^{\old}}}(s, a)$ будет смещённой. Но важно, что в оценке $\Psi$ будут заложены сэмплы из собранных при помощи $\textcolor{ChadPurple}{\pi^{\old}}$ данных: награды и состояния за будущие шаги. Для каждой пары $s, a$ мы увидим лишь по одному сэмплу будущего, и как бы мы оценку $\Psi(s, a)$ ни строили, эти сэмплы $s', a', \dots$ у нас ровно в одном экземпляре. Мы не хотим, ест-но, несколько раз $\textcolor{ChadPurple}{\pi^{\old}}$ в среду отправлять --- это бессмысленно, в on-policy режиме всегда выгоднее если и отправлять в среду, то самую свежую стратегию, переходя таким образом к очередному шагу алгоритма. А значит, что при фиксированных данных при оптимизации суррогатной функции мы не жадный policy improvement стратегии $\textcolor{ChadPurple}{\pi^{\old}}$ проведём, а полную ерунду: будем искать $\argmax\limits_{a} \Psi(s, a)$, то есть перекладывать всю вероятность в те действия, где оценка (!) Advantage случилась положительная, и убирать вероятность оттуда, где оценка Advantage случилась отрицательная. Этот эффект можно охарактеризовать как <<переобучение под сэмплы>>.

\begin{example}
Утрированный пример. Текущая стратегия сделала два шага в среде, собрала $s_1, a_1, s_2, a_2$. Оценки Advantage получились следующими: $\Psi(s_1, a_1) \HM= 1$, $\Psi(s_2, a_2) \HM= -1$. Теперь если мы по сэмплам оценили суррогатную функцию \eqref{surrogate} и начали её оптимизировать по параметрам стратегии, то получится:
$$\frac{\textcolor{ChadBlue}{\pi_\theta}(a_1 \mid s_1)}{\textcolor{ChadPurple}{\pi^{\old}}(a_1 \mid s_1)} (+1) + \frac{\textcolor{ChadBlue}{\pi_\theta}(a_2 \mid s_2)}{\textcolor{ChadPurple}{\pi^{\old}}(a_2 \mid s_2)} (-1) \to \max_{\textcolor{ChadBlue}{\theta}}$$

Знаменатели дробей можно считать какими-то положительными числами, поэтому в итоге $\textcolor{ChadBlue}{\pi_\theta}(a_2 \mid s_2)$ полетит в ноль, $\textcolor{ChadBlue}{\pi_\theta}(a_1 \mid s_1)$ --- или в единицу, если пространство действий дискретно, или вообще в бесконечность, если пространство действий непрерывно.
\end{example}

То есть на тех парах $s, a$, где оценка Advantage положительна, стратегия начнёт улетать к вырожденной, а вероятности на парах $s, a$ с отрицательной оценкой Advantage начнут уплывать в ноль. В итоге начинают регулярно встречаться взрывающиеся или затухающие importance sampling коэффициенты $\frac{\textcolor{ChadBlue}{\pi_\theta}(a \mid s)}{\textcolor{ChadPurple}{\pi^{\old}}(a \mid s)}$, мини-батч становится несбалансированным и в плане <<весов>> объектов. И регуляризатор в виде KL-дивергенции из нижней оценки, поставленный в виде жёсткого ограничения (<<trust region>>-а) в задачу оптимизации \eqref{trustregion}, защищал нас от этого эффекта в TRPO.

Предлагается полечить костылём: обрезать. Обозначим importance sampling коэффициент как
$$\rho(\textcolor{ChadBlue}{\theta}) \coloneqq \frac{\textcolor{ChadBlue}{\pi_\theta}(a \mid s)}{\textcolor{ChadPurple}{\pi^{\old}}(a \mid s)}$$
и обрежем его как
$$\rho^{\clip}(\textcolor{ChadBlue}{\theta}) \coloneqq \clip(\rho(\textcolor{ChadBlue}{\theta}), 1 - \epsilon, 1 + \epsilon),$$
где $\epsilon \in (0, 1)$ --- гиперпараметр (типичный выбор --- 0.1 или 0.2). Рассмотрим альтернативную функцию потерь:
\begin{equation*}
\E_{s \sim \textcolor{ChadPurple}{d_{\pi^{\old}}}(s)} \E_{a \sim \textcolor{ChadPurple}{\pi^{\old}}(a \mid s)} \left[ \rho^{\clip}(\textcolor{ChadBlue}{\theta}) \textcolor{ChadPurple}{A^{\pi^{\old}}}(s, a) - C \sqrt{\KL(\textcolor{ChadPurple}{\pi^{\old}} \parallel \textcolor{ChadBlue}{\pi_\theta})} \right] \to \max_{\textcolor{ChadBlue}{\theta}}
\end{equation*}

\begin{wrapfigure}{r}{0.3\textwidth}
\vspace{-0.5cm}
\centering
\includegraphics[width=0.3\textwidth]{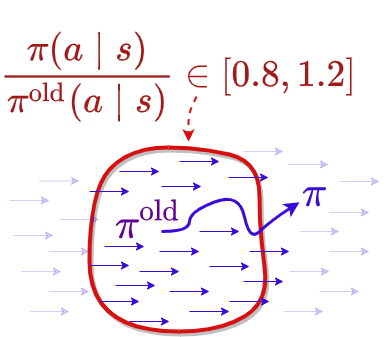}
\vspace{-0.5cm}
\end{wrapfigure}
Что случилось с градиентами при таком изменении? Если происходит обрезка, то есть если $\rho(\textcolor{ChadBlue}{\theta})$ не попадает в указанный диапазон $[1 - \epsilon, 1 + \epsilon]$, то градиент основного слагаемого, как легко видеть, зануляется. Иначе он остаётся без изменений:
$$\textcolor{ChadBlue}{\nabla_\theta} \rho^{\clip}(\textcolor{ChadBlue}{\theta}) = 
\begin{cases}
0 \quad &\rho(\textcolor{ChadBlue}{\theta}) \not\in [1 - \epsilon, 1 + \epsilon] \\
\textcolor{ChadBlue}{\nabla_\theta} \rho(\textcolor{ChadBlue}{\theta}) \quad &\rho(\textcolor{ChadBlue}{\theta}) \in [1 - \epsilon, 1 + \epsilon] \\
\end{cases}
$$
Таким образом, подобный клиппинг --- <<мягкий trust region>>: как только стратегия слишком отдаляется от $\textcolor{ChadPurple}{\pi^{\old}}$ на данной паре $s, a$, градиенты <<перестают>> обновлять эту пару. Это не означает, что стратегия в этой паре не продолжит меняться: с ней потенциально в градиентном спуске может происходить всё что угодно, она может продолжать изменяться за счёт <<схожих>> пар $s, a$, например, которые ещё остаются <<внутри trust region-а>> или, в конце концов, из-за моментума в алгоритме стохастической оптимизации\footnote{в RL, как и в глубоком обучении, обычный стохастический градиентный спуск не используется, а вместо этого используется, например, Adam.}.

Замена функции потерь лишила нас в очередной раз <<гарантий нижней оценки>>: хотелось бы, чтобы при достаточно большой константе $C$ эти гарантии оставались. Поэтому мы не просто заменим функцию потерь на версию с обрезкой, а возьмём минимум между ними: так мы сохраним свойство нижней оценки и гарантируем, что функционал \eqref{lowerboundunconstrained} не увеличился: 
\begin{equation}\label{PPOlowerbound}
\E_{s \sim \textcolor{ChadPurple}{d_{\pi^{\old}}}(s)} \E_{a \sim \textcolor{ChadPurple}{\pi^{\old}}(a \mid s)}  \left[ \min \left( \rho(\textcolor{ChadBlue}{\theta}) \textcolor{ChadPurple}{A^{\pi^{\old}}}(s, a), \rho^{\clip}(\textcolor{ChadBlue}{\theta}) \textcolor{ChadPurple}{A^{\pi^{\old}}}(s, a)\right) - C \sqrt{\KL(\textcolor{ChadPurple}{\pi^{\old}} \parallel \textcolor{ChadBlue}{\pi_\theta})} \right] \to \max_\theta
\end{equation}

Интуиция нижней оценки здесь, на самом деле, под очень большим вопросом. Авторы алгоритма на этом этапе в ablation studies внезапно обнаружили, что на слагаемое с $\KL$-дивергенцией можно внезапно забить (выставить $C \HM= 0$), и эмпирические результаты не изменятся. Обычно в имплементациях PPO $\KL$-дивергенции в функционале по дефолту нет\footnote{если же его всё-таки оставляют, то обычно без квадратного корня.}, и итого оптимизируемый функционал выглядит просто вот так:
\begin{equation}\label{PPOobjective}
\E_{s \sim \textcolor{ChadPurple}{d_{\pi^{\old}}}(s)} \E_{a \sim \textcolor{ChadPurple}{\pi^{\old}}(a \mid s)} \min \left( \rho(\textcolor{ChadBlue}{\theta}) \textcolor{ChadPurple}{A^{\pi^{\old}}}(s, a), \rho^{\clip}(\textcolor{ChadBlue}{\theta}) \textcolor{ChadPurple}{A^{\pi^{\old}}}(s, a)\right) \to \max_\theta
\end{equation}

На этом месте гробик связи между алгоритмом и теорией нижней оценки засыпается землёй. Но что же произошло? Мы ввели обрезку (имеющую некоторый смысл trust region-а) и взятие минимума между двумя функциями потерь; если обрезка так хорошо <<смоделировала>> trust region, что регуляризатор (слагаемое с $\KL$-дивергенцией) и не нужен, то какой физический смысл имеет минимум? 

Итак, давайте поймём, какую роль в интуиции trust region-а играет взятие минимума, посмотрев на градиенты для одной пары $s, a$. Введение минимума всё ещё <<выкидывает>> из градиентов функционала пары $s, a$, на которых коэффициент $r(\theta)$ не близок к 1; или же оставляет градиент без изменений. Зануление градиента происходит в случае, если происходит сразу два события: минимум достигается на <<обрезанной>> версии градиента, и importance sampling вес вышел за границы. Рассмотрим всевозможные случаи:

$$\textcolor{ChadBlue}{\nabla_\theta} \min \left( \rho(\textcolor{ChadBlue}{\theta}) \textcolor{ChadPurple}{A^{\pi^{\old}}}(s, a), \rho^{\clip}(\textcolor{ChadBlue}{\theta}) \textcolor{ChadPurple}{A^{\pi^{\old}}}(s, a)\right) = 
\begin{cases}
0 \qquad & \rho(\textcolor{ChadBlue}{\theta}) > 1 + \epsilon, \quad \textcolor{ChadPurple}{A^{\pi^{\old}}}(s, a) > 0 \\
\textcolor{ChadBlue}{\nabla_\theta} \rho(\textcolor{ChadBlue}{\theta}) \quad &\rho(\textcolor{ChadBlue}{\theta}) < 1 + \epsilon, \quad \textcolor{ChadPurple}{A^{\pi^{\old}}}(s, a) > 0 \\
0 \qquad & \rho(\textcolor{ChadBlue}{\theta}) < 1 - \epsilon, \quad \textcolor{ChadPurple}{A^{\pi^{\old}}}(s, a) < 0 \\
\textcolor{ChadBlue}{\nabla_\theta} \rho(\textcolor{ChadBlue}{\theta}) \quad &\rho(\textcolor{ChadBlue}{\theta}) > 1 - \epsilon, \quad \textcolor{ChadPurple}{A^{\pi^{\old}}}(s, a) < 0 \\
\end{cases}
$$

\needspace{13\baselineskip}
\begin{wrapfigure}{r}{0.3\textwidth}
\vspace{-0.4cm}
\centering
\includegraphics[width=0.3\textwidth]{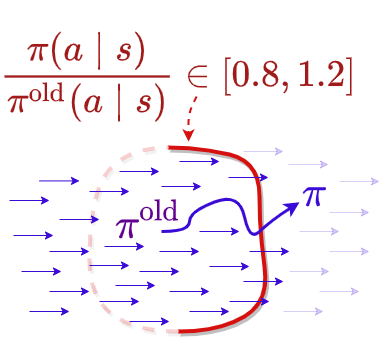}
\vspace{-0.4cm}
\end{wrapfigure}
Внимательно вглядевшись в эту <<таблицу>>, становится понятно, что происходит. Если оценка критика положительна, градиенты говорят увеличивать $\textcolor{ChadBlue}{\pi_\theta}(a \mid s)$; importance sampling вес повышается, и в какой-то момент случится обрезка по $1 + \epsilon$. Но если оценка критика положительна, и градиенты указывают увеличивать вероятность, а она по какой-то причине уменьшилась (такое вполне возможно в процессе оптимизации) и даже <<вылетела за границы trust region>>, то градиенты не зануляются: она вылетела <<не с той стороны>>! Симметричная ситуация случится при отрицательной оценке критика: вероятность должна уменьшаться, но сильно за счёт обрезки уменьшиться она не может, а за счёт оператора минимума при случайном увеличении градиенты продолжат тянуть вероятности <<к барьеру>>. Мы получили этакий <<полуоткрытый trust region>>.

\subsection{Clipped Value Loss}

\begin{wrapfigure}{r}{0.4\textwidth}
\vspace{-0.5cm}
\centering
\includegraphics[width=0.4\textwidth]{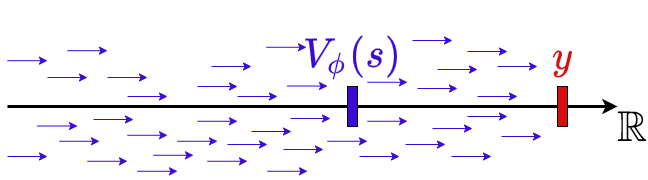}
\end{wrapfigure}
Применим аналогичную идею <<полуоткрытого trust region-а>> для лосса критика. Пусть для данного состояния $s$ наша аппроксимация V-функции выдаёт $V_{\phi}(s)$, а таргет равен $y$. Обычно мы бы минимизировали MSE:
$$\Loss_1(\phi) \coloneqq (V_{\phi}(s) - y)^2$$

Однако если таргет $y$ содержит в себе в том числе какие-то Монте-Карло оценки и переиспользуется <<несколько раз>> из небольшого датасета, то мы не хотим на него переобучаться. Пусть на момент сбора датасета критик выдавал для данного состояния $V^{\old}(s)$; таргет указывает лишь направление, в котором мы хотим изменить значение выхода критика, но, как и в любой стохастической аппроксимации, мы не хотим <<заменять жёстко>> выход $V_{\phi}(s)$ на $y$, а лишь сдвинуть в его сторону. Для этого добавим и вычтем в MSE $V^{\old}(s)$:
$$\Loss_1(\phi) = (V_{\phi}(s) - V^{\old}(s) - (y - V^{\old}(s)))^2$$

\begin{wrapfigure}{r}{0.4\textwidth}
\centering
\includegraphics[width=0.4\textwidth]{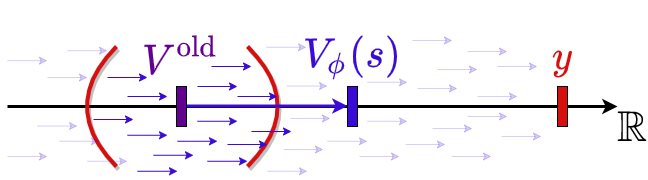}
\end{wrapfigure}
Теперь мы сравниваем не выход сети с таргетом, а изменение значения выхода критика с <<желаемым изменением>> (на самом деле, просто оценкой Advantage; $y \HM- V^{\old}(s)$ используется в качестве аппроксимации $\Psi(s, a)$). Введём другую функцию потерь, с <<обрезкой>>, которая построит <<мягкий trust-region>> и будет занулять градиенты, как только $V_{\phi}(s)$ станет непохожим на $V^{\old}(s)$:
$$\Loss_2(\phi) \coloneqq (\clip(V_{\phi}(s) - V^{\old}(s), \epsilon, -\epsilon) - (y - V^{\old}(s)))^2,$$
где $\epsilon$ --- гиперпараметр. Аналогично лоссу актёра, градиенты в такой функции потерь просто будут зануляться в ситуациях, когда происходит обрезка.

\begin{wrapfigure}{r}{0.4\textwidth}
\vspace{-0.7cm}
\centering
\includegraphics[width=0.4\textwidth]{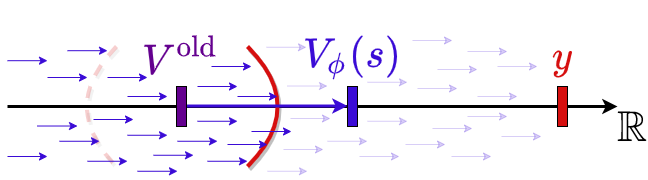}
\end{wrapfigure}
Наконец, чтобы <<открыть>> trust-region с одной стороны, нужно просто, по аналогии с лоссом актёра, взять максимум от этих двух функций потерь (для актёра мы брали минимум, поскольку там велась максимизация, а здесь лосс минимизируется):
$$\Loss(\phi) \coloneqq \max(\Loss_1(\phi), \Loss_2(\phi))$$
Непосредственной проверкой легко убедиться, что градиенты будут зануляться, только если наш критик вышел из trust-region-а <<с правильной стороны>>; а так, градиенты будут тянуть нашего критика к допустимому барьеру.

\begin{remark}
Считается, что обрезка функционала для критика менее существенна, чем для актёра. Если мы не обрежем функционал для актёра, и тот <<сломается>>, переобучившись под сэмплы, это может сломать весь оптимизационный процесс, поскольку стратегия используется для сбора данных. Но если немного сломался критик, то это не так сильно отразится на процессе обучения, потому что актёр не опирается целиком на оценки критика. Тем не менее, раз такая возможность <<защищиться>> от переобучения под сэмплы в критике есть, то стоит ей пользоваться. Недостаток --- в необходимости подобрать соответствующее масштабу V-функции значение гиперпараметра обрезки $\epsilon$.
\end{remark}

\subsection{Proximal Policy Optimization (PPO)}

Итоговая процедура работы алгоритма следующая. Собираем большой роллаут (порядка хотя бы 1000 шагов). Версия стратегии, использовавшаяся для сбора, обозначается как $\textcolor{ChadPurple}{\pi^{\old}}$, и вероятности, с которыми она выбирала действия, сохраняются. Для всех пар $s, a$ высчитываются оценки Q-функции и Advantage методом GAE \eqref{truncatedGAE}. К собранным данным относимся как к датасету, из которого можно брать пары $s \HM\sim \textcolor{ChadPurple}{d_{\pi^{\old}}}(s), a \HM\sim \textcolor{ChadPurple}{\pi^{\old}}(a \mid s)$ и считать Монте-Карло оценку градиента \eqref{PPOlowerbound}. Размер батча при сэмплировании из датасета при этом обычный для градиентных методов первого порядка, условно такой же, как был бы в A2C. По датасету нужно пройтись при этом несколько раз, теоретически --- хочется как можно больше, но понятно, что чем больше расходится $\textcolor{ChadBlue}{\pi_\theta}$ и $\textcolor{ChadPurple}{\pi^{\old}}$, тем менее эффективна <<нижняя оценка>> и тем больше данных будет резать наш клиппинг. Соответственно, количество эпох --- сколько раз пройтись по датасету --- является ключевым гиперпараметром. Существенно, сколько именно шагов градиентного спуска будет сделано по одному датасету; это более важно, чем размер мини-батчей, поскольку именно первое больше влияет на то, когда мы начнём <<вываливаться>> из trust region-а, то есть когда отступим от стратегии сбора данных достаточно далеко.

\begin{algorithm}[label = PPOalgorithm]{Proximal Policy Optimization (PPO)}
\textbf{Гиперпараметры:} $M$ --- количество параллельных сред, $N$ --- длина роллаутов, $B$ --- размер мини-батчей, $\mathrm{n\_epochs}$ --- количество эпох, $\lambda$ --- параметр GAE-оценки, $\epsilon$ --- параметр обрезки для актёра, $\hat{\epsilon}$ --- параметр обрезки для критика, $V_\phi(s)$ --- нейросеть с параметрами $\phi$, $\pi_{\theta}(a \mid s)$ --- нейросеть для стратегии с параметрами $\theta$, $\alpha$ --- коэф. масштабирования лосса критика, SGD оптимизатор.

\vspace{0.3cm}
Инициализировать $\theta, \phi$ \\
\textbf{На каждом шаге:}
\begin{enumerate}
    \item в каждой параллельной среде собрать роллаут длины $N$, используя стратегию $\pi_{\theta}$, сохраняя вероятности выбора действий как $\pi^{\old}(a \mid s)$, а выход критика на встреченных состояниях как $V^{\old}(s) \leftarrow V_{\phi}(s)$
    \item для каждой пары $s, a$ из роллаутов посчитать одношаговую оценку Advantage:
    $$\Psi_{(1)}(s, a) \coloneqq r + \gamma (1 - \done') V_{\phi}(s') - V_{\phi}(s)$$
    \item посчитать GAE-оценку:
    $$\Psi_{\mathrm{GAE}}(s_{N-1}, a_{N-1}) \coloneqq \Psi_{(1)}(s_{N-1}, a_{N-1})$$
    \item для $t$ от $N - 2$ до 0:
    \begin{itemize}
    \item $\Psi_{\mathrm{GAE}}(s_t, a_t) \coloneqq \Psi_{(1)}(s_t, a_t) + \gamma \lambda (1 - \done_t) \Psi_{\mathrm{GAE}}(s_{t+1}, a_{t+1})$
    \end{itemize}
    \item посчитать таргет для критика:
    $$y(s) \coloneqq \Psi_{\mathrm{GAE}}(s, a) + V_{\phi}(s)$$
    \item составить датасет из шестёрок $(s, a, \Psi_{\mathrm{GAE}}(s, a), y(s), \pi^{\old}(a \mid s), V^{\old}(s))$
    \item выполнить $\mathrm{n\_epochs}$ проходов по роллауту, генерируя мини-батчи пятёрок $\T \coloneqq (s, a, \Psi_{\mathrm{GAE}}(s, a), y(s), \pi^{\old}(a \mid s), V^{\old}(s))$ размером $B$; \textbf{для каждого мини-батча:}
    \begin{itemize}
    \item вычислить лосс критика:
    $$\Loss_1(\T, \phi) \coloneqq \left( y(s) - V_\phi(s) \right) ^2$$
    $$\Loss_2(\T, \phi) \coloneqq \left( y(s) - V^{\old}(s) - \clip(V_\phi(s) - V^{\old}(s), -\hat{\epsilon}, \hat{\epsilon}) \right) ^2$$
    $$\Loss^{\critic}(\phi) \coloneqq \frac{1}{B}\sum_{\T} \max(\Loss_1(\T, \phi), \Loss_2(\T, \phi))$$
    \item сделать шаг градиентного спуска по $\phi$, используя $\nabla_\phi \Loss^{\critic}(\phi)$
    \item нормализовать $\Psi_{\mathrm{GAE}}(s, a)$ по батчу, чтобы в среднем значения равнялись 0, а дисперсия --- 1.
    \item посчитать коэффициенты в importance sampling:
    $$r_\theta(\T) \coloneqq \frac{\pi_\theta(a \mid s)}{\pi^{\old}(a \mid s)}$$
    \item посчитать обрезанную версию градиентов:
    $$r_\theta^{\clip}(\T) \coloneqq \clip(r_\theta(\T), 1 - \epsilon, 1 + \epsilon)$$
    \item вычислить градиент для актёра:
    $$\nabla^{\actor}_\theta \coloneqq \frac{1}{B}\sum_{\T} \nabla_\theta \min \left( r_\theta(\T)\Psi_{\mathrm{GAE}}(s, a), r_\theta^{\clip}(\T)\Psi_{\mathrm{GAE}}(s, a) \right) $$
    \item сделать шаг градиентного подъёма по $\theta$, используя $\nabla^{\actor}_\theta$
    \end{itemize}
\end{enumerate}
\end{algorithm}

\begin{remark}
В Atari играх хорошим значением гиперпараметра <<число эпох>> считается 3, а для задач непрерывного управления --- 10. Возможность несколько раз <<увидеть>> переход (пусть даже всего 3) вместо одного даёт существенный прирост эффективности алгоритма PPO по сравнению с A2C, поэтому на практике осмысленно использовать именно его, если нужен on-policy алгоритм.  
\end{remark}

\begin{remark}
PPO использовался для \href{https://openai.com/blog/openai-five/}{великих достижений в Dota} и считается более-менее устоявшейся SOTA в RL: если нет проблем с симулятором и можно использовать on-policy алгоритмы --- стоит начать с PPO, но причины, по которым он так круто работает, полностью не ясны. В частности, \href{https://openreview.net/forum?id=r1etN1rtPB}{последовавшие исследования} указывают на то, что клиппинг --- не основная причина успеха, а кроется она в целом наборе удачных инженерных хаков в официальной реализации алгоритма от OpenAI. Ключевая из них приведена в описании алгоритма --- нормализация $\Psi_{\mathrm{GAE}}(s, a)$ по батчу; мы знаем, что в среднем Advantage должен быть равен нулю, и поэтому можем его центрировать. Деление же на стандартное отклонение Advantage-оценок позволяет <<отнормировать>> масштаб функции награды внутри самого алгоритма.
\end{remark}

\chapter{Continuous control}\label{continuouscontrolchapter}

В этой главе будут рассмотрены алгоритмы для задачи непрерывного управления ($|\A| \subseteq [-1, 1]^A$), развивающие идеи Value-based и Policy gradient подходов. Мы увидим, что эти два подхода имеют много общего и отчасти даже эквивалентны, в обоих случаях получив схемы, моделирующие алгоритм Policy Iteration.

\section{DDPG}\label{DDPGsection}

\subsection{Вывод из Deep Q-learning}

Схема DQN \ref{DQNalgorithm} имела принципиальный недостаток: мы не могли работать с непрерывными пространствами действий в силу необходимости постоянно считать операторы максимума и аргмаксимума
$$\max_a Q_{\theta}(s, a)$$
как для жадного выбора действия, так и для построения таргета в задаче регрессии. 

\needspace{7\baselineskip}
\begin{wrapfigure}{r}{0.35\textwidth}
\vspace{-0.3cm}
\centering
\includegraphics[width=0.3\textwidth]{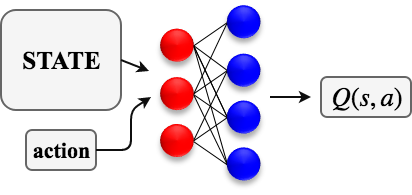}
\vspace{-0.3cm}
\end{wrapfigure}

Единственная возможная архитектура модели --- приём действий на вход вместе с состояниями, тогда поиск аргмаксимума проводить, в целом, можно, но дорого: инициализируем $a^0$ случайно, и устраиваем градиентный подъём по входу в модель:
$$a^{k+1} = a^k + \alpha \left. \nabla_a Q_{\theta}(s, a) \right|_{a = a^k}$$

Понятно, что такую процедуру устраивать по несколько раз за шаг дороговато. Однако, в глубинном обучении для таких проблем есть мега-универсальное решение: давайте задачу поиска аргмаксимума тоже аппроксимируем другой нейросетью! Максимум тогда можно будет считать, просто подставив в $Q_{\theta}$ вместо действия приближение этого аргмаксимума.

Итак, пусть $\pi_{\omega}(s)$ принимает на вход состояние $s$ и выдаёт аргмаксимум текущей аппроксимации Q-функции, то есть будем добиваться $\pi_{\omega}(s) \HM\approx \argmax\limits_a Q_{\theta}(s, a)$. Понятно, как обучать такую сеть:
$$Q_\theta(s, \pi_{\omega}(s)) \to \max_{\omega}$$

Весь алгоритм DQN оставляем неизменным с единственной модификацией, что на каждом батче также нужно сделать шаг оптимизации $\omega$. При этом каждый раз, когда в схеме необходимо считать максимум или аргмаксимум $Q_\theta$, используется $\pi_{\omega}(s)$.

В стандартном алгоритме DQN нам было необходимо считать $\max\limits_{a'} Q_{\theta^{-}}(s', a')$, и в дефолтной версии алгоритма использовалась таргет-сеть. Технически это означает, что для таргет-сети $Q_{\theta^{-}}(s', a')$ нам тоже нужно знать аргмаксимум, поэтому можно хранить старую версию вспомогательной функции $\pi_{\omega^{-}}(s)$. 

\begin{remark}
Считается, что это не так принципиально, поскольку использование <<свежей>> $\pi_{\omega}(s)$ при подсчёте таргета соответствует моделированию Double оценки (см. раздел \ref{subsec:doubledqn}). Все остальные модификации DQN также применимы.
\end{remark}

Итого мы получили, что для жадного выбора действия используется $\pi_{\omega}(s)$ (отсюда такое обозначение этой <<вспомогательной>> функции --- это фактически стратегия); а таргет для перехода $\T \coloneqq (s, a, r, s')$ вычисляется по формуле
$$y(\T) \coloneqq r + \gamma Q_{\theta^{-}}(s', \argmax_{a'} Q_{\theta^{-}}(s', a')) \approx r + \gamma Q_{\theta^{-}}(s', \pi_{\omega^{-}}(s))$$

Такой алгоритм называется Deep Deterministic Policy Gradient (DDPG), и название может сбить с толку: а причём здесь policy gradient?

\subsection{Вывод из Policy Gradient}


В Policy Gradient алгоритмах мы получили формулу градиента нашего функционала, <<релаксировав>> задачу и перейдя к оптимизации в пространстве стохастических стратегий. Если пространство действий непрерывно, то такая релаксация на самом деле не обязательна. Предположим\footnote{даже если это не так, в будущем мы всё равно будем приближать эти Q-функции нейросетями, для которых всегда справедлива дифференцируемость по входу.} дифференцируемость любых Q-функций $Q^{\pi}(s, a)$ по действиям $a$. Попробуем посчитать градиент по параметрам стратегии в случае детерминированной стратегии $\pi_\theta(s)$:

\begin{theoremBox}[label=th:dpg]{Deterministic Policy Gradient} В непрерывных пространствах действий в предположении дифференцируемости Q-функций по действиям:
\begin{equation}\label{dpg}
\nabla_\theta J(\pi_\theta) = \frac{1}{1 - \gamma}\E_{d_{\pi_\theta}(s)} \nabla_\theta \pi_\theta(s) \left. \nabla_a Q^{\pi}(s, a) \right|_{a = \pi_\theta(s)}
\end{equation}
\beginproof
$$\nabla_\theta V^{\pi_\theta}(s) = \{\text{VQ уравнение \eqref{VQ}}\} = \nabla_\theta \E_{a \sim \pi_\theta(s)} Q^{\pi_\theta}(s, a) = \nabla_\theta Q^{\pi_\theta}(s, \pi_\theta(s)) = (*)$$
Заметим, что в последнем выражении при малом изменении $\theta$ поменяется не только $\pi_\theta(s)$, но и сама оценочная функция $Q^{\pi_\theta}$. Считая, что якобиан функции $\R^n \to \R^m$ имеет размерность $n \times m$, и обозначая размерность действий как $A$, а размерность параметров $\theta$ буквой $d$, получаем следующие размерности градиентов:
$$\nabla_\theta Q^{\pi_\theta}(s, a) \in \R^{d \times 1} \quad \nabla_a Q^{\pi_\theta}(s, a) \in \R^{A \times 1} \quad \nabla_\theta \pi_\theta(s) \in \R^{d \times A}$$
Тогда продолжение вычисления выглядит так:
$$(*) = \nabla_\theta \pi_\theta(s) \left. \nabla_a Q^{\pi}(s, a) \right|_{a = \pi_\theta(s)} + \left. \nabla_\theta Q^{\pi_\theta}(s, a) \right|_{a = \pi_\theta(s)},$$
где последнее слагаемое --- якобиан $Q^{\pi_\theta}$ при фиксированном $a$ по параметрам стратегии $\pi_\theta$, которую он оценивает. Отдельно это слагаемое имеет вид:
$$\nabla_\theta Q^{\pi_\theta}(s, a) = \{\text{QV уравнение \eqref{QV}} \} = \gamma \E_{s'} \nabla_\theta V^{\pi_\theta}(s')$$

Получаем рекурсивную формулу, аккуратно собирая которую, получим:
$$\nabla_\theta V^{\pi_\theta}(s) = \E_{\Traj \sim \pi \mid s_0 = s} \sum_{t \ge 0} \gamma^t \nabla_\theta \pi_\theta(s_t) \left. \nabla_a Q^{\pi}(s_t, a) \right|_{a = \pi_\theta(s_t)}$$

Осталось только применить теорему \ref{th:decoupling_stoch} об эквивалентной форме мат.ожидания по траекториям для 
\begin{equation*}
f(s, a) = \nabla_\theta \pi_\theta(s) \left. \nabla_a Q^{\pi}(s, a) \right|_{a = \pi_\theta(s)}   \tagqed
\end{equation*}
\end{theoremBox}

Сразу построим суррогатную функцию для такой формулы градиента:
$$\mathcal{L}_{\textcolor{ChadPurple}{\tilde{\pi}}}(\textcolor{ChadBlue}{\theta}) \coloneqq \frac{1}{1 - \gamma}\E_{\textcolor{ChadPurple}{d_{\tilde{\pi}}(s)}} Q^{\textcolor{ChadPurple}{\tilde{\pi}}}(s, \textcolor{ChadBlue}{\pi_{\theta}}(s))$$
Действительно, если мы посчитаем градиент этой функции по $\theta$, то мы просто получим формулу chain rule для оптимизации параметров стратегии через градиент Q-функции по действиям. Иными словами, градиент по параметрам детерминированной стратегии указывает просто проводить policy improvement: выбирать те действия, для которых Q-функция больше, используя её градиент по действиям.

Если мы хотим построить Actor-Critic схему, воспользовавшись такой формулой, нам придётся аппроксимировать Q-функцию $Q_\omega(s, a) \HM \approx Q^\pi(s, a)$ и явно использовать её градиент по действиям, надеясь на то, что $\nabla_a Q_\omega(s, a) \HM\approx \nabla_a Q^\pi(s, a)$. Таким образом, обойтись обучением лишь V-функции или пользоваться многошаговыми оценками не получится, а качество обучения стратегии будет упираться в качество обучения критика, поэтому всеми преимуществами on-policy подхода в такой формуле воспользоваться не удастся.

Но можем ли мы воспользоваться формулой в off-policy режиме? Вообще говоря, нет, поскольку состояния должны приходить согласно формуле из распределения $d_{\pi_\theta}(s)$. Однако, как мы обсуждали в разделе \ref{subsec:pg_is_pi}, мы понимаем, что если мы будем оптимизировать Q-функцию по действиям, то как бы мы это не делали (из какого бы распределения не брали состояния, в которых мы изменяем стратегию), мы всё равно сможем увеличить значение нашего функционала, поскольку мы проводим policy improvement. Это означает, что если мы в формуле \eqref{dpg} заменим $d_{\pi_\theta}(s)$ на что-либо другое, полученная формула <<градиента>> будет всё равно направлением улучшения стратегии, пусть и не направлением локально максимального увеличения функционала, что верно для честного градиента. Итого, будем сэмплировать батч состояний из реплей буфера и делать шаг градиентного подъёма: 
$$\theta \leftarrow \theta + \alpha \E_{s} \nabla_\theta \pi_\theta(s) \left. \nabla_a Q^{\pi}(s, a) \right|_{a = \pi_\theta(s)},$$
где состояния $s$ приходят из произвольного распределения (например, из реплей буфера). Это эквивалентно одному шагу градиентной оптимизации суррогатной функции:
\begin{equation}\label{ddpg_actor}
\E_s Q^{\pi}(s, \pi_{\theta}(s)) \to \max_{\theta}
\end{equation}

Q-функцию $Q_\omega(s, a) \HM \approx Q^\pi(s, a)$, необходимую для такой оптимизации, будем тоже учить в off-policy режиме с одношаговых оценок: ему для данной пары $s,a$ требуется лишь сэмпл $s'$, поэтому такого критика можно обучать по переходам $\T \coloneqq (s, a, r, s')$ из буфера на таргеты
$$y(\T) \coloneqq r + \gamma Q_{\omega^{-}}\left( s', \pi(s') \right)$$

Одношаговые таргеты имеют сильное смещение (сильно опираются на выход нашей же нейросети), и поэтому для стабилизации процесса требуется использование таргет-сетей. Тут-то и можно, заметить, что...

\subsection{Связь между схемами}

\begin{theorem}
Предыдущие две схемы (вывод через DQN и через Policy Gradient) эквивалентны полностью.
\begin{proof}
Методом пристального взгляда.
\end{proof}
\end{theorem}

Итак, DQN для непрерывных действий и Policy Gradient для детерминированных стратегий --- это одно и то же. Поймём, как так случилось.

Мы двумя путями\footnote{что в очередной раз означает, что между Value-based подходом и Policy Gradient подходом есть тесная связь.} пришли к Policy Iteration схеме \ref{policyiteration}. Действительно: мы параллельно ведём два оптимизационных процесса: оцениваем $Q \HM \approx Q^\pi$ для текущей стратегии $\pi$ и учим $\pi(s) \leftarrow \argmax\limits_a Q(s, a)$, то есть проводим Policy Evaluation и Policy Improvement. При этом на этапе Policy Improvement мы делаем апдейт стратегии сразу для всех состояний, и никаких требований на <<распределение>> состояний мы можем не накладывать в силу теоремы \ref{th:policyimprovement} о Policy Improvement. 

Таким образом, обоснование, почему в выводе через Policy Gradient мы можем забить на $d_\pi(s)$ и брать состояния из буфера, можно сформулировать так: мы отказываемся от Policy Gradient подхода, в котором мы оптимизируем функционал \eqref{goal} напрямую, и переходим к Policy Iteration схеме \ref{policyiteration}.

\begin{remark}
А ещё подметим, что схема шибко похожа на GAN: критик в этом алгоритме <<учит>> функцию потерь для стратегии. Так что вполне естественно, что мы принципиально используем непрерывность пространства действий. Эта аналогия также объясняет, почему схема DDPG нестабильна; как только что-то ломается в одной из двух оптимизируемых функций (критике или актёре), другому тут же становится плохо. Поэтому схема жутко чувствительна к гиперпараметрам; пожалуй, это один из самых нестабильных алгоритмов RL.
\end{remark}

\subsection{Ornstein--Uhlenbeck Noise}

В рассмотренной схеме из-за использования детерминированной стратегии, как и в DQN, возникает проблема exploration-exploitation-а. В непрерывных пространствах действий вместо $\eps$-жадной стратегии возможно добавлять к выбранным стратегией действиям шум из нормального распределения:
$$a_t \coloneqq \pi(s_t) + \eps_t, \qquad \eps_t \sim \N(0, \sigma^2I)$$
Гиперпараметр $\sigma$, контролирующий магнитуду впрыскиваемого шума, нужно подбирать, его также можно, например, постепенно затухать к нулю с ходом обучения. Однако, такое впрыскивание шума предполагает, что исследование в соседние моменты времени независимо.

\begin{example}
Если действия робота --- это направление движения (например, поворот руля управляемой машины), а один шаг в среде это доля секунды, странно проводить исследования, рандомно <<подёргиваясь>> пару раз в секунду. Хочется целенаправленно смещать траекторию: если мы решили в целях исследования повернуть руль чуть правее, чем говорит наша детерминированная стратегия, следует сохранить это смещение руля вправо и в дальнейшем. Для моделирования этого шум должен быть скоррелированным: поэтому вместо независимого шума имеет смысл добавлять случайный процесс, колеблящийся вокруг нуля.
\end{example}

\begin{definition}
\emph{Шум Орнштейна — Уленбека} (Ornstein–Uhlenbeck noise), в начале эпизода инициализированный нулём, задаётся рекурсивно как:
$$\eps_{t + 1} \coloneqq \alpha \eps_t + \N(0, \sigma^2I),$$
где $\alpha \le 1$ и $\sigma$ --- гиперпараметры.
\end{definition}

По сути, это просто кумулятивный шум, который с коэффициентом $\alpha$ прибивается к нулю. Если $\alpha \HM= 0$, получаем обычный независимый шум из нормального распределения. Одно из преимуществ такого эксплорейшна --- считается, что его параметры можно со временем не менять, то есть даже при околооптимальном поведении такой шум будет исследовать разумные альтернативы вместо рандомных подёргиваний.

\begin{example}
На рисунке приведён пример поведения процесса Орнштейна-Уленбека для $\alpha \HM= 0.9, \sigma \HM= 1$.
\begin{center}
    \includegraphics[width=\textwidth]{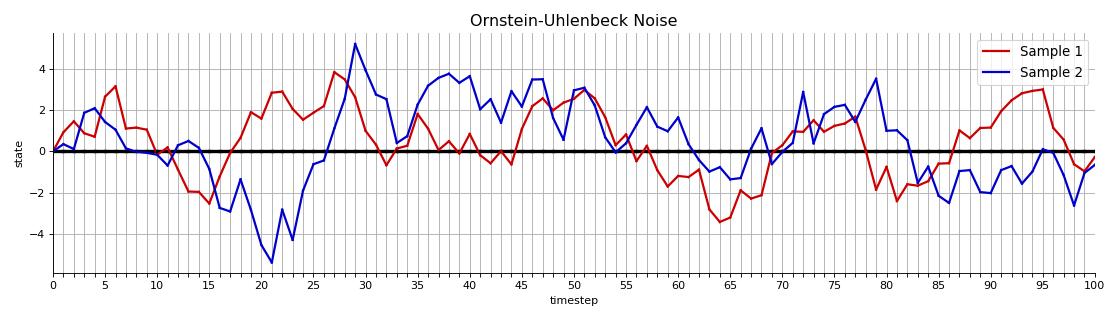}
\end{center}
\end{example}

\subsection{Deep Deterministic Policy Gradient (DDPG)}

Мы собрали алгоритм DDPG --- off-policy алгоритм для непрерывных пространств действий. Несмотря на название, по свойствам этот алгоритм похож именно на DQN, и обладает аналогичными недостатками и преимуществами.

\begin{algorithm}[label = DDPGalgorithm]{Deep Deterministic Policy Gradient (DDPG)}
\textbf{Гиперпараметры:} $B$ --- размер мини-батчей, $\beta$ --- коэф. экспоненциального сглаживания для таргет-сеток, $\alpha, \sigma$ --- параметры шума, $Q_{\theta}(s, a)$ --- нейросетка с параметрами $\theta$, $\pi_{\omega}(s)$ --- детерминированная стратегия с параметрами $\omega$, SGD-оптимизаторы.

\vspace{0.3cm}
Инициализировать $\theta, \omega$ произвольно \\
Положить $\theta^- \coloneqq \theta$ \\
Положить $\omega^- \coloneqq \omega$ \\
Инициализировать шум $\eps \coloneqq 0$ \\
Пронаблюдать $s_0$ \\
\textbf{На очередном шаге $t$:}
\begin{enumerate}
    \item обновить шум $\eps \leftarrow \alpha \eps + \hat{\eps}$, где $\hat{\eps} \sim \N(0, \sigma^2 I)$
    \item выбрать $a_t \coloneqq \pi_\omega(s_t) + \eps$
    \item пронаблюдать $r_t$,  $s_{t+1}$, $\done_{t+1}$
    \item добавить пятёрку $(s_t, a_t, r_t, s_{t+1}, \done_{t+1})$ в реплей буфер
    \item засэмплировать мини-батч размера $B$ из буфера
    \item сделать один шаг градиентного подъёма по $\omega$:
    $$\frac{1}{B}\sum_{s \in B} Q_{\theta}(s, \pi_\omega(s)) \to \max_{\omega}$$
    \item для каждого перехода $\T \coloneqq (s, a, r, s', \done)$ посчитать таргет:
    $$y(\T ) \coloneqq r + \gamma (1 - \done) Q_{\theta^{-}} \left( s', \pi_{\omega^{-}} \left( s' \right) \right)$$
    \item сделать один шаг градиентного спуска по $\theta$:
    $$\frac{1}{B}\sum_{\T} \left( Q_{\theta}(s, a) - y(\T ) \right) ^2 \to \min_\theta$$
    \item если $t \operatorname{mod} K = 0$: 
        $$\theta^{-} \gets (1 - \beta) \theta^{-} + \beta \theta$$
        $$\omega^{-} \gets (1 - \beta) \omega^{-} + \beta \omega$$
\end{enumerate}
\end{algorithm}

\begin{remark}
DDPG за счёт off-policy режима может оказаться эффективнее PPO в задаче локомоции \ref{ex:locomotion}. В этой задаче нет тех проблем, из-за которых off-policy обучение может сломаться или плохо работать: там нет сильно отложенного сигнала (плохое действие --- и существо сразу падает, хорошее действие --- и существо продвинется вперёд и получит положительное подкрепление), Q-функция как функция от действий не похожа на плато (интуитивно, в большинстве состояний есть <<хорошие>> действия с высоким значением $Q(s, a)$, которые позволяют существу продолжать бежать, и <<плохие>> с низким значением, которые нарушают баланс устойчивости существа и приводят к его падению), а функция награды очень плотная и информативная, из-за чего полезно иметь возможность все переходы много раз из буфера вспоминать и обучаться восстанавливать хотя бы $r(s, a)$, содержащуюся в одношаговых таргетах, из разных пар $s, a$.
\end{remark}

\subsection{Twin Delayed DDPG (TD3)}\label{subsec:td3}

TD3 --- набор из трёх эвристических костылей, которые можно навесить над DDPG для существенного повышения стабильности происходящих процессов.

Одна из главных проблем DDPG --- унаследованная от DQN проблема переоценивания (см. раздел \ref{subsec:overestimation}). Хотя сейчас в формулах в явном виде не присутствует оператор максимума, на самом деле он всё равно есть: наш актёр учится <<взламывать>> критика и находить те действия, где погрешность нашей аппроксимации истинной Q-функции положительна. Поэтому оценка $Q_\theta (s, \pi_\omega(s))$ практически всегда завышенно оценивает $\max\limits_a Q^\pi(s, a)$, и если это завышение попадает в целевую переменную для обучения модели Q-функции, начинается цепная реакция. 

\needspace{7\baselineskip}
\begin{wrapfigure}[9]{r}{0.35\textwidth}
\centering
\includegraphics[width=0.35\textwidth]{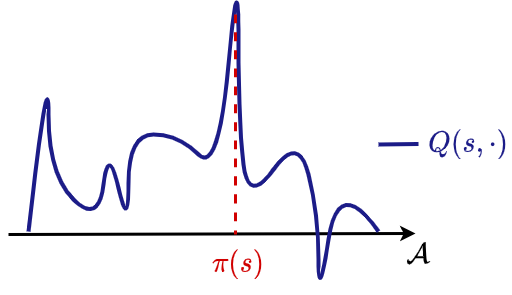}
\end{wrapfigure}

Более того, в пространстве действий могут обнаружиться узкие области, в которых наша неидеальная нейросетевая аппроксимация Q-функции имеет всплеск; актёр может научиться <<взламывать>> критика, используя эти всплески. Поэтому первая идея борьбы с этим эффектом заключается в том, чтобы при построении таргета зашумить выход нашей стратегии при помощи некоторого шума. Шум здесь не должен быть очень большим по модулю (важно, что этот шум не имеет смысл <<исследований>>): обычно сэмплируют зашумление из гауссианы и обрезают, чтобы получить что-то не очень большое по значению:
$$\eps' \sim \clip(\N(0, \hat{\sigma}I), -c, c)$$

Во-вторых, воспользуемся Twin-трюком, который мы обсуждали в разделе \ref{subsec:clippedtwin}, а то есть будем обучать две Q-функции по общему буферу и использовать в таргетах минимум из двух оценок критиков. Таргет-сеть обычно заводят для каждой из двух копий; а вот стратегию предлагается оставить для них общую, то есть использовать одно и то же <<приближение аргмакса>> для обоих Q-функций, поскольку в такой схеме она моделируется отдельной нейросетью. Итого формула таргета получается общая для двух критиков:
$$y(\T ) \coloneqq r + \gamma \min_{i \in \{1, 2\}}Q_{\theta_i^{-}}( s', \pi_{\omega^{-}}(s') + \eps')$$

При обучении актёра можно как оставить оптимизацию при помощи первого из двух критиков
$$Q_{\theta_1}(s, \pi_\omega(s)) \to \max_{\omega},$$
и тогда актёр не будет видеть одну из Q-функций (и там, где у первого критика будет взломанное актёром завышение, у второго критика, мы надеемся, завышение или занижение будет условно равновероятно), так и использовать снова минимум из двух критиков:
$$\min_{i \in \{1, 2\}} Q_{\theta_i}(s, \pi_\omega(s)) \to \max_{\omega}$$

Наконец, в-третьих, обновление весов стратегии будем делать реже, чем обновление весов Q-функции: это позволит <<поучить>> Q-функцию приближать $Q^\pi$ именно для текущей, свежей версии $\pi$, прежде чем использовать её градиент для оптимизации стратегии; здесь наблюдается полная аналогия с GAN-ами, где тоже иногда помогают подобные фокусы.

\begin{remark}
Третья эвристика кажется наименее существенной, поскольку авторы предлагали делать $N \HM= 2$ шагов обучения критика на один шаг обучения актёра.
\end{remark}

\begin{algorithm}[label = TD3]{Twin Delayed DDPG (TD3)}
\textbf{Гиперпараметры:} $B$ --- размер мини-батчей, $N$ --- периодичность обновления весов стратегии, $\alpha, \sigma$ --- параметры шума, $\hat{\sigma}, c$ --- параметры шума для добавки к действиям для таргета, $\beta$ --- коэф. экспоненциального сглаживания для таргет-сеток, $Q_{\theta_1}(s, a), Q_{\theta_2}(s, a)$ --- нейросетки с параметрами $\theta_1, \theta_2$, $\pi_{\omega}(s)$ --- детерминированная стратегия с параметрами $\omega$, SGD-оптимизаторы.

\vspace{0.3cm}
Инициализировать $\theta_1, \theta_2, \omega$ произвольно \\
Инициализировать таргет-сетки $\theta_1^- \coloneqq \theta_1, \theta_2^- \coloneqq \theta_2, \omega^- \coloneqq \omega$ \\
Инициализировать шум $\eps_0 \coloneqq 0$ \\
Пронаблюдать $s_0$ \\
\textbf{На очередном шаге $t$:}
\begin{enumerate}
    \item посчитать шум $\eps_{t} \coloneqq \alpha \eps_{t-1} + \eps$, где $\eps \sim \N(0, \sigma^2 I)$
    \item выбрать $a_t \coloneqq \pi_\omega(s_t) + \eps_{t}$
    \item пронаблюдать $r_t$,  $s_{t+1}$, $\done_{t+1}$
    \item добавить пятёрку $(s_t, a_t, r_t, s_{t+1}, \done_{t+1})$ в реплей буфер
    \item засэмплировать мини-батч размера $B$ из буфера
    \item для каждого перехода $\T \coloneqq (s, a, r, s', \done)$ посчитать таргет:
    $$\eps' \sim \clip(\N(0, \hat{\sigma}I), -c, c)$$
    $$y(\T ) \coloneqq r + \gamma (1 - \done) \min_{i \in \{1, 2\}}Q_{\theta_i^{-}}(s', \pi_{\omega^{-}}(s') + \eps')$$
    \item сделать один шаг градиентного спуска по $\theta_1$ и $\theta_2$:
    $$\frac{1}{B}\sum_{\T} \left( Q_{\theta_1}(s, a) - y(\T ) \right) ^2 \to \min_{\theta_1}$$
    $$\frac{1}{B}\sum_{\T} \left( Q_{\theta_2}(s, a) - y(\T ) \right) ^2 \to \min_{\theta_2}$$
    \item \textbf{если $t \operatorname{mod} N = 0$}:
    \begin{itemize}
        \item сделать один шаг градиентного подъёма по $\omega$:
        $$\frac{1}{B}\sum_{s \in B} Q_{\theta_1}(s, \pi_\omega(s)) \to \max_{\omega}$$
        \item обновить таргет-сети:
        $$\theta^{-}_1 \gets (1 - \beta) \theta^{-}_1 + \beta \theta_1$$
        $$\theta^{-}_2 \gets (1 - \beta) \theta^{-}_2 + \beta \theta_2$$
        $$\omega^{-}   \gets (1 - \beta) \omega^{-}   + \beta \omega$$
    \end{itemize}
\end{enumerate}
\end{algorithm}

\subsection{Обучение стохастичных политик}\label{subsec:stochpolicies}

Среди преимуществ on-policy режима, которые мы потеряли в off-policy алгоритмах, мы упоминали возможность обучать стохастичные политики. Действительно, ряд наших проблем с нестабильностью в DDPG был связан с тем, что мы учим детерминированную стратегию: нужны костыли для решения проблемы exploration-а, актёр может <<взламывать>> нашу аппроксимацию критика и приводить к overestimation-у, и всё это приходится лечить очередной порцией эвристик. Когда же в on-policy подходе мы можем получить, если так можно сказать, <<естественный>> exploration за счёт обучения стохастичной стратегии. Обязательно ли в off-policy режиме учить именно детерминированную стратегию?

На самом деле нет. В теории есть возможность обучать и стохастического актёра, хотя точные формулы градиента будут немного отличаться в зависимости от выбранной параметризации политики. Итак, допустим мы моделируем актёра $\pi_\theta(a \HM\mid s)$ в классе стохастичных стратегий.

\begin{definition}
Скажем, что для параметризации $\pi_\theta(a \HM\mid s)$ применим \emph{репараметризационный трюк} (reparameterization trick), если сэмплирование $a \HM\sim \pi_\theta(a \mid s)$ эквивалентно сэмплированию шума из некоторого не зависящего от параметров распределения $\eps \HM\sim p(\eps)$ и его дальнейшего детерминированного преобразования $a \HM= f_{\theta}(s, \eps)$.
\end{definition}

\begin{exampleBox}[label=ex:gaussian_policy]{}
Пусть наша стратегия параметризована нормальным распределением:
$$\pi_\theta(a \HM\mid s) \HM\coloneqq \N(\mu_\theta(s), \sigma_\theta(s)^2I)$$
Тогда для неё применим репараметризационный трюк: сэмплирование действий эквивалентно $a \coloneqq \mu_\theta(s) \HM+ \eps \HM\odot \sigma_\theta(s)$, где $\eps \HM\sim \N(0, I)$, $\odot$ --- поэлементное перемножение.
\end{exampleBox}

\begin{example}
Семейство детерминированных стратегий $\pi_\theta(s)$ тоже можно считать таким <<вырожденным>> примером параметризаций, для которой можно проворачивать репараметризационный трюк: просто шум $\eps$ считаем <<пустым>>.
\end{example}

Можно заметить, что если для семейства стратегий применим репараметризационный трюк, то в выводе формулы Policy Gradient можно пользоваться им вместо REINFORCE: фактически, в теореме \ref{th:dpg} мы этим воспользовались.

\begin{theorem}
Если для $\pi_\theta(a \HM\mid s)$ применим репараметризационный трюк, то:
\begin{equation}\label{rt_dpg}
\nabla_\theta J(\pi_\theta) = \frac{1}{1 - \gamma}\E_{d_{\pi_\theta}(s)} \E_{\eps \HM\sim p(\eps)} \nabla_\theta f_\theta(s, \eps) \left. \nabla_a Q^{\pi}(s, a) \right|_{a = f_\theta(s, \eps)}
\end{equation}
\begin{proof}
Полностью полностью повторяет вывод теоремы \ref{th:dpg}.
\end{proof}
\end{theorem}

Таким образом, все идеи DDPG расширяются на этот случай. Убирая из формулы градиента частоты посещения состояний и переходя к policy iteration схеме, получаем следующий функционал:
\begin{equation}\label{general_pi_ddpg}
\E_s \E_{a \sim \pi_\theta(a \mid s)} Q_\omega(s, a) \to \max_{\theta},
\end{equation}
где состояния берутся из буфера. В силу репараметризационного трюка мы легко справимся со взятием градиента на практике:
\begin{equation*}
\nabla_\theta \E_{a \sim \pi_\theta(a \mid s)} Q_\omega(s, a) = \E_{\eps \sim p(\eps)} \nabla_\theta Q_\omega(s, f_{\theta}(s, \eps))
\end{equation*}
Теперь стохастическая оценка градиента по $\theta$ получается напрямую. Понятно, что мы получили обобщение формулы \eqref{ddpg_actor}; мы оптимизируем актёра при помощи градиентов из критика, но дополнительно впрыскиваем шум в действия для получения стохастичной стратегии. 

Недостатком гауссианы является то, что она унимодальна, что может приводить к бедам.

\begin{exampleBox}[righthand ratio=0.35, sidebyside, sidebyside align=center, lower separated=false]{}
Представьте, что вы хотите объехать дерево. Вы можете объехать его справа, можете слева. Критик сообщает вам высокие значения и слева, и справа, и оптимизируя \eqref{general_pi_ddpg} в классе гауссиан, можно получить стратегию, которая с наибольшей вероятностью выбирает действие <<врезаться в дерево>>.

\tcblower
\includegraphics[width=\textwidth]{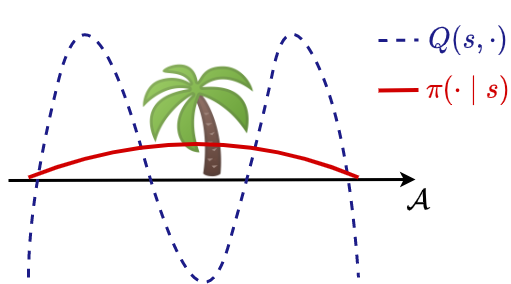}
\end{exampleBox}

Поэтому мы можем захотеть выбрать более сложную параметризацию стратегии, для которой не применим репараметризационный трюк.

\begin{exampleBox}[label=ex:gaussianmixture_policy, righthand ratio=0.35, sidebyside, sidebyside align=center, lower separated=false]{}
Например, можно использовать смесь гауссиан. Тогда при использовании $K$ компонент смеси актёр для данного состояния $s$ выдаёт следующие величины: $K$ суммирующихся в единицу чисел $w_i(s, \theta)$, а также $K$ векторов $\mu_i(s, \theta), \sigma_i(s, \theta)$, где $i \in {1, 2, \dots, K}$. Итоговое распределение полагается
$$\pi_\theta(a \mid s) \coloneqq \sum_{i = 1}^K w_i(s, \theta) \N(\mu_i(s, \theta), \sigma_i(s, \theta)^2I)$$

\tcblower
\includegraphics[width=\textwidth]{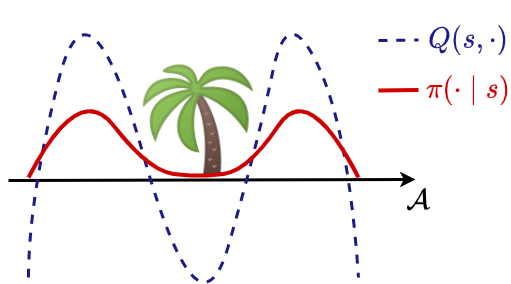}
\end{exampleBox}

Тогда для оптимизации \eqref{general_pi_ddpg} нам придётся использовать REINFORCE, обладающий более высокой дисперсией оценок, для сбивания которой необходимо использование бэйзлайна:

\begin{proposition}
Градиент \eqref{general_pi_ddpg} по параметрам актёра $\theta$ равен
$$\nabla_\theta = \E_{a \sim \pi_\theta(a \mid s)} \nabla_{\theta} \log \pi_\theta(a \mid s) \left( Q_\omega(s, a)(s, a) - b(s) \right),$$
где $b(s)$ --- бэйзлайн, произвольная функция от состояний.
\end{proposition}

Естественно, эту формулу можно интерпретировать как то, что в обычной формуле policy gradient \eqref{pgt_baseline} мы аналогично DDPG <<отказались>> от сэмплирования состояний из частот посещения текущей стратегии ради off-policy режима работы. Ну и хорошим бэйзлайном будет, соответственно, $V_\omega(s) \coloneqq \E_{a \sim \pi_\theta(a \mid s)} Q_\omega(s, a)$. Чтобы посчитать такое значение, можно либо использовать Монте-Карло оценку, либо учить отдельную нейросеть, аппроксимирующую V-функцию (целевой переменной для входа $s$ тогда будет $Q_\omega(s, a)$, где $a \HM\sim \pi_\theta(a \HM\mid s)$).

Обсудим ещё одну тонкость. В задачах непрерывного управления мы обычно работаем с пространством действий $|\A| \subseteq [-1, 1]^A$, и хотелось бы, чтобы наши семейства стратегий тоже выдавали действия именно из этого диапазона. Когда мы работали с детерминированными стратегиями, можно было учесть это в параметризации простым навешиванием гиперболического тангенса $\tanh$ на последний слой сети. А что, если мы используем гауссиану или смесь гауссиан?

\begin{remark}
Как ни странно, игнорирование этого нюанса может всё равно сработать на практике. То есть, актёр моделирует гауссиану, и считает, что он отправил сэмплы $a$ в среду, которые могут выходить за требуемый диапазон. Неожиданное преимущество такого подхода в том, что иногда в задачах <<оптимальные>> действия находятся на границе и равны +1 или -1 в каких-то компонентах, и тогда такая необрезанная гауссиана может довольно удобно <<часто>> сэмплировать подобные действия. Тем не менее, чаще имеет смысл всё же учесть домен в параметризации. 
\end{remark}

Тогда можно применять гиперболический тангенс к сэмплам из <<необрезанной>> модели. То есть, стратегия объявляется следующей: $\pi_\theta(a \HM \mid s) \coloneqq \tanh(u)$, где $u \HM \sim \mu_\theta(u \HM \mid s)$, и $u \HM \in \R^A$. Понятно, как тогда применять репараметризационный трюк, но если мы пользуемся этой идеей вместе с REINFORCE (например, для обучения смеси гауссиан или в on-policy алгоритмах), то нам нужно уметь считать $\log \pi_\theta(a \HM \mid s)$. Для этого нужно вспоминать правило замены переменных в плотностях (см., например, \href{https://ru.wikipedia.org/wiki/Плотность_вероятности#Плотность_преобразования_случайной_величины}{википедию}).

\begin{theorem}[Формула замены переменной в плотности]
Пусть $\pi_\theta(a \HM \mid s) \coloneqq g(u)$, где $g \colon \R^A \to \R^A$, и $u \HM \sim \mu_\theta(u \HM \mid s)$. Тогда:

\begin{equation}\label{changeofvariabledensity}
\log \pi_\theta(a \HM \mid s) = \log \mu_\theta(a \HM \mid s) - \log | \det \nabla_u g | 
\end{equation}

\begin{proof}[Без доказательства]
\end{proof}
\end{theorem}

\begin{example}
Например, для функции $g(u) \coloneqq \tanh (u)$ якобиан $\nabla_u g$ есть диагональная матрица (поскольку преобразование поэлементное), и его определитель равен покомпонентному произведению:
\begin{align*}
\det \nabla_u g(u) &= \prod_{i=1}^{A} \nabla_{u_i} \tanh(u_i) = \\
= \{ \text{производная гиперболического тангенса} \} &= \prod_{i=1}^A (1 - \tanh^2(u_i))    
\end{align*}
Заметим, что все компоненты положительные ($\tanh(u_i) \HM \le 1$), поэтому модуль из формулы \eqref{changeofvariabledensity} брать не нужно. Подставляя в \eqref{changeofvariabledensity}, получаем окончательно:
$$\log \pi_\theta(a \HM \mid s) = \log \mu_\theta(a \HM \mid s) - \sum_{i=1}^A \log (1 - \tanh^2(u_i)) $$
\end{example}

Итак, теоретически возможность обучать стохастичные стратегии в off-policy режиме есть. Что тогда мешает в DDPG воспользоваться этим и использовать гауссиану или смесь гауссиан? Дело в том, что мы помним, что направление оптимизации актёра --- детерминированная стратегия: мы идём в сторону жадной стратегии $\pi_\theta(s) \HM= \argmax\limits_a Q_\omega(s, a)$. Схлопывание стохастичной стратегии в детерминированную чревато численными проблемами: например, для гауссианы дисперсия начнёт уходить в ноль, градиенты могут начать взрываться.

Можно, конечно, попытаться как-то побороться с этим эффектом, например, при помощи эвристики, которой часто пользуются в on-policy методах --- добавкой энтропийного лосса. Напомним: чем больше значение энтропии для распределения, тем оно <<менее вырожденное>>:

\begin{definition}
Энтропией распределения $\pi(a)$ называется
\begin{equation}\label{entropydef}
\entropy(\pi(a)) \coloneqq - \E_{\pi(a)} \log \pi(a)
\end{equation}
\end{definition}

И в policy gradient алгоритмах часто в формулу градиента добавляют слагаемое $\nabla_{\theta} \entropy(\pi_\theta(\cdot \mid s))$, которое поощряет высокую энтропию стратегии. Однако в on-policy режиме Q-функция заменялась на оценку и была стохастичной. В off-policy же актёр имеет куда больше шансов <<переобучиться>> под критика, и энтропийный лосс придётся тогда выставлять с большим коэффициентом.

Вместо подобных плясок с бубном хотелось бы, чтобы подобных проблем в принципе не возникало. Конечно, детерминированность оптимальной стратегии --- особенность постановки задачи RL, и поэтому если мы хотим, чтобы таких эффектов в оптимизационных процессах не было, нам придётся найти какую-то альтернативную постановку задачи. Оказывается, такая альтернативная формулировка есть, и она бывает крайне удобна. В ней оптимальные стратегии уже будут стохастичны, и ряд численных проблем, а также проблем с exploration-ом, отпадёт; в частности, она <<обоснует>> добавку градиента энтропии в формулу градиента.
\section{Soft Actor-Critic}\label{SACsection}

\subsection{Maximum Entropy RL}\label{maximumentropyrlsubsection}

\begin{definition}
Задачей \emph{Maximum Entropy RL} является максимизация функционала
\begin{equation}\label{merl}
    J_{\soft}(\pi) \coloneqq \E_{\Traj \sim \pi} \sum_{t \ge 0} \gamma^t \left[ r_t + \alpha \entropy(\pi(\cdot \mid s)) \right] \to \max_\pi
\end{equation}
где $\alpha$ --- гиперпараметр, называемый \emph{температурой} (temperature).
\end{definition}

\begin{proposition}
Задача \eqref{merl} эквивалентна
\begin{equation*}
    J_{\soft}(\pi) \coloneqq \E_{\Traj \sim \pi} \sum_{t \ge 0} \gamma^t \left[ r_t - \alpha \log \pi(a_t \mid s_t) \right] \to \max_\pi,
\end{equation*}
\begin{proof}
Следует из определения энтропии \eqref{entropydef}; ведь мат.ожидание по действиям присутствует в мат.ожидании по траекториям.
\end{proof}
\end{proposition}

Интуиция такого функционала проста: мы говорим, что хотим не просто оптимальную стратегию, а самую стохастичную около-оптимальную стратегию (коэффициент $\alpha$ масштабирует добавку и определяет важность добавленного слагаемого). Цель --- избегать локальных оптимумов в среде. 

\begin{exampleBox}[righthand ratio=0.3, sidebyside, sidebyside align=center, lower separated=false]{}
Представим, что у агента есть два пути, и по мере углубления награда на каждом пути одинаково растёт. Пусть первый путь заканчивается тупиком и суммарно позволяет набрать не более +100, а на втором тупик стоит чуть дальше и даёт +110. Во время обучения агент может уловить награду вдоль первого пути и учиться углубляться в него, игнорируя исследование второго пути, даже если агент умеет набирать там награду как на первом; за счёт бонуса за наиболее стохастичную стратегию агент мотивирован в течение обучения в начале эпизодов случайно выбирать между двумя путями. То есть, энтропийный бонус помогает избегать подобных <<локальных максимумов>> в среде.

\tcblower
\includegraphics[width=\textwidth]{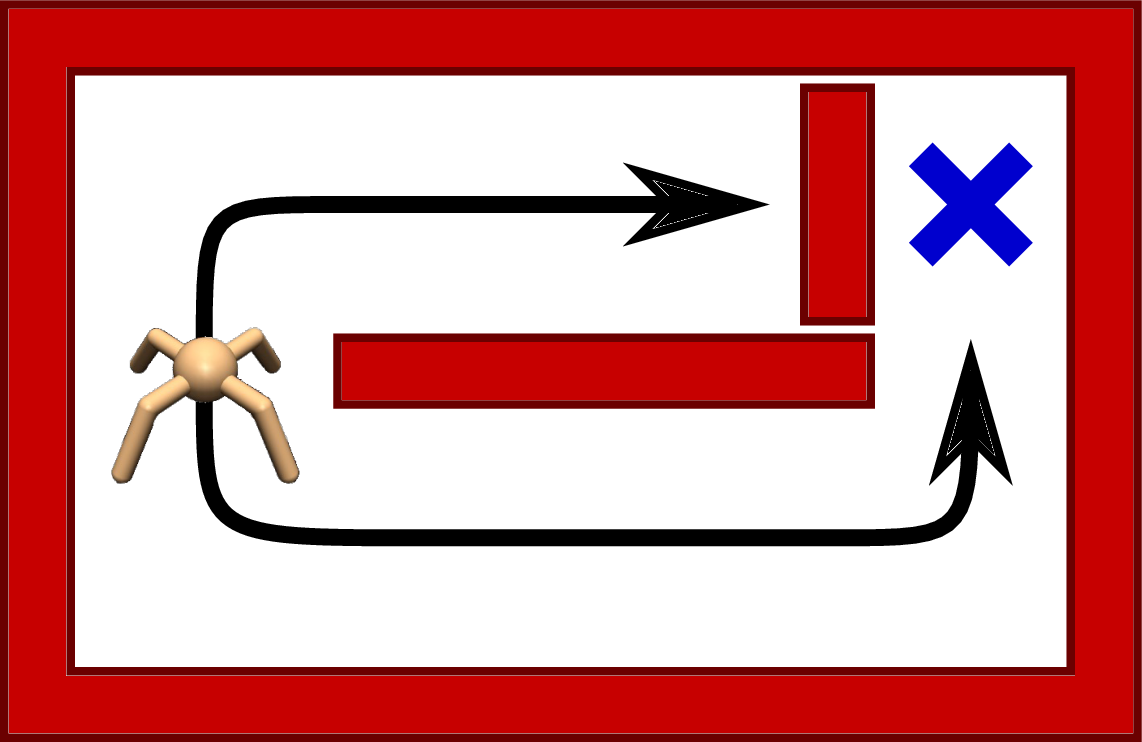}
\end{exampleBox}

Можно считать, что в данном фреймворке мы на самом деле лишь чуть-чуть модифицировали награду в среде:
\begin{equation}\label{softreward}
r_{\soft}(s, a) \coloneqq r(s, a) - \alpha \log \pi(a \mid s)
\end{equation}
Построенную теорию теперь нужно перепроверять, поскольку награда $r$ стала зависеть напрямую от вероятностей, выдаваемых стратегией (такого в формализме MDP не предполагалось), да и очевидно, что не все утверждения переносятся на новую постановку:

\begin{proposition}
Оптимальной детерминированной стратегии может не существовать.
\begin{proof}
В MDP, где награда всегда 0, оптимальна стратегия с максимальной энтропией.
\end{proof}
\end{proposition}

Однако принцип Policy Iteration --- чередование этапов оценивания и улучшения стратегии --- остаётся для этой альтернативной постановки задачи неизменным. Чтобы подчеркнуть, что речь идёт именно о постановке Maximum Entropy RL, ко всем терминам из обычной теории добавляют слово \emph{soft} (<<мягкие>>), подразумевая, что энтропийный бонус <<сглаживает>> стратегию, предупреждая её вырождение в детерминированную. Поэтому будем обозначать оценочные функции в рамках данного фреймворка как $Q^\pi_{\soft}, V^\pi_{\soft}$. Также без ограничения общности далее будем считать $\alpha \HM= 1$, так как всегда можно перемасштабировать награду $r(s, a)$ без изменения задачи.

\begin{remark}
Гиперпараметр температуры является основным недостатком идеи Maximum Entropy RL --- на практике его обычно очень сложно подбирать, а он существенно влияет на то, какая стратегия получится в итоге обучения.
\end{remark}

Мы также должны договориться о том, в какой момент во время взаимодействия приходит энтропийный бонус. Формула \eqref{softreward} предполагает, что после выбора действия $a$ из состояния $s$ помимо награды $r(s, a)$ агент ещё поощряется $\log \pi(a \mid s)$; однако мы договоримся по-другому\footnote{такая договорённость удобнее, поскольку после сэмплирования действия $a$ в состоянии $s$ неудобно учитывать зависимость награды от распределения $\pi(\cdot \HM\mid s)$.}. Будем считать, что агент, приходя в некоторое состояние $s$, получает бонус в размере $\entropy(\pi(\cdot \mid s)) \HM\coloneqq - \E_{a \sim \pi(a \mid s)} \log \pi(a \mid s)$ (если состояние не терминальное), затем сэмплирует действие $a$, получает бонус $r(s, a)$ и потом наблюдает $s'$. В такой договорённости уравнения для мягких оценочных функций выглядят так:

\begin{theorem}[Мягкие уравнения связи]
\begin{equation}\label{softQV}
Q^\pi_{\soft}(s, a) = r(s, a) + \gamma \E_{s'} V^\pi_{\soft}(s')
\end{equation}
\begin{equation}\label{softVQ}
V^\pi_{\soft}(s) = \E_{\pi(a \mid s)} \left[ Q^\pi_{\soft}(s, a) - \log \pi(a \mid s) \right]
\end{equation}
\begin{proof}
По определению с учётом договорённости.
\end{proof}
\end{theorem}

\begin{theorem}[Мягкие уравнения Беллмана (soft Bellman equations)]
\begin{equation}\label{softQQ}
Q^\pi_{\soft}(s, a) = r(s, a) + \gamma \E_{s'} \E_{a'} \left[ Q^\pi_{\soft}(s', a') - \log \pi(a' \mid s') \right]
\end{equation}
\begin{equation}\label{softVV}
V^\pi_{\soft}(s) = \E_{a} \left[ r(s, a) - \log \pi(a \mid s) + \gamma \E_{s'} V^\pi_{\soft}(s') \right]
\end{equation}
\end{theorem}

\subsection{Soft Policy Evaluation}

Имея на руках мягкие уравнения Беллмана, можно сразу же построить алгоритм оценивания стратегии (\emph{Soft Policy Evaluation}), обучения мягкой Q- или V-функции по заданной стратегии $\pi$. Технически, необходимо только проверить, что правые части мягких уравнений Беллмана являются сжатиями:

\begin{theorem}
Операторы, стоящие в правой части мягких уравнений Беллмана, являются сжимающими с коэффициентом $\gamma$ по метрике
$$\rho(V_1, V_2) \coloneqq \max\limits_{s} | V_1(s) - V_2(s) |$$
\begin{proof}
Покажем для мягкой V-функции. Пусть $\B_{\soft}$ --- оператор, стоящий в правой части \eqref{softVV}, и пусть даны две V-функции $V_1, V_2$.
Тогда:
$$
| [\B_{\soft} V_1](s) - [\B_{\soft} V_2](s) | = \gamma | \E_{a} \E_{s'} \left[ V_1(s') - V_2(s') \right] |,
$$
поскольку энтропия стратегии $\pi$ вместе с наградой за шаг одинакова для $V_1$ и $V_2$ и потому сокращается. Дальше, как и для обычных V-функций, можно просто оценить это выражение сверху $\gamma \rho(V_1, V_2)$, заканчивая доказательство.
\end{proof}
\end{theorem}

\begin{proposition}
Метод простой итерации сходится к единственному решению мягких уравнений Беллмана из любого начального приближения.
\end{proposition}

Итак, мы уже можем сразу построить процедуру обучения критика. Рассмотрим обучение $Q_\omega(s, a) \HM \approx Q^\pi_{\soft}(s, a)$ с одношаговых оценок в off-policy режиме, то есть будем просто решать мягкое уравнение Беллмана \eqref{softQQ}. Тогда для заданного перехода $\T \HM= (s, a, r, s')$ целевая переменная строится как
$$y(\T) \coloneqq r + \gamma \E_{a' \sim \pi(a' \mid s')} \left[ Q_\omega(s', a') - \log \pi(a' \mid s') \right]$$

В непрерывных пространствах действий взять мат.ожидание по $a'$ аналитически может не удастся (например, если $\pi$ моделируется гауссианой). Можно, как всегда, тоже заменить его на Монте-Карло оценку (по возможности, взяв аналитически энтропию). Но при желании есть возможность ещё немножко сбить дисперсию, заведя вспомогательную нейросеть для аппроксимации $V_\psi(s) \HM \approx V^\pi_{\soft}(s)$: то есть, фактически, просто учить мат.ожидание $\E_{a'}$.

В таком варианте таргеты для критика (Q-функция с параметрами $\omega$) и для вспомогательной V-функции выглядят следующим образом: для заданного перехода $\T \HM= (s, a, r, s')$, взятого из буфера, генерируем $a_\pi$ из текущей версии стратегии $\pi$ и запоминаем вероятность $\pi(a_\pi \HM\mid s)$, после чего вычисляем несмещённые оценки правых частей уравнений связи \eqref{softVQ} и \eqref{softQV}:
$$y_Q(\T) \coloneqq r + \gamma V_\psi(s')$$
$$y_V(\T) \coloneqq Q_\omega(s, a_\pi) - \log \pi(a_\pi \mid s)$$
И с такими таргетами, как обычно, решаем задачу регрессии с функцией потерь MSE. Самое главное --- соответствующее действие $a_\pi$ для таргета $y_V$ брать не из буфера, а сгенерировать из текущей версии стратегии, поскольку иначе мы некорректно оценим среднее $\E_{a \sim \pi}$.

Естественно, в любых одношаговых таргетах оценка очень смещённая, и поэтому для стабилизации полезны две стандартные эвристики --- таргет-сети и clipped double оценки (см. раздел \ref{subsec:clippedtwin}). В реализациях места, где используются таргет-сети, могут различаться, поскольку однозначно сложно сказать, где их лучше всего использовать, но они точно должны быть; без таргет-сетей эти алгоритмы не заработают из-за сильной смещённости всех оценок.

\subsection{Soft Policy Improvement}

Теперь обсудим, как обучать актёра. Напомним, что вся теория оптимизации стратегии сводится к policy improvement и нам нужен аналог теоремы \ref{th:policyimprovement}. Построить этот аналог можно из простых соображений: вот нам дана стратегия $\textcolor{ChadBlue}{\pi_1}$ и её мягкая оценочная функция $\textcolor{ChadBlue}{Q^{\pi_1}_{\soft}}(s, a)$. Как можно улучшить эту стратегию из состояния $s$? Сейчас $\textcolor{ChadBlue}{\pi_1}$ набирает из $s$ в среднем $\textcolor{ChadBlue}{V^{\pi_1}_{\soft}}(s)$. А сколько мы будем в среднем набирать, если сейчас в состоянии $s$ сменим стратегию на $\textcolor{ChadPurple}{\pi_2}$ (теперь с учётом энтропийного бонуса, который мы тогда получим в $s$ за эту новую стратегию $\textcolor{ChadPurple}{\pi_2}$), а в будущем будем вести себя ну по крайней мере не хуже, чем стратегия $\textcolor{ChadBlue}{\pi_1}$?

\begin{theoremBox}[label=th:softpi]{Soft Policy Improvement}
Пусть стратегии $\textcolor{ChadBlue}{\pi_1}$ и $\textcolor{ChadPurple}{\pi_2}$ таковы, что для всех состояний $s$ выполняется:
$$\E_{\textcolor{ChadPurple}{\pi_2}(a \mid s)} \textcolor{ChadBlue}{Q^{\pi_1}_{\soft}}(s, a) + \entropy(\textcolor{ChadPurple}{\pi_2}(\cdot \mid s)) \ge \textcolor{ChadBlue}{V^{\pi_1}_{\soft}}(s),$$
тогда $\textcolor{ChadPurple}{\pi_2} \succeq \textcolor{ChadBlue}{\pi_1}$ с учётом энтропийного бонуса; если хотя бы для одного $s$ неравенство выполнено строго, то $\textcolor{ChadPurple}{\pi_2} \succ \textcolor{ChadBlue}{\pi_1}$.
\begin{proof}
Полностью аналогично доказательству в обычном случае (теорема \ref{th:policyimprovement}).
Покажем, что $\textcolor{ChadPurple}{V^{\pi_2}_{\soft}}(s) \ge \textcolor{ChadBlue}{V^{\pi_1}_{\soft}}(s)$ для любого $s$:
\begin{align*}
\textcolor{ChadBlue}{V^{\pi_1}_{\soft}}(s) \le \{ \text{по построению $\textcolor{ChadPurple}{\pi_2}$} \} &\le \E_{\textcolor{ChadPurple}{\pi_2}(a \mid s)} \textcolor{ChadBlue}{Q^{\pi_1}_{\soft}}(s, a) + \entropy(\textcolor{ChadPurple}{\pi_2}(\cdot \mid s)) = \\
= \{ \text{связь QV \eqref{softQV}} \} &= \E_{\textcolor{ChadPurple}{\pi_2}(a \mid s)} \left[ r  + \entropy(\textcolor{ChadPurple}{\pi_2}(\cdot \mid s)) + \gamma \E_{s'} \textcolor{ChadBlue}{V^{\pi_1}_{\soft}}(s') \right]  \le \\
\le \{ \text{применяем это же неравенство рекурсивно} \} &= \E_{\textcolor{ChadPurple}{\pi_2}(a \mid s)} \Bigl[ r + \entropy(\textcolor{ChadPurple}{\pi_2}(\cdot \mid s)) + \\ &+ \E_{s'} \E_{\textcolor{ChadPurple}{\pi_2}(a' \mid s')} \left[ \gamma r' + \gamma \entropy(\textcolor{ChadPurple}{\pi_2}(\cdot \mid s')) + \gamma^2 \E_{s''} \textcolor{ChadBlue}{V^{\pi_1}_{\soft}}(s'') \right] \Bigr] \le \\
\le \{ \text{раскручиваем цепочку далее} \} &\le \dots \le \E_{\textcolor{ChadPurple}{\Traj \sim \pi_2} \mid s_0 = s} \sum_{t \ge 0} \gamma^t r_t + \gamma^t \entropy(\textcolor{ChadPurple}{\pi_2}(\cdot \mid s_t)) = \\
= \{ \text{по определению мягкой V-функции} \} &= \textcolor{ChadPurple}{V^{\pi_2}_{\soft}}(s)
\end{align*}
Если для какого-то $s$ неравенство из условия теоремы было выполнено строго, то для него первое неравенство в этой цепочке рассуждений выполняется строго, и, значит, $\textcolor{ChadPurple}{V^{\pi_2}_{\soft}}(s) > \textcolor{ChadBlue}{V^{\pi_1}_{\soft}}(s)$.
\end{proof}
\end{theoremBox}

Таким образом, аналог общей схемы Generalized Policy Iteration в задаче Maximum Entropy RL выглядит так:
\begin{itemize}
    \item \textbf{Soft Policy Evaluation} заключается в обучении аппроксимации мягкой оценочной функции $Q_\omega \HM \approx Q^{\pi}_{\soft}$ для текущей стратегии $\pi$;
    \item \textbf{Soft Policy Improvement} заключается в оптимизации следующего функционала (для разных состояний $s$):
    \begin{equation}\label{softpolicyimprov}
    \E_{\pi(a \mid s)} Q_\omega(s, a) + \entropy(\pi(\cdot \mid s)) \to \max_{\pi}
    \end{equation}
\end{itemize}

Соответственно, для обучения актёра, заданного параметрической стратегией $\pi_\theta(a \HM\mid s)$, нужно просто оптимизировать не среднее значение Q-функции, а учесть дополнительно добавку энтропийного бонуса в рассматриваемом состоянии. У задачи оптимизации \eqref{softpolicyimprov} есть один важный альтернативный вид:

\begin{theoremBox}[label=softpi_is_kl_minimization]{}
Задача \eqref{softpolicyimprov} эквивалентна задаче
$$\KL(\pi_\theta(\cdot \mid s) \parallel \exp Q_\omega(s, \cdot)) \to \min_{\theta},$$
где $\exp Q_\omega(s, \cdot)$ --- ненормированное распределение над действиями.
\beginproof
Обозначим нормировочную константу распределения $\exp Q_\omega(s, \cdot)$ как $Z_\omega(s) \HM\coloneqq \int\limits_\A \exp Q\omega(s, a) \diff a$. Тогда:
$$\KL(\pi_\theta(\cdot \mid s) \parallel \exp Q_\omega(s, \cdot)) = \E_{a \sim \pi_\theta(a \mid s)} \log \pi_\theta(a \mid s) - Q_\omega(s, a) + \log Z_\omega(s)$$
Осталось заметить, что при домножении на минус получим \eqref{softpolicyimprov}: первое слагаемое есть энтропия, а третье слагаемое не зависит от оптимизируемых параметров $\theta$:
\begin{equation*}
\E_{a \sim \pi_\theta(a \mid s)} \log Z_\omega(s) = \log Z_\omega(s) = \const(\theta)    \tagqed
\end{equation*}
\end{theoremBox}

\begin{theorem}[Вид жадной стратегии]
Максимальное значение \eqref{softpolicyimprov} достигается на стратегии
\begin{equation}\label{softgreedy}
\pi(a \mid s) \propto \exp Q_\omega(s, a)
\end{equation}
\begin{proof}
Следует из теоремы \ref{softpi_is_kl_minimization}, поскольку минимум KL-дивергенции достигается в нуле на совпадающих распределениях.
\end{proof}
\end{theorem}

Таким образом мы показали, что актёр просто пытается выучить распределение $\pi_\theta(a \HM\mid s) \HM\propto \exp{Q_\omega(s, \cdot)}$ --- больцмановскую стратегию по отношению к Q-функции \eqref{boltzman}. Мы сталкивались с ней, когда обсуждали способы исследования. Соответственно, в Maximum Entropy RL понятие жадной стратегии немножко другое: нужно не аргмакс по действиям брать, а софтмакс.

На практике второе слагаемое функционала \eqref{softpolicyimprov} --- энтропию $\pi_\theta(a \mid s)$ --- обычно можно рассчитать аналитически для выбранной параметризации $\pi_{\theta}(a \HM\mid s)$. Первое же слагаемое берётся так, как мы обсуждали в разделе \ref{subsec:stochpolicies}: при помощи репараметризационного трюка, если такая возможность есть, и при помощи REINFORCE иначе.

\begin{exampleBox}[label=ex:gaussianactor_sac]{}
Пусть наша стратегия параметризована гауссианой (см. пример \ref{ex:gaussian_policy}). Для неё можно проводить репараметризационный трюк и также можно аналитически посчитать энтропию:
$$\entropy(\N(\mu, \sigma^2I)) = \sum_{i=1}^{A} \log \sigma_i$$

Итого, формула \eqref{softpolicyimprov} в таком случае получается следующей:
$$\E_{\eps \sim \N(0, I)} Q_\omega(s, \mu_\theta(s) + \sigma_\theta(s) \odot \eps) + \sum_{i=1}^A \log \sigma_i(s, \theta) \to \max_\theta $$
\end{exampleBox}

\begin{exampleBox}[label=ex:gaussianmixtureactor_sac]{}
Пусть наша стратегия параметризована смесью гауссиан (см. пример \ref{ex:gaussianmixture_policy}). Тогда для неё не применим репараметризационный трюк, и сложно аналитически посчитать энтропию. Тогда придётся применять REINFORCE, и  формула градиента \eqref{softpolicyimprov} получается следующей:
$$\nabla_\theta = \E_{a \sim \pi_\theta(a \mid s)} \nabla_\theta \log \pi_\theta(a \mid s) \left[ Q_\omega(s, a) - \log \pi_\theta(a \mid s) - b(s) \right],$$
где $b(s)$ --- бэйзлайн, произвольная функция от состояний. Имеет смысл делать её близкой к среднему значению $Q_\omega(s, a) \HM- \log \pi_\theta(a \mid s)$, то есть хорошо выбирать $b(s) \HM\coloneqq V_\omega(s) = \E_{a} \left[ Q_\omega(s, a) \HM- \log \pi_\theta(a \mid s) \right]$.
\end{exampleBox}

\subsection{Soft Actor-Critic (SAC)}

Мы уже можем собрать алгоритм-аналог DDPG для задачи Maximum Entropy RL. В этом алгоритме по аналогии с DDPG есть актёр и критик, критик учится оценивать актёра по одношаговым таргетам, а актёр обучается моделированием soft policy improvement-а. Таким образом, схема работает в off-policy режиме.

Перечислим ряд деталей, с которыми в целом можно играться в этом алгоритме\footnote{существует несколько версий этого алгоритма; в частности, есть версия, где Q-функции учатся через Q-функции, и вспомогательная V-сеть не используется. Тогда нужно заводить таргет-сети для двух Q-сетей.}. При обучении критика используется вспомогательная V-функция, для стабилизации можно ограничиться таргет-сетью (с параметрами $\psi^{-}$) только для V-функции:
$$y_Q(\T) \coloneqq r + \gamma V_{\psi^{-}}(s'),$$
а для борьбы с overestimation bias в таргете для V-функции используется twin оценка аналогично алгоритму TD3 (см.~раздел \ref{subsec:td3}), то есть обучается две Q-функции с параметрами $\omega_1, \omega_2$, и таргет $y_V$ строится по следующей формуле: 
$$y_V(\T) \coloneqq \min_{i = {1, 2}} Q_{\omega_i}(s, a_\pi) - \log \pi(a \mid s),$$
где $a_\pi \sim \pi_\theta(a \mid s)$.

Наконец, в функционале \eqref{softpolicyimprov} актёру предлагается <<взламывать>> минимум из двух Q-функций, то есть использовать twin-оценку и в этой формуле:
$$\E_{\pi_\theta(a \mid s)} \min_{i=1,2} Q_{\omega_i}(s, a) + \entropy(\pi(\cdot \mid s)) \to \max_{\pi}$$

Далее схема приведена для простоты для случая, если актёр моделирует гауссиану (см.~пример \ref{ex:gaussianactor_sac}). Конечно, возможно использование и для смеси гауссиан (см.~пример \ref{ex:gaussianmixture_policy}) при замене в функционале актёра репараметризационного трюка на REINFORCE. 

\begin{algorithm}[label = SACalgorithm]{Soft Actor-Critic (SAC)}
\textbf{Гиперпараметры:} $B$ --- размер мини-батчей, $\beta$ --- параметр экспоненциального сглаживания таргет-сети, $\alpha$ --- температура, $\pi_{\theta}(a \mid s) \coloneqq \N(\mu_{\theta}(s), \sigma_{\theta}(s)^2 I)$ --- гауссова стратегия с параметрами $\theta$, $Q_{\omega_1}(s, a), Q_{\omega_2}(s, a)$ --- две нейросетки с параметрами $\omega_1$ и $\omega_2$, $V_\psi(s)$ --- нейросетка с параметрами $\psi$, SGD-оптимизаторы.

\vspace{0.3cm}
Инициализировать $\theta, \omega_1, \omega_2, \psi$ произвольно \\
Инициализировать таргет-сеть $\psi^- \coloneqq \psi$ \\
Пронаблюдать $s_0$ \\
\textbf{На очередном шаге $t$:}
\begin{enumerate}
    \item выбрать $a_t \sim \pi_\theta(a_t \mid s_t)$
    \item пронаблюдать $r_t$,  $s_{t+1}$, $\done_{t+1}$
    \item добавить пятёрку $(s_t, a_t, r_t, s_{t+1}, \done_{t+1})$ в реплей буфер
    \item засэмплировать мини-батч размера $B$ из буфера
    \item для каждого $s$ из батча засэмплировать шума $\eps(s) \sim \N(0, I)$ и посчитать $\mu(s, \theta)$, $\sigma(s, \theta)$ стратегии $\pi_\theta$
    \item посчитать оценку градиента по параметрам стратегии:
    $$\nabla_\theta \coloneqq \frac{1}{B}\sum_{s \in B} \nabla_\theta \left[ \alpha \sum_{i=1}^A \log \sigma_i(s, \theta) + \min_{i=1,2} Q_{\omega_i}(s, \mu_\theta(s) + \sigma_\theta(s) \odot \eps(s)) \right]$$
    \item делаем шаг градиентного подъёма по $\theta$, используя $\nabla_\theta$
    \item для каждого перехода $\T \coloneqq (s, a, r, s', \done)$ засэмплировать $a_{\pi} \sim \pi_\theta(a_{\pi} \mid s)$ и сохранить вероятности $\pi_\theta(a_\pi \mid s)$
    \item посчитать таргеты:
    $$y_V(\T ) \coloneqq \min_{i = 1, 2} Q_{\omega_i} ( s, a_\pi ) - \alpha \log \pi_\theta(a_\pi \mid s)$$
    $$y_Q(\T ) \coloneqq r + \gamma V_{\psi^{-}}(s')$$
    \item посчитать лоссы:
    $$\Loss_V(\psi) \coloneqq \frac{1}{B}\sum_{\T} \left( V_{\psi}(s') - y_V(\T ) \right) ^2$$
    $$\Loss_{Q1}(\omega_1) \coloneqq \frac{1}{B}\sum_{\T} \left( Q_{\omega_1}(s, a) - y_Q(\T ) \right) ^2$$
    $$\Loss_{Q2}(\omega_2) \coloneqq \frac{1}{B}\sum_{\T} \left( Q_{\omega_2}(s, a) - y_Q(\T ) \right) ^2$$
    \item делаем шаг градиентного спуска по $\psi$, $\omega_1$ и $\omega_2$, используя $\nabla_\psi \Loss_V(\psi)$, $\nabla_\omega \Loss_{Q1}(\omega_1)$ и $\nabla_{\omega_2} \Loss_{Q2}(\omega_2)$ соответственно
    \item обновляем таргет-сеть: $\psi^{-} \leftarrow (1 - \beta) \psi^{-} + \beta \psi$
\end{enumerate}
\end{algorithm}

Можно провести sanity check, убедившись, что при $\alpha \HM\to 0$, мы получаем обычный DDPG (с twin-трюком и для стохастичной стратегии).

\begin{remark}
Консенсуса по поводу того, кто круче --- TD3 (алгоритм \ref{TD3}) или SAC, нету. Считается, что они оба хорошо работают, по крайней мере лучше DDPG, и на разных задачах может победить как первый, так и второй. Недостатком TD3 является некоторая <<костыльность>> предложенных эвристик, а недостатком SAC --- неудобный гиперпараметр температуры.
\end{remark}

\subsection{Аналоги других алгоритмов}

В обычном RL оптимальной стратегией была жадная по отношению к оптимальной Q-функции, то есть та, для которой нельзя провести policy improvement ни в одном состоянии. Аналогичные рассуждения верны и для Maximum Entropy RL, разве что форма жадной стратегии \eqref{softgreedy} теперь другая.

Сформулируем критерий оптимальности Беллмана в задаче Maximum Entropy RL. По аналогии с обычным случаем, введём оптимальные оценочные функции:
$$V^*_{\soft}(s) \coloneqq \max_{\pi} V^{\pi}_{\soft}(s)$$
$$Q^*_{\soft}(s, a) \coloneqq \max_{\pi} Q^{\pi}_{\soft}(s, a)$$

\begin{theorem}[Вид оптимальной стратегии для Maximum Entropy RL]
Оптимальной является единственная стратегия
$$\pi(a \mid s) \propto \exp{Q^*_{\soft}(s, a)}$$
\begin{proof} Проводим рассуждения аналогичные теореме \ref{th:nonstatbellmancriterion}. Откажемся от стационарности и будем рассматривать задачу поиска оптимальной стратегии $\pi_t(a \mid s_0)$ в предположении, что в будущем мы сможем набрать максимально возможную награду $Q^*_{\soft}(s, a, t)$:
$$\E_{\pi_t(a \mid s)} \left[ Q^*_{\soft}(s, a, t) - \log \pi_t(a \mid s) \right] \to \max\limits_{\pi_t(a \mid s)}
$$

Аналогично теореме \ref{softpi_is_kl_minimization} про soft policy improvement, можно заметить, что с точностью до константы, не зависящей от $\pi_t$, оптимизируемое выражение есть:
$$-\KL \left( \pi_t(a \mid s) \parallel \frac{\exp Q^*_{\soft}(s, a, t)}{Z(s, t)} \right) \to \max\limits_{\pi_t(a \HM\mid s)},$$
где $Z(s, t)$ --- нормировочная константа $\exp Q^*_{\soft}(s, a, t)$. Понятно, что оптимум достигается в нуле на $\pi_t(a \HM\mid s)$, совпадающей с этим распределением.

Дальнейшее рассуждение строится как раньше: $Q^*_{\soft}$ от времени не зависит по определению (аналогично утв. \ref{pr:nonstat_optimal_are_stat}), поэтому оптимальная стратегия получается стационарной, следовательно
\begin{equation*}
\pi(a \mid s) \propto \exp Q^*_{\soft}(s, a)
\end{equation*}
Заметим, что в силу однозначного определения $Q^*_{\soft}(s, a)$, такая стратегия в принципе единственна в отличие от обычной задачи RL.
\end{proof}
\end{theorem}




Имея на руках вид оптимальной стратегии, мы можем получить уравнения оптимальности:

\begin{theorem}
\begin{equation}\label{softV*Q*}
V^*_{\soft}(s) = \log \int\limits_{\A} \exp Q^*_{\soft}(s, a) \diff a
\end{equation}
\begin{proof}
Мы знаем, что оптимальная стратегия имеет вид $\pi^*(a \mid s) \HM= \frac{\exp{Q^*_{\soft}(s, a)}}{Z(s)}$, где
$$Z(s) \coloneqq \int\limits_{\A} \exp Q^*_{\soft}(s, a) \diff a$$
является нормировочной константой. Посчитаем энтропию такого распределения:
$$-\E_{\pi^*(a \mid s)} \log \pi^*(a \mid s) = \int\limits_{\A} \frac{\exp{Q^*_{\soft}(s, a)}}{Z(s)} \left(\log Z(s) - Q^*_{\soft}(s, a) \right) \diff a = \log Z(s) - \E_{\pi^*(a \mid s)} Q^*_{\soft}(s, a)$$

Подставим оптимальную стратегию в мягкое VQ уравнение \eqref{softVQ}, которое справедливо в том числе и для оптимальной стратегии:
\begin{align*}
V^*_{\soft}(s) &= \E_{\pi^*(a \mid s)} \left[ Q^*_{\soft}(s, a) - \log \pi^*(a \mid s) \right] = \\ &= \E_{\pi^*(a \mid s)} Q^*_{\soft}(s, a) - \E_{\pi^*(a \mid s)} Q^*_{\soft}(s, a) + \log Z(s) = \\ &= \log Z(s)
\end{align*}
Вспоминая определение $Z(s)$, получаем доказываемое.
\end{proof}
\end{theorem}

\begin{proposition}
$$Q^*_{\soft}(s, a) = r(s, a) + \gamma \E_{s'} V^*_{\soft}(s')$$
\end{proposition}

\begin{proposition}[Мягкое уравнение оптимальности Беллмана (soft Bellman optimality equations)]
\begin{equation}\label{softQ*Q*}
Q^*_{\soft}(s, a) = r(s, a) + \gamma \E_{s'} \log \int\limits_{\A} \exp Q^*_{\soft}(s', a') \diff a'
\end{equation}
\end{proposition}

\begin{theorem}
Оператор, стоящий в правой части уравнения \eqref{softQ*Q*}, является сжимающим с коэффициентом $\gamma$, и, следовательно, метод простой итерации решения этой системы уравнений сходится из любого начального приближения к единственной неподвижной точке.
\begin{proof}
Пусть даны две Q-функции, $Q_1, Q_2$, и пусть 
$$\rho(Q_1, Q_2) \coloneqq \max\limits_{s, a} | Q_1(s, a) \HM- Q_2(s, a) | \HM< \eps$$
Тогда:
$$
\log \int\limits_{\A} \exp Q_1(s, a) \diff a \le \log \int\limits_{\A} \exp (Q_2(s, a) + \eps) \diff a = \eps + \log \int\limits_{\A} \exp Q_2(s, a) \diff a
$$
Аналогично можно показать, что 
$$
\log \int\limits_{\A} \exp Q_1(s, a) \diff a \ge -\eps + \log \int\limits_{\A} \exp Q_2(s, a) \diff a
$$

Пусть $\B_{\soft}$ --- оператор, стоящий в правой части \eqref{softQ*Q*}. Тогда:
\begin{align*}
| [\B_{\soft} Q_1](s, a) - [\B_{\soft} Q_2](s, a) | &= \gamma | \E_{s'} \left( \log \int\limits_{\A} \exp Q_1(s', a') \diff a' - \log \int\limits_{\A} \exp Q_2(s', a') \diff a' \right) | \le \\ 
&\le \gamma \eps
\end{align*}
Таким образом, $\rho(\B_{\soft} Q_1, \B_{\soft} Q_2)$ уменьшилось по крайней мере в $\gamma$ раз по сравнению с $\rho(Q_1, Q_2)$.
\end{proof}
\end{theorem}

\begin{proposition}
Если $\pi(a \mid s) \propto \exp Q^{\pi}_{\soft}(s, a)$, то она оптимальна.
\begin{proof}
Q-функция такой стратегии удовлетворяет мягкому уравнению оптимальности Беллмана \eqref{softQ*Q*} и в силу единственности его решения совпадает с $Q^*_{\soft}(s, a)$.
\end{proof}
\end{proposition}

Теперь можно построить аналог DQN для задачи Maximum Entropy RL, называемым \emph{Soft Q-learning}. В нём нет отдельной модели актёра, и текущая стратегия просто полагается жадной по отношению к текущему критику. Это также можно интерпретировать как моделирование \emph{Soft Value Iteration} --- решение мягкого уравнения оптимальности \eqref{softQ*Q*}. Тогда таргет для перехода $\T \coloneqq (s, a, r, s')$ строится как
$$y(\T) \coloneqq r + \gamma \log \int\limits_{\A} \exp Q_{\theta^{-}}(s', a') \diff a',$$
где $\theta^{-}$ --- параметры таргет-сети. Интересно, что это соответствует <<учёту>> Больцмановской стратегии исследования внутри оценочной функции (нечто похожее мы делали в табличном алгоритме SARSA в разделе \ref{subsec:sarsa}). Однако, такой подход применим, как и в DQN, только для дискретных пространств действий, поскольку иначе стоящий в формуле интеграл мы аналитически не возьмём. Если мы попробуем завести отдельную нейросеть, чтобы приближать больцмановскую стратегию --- получим SAC.

И, наконец, обсудим аналоги Policy Gradient алгоритмов для Maximum Entropy RL сеттинга. Многошаговые уравнения Беллмана выводятся тривиально --- достаточно учесть в наградах энтропийные бонусы. Ну а аналог формулы policy gradient тоже можно легко угадать; мы же знаем, что если из формулы градиента удалить мат.ожидание по частотам посещения состояний текущей стратегии, мы получаем формулу policy improvement-а. Это свойство в Maximum Entropy RL сохраняется, то есть можно просто к формуле soft policy improvement \eqref{softpolicyimprov} добавить мат.ожидание по частотам посещения состояний. Выпишем, что получится для метода REINFORCE:

\begin{theorem}[Soft Policy Gradient]
$$\nabla_{\theta} J_{\soft}(\theta) = \frac{1}{1 - \gamma} \E_{s \sim d_{\pi_\theta}(s)} \E_{a \sim \pi_\theta(a \mid s)} \nabla_\theta \log \pi_\theta(a \mid s) Q^{\pi}_{\soft}(s, a) + \alpha \nabla_\theta \entropy(\pi_\theta(\cdot \mid s))$$
\begin{proof}
Аналогично доказательству в обычном случае.
\end{proof}
\end{theorem}

Другими словами, сеттинг Maximum Entropy RL <<объясняет>> добавку энтропийного лосса при обучении актёра, но здесь ваджно помнить, что поменялось определение Q-функции: мягкая оценочная функция учитывает получение в будущем энтропийных бонусов (поэтому добавка энтропийного регуляризатора в обычных policy gradient алгоритмах остаётся эвристикой).

\chapter{Model-based}\label{modelbasedchapter}

В этой главе мы рассмотрим ситуацию, когда нам дана или мы готовы учить модель динамики среды.

\section{Бандиты}\label{sec:bandistssection}

\subsection{Задача многоруких бандитов}

Рассмотрим сильно упрощённую задачу RL, где эпизод заканчивается после первого шага.

\begin{example}
В игровом зале стоят в ряд $|\A|$ игровых автоматов (<<одноруких бандитов>>): в каждый можно отправить монетку, дёрнуть за ручку, и тогда автомат с некоторой вероятностью выдаст вам приз. Вероятность приза фиксирована для каждого автомата (зависит от настроек, выставленных владельцами зала...), но может различаться между автоматами. 

\needspace{5\baselineskip}
\begin{wrapfigure}{r}{0.25\textwidth}
\vspace{-0.6cm}
\centering
\includegraphics[width=0.25\textwidth]{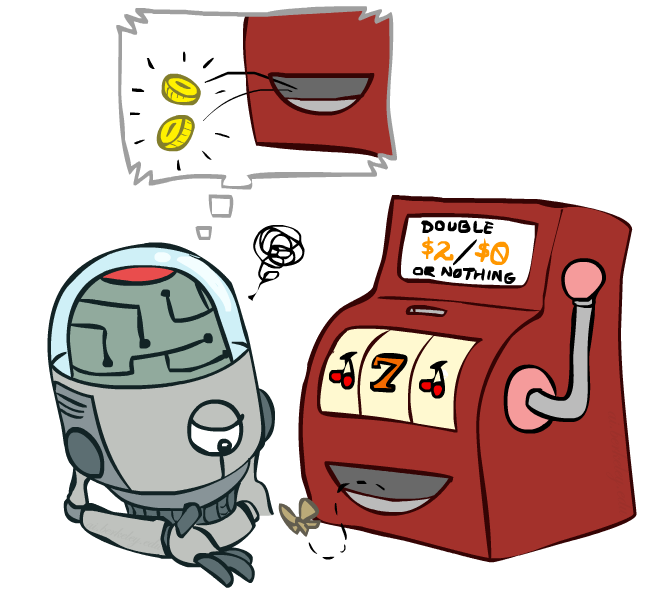}
\vspace{-0.3cm}
\end{wrapfigure}

Игроку (агенту) доступна только информация о том, получил ли он приз или нет, и всё, что он может --- это пробовать дёргать за ручки разных автоматов. Задача --- как можно быстрее выучить, какую ручку дёргать наиболее выгодно на основе накапливаемого опыта. Однако, без знания истинных вероятностей у игрока возникает вопрос: действительно ли тот автомат, который согласно его опыту выдавал приз чаще, является самым хорошим, или ему просто везло? Или с автоматом, который редко выдавал приз, просто было сыграно слишком мало игр? Налицо проблема exploration-exploitation дилеммы, которую в таком упрощённом виде легче теоретически анализировать.
\end{example}

Формально мы рассмотрим задачу RL для MDP, в котором нет состояний (<<есть только одно состояние>>). Агент выбирает действие при помощи стратегии $\pi(a)$ и получает какой-то приз из автомата с выбранным номером, после чего эпизод заканчивается. Формально мы будем интерпретировать это как стохастичную функцию награды, которая зависит только от выбранного игроком действия.

\begin{definition}
\emph{Многорукий бандит} (multi-armed bandit) --- это $A$ распределений на вещественной оси $p(r \mid a)$, $a \in \{1,2,3 \dots A\}$.
\end{definition}

Понятно, что в данной задаче Q-функция не зависит от стратегии и всегда равна
$$Q(a) = \E_{p(r \mid a)} r,$$
а оптимальная V-функция --- скаляр, равный Q-функции оптимального действия, <<наилучшего автомата>>:
$$V^* = \max\limits_a Q(a)$$

После каждого эпизода (т.е. после каждого шага) мы вольны улучшить свою стратегию, поэтому с каждым эпизодом наша стратегия меняется. Полный процесс, соответственно, задан следующим образом: на $k$-ом эпизоде мы сэмплируем $a_k \sim \pi_k(a)$ и наблюдаем награду $r_k \HM\sim p(r \HM\mid a_k)$, затем как-то меняем свою стратегию и получаем $\pi_{k+1}$. 

Допустим, всего проводится $T$ итераций обучения (играется $T$ эпизодов). Если бы мы знали оптимальную ручку, мы бы сыграли все $T$ эпизодов именно с ней, получив $TV^*$ награды. Но нам придётся сколько-то итераций потратить на поиск этой самой ручки --- на <<исследования>> --- и за счёт этого мы проиграем за каждое неоптимальное действие $a_k$ в среднем объёме $V^* \HM- Q(a_k)$.

\begin{definition}
\emph{Сожалением} (regret) за $T$ эпизодов называется величина
$$\Regret_T \coloneqq T V^* - \sum_{k=0}^T Q(a_k)$$
\end{definition}

Задача многорукого бандита --- поиск такой процедуры обучения $\pi$, которая минимизировала бы наши средние сожаления. В первую очередь здесь интересны асимптотически оптимальные стратегии с точки зрения сожалений: что $\lim\limits_{T \to \infty} \E \Regret_T$, где мат.ожидание берётся по $\pi_1, \pi_2 \dots$, ведёт себя в некотором смысле наилучшим возможным образом.

\subsection{Простое решение}

Единственный способ оценивать $Q(a)$ --- по Монте-Карло. На $k$-ом шаге мы можем оценить каждую Q-функцию как
$$Q_k(a) \coloneqq \frac{\sum_k r_k [a_k = a]}{\sum_k [a_k = a]}$$

Как мы обсуждали в разделе про экспоненциальное сглаживание, обновление такой Монте-Карло оценки $Q_k(a)$ через счётчики по сути и есть обучение: пусть $n_k(a)$ --- счётчик, сколько раз мы выбирали действие $a$ за все эпизоды, тогда:
\begin{align*}
Q_k(a_k) &= \left( 1 - \frac{1}{n_k(a_k)} \right)Q_{k-1}(a_k) + \frac{1}{n_k(a_k)} r_k = \\
&= Q_{k-1}(a_k) + \frac{1}{n_k(a_k)} \left( r_k - Q_{k-1}(a_k) \right)
\end{align*}

Весь вопрос заключается в том, как на основе этих оценок и знания, сколько раз какое действие было попробовано --- величины $n_k(a_k)$ --- выбирать действия. Понятно, что жадный выбор может завести нас в ситуацию, когда мы будем всё время играть с не самой оптимальной ручкой.

\begin{proposition}
При любом алгоритме скорость роста сожалений не более чем линейная: для некоторой константы $C$
$$\E \Regret_T \le CT$$
\begin{proof}
Рассмотрим худшую стратегию, которая всегда выбирает худшее действие с наибольшими сожалениями $\max\limits_a \left( V^* - Q(a) \right)$. Тогда сожаления за $T$ эпизодов будут равны
$$\Regret_T = T \max\limits_a \left( V^* - Q(a) \right)$$
Понятно, что это верхняя оценка на сожаления любого алгоритма, и $\max\limits_a \left( V^* - Q(a) \right)$ --- константа $C$ в линейной скорости роста.
\end{proof}
\end{proposition}

\begin{proposition}
При жадном выборе сожаления растут с линейной скоростью: для некоторой константы $C$
$$\E \Regret_T \ge CT$$
\begin{proof}
С некоторой вероятностью $\alpha$ жадный алгоритм застрянет на постоянном выборе автомата $a$ с ненулевым регретом $V^* - Q(a)$ и продолжит выбирать его до бесконечности. Тогда в среднем регрете появится слагаемое $T \alpha (V^* - Q(a))$, следовательно как минимум можно в качестве константы $C$ выбрать $\alpha (V^* - Q(a))$.
\end{proof}
\end{proposition}

Наивное решение --- решать проблему эксплорейшна при помощи $\eps$-жадной стратегии \ref{egreedy}.

\begin{algorithm}{Наивный бандит}
\textbf{Гиперпараметры:} $\eps(k)$ --- стратегия исследования

\vspace{0.3cm}
Инициализировать $Q_0(a)$ произвольно \\
Обнулить счётчики $n_0(a) \coloneqq 0$ \\
\textbf{На очередном шаге $k$:}
\begin{enumerate}
    \item выбрать $a_k$ случайно с вероятностью $\eps(k)$, иначе $a_k \coloneqq \argmax\limits_{a_k} Q_k(a_k)$
    \item увеличить счётчик $n_k(a) \coloneqq n_{k-1}(a) + [a_k = a]$
    \item пронаблюдать $r_k$
    \item обновить $Q_k(a) \coloneqq Q_{k-1}(a) + [a_k = a] \frac{1}{n_k(a)} \left( r_k - Q_{k-1}(a) \right)$
\end{enumerate}
\end{algorithm}

\begin{proposition}
При $\eps$-жадном выборе сожаления всё равно растут с линейной скоростью: для некоторой константы $C$
$$\E \Regret_T \ge CT$$
\begin{proof}
Поскольку на каждом шаге с вероятностью $\frac{\eps}{|\A|}$ мы выбираем некоторое неоптимальное действие $a$ с ненулевым регретом $V^* - Q(a)$, в среднем регрете появится слагаемое $T \frac{\eps}{|\A|} (V^* - Q(a))$, следовательно как минимум можно в качестве константы $C$ выбрать $\frac{\eps}{|\A|} (V^* - Q(a))$.
\end{proof}
\end{proposition}

\begin{remark}
В случае нестационарных бандитов распределения $p(r \HM\mid a)$ тоже меняются со временем и куда-то плывут. Наивное решение в такой ситуации можно модифицировать костыльми. Во-первых, $\eps$ нельзя уменьшать к нулю, необходимо постоянно пробовать различные действия, поэтому можно выставить $\eps$ в константу. Во-вторых, вместо счётчиков будем обновлять информацию о Q-функции через экспоненциальное сглаживание:
$$Q_k(a_k) = Q_{k-1}(a_k) + \alpha \left( r_k - Q_{k-1}(a_k) \right)$$
где $\alpha < 1$ --- константный гиперпараметр.
\end{remark}

\subsection{Теорема Лаи-Роббинса}

Можно ли придумать что-то лучшее, чем линейная скорость роста регрета? Перепишем регрет чуть-чуть по-другому:

\begin{proposition}
\begin{equation}\label{regretalt}
\Regret_T = \sum_{a} n_T(a) (V^* - Q(a))
\end{equation}
\end{proposition}

Понятно, что действия с большим регретом $V^* - Q(a_k)$ нужно выбирать как можно меньше, то есть асимптотически уменьшать счётчик $n_T(a)$. С другой стороны, ясно, что если распределения $p(r \mid a^*)$, где $a^*$ --- оптимальная ручка, и $p(r \mid a)$ для некоторого другого действия $a$ очень похожи, то различить эти два автомата будет тяжело (просто потому, что если распределения похожи, то и сэмплы из них будут очень похожи). Также ясно, что если распределения похожи, то есть надежда, что у них будут очень похожи средние, то есть регрет для такого действия $a$ будет маленьким. И наоборот: если регрет за действие большой, у автомата наверняка сильно другое распределение, нежели чем у оптимального автомата, и тогда их наверняка можно как-то просто различить по выборкам сэмплов. Оказывается, можно получить следующую нижнюю оценку на асимптотическое поведение любого алгоритма:

\begin{theorem}[Теорема Лаи-Роббинса (Lai-Robbins theorem)]
$$\E \Regret_T \ge \log T \sum_{a \ne a^*} \frac{V^* - Q(a)}{\KL(p(r \mid a) \parallel p(r \mid a^*))}$$
\begin{proof}[Без доказательства]
\end{proof}
\end{theorem}

Весьма сильное утверждение: оно говорит, что теоретически нельзя построить алгоритм с лучшим асимптотическим поведением регрета, чем $\log T$. Константа имеет понятный физический смысл: каждый автомат вносит свой вклад в эту константу, пропорциональный неоптимальности действия (числитель) и обратно пропорциональный похожести распределения награды в этом автомате с оптимальным автоматом (знаменатель). 

\begin{definition}
Будем говорить, что алгоритм \emph{асимптотически оптимален}, если он имеет логарифмическую скорость роста среднего регрета.
\end{definition}

\subsection{Upper Confidence Bound (UCB)}\label{subsec:ucb}

Попробуем поискать хорошую эвристику исследования среди алгоритмов следующего вида: на очередном шаге $k$ будем выбирать действие по следующей формуле:
\begin{equation}\label{ucb_exploration}
a_k \coloneqq \argmax_a \left[ Q_k(a) + U_k(a) \right],
\end{equation}
где $U_k(a)$ --- некоторая положительная добавка, имеющая смысл \emph{бонуса за исследования} (exploration bonus). То, что добавка должна быть положительна, следует из принципа \emph{оптимизма перед неопределённостью} (optimism in the face of uncertainty).

\begin{example}
Представьте, что вы идёте мимо пещеры, в которую вы никогда не заходили, и ваша оценка Q-функции для действия <<зайти в пещеру>> ниже, чем оценка других действий. Если алгоритм исследования таков, что ваше значение Q-функции занижается, то может возникнуть ситуация, что вы никогда не зайдёте в пещеру и не узнаете, что там. Если бы вы были уверены в идеальности ваших оценок, вы бы имели гарантии неоптимальности действия <<зайти в пещеру>>; но в реальности оценки никогда не бывают точны, и поэтому алгоритму исследований следует <<завышать>> те значения Q-функции, которые потенциально посчитаны ошибочно. При этом, чем больше неопределённость в их значениях, тем больше должно быть завышение.
\end{example}

Строить добавку нужно из соображений, вытекающих из формы регрета \eqref{regretalt}. Добавка должна быть маленькая, если данное действие было выбрано уже много раз, и наша неопределённость в знаниях о среднем значении $Q(a)$ достаточно точные, или же если нам кажется, что регрет для этого действия близок к нулю.

Идея \emph{upper confidence bounds} (UCB) алгоритмов следующая: давайте выбором $U_k(a)$ прогарантируем, что
$$Q(a) \le Q_k(a) + U_k(a)$$
с очень высокой вероятностью, близкой к единице, то есть, другими словами, построим \emph{доверительный интервал} (confidence interval) и возьмём его верхнюю границу. Такой $U_k(a)$ будет обратно пропорционален $n_k(a)$, ведь граница будет сжиматься к эмпирическому среднему. Жадный выбор $\argmax\limits_a Q_k(a)$, интуитивно, будет выбираться часто; его счётчик будет увеличиваться, и exploration bonus для него будет уменьшаться; тогда с достаточно маленькой вероятностью мы выберем при помощи формулы \eqref{ucb_exploration} действительно неоптимальное действие, для которого добавка $U_k(a)$ в силу его редкого выбора большая, и эта вероятность будет тем меньше, чем меньше эмпирическое среднее для этого неоптимального действия.

\begin{exampleBox}[righthand ratio=0.4, sidebyside, sidebyside align=center, lower separated=false]{}
На картинке справа изображены <<свечки>>: для каждого из четырёх автоматов указаны средние (оценки $Q_k(a)$), а также верхние и нижние границы доверительного интервала. Хотя автомат B кажется наиболее выгодным, мы уже достаточно часто играли с ним, поэтому его верхняя граница интервала доверия не так далека от среднего; а вот с автоматом A мы играли редко и поэтому на очередном шаге решим выбрать именно его.

\tcblower
\includegraphics[width=\textwidth]{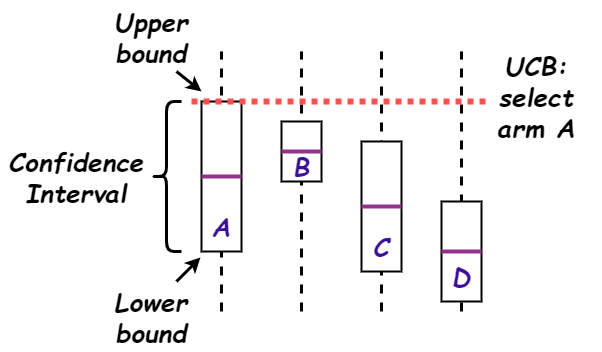}
\end{exampleBox}

Доверительный интервал можно построить при помощи следующего неравенства:

\begin{theorem}[Неравенство Хёфдинга]
\setcounter{footnote}{1}
Пусть $X_1 \dots X_n$ --- i.i.d выборка из распределения на домене\footnote{мы всегда предполагаем ограниченность наград; для удобства записи будем считать, что диапазон награды --- $[0, 1]$.} $[0, 1]$ с истинным средним $\mu$, $\hat{\mu} \coloneqq \frac{1}{N} \sum_i^N X_i$ --- выборочная оценка среднего. Тогда $\forall u > 0$:
$$\Prob(\mu \ge \hat{\mu} + u) \le e^{-2nu^2}$$
\begin{proof}[Без доказательства]
\end{proof}
\end{theorem}

Мы можем переформулировать эту теорему в терминах доверительного интервала:

\begin{proposition}
Для любого $\delta$ с вероятностью $1 - \delta$ истинное значение $Q(a)$ не превосходит $Q_k(a) \HM+ U_k(a)$, где:
$$U_k(a) \coloneqq \sqrt{\frac{-\ln \delta}{2n_k(a)}}$$
\begin{proof}
В силу неравенства Хёфдинга:
$$\Prob (Q(a) \ge Q_k(a) + U_k(a)) \le \exp^{-2 n_k(a) U_k(a)^2}$$
Возьмём заданное $\delta$ и прогарантируем, что $\exp^{-2 n_k(a) U_k(a)^2} \HM= \delta$. Разрешая это равенство относительно $U_k(a)$, получаем доказываемое.
\end{proof}
\end{proposition}

В качестве $\delta$ на $k$-ом шаге будем выбирать, например, $\frac{1}{k^c}$, где $c$ --- гиперпараметр. Таким образом, истинное значение будет всё с большей вероятностью оказываться внутри доверительного интервала. Итого в алгоритме UCB предлагается следующая формула выбора действия на очередном $k$-ом шаге:
\begin{equation}\label{UCB}
a_k = \argmax_a \left[ Q_k(a) + c \sqrt{\frac{\log k}{n_k(a)}} \right]
\end{equation}

Получается интерпретируемая формула: числитель $\sqrt{\log k}$ гарантирует, что если некоторое действие давно не выбиралось (счётчик $n_k(a)$ не меняется с увеличением числа итераций $k$), то добавленное слагаемое будет неограниченно расти и в какой-то момент промотивирует попробовать данное действие ещё раз.

\begin{theorem}
UCB-алгоритм \eqref{UCB} асимптотически оптимален.
\begin{proof}[Без доказательства]\end{proof}
\end{theorem}

\subsection{Сэмплирование Томпсона}

Мы далее попробуем учить степень нашей неопределённости в знаниях об $Q(a)$, моделируя всё неизвестное распределение $p(r \HM\mid a)$, а не только среднюю награду. Таким образом, мы перейдём к model-based подходу в RL, в котором динамику среды предлагается пытаться учить. В случае с бандитами под динамикой среды подразумеваются распределения награды для каждого автомата $p(r \HM\mid a)$, для чего мы воспользуемся стандартным байесовским подходом.

Введём предположение, что распределения наград $p(r \HM\mid a)$ принадлежат некоторому параметрическому семейству; для каждого автомата заданы значения параметров этого семейства $\theta_a \HM\in \Theta$, и награда генерируется из распределения $p(r \HM\mid \theta_a)$. Однако, истинные значения $\theta_a$ нам неизвестны.

Вместо оценки максимального правдоподобия на $\theta_a$ будем делать \emph{байесовский вывод}. Зададимся некоторым \emph{априорным распределением} (<<прайором>>) $p(\theta_a)$ для каждого действия $a$ (сюда мы можем в том числе положить какую-то дополнительную априорную информацию о том, какие автоматы заведомо лучше), и после очередного эпизода игры с автоматом $a$ с исходом $r_k$ будем обновлять распределение над $\theta_a$ по формуле Байеса на \emph{апостериорное распределение}:
\begin{equation*}
p(\theta_a) \leftarrow p(\theta_a \mid r_k) \propto p(r_k \mid \theta_a)p(\theta_a)
\end{equation*}

Таким образом мы аккумулируем всю информацию о значениях $\theta_a$, полученную из всех имеющихся сэмплов награды и исходного априорного распределения. Средний выигрыш из автомата $a$, соответственно равный $\E p(r \mid \theta_a)$, теперь для нас будет являться случайной величиной, поскольку $\theta_a$ --- случайное, с распределением $p(\theta_a)$.

\begin{example}[Bernoulli-бандиты]
Положим, что автоматы выдают только награду 0 или 1, то есть что истинное распределение есть распределение Бернулли с вероятностью $\theta_a$:
$$p(r \mid a) \coloneqq \operatorname{Bernoulli}(r \mid \theta_a) = \theta_a^{\mathbb{I}[r = 1]}(1 - \theta_a)^{\mathbb{I}[r = 0]} = \theta_a^{r}(1 - \theta_a)^{1 - r}$$

Априорное распределение зададим при помощи \href{https://en.wikipedia.org/wiki/Beta_distribution}{Бета-распределения}\footnote[1]{это распределение является \emph{сопряжённым} (conjugate) к Бернулли, что означает, что после применения формулы Байеса наше распределение останется в этом же семействе распределений.} распределения, а то есть:
$$p(\theta_a) \coloneqq \operatorname{Beta}(\theta_a \mid \alpha, \beta) \propto \theta_a^{\alpha - 1}(1 - \theta_a)^{\beta - 1}$$
где $\alpha$ и $\beta$ --- некоторые параметры (свои для каждого автомата $a$). Мы можем обновлять эти знания при помощи новых сэмплов $r_k \HM\sim p(r \mid a)$, проводя байесовский вывод. Применяем формулу Байеса:
$$p(\theta_a \mid r) \propto \underbrace{\theta_a^{r_k}(1 - \theta_a)^{1 - r_k}}_{p(r_k \mid \theta_a)}\underbrace{\theta_a^{\alpha - 1}(1 - \theta_a)^{\alpha - 1}}_{p(\theta_a)} = \operatorname{Beta}(\theta_a \mid \alpha + r_k, \beta + 1 - r_k)$$
Таким образом, для обновления параметров $\alpha, \beta$ достаточно увеличить $\alpha$ на $r_k$, а $\beta$ --- на $1 - r_k$.

При заданном $\theta_a$ мы можем посчитать среднее значение выигрыша, ценность автомата $a$: для случайной Бернуллиевской величины с параметром $\theta_a$ среднее, сообственно, совпадает с $\theta_a$: 
$$\hat{Q}(a) = \E p(r \mid \theta_a) = \theta_a$$

Тогда, имея $\alpha$ и $\beta$, мы всегда можем посчитать мат.ожидание выигрыша в нашей модели:
$$\E_{\theta_a} \hat{Q}(a) = \E_{\operatorname{Beta}(\theta_a \mid \alpha, \beta)} \theta_a = \frac{\alpha}{\alpha + \beta}$$
Как видно, оно не отличается от Монте-Карло оценки Q-функции: $\alpha$ имеет смысл <<успехов>>, когда Бернулливская величина выкинула нам единичку, а $\beta$ --- число неуспехов, нулей, и формула обновления этих параметров в точности соответствует подсчёту этих счётчиков; при выборе прайора $\alpha \HM= \beta \HM= 0$ мы получим в формуле мат.ожидания отношение числа успехов к общему числу попыток. 

Однако теперь у нас есть не только оценка среднего значения $\theta_a$, но и вероятности для каждого его возможного значения. На рисунке далее приведены $p(\theta_a)$ для разных значений $\alpha$ (<<успехов>>) и $\beta$ (<<неудач>>) при прайоре $\alpha \HM= \beta \HM= 1$:

\begin{center}
    \includegraphics[width=\textwidth]{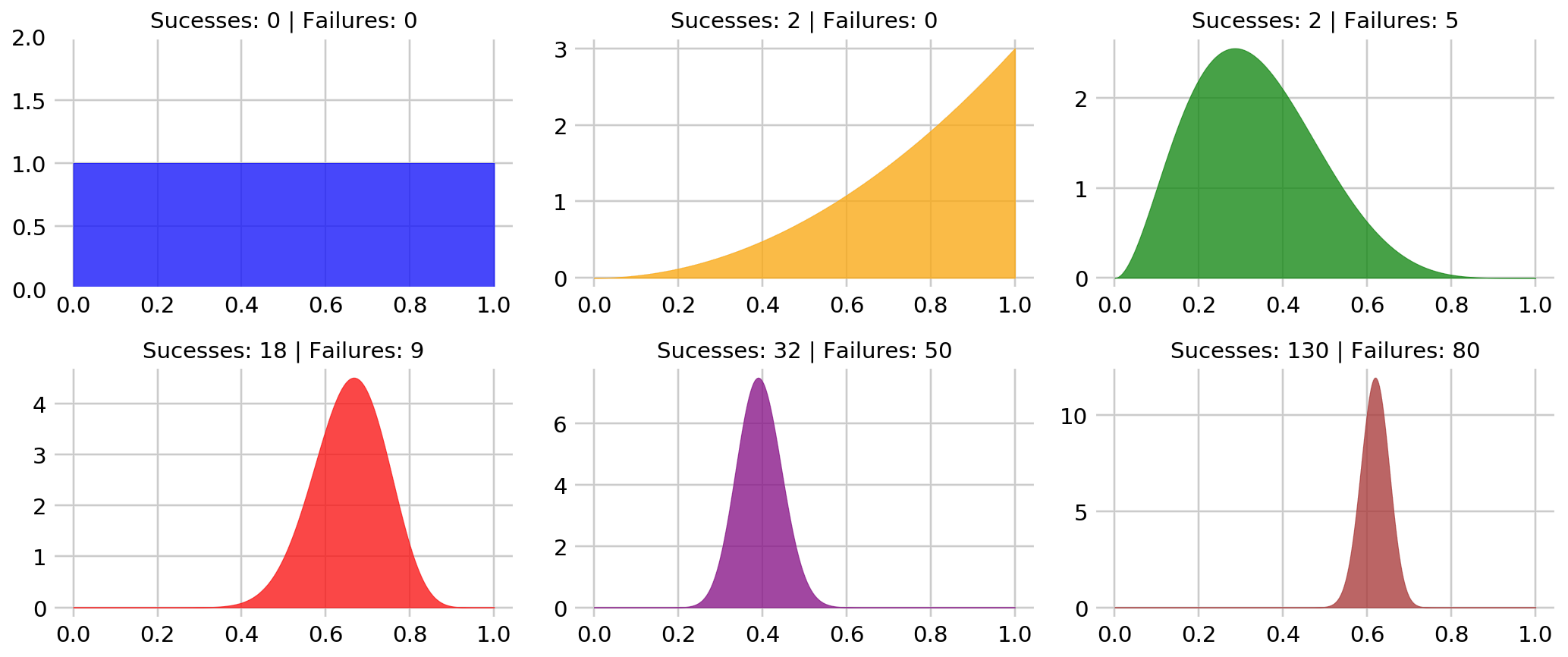}
\end{center}
\end{example}

Как пользоваться полученным апостериорным распределением? Мы можем реализовать идею, которая называется \emph{probability matching}. Согласно текущим $p(\theta_a)$ жадный выбор действия имеет вид
$$a \HM\coloneqq \argmax_a \E_{\theta_a \sim p(\theta_a)} \E p(r \mid \theta_a),$$
но теперь в нашей модели есть некоторая вероятность и того, что такое действие на самом деле неоптимально. Мы можем посчитать эту вероятность, то есть с учётом распределений $\theta_a$ посчитать вероятность, что хотя бы для одного другого автомата $\hat{a} \ne a$
$$\E p(r \mid \theta_{\hat{a}}) > \E p(r \mid \theta_{a}),$$
и посмотреть, с какой вероятностью мы ошибёмся. Аналогично можно для любого действия посчитать вероятность того, что на самом деле оптимальным является оно. Предлагается ровно с такими вероятностями, собственно, и выбирать автомат на очередном шаге.

\begin{definition}
\emph{Сэмплированием Томпсона} (Thompson Sampling) называется процедура, при котором на $k$-ом шаге при решении задачи многоруких бандитов действие $a$ выбирается с вероятностью того, что оно оптимально в рамках выученных моделей $p(\theta_a)$:
$$\pi(a) \coloneqq \Prob \left( \E p(r \mid \theta_a) = \max_{\hat{a}} \E p(r \mid \theta_{\hat{a}}) \right)$$
\end{definition}

Посчитать такие вероятности может быть нетривиально, зато легко из такого распределения сэмплировать (отсюда название\footnote{забавный факт: Томпсон вообще первый рассмотрел задачу многоруких бандитов, и в качестве эвристического решения предложил сэмплировать из моделей и брать аргмаксимум из сэмплов. То есть, можно сказать, сэмплирование Томпсона было первым придуманным способом решения задачи. А позже через почти век выяснилось, что это решение асимптотически оптимально.}). Действительно: просто для каждого действия $a$ засэмплируем $\theta_a$ из текущего $p(\theta_a)$, и, соответственно, выберем автомат с наибольшим мат.ожиданием согласно выпавшим $\theta_a$. 

\begin{example}
Допустим, мы храним информацию о распределениях $p(r \mid a)$ в Бета-распределениях, как в предыдущем примере. Для всех действий сэмплируем $\theta_a \sim \operatorname{Beta}(\theta_a \mid \alpha_a, \beta_a)$ при выученных параметрах $\alpha_a, \beta_a$, и считаем мат.ожидание соответствующего распределения Бернулли при выпавших параметрах $\theta_a$ --- для распределения Бернулли среднее совпадает с $\theta_a$ в точности. Таким образом, мы выберем автомат, для которого выпало наибольшее значение сэмпла $\theta_a$.
\end{example}

Идея в том, что для автоматов, которые мы пробовали часто, дисперсия апостериорного распределения будет маленькой и сэмпл почти всегда будет в точности равен текущей оценке Q-функции. Для автоматов, которые мы пробовали редко в силу маленького реварда, дисперсия будет большая и иногда сэмплы будут оказываться лучше текущего самого хорошего автомата, что заставит нас поэксплорить данный автомат ещё.

\begin{algorithm}{Beta-Bernoulli Бандит с сэмплированием Топмсона}
\textbf{Гиперпараметры:} $\alpha, \beta$ --- параметры прайора.

\vspace{0.3cm}
Обнулить счётчики $\alpha_0(a) \coloneqq \alpha, \beta_0(a) \coloneqq \beta$ \\
\textbf{На очередном шаге $k$:}
\begin{enumerate}
    \item сгенерировать $\theta_a \sim \operatorname{Beta}(\alpha_k(a), \beta_k(a))$
    \item выбрать $a_k \coloneqq \argmax\limits_{a} \theta_a$
    \item пронаблюдать $r_k \in \{0, 1\}$
    \item обновить счётчик $\alpha_k(a) \coloneqq \alpha_{k-1}(a) + [a_k = a][r_k = 1]$
    \item обновить счётчик $\beta_k(a) \coloneqq \beta_{k-1}(a) + [a_k = a][r_k = 0]$
\end{enumerate}
\end{algorithm}

\begin{theorem}
Сэмплирование Томпсона асимптотически оптимально.
\begin{proof}[Без доказательства]\end{proof}
\end{theorem}

\begin{example}
Поиграться с визуализацией этого алгоритма можно \href{https://learnforeverlearn.com/bandits/}{здесь}.
\end{example}

\subsection{Обобщение на табличные MDP}

Теория бандитов подсказывает, как можно <<правильно>> разрешать exploration-exploitation дилемму. К сожалению, эти идеи плохо масштабируются на произвольные MDP. Подход на основе UCB-счётчиков \eqref{UCB} можно эвристически моделировать в сложных MDP, что мы обсудим позже в главе \ref{subsec:intrinsic_motivation}. Сэмплирование Томпсона же может оказаться удобным способом справиться с некоторыми практическими задачами табличного RL.

Итак, пусть задано табличное MDP с конечным числом состояний и действий. Будем в байесовском смысле учить функцию переходов и функцию награду, то есть моделировать вероятности $p(r \HM \mid \theta_{s, a}) \HM \approx p(r \HM \mid s, a)$ и $p(s' \HM \mid \phi_{s, a}) \HM \approx p(s' \HM \mid s, a)$ и обновлять $\theta$ и $\phi$ по правилу Байеса. Для этого для каждой пары $s, a$ мы будем хранить <<текущее>> распределение $p(\theta_{s, a})$ и $p(\phi_{s, a})$, обусловленные на всю встретившуюся историю. Фактически это будут вероятности, с которыми истинное MDP живёт по тем или иным законам. После очередного перехода $(s, a, r, s')$ будем обновлять распределения над параметрами по формулам
\begin{equation*}
p(\theta_{s, a}) \leftarrow p(\theta_{s, a} \mid r) \propto p(r \mid \theta_{s, a})p(\theta_{s, a})
\end{equation*}
\begin{equation*}
p(\phi_{s, a}) \leftarrow p(\phi_{s, a} \mid s') \propto p(s' \mid \phi_{s, a})p(\phi_{s, a})
\end{equation*}

Далее на очередном шаге мы будем из такого <<распределения в пространстве MDP>> сэмплировать MDP: $\forall s, a \colon$
$$\theta_{s, a} \sim p(\theta_{s, a}), \quad \phi_{s, a} \sim p(\phi_{s, a})$$
Конкретные значения $\theta$ и $\phi$ теперь задают какое-то одно конкретное MDP, в котором мы можем любым алгоритмом динамического программирования (например, Value Iteration) найти оптимальную стратегию $\pi^*(a \mid s, \theta, \phi)$. Ей и будем пользоваться, чтобы сыграть, например, очередной эпизод игры и собирать новые данные для уточнения модели функции переходов и функции наград.

Такая процедура называется \emph{posterior sampling}, и является аналогом сэмплирования Томпсона для табличных MDP: мы будем выбирать действие с той вероятностью, с которой оно является оптимальным, обусловленную на всю историю нашего взаимодействия с настоящим MDP. Понятнее идея может стать на следующем примере.

\begin{exampleBox}[label=ex:onlineshortestpath]{Online Shortest Path}
Каждый день с утра агент едет на машине на работу. Карта представляет собой граф из нескольких вершин, некоторые из которых соединены рёбрами. Агент хочет кратчайшим маршрутом добираться из вершины <<дом>> в вершину <<работа>>. Такая задача решалась бы алгоритмом поиска кратчайшего пути в графе, если бы агенту были известны точные значения, сколько времени нужно на проезд по каждому ребру. Но это какие-то случайные величины, и каждый раз, когда агент проезжает по тому или иному ребру $e$, он тратит случайное время $r \HM \sim p(r \mid e)$. Как ездить на работу так, чтобы постепенно, с течением процесса суммарное время (<<регрет>>), которое агент тратил на поездки, было всё меньше и меньше?

В этой задаче вершины можно считать состояниями $s$, а рёбра можно считать парами $s, a$. Агент, допустим, знает карту, то есть знает, что <<функция переходов>> --- детерминированная, и знает $p(s' \HM \mid s, a)$, но не знает распределение наград. Поэтому в этой задаче достаточно учить лишь второе.

Итак, что говорит сэмплирование Томпсона: а давайте мы для каждого ребра $e$ заведём модель $p(r \HM \mid \theta_e)$, и будем в байесовском смысле учить $\theta_e$. Здесь $r$ --- вещественное, поэтому понадобится проводить байесовский вывод для непрерывных распределений\footnote[*]{например, если моделировать $p(r \mid \theta_e)$ в виде нормального распределения с настраиваемыми средним и дисперсией, $p(r \mid \theta_e) \HM \coloneqq \N(\mu_e, \sigma^2_e)$, то тогда в качестве сопряжённого распределения, как подсказывает  \href{https://en.wikipedia.org/wiki/Conjugate_prior\#When_likelihood_function_is_a_continuous_distribution}{википедия}, понадобится выбрать \href{https://en.wikipedia.org/wiki/Normal-gamma_distribution}{гамма-нормальное распределение}. Формулы вывода тогда можно посмотреть, например, \href{https://en.wikipedia.org/wiki/Normal-gamma_distribution\#Posterior_distribution_of_the_parameters}{здесь}.}. Допустим, мы выбрали какое-то семейство $p(r \mid \theta_e)$, завели какой-то прайор $p(\theta_e)$ и научились обновлять апостериорное распределение по формуле Байеса. 

Тогда перед очередным эпизодом мы сэмплируем $\theta_e \HM\sim p(\theta_e)$, и теперь у нас есть конкретное MDP с конкретной моделью наград. Мы можем посчитать среднее $\E p(r \mid \theta_e)$ и считать, что в среднем ровно столько времени потребуется для проезда по ребру $e$. Затем мы в графе можем найти кратчайший маршрут (поиск кратчайшего маршрута в общем-то является частным случаем общего алгоритма динамического программирования Value Iteration), и алгоритм предлагает отправиться именно по нему. Собранные награды за шаг --- затраченное время для проезда по каждому ребру --- используются для уточнения распределений $p(\theta_e)$, и модель всё улучшается.
\end{exampleBox}
\section{Обучаемые модели окружения}

\subsection{Планирование}

Понятно, что model-free алгоритмы, рассмотренные ранее, действуют не совсем так, как размышляет человек. Принимая решения, мы принимаем в учёт наши предсказания о том, в какое будущее выльется то или иное действие, и для построения предсказаний используем наши представления о том, как работает окружающий нас мир.

Итак, вернёмся к общей постановке задачи RL, и сейчас допустим, что нам доступен симулятор $p(s', r \mid s, a)$ для произвольных $s, a$ в произвольный момент времени.

В табличном сеттинге знание $p(s', r \mid s, a)$ позволяет применять алгоритмы динамического программирования Value Iteration и Policy Iteration. Однако, в табличном сеттинге мы умеем считать все интегралы --- мат.~ожидания по функции переходах; а в сложных средах знание функции переходов эти интегралы брать не позволяют.

Понятное дело, что, имея доступ к точному симулятору, мы можем, например, запустить какой-нибудь model-free алгоритм для обучения $\pi$, но нас интересует, можем ли мы непосредственно <<воспользоваться>> симулятором внутри самой стратегии, или даже придумать стратегию, использующую только симулятор. Раз мы полностью знаем правила игры, то, выбирая действие в некотором состоянии $s_0$, мы можем играть <<виртуальные>> игры, заглядывая в своё будущее. 

\begin{definition}
Для данного состояния $s_0$ набор действий $a_0, a_1, a_2, \dots$ будем называть \emph{планом} (plan).
\end{definition}

\needspace{5\baselineskip}
\begin{wrapfigure}{r}{0.4\textwidth}
\vspace{-0.5cm}
\centering
\includegraphics[width=0.4\textwidth]{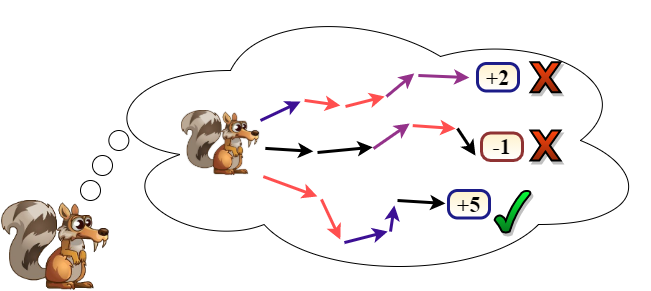}
\vspace{-0.5cm}
\end{wrapfigure}

Итак, допустим мы сидим в некотором $s_0$ и хотим найти хорошее действие $a_0$. Мы можем выбрать какое-нибудь $a_0$, засэмплировать $s_1$, выбрать как-то $a_1$, и так построить какой-нибудь план, для которого у нас будет сэмпл будущей награды. Можем так сделать, скажем, много раз. Нас ждёт две проблемы: у нас не получится рассмотреть всевозможные варианты будущего, и мы не сможем посчитать интеграл средней награды для одного плана; мы можем получить лишь сэмпл или Монте-Карло оценку по нескольким сэмплам.

\begin{example}
Какой типичный способ построения ИИ для ботов в играх: для текущего состояния $s$, в котором нужно выбрать действие, рассматривается несколько вариантов дальнейших действий на ближайшие шагов сто (при этом желательно заведомо неоптимальные варианты отсекать, чтобы сократить перебор); для каждого плана проводится симуляция (или несколько, если правила игры стохастичны и время позволяет). Играть до конца эпизода обычно дороговато, поэтому состояние $\hat{s}$, на котором симуляцию игры прервали, придётся оценивать при помощи какой-то эвристичной оценки $V^*(\hat{s})$.
\end{example}

Подобные алгоритмы могут <<спланировать>> свои будущие действия, а не выучить непосредственно зависимость $\pi(a \HM\mid s)$ (<<обучить стратегию>>).

\begin{definition}
Задача
\begin{equation}\label{planning}
\argmax_{a_0, a_1, a_2 \dots} \E_{\Traj \mid s_0, a_0, a_1, a_2 \dots} R(\Traj)
\end{equation}
при доступной функции переходов $p(s' \mid s, a)$ и функции награды называется \emph{планированием} (planning).
\end{definition}

В такой концепции понятия обучения нет. Для детерминированных MDP задача планирования по определению даёт оптимальный план, если задача \eqref{planning} решена точно, но для стохастичных MDP это неверно: 

\begin{theoremBox}[label=th:planningnotoptimal]{Неоптимальность планирования}
Пусть MDP стохастично и $a_0, a_1, a_2 \dots$ --- точное решение задачи \eqref{planning} для $s_0$. Тогда может быть такое, что действие $a_0$ неоптимально ($Q^*(s_0, a_0) < Q^*(s_0, a)$ для некоторого $a \ne a_0$).

\needspace{5\baselineskip}
\begin{wrapfigure}{r}{0.3\textwidth}
\centering
\includegraphics[width=0.3\textwidth]{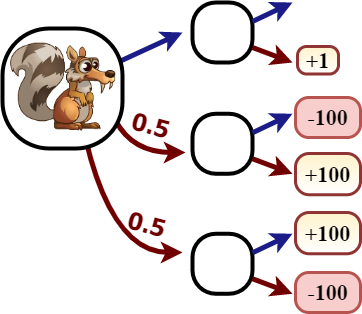}
\vspace{-0.3cm}
\end{wrapfigure}
\beginproof
Посмотрим, чему равна средняя награда для каждого плана для MDP с рисунка. 

\begin{adjustwidth}{0.2\textwidth}{}
\begin{tabular}{cc}
План & Ценность плана \\
 \hline
(\colorsquare{ChadBlue}, \colorsquare{ChadRed})      & +1      \\ 
(\colorsquare{ChadBlue}, \colorsquare{ChadBlue})     & 0       \\
(\colorsquare{ChadRed}, \colorsquare{ChadRed})       & $\frac{1}{2}(+100 - 100) = 0$       \\
(\colorsquare{ChadRed}, \colorsquare{ChadBlue})      & $\frac{1}{2}(+100 - 100) = 0$       \\
\end{tabular}
\end{adjustwidth}

Значит, будет выбран <<безопасный>> вариант в виде плана (\colorsquare{ChadBlue}, \colorsquare{ChadRed}). Конечно же, оптимально выбрать в начале действие \colorsquare{ChadRed} и далее в зависимости от состояния выбрать то действие, которое даст +100, таким образом на самом деле $V^*(s_0) \HM= +100$.   \qed
\end{theoremBox}


Другими словами, планирование --- разумный подход в детерминированных средах, когда будущее предопределено. Поэтому большинство планировщиков, которые мы будем обсуждать далее, заточены именно под детерминированный случай. Важно, что даже в такой ситуации решать задачу \eqref{planning} точно, конечно же, не получится, и наше решение будет лишь приближённым. 

По этим двум причинам планировщик иметь смысл перезапускать на каждом шаге. Обычно алгоритм решения задачи планирования используется как просто <<очень вычислительно дорогая>> стратегия: сидя в $s_0$, алгоритм планирует совершить действия $a_0, a_1, a_2 \dots$; в реальной среде совершается лишь действие $a_0$, после чего для $s_1 \sim p(s_1 \mid s_0, a_0)$ алгоритм планирования запускается ещё раз (важно, что тут он может использовать какую-то информацию с прошлого процесса планирования) для определения $a_1$ и так далее. Такая процедура называется \emph{планированием с обратной связью}. Планировщик, в частности, может понимать, что он не рассмотрел всевозможные исходы будущего, и из этих соображений выдать вероятностное распределение $\pi(a_0 \HM\mid s_0)$; тогда стратегия, полученная при помощи такого планировщика, будет стохастична.

\begin{example}[Мета-эвристики как планировщики]
Для решения \eqref{planning} можно использовать, например, случайный поиск. Сидя в $s_0$, засэмплируем 100 случайных планов и при помощи симулятора посчитаем для каждого из них сэмплы $R(\Traj)$; выберем то $a_0$, для которого нашлась траектория с максимальным сэмплом reward-to-go. Если среда детерминирована, то $a_0$ гарантировано позволяет получить такую награду; что не означает, что мы нашли наилучший план, и оптимальным может оказаться другое действие. Если среда стохастична, то даже гарантий получить <<найденный>> reward-to-go нет, поскольку это лишь сэмпл и нам могло в симуляции просто повезти. Выбранное $a_0$ отправляется в среду; для сэмпла $s_1$ из реальной среды процесс планирования запускается заново. 


\end{example}

То есть, обучения в такой концепции нет, но каждое принятие решения алгоритмом (каждый <<запуск стратегии>>) --- это целый процесс оптимизации, что обычно очень дорого.

\subsection{Модели мира (World Models)}

В реальной жизни в рамках нашей общей постановки задачи функция переходов и награды $p(s', r \mid s, a)$ нам неизвестна. Тем не менее, у нас есть опция как-то попытаться её выучить и как-то использовать для улучшения наших алгоритмов. Открывается два вопроса: как учить и как использовать.

\begin{definition}
\emph{Моделью мира} (world model) называется любая модель, явно или неявно обучающаяся модели динамики среды.
\end{definition}

Под <<явно или неявно>> подразумевается, что это не обязательно должна быть функция, по состоянию и действию возвращающая следующее состояние, хотя, конечно, это самая стандартная опция.

\begin{remark}
Модель мира также может сжимать описание состояний в некоторое латентное пространство, которое далее может использовать стратегия. Ещё можно нарушить end-to-end парадигму и в целом отдельно выучить автокодировщик, который сожмёт описание состояний в некоторый компактный векторочек; RL же применять только для обучения самой стратегии, которая будет принимать на вход такое компактное описание состояний. \href{https://worldmodels.github.io/assets/mp4/carracing_vae_compare.mp4}{Пример подобного сжатия}.
\end{remark}

Простейший способ <<использовать>> выученную модель --- взять в качестве стратегии какой-нибудь планировщик, который будет работать с обученной аппроксимацией модели мира будто это истинная модель. Нюанс такого подхода заключается в том, что планировщик по определению <<хорош>> только если модель мира точна. Если модель мира обучается только на тех примерах, которые мы встречаем в нашем опыте, то мы рискуем познать только тот мир, который обозревает наша текущая стратегия. До <<генеральной совокупности>> подобному опыту может быть далеко, и поэтому необходимо постоянно дообучать модель мира после каждого улучшения стратегии.

\begin{algorithm}[label = alg:generalmodelbased]{Общая схема Model-based подхода}
\textbf{Гиперпараметры:} Планировщик, модель мира.

\vspace{0.3cm}
Проинициализировать стратегию $\pi_0(a \mid s)$ случайно. \\
Проинициализировать модель мира случайно. \\
Проинициализировать датасет пустым множеством. \\
\textbf{На $k$-ом шаге:}
\begin{enumerate}
    \item Повзаимодействовать со средой при помощи $\pi_k$, добавив встреченные переходы $\{s, a, r, s'\}$ в датасет.
    \item Провести дообучение модели мира на собранном датасете:
    \item Получить $\pi_k$ из планировщика, используя текущую модель мира.
\end{enumerate}
\end{algorithm}

Под словами <<получить $\pi_k$ из планировщика>> здесь понимается, что далее в качестве стратегии $\pi_k$ используется запуск планировщика, который кушает текущее состояние, рассматривает какие-то планы и выдаёт первое действие наилучшего плана $a_0$ или какое-то распределение в пространстве действий, <<вероятности того, что $a_0$ соответствует оптимальному плану>>.

\subsection{Модель прямой динамики}

\begin{definition}
Модель функции переходов $p_\theta(s' \mid s, a)$ называется \emph{моделью прямой динамики} (forward dynamics model).
\end{definition}

\needspace{5\baselineskip}
\begin{wrapfigure}{r}{0.6\textwidth}
\vspace{-0.5cm}
\centering
\includegraphics[width=0.6\textwidth]{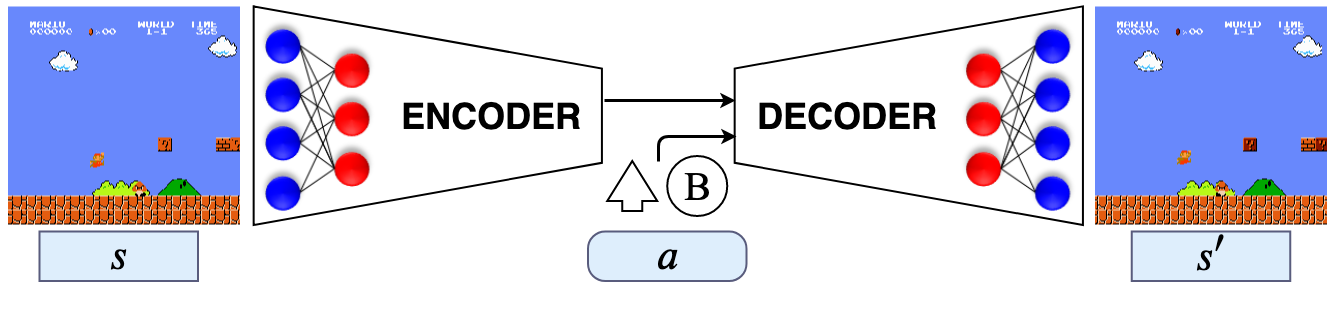}
\vspace{-0.5cm}
\end{wrapfigure}

Учить генеративную модель может быть дороговато, и простым удешевлением является обучение детерминированного приближения $s' \approx f_\theta(s, a)$. Это можно делать по любым доступным траекториям, собранным любым способом:
$$\sum_{s, a, s'} \|f_\theta(s, a) - s'\|^2 \to \min_\theta$$
$$\sum_{s, a, r} \left( r_\psi(s, a) - r \right)^2 \to \min_\psi$$

Этот лобовой подход многим плох. В сложных средах описание состояний обычно содержит огромное количество никак не связанной с задачами агента информацией; например, декоративные элементы в видеоиграх. Обучение подобной $f_\theta$ сопряжено с бессмысленным изучением этой информации. Наконец, если состояния $s$ --- картинки, то в текущей формуле сгенерированное изображение сравнивается с реальным $s'$ по l2-метрике; для изображений это не сильно осмысленно, конечно, и нужно придумывать что-то другое.


\begin{remark}
Иногда в задачах робототехники можно в модель прямой динамики внести <<inductive bias>>, используя знания о том, что модель динамики представляет собой какое-то дифференциальное уравнение. Тогда могут искаться параметры функций, стоящих внутри уравнений связи, а в качестве итоговой модели использоваться какая-то дискретизация <<обученного>> диффура.
\end{remark}

\subsection{Сновидения}

Наличие модели прямой динамики $p_\theta(s', r \mid s, a)$ позволяет нам при помощи текущей стратегии $\pi$ генерировать траектории, используя полностью наши <<внутренние модели>> и никак не используя реальную внешнюю среду. Это означает, что в случае обучаемой модели динамики мы теоретически можем начать симулировать опыт при помощи, условно, нашей генеративной модели, и обучать на нём model-free алгоритм.

\begin{definition}
\emph{Сновидениями} (dreaming) называется обучение агента на опыте, собранном при помощи приближения динамики среды $p_\theta(s', r \mid s, a)$.
\end{definition}

Далеко мы так, конечно, не уедем, поскольку наша генеративная модель вряд ли будет идеальна: регулярно нужно <<просыпаться>> и отправляться в реальную среду собирать сэмплы для улучшения как минимум модели мира. Тем не менее, использование снов может существенно повысить sample efficiency алгоритмов: потребуется меньше реальных взаимодействий со средой, и большую часть процесса обучения самой стратегии можно переложить на сны. Цена --- в вычислительной неэффективности: траекторий во сне из-за отличий от реальных потребуется генерировать довольно много. То есть, хотя шагов взаимодействия с настоящей средой станет меньше, время обучения алгоритма (wall-clock time) наоборот увеличится. 


\section{Планирование для дискретного управления}\label{mctssection}

\subsection{Monte-Carlo Tree Search (MCTS)}

Попробуем придумать алгоритм планирования \eqref{planning} в предположении идеального симулятора для MDP с дискретным пространством действий.

Понятно, что для детерминированных MDP с функцией переходов $s' \HM= f(s, a)$, чисто теоретически, мы можем просто построить полное дерево игры. Заведём дерево, в котором каждому узлу будет соответствовать состояние, дугам --- действия, и скажем, что узел, соответствующий $s'$ --- потомок $s$ по дуге $a$, если $s' \HM= f(s, a)$.

У нас есть две проблемы. Во-первых, в сложных MDP такое дерево экспоненциально большое и целиком мы его не построим. Во-вторых, наши MDP, вообще говоря, стохастичны. Мы могли бы ввести много рёбер из $s$, соответствующих одному и тому же действию $a$ и <<подписать>> над каждым вероятности перехода $p(s' \HM\mid s, a)$, однако наша функция переходов $p(s' \HM\mid s, a)$ может быть произвольным распределением, да и вообще пространства состояний $\St$ могут быть в том числе континуальные или очень большие. Поэтому давайте введём понятие дерева чуть-чуть по-другому\footnote{распространены объяснения MCTS в контексте детерминированных MDP, и поэтому про узлы постоянно говорят как про состояния; однако на самом деле ход рассуждений в принципе обобщается на стохастические MDP с тем замечанием, что в силу утверждения \ref{th:planningnotoptimal} в стохастичных средах планирование \eqref{planning} неоптимально.}:

\begin{definition}
\emph{Деревом MDP} с корнем в состоянии $s_0$ будем называть дерево, где каждой дуге соответствует действие $a$; узел на $t$-ом уровне дереве соответствуют плану $a_0, a_1 \dots a_{t-1}$, соответствующему пути от корня.
\end{definition}

Нам нужно продумать эвристики, как сокращать перебор поиска по такому дереву, и как его строить в ситуациях, когда эта задача экспоненциально сложная и у нас ограничены вычислительные ресурсы. Мы сейчас рассмотрим некоторую очень общую схему, как можно такую эвристику строить.

Будем строить дерево MDP с корнем в $s_0$ постепенно. Изначально, на первой итерации, наше дерево пусто --- есть только корень. Узлы дерева будем обозначать красивым символом $\aleph$. Дуги тогда однозначно задаются парой $\aleph, a$, где $\aleph$ --- узел дерева, откуда исходит дуга, $a$ --- действие, которому дуга соответствует. Для каждой дуги $(\aleph, a)$ мы будем также хранить некоторую вспомогательную информацию; самый типичный вариант --- это счётчики прохождения по данной дуге $n(\aleph, a)$ и ценность\footnote{хотя принято обозначать ценности через $V$ или $Q$, это не ценности каких-то состояний: в контексте стохастичных MDP это ценность плана, который привёл нас в это состояние; если угодно, можно обозначить этот счётчик как $Q(s_0, a_0, a_1, a_2 \dots a_t) \HM= \E_{\Traj \mid s_0, a_0, a_1 \dots a_t} R(\Traj)$. Но в детерминированных средах они, конечно же, будут соответствовать оценочным функциям в стандартном понимании.} $Q(\aleph, a)$, скаляр.

Один шаг процедуры \emph{Monte-Carlo Tree Search} (MCTS) заключается в проведении четырёх этапов: Selection, Expansion, Simulation и Update. Введём их по порядку.

\begin{definition}
На шаге \emph{Selection} стартуем в корне, которому соответствует текущее состояние в реальной игре $s_0$. При помощи некоторой стратегии, которую будем называть \emph{TreePolicy}, и которая использует данные на дугах, выбираем действие $a_0$; спускаемся по дереву на уровень глубже по дуге, соответствующей $a_0$, и сэмплируем $s_1, r_1$ из $p(s_1, r_1 \HM\mid s_0, a_0)$. Повторяем процедуру до тех пор, пока не попадём в некоторый лист дерева на уровне $t$. Мы запоминаем (знаем) всю траекторию от корня до листа, то есть фактически выбираем таким образом начало некоторого плана $a_0, a_1 \HM\dots a_{t-1}$ для рассмотрения, и заодно по Монте-Карло генерируем начало возможной траектории, соответствующей этому плану; в результате этой процедуры, мы попадаем в некоторый лист нашего текущего дерева и заодно симулируем для него состояние $s_t$.
\end{definition}

На этом шаге задачей TreePolicy является выбор того плана, для которого мы будем дальше строить дерево. То есть, допустим, мы сидим в некотором узле $\aleph$, не являющемся листом, из которого исходит $|A|$ дуг, и для каждого возможного пути знаем приблизительную оценку качества $Q(\aleph, a)$ и то, сколько раз мы уже пробовали ходить по данной ветке --- $n(\aleph, a)$. Задача этой части эвристики --- найти самый перспективный путь в дереве, используя эти статистики. С одной стороны, нужно искать оптимальный план в той ветке, где оценка качества велика; однако, если по каким-то веткам мы ходили редко, то высока вероятность, что там нам не повезло с сэмплом награды или мы просто недостаточно раскрутили там дерево. Налицо проблема многорукого бандита (см. раздел \ref{sec:bandistssection}); поэтому типичные TreePolicy вдохновлены\footnote{тем не менее, применение бандитов здесь --- тоже эвристика, хотя бы потому, что из-за постепенного раскрытия дерева вдоль каждой ветки <<распределение $p(r \mid a)$>>, из которого приходит дальнейший reward-to-go, меняется.} решениями этой задачи.

\begin{example}[Пример TreePolicy] Воспользуемся UCB-бандитом \eqref{UCB}, формула которого принимает следующий вид:
$$a \coloneqq \argmax_a \left[ Q(\aleph, a) + C \sqrt{\frac{\log n(\aleph)}{n(\aleph, a)}}\right]$$
где $n(\aleph) \coloneqq \sum\limits_a n(\aleph, a)$ --- это счётчик посещений узла, $C$ --- гиперпараметр. Действительно: $n(\aleph)$ --- сколько <<эпизодов игры>> мы провели с этим многоруким бандитом, а $n(\aleph, a)$ --- сколько раз выбирали в этом месте действие $a$. 
\end{example}

\begin{definition}
    На шаге \emph{Expansion} в выбранном на предыдущем этапе листе создаём для каждого действия $a_t \HM\in \A$ по новому листу, соответствующему выбору этого действия: таким образом, мы расширяем дерево вдоль выбранной ветки <<на один шаг вперёд>>.
\end{definition}

\begin{remark}
Если число действий $|\A|$ велико, то имеет смысл определять поднабор действий случайным образом и создавать листы только для него. Тогда этап Selection заканчивается как только, например, TreePolicy сэмплирует действие, для которого ещё не существует дуги.
\end{remark}

\begin{definition}
    Шаг \emph{Simulation} или \emph{Evaluation} заключается в построении некоторой эвристичной оценки reward-to-go для каждого нового построенного на предыдущем шаге листа.
\end{definition}

\needspace{5\baselineskip}
\begin{wrapfigure}{r}{0.5\textwidth}
\vspace{-0.3cm}
\centering
\includegraphics[width=0.5\textwidth]{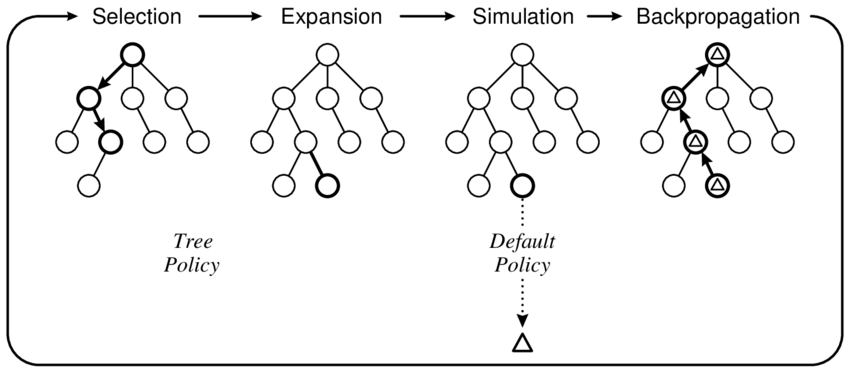}
\vspace{-0.3cm}
\end{wrapfigure}

В этом месте, конечно, можно использовать какие-нибудь handcrafted-эвристики, оценивающие очень примерно reward-to-go до конца эпизода (поэтому у этого шага есть второе название Evaluation), но универсальный вариант получить её --- это для каждого $a_t$ просимулировать игру (или несколько) из $s_t, a_t$ до конца эпизода, выбирая действия при помощи некоторой другой стратегии, которую назовём \emph{DefaultPolicy}, и которая уже не может использовать никакой информации в узлах дерева (эти узлы мы ещё просто-напросто не построили). Итого, мы получаем сэмплы reward-to-go для $|\A|$ планов, которые начинаются с $a_0, a_1 \dots a_{t-1}$, на $t$-ом месте действие варьируется, и для каждого варианта у нас есть одна (или несколько) новая Монте-Карло оценка награды.

\begin{example}
Пример типичной DefaultPolicy --- это банально случайная стратегия. 
\end{example}

\begin{definition}
    Шаг \emph{Update} или \emph{Backpropagation}: обновление счётчиков и оценок во всех дугах дерева, по которым мы проходили на данном шаге при помощи полученных на шаге симуляции новых Монте-Карло оценок.
\end{definition}

\begin{example}
Процесс обновления счётчиков прост: в свежераскрытом листе для новых дуг инициализируем $n(\aleph, a) \coloneqq 1, Q(\aleph, a) \coloneqq \hat{V}_a$, где $\hat{V}_a$ --- reward-to-go, полученный на шаге Simulation для действия $a$. Далее, рассмотрим пару $d, a$ где-то в рассмотренной ветке дерева; во-первых, увеличиваем счётчик посещения этой дуги на единицу. Для обновления оценки $Q$ просто учтём новый сэмпл Монте-Карло оценки: мы симулировали награды за шаг, пока шли по дереву, и поэтому можем посчитать reward-to-go до момента прихода в лист. В листе же можно считать, что мы выбирали действие случайным образом, и поэтому усредним все $\hat{V}_a$ по действиям. Таким образом, если $\hat{V}$ --- полученный новый сэмпл reward-to-go, $Q(\aleph, a)$ обновляется как:
$$Q(\aleph, a) \leftarrow Q(\aleph, a) + \frac{1}{n(\aleph, a)}(\hat{V} - Q(\aleph, a))$$
\end{example}

Интуиция, почему это работает: чем больше мы строим дерево, тем меньше зависим от эвристики для этапа Simulation; в пределе после бесконечного числа итераций мы условно построим дерево игры целиком. Чем меньше эта зависимость, тем чаще TreePolicy выбирает хорошие действия. Эвристика из Simulation же лишь указывает на те узлы, куда MCTS быстрее отправится детальнее строить подробное дерево разбора игры, и чем эта эвристика лучше, тем удачнее пройдёт ограниченный перебор.

Вполне можно считать, что расчёт $Q(\aleph, a)$ при помощи Монте-Карло оценок --- это Policy Evaluation, а поиск по дереву при помощи многоруких бандитов --- Policy Improvement.

\subsection{Применение MCTS}

MCTS можно <<предобучить>> перед использованием: проводим много шагов MCTS-процедуры, считая, что корню соответствует $s_0$ (если распределение на начальное состояние стохастично, придётся сэмплировать $s_0$: но тогда, в силу теоремы \ref{th:planningnotoptimal}, мы будем искать план, хороший в среднем для возможных исходов). На каждом шаге MCTS-процедуры ровно один лист в дереве <<раскрывается>> и для каждого пути симулируется одна или несколько игр до конца эпизодов. Полностью дерево для всего MDP мы скорее всего не построим, и однажды придётся остановиться.

Дальше мы хотим нашу стратегию отправить <<играть>> с реальной средой, и тут возможны варианты. Во-первых, мы можем дерево больше не трогать и просто использовать, например, TreePolicy (возможно, с <<выключенным>> эксплорейшном): а то есть, изначально нам дали $s_0$, которое, условно, соответствует корню. Мы выбираем действие при помощи TreePolicy и отправляем его в реальную среду; после этого мы можем в нашем дереве спустится по дуге, соответствующей выбранному действию, и на следующем шаге воспользоваться TreePolicy в этом узле. Однако, во время <<использования>> подобной стратегии нам могут встретиться состояния, для которых в дереве ещё нет узлов; тогда либо придётся выбирать действия случайно, либо проводить новые шаги MCTS-процедуры.

По этой причине, так обычно не делают; считается, что в общем случае, MCTS не строит дерево игры один раз и потом использует его во всех играх, а сама по себе является стратегией выбора действия в данном состоянии: просто долгой и тяжёлой в связи с переборной природой. То есть, перед каждой отправкой итогового действия в среду, нужно провести <<ещё парочку>> итераций MCTS-процедуры. Это означает, что, во-первых, можно вообще предобучения не проводить, а перед каждым выбором действия для реальной среды делать, там, 1000 шагов MCTS-процедуры, а во-вторых, после очередного реального шага, спускаясь на один узел вниз в дереве, мы можем оставлять только поддерево того узла, в который пришли, для оптимизации (наверх по дереву мы никогда не идём --- прошлое уже не изменить).

Итоговое действие для реальной среды не обязательно выбирать при помощи TreePolicy (в нём есть учёт эксплорейшна, который при использовании стратегии нам не нужен). Распространённый вариант следующий: выбирать с большей вероятностью те действия, которые MCTS исследовал чаще всего в ходе всей процедуры:
$$\pi(a_0 \mid s_0) \propto n(\aleph_0, a_0)^T$$
где $T$ --- температура, очередной гиперпараметр.

\begin{remark}
При <<использовании>> стратегии в бою обычно используют жёсткий вариант: $a \coloneqq \argmax\limits_a n(\aleph_0, a)$.
\end{remark}

\begin{algorithm}[label=alg:mcts]{Стратегия MCTS (одна из вариаций)}
\textbf{Вход:} $s_0$ --- текущее состояние реальной среды, $p(s', r \mid s, a)$ --- симулятор, $C$ --- гиперпараметр UCB-бандита, $\aleph_0$ --- корень текущего дерева в памяти с хранением счётков $n(\aleph, a)$ и оценок $Q(\aleph, a)$ в дугах, $K$ --- число шагов, $T$ --- температура.

\vspace{0.3cm}
\textbf{На $k$-ом шаге из $K$:}
\begin{enumerate}
    \item садимся в корень: $\aleph \coloneqq \aleph_0, s \coloneqq s_0$.
    \item инициализируем траекторию симуляции: $\Traj \coloneqq (s_0)$
    \item \textbf{пока $\aleph$ --- не лист:}
        \begin{itemize}
        \item выбираем ветку, куда пойти:
        $$n(\aleph) \coloneqq \sum_a n(\aleph, a)$$
        $$a \coloneqq \argmax\limits_a \left[ Q(\aleph, a) + C \sqrt{\frac{\log n(\aleph)}{n(\aleph, a)}}\right]$$
        \item генерируем $s', r \sim p(s', r \mid s, a)$
        \item сохраняем $a, r, s'$ в $\Traj$
        \item спускаемся по дереву: $\aleph \leftarrow \operatorname{child}(\aleph, a)$, $s \leftarrow s'$
        \end{itemize}
    \item \textbf{для каждого $a \in \A$:}
        \begin{itemize}
        \item создаём узел $\hat{\aleph}$ --- ребёнка $\aleph$ с дугой для действия $a$
        \item симулируем $\Traj_a \sim \pi^{\mathrm{random}} \mid s, a$, где $\pi^{\mathrm{random}}$ --- случайная стратегия
        \item инициализируем $n(\hat{\aleph}, a) \coloneqq 1, Q(\hat{\aleph}, a) \coloneqq R(\Traj_a)$
        \end{itemize}
    \item \textbf{для каждой посещённой дуги $\aleph, a$:}
        \begin{itemize}
        \item считаем $\hat{V}$ --- суммарный reward-to-go в траектории $\Traj$, полученный после посещения данной дуги, где награда после посещения листа оценена как $\frac{1}{|\A|} \sum_{a} R(\Traj_a)$.
        \item $Q(\aleph, a) \leftarrow Q(\aleph, a) + \frac{1}{n(\aleph, a)}(\hat{V} - Q(\aleph, a))$
        \item $n(\aleph, a) \leftarrow n(\aleph, a) + 1$
        \end{itemize}
\end{enumerate}

\vspace{0.3cm}
\textbf{Выход:} стратегия $\pi(a_0 \mid s_0) \propto n(\aleph_0, a_0)^T$
\end{algorithm}

Итого, мы получили очень дорогостоящую по времени, разумную по памяти, но работающую процедуру поиска хороших действий в игре.

\begin{exampleBox}[label=ex:mcts]{}
Попробуем провизуализировать шаг игры при помощи MCTS стратегии. Перед выбором действия будем делать 4 шага MCTS процедуры; выбирать действие для реальной среды будем при помощи $\pi^{\mathrm{MTCS}}(a \HM\mid s) \propto n(\aleph, a)$, где $\aleph$ --- корень дерева.

\begin{center}
\animategraphics[controls, width=\linewidth]{1}{Images/MCTS/MCTS-}{0}{23}
\end{center}
\end{exampleBox}

\subsection{Дистилляция MCTS}

Появляется прекрасная идея: \emph{дистиллировать} MCTS в нейронку. Практически, мы займёмся имитационным обучением с MCTS в качестве эксперта. Для этого мы уже будем проводить этап обучения. Для обучения мы сыграем при помощи вышеописанной стратегии MCTS много игр, и сохраним их траектории\footnote{замечание: траектории именно реальных игр, в которых каждое действие выбиралось в результате, там, 1000 шагов MCTS процедуры, на каждом из которых MCTS делало кучу симуляций при помощи своей DefaultPolicy; эти симуляции внутри MCTS мы нигде не записываем, поскольку нас интересуют только хорошие действия по результатам перебора.} $\Traj$. Далее решаем задачу классификации: пытаемся по состоянию $s$ нейросеткой выдавать действия, которые выбрала бы MCTS. В идеале мы получим стратегию, работающую не хуже MCTS, но при этом скорость прямого прохода по сети будет намного, намного выше.

Схожая альтернатива --- пытаться выдавать нейросеткой-стратегией $p_{\theta}(a \mid s)$ то же распределение, которое выдаёт долгий MCTS перебор; назовём это <<целевое распределение>> $\pi^{\mathrm{MCTS}}(a \mid s)$. Тогда функция потерь выглядит так:
\begin{equation}\label{distill_mcts}
\E_{s} \KL(\pi^{\mathrm{MCTS}}(a \mid s) \parallel p_{\theta}(a \mid s)) = \E_{s} \sum_a \pi^{\mathrm{MCTS}}(a \mid s) \log p_{\theta}(a \mid s) + \const(\theta) \to \min_{\theta},
\end{equation}
где мат.ожидание по $s$ берётся из буфера, из кучи сыгранных при помощи MCTS игр.

И тут возникает желание сделать следующий шаг: если у нас есть нейросеть, которая знает, какие действия в каком состоянии хорошие за счёт моделирования результата MCTS-перебора, может быть мы можем в новой игре воспользоваться ею внутри самого MCTS, использовать эту нейросеть $p_{\theta}(a \mid s)$ для ускорения перебора? Тогда мы наверняка сможем построить ещё более хорошую стратегию, дистиллировать её в нейронку, построить ещё более хорошую стратегию, дистиллировать её в нейронку, и так далее. Итак, появляется идея объединить MCTS с нейросетками.

\subsection{AlphaZero}

В алгоритме AlphaZero\footnote{AlphaZero исторически был обобщением алгоритма AlphaGo для игры в \href{https://ru.wikipedia.org/wiki/Го}{го} как общей процедуры обучения стратегии для произвольной игры с доступным симулятором. Для случая игры двух игроков вроде шахмат и го, для которого она изначально и строилась, предполагается, что в симуляторе за противника играет просто недавняя версия текущей MCTS-стратегии; он также строился для детермированных сред с ненулевой наградой лишь в конце эпизода по типу <<выиграл-проиграл>>, но мы далее опишем сразу чуть-чуть более общую схему.} вводится нейросеть с параметрами $\theta$, принимающая на вход состояние $s$, с двумя головами. Одна выдаёт распределение на действиях $p_{\theta}(a \mid s)$ (<<вспомогательную стратегию>> или <<дистилированную MCTS>>), другая выдаёт скалярную оценку текущего состояния $V_\theta(s)$. Сразу скажем, что $p$ --- лишь вспомогательная стратегия, и хотя ею вполне можно будет пользоваться для игры по итогам обучения, но наша итоговая стратегия всё-таки будет перебирать ходы при помощи MCTS и поэтому будет потенциально лучше. Цель $p_{\theta}(a \mid s)$ --- запоминать с прошлых игр, какие действия были хорошими, ускоряя перебор, а цель $V_{\theta}(s)$ --- запоминать reward-to-go, чтобы не было необходимости проводить симуляции при помощи случайной стратегии на этапе Simulation; вместо симуляции мы теперь просто будем вызывать эту нейросетку.

Итак, за основу алгоритма берётся схема MCTS-стратегии, описанная в алгоритме \ref{alg:mcts}. Как и раньше, в каждом узле $\aleph$ для каждого действия хранится оценка ценности $Q(\aleph, a)$ и счётчик, сколько раз какое действие было попробовано $n(\aleph, a)$; при выборе очередного действия для состояния $s$ повторяется, там, 1600 раз наши четыре этапа MCTS схемы, а каждый эпизод дерево начинает строиться с нуля. 

Изменений в схеме ровно два: на этапе Simulation мы вместо того, чтобы играть случайной стратегией до конца игры, просто воспользуемся нашей моделью $\hat{V}_a \coloneqq V_{\theta}(\hat{s})$ как оценкой дальнейшего reward-to-go (здесь $\hat{s}$ --- симулированное состояние после выбора оцениваемого действия $a$ в листе). Счётчики и скаляры $Q(\aleph, a)$ обновляются как раньше.

Второе изменение --- в формуле TreePolicy, которая принимает следующий вид:
$$a \coloneqq \argmax\limits_a \left[ Q(\aleph, a) + C p_{\theta}(a \mid s) \frac{\sqrt{ n(\aleph)}}{n(\aleph, a) + 1} \right]$$
Здесь $C$ --- гиперпараметр; бонус за исследования немного не похож на UCB-бандита, но имеет примерно тот же смысл (он обратно пропорционален числу выборов действия, а числитель гарантирует, что это слагаемое неограниченно растёт при фиксированном знаменателе и промотивирует выбор любого сколь угодно неоптимального действия рано или поздно). Ключевое изменение --- домножение бонуса на $p_{\theta}(a \mid s)$. Если действие в прошлых играх никогда не выбиралось по итогам перебора, дистилляция в нейросетку приведёт к низкой вероятности $p_{\theta}(a \mid s)$, и такое плохое действие будет выбираться редко. Это позволяет существенно сократить перебор для сред с большим числом\footnote{например, в том же го число доступных действий достигает $19^2$.} действий $|\A|$: в новых состояниях дерева, где все оценки, условно, $Q(\aleph, a) \HM= 0, n(\aleph, a) \HM= 0$, мы вместо случайного тыканья и исследования всех вариантов будем больше опираться именно на те действия, которые показались хорошими нейросетке. Запоминание информации из предыдущих эпизодов обучения направляет раскрытие дерева <<в хорошую область>>.

Как и раньше, результатом работы MCTS является стратегия $\pi^{\mathrm{MCTS}}(a \mid s) \propto n(\aleph_0, a)^T$, где $T$ --- гиперпараметр температуры, $\aleph_0$ --- текущий корень. В ответ на наше действие $a$, сэмплированное из этой стратегии, получаем истинный сэмпл $s'$ из реальной среды и оставляем от нашего дерева лишь поддерево, соответствующее $\operatorname{child}(\aleph_0, a)$.

При помощи такой MCTS-стратегии с текущими параметрами нейросети $\theta$ играем много-много игр и сохраняем записи реальных траекторий; для каждого состояния $s$ запомним $\pi^{\mathrm{MCTS}}(a \mid s)$ и считаем реальный reward-to-go, который обозначим за $z$. Дальше батчами учим $V_\theta$ воспроизводить среднее $z$, минимизируя MSE, а $p_{\theta}(a \mid s)$ --- воспроизводить $\pi^{\mathrm{MCTS}}$ аналогично \eqref{distill_mcts}. Ещё добавлен регуляризатор на параметры, итого:
$$\Loss(\theta) \coloneqq \E_{s} \left[ (z - V_\theta(s))^2 - \sum_a \pi(a \mid s) \log p_{\theta}(a \mid s) + \alpha \| \theta \|^2 \right] \to \min_\theta,$$
где в мат.ожидании $s$ берутся из буфера, $\alpha$ --- гиперпараметр для регуляризации. После многих итераций дообучения нейросети снова играем много игр при помощи MCTS с новыми параметрами нейросетки $\theta$, снова сохраняем записи реальных игр и докидываем их в буфер, и так далее.

В буфере можно хранить в том числе довольно старые игры, поскольку $\pi^{\mathrm{MCTS}}(a \mid s)$ можно считать в некотором смысле <<хорошей>> стратегией даже если $V_\theta(s)$ выдаёт ерунду, а $p_{\theta}(a \mid s)$ условно случайная. Действительно, по мере раскрытия дерева первое <<перестаёт>> использоваться, а второе влияет лишь на исследования внутри самого дерева. Но чем лучше будут эти нейросетки, тем лучше пройдёт MCTS-поиск. Понятно, что $V_{\theta}(s)$, обучаясь на reward-to-go, учит V-функцию для MCTS-стратегии, породившей сэмпл $z$, но ей в таком алгоритме не существенно быть критиком именно самой свежей MCTS-стратегии: ведь в исходной схеме MCTS мы на этапе симуляции так вообще доигрывали игру случайной стратегией.

\begin{example}[AlphaGo Zero]
Основные детали применения данного алгоритма для го можно посмотреть на \href{https://miro.medium.com/max/2000/1*0pn33bETjYOimWjlqDLLNw.png}{этой картинке}.
\end{example}

\subsection{$\mu$-Zero}

Обобщим\footnote{для полноты описания алгоритма, в оригинальной статье введён ещё ряд непринципиальных изменений в AlphaZero: используется ещё более длинная формула для TreePolicy сразу с несколькими гиперпараметрами, а также вводится обобщение алгоритма на случай частично наблюдаемых сред (см. раздел \ref{sec:PoMDP}).} AlphaZero на случай, когда симулятор нам недоступен. Параметры всех нейросетевых моделей будем обозначать $\theta$, в конечном счёте все части модели будут обучаться end-to-end единой функцией потерь.

На вход стратегии будет приходить реальное состояние среды $s_0$. Оно отправляется в кодировщик: нейросеть $h_{\theta}(s_0)$, которая вернёт латентное представление состояния $\aleph_0$ --- некоторый компактный вещественный вектор. Далее мы для определения стратегии запускаем MCTS-процедуру, однако ей нужен симулятор. В качестве симулятора мы заведём нейросеть $g_{\theta}(\aleph, a)$, которая по данному латентному представлению состояния $\aleph$ и действию $a$ детерминировано выдаст латентное представление следующего состояния $\aleph'$; а также нейросеть $\hat{r}_{\theta}(\aleph, a)$ для моделирования награды за один данный шаг. Поскольку функция $g_{\theta}$ детерминирована, пересчитывать её значение каждый раз, когда MCTS идёт вдоль этой ветки, не понадобится. Нейросети $V_{\theta}$ и $p_{\theta}$ теперь будут тоже принимать на вход компактное латентное представление $\aleph$; это позволит нам внутри дерева запускать их в произвольных узлах.

Осталось понять, как обучать нововведённые функции $g_{\theta}, \hat{r}_\theta, h_{\theta}$. Как и в AlphaZero, соберём датасет из нескольких игр текущей версии модели, и дальше проведём несколько итераций улучшения всех параметров $\theta$. Мы сможем сделать это end-to-end. Возьмём некоторый момент реальной игры с состоянием $s$, reward-to-go с этого момента $z$ и выданной MCTS <<хорошей стратегией>> $\pi^{\mathrm{MCTS}}$. Тогда функция потерь из AlphaZero будет в том числе зависеть от параметров функции $h_{\theta}$:
$$(z - V_\theta(h_{\theta}(s)))^2 - \sum_a \pi^{\mathrm{MCTS}}(a \mid s) \log p_\theta(a \mid h_{\theta}(s)) + \alpha \| \theta \|^2 \to \min_{\theta}$$

Добавить слагаемое для обучения модели функции награды, которая должна воспроизводить награду $r(s, a)$ за один шаг, несложно:
$$\left( r(s, a) - \hat{r}_{\theta}(h_{\theta}(s), a) \right)^2 \to \min_{\theta}$$

Наконец, осталось разобраться с обучением приближения динамики $g_{\theta}$. Тут есть первое соображение: давайте заглянем на один момент в будущее и посмотрим на реальное $s' \HM\sim p(s' \mid s, a)$. Мы могли бы воспользоваться этим сэмплом и явно попросить нашу динамику предсказывать латентное представление будущего: 
$$\|\underbrace{g_{\theta}(h_{\theta}(s), a)}_{\text{\shortstack{предсказанное \\ будущее}}} - \underbrace{h_{\theta}(s')}_{\text{\shortstack{реальное \\ будущее}}}\|_2^2 \to \min_{\theta}$$
Мы так делать не будем (хотя могли бы). Такое слагаемое функции потерь мотивирует латентное представление быть в некотором смысле <<простым>> для предсказания, и эта простота достигается за счёт ухудшения семантического смысла: какие-то важные фичи для $V_\theta, p_\theta$ <<теряются>>, поскольку потеря в лоссе для них меньше, чем потеря в лоссе предсказания усложнённого представления $h_{\theta}(s)$. Нам же важно другое: чтобы латентное представление было <<достаточным>> для предсказания $V_\theta, p_\theta$, будущих наград и хорошей стратегии. Если это достигается, то мы спокойно согласимся как с <<несогласованностью>> латентного описания (когда результат применения модели динамики $g_{\theta}$ не соответствует латентному описанию реального $s'$, хотя всё равно позволяет предсказывать хороший reward-to-go и стратегию для этого $s'$), так и с потерей некоторой информации, несущественной для агента.

Поэтому мы будем обучать $g_{\theta}$ по-другому. Для $s'$ мы знаем целевые переменные для функций $V_\theta, p_\theta, \hat{r}_\theta$. Давайте добавим их лоссы для $s'$, но входное латентное представление для $s'$ получим не при помощи $h_{\theta}(s')$, а при помощи $g_{\theta}(h_{\theta}(s), a)$. То есть пусть $z'$ --- reward-to-go со следующего состояния в реальной игре, $\pi^{\mathrm{MCTS}}(a' \HM\mid s')$ --- результат работы MCTS на следующем шаге $s'$, тогда добавим такое слагаемое в функцию потерь:
\begin{equation*}
(z' - V_\theta(\aleph'_{\theta}))^2 - \sum_{\hat{a}} \pi^{\mathrm{MCTS}}(\hat{a} \mid s') \log p_\theta(\hat{a} \mid \aleph'_{\theta}) + \left( r' - \hat{r}_{\theta}(\aleph'_{\theta}, a') \right)^2 + \alpha \| \theta \|^2 \to \min_{\theta},
\end{equation*}
где $\aleph'_{\theta} \coloneqq g_{\theta}(h_{\theta}(s), a)$.

Заметим, что мы можем взять из буфера целый роллаут длины $K$, и представление для $s^{(k)}$ получать как $k$ преобразований нашей динамикой $g$ представления первого состояния $h_{\theta}(s_0)$; аналогично, для него считать лоссы обучения $V_\theta, p_\theta$ (а также $\hat{r}_{\theta}$). Все выполненные при этом действия следует брать из роллаута, реальной записи игры. Полученные лоссы все суммируются; формальная функция потерь получается следующая:
\begin{equation}\label{muzero}
\sum_{k=0}^K \left[ (z^{(k)} - V_\theta(\aleph^{(k)}_{\theta}))^2 - \sum_{\hat{a}} \pi^{\mathrm{MCTS}}(\hat{a} \mid s^{(k)}) \log p_\theta(\hat{a} \mid \aleph^{(k)}_{\theta}) + \left( r^{(k)} - \hat{r}_{\theta}(\aleph^{(k)}_{\theta}, a^{(k)}) \right)^2 \right] + \alpha \| \theta \|^2 \to \min_{\theta},
\end{equation}
где $z^{(k)}$ --- reward-to-go, начиная с $k$-го шага после рассматриваемого момента времени $s_0$, латентное представление начального состояния определяется кодировщиком $\aleph^{(0)}_{\theta} \HM\equiv \aleph \HM\coloneqq h_{\theta}(s)$, а последующие --- динамикой: $\aleph^{(k)}_{\theta} \HM \coloneqq g_{\theta}(\aleph^{(k - 1)}_{\theta}, a^{(k-1)})$.

Итоговую схему проще представить на картинке:
\begin{center}
    \includegraphics[width=0.9\textwidth]{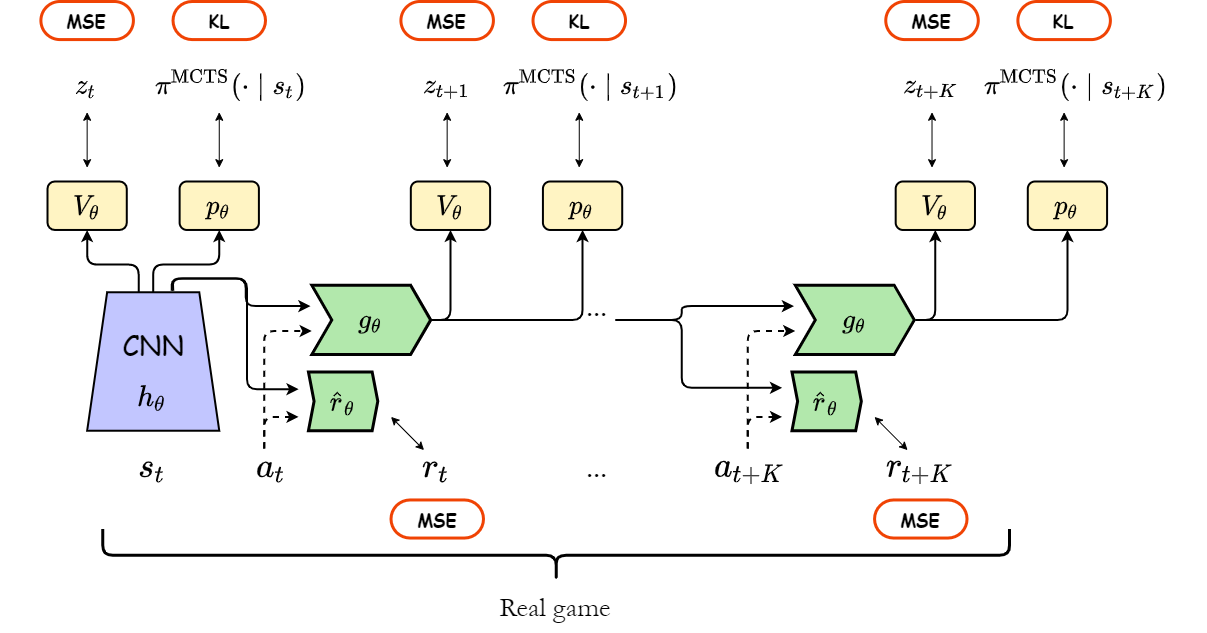}
\end{center}

Подумаем, как полученный алгоритм $\mu$-Zero будет обучаться, то есть <<в каком порядке>> будут улучшаться нейросетки. Понятно, что MCTS, в котором динамика среды заменена на необученную нейросетку, ничего разумного выдавать не будет. Однако для обучения модели награды за шаг $r_{\theta}$ в любом датасете у нас всегда будет ground truth, поэтому эта сетка постепенно будет обучаться, и с ходом этого обучения $h_{\theta}$ будет постепенно строить осмысленное латентное представление состояний. Наше приближение $V_{\theta}$, обучающееся на reward-to-go, тоже учится воспроизводить V-функцию <<текущей MCTS-стратегии>>, сколь бы плохой она ни была, и мы помним, что оценочная функция любой стратегии --- это путь к её улучшению. Поскольку мы учимся предсказывать будущие reward-to-go в том числе по преобразованному моделью динамики латентному представлению в слагаемом \eqref{muzero}, модель $g_{\theta}$ научится выдавать такое представление, по которому мы умеем хорошо предсказывать будущие награды. Именно их мы и используем на этапе Simulation в листьях дерева; дерево начнёт ходить в те ветки, где reward-to-go больше, и проводить таким образом policy improvement. Дальше эта более хорошая стратегия будет дистиллироваться в нейронку $p_{\theta}$, а для этого ещё более информативным будет становиться как латентное представление, так и его преобразование моделью динамики.

\begin{remark}
Число итераций на этапе планирования для каждого выбора действий для реальной среды здесь имеет поставить сильно меньше, чем в AlphaZero, поскольку понятно, что перебор с неидеальным симулятором менее осмысленный, чем с идеальным. Тем не менее, универсальность алгоритма $\mu$-Zero противостоит его огромной вычислительной сложности: необходимо делать огромное количество итераций MCTS процедуры, требующей постоянных вызовов нейросетей. 
\end{remark}
\section{Планирование для непрерывного управления}\label{lqrsection}

\subsection{Прямое дифференцирование}


Построим планировщик для непрерывных пространств действий. Предположим, что модель динамики среды и награды (то есть все функции из постановки задачи) нам известны, причём даны не просто как симулятор, а как дифференцируемые и, главное, детерминированные функции. Для упрощения повествования будем считать, что эпизод состоит из $T$ шагов, $\gamma \HM= 1$. Дополнительно рассматривая случай непрерывного пространства действий $\A \HM\equiv \R^d$, мы получим ситуацию, в которой мы можем просто дифференцировать вдоль оси времени и оптимизировать награду по действиям <<напрямую>>.

В силу детерминированности функции переходов наш выбор действий полностью определяет траекторию. То есть искать будем даже не оптимальную детерминированную стратегию, а \emph{оптимальное управление} --- набор действий $a_1, a_2 \dots a_T$, которые приведут к наилучшей траектории. 

Поскольку мы здесь немного уходим в мир оптимального управления, для соблюдения каноничности не будем предполагать однородность: функции награды и динамики среды дополнительно зависят от дискретной переменной времени $t \in \{1 \dots T\}$. 

Итого, рассматриваем следующую задачу:
\begin{equation}\label{LQRtask}
\begin{cases}
\sum_t^T r_t(s_t, a_t) \to \max\limits_{a_1 \dots a_T} \\
s_t = f_t(s_{t-1}, a_{t-1})
\end{cases}
\end{equation}

Сразу заметим, что при сделанных предположениях можно попытаться решать задачу <<в лоб>>. Промоделируем детерминированную стратегию  нейросеткой $\pi_t(s, \theta)$ с параметрами $\theta$, и рассмотрим такой вычислительный граф:

\begin{center}
    \includegraphics[width=\textwidth]{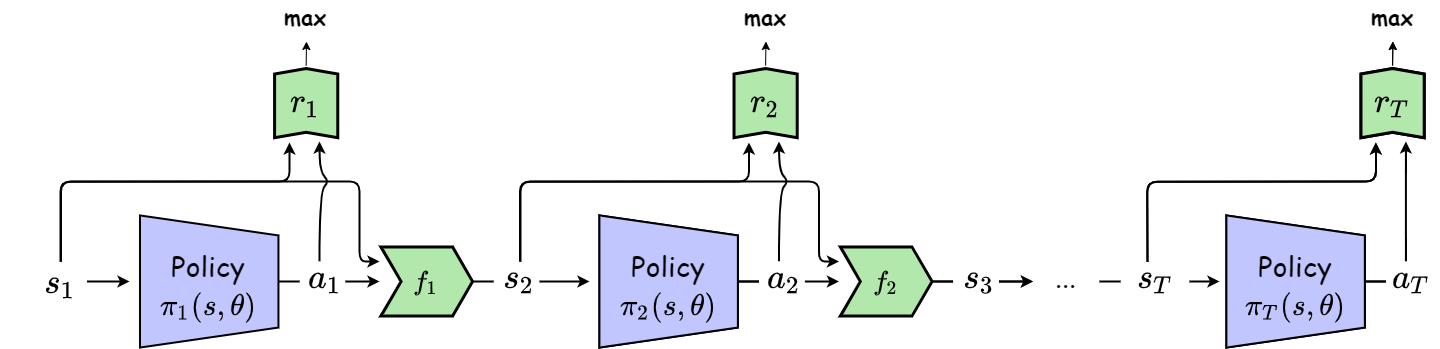}
\end{center}

Здесь суммарная награда --- дифференцируемая функция от $\theta$, поэтому параметры стратегии можно оптимизировать напрямую. Таким образом, у нас получилась <<дифференцируемая>> задача динамического программирования. Однако, если $T$ велико, то градиент вдоль такого вычислительного графа будет подвержен затуханию, а сам такой поиск в пространстве траектории будет содержать множество локальных минимумов.

Нам хочется построить, если угодно, более специализированный метод оптимизации для задач вида \ref{LQRtask}. Как мы обычно действуем при построении методов оптимизации: вот у нас есть некоторая сложная функция, которую мы хотим оптимизировать, возможно, с ограничениями (а аналитически ничего не считается). Заводим некоторое приближение решения, в данном случае --- какой-то план $a_1, a_2 \dots a_T$. Затем приблизим в окрестности этой точки оптимизируемую функцию какой-то простой моделью, с которой мы умеем работать, за счёт простоты модели делаем какой-то шаг и попадаем в новую точку, новый план $a_1, a_2 \dots a_T$. Оказывается, что в контексте задачи \ref{LQRtask} мы можем поступить умнее, чем просто подставить ограничения внутрь оптимизируемого функционала и приблизить всё линейной моделью, как мы делаем при градиентном спуске. Оказывается, если мы заменим функции наград на квадратичное приближение, а ограничение --- функцию динамики --- на линейное приближение, то мы сможем за счёт структуры задачи просто методом динамического программирования аналитически решить задачу! Тогда <<проведение шага>> для нас будет просто сдвиг в точку нового оптимального плана для подобной локальной аппроксимации. 

\subsection{Linear Quadratic Regulator (LQR)}

Обсудим, как можно аналитически найти решение задачи в ситуации, когда $f, r$ --- <<простые>> функции, а именно линейная и квадратичная соответственно. Пусть функция $f$ --- линейна, функция $r$ --- квадратична (hence the name). Введём соответствующие обозначения:
\begin{equation}\label{lineardynamics}
f_t(s, a) = F_t \begin{bmatrix} s \\ a \end{bmatrix} + f_t
\end{equation}
$$r_t(s, a) = \frac{1}{2} \begin{bmatrix} s \\ a \end{bmatrix}^T R_t \begin{bmatrix} s \\ a \end{bmatrix} + \begin{bmatrix} s \\ a \end{bmatrix}^T r_t$$

Напоминаем, что выход функции $f$ --- следующее состояние (т.е. вектор), поэтому $F_t$ --- матрица, $f_t$ --- вектор. Выход функции $r$ --- скаляр, поэтому $R_t$ --- матрица, $r_t$ --- вектор. Свободный член квадратичной формы не пишем, потому что он не зависит от состояний и действий (траектории), поэтому при оптимизации это константная неоптимизируемая награда, и она несущественна. Матрицы $F_t, R_t$ и вектора $r_t, f_t$ считаем известными. 

Нам понадобится ещё чуть-чуть обозначений. Будем считать, что блоки матрицы $R_t$ и блоки вектора $r_t$ выглядят следующим образом:
$$R_t \coloneqq \begin{bmatrix} R_{t, s, s} & R_{t, s, a} \\ R_{t, a, s} & R_{t, a, a} \end{bmatrix}; \qquad r_t \coloneqq \begin{bmatrix} r_{t, s} \\ r_{t, a} \end{bmatrix}$$
Для упрощения выкладок также будем полагать, что $R_{t, s, a} \HM= R^T_{t, a, s}$.

Поймём, что мы можем решить задачу обычным динамическим программированием <<с конца>>.
\begin{theorem}
Оптимальное действие в последний момент времени $a_T^*$ --- линейная форма от состояния $s_T$.
\begin{proof}
Рассмотрим последний момент времени $T$ и распишем оптимальную Q-функцию (в силу отказа от однородности оценочные функции также зависят от времени):
$$Q^*_T(s_T, a_T) = r_T(s_T, a_T) = \frac{1}{2} \begin{bmatrix} s_T \\ a_T \end{bmatrix}^T R_T \begin{bmatrix} s_T \\ a_T \end{bmatrix} + \begin{bmatrix} s_T \\ a_T \end{bmatrix}^T r_T$$

И всё, потому что на сим эпизод заканчивается и награды больше не будет. Мы легко можем найти оптимальное действие, если на последнем шаге оказались в состоянии $s_T$, промаксимизировав $Q^*_T$ по действию $a_T$:
$$a^*_T = \argmax_{a_T} Q^*_T(s_T, a_T)$$

Ищем оптимум квадратичной формы\footnote[*]{формально здесь нужно оговариваться, что на матрицу $R_t$ и вектор $r_t$ надо накладывать некоторые ограничения, например, что эта квадратичная форма отрицательно определена и у неё есть искомый глобальный максимум. Здесь и далее мы на условиях регулярности останавливаться не будем.}, приравняв градиент $Q^*_T$ по действиям к нулю:
$$\nabla_{a_T} Q^*_T(s_T, a_T) = R_{T, a, a} a_T + R_{T, a, s}s_T + r_{T, a} = 0$$
$$a^*_T = -R_{T, a, a}^{-1} \left( R_{T, a, s}s_T + r_{T, a} \right)$$

Видно, что оптимальное действие --- линейная форма от последнего состояния. 
\end{proof}
\end{theorem}

\begin{remark}
Видно, что придётся обращать матрицу $R_{T, a, a}$, но размерность пространства действий в задачах такой постановки имеет обычно разумные размеры (обычно не более 100), поэтому мы не боимся.
\end{remark}

Введём обозначения, чтобы привести форму $a^*_T$ к стандартному виду:
$$K_T \coloneqq -R_{T, a, a}^{-1} R_{T, a, s}, \qquad k_T \coloneqq -R^{-1}_{T, a, a} r_{T, a}$$
$$a^*_T = K_T s_T + k_T$$

\begin{theorem}
Ценность последнего состояния $V^*(s_T)$ --- квадратичная форма от состояния $s_T$.
\beginproof
По определению, в силу детерминированности среды, $V^*(s_T) = Q^*(s_T, a^*_T)$; осталось заметить, что подстановка линейной формы в линейную даст квадратичную:
\begin{align*}
V^*(s_T) &= Q^*(s_T, a^*_T) = \frac{1}{2} \begin{bmatrix} s_T \\ a^*_T \end{bmatrix}^T R_T \begin{bmatrix} s_T \\ a^*_T \end{bmatrix} + \begin{bmatrix} s_T \\ a^*_T \end{bmatrix}^T r_T = \\
&= \frac{1}{2} \begin{bmatrix} s_T \\ K_T s_T + k_T \end{bmatrix}^T R_T \begin{bmatrix} s_T \\ K_T s_T + k_T \end{bmatrix} + \begin{bmatrix} s_T \\ K_T s_T + k_T \end{bmatrix}^T r_T = (*)
\end{align*}

Просто перегруппируя слагаемые, видим квадратичную форму от состояний $s_T$. Сразу запишем её в каноничных обозначениях:
$$V_T \coloneqq R_{T, s, s} + K_T^TR_{T, a, a}K_T + K_T^TR_{T, a, s} + R_{T, s, a}K_T$$
$$v_T \coloneqq R_{T, s, a}k_T + r_T^T K_T^T r_T$$
\begin{equation*}
(*) = \frac{1}{2} s_T^T V_T s_T + v_T^T s_T,    \tagqed
\end{equation*}
\end{theorem}

Осталось понять, что мы можем раскручивать это к началу времён, не выходя из квадратичных форм. Так и есть.

\begin{theorem}
Оптимальные оценочные функции $Q_t^*$ --- квадратичные формы от $\begin{bmatrix} s_t \\ a_t \end{bmatrix}$, оптимальные действия $a_t^*$ --- линейные формы от $s_t$, а оптимальные оценочные функции $V_t^*$ --- квадратичные формы от $s_t$.
\begin{proof}
Из определений и детерминированности:
$$Q^*_{T-1}(s_{T-1}, a_{T-1}) = r_{T-1}(s_{T-1}, a_{T-1}) + V^*_T(s_T) = r_{T-1}(s_{T-1}, a_{T-1}) + V^*_T \left( F_T \begin{bmatrix} s_{T-1} \\ a_{T-1} \end{bmatrix} + f_T \right) = (*)$$

В последнее слагаемое подставили линейную форму динамики среды \eqref{lineardynamics}. При этом подстановка линейной формы в квадратичную оставит её квадратичной.

\begin{align*}
    (*) &= r_{T-1}(s_{T-1}, a_{T-1}) + \frac{1}{2} \left[ F_T \begin{bmatrix} s_{T-1} \\ a_{T-1} \end{bmatrix} + f_T \right]^T V_T \left[ F_T \begin{bmatrix} s_{T-1} \\ a_{T-1} \end{bmatrix} + f_T \right] + v_T^T \left[ F_T \begin{bmatrix} s_{T-1} \\ a_{T-1} \end{bmatrix} + f_T \right] = \\
    &= r_{T-1}(s_{T-1}, a_{T-1}) + \frac{1}{2} \begin{bmatrix} s_{T-1} \\ a_{T-1} \end{bmatrix}^T F_T^TV_TF_T \begin{bmatrix} s_{T-1} \\ a_{T-1} \end{bmatrix} + \begin{bmatrix} s_{T-1} \\ a_{T-1} \end{bmatrix}^T \left( F_T^T V_T f_T + F_T^T v_T \right) + \const(a_{T-1})
\end{align*}

Переписываем в виде квадратичной формы:
$$Q^*_{T-1}(s_{T-1}, a_{T-1}) = \frac{1}{2} \begin{bmatrix} s_{T-1} \\ a_{T-1} \end{bmatrix}^T Q_{T-1} \begin{bmatrix} s_{T-1} \\ a_{T-1} \end{bmatrix} + \begin{bmatrix} s_{T-1} \\ a_{T-1} \end{bmatrix}^T q_{T-1} + \const(a_{T-1})$$
$$Q_{T-1} \coloneqq R_{T-1} + F_T^TV_TF_T$$
$$q_{T-1} \coloneqq r_{T-1} + F_T^T V_T f_T + F_T^T v_T$$

Коли это снова квадратичная формула, мы можем посчитать оптимальное действие, которое будет линейной формой:
$$a_{T-1} = K_{T-1}s_{T-1} + k_{t-1}$$
$$K_{T-1} \coloneqq -Q_{T-1, a, a}^{-1} Q_{T-1, a, s}, \qquad k_{T-1} \coloneqq -Q^{-1}_{T-1, a, a} q_{T-1,a}$$

Подставляем её в $V^*_{T-1}$ и получаем квадратичную и так далее.
\end{proof}
\end{theorem}

Собираем всё вместе в единую схему.

\begin{algorithm}{Обратный проход в LQR}
\textbf{Вход:} $F_t, f_t$ --- функция динамики, $R_t, r_t$ --- функция награды.

\vspace{0.3cm}
Инициализировать $V_{T+1} = 0_{|\St| \times |\St|}, v_{T+1} = 0_{|\St|}$ \\
\textbf{for $t$ от $T$ до 1:}
\begin{enumerate}
    \item считаем Q-функцию:
    $$Q_t \coloneqq R_t + F_{t+1}^TV_{t+1}F_{t+1}$$
    $$q_t \coloneqq r_t + F_{t+1}^TV_{t+1}f_{t+1} + F_{t+1}^T v_{t+1}$$
    \item считаем оптимальную стратегию:
    $$K_t \coloneqq -Q_{t, a, a}^{-1} Q_{t, a, s}$$
    $$k_t \coloneqq -Q^{-1}_{t, a, a} q_{t,a}$$
    \item считаем V-функцию:
    $$V_t \coloneqq Q_{t, s, s} + K_t^TQ_{t, a, a}K_t + K_t^TQ_{t, a, s} + Q_{t, s, a}K_t$$
    $$v_t \coloneqq Q_{t, s, a}k_T + q_t^T K_t^T q_t$$
\end{enumerate}

\vspace{0.3cm}
\textbf{Выход:} $\pi_t(s) \coloneqq K_ts + k_t$
\end{algorithm}

Поскольку в рамках сделанных предположений мы на самом деле детерминированно знаем, в каких состояниях окажемся (считаем стартовое состояние $s_1$ также известным), мы можем просто вывести оптимальное управление:
\begin{algorithm}{Прямой проход в LQR}
\textbf{Вход:} $K_t, k_t$ --- стратегия с прямого прохода, $f$ --- функция динамики.

\vspace{0.3cm}
\textbf{for $t$ от 1 до $T$:}
\begin{enumerate}
    \item $a_t = K_t s_t + k_t$
    \item $s_{t+1} = f(s_t, a_t)$
\end{enumerate}

\vspace{0.3cm}
\textbf{Выход:} $a_1, a_2, \dots a_T$ --- план.
\end{algorithm}

\subsection{Случай шумной функции перехода}

LQR обобщается на случай недетермнированных сред, если предположить, что динамика среды --- нормальное распределение с линейной функцией от предыдущих состояний-действий и какой-то фиксированной (зависящей только от момента времени $t$, но не состояний-действий) матрицей ковариации:
\begin{equation}\label{noisyLQRlineardynamics}
p(s_{t} \mid s_{t-1}, a_{t-1}) \coloneqq \N \left( f_t(s_{t-1}, a_{t-1}), \Sigma_t \right) 
\end{equation}

Формулу \eqref{noisyLQRlineardynamics} можно интерпретировать как <<зашумление сенсоров>>, причём зашумление не зависит от того, в какой области пространства состояний мы оказались.

\begin{theorem}
В предположении \eqref{noisyLQRlineardynamics} схема LQR остаётся неизменной.
\begin{proof}
Единственное, что поменялось --- это зависимость V-функции от Q-функции:
$$Q^*_{t-1}(s_{t-1}, a_{t-1}) = r_{t-1}(s_{t-1}, a_{t-1}) + \E_{s_t} V_t^*(s_t)$$
Покажем, что формулы для матрицы $Q_{t-1}$ и вектора $q_{t-1}$ не поменялись, а изменилась только константа (которая на вывод оптимальной стратегии не влияет).

$$\E_{s_t} V^*_t(s_t) = \E_{s_t} \left[ \frac{1}{2} s_t^TV_ts_t + v_t^T s_t + \const(s_t) \right] = (*)$$

Итак, нужно взять мат.ожидание квадратичной формы по гауссиане. Лезем\footnote[*]{вывод не так сложен, нужно лишь применить трюк, что след скаляра равен скаляру: $$\E_{s_t \sim \N(\mu_t, \Sigma_t)} s_t^TV_ts_t = \E_{s_t \sim \N(\mu_t, \Sigma_t)} \Tr(s_t^TV_ts_t) = (*),$$ затем применить свойство следа $\Tr(ABC) = \Tr(BCA)$ и получить $$(*) = \E_{s_t \sim \N(\mu_t, \Sigma_t)} \Tr(V_ts_ts_t^T).$$ Осталось занести мат.ожидание внутрь следа (след --- линейная операция), и вспомнить, что мат.ожидание $s_t s_t^T$ по гауссиане есть $\Sigma + \mu \mu^T$.} в \href{https://www.math.uwaterloo.ca/~hwolkowi/matrixcookbook.pdf}{matrix cookbook} (ур. (318)) и видим:
$$\E_{s_t \sim \N(\mu_t, \Sigma_t)} s_t^TV_ts_t = \operatorname{Tr}(V_t \Sigma_t) + \mu_t^TV_t\mu_t$$
где $\mu_t \coloneqq f_t(s_{t-1}, a_{t-1})$ --- среднее гауссианы, $\operatorname{Tr}$ --- операция взятия \href{https://ru.wikipedia.org/wiki/\%D0\%A1\%D0\%BB\%D0\%B5\%D0\%B4_\%D0\%BC\%D0\%B0\%D1\%82\%D1\%80\%D0\%B8\%D1\%86\%D1\%8B}{следа}.

Итого получим:
$$(*) = \frac{1}{2} \operatorname{Tr}(V_t \Sigma_t) + \frac{1}{2} \mu_t^TV_t\mu_t + v_t^T \mu_t + \const(s_t) = V^*_t(\mu_t) + \const(s_t)$$
то есть поменялась исключительно константа, которая на оптимальное управление не влияет.
\end{proof}
\end{theorem}

\subsection{Iterative LQR (iLQR)}

Вернёмся к детерминированному случаю. Что делать, если функции $f, r$ нам известны, но не являются линейными и квадратичными соответственно? Мы хотели воспользоваться локальным приближением этих функций в окрестности некоторого текущего плана, и в качестве приближений будем использовать разложение в ряд Тейлора до первого и до второго члена соответственно.

Итак, возьмём какой-нибудь план и просчитаем честным прямым проходом точные значения состояний, в которых мы окажемся. Для полученной траектории $\hat{s}_1, \hat{a}_1, \hat{s}_2 \dots \hat{s}_T, \hat{a}_T$ разложим функции переходов и награды в Тейлора следующим образом:
$$f_t(s_{t-1}, a_{t-1}) \approx f_t(\hat{s}_{t-1}, \hat{a}_{t-1}) + \nabla_{s, a} f(\hat{s}_{t-1}, \hat{a}_{t-1})^T \begin{bmatrix} s_{t-1} - \hat{s}_{t-1} \\ a_{t-1} - \hat{a}_{t-1} \end{bmatrix}$$
$$r_t(s_t, a_t) \approx \frac{1}{2}\begin{bmatrix} s_t - \hat{s}_t \\ a_t - \hat{a}_t \end{bmatrix}^T \nabla^2_{s, a} r_t(\hat{s}_{t}, \hat{a}_{t})\begin{bmatrix} s_t - \hat{s}_t \\ a_t - \hat{a}_t \end{bmatrix} + \nabla_{s, a} r(\hat{s}_t, \hat{a}_t)^T \begin{bmatrix} s_t - \hat{s}_t \\ a_t - \hat{a}_t \end{bmatrix}$$

Вводим обозначения аналогично LQR:

\begin{equation}\label{TaylorFforLQR}
F_t \coloneqq \nabla_{s, a} f(\hat{s}_{t-1}, \hat{a}_{t-1}) \quad f_t \coloneqq f_t(\hat{s}_{t-1}, \hat{a}_{t-1})
\end{equation}
\begin{equation}\label{TaylorRforLQR}
R_t \coloneqq \nabla^2_{s, a} r_t(\hat{s}_{t}, \hat{a}_{t}) \quad r_t \coloneqq \nabla_{s, a} r(\hat{s}_t, \hat{a}_t)
\end{equation}

Используя LQR с такой приближённой моделью динамики и награды, мы можем пересчитать оптимальную траекторию, проделав backward pass, а затем и forward pass (причём на шаге forward pass, естественно, пользуясь истинными точными функциями $f$, $r$, а не их разложениями в Тейлора). Технически, нужно только учесть, что в этой рассматриваемой задаче мы <<перецентрировали>> пространства состояний и действий: LQR тут на шаге $t$ работает с <<новым>> пространством состояний $s_t \HM- \hat{s}_t$ и <<новым>> пространством действий $a_t \HM- \hat{a}_t$. Поэтому, чтобы окончательно получить оптимальную траекторию на этапе forward pass помимо использования истинной функции переходов нужно ещё учесть эти центрирования при вызове стратегии:
$$a_t = K_t (s_t - \hat{s}_t) + k_t + \hat{a}_t$$
Это соответствует простому учёту добавленных слагаемых в свободном векторе:
\begin{equation}\label{iLQR_correction}
k_t \HM\leftarrow k_t - K_t \hat{s}_t + \hat{a}_t
\end{equation}

В полученной траектории снова раскладываем модели в Тейлора, и так далее до удовлетворения.

\begin{algorithm}{Iterative LQR (iLQR)}
\textbf{Вход:} $f$ --- функция динамики, $r$ --- функция награды.

\vspace{0.3cm}
Проинициализировать траекторию $\hat{s}_1, \hat{a}_1, \hat{s}_2 \dots \hat{s}_T, \hat{a}_T$ при помощи случайной стратегии. \\ 
\textbf{На каждом шаге:}
\begin{enumerate}
    \item Получить $F_t, f_t, R_t, r_t$ по формулам \eqref{TaylorFforLQR} и \eqref{TaylorRforLQR}
    \item Получить матрицы $K_t, k_t$ при помощи алгоритма LQR с матрицами динамики $F_t, f_t$ и награды $R_t, r_t$
    \item Учесть коррекцию \eqref{iLQR_correction}
    \item При помощи стратегии $\pi(s_t) = K_t s_t + k_t$ заспавнить траекторию $\hat{s}_1, \hat{a}_1, \hat{s}_2 \dots \hat{s}_T, \hat{a}_T$, используя честный прямой проход с использованием точных функций $f, r$.
\end{enumerate}

\vspace{0.3cm}
\textbf{Выход:} $\hat{a}_1, \hat{a}_2, \dots \hat{a}_T$ --- план.
\end{algorithm}

Мы построили планировщик для непрерывных пространств действий с произвольной (дифференцируемой) динамикой. Мы находимся в некотором состоянии, предполагаем свою будущую траекторию; рассматриваем некоторое приближение поведения среды, которое достаточно точно для предположенной траектории (это разложение в ряд Тейлора); находим оптимальную траекторию при помощи LQR; получаем новую траекторию; рассматриваем приближение поведения среды для новой траектории и так далее до сходимости.

В частности, если модель среды нам неизвестна, iLQR можно применять в качестве планировщика для обученных (например, нейросетевых) приближений динамики среды --- см. общую схему model-based подхода \ref{alg:generalmodelbased}.

\chapter{Next Stage}

Если уж мы рассматриваем задачу RL как попытку создания алгоритма <<искусственного интеллекта>>, то мы должны дополнительно учесть следующие три факта:
\begin{itemize}
    \item понятно, что в одной и той же среде агент может ставить себе совершенно разные задачи; интеллектуальное обучение должно позволять обобщать решения одних задач на другие, решать сложные задачи, состоящие из составных частей, и, наконец, уметь самостоятельно ставить самому себе <<промежуточные>> задачи.
    \item в общем случае, текущее наблюдение среды не описывает её состояние полностью, и агент, во-первых, должен обладать модулем памяти для запоминания предыдущих наблюдений, во-вторых, действовать в условиях неопределённости.
    \item наконец, в среде могут присутствовать другие агенты, которые могут иметь как схожие, так и противоположные цели, передавать вспомогательную информацию или, в частности, играть роль эксперта, демонстрирующих оптимальное (или полезное) поведение.
\end{itemize}
В этой главе мы обсудим ряд избранных идей в направлении этих соображений, которые не вписались в предыдущее повествование. 

\section{Имитационное обучение}\label{sec:imitationlearning}

\subsection{Клонирование поведения}

Зачастую продемонстрировать то, как надо решать задачу, проще, чем описать её в терминах награды.

\begin{example}
Чтобы обучить self-driving car, проще посадить реального водителя за руль и попросить собрать примеры траекторий, чем описать функцию награды, описывающую правила движения.
\end{example}

\begin{example}
Чтобы обучить робота переливать воду из стакана в стакан, проще не придумать функцию награды, описывающую такую задачу, а взять руку робота и несколько раз, <<держа его за ручку>>, перелить воду из стакана в стакан.
\end{example}

Допустим, некоторый эксперт $\pi^{expert}$ повзаимодействовал со средой и собрал для нас набор траекторий $(\Traj)$. Если мы уверены в крутизне эксперта (то есть готовы считать его стратегию оптимальной или в достаточной степени около-оптимальной), то задача обучения собственной стратегии $\pi_{\theta}$ на первый взгляд сводится к задаче обучения с учителем:

\begin{definition}
\emph{Клонированием поведения} (behavioral cloning) называется обучение стратегии воспроизводить действия эксперта:
\begin{equation}\label{behavioralcloning}
\sum_{\Traj} \sum_{s, a \in \Traj} \log \pi_{\theta}(a \mid s) \to \max_{\theta}
\end{equation}
\end{definition}

В случае успеха, мы можем надеяться на то, что наша стратегия $\pi_{\theta}$ будет вести себя не хуже эксперта. При достаточном количестве данных и достаточной крутизне эксперта такой вариант вполне может сработать. При этом мы можем даже не знать или не заниматься придумыванием функции награды: она никак не участвует в обучении.

\begin{remark}
Кстати, учиться выбирать те действия из состояний, которые привели к наиболее удачным сессиям (эпизодам с наибольшей наградой) может быть полезно всегда (в частности, на этом основан кросс-энтропийный метод \ref{alg:cem}). Например, в policy gradient алгоритмах можно пробовать запоминать эпизоды обучения, в которых агенту удалось набрать больше всего награды, и с некоторым весом использовать градиент \eqref{behavioralcloning}, посчитанный по этим лучшим траекториям, для обучения актёра. Такой подход называется \emph{самоимитацией} (self-imitation): просто учимся воспроизводить свои собственные же наилучшие попытки.
\end{remark}

\needspace{5\baselineskip}
\begin{wrapfigure}{r}{0.35\textwidth}
\vspace{-0.4cm}
\centering
\includegraphics[width=0.35\textwidth]{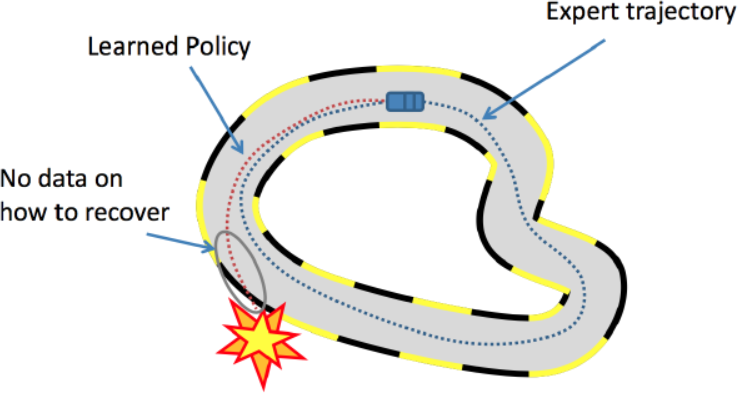}
\vspace{-0.5cm}
\end{wrapfigure}

Однако, подобная <<дистилляция знаний>> одного актёра в другого не учитывает то, что мы работаем с последовательным принятием решений. Как только наша стратегия чуть-чуть отклонится от экспертного поведения (а это, так или иначе, неизбежно) или просто получит необычный отклик от среды, она может оказаться в той области пространства состояний, где эксперта никогда не было, и примеров правильного поведения в обучающей выборке не встречалось. Тогда модель будет предсказывать действия для состояний, которых не было при обучении (эту проблему называют \emph{covariate shift}), и в такой момент клонированное поведение может выдать сколько угодно безумные результаты.

\begin{remark}
Так или иначе, клонирование поведение может сыграть роль отличной инициализации, существенно ускоряющей процесс обучения. Если для off-policy обучения имеет смысл просто докинуть экспертные траектории в буфер, то для on-policy алгоритмов при наличии экспертных траекторий имеет смысл предобучить стратегию при помощи клонирования поведения. 
\end{remark}

Во многих задачах можно побороться с этой проблемой при помощи более <<тщательного>> или <<хитрого>> процесса сбора данных; например, специально закатываясь в те области MDP, куда может заехать <<сломанная>> стратегия.

\begin{example}[DAgger]
Одна из универсальных идей звучит так: после клонирования поведения запустить полученную стратегию в среду, собрать набор тех состояний, которые она посетила, и попросить эксперта <<разметить>> их: выбрать оптимальные действия. Но такой алгоритм предполагает, что у нас есть подобное <<средство разметки>>, за счёт которого задача и сводится к обучению с учителем.
\end{example}

\begin{exampleBox}[label=ex:quadrocopter]{Quadcopter Navigation in the Forest}
Иногда возможно извернуться и придумать какое-нибудь ухищрение, как всё-таки получить в RL <<правильные ответы>>. Например, \href{https://www.youtube.com/watch?v=umRdt3zGgpU&ab_channel=AAAIVideoCompetition}{в этом примере} квадрокоптер учится лететь вдоль лесной тропинки, выбирая на каждом шаге из трёх действий: влево, вперёд или вправо. Чтобы собрать <<обучающую выборку>> для него, человек надел шлем с тремя камерами, смотрящими вправо, вперёд и влево, и пошёл по центру лесной тропинки. Собирается такой датасет: для камеры, смотрящей влево, правильный ответ --- действие <<вправо>>, для центральной --- <<вперёд>>, для правой --- <<влево>>. Всё свелось к обучению классификатора.
\end{exampleBox}

\subsection{Обратное обучение с подкреплением (Inverse RL)}\label{subsec_irl}

Почему клонирование поведения --- неидеальное решение? На самом деле, в разных областях пространства состояний у наших решений различается <<критичность>>: если мы ошибёмся в одном месте, наше будущее поменяется не так сильно, и эксперт в таких областях сам может выбирать довольно разнообразные действия, а в других местах от нашего решения существенно зависит дальнейшая награда, и в них крайне важно безошибочно выбрать то же действие, что и эксперт. Это означает, что функция награды как обучающий сигнал более информативен. Но зачастую функция награды нам напрямую недоступна. 

\begin{definition}
Задачей \emph{обратного обучения с подкреплением} (inverse reinforcement learning, IRL) называется задача по набору траекторий $(\Traj)$ оптимального агента восстановить функцию награды, которую он максимизирует.
\end{definition}

За счёт обратного обучения с подкреплением можно попытаться получить функцию награды, которая позволит обучить при помощи всего арсенала RL алгоритмов куда более близкую к оптимальной стратегию, чем простое клонирование поведения. Однако понятно, что задача обратного обучения с подкреплением <<некорректна>> в том смысле, что допускает всякие дурацкие решения: например, $r(s, a) \HM= 0$ всегда <<подходит>> в качестве ответа. Какой критерий качества в этой задаче, то есть как понять, насколько адекватна придуманная функция награды?

\begin{example}
Рассмотрим клеточный мир, в котором агент может ходить вправо-влево-вниз-вверх. Для простоты также допустим, что функция награды --- детерминированная, зависит только от состояний, и на клетках одного цвета её значения совпадают. Что мы тогда можем о ней сказать, имея на руках одну экспертную траекторию, порождённую оптимальной стратегией?
\begin{center}
    \includegraphics[width=0.7\textwidth]{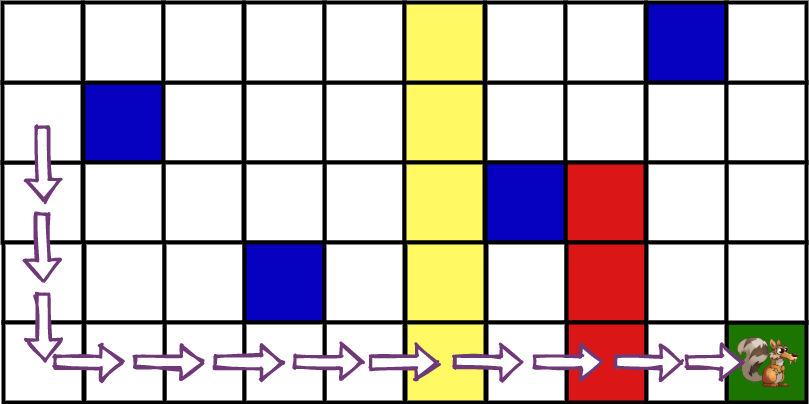}
\end{center}
На самом деле, не так много. Агент, видимо, стремился в зелёную клетку; наверное, за неё полагается положительная награда. Через красную клетку агент прошёл, хотя мог бы обойти за счёт более позднего попадания в зелёную клетку; видимо, это того не стоило, и за красную клетку награда, если и отрицательная, то совсем маленькая. Могла ли она быть положительной? Тогда бы эксперт, наверное, походил бы по красным клеткам; видимо, награда за зелёную сильно выгоднее. Синие клетки агент избегал; видимо, они или дают штраф, или ноль, поскольку если бы они давали бонус, то было бы выгодно добираться до зелёной клетки-цели через них. Наконец, про жёлтые клетки сказать почти ничего нельзя: агенту пришлось бы в любом случае пройти через них, чтобы добраться до зелёной клетки, и поэтому в них может быть как штраф (но не очень большой), так и бонус (но тоже не очень большой).
\end{example}

Ещё одна проблема задачи в том, что эксперт, обычно, не является абсолютно оптимальным. Люди не перемещаются по комнате до целей по строго прямой линии, а ходы шахматного эксперта не являются абсолютной истиной оптимальных действий.

Для упрощения формул будем везде далее полагать $\gamma \HM= 1$. Введём следующее предположение: оптимальная стратегия $\pi^*$ стохастична и генерирует такие траектории, что:
\begin{equation}\label{meirl}
p(\Traj \mid \pi^*) \propto e^{R(\Traj)} \prod_{t \ge 0}p(s_{t+1} \mid s_t, a_t)
\end{equation}

Иначе говоря, вероятность оптимальной стратегии сгенерировать ту или иную траекторию пропорциональна суммарной кумулятивной награде. Мы считаем, что стратегия эксперта, сгенерировавшего для нас датасет, удовлетворяет этому предположению. Такое предположение позволяет нам записать правдоподобие траекторий: насколько вероятно увидеть траекторию $\Traj$, если она пришла из оптимального эксперта, максимизировавшего награду в терминах \eqref{meirl}.

Откуда это предположение свалилось? На самом деле, это уже знакомый нам Maximum Entropy RL. Действительно: рассмотрим задачу поиска стратегии, которая порождает траектории из распределения \eqref{meirl}:
\begin{equation}\label{meirl_kl}
    \KL(p(\Traj \mid \pi) \parallel p(\Traj \mid \pi^*)) \to \min_{\pi}
\end{equation}

\begin{theoremBox}[label=th:inverserlassumptionexplanation]{}
Задача \eqref{meirl_kl} эквивалентна задаче Maximum Entropy RL \eqref{merl}.
\begin{proof}
Распишем \eqref{meirl_kl}:
\begin{align}
    \KL(p(\Traj \mid \pi) \parallel p(\Traj \mid \pi^*)) &= \E_{\Traj \sim \pi} \overbrace{\sum_{t \ge 0} \log \pi(a_t \mid s_t) + \log p(s_{t+1} \mid s_t, a_t)}^{\text{$\log p(\Traj \mid \pi)$}} - \\ &- \E_{\Traj \sim \pi} \underbrace{\sum_{t \ge 0} \log p(s_{t+1} \mid s_t, a_t) - r_t - \const(\pi)}_{\text{$\log p(\Traj \mid \pi^*)$ из \eqref{meirl}}},
\end{align}
где $\const(\pi)$ --- нормировочная константа распределения \eqref{meirl}. Убирая сокращающиеся логарифмы вероятностей переходов и домножая на минус единицу, получаем:
$$\E_{\Traj \sim \pi} \sum_{t \ge 0} \left[ r_t - \log \pi(a_t \mid s_t) \right] \to \max_{\pi},$$
что есть в точности Maximum Entropy RL.
\end{proof}
\end{theoremBox}

Оговоримся, что на самом деле у нас нет гарантий, что мы сможем проминимизировать KL-дивергенцию до точного нуля. Другими словами, нельзя сказать, что оптимальная стратегия в Max. Entropy RL обязательно удовлетворяет свойству \eqref{meirl}.

\begin{proposition}
Ноль в задаче \eqref{meirl_kl} не обязательно достижим.
\begin{proof}[Контрпример]
Рассмотрим MDP с единственным состоянием, в котором от действий ничего не зависит, но с вероятностью 0.5 агент получает $+\log 2$, а с вероятностью 0.5 агент получает $+0$, после чего игра заканчивается. Распределение \eqref{meirl} говорит, что оптимальная стратегия так выбирает действия, что траектория с наградой $+\log 2$ встречает с вероятностью $\frac{2}{3}$, а с наградой $+0$ --- с вероятностью $\frac{1}{3}$, однако это невозможно: любая стратегия будет получать их с вероятностями 0.5.
\end{proof}
\end{proposition}

Таким образом, теорема \ref{th:inverserlassumptionexplanation} говорит, что в Maximum Entropy RL можно считать, что оптимизация движет нашу стратегию в сторону \eqref{meirl_kl}, и поэтому это предположение для моделирования экспертного поведения можно считать разумным, но в строгом смысле оно для оптимальных стратегий не выполняется.

\subsection{Guided Cost Learning}

Аппроксимируем функцию награды нейросетью $r_\theta(s, a)$. Тогда правдоподобие одной траектории в предположении \eqref{meirl} равно:
$$p_\theta(\Traj \mid \pi^*) = \frac{e^{R_{\theta}(\Traj)} \prod\limits_{t \ge 0}p(s_{t+1} \mid s_t, a_t)}{Z(\theta)}$$
где $R_{\theta}(\Traj) \coloneqq \sum\limits_{s, a \in \Traj} r_{\theta}(s, a)$ --- текущая аппроксимация кумулятивной награды, а нормировочная константа, внимание, зависит от параметров нейросети $\theta$:

\begin{equation}\label{irl_partionfunction}
Z(\theta) \coloneqq \int\limits_{\Traj} e^{R_{\theta}(\Traj)} \prod_{t \ge 0}p(s_{t+1} \mid s_t, a_t) \diff \Traj
\end{equation}

Рассмотрим логарифм правдоподобия:
\begin{equation}\label{irl_loglikelihood}
\log p_\theta(\Traj \mid \pi^*) = R_{\theta}(\Traj) + \underbrace{\sum_{t \ge 0} \log p(s_{t+1} \mid s_t, a_t)}_{\const(\theta)} - \log Z(\theta)
\end{equation}

Максимизация логарифма правдоподобия имеет интересную интерпретацию: нам нужно выдавать большую награду на тех состояниях, которые эксперт посетил (первое слагаемое) и маленькую награду <<всюду в остальных местах>> (второе слагаемое). Тогда интеграл $\log Z(\theta)$ будет уменьшаться.

Как оптимизировать это правдоподобие? С первым слагаемым всё понятно: награда, моделируемая нейросеткой, дифференцируема по параметрам. Проблема заключается в нормировочном слагаемом.

Нам понадобится следующий фокус. Пусть у нас есть некоторая функция награды с параметрами $\theta$. Пусть $\pi^*_{[\theta]}$ --- стратегия, которая оптимально (в терминах Maximum Entropy фреймворка, в рамках предположения \eqref{meirl}) оптимизирует вот эту награду, которую мы предлагаем с текущими параметрами $\theta$. Такая стратегия по определению будет в среде генерировать траектории из распределения
\begin{equation}\label{irl_currentoptimalpolicy}
p(\Traj \mid \pi^*_{[\theta]}) \coloneqq \frac{e^{R_{\theta}(\Traj)} \prod\limits_{t \ge 0}p(s_{t+1} \mid s_t, a_t)}{Z(\theta)}
\end{equation}
и это в точности есть правдоподобие траекторий эксперта при текущих значениях параметров $\theta$.

\begin{theorem}[Guided Cost Learning]
Градиент для оптимизации правдоподобия \eqref{irl_loglikelihood} траекторий эксперта по параметрам функции награды $\theta$ равен:
\begin{equation}\label{costguidedlearning}
    \E_{\Traj \sim \pi^*} \nabla_{\theta} R_{\theta}(\Traj) - \E_{\Traj \sim \pi^*_{[\theta]}} \nabla_{\theta} R_{\theta}(\Traj)
\end{equation}
\beginproof
Рассмотрим градиент логарифма правдоподобия одной траектории:
$$\nabla \log p_\theta(\Traj \mid \pi^*) = \nabla R_{\theta}(\Traj) - \nabla \log Z(\theta)$$

Мы хотим оптимизировать правдоподобие в среднем по траекториям эксперта, однако нам нужен градиент нормировочной константы (общий для всех траекторий эксперта, поэтому из мат.ожидания по ним эту константу можно вынести). Рассмотрим дифференцирование нормировочной константы отдельно:
\begin{align*}
\nabla \log Z(\theta) &= \\
= \{ \text{градиент логарифма} \} &= \frac{1}{Z(\theta)} \nabla Z(\theta) = \\
= \{ \text{определение $Z(\theta)$ \eqref{irl_partionfunction}} \} &= \frac{1}{Z(\theta)} \int\limits_{\Traj} \nabla_{\theta} e^{R_{\theta}(\Traj)} \prod_{t \ge 0}p(s_{t+1} \mid s_t, a_t) \diff \Traj = \\
= \{ \text{дифференцируем экспоненту} \} &= \frac{1}{Z(\theta)} \int\limits_{\Traj} e^{R_{\theta}(\Traj)} \prod_{t \ge 0}p(s_{t+1} \mid s_t, a_t) \nabla_{\theta} R_{\theta}(\Traj) \diff \Traj = \\
= \{ \text{определение $\pi^*_{[\theta]}$ \eqref{irl_currentoptimalpolicy} } \} &= \E_{\Traj \sim p(\Traj \mid \pi^*_{[\theta]})} \nabla R_{\theta}(\Traj)    \tagqed
\end{align*}
\end{theorem}

Отсюда напрашивается такая забавная идея. С первым мат.ожиданием в формуле градиента всё понятно: мы максимизируем награду в тех парах состояние-действие, которые встречались у эксперта. Затем мы берём и при помощи любого алгоритма Maximum Entropy RL оптимизируем нашу текущую награду $r_\theta(s, a)$, получая оптимальную относительно нашей текущей функции награды стратегию $\pi^*_{[\theta]}$ (точнее, её некоторое приближение). Отправляем её в реальную среду и собираем траектории $\Traj \sim \pi^*_{[\theta]}$. И чтобы получить второе слагаемое градиента, говорим, что в состояниях-действиях, которые встретились в траекториях этого <<псевдо-эксперта>>, награду нужно минимизировать. Если наша награда стала настоящей, псевдо-эксперт сойдётся к эксперту, и градиент станет нулевым.

Конечно, мы не будем обучать RL-алгоритм для оптимизации стратегии $\pi^*_{[\theta]}$ до сходимости, и вместо этого будем чередовать шаг оптимизации по параметрам $\theta$ функции награды при помощи формулы \eqref{costguidedlearning} (где второе мат.ожидание оценено при помощи сэмплов из текущей $\pi^*_{[\theta]}$) и шаг оптимизации параметров самой $\pi^*_{[\theta]}$ при помощи RL-алгоритма. В пределе мы потенциально сойдёмся к той функции награды, которой пользовался истинный эксперт, и заодно к стратегии, её максимизирующей.

\subsection{Generative Adversarial Imitation Learning (GAIL)}

Guided Cost Learning со своим чередованием двух шагов оптимизации напоминает игру двух игроков. Оптимизация по параметрам награды говорит уменьшать слагаемое, соответствующее нормировочной константе, а оптимизация стратегии --- увеличивать. У этого есть важная интерпретация, которая помогает понять, как на самом деле нужно обучаться с экспертных траекторий.

Сначала мы чуть-чуть перепишем нашу задачу оптимизации по $\pi$ и по $r$.

\begin{proposition}
Оптимизация функции награды по формуле \eqref{costguidedlearning} соответствует оптимизации следующего функционала:
\begin{equation}\label{irl_dual}
\E_{\Traj \sim \pi^*} R(\Traj) - \max_{\pi} \E_{\Traj \sim \pi} \sum_{t \ge 0} \left( r(s_t, a_t) + \entropy(\pi(\cdot \mid s_t)) \right) \to \max_{r}
\end{equation}
\begin{proof}[Пояснение]
Градиент максимума от функции есть градиент этой функции в точке максимума, а энтропия $\entropy(\pi(\cdot \mid s_t))$ не зависит от $r$, поэтому градиент второго слагаемого совпадает со вторым слагаемым \eqref{costguidedlearning}.
\end{proof}
\end{proposition}

\begin{definition}
\emph{Occupancy measure} для стратегии $\pi$ будем называть
\begin{equation}\label{occupancymeasure}
\rho_\pi(s, a) \coloneqq \pi(a \mid s)d_{\pi}(s),
\end{equation}
где $d_\pi(s)$ --- частоты посещения состояний \eqref{svd}.
\end{definition}

По определению, мат.ожидания по траектории с таким обозначением можно записывать как
$$\E_{\Traj \sim \pi} \sum_{t \ge 0} f(s_t, a_t) = \E_{\rho_{\pi}} f(s, a)$$

Давайте воспользуемся этим обозначением в \eqref{irl_dual}. Раскрывая определение мат.ожидания и переписывая оптимизацию по $\pi$ как минимизацию, получаем такую <<игру>> (читать --- поиск седловой точки):
\begin{equation}\label{game_pi_r}
\min\limits_{r} \max\limits_{\pi} \E_{\rho_{\pi}} r(s, a) - \E_{\rho_{\pi^*}} r(s, a) + \E_{\Traj \sim \pi} \sum_{t \ge 0} \entropy(\pi(\cdot \mid s_t))
\end{equation}

Сразу сделаем одно наблюдение: по определению occupancy measure однозначно задаёт стратегию $\pi$:

\begin{proposition}
\begin{equation}\label{policy_from_om}
\pi(a \mid s) = \frac{\rho_\pi(s, a)}{\int\limits_{\A} \rho_\pi(s, a) \diff a}
\end{equation}
\end{proposition}

Значит, можно поиск стратегии интерпретировать как поиск occupancy measure: действительно, в \eqref{game_pi_r} последнее слагаемое --- суммарный энтропийный бонус стратегии $\pi$ --- тоже можно переписать в терминах $\rho_\pi(s, a)$. Обозначим его как
$$\tilde{\entropy}(\rho_\pi) \coloneqq \E_{\Traj \sim \pi} \sum_{t \ge 0} \entropy(\pi(\cdot \mid s_t)) = \E_{\rho_\pi} -\log \pi(a \mid s) = \{ \text{\eqref{policy_from_om}} \} = \E_{\rho_\pi} -\log \frac{\rho_\pi(s, a)}{\int\limits_{\A} \rho_\pi(s, a) \diff a}$$

Получаем такой взгляд на процесс обучения:
\begin{equation}\label{game_pi_rho}
\min\limits_{r} \max\limits_{\rho_\pi} \E_{\rho_{\pi}} r(s, a) - \E_{\rho_{\pi^*}} r(s, a) + \tilde{\entropy}(\rho_\pi)
\end{equation}

Глядя на \eqref{game_pi_rho}, видно, что единственная возможная седловая точка --- это случай $\rho_{\pi} \HM\equiv \rho_{\pi^*}$: относительно $r(s, a)$ функционал внутри линеен\footnote{поскольку можно переписать мат.ожидания в следующем виде:
$$\E_{\rho_{\pi}} r(s, a) - \E_{\rho_{\pi^*}} r(s, a) = \int\limits_{\St} \int\limits_{\A} \left(\rho_{\pi}(s, a) - \rho_{\pi^*}(s, a)\right) r(s, a) \diff s \diff a$$
Видно, что это фактически линейный функционал $\langle \rho_{\pi} \HM- \rho_{\pi^*},  r\rangle$ в пространстве функций $\St \HM\times \A \HM\to \R$.}, как только в какой-то паре $s, a$ разница $\rho_{\pi} \HM - \rho_{\pi^*} \HM\neq 0$, функция награды может начать выдавать бесконечно большие значения соответствующего знака. На энтропийное слагаемое $\tilde{\entropy}(\rho_\pi)$ можно смотреть как на регуляризатор по частотам посещения пар состояния-действия. Минимизация по $\rho_\pi$ с точностью до этого энтропийного регуляризатора --- это поиск такого $\rho_\pi(s, a)$, который наиболее похож на $\rho_{\pi^*}(s, a)$.

Другими словами: задача обучения по экспертным траекториям заключается не в том, чтобы матчить нашу стратегию со стратегией эксперта, а в том, чтобы заматчить occupancy measure --- частоты посещений пар состояние-действие. Это сильно разные вещи: выбор одних действий в одной области пространства состояний влияет на occupancy measure в других! Если мы матчим стратегии, то, вероятно, мы просто во всех состояниях пытаемся сделать нашу стратегию похожей на стратегию эксперта. А на самом деле мы хотим с одинаковой с экспертом частотой во все состояния попадать! И это значит, что имитационное обучение --- это матчинг распределений $p(\Traj \HM\mid \pi^*)$ и $p(\Traj \HM\mid \pi)$!

Это рассуждение открывает следующую идею: давайте при обучении награды мы будем напрямую пытаться отличать распределения $\rho_{\pi^*}(s, a)$ и $\rho_{\pi}(s, a)$. Как различать распределения? Вспоминаем генеративно-состязательные сети (GAN). Там дискриминатор неявно учил различать, приходят ли сэмплы из одного класса или из другого, просто решая задачу бинарной классификации.

Но как GAN-ы связаны с обратным обучением с подкреплением, с обучением награды? Оказывается, это одно и то же. Чтобы увидеть это в чистом виде, сделаем следующий важный шаг: мы добавим регуляризатор и для оптимизации по $r$. Это можно мотивировать тем, что, как мы обсуждали ранее, задача обратного обучения с подкреплением <<некорректна>> и может иметь много разных решений. Итак, добавим некоторое слагаемое $\psi(r)$, которое будет смотреть на нашу выдаваемую функцию награды и некоторым образом её дополнительно штрафовать:

\begin{equation}\label{game_pi_rho_reg}
\min\limits_{r} \max\limits_{\rho_\pi} \psi(r) + \E_{\rho_{\pi}} r(s, a) - \E_{\rho_{\pi^*}} r(s, a) + \tilde{\entropy}(\rho_\pi)
\end{equation}

Оказывается, ванильный GAN, различающий сэмплы пар $s, a$ из распределений $p(\Traj \HM\mid \pi^*)$ и $p(\Traj \HM\mid \pi)$, соответствует просто определённому выбору регуляризатора!

\begin{theorem}
Выберем в \eqref{game_pi_rho_reg} следующий регуляризатор:
\begin{equation}\label{gail_reg}
\psi(r) \coloneqq \E_{\rho_{\pi^*}} r(s, a) + \log (1 - e^{-r(s, a)}),
\end{equation}
где штраф полагается бесконечно большим, если $r(s, a) \HM\le 0$. Тогда задача \eqref{game_pi_rho_reg} примет вид:
\begin{equation}\label{GAIL}
\min\limits_{D} \max\limits_{\pi} -\E_{\rho_{\pi^*}} \log (1 - D(s, a)) - \E_{\rho_{\pi}} \log D(s, a) + \tilde{\entropy}(\rho_\pi),
\end{equation}
где $D(s, a) \in (0, 1)$.

\beginproof
Подставим в \eqref{game_pi_rho_reg} выбранный регуляризатор:
\begin{align*}
&\, \min\limits_{r} \max\limits_{\pi} \psi(r) + \E_{\rho_\pi} r(s, a) - \E_{\rho_{\pi^*}} r(s, a) + \tilde{\entropy}(\rho_\pi) = \\
&= \min\limits_{r} \max\limits_{\pi} \E_{\rho_{\pi^*}} \log (1 - e^{-r(s, a)}) + \E_{\rho_\pi} r(s, a) + \tilde{\entropy}(\rho_\pi)
\end{align*}

Сделаем замену переменных: вместо оптимизации по $r(s, a) \HM > 0$ будем оптимизировать по $D(s, a)$, где $D(s, a) \HM \coloneqq e^{-r(s, a)}$ --- произвольное число в диапазоне (0, 1). Тогда $r(s, a) \HM= -\log(D(s, a))$.
\begin{equation*}
\min\limits_{D} \max\limits_{\pi} -\E_{\rho_{\pi^*}} \log (1 - D(s, a)) - \E_{\rho_\pi} \log D(s, a) + \tilde{\entropy}(\rho_\pi)   \tagqed
\end{equation*}
\end{theorem}

Итак, что мы получили в формуле \eqref{GAIL}: вместо награды будем обучать дискриминатор, решающий задачу бинарной классификации, где пары $s, a$ из $\rho_{\pi^*}$ образуют класс 1, а пары $s, a$ из $\rho_\pi$ --- класс 0. Действительно, оптимизация \eqref{GAIL} по $D$ при фиксированной $\pi$ выглядит так:
$$\E_{\rho_{\pi^*}} \log (1 - D(s, a)) + \E_{\rho_{\pi}} \log D(s, a) \to \max\limits_{D},$$
то есть мы просто учимся отличать пары $s, a$, порождённые (встреченные) экспертом от тех, что встречает наша текущая стратегия. 

Оптимизация же по $\pi$ (давайте вернёмся к оптимизации по стратегии) при фиксированном <<дискриминаторе>> $D$ выглядит так:
$$
\E_{\Traj \sim \pi} \sum_{t \ge 0} \left[ -\log D(s_t, a_t) + \entropy(\pi(\cdot \mid s_t)) \right] \to \max\limits_{\pi}
$$
То есть величина $-\log D(s, a)$ и будет являться наградой за шаг: именно такую кумулятивную дисконтированную награду (плюс энтропийный бонус) оптимизирует стратегия $\pi$, она же <<генератор>> в этой схеме. Такой генератор учится порождать $s, a$, которые дискриминатор-награда не отличает от экспертных пар, только в отличие от обычного генератора здесь наши сгенерированные действия влияют на следующие состояния, и поэтому мы генерируем не <<отдельные>> сэмплы $s, a$, а целые цепочки траекторий $\Traj \HM\sim \pi$. 

\subsection{Generative Adversarial Imitation from Observation (GAIfO)}

Часто эксперт предоставляет нам траектории, в которых отсутствует информация о совершённых агентом действиях: есть лишь цепочки состояний $s_0, s_1, s_2, \dots$, которые посещал эксперт. Такая задача называется \emph{имитационным обучением по наблюдениям} (imitation learning from observations).

\begin{example}
Допустим, у вас есть покадровые анимации того, как персонаж делает сальто. Вы хотите научить робота с такими же конечностями делать тоже самое. В анимациях понятно, в каких координатах находились все конечности персонажа в каждый момент времени, и можно считать, что робот при выполнении задачи должен <<посещать>> те же цепочки состояний. Однако в анимации действий нету --- вы не знаете, как нужно управлять роботом, чтобы получить ту же траекторию в реальной среде.
\end{example}

Оказывается, идея GAIL очень легко и просто расширяется на такую ситуацию. Мы просто хотим попадать в те же состояния, в которые попадал эксперт, поэтому дискриминатором $D(s) \HM \in (0, 1)$ теперь будем пытаться различать состояния --- порождены ли они экспертом $s \HM\sim d_{\pi^*}(s)$ или же нашей текущей стратегией $s \HM\sim d_{\pi}(s)$.
$$
\min\limits_{D} \max\limits_{\pi} -\E_{d_{\pi^*}(s)} \log (1 - D(s)) - \E_{d_{\pi}(s)} \log D(s) + \E_{d_\pi(s)}\entropy(\pi(\cdot \mid s))
$$

В методе Generative Adversarial Imitation from Observation (GAIfO) предлагается сделать хитрость, и различать не состояния, а пары <<состояние-следующее состояние>> $s, s'$. Другими словами, награда полагается зависящей от пары состояний $r(s, s')$, а дискриминатор $D(s, s')$ учится различать именно пары $s, s'$ из экспертных траекторий и из порождаемых траекторий. Формально говоря, вводится альтернативное определение occupancy measure как вероятность встретить ту или иную пару $s, s'$ в траекториях из стратегии $\pi$:
$$
\nu_\pi(s, s') \coloneqq d_\pi(s) \int_\A p(s' \mid s, a) \pi(a \mid s) \diff a
$$

Тогда минимаксная задача оптимизации принимает следующий вид:
$$
\min\limits_{D} \max\limits_{\pi} -\E_{\nu_{\pi^*}} \log (1 - D(s, s')) - \E_{\nu_\pi} \log D(s, s') + \E_{d_\pi(s)}\entropy(\pi(\cdot \mid s))
$$

\section{Внутренняя мотивация}\label{subsec:intrinsic_motivation}

\subsection{Вспомогательные задачи}

В обучении с подкреплением типична ситуация разреженной функции награды, когда агенту редко поступает сигнал от среды. Например, в худшем случае функция награды представляет собой константу, имеющую смысл <<штрафа за потерю времени>>, а в конце эпизода нам приходит условно +1, если задача была успешно решена.  

\begin{definition}
Задачей \emph{поиска} будем называть задачу со следующей функцией награды:
\begin{equation}\label{searchreward}
r(s, a) = \begin{cases}
+1 \quad & s \in \St^+ \\
\const \quad & s \not\in \St^+ \\
\end{cases}    
\end{equation}
где $\St^+$ --- множество терминальных состояний, $\const \le 0$ --- штраф за потерю времени.
\end{definition}

Задача поиска сложна тем, что сигнала от среды нет. Пока мы не решим задачу, оптимизируемый функционал представляет собой плато, а наши RL алгоритмы, как и любые методы оптимизации, работают за счёт разницы в сигнале.

Что в таких ситуациях делать? Первая возможность --- учить модель динамики среды, если она неизвестна. Найти сигнал от среды это скорее всего не поможет, поскольку экспоненциальный перебор всевозможных будущих траекторий не сильно лучше случайного блуждания в среде.

Мы придумаем себе другую, вспомогательную задачу, которую мы сможем решать в \emph{self-supervised} режиме --- режиме, не требующем никакого сигнала от среды, никакой <<разметки>>.

\begin{definition}
Для данной среды $(\St, \A, \Trans)$ \emph{вспомогательной задачей} называется задача обучения с подкреплением для MDP $(\St, \A, \Trans, r^{\intr})$, где $r^{\intr}$ --- \emph{внутренняя мотивация} (intrinsic motivation) или \emph{внутренняя награда} (intrinsic reward function), определяемая самим агентом.
\end{definition}

Слова <<определяемая самим агентом>> означает, что эта функция нам не дана. Исходная задача, которую мы хотим решить, состоит в оптимизации в данной среде некоторой \emph{внешней функцией награды} (extrinsic reward function) $r^{\extr}$, также называемой \emph{внешней мотивацией} (extrinsic motivation). Однако, если эта функция награды, например, всегда константна, как в задаче поиска \eqref{searchreward}, то нам необходимо откуда-то взять какой-то другой обучающий сигнал. Этот сигнал нам придётся придумать <<самим себе>>, при помощи нового модуля в обучающейся системе.

Какой обучающий сигнал мы хотим получить? Во-первых, плотный, чтобы было, на чём обучаться базовому алгоритму. Во-вторых, осмысленный: награждающий за развитие каких-то <<полезных>> в среде навыков, которые могут пригодиться для взаимодействия, условно, вне зависимости от того, какой на самом деле окажется та внешне мотивированная задача, которую агент призван решать.

\begin{example}
Идея, можно сказать, вдохновлена подобной <<внутренней наградой>> человека, поощряющей такое поведение, как игра, любопытство, стремление к познанию. В ходе игрового поведения, человек улучшает своё представление о том, как работает мир вокруг него, по каким законам он устроен и как он своими действиями может влиять на будущее состояние мира и вызывать те или иные явления. Такое более интеллектуальное <<самообучение>> не только ускоряет поиск сигнала от среды, но и позволяет агенту выработать широкий набор умений, которые скорее всего окажутся полезны для достижения заданной внешней наградой цели, какой бы она ни оказалась.
\end{example}

Придумать в общем случае такой сигнал непросто, и могут возникать ситуации, когда внутренняя мотивация <<промотивирует>> агента <<залипнуть>> в какой-то области среды.

\begin{definition}
Проблемой \emph{прокрастинации} (procrastination) называется ситуация, когда в некоторой области в среде внутренняя мотивация выдаёт высокий (и не снижающийся с течением обучения) сигнал, перебивающий остальные мотивации агента.
\end{definition}

\subsection{Совмещение мотиваций}

По умолчанию всегда считается, что внешняя и внутренняя мотивация складываются:
\begin{equation}\label{sumgoal}
r(s, a) \coloneqq r^{\extr}(s, a) + \alpha r^{\intr}(s, a),
\end{equation}
где $\alpha$ --- масштабирующий гиперпараметр. Для простоты далее будем считать $\alpha \HM= 1$.

Заданная так награду можно оптимизировать любым <<базовым>> алгоритмом RL, ничего не подозревающим о разложении. Но понятно, что агенту доступно каждое слагаемое по отдельности (коли внешняя награда выдаётся средой, а внутренняя генерируется внутри самого алгоритма); как мы можем это использовать?

Пусть оценочные функции с индексом $\extr$ соответствуют оценочным функциям для внешней мотивации, с индексом $\intr$ --- внутренней мотивации, без индекса --- суммарной мотивации. Тогда:
\begin{proposition}
\begin{equation}\label{valuefunctionssum}
\begin{aligned}
Q^\pi(s, a) &= Q^\pi_{\extr}(s, a) + Q^\pi_{\intr}(s, a) \\
V^\pi(s) &= V^\pi_{\extr}(s) + V^\pi_{\intr}(s)
\end{aligned}
\end{equation}
\begin{proof}
По определению.
\end{proof}
\end{proposition}

\needspace{5\baselineskip}
\begin{wrapfigure}{r}{0.35\textwidth}
\centering
\includegraphics[width=0.35\textwidth]{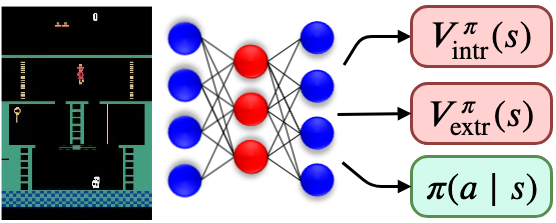}
\end{wrapfigure}

Итак, если в алгоритме учится оценочная функция $V^\pi$ или $Q^\pi$, то можно учить оценочные функции каждого слагаемого в награде по отдельности, например, в виде отдельной головы для каждой мотивации. 

\begin{remark}
Этим фактом можно также пользоваться в ситуациях, когда награда представлена в виде суммы нескольких слагаемых (очень типичная ситуация), и агенту доступно полное разложение на эти слагаемые. Это позволит по ходу оптимизационного процесса <<включать>> и <<отключать>> мотивации, что бывает очень удобно.
\end{remark}

Трюк можно применять даже в value-based методах, где мы учим $Q^*$, поскольку тот процесс можно интерпретировать как оценивание (обучение $Q^\pi$) текущей стратегии $\pi \HM= \argmax\limits_a Q(s, a)$, хотя формально для истинной оптимальной оценочной функции $Q^*$ разложение неверно: 

\begin{proposition}
В общем случае аналогичное разложение для $Q^*, V^*$ неверно.
\begin{proof}
Максимум суммы может быть меньше суммы максимумов. Действительно, пусть первое действие даёт внешнюю +1, а второе действие даёт внутреннюю +1 (иначе по нулям); тогда оптимальные оценочные функции равны $V^*_{\intr}(s) = V^*_{\extr}(s) = +1$, но оптимальная V-функция для суммы наград равна не 2, а 1, поскольку выбрать одновременно два действия нельзя и между мотивациями придётся выбирать.
\end{proof}
\end{proposition}

В Distributional-подходе при использовании этой идеи придётся предполагать независимость внутренней и внешней мотивации (что практически всегда неверно).

\begin{proposition}
В общем случае аналогичное разложение для $\Z^\pi$ неверно.
\begin{proof}
Это связано с тем, что внутренняя и внешняя мотивация могут быть скоррелированы, и этой информации при доступе к отдельным $\Z^\pi_{\extr}$ и $\Z^\pi_{\intr}$ у нас нет. Действительно: пусть известно, что для некоторых $s, a$ с вероятностью 0.5 $\Z^\pi_{\extr}(s, a)$ принимает значение +1, а иначе 0, и тоже самое для $\Z^\pi_{\intr}(s, a)$. Мы могли бы сказать, что распределение суммы $\Z^\pi_{\extr}(s, a) + \Z^\pi_{\intr}(s, a)$ имеет такой вид: с вероятностью 0.25 мы получим +2, с вероятностью 0.5 получим +1, и с вероятностью 0.25 не получим ничего. Однако, тогда мы предполагаем независимость этих случайных величин, а это может быть неправдой: например, вдруг мы получаем внешнюю +1 тогда и только тогда, когда получаем внутреннюю +1. Тогда $\Z^\pi(s, a)$ суммы мотиваций есть на самом деле +2 с вероятностью 0.5 и +0 иначе.
\end{proof}
\end{proposition}

\begin{proposition}
В общем случае аналогичное разложение для $\Z^*$ неверно.
\end{proposition}

Разложение также позволяет использовать разные коэффициенты дисконтирования для разных мотиваций. Допустим, агент учит отдельно оценочную функцию будущей внутренней мотивации $V^{\intr}$ и будущей внешней мотивации $V^{\extr}$. Тогда можно использовать разное дисконтирование $\gamma^{\intr}$ и $\gamma^{\extr}$ при их обучении. Для примера рассмотрим одношаговые таргеты:
$$y^\extr \coloneqq r + \gamma^{\extr} V^{\extr}(s')$$
$$y^{\intr} \coloneqq r + \gamma^{\intr} V^{\intr}(s')$$

Разложение также позволяет во внутренней мотивации игнорировать понятия эпизодов: это означает, что агент рассматривает весь процесс решения вспомогательной задачи как один большой никогда не заканчивающийся эпизод, не используя флаги $\done$ из среды при обучении $V^{\intr}$. Такой агент учитывает, что если он окажется в терминальном состоянии, то произойдёт сброс среды, и он окажется в стартовом состоянии $s_0 \HM\sim p(s_0)$, для которого значение $V^{\intr}(s)$ может быть высоким. Это может мотивировать агента прерывать текущий <<неудачный>> эпизод ради того, чтобы начать новый, что иногда может быть полезно.

\vspace{0.2cm}
\begin{example}[Агент в яме]
В среде существует некоторая труднодоступная область, которую агент внутренне мотивирован посетить. Агент предпринимает попытку добраться до области, но падает в яму. Оценочная функция для внешней мотивации знает, что из ямы уже не выбраться, и оценивает действие <<закончить эпизод>> как негативное (<<смерть>>), остальные действия как безрезультатные (+0). Оценочной функции для внутренней мотивации, допустим, известно, что из ямы уже не выбраться, и, в случае, если она обучается эпизодично, оценивает любые действия как безрезультатные (+0). Если же оценочная функция игнорирует понятие эпизодов, агент знает, что он может произвести сброс среды и, в частности, попробовать ещё раз добраться до труднодоступной области. Это может промотивировать агента не сидеть в яме, оттягивая негативный, но неизбежный эффект смерти, а приступить к следующему эпизоду обучения.
\end{example}

\subsection{Exploration Bonuses}

Как строить $r^{\intr}$? Что должен делать агент, который попал в среду, и не получает никакого внешнего сигнала? На этот вопрос можно ответить по-разному.

\begin{exampleBox}[label=ex:chaosminimization]{Минимизация хаоса (chaos minimization)}
Нужно искать наиболее <<безопасные>>, стабильные области пространства состояний, где будущее наиболее предсказуемо, <<избегать сюрпризов>>. Интуицией в такой вспомогательной задаче является идея о том, что многие изобретения человечества были созданы для защиты от сюрпризов, которые, зачастую, неприятны. Существуют среды, например, тетрис, где задача <<минимизации хаоса>> коррелирует с задачей самой игры; агент, решающий такую вспомогательную задачу без доступа к внешней функции награды, <<неявно>> решает исходную задачу.
\end{exampleBox}

Мы далее рассмотрим другой ответ --- заниматься исследованием. Отчасти эта задача полностью противоположна минимизации хаоса: однако задачи не противоречат друг другу, ведь чтобы найти самую <<спокойную>> область среды и научиться до неё добираться, агенту потребуется заняться исследованием окружения и поиском этих самых стабильных областей. В этом смысле, задача исследования, вероятно, является наиболее универсальной вспомогательной задачей, которую агент может себе поставить в среде.

Мы постоянно встречались с дилеммой исследования-использования по ходу пьесы, однако теперь, когда оптимизировать внешний сигнал у нас не получается ввиду его отсутствия, <<использовать>> нам абсолютно нечего, и дилемма не стоит. Итак, считаем, что нам дана среда и есть задача <<заисследовать её>>. Как формализовать такую задачу в терминах награды?

\needspace{5\baselineskip}
\begin{wrapfigure}{r}{0.3\textwidth}
\vspace{-0.4cm}
\centering
\includegraphics[width=0.3\textwidth]{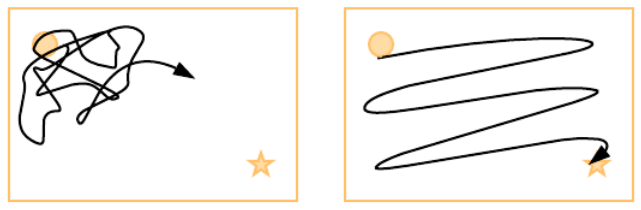}
\vspace{-0.5cm}
\end{wrapfigure}

Понятно, что случайная стратегия, которой мы часто пользовались до этого для <<исследований>>, является теоретическим решением (например, с точки зрения теоремы \ref{th:TDconvergencetheorem}). Но интуитивно, куда более оптимальным поведением является некий интеллектуальный перебор состояний в среде.

Мы уже встречались с \emph{исследовательскими бонусами} (exploration bonus) в контексте UCB-бандитов (раздел \ref{subsec:ucb}): там мы добавляли к нашей оценке Q-функции некоторое слагаемое, имевшее смысл <<награды за то, что действие редко пробовалось в прошлом>>. Наша внутренняя мотивация тоже есть такая добавка, только теперь она должна оценивать новизну посещаемых областей в среде.

Попробуем исходить из схожих соображений: будем награждать агента за посещения тех состояний, в которых он был редко. Мы можем это сделать двумя способами.

\begin{definition}
Пусть $h(s) \colon \St \to \{0, 1 \dots N\}$ --- некоторая хэш-функция состояний, называемая \emph{оракулом} (oracle), и $n(i)$ --- счётчик, сколько раз за время всего обучения нам встретились состояния с хэшем $i$. Тогда
$$r_{\intr}(s, a) \coloneqq \frac{1}{n(h(s))}$$
называется \emph{нестационарным} исследовательским бонусом; награда
$$r_{\intr}(s_t, a_t) \coloneqq \mathbb{I}[\forall t' < t \colon s_t \neq s_{t'}],$$
то есть награждение +1, если мы попали в состояние, хэш для которого $h(s_t)$ не встречался до этого в течение данного эпизода, называется \emph{эпизодичным} исследовательским бонусом.
\end{definition}

Нестационарные исследовательские бонусы затухают с ходом обучения; в пределе мы, надеемся, посетим все состояния достаточное число раз, внутренняя мотивация затухнет и мы переключимся на внешнюю мотивацию. Плохо это тем, что такая мотивация нарушает стационарность формализма MDP, так как с ходом обучения меняются счётчики посещения. Это довольно типично, что внутренняя мотивация нестационарна: модуль внутренней мотивации принципиально есть часть обучающейся системы, и он тоже постепенно <<обучается>>, следовательно, меняется. Для нас это значит, что нужно будет использовать on-policy алгоритмы для обучения на такой сигнал.

Эпизодичные бонусы, конечно же, можно считать модификацией функции награды, и поэтому подобные оракулы можно считать <<ручными эвристиками>>. Агент в том числе по итогам обучения научится в ходе одного эпизода <<бегать по всему MDP>>. Это, однако, вполне может быть полезно в каких-нибудь лабиринтах или задачах, где агенту нужно что-то где-то найти в течение самой игры. Проблема эпизодичных бонусов в том, что они формально нарушают предположение о полной наблюдаемости пространства состояний: функция награды зависит от всей прошлой истории посещения состояний в течение эпизода, и нам по-хорошему нужно переходить в формализм PoMDP.

\begin{exampleBox}[righthand ratio=0.25, sidebyside, sidebyside align=center, lower separated=false]{}
В табличных MDP, где $|\St| < +\infty$, хэш-функция по сути не нужна: $h(s) = s$. В ряде сред подобный оракул можно придумать эвристически; например, если у агента есть понятие <<координат>>, можно разделить пространство сеточкой, то есть поделив его на условные блоки, и награждать агента за <<посещение большого числа блоков>>. 

\tcblower
\includegraphics[width=\textwidth]{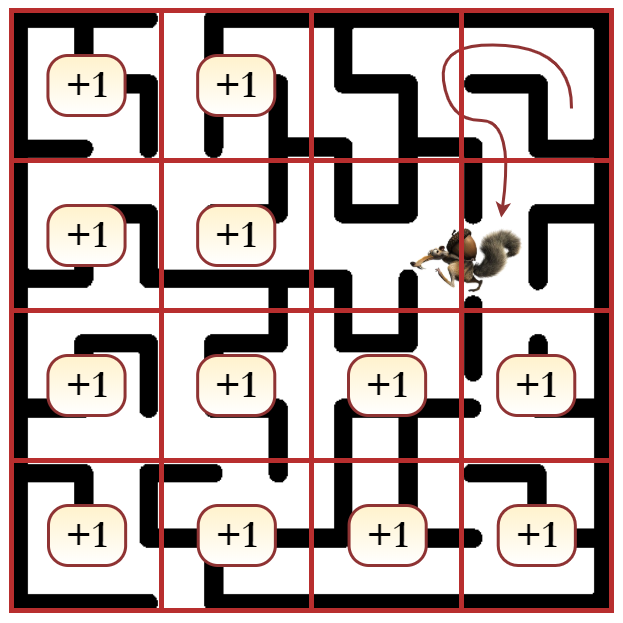}
\end{exampleBox}

\needspace{5\baselineskip}
\subsection{Дистилляция случайной сетки (RND)}

\begin{wrapfigure}{r}{0.55\textwidth}
\vspace{-0.4cm}
\centering
\includegraphics[width=0.55\textwidth]{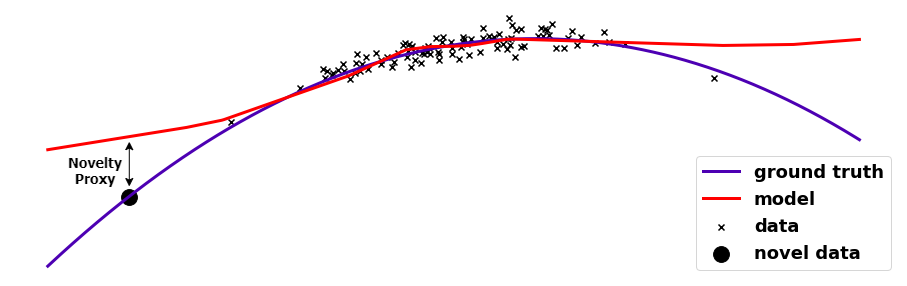}
\vspace{-0.5cm}
\end{wrapfigure}

Понятно, что в общем случае хэш-функцию придумать сложно, и нам нужен какой-то более универсальный способ оценки новизны. Мы воспользуемся трюком из машинного обучения: известно, что если некоторая модель обучалась решать задачу регрессии, то на входных данных, не похожих на примеры из обучающей выборки, ошибка её прогнозов будет больше. Отличие предсказания модели от истинного значения целевой переменной может быть использовано как приближение \emph{оценки новизны} (novelty esimation) или \emph{аномальности} данных.

Давайте возьмём какую-нибудь случайную задачу регрессии с состояниями в качестве входов. Нам важно лишь, что значение целевой переменной должно определяться по входному состоянию детерминировано и стационарно: ведь если таргет не детерминирован, то ошибка предсказывающей модели для состояния $s$ будет (если, например, модель обучается на минимизацию MSE) в среднем равна дисперсии. В такой ситуации сигнал внутренней мотивации будет выше в тех областях пространства состояний, где дисперсия целевой переменной выше. Нестационарность, очевидно, нарушает идею подхода, поскольку ошибка модели будет связана с изменением целевой переменной, а не новизной входного состояния. 

Пусть $\phi \colon \St \to \R^d$ --- некоторая функция, строящая эмбеддинги для состояний; например, \emph{случайная сеть} (random network) --- нейросеть со случайно инициализированными весами, которые не обучаются и никак не изменяются. Пусть $f \colon \St \to \R^d$ учится предсказывать выход $\phi$, используя в качестве обучающей выборки её значения на встречавшихся в ходе обучения состояниях.

\begin{definition}
Задача обучения одной нейросети $f$ на входах-выходах другой (заданной, фиксированной) нейросети $\phi$
$$\E_s \|f(s) - \phi(s)\|_2^2 \to \min_{f},$$
где мат.ожидание $\E_s$ берётся по произвольному буферу, называется \emph{дистилляцией} (distillation).
\end{definition}

Коли ошибка выше там, где новее состояния, мы можем использовать это значение в качестве обучающего сигнала:
$$r^{\intr}(s) \coloneqq \|f(s) - \phi(s)\|_2^2$$
Интуиция понятна: если мы <<впервые>> увидели состояние $s$, модель $f$ ещё ни разу не видела, какой эмбеддинг выдаёт на нём наша случайная сеть $\phi$, и поэтому скорее всего ошибётся. Такой сигнал будет мотивировать агента отправиться в ту область среды, где обучающаяся нейросеть плохо предсказывает выход случайно проинициализированной нейросети.

\begin{remark}
Здесь и в аналогичных местах далее использование MSE не принципиально (можно использовать любую функцию потерь для регрессии). В частности, можно одну метрику использовать для функции потерь, и другую --- для расчёта внутренней мотивации.
\end{remark}


\begin{remark}
Авторы также предложили нормировать сигнал внутренней мотивации на его бегущее среднее отклонение, то есть <<в среднем>> выдавать некоторую константу. Это существенно упрощает масштабирование внутренней мотивации относительно внешней, но тогда <<исследование>> не будет естественно затухать с ходом обучения.
\end{remark}

\subsection{Любопытство}

\begin{definition}
\emph{Любопытством} (curiosity) называется ошибка модели мира агента.
\end{definition}

Ошибка свидетельствует о том, что агент не всё знает о той области пространства состояний, в которой был произведён неверный прогноз. Использование любопытства как внутренней мотивации приводит к тому, что агент стремится в те области среды, которые ему <<непонятны>>, в частности те, которые для агента новы, и <<дообучить>> своё представление о мире. Зачастую выбор действий, максимизирующих ошибку модели мира приводит к \emph{возникновению игрового поведения} (emergence of playing behavior), когда агент <<играет>> с доступными для взаимодействия предметами.

Любопытство и построение внутренней мотивации на основе новизны состояний немного различаются. То, что состояние ново, не означает, что оно <<незаисследовано>>, что мы ничего о нём не знаем; мы вполне можем обобщиться за счёт прошлого опыта и спокойно ориентироваться даже в том состоянии, которое технически увидели впервые.

\begin{example}[Бесконечные двери]
В ряд стоит бесконечное количество одинаковых дверей, за каждой из которых ничего нет. Агенту доступен на вход, помимо прочего, номер двери. Сигнал, построенный на основе оценки новизны состояний, будет поощрять обнаружение новых дверей, поскольку
их номер <<нов>> по сравнению со старыми. Любопытство же через некоторое время научится предсказывать, что при движении вдоль ряда агенту будут встречаться точно такие же двери, как и раньше, и что за дверьми ничего нет; сигнал затухнет. При этом, любопытство среагирует на любое изменение в описании двери (если все двери были, например, красными, а тут вдруг бац, и дверь синяя), или если за очередной дверью окажется что-то непредсказуемое.

\begin{center}
\vspace{0.2cm}
    \includegraphics[width=0.7\textwidth]{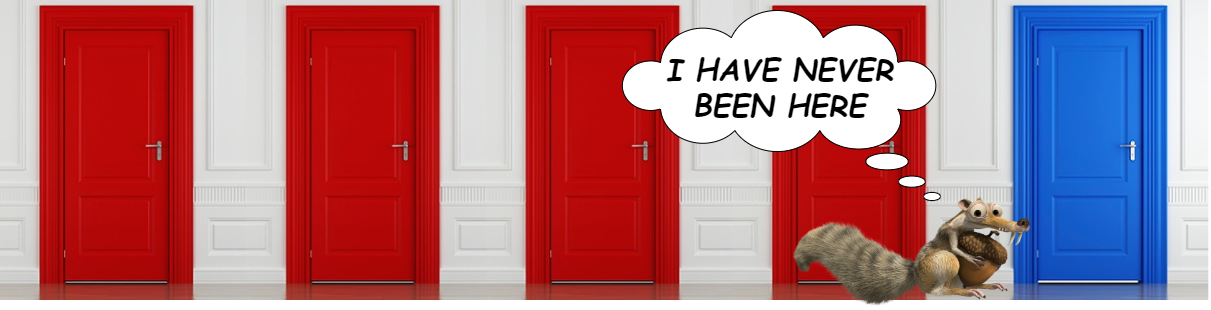}
    \includegraphics[width=0.7\textwidth]{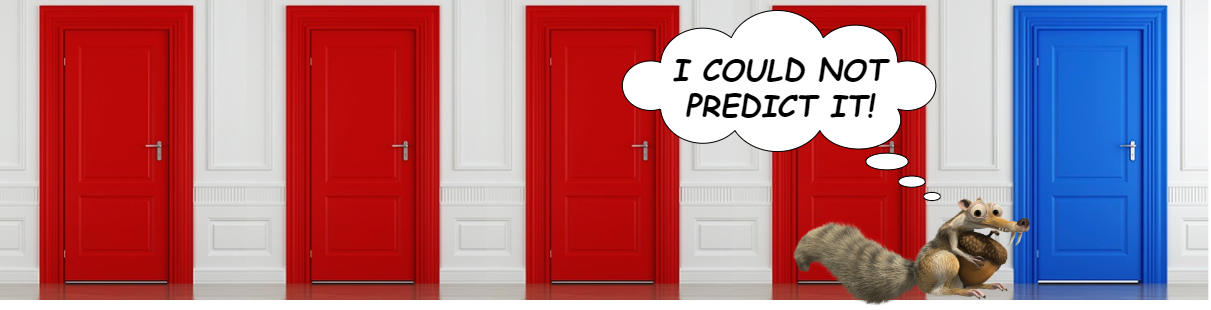}
\vspace{-0.2cm}
\end{center}

На самом деле, разница достаточно условна: предполагается, что модель предсказания будущего способна «обобщаться» на новые состояния, что, в целом, может оказаться верным и для модели оценки новизны. Так, в приведённом примере модель оценки новизны также может перестать оценивать увеличение номера двери как «новые» состояния, и тогда поведение этих двух видов внутренней мотивации станет схожим.

\end{example}

Пусть внутри агента строится модель прямой динамики:
$$\E_{s,a,s'} \|f(s, a) - s'\|^2_2 \to \min_f$$
Тогда во время очередного обучающего эпизода ошибка модели $f$ на каждом шаге может рассматриваться как любопытство и задавать внутреннюю мотивацию агента:
\begin{equation}\label{curiosity}
r^{\intr}(s, a, s') \coloneqq \|f(s, a) - s'\|^2_2
\end{equation}

Такой сигнал будет мотивировать агента искать не новые области, а те, в которых он не понимает, как работает среда. В некоторых средах такой сигнал, однако, может привести к прокрастинации.

\begin{definition}
\emph{Шумным телевизором} (noisy TV) в среде называются принципиально непредсказуемые в силу стохастичности функции переходов явления.
\end{definition}

\begin{exampleBox}[righthand ratio=0.2, sidebyside, sidebyside align=center, lower separated=false]{Шумный телевизор}
В среде стоит сломавшийся телевизор, демонстрирующий гауссовский шум. На каждом шаге шум сэмплируется заново. Предсказывать следующее значение шума по предыдущему в силу независимости сэмплов невозможно. В рамках концепции любопытства, агент мотивирован найти подобный <<шумный телевизор>> в среде (что может быть непросто) и получать наслаждение от непредсказуемости своих будущих наблюдений.

\tcblower
\includegraphics[width=\textwidth]{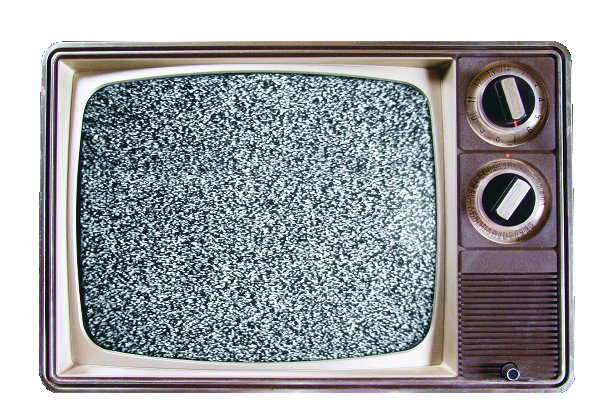}
\end{exampleBox}

Шумные телевизоры по определению нерелевантны: они не имеют отношения к истинной задаче, и <<отвлекают>> агента, когда ошибка модели мира принципиально не снижаема. Агент, мотивированный любопытством находить в среде шумные телевизоры --- это типичный пример прокрастинации.

\begin{example}
<<Шумные телевизоры>> могут принимать самые разные формы: например, у агента в некоторой области пространства состояний сломались сенсоры, и зашумляются гауссовским шумом. Или, например, среда отправляет агента в одну комнату или в другую с вероятностью 0.5; агент не может предсказать, куда именно его перекинет. В видеоиграх могут встречаться всякие декоративные рандомизированные спецэффекты, не имеющие отношения к самой задаче.
\end{example}

С одной стороны, проблема проистекает из того, что мы приближаем стохастичную динамику среды детерминированной моделью. Если стохастика среды существенно влияет на будущее агента и его путь к целевым состояниям, знание и понимание вероятностей различных исходов и их состав, очевидно, является ценным знанием об окружающем мире. Но если задуматься, строить генеративную модель сложного мира --- по сути, построить модель вселенной --- очень сложная задача, и шумные телевизоры скорее всего задаются сложным распределением, которое в принципе будет плохо поддаваться изучению. В совокупности с нерелевантностью шумных телевизоров, смысла заниматься этим нет, и хочется как-то избежать попыток предсказывать будущие состояния шумных телевизоров вовсе\footnote{поэтому считается, что сигналом любопытства должна быть не столько ошибка модели, сколько её изменение после дообучения. Иными словами, если модель продолжает ошибаться, <<понимать явление>> в среде не удаётся, и тогда необходимо почувствовать разочарование (убрать мотивационный сигнал) и бросить силы на исследование других областей среды. Тогда, если возле шумного телевизора ошибка модели мира не падает, агент перестанет на него отвлекаться. Однако построить масштабируемый алгоритм, эффективно замеряющий, как много информации о модели мира предоставил очередной переход $(s, a, s')$, чтобы превратить это значение во внутреннюю награду, довольно сложно, и далее мы разберём модуль любопытства, который основан на определении через абсолютное значение ошибки модели.}.

\begin{example}
Человек не пытается моделировать все без исключения окружающие сложные процессы, например, предсказывать поведение (траектории) всех наблюдаемых капель дождя или опадающих листьев. Вероятно, именно поэтому на них так легко <<залипнуть>>.
\end{example}


\subsection{Модель обратной динамики}

\begin{definition}
Модель, аппроксимирующая $p(a \HM\mid s, s')$, называется \emph{моделью обратной динамики} (inverse dynamics model).
\end{definition}

Такая модель тоже представляет собой пример модели мира. Агент, делая шаг в среде, может проверить, а правда ли он может по текущему состоянию $s'$ и предыдущему состоянию $s$ восстановить только что выбранное им действие. Отрицательный ответ может соответствовать ситуации, когда агент открыл новые явления в среде, попал в новую область пространства состояний. В таком случае, ошибка модели обратной динамики может быть использована в качестве любопытства.

Задача предсказывать действие по состоянию и следующему состоянию --- самая обычная задача классификации для дискретных пространств действий и задача регрессии для непрерывных пространств. Преимуществом модели обратной динамики является то, что построение модели $\St \HM\times \St \HM\to \A$ сопоставимо по сложности с моделями, использующимися внутри основного RL-алгоритма: не требуется построение моделей, выдающих объекты из $\St$.

Для модели обратной динамики проблема <<шумных телевизоров>> в среде заменяется симметричной проблемой в пространстве действий $\A$. Из формализма MDP по формуле Байеса следует:
\begin{equation}\label{bayesdynamics}
p(a \mid s, s') = \frac{p(s' \mid s, a)\pi(a \mid s)}{\int\limits_\A p(s' \mid s, a)\pi(a \mid s) \diff a}
\end{equation}

Из формулы \eqref{bayesdynamics} понятно, что, во-первых, искомое распределение существенно зависит от используемой для порождения выборки стратегии $\pi$. Во-вторых, во многих средах пространство действий для некоторых областей $\St$ может содержать принципиально неразличимые действия. Например, если пространство действий непрерывно, то мы скорее всего опять будем использовать детерминированную модель, а в дискретных пространствах действий, когда мы решаем задачу классификации, использование ошибки $-\log q(a \HM\mid s, s')$, где $q$ --- наше приближение \eqref{bayesdynamics}, вовсе не будет выдавать нулевую ошибку даже если мы выучили \eqref{bayesdynamics} идеально.

\begin{example}
Например, если в данном состоянии $s$ два действия в принципе эквивалентны, то есть $p(s' \HM\mid s, a_1) \HM\equiv p(s' \HM\mid s, a_2)$, то агент будет мотивирован, находясь в $s$, совершать действия $a_1, a_2$, поскольку его модель обратной динамики не сможет их различать, и классификатор будет размазывать вероятности между ними. Это типичная ситуация в видеоиграх, когда часто несколько комбинаций кнопок (считающиеся разными действиями) эквивалентны.
\end{example}

\needspace{8\baselineskip}
\begin{wrapfigure}{l}{0.5\textwidth}
\centering
\includegraphics[width=0.5\textwidth]{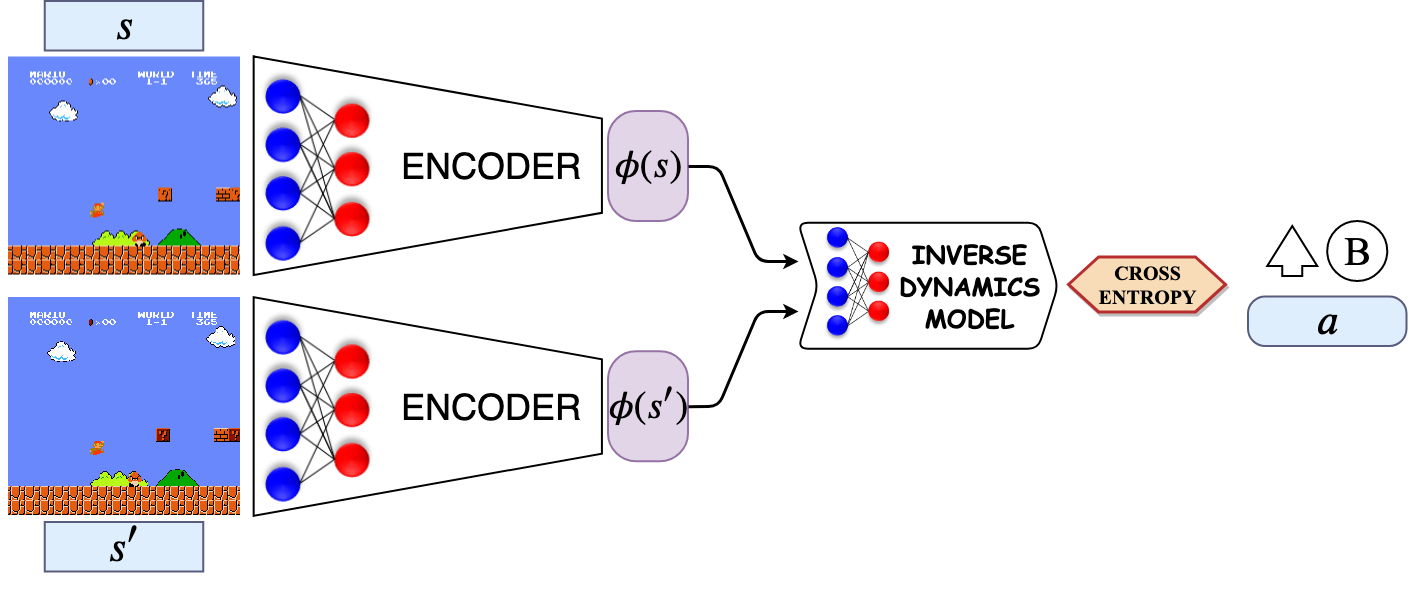}
\vspace{-0.4cm}
\end{wrapfigure}

Поэтому модель обратной динамики для любопытства обычно не используют. У неё, однако, есть другое, куда более интересное и полезное применение. Рассмотрим модель обратной динамики с <<сиамской>> архитектурой; для непрерывных пространств действий задача выглядит так:
\begin{equation}\label{siaminversemodel_continuous}
\E_{s, a, s'} \|g(\phi(s), \phi(s')) - a\|^2_2 \to \min_{g, \phi},
\end{equation}
а для классификации как
\begin{equation}\label{siaminversemodel_discrete}
\E_{s, a, s'} \log g(a \mid \phi(s), \phi(s')) \to \max_{g, \phi},
\end{equation}
где $\phi(s) \colon \St \to \R^d$ строит некоторое латентное описание состояний, а затем функция $g$ пытается восстановить действие $a$, случившиеся между двумя состояниями, по их латентным описаниям. Что можно сказать о том представлении состояний, которое выучит функция $\phi(s)$? Можно ожидать, что она будет оставлять от состояний только информацию, необходимую для предсказания промежуточных действий. Скорее всего, эта информация будет соответствовать описанию только тех объектов в среде, с которыми агент может непосредственно провзаимодействовать, по изменению состояний которых можно судить о том, какое действие совершил агент. Таким образом, $\phi(s)$ будет выдавать описание состояний, очищенное от нерелевантных для агента явлений.

\begin{definition}
Латентные представления состояний, которые учит функция $\phi(s)$ из задачи \eqref{siaminversemodel_continuous} или \eqref{siaminversemodel_discrete}, называются \emph{контролируемым состоянием} (controlable state), а сама функция $\phi(s)$ --- \emph{фильтром} (filter).
\end{definition}

\begin{example}
Представьте, что в следующем кадре видеоигры с вероятностью 0.1 моргает декоративное солнышко. Если его моргание никак не связано с агентом и действиями, которые агент выбирает, то это типичный шумный телевизор. Модель прямой динамики пыталась бы безуспешно предсказывать его моргание, из-за чего в предсказаниях будущих состояний всегда была бы некоторая ошибка. Модели же обратной динамики не нужно ничего знать о солнышке, чтобы предсказывать промежуточные действия, поэтому модель быстро научится игнорировать солнышко во входных данных; в описаниях $\phi(s)$ информации о солнышке не будет. Таким образом, фильтр очистит описание состояний от этого лишнего элемента.

А что, если солнышко всё-таки как-то связано с агентом? Тогда в зависимости от состояния солнца --- назовём его $c$, это часть информации внутри $s$ --- при каком-то действии $a$ функция переходов отличается от другого действия $\hat{a}$: $p(s' \HM \mid c, a) \neq p(s' \HM \mid c, \hat{a})$. Тогда классификатору или регрессору в модели обратной динамики будет необходимо оставить информацию $c$ внутри латентного представления $\phi(s)$, чтобы снизить ошибку при различении $a$ и $\hat{a}$.
\end{example}

\subsection{Внутренний модуль любопытства (ICM)}

Свойство модели обратной динамики наводит на мысль, как можно построить защиту от шумных телевизоров. Для этого мы возьмём описание состояний и <<почистим>> их от шумных телевизоров при помощи фильтра $\phi(s)$, который переведёт описание состояния в некоторое латентное пространство, хранящее лишь частичную информацию о входе. А дальше модель прямой динамики построим в таком <<отфильтрованном>> латентном представлении:
\begin{equation}\label{ICMforward}
\E_{s, a, s'} \|f(\phi(s), a) - \phi(s')\|^2_2 \to \min_{f}
\end{equation}

Авторы алгоритма ICM предлагают обе модели обучать совместно, то есть в частности оптимизировать \eqref{ICMforward} по параметрам фильтра $\phi$. Итого модель ICM выглядит следующим образом (рассмотрим для примера случай непрерывных действий):
$$\E_{s, a, s'} \left[ \|g(\phi(s), \phi(s')) - a\|^2_2 + \alpha \|f(\phi(s), a) - \phi(s')\|^2_2 \right] \to \min_{f, g, \phi}$$
где $s, a, s'$ --- произвольные тройки из любого буфера, $\alpha$ --- масштабирующий гиперпараметр. 

\needspace{10\baselineskip}
\begin{wrapfigure}{r}{0.5\textwidth}
\vspace{-0.4cm}
\centering
\includegraphics[width=0.5\textwidth]{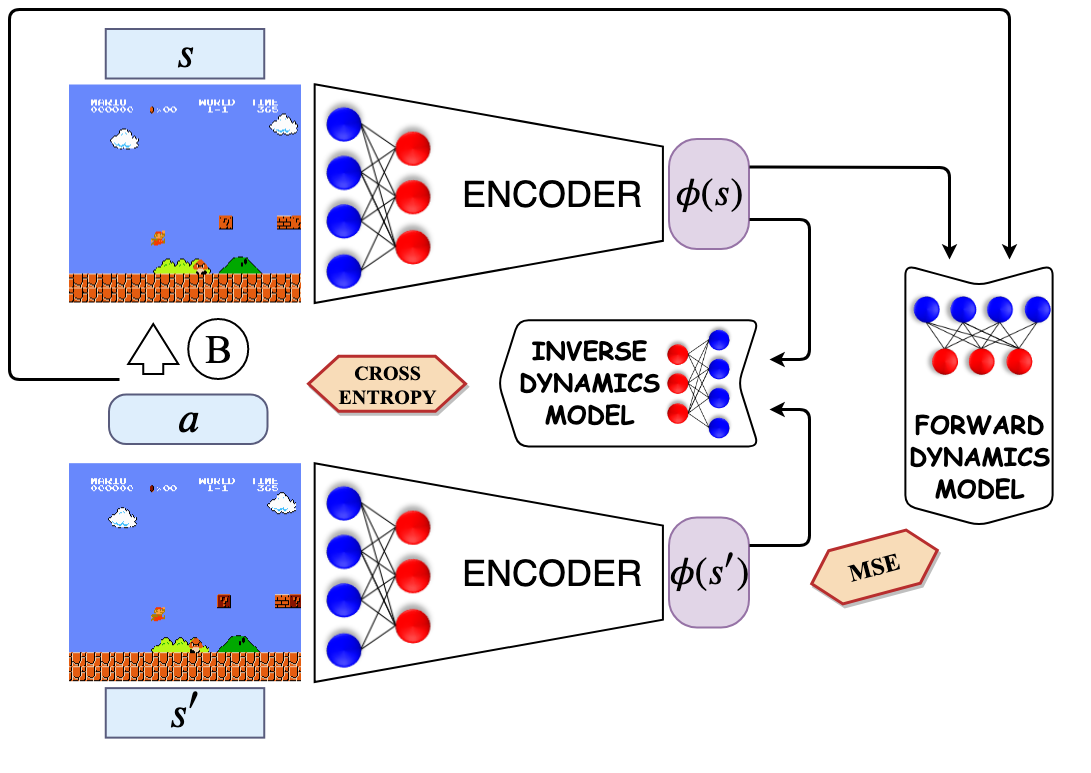}
\vspace{-0.5cm}
\end{wrapfigure}

В этой задаче оптимизации градиенты нигде не останавливаются: это значит, что от фильтра $\phi$ требуется построение таких представлений, в рамках которых модели прямой динамики $f$ <<проще всего>> предсказывать будущее. Понятно, что, если бы первого слагаемого (задачи обратной динамики) в таком функционале не было бы, оптимальным решением было бы $\phi(s) \HM= \const$. Можно считать, что модель прямой динамики выступает регуляризатором для модели обратной динамики: представления $\phi(s)$ одновременно должны содержать достаточно информации для предсказания выбранных действий и при этом быть максимально простыми.

В качестве внутренней мотивации используется только ошибка модели прямой динамики:
\begin{equation*}
r^{\intr}(s, a, s') \coloneqq \|f(\phi(s), a) - \phi(s')\|^2_2
\end{equation*}

Решает ли ICM проблему шумных телевизоров? Теоретически, можно рассчитывать на то, что модель обратной динамики отфильтрует те шумные телевизоры, с которыми агент не может провзаимодействовать. Но если у агента есть <<пульт>> от телевизора, если случайное непредсказуемое явление можно <<затриггерить>> определёнными действиями, то ICM всё равно приведёт к прокрастинации.

\begin{exampleBox}[label=ex:controltv]{Управляемый шумный телевизор}
В среде присутствует телевизор, демонстрирующий изображение (возможно, осмысленное!) из разнообразного бесконечного набора. У агента есть пульт от телевизора, то есть специальное действие $\hat{a}$, при выборе которого изображение на телевизоре сменяется на случайное из набора. По факту смены изображения между $s$ в $s'$ модель обратной динамики может сделать однозначный вывод о том, что между состояниями было выбрано именно действие $\hat{a}$, следовательно в представлении $\phi(s)$ останется информация о содержимом телевизора. При этом, в силу случайности выбора изображения из набора, предсказать контент телевизора по предыдущему изображению и факту нажатия на пульт невозможно. Следовательно, модель прямой динамики в ICM будет ошибаться на тройках, содержащих $\hat{a}$, и агент будет мотивирован бесконечно выбирать действие $\hat{a}$.
\end{exampleBox}


Шумные телевизоры, с которыми агент может провзаимодействовать, представляют собой любые стохастичные явления в среде, которые агент может <<запускать>> своими действиями. Это не столько проблема самого алгоритма ICM, сколько концептуальная проблема любопытства. Наличие у агента возможности взаимодействовать с шумным телевизором потенциально означает, что некоторая комбинация действий приводит к решению искомой задачи в среде (высокой внешней награде), а сам шумный телевизор --- связан с задачей (в примере \ref{ex:controltv} с управляемым шумным телевизором задача гипотетически могла заключаться в поиске определённого изображения из набора, или совершении определённой комбинации действий при определённых условиях на текущее отображаемое изображение). При этом любопытство, как и любая внутренняя мотивация, вводится из соображений, что априорных знаний об истинной задаче агента не дано, и произвольные явления в среде должны быть исследованы.

\begin{remark}
В большинстве традиционных задач для тестирования алгоритмов RL такие управляемые телевизоры обычно отсутствуют; обычно, для того, чтобы <<сломать>> ICM, нужно строить специальную среду. Но понятно, что чем сложнее и реалистичнее рассматриваются задачи, тем больше вероятность натолкнуться на такое явление.
\end{remark}





\section{Multi-task RL}\label{sec:multitask}

\subsection{Многозадачность}

Понятно, что в одной и той же среде мы можем решать много разных задач. Для одной и той же среды каждая задача --- в рамках нашего формализма MDP --- может быть задана при помощи своей функции награды.

\begin{definition}
Для данной среды $(\St, \A, \Trans)$ \emph{набором подзадач} будем называть множество $\G$, элементы $g \in \G$ которого имеют уникальное признаковое описание, для которых задана функция награды $r^g(s, a)$ (и, возможно, свой набор терминальных состояний $\St^+_g$).
\end{definition}

Пусть наша цель --- научиться в среде решать не одну задачу, а много. Сформулируем задачу так. Нам дан набор подзадач и некоторое распределение над задачами $p(g)$. Будем считать, что в начале эпизода нам сэмплируется случайная задача, и дальше в этом эпизоде мы должны решать её. Оптимизировать будем среднюю награду <<по задачам>>:
\begin{equation}\label{multitask}
\E_{g \sim p(g)} \E_{\Traj \sim \pi} \sum_{t \ge 0}\gamma^t r^g_t \to \max_{\pi}
\end{equation}

\begin{example}
Типичный пример $g$ --- координаты целевой точки, до которой нужно добраться, или в которую нужно что-то переместить.
\end{example}

\begin{example}
Например, если вы учите робота ходить, то вы можете в качестве $g$ брать направления движения. Тогда функция награды будет поощрять робота за, например, перемещение по вектору $g$, а сам этот вектор будет генерироваться случайно в начале эпизода. Решая задачу \eqref{multitask}, робот научится ходить во все стороны, и целевое направление можно будет подавать ему на вход!
\end{example}

\begin{theorem}
Задачу \eqref{multitask} можно свести к обычному MDP, добавив в состояния информацию о текущей решаемой задаче.
\begin{proof}
Действительно, пусть $\hat{\St} \coloneqq \St \times \G$. Пусть начальное состояние будет определяться стохастично как $(s_0, g)$, где $s_0$ --- начальное состояние нашей исходной среды (без ограничения общности считаем его фиксированным), $g \sim p(g)$. В функцию переходов $\hat{\Trans}$ просто добавим информацию о текущей решаемой задачи:
\begin{equation}\label{transition_indep_from_goal}
p(s' \mid s, g, a) \coloneqq p(s' \mid s, a), \quad g' \HM\coloneqq g
\end{equation} 
Функцию награды определим как $\hat{r}(s, g, a) \coloneqq r^g(s, a)$. Наконец, терминальными определим все пары $\{ (s, g) \mid g \in \St^+_g\}$. Тогда оптимизируемый функционал для MDP $(\hat{\St}, \A, \hat{\Trans}, \hat{r})$ в точности совпадает с \eqref{multitask}.
\end{proof}
\end{theorem}

Итак, придуманная постановка задачи просто переводит нас в другое MDP: формально <<ничего не изменилось>>. Но мы поняли важную вещь: решение многих задач в среде <<параллельно>> эквивалентно хранению в состояниях информации о текущей решаемой задачи. Коли так, и если у задач есть признаковое описание (это было важное предположение), то мы можем искать оценочные функции всего набора совместно: например, моделировать $Q^*(s, g, a)$, где описание $g$ решаемой задачи поступает модели на вход вместе с $s$. Это в точности эквивалентно тому, что $g$ хранилось бы как часть описание состояний; тогда оно тоже бы поступало на вход оценочным функциям.

\begin{definition}
Модель для аппроксимации оценочной функций сразу для набора подзадач называется \emph{универсальной оценочной функцией} (universal value function). Аналогично можно рассматривать <<\emph{универсальную стратегию}>> $\pi(a \mid s, g)$.
\end{definition}

\subsection{Мета-контроллеры}

Обсудим, как можно формализм мультизадачности применить в обычной задаче RL. Часто в алгоритме у нас встречаются важные гиперпараметры, которые трудно заранее подобрать. Выделим два примера: коэффициент дисконтирования $\gamma$ и коэффициент масштабирования внутренней мотивации $\alpha$ из уравнения \eqref{sumgoal}. Вместо того, чтобы подбирать эти параметры <<по сеточке>>, мы можем запустить multi-task RL, где $g$ --- пара $\gamma, \alpha$. Далее, вместо оценочных функций, например, $Q^*(s, a)$, будем учить универсальные оценочные функции $Q^*(s, a, \gamma, \alpha)$. Такая функция будет выдавать для данной пары $s, a$ будущую награду с учётом поданного на вход дисконтирования $\gamma$ и масштаба бонуса внутренней мотивации $\alpha$. Такое обобщение хорошо и само по себе, поскольку снабдит модель оценочной функции вспомогательными градиентами.

Мета-контроллеры могут помочь автоматически определить, для каких именно значений гиперпараметров $\gamma, \alpha$ алгоритму легче всего максимизировать среднюю внешнюю награду за эпизод. Для этого перед каждым очередным запуском эпизода обучения многорукий бандит (раздел \ref{sec:bandistssection}) выбирает $\gamma, \alpha$, которые будут использоваться в текущем эпизоде обучения для сбора данных. После окончания каждого эпизода бандит считает наградой, полученной <<из данного автомата>> суммарную (не дисконтированную) внешнюю награду, то есть истинное значение счёта игры.

Этот трюк может успешно применяться для автоматического подбора гиперпараметров прямо по ходу самого обучения. Можно смотреть на это так: будто в формуле \eqref{multitask} мы начинаем учить <<хорошее>> $p(g)$.

\subsection{Переразметка траекторий}\label{subsec:hindsight}

Рассмотрим обучение универсальной оптимальной Q-функции $Q^*(s', g, a')$ для набора подзадач $\G$. Для данной тройки $s, g, a$ при имеющемся сэмпле следующего состояния $s'$ мы можем обучаться на стандартный одношаговый таргет:
$$y(s, g, a) \coloneqq r^g(s, a) + \gamma \max_{a'} Q(s', g, a')$$

Сделаем важное наблюдение: допустим, нам известна функция награды $r$. Это довольно типичная ситуация в тех случаях, когда каким-то образом задан целый набор подзадач в среде, но при необходимости её также можно учить по собираемым сэмплам. Тогда заметим, что мы можем для имеющегося сэмпла $s' \sim p(s' \mid s, a)$ посчитать значение целевой переменной $y(s, \hat{g}, a)$ для любых $\hat{g} \in \G$. Действительно: в силу \eqref{transition_indep_from_goal} вне зависимости от того, какую цель преследовал агент, собравший переход $(s, g, a, r, s', \done)$, информацию о следующем состоянии $s'$ --- сэмпл из функции переходов --- можно использовать для обучения оценочной функции для любой задачи:
$$y(s, \hat{g}, a) \coloneqq r^{\hat{g}}(s, a) + \gamma \max_{a'} Q(s', \hat{g}, a')$$

\begin{example}
Допустим, вы стояли перед дверью ($s$), хотели пиццу ($g$), открыли дверь ($a$) и дверь открылась ($s'$). Тогда если бы вы, стоя перед дверью ($s$), хотели бы мороженное ($\hat{g}$) и открыли дверь ($a$), дверь всё равно бы открылась ($s'$ не изменился).
\end{example}

Это открывает путь к \emph{трансферу знаний} (transfer learning): опыт, собранный при решении одной задачи, можно использовать для обучения решению других задач. Понятно, что мы так можем <<переразметить>> не только переход, но целую траекторию.

\begin{definition}
Замена в собранной траектории цели $g$ на другую цель $\hat{g}$ и наград $r^g(s, a)$ на $r^{\hat{g}}(s, a)$ называется \emph{переразметкой траектории} (trajectory relabeling). 
\end{definition}

В простейшем случае, если размерность пространства задач $\G$ не очень большое, можно <<сохранить>> в буфере (или использовать для on-policy обучения в зависимости от использующегося алгоритма) переразмеченные траектории для всех $\hat{g}$. Однако, если $\G$ континуально, или поднабор задач богатый, то необходимо выбирать, для каких $\hat{g}$ проводить переразметку. Например, можно посэмплировать $\hat{g}$ случайно; можно ли придумать что-то умнее?

\subsection{Hindsight Experience Replay (HER)}

\needspace{5\baselineskip}
\begin{wrapfigure}{l}{0.3\textwidth}
\vspace{-0.7cm}
\centering
\includegraphics[width=0.3\textwidth]{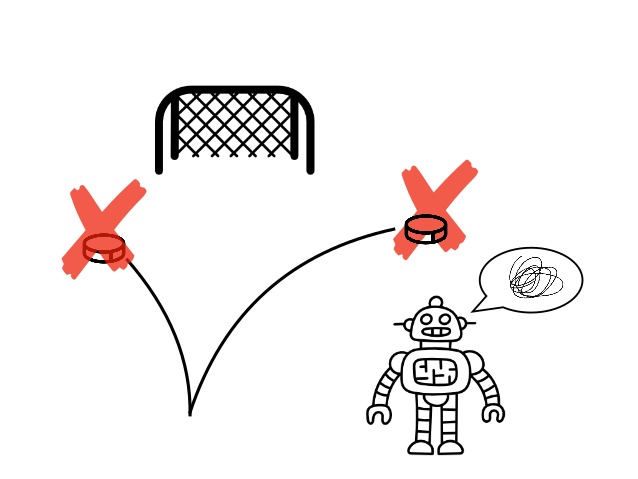}
\vspace{-0.7cm}
\end{wrapfigure}

Идею hindisght-переразметки сначала обсудим на примере частного случая задачи RL, задачи поиска \eqref{searchreward}. Мы уже обсуждали как можно справляться с этой задачей при помощи внутренней мотивации (раздел \ref{subsec:intrinsic_motivation}); сейчас мы сможем при помощи формализма мультизадачности придумать ещё один интересный способ.

Пусть мы предприняли попытку достичь цели и, через некоторое число шагов, остановились в состоянии $\bar{s}$, так и не справившись с задачей. Очередной отрицательный пример с константной наградой не позволяет нам начать ничему обучаться... хотя постойте-ка. Мы же достигли состояния $\bar{s}$! Давайте положим в него виртуальный тортик. За виртуальный тортик выдадим себе виртуальную +1. Теперь у нас есть положительный пример: если бы мы хотели достичь состояния $\bar{s}$, достичь виртуального тортика, то в ходе этой последней попытки мы всё делали правильно. Другими словами, мы говорим: я так и задумывал изначально, я с самого начала хотел добраться до тортика. В английском языке подобному <<мышлению задним числом>> соответствует выражение <<\emph{in hindsight}>>, и это слово часто используется для названия алгоритмов с этой идеей. 

\needspace{5\baselineskip}
\begin{wrapfigure}[6]{r}{0.35\textwidth}
\vspace{-0.5cm}
\begin{adjustwidth}{-0.4cm}{}
\includegraphics[width=0.4\textwidth]{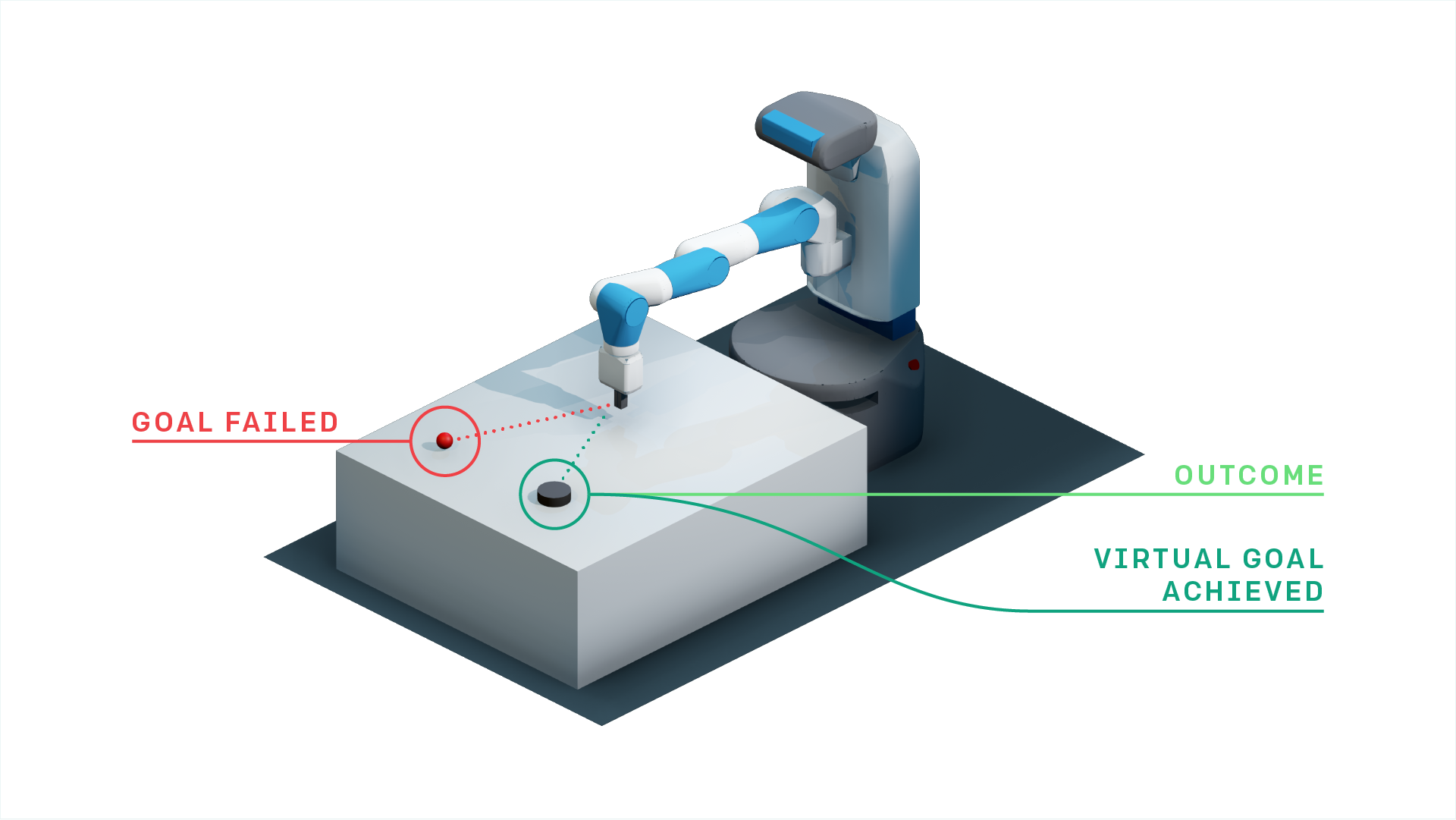}
\end{adjustwidth}
\end{wrapfigure}

\emph{Hindsight Experience Replay} может быть формализован следующим образом. Формально мы зададимся набором подзадач, и все подзадачи будут являться задачи поиска \eqref{searchreward}. Тогда каждая функция награды $r^g$ однозначно задаётся множеством терминальных состояний --- подмножеством $\St^+_g \subseteq \St$. 

\begin{definition}
Задачей \emph{навигации} (navigation) в среде с пространством состояний $\St$ будем называть набор подзадач $\G \equiv \St$, в котором для решения задачи $g \in \St$ необходимо достичь состояния из $\St^+_g$ --- близких по некоторой метрике состояний к $g$.
\end{definition}

\begin{example}
Самый простой и доступный всегда вариант --- взять вырожденную метрику, и таким образом рассматривать $\St^+_g \coloneqq \{g\}$. В детерминированных средах такая задача даже осмысленна, но дальнейший алгоритм сработает с такой метрикой и для стохастичных сред (поскольку in hindisight мы всегда берём реально достигнутые состояния). Но во многих задачах естественным образом возникают и другие метрики. Например, если вы перемещаете шайбу, и цель задаётся координатами, то можно в качестве метрики взять расстояние от шайбы до цели. 
\end{example}

Навигация --- это в некотором смысле <<полный>> набор подзадач: для любого состояния $s$ найдётся задача, для которой оно является терминальным (победным): $\exists \St^+_g \in \G \colon s \in \St^+_g$. Это важно для нас, чтобы мы для любой траектории могли найти задачу $g \HM\in \G$, <<решением>> которой эта траектория является.

Допустим, обучается какой-либо off-policy алгоритм, работающий с реплей буфером. Все модели в алгоритме, принимающие на вход $s$, теперь также будут принимать описание задачи $g$. Будем сэмплировать $g \sim p(g)$, как и раньше, в начале эпизода; или же, если этот вариант нам недоступен, пытаться решить исходную задачу с <<истинной>> целью, которую обозначим за $g^*$. 

\begin{remark}
Допустим, нам поставили весьма конкретную задачу научиться достигать $\St^+ \subseteq \St$. Возможно, мы не знаем даже описания целевых терминальных состояний (<<выйти из лабиринта>>, а где выход --- непонятно); тогда положим описание этой <<истинной>> задачи каким-нибудь специальным вектором $g^*$.
\end{remark}

\needspace{12\baselineskip}
\begin{wrapfigure}{r}{0.35\textwidth}
\vspace{-0.4cm}
\centering
\animategraphics[controls, width=\linewidth]{1}{Images/HER/HER-}{1}{5}
\vspace{-0.5cm}
\end{wrapfigure}

Допустим, на очередном шаге мы не смогли решить задачу $g^*$, и очередной эпизод закончился в состоянии $\bar{s}$. Собранные переходы с $r = 0$ и $\done = 0$ отправились в буфер. Но мы понимаем, что если бы мы хотели бы решить задачу $g \colon \bar{s} \in \St^+_g$, то данная траектория была победной. Тогда на последнем шаге, в конце эпизода, мы получили бы награду +1 и индикатор завершения эпизода $\done = 1$. Итак, мы можем переразметить траекторию новой целью $g$, и добавить переразмеченную траекторию в буфер.

В итоге, если раньше в буфер попадали только примеры с нулевым (константным) сигналом, и обучение было невозможно, теперь же у нас есть примеры с информативным сигналом.  

Здесь важно озаботиться богатством реплей буфера для решения вспомогательных задач. Нужно иметь примеры не только правильных последних шагов, но и всех остальных; важно также не перебить вспомогательными задачами буфер и оставить примеры с целью $g^*$, чтобы алгоритм в конечном счёте научился решать и её.

\begin{exampleBox}[righthand ratio=0.3, sidebyside, sidebyside align=center, lower separated=false]{}
Допустим, вы хотите научиться бить по шайбе так, чтобы она попадала в ворота. Вы бьёте по шайбе, но не попадаете; информативного сигнала нет. Тем не менее, люди каким-то образом способны учиться на подобных ошибках, не имея успешных примеров. Вероятно, мы в реальности мыслим в терминах <<левее-правее>>, которых в формализме MDP нет. С точки зрения HER, мы просто учимся попадать в ту же точку, в которую действительно попали, поскольку у нас есть пример того, как это делается; <<если бы ворота были в этой точке, я бы попал>>. И так, обучаясь попадать в различные точки, модель может обобщится на разные $g$, и однажды попадёт в том числе в реальные ворота $g^*$.

\tcblower
\animategraphics[controls, width=\linewidth]{1}{Images/HERres/HERres}{5}{9}
\end{exampleBox}

\subsection{Hindsight Relabeling}

Обобщим идею HER на произвольные multi-task задачи. Понятно, что примеры <<хорошего>> решения задачи нам намного ценнее примеров неудач: переразметкой мы боремся с тем, что в классическом машинном обучении назвали бы <<несбалансированностью выборки>>, увеличивая число примеров <<более редкого класса>> --- положительного опыта, когда агент набирает больше награды, чем уже умеет набирать.

Итак, дан произвольный набор подзадач $r(s, a, g)$. Мы преследовали какую-то цель (неважно какую) и породили траекторию $\Traj$. Обозначим $R^g(\Traj)$ суммарную награду за траекторию с точки зрения задачи $g$. Вопрос: для каких $g \in \G$ имеет смысл переразметить эту траекторию? 

\begin{exampleBox}[label=ex:multitask]{}
Допустим, мы собирали орешки и породили такую траекторию:

\begin{center}
    \includegraphics[width=0.7\textwidth]{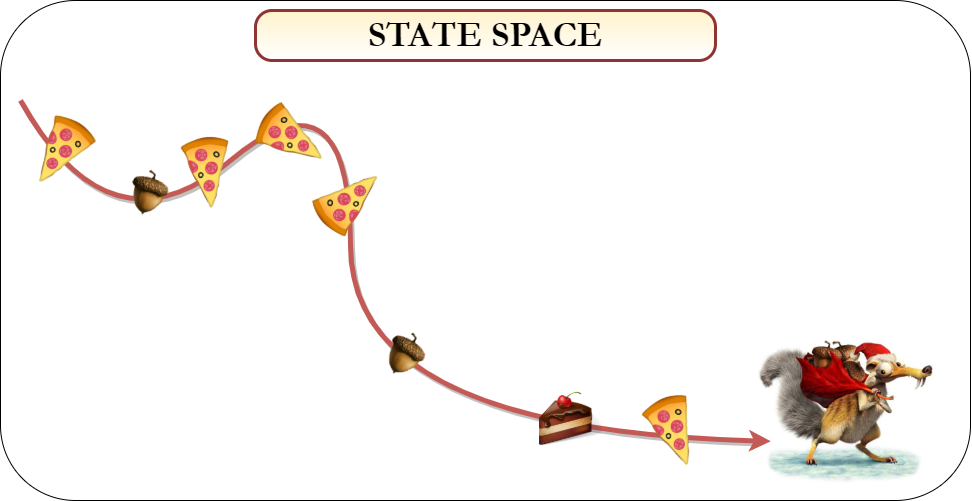}
\end{center}

Для какой задачи данная траектория является <<информативной>>, то есть примером хорошего решения? Очевидный ответ <<для пицц>> внезапно неверен: да, мы собрали пицц больше чем орешков, но вдруг собрать 5 пицц --- это очень мало, и такая траектория наоборот является плохой с точки зрения задачи сбора пицц?

А ещё данная траектория является успешным примером задачи <<избежать львов>>, поскольку ни одного льва в траектории не было. То есть важно не сколько награды $R^g(\Traj)$ мы собрали, а сравнение этого значения с тем, сколько может набрать хорошая стратегия решения задачи $g$.
\end{exampleBox}

Переформулируем вопрос: для какой задачи $g$ из нашего параметрического семейства $r(s, a, g)$ наша траектория является экспертной? Итак, ключевое наблюдение: поиск хороших $g$ для переразметки --- это задача обратного обучения с подкреплением, которую мы обсуждали в разделе \ref{subsec_irl}. В идеях Maximum Entropy Inverse RL и лежит ответ на наш вопрос.

С точки зрения Maximum Entropy подхода (формулы \eqref{meirl}), траектория $\Traj$ является экспертной для задачи $g$ с вероятностью
$$p(\Traj \mid g) \coloneqq \frac{e^{R^g(\Traj)} \prod_{t \ge 0}p(s_{t+1} \mid s_t, a_t)}{Z(g)}$$
Заметим, что нормировочная константа для каждого $g$ своя. Рассмотрим в качестве прайора $p(g)$, из которого нам, например, приходят задачи в начале эпизодов или возьмём какое-нибудь равномерное распределение. Тогда по формуле Байеса:
\begin{equation}\label{bayes_relabeling}
p(g \mid \Traj) \propto p(\Traj \mid g)p(g) = \frac{p(g) e^{R^g(\Traj)}}{Z(g)}
\end{equation}
Часть с вероятностями переходов сократились с нормировочной константой, поскольку они общие для всех $g$. Эта формула и говорит нам, что нужно использовать для поиска хорошей переразметки $g$ не просто задачи, для которых мы собрали большую суммарную награду $R^g(\Traj)$, а отнормированную на соответствующий интеграл $Z(g)$. Именно с вероятностями \eqref{bayes_relabeling} цель $g$ можно считать <<экспертной>>.

Формула \eqref{bayes_relabeling} даёт теоретический ответ на наш вопрос, но как использовать её на практике, то есть как сэмплировать из такого распределения? Здесь придётся ограничиться эвристиками. Если $\G$ конечно и мало, то мы можем подсчитать для каждого $g$ все суммарные награды $R^g(\Traj)$, но с нормировочной константой дела обстоят плохо: 
$$Z(g) = \int\limits_{\Traj} e^{R^g(\Traj)} \prod_{t \ge 0}p(s_{t+1} \mid s_t, a_t) \diff \Traj,$$
где траектории, вообще говоря, должны быть порождены оптимальной стратегией, решающей задачу $g$; мы можем попробовать считать, что текущая универсальная стратегия $\pi(a \mid s, g)$ и является экспертной для искомого $g$. Однако практические эвристики здесь сводятся к ещё большим упрощениям: например, к тому, чтобы взять какие-нибудь траектории из буфера $\Traj_1, \Traj_2 \dots \Traj_M$ и посчитать примерно среднюю награду, которую мы набираем как-нибудь так:
$$Z(g) \approx \frac{1}{M} \sum_{i = 1}^M e^{R^g(\Traj_i)}$$
Если $\G$ велико или, например, континуально, то просто для поиска хороших $g$ сэмплируется случайно несколько целей-кандидатов, и поиск хороших целей для переразметки проводится среди них.

\begin{example}
Продолжим пример \ref{ex:multitask} и посмотрим на несколько других встречавшихся траекторий из буфера.
\begin{center}
    \includegraphics[width=0.7\textwidth]{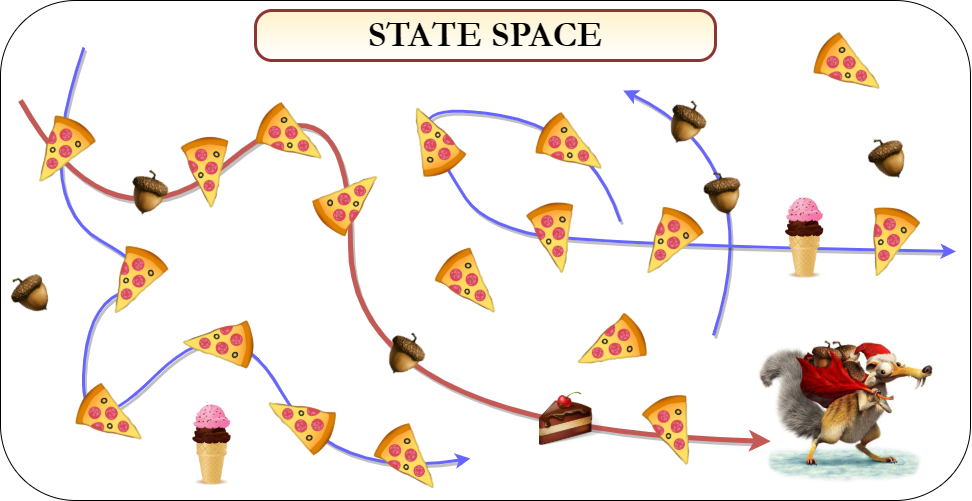}
\end{center}
Теперь мы видим, что собрать 5 пицц --- не такое редкое явление; наша собственная стратегия когда-то уже получала даже больше. А вот до тортиков мы в засэмплированных траекториях никогда не добирались, и поэтому нашу траекторию имеет смысл переразметить для задачи <<добраться до тортика>>. Таким образом у нас в буфере появится пример сбора целого 1 тортика, что, согласно примерам других траекторий, <<много>> для этой задачи.
\end{example}
\section{Иерархическое обучение с подкреплением}

\subsection{Опции}

Представьте, что ваша задача --- приготовить суп. Пытаться решить задачу методом проб и ошибок вам не придётся: у вас есть рецепт. Однако, беда: в рецепте первым пунктом написано <<возьмите чистую кастрюлю>>... И учиться брать чистую кастрюлю, видимо, придётся методом проб и ошибок.

\needspace{5\baselineskip}
\begin{wrapfigure}{r}{0.5\textwidth}
\vspace{-0.4cm}
\centering
\includegraphics[width=0.5\textwidth]{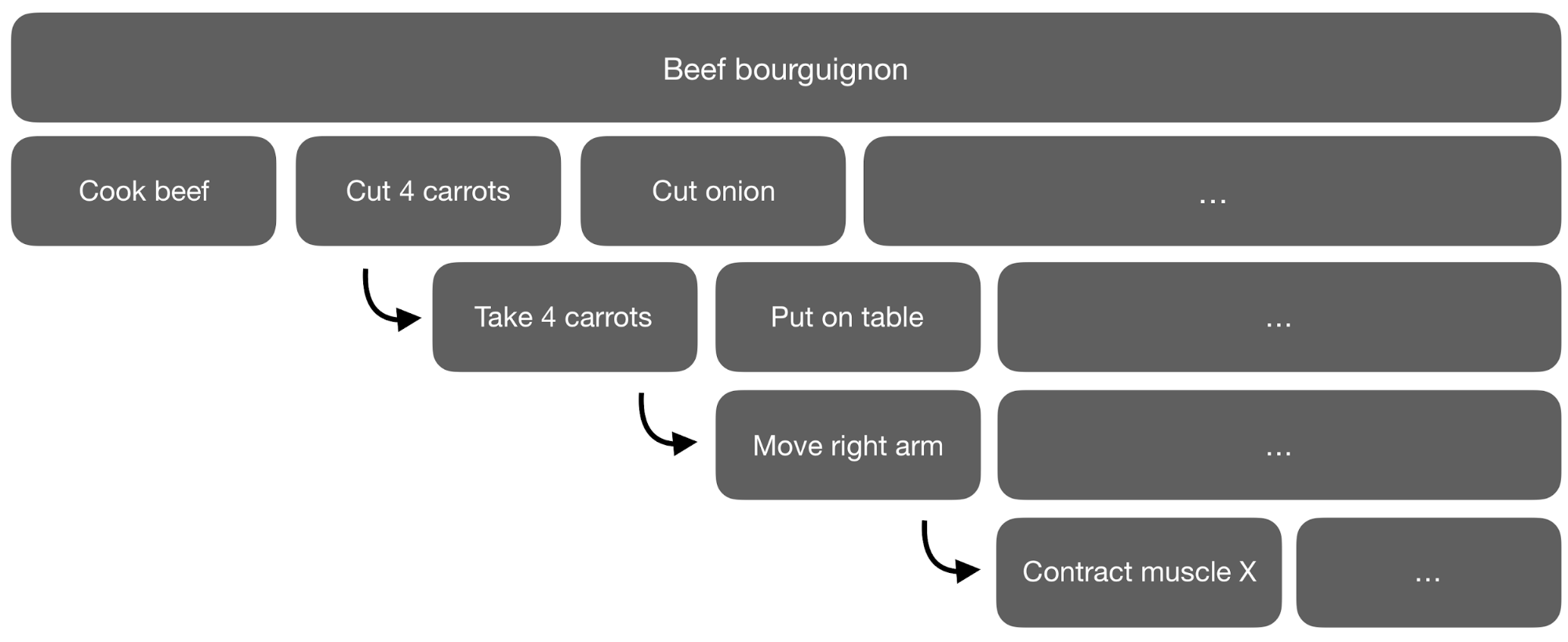}
\vspace{-0.5cm}
\end{wrapfigure}

Большинство задач в сложных средах делятся на подзадачи. <<Рецепта>>, позволяющего разложить сложную задачу на простые, однако, в общем случае нет, и агенту нужно не просто научиться решать набор подзадач, но и определять, какие именно подзадачи требуется выполнить для успеха.

Мы можем не вводить понятие подзадач или иерархичности и пытаться оптимизировать суммарную награду напрямую, как делали до этого. Но тогда наш рецепт для приготовления супа будет выглядеть примерно так: <<сожмите вот этот перечень мышц, теперь вот этот, так-так-так, ваша рука начала подниматься...>>. Введение иерархий позволит агенту смотреть на задачу на более абстрактном уровне --- на уровне выбора подзадач. Важно, что на этом уровне будет совершенно другой <<масштаб времени>>: число последовательных решений для решения всей задачи существенно сократится. Такое свойство <<высокоуровневых стратегий>> называется \emph{temporal abstraction}. Концептуально, именно на этом уровне агент должен заботиться о награде, описывающей основную задачу. Пока что, для начала, мы ограничимся чуть менее амбициозной задачей: давайте как-нибудь сделаем нашу стратегию <<многоуровневой>>, например, следующим образом.

\begin{definition}
Для данного MDP \emph{опцией} (option) $g$ называется пара\footnote[*]{в оригинале дополнительно рассматривалось множество $I_g \subseteq \St$ тех состояний, в которых опцию можно начать выполнять, однако в дальнейшем повествовании ситуация $I_g \ne \St$ нам не встретится.} $(\pi_g, \beta_g)$:
\begin{itemize}
    \item $\pi_g$ --- стратегия для исходного MDP.
    \item $\beta_g \colon \St \to [0, 1]$ --- \emph{политика терминальности} (termination policy).
\end{itemize}
\end{definition}

Пусть у нас есть множество опций $\G$, то есть даны или несколько разных стратегий $\pi_g$, или просто универсальная стратегия $\pi(a \mid s, g)$. Эти стратегии, работающие <<с исходным>> MDP на уровне \emph{примитивных действий} (primitive actions) (элементов $\A$), будем далее также называть \emph{рабочими} (workers). Также мы заводим <<высокоуровневую>> стратегию, которую далее будем называть \emph{менеджером} (manager) или \emph{мастер-стратегией} (master policy) $\manager{\pi}(g \mid s)$. Всё, что относится к менеджеру, будем помечать звёздочкой $\star$ над буквами. Высокоуровневые действия также ещё называют \emph{макро-действиями}, а примитивные действия --- \emph{микро-действиями}.

На очередном шаге менеджер, исходя из текущего состояния $s_t$, выбирает опцию, которая будет далее работать в среде: $g_t \sim \manager{\pi}(g_t \mid s_t)$. Выбранный рабочий генерирует примитивное действие: $a_t \sim \pi(a_t \mid s_t, g_t)$. Среда переходит в новое состояние $s_{t+1} \sim p(s_{t+1} \mid s_t, a_t)$. И в этот момент вызывается политика терминальности опции $g$: с вероятностью $\beta_{t+1} \sim \mathrm{Bernoulli}(\beta_g(s_{t+1}))$ рабочий завершает свою работу и снова передаёт решение менеджеру (тот снова выбирает следующего рабочего, и так далее). Если же политика терминальности не срабатывает, менеджер не вызывается, и взаимодействовать со средой продолжает рабочий $\pi_{g_t}$.

Для удобства будем считать, что в рамках такого фреймворка в траекториях хранятся не только примитивные действия, но и решения менеджера вместе с решениями политики терминальности:
$$\Traj \coloneqq (g_0, a_0, r_0, s_1, \done_1, \beta_1, g_1, a_1, r_1, s_2, \done_2, \beta_2 \dots )$$
причём $g' = g$ с вероятностью 1, если $\beta' = 1$, и $\beta_0 = 0$. Таким образом, вероятностная модель порождения траекторий задана так:
$$\begin{cases}
g_t \sim \manager{\pi}(g_t \mid s_t) \quad & \text{если $\beta_t = 0$} \\
g_t \coloneqq g_{t-1} \quad & \text{если $\beta_t = 1$} \\
\end{cases}$$
$$a_t \sim \pi(a_t \mid s_t, g_t)$$
$$s_{t+1} \sim p(s_{t+1} \mid s_t, a_t)$$
$$\beta_{t} \sim \mathrm{Bernoulli}(\beta_g(s_{t})) \quad (t > 0)$$

Мы поставим себе задачу обучать эту конструкцию end-to-end, то есть оптимизировать параметры стратегии менеджера, стратегий рабочих (опций) и политик терминальности опций напрямую с единственной целью максимизировать среднюю награду.


\subsection{Semi-MDP}

Как нам это всё чудо обучать? Начнём с менеджера. Для простоты будем считать, что набор опций нам дан, то есть даны распределения $\pi_g(a \mid s)$ и $\beta_g(s)$; хотим научиться обучать $\manager{\pi}(g \mid s)$. Попробуем понять, не живёт ли он в каком-то MDP. Поскольку мы решили, что на вход он получает текущее состояние $\St$, то его пространство состояний такое же, как и в исходном MDP. Его пространством действий являются макро-действия $\G$. На каждом шаге менеджер выбирает $g \in \G$, после чего среда для менеджера работает так: стратегия-рабочий для выбранного $g$ бегает в настоящей среде и решает поставленную подзадачу, переводя среду в новое состояние $\manager{s}'$. То, в каком состоянии окажется среда по итогу, полностью задано вероятностной моделью --- для менеджера это можно рассматривать как функцию переходов $\manager{p}(\manager{s}' \mid s, g)$. Наконец, награда для менеджера за этот шаг есть сумма собранной за время работы рабочего награды основной цели:
\begin{equation}\label{managerreward}
\manager{r}(s, a) \coloneqq \sum_{t=0}^{\tau} \gamma^t r_t
\end{equation}
где $\tau$ --- число шагов, затраченных рабочим на решение задачи. И вот тут возникает нюанс.

Награду, который менеджер получит за второй шаг, необходимо дисконтировать на $\gamma^{\tau}$, поскольку в настоящем MDP прошло $\tau$ шагов. Более того, это $\tau$ --- случайная величина; рабочий мог потратить на решение как 10 шагов, так и 100. А значит, менеджер живём не совсем в MDP; для него действия имеют различную продолжительность по времени. Если раньше в MDP все действия, можно считать, гарантированно <<выполнялись>> один шаг, то теперь, для менеджера, это не так.

\begin{definition}
MDP называется \emph{полумарковским} (Semi-Markov decision process, sMDP), если его функция перехода $p(s', \tau \mid s, a)$ помимо следующего состояния возвращает время $\tau > 0$, <<затраченное>> на выполнение данного шага. 
\end{definition}

В общем случае в sMDP время шагов может быть и вещественным числом (в частности, так можно учитывать, что среда взаимодействует с агентом в режиме <<реального времени>>), но в нашем случае $\tau > 0$ --- всегда натуральное число. Итак, при работе в sMDP в траекториях дополнительно необходимо хранить время каждого шага $\tau_t$; награда за $t$-ый шаг дисконтируется не на $\gamma^t$, а на $\gamma^{\sum_{\hat{t} \ge 0}^t \tau_{\hat{t}}}$.

Ранее рассматриваемая теория достаточно естественно обобщается на данный кейс. Например, уравнения Беллмана должны учитывать время в дисконтировании; например, для Q-функции:
$$Q^\pi(s, a) = r(s, a) + \E_{s', \tau} \gamma^\tau \E_{a'} Q^\pi(s', a')$$

Адаптация Q-learning для sMDP выглядит соответствующе: для перехода $(s, a, r, \tau, s', \done)$ обновление выглядит так:
$$Q(s, a) \leftarrow (1 - \alpha) Q(s, a) + \alpha \left(r + \gamma^\tau \max_{a'} Q(s', a') \right)$$
То есть по прошедшему времени $\tau$ мы тоже стохастически аппроксимируем: используем сэмпл из буфера вместе с сэмплом $s'$.

\subsection{Оценочная функция по прибытию (U-функция)}

Итак, менеджер живёт в полумарковском процессе принятия решений sMDP $(\St, \G, \manager{\Trans}, \manager{r})$, где $\manager{r}$ определено \eqref{managerreward}, а функция переходов\footnote{мы, в принципе, можем полностью расписать это распределение $\manager{p}(\manager{s}', \tau \mid s, g)$, но это получится громоздко: нужно выинтегрировать $a_0, s_1, a_1, \dots s_{\tau - 1}, a_{\tau - 1}$, учесть вероятность попадания в $p(\manager{s}' \mid s_{\tau - 1}, a_{\tau - 1})$, срабатывание политики терминальности $\beta_g(\manager{s}')$ и несрабатывание политики терминальности для $s_1, s_2 \dots s_{\tau - 1}$ внутри интеграла.} $\manager{\Trans}$ задаётся процессом взаимодействия выбранного рабочего со средой с завершением по триггеру соответствующей политики терминальности. Можно обучать оценочные функции менеджера аналогом Q-learning для такого sMDP, но тогда для одного обновления понадобится получать один переход --- ждать, пока сработает политика терминальности выбранного рабочего. Оказывается, это необязательно. 

Вместо этого удобнее работать с теорией на уровне одношаговых рекурсивных соотношений между величинами, где один шаг --- это генерация одной случайной величины. Раньше в обычных MDP мы как делали: сгенерировали действие, и вся будущая награда --- это Q-функция. Среда ответила нам сэмплом $s'$ (ещё наградой за шаг и флагом $\done$, но считаем, что это идёт <<в одном комплекте>>) --- и дальнейшая награда есть V-функция. Это было очень удобно, поскольку все эти оценочные функции легко выражались между собой. Нам надо поступить также для траектории, состоящей из случайных величин $g, a, s', \beta'$: выбор менеджера, выбор стратегии, отклик среды, выбор политики терминальности.

По определению, $\manager{Q}(s, g)$ обозначает следующее: менеджер сидел в некотором в состоянии $s$ и выбрал опцию $g$, или, что не существенно, опция $g$ уже была выбрана на предыдущем шаге, а политика терминальности не затриггерилась (в любом случае, в состоянии $s$ активировалась опция $g$), и функция возвращает среднюю будущую награду. Мы опустим здесь верхний индекс $\pi$, подразумевая, что мы считаем оценочную функцию для всего комплекта подконтрольных нам распределений: стратегии менеджера, рабочего и политик терминальности. Попробуем получить одношаговое рекурсивное соотношение для $\manager{Q}(s, g)$, связав её с оценочной функцией для следующей случайной величины --- выбором действия $a$ рабочим. Каким рабочим? Ну, раз активирована опция $g$, то однозначно рабочим $\pi(a \mid s, g)$. 

Мы уже поняли, что на стратегию рабочих можно смотреть как на универсальную стратегию в MDP с пространством состояний $\St \times \G$; 
мы можем определить $V(s, g)$ и $Q(s, g, a)$ --- универсальные оценочные функции рабочих --- как оценочные функции в таком MDP, как будущие награды после реализации поступающих на вход случайных величин. Тогда в силу структуры вероятностной модели:
\begin{proposition}
\begin{equation}\label{managerQworkerVworkerQ}
\manager{Q}(s, g) = V(s, g) = \E_{\pi_g(a \mid s)} Q(s, a, g)
\end{equation}
\end{proposition}

Попробуем пойти дальше и построить одношаговое соотношение для $Q(s, a, g)$. Что случится после выбора действия $a$ рабочим $\pi_g$? Среда выдаст награду за шаг, перейдёт в следующее состояние $s'$, и вот тут дальше случится загвоздка: будет генерироваться бернулиевская $\beta_g$. Если политика терминальности затриггерилась, то первым следующем шагом генерации траектории будет сэмплирование действия менеджером, и мы по определению получим столько, сколько выдаёт V-функция менеджера $\manager{V}(s')$. Здесь важно, что $\manager{V}(s)$ есть хвост награды по траектории после попадания в $s$ при условии (!) срабатывания триггера $\beta \HM= 1$, она предполагает, что первым шагом менеджер сможет выбрать новую опцию. Но если $\beta \HM= 0$, то действие выбирает текущий рабочий, и тогда будущая награда равна $V(s, g)$ (которое, как мы уже разобрались в \eqref{managerQworkerVworkerQ}, совпадает с $\manager{Q}(s, g)$). Работать с этим неудобно, поскольку мы не знаем, сработал ли триггер или нет, и поэтому для удобства вводится вспомогательная U-функция --- это хвост награды по траектории после попадания в состояние $s$ без каких-либо дополнительных ограничений.

\begin{definition}
Для sMDP, заданного MDP с набором опций $\{\pi_g, \beta_g\}$, U-функцией или \emph{оценочной функцией по прибытию} (state-value function upon arrival) называется
\begin{equation}\label{Ufunction}
U(s', g) \coloneqq (1 - \beta_g(s'))\manager{Q}(s', g) + \beta_g(s') \manager{V}(s')
\end{equation}
\end{definition}

Что это за формула: на вход U-функция получает состояние, в которое <<входит>> агент, и текущую опцию. Дальше расписано мат.ожидание по срабатыванию триггера. Если политика терминальности не срабатывает, следующей опцией снова будет $g$, поэтому дальнейшяя награда эквивалентна $\manager{Q}(s', g)$. Иначе выбор переходит к менеджеру и мы получим $\manager{V}(s')$. Таким образом, по определению:

\begin{proposition}
\begin{equation}\label{QU}
Q(s, g, a) = r(s, a) + \gamma \E_{s'} U(s', g)
\end{equation}
\end{proposition}



\subsection{Intra-option обучение}

Термин <<\emph{intra-option}>> означает, что мы можем за счёт таких соотношений обучать функции менеджера, используя информацию после выполнения каждого микро-действия (после получения информации о $s'$) вне зависимости от того, запускалась ли на данном шаге в принципе стратегия-менеджер или нет. Интуитивно основное соображение выглядит так: если стратегия-рабочий сделала один шаг в среде, а политика терминальности не сработала, то это всё равно что в этот момент менеджер снова выбрал того же самого рабочего.

Допустим, мы хотим обучить $\manager{Q}^*(s, g)$ --- оптимальную Q-функцию менеджера --- в предположении, что политика терминальности и стратегии рабочего зафиксированы и не меняются. Попробуем составить уравнение оптимальности для неё в аналогии с обычными уравнениями.

\begin{definition}
Для sMDP, заданного MDP с фиксированным набором опций $\{\pi_g, \beta_g\}$, \emph{оптимальной U-функцией} называется
\begin{equation}\label{U*function}
U^*(s', g) \coloneqq (1 - \beta_g(s'))\manager{Q}^*(s', g) + \beta_g(s') \max\limits_{g'} \manager{Q}^*(s', g')
\end{equation}
\end{definition}

В этом определении мы по сути просто взяли формулу обычной U-функции \eqref{Ufunction} и заменили в ней $\manager{V}(s')$ на $\max\limits_{g'} \manager{Q}^*(s', g')$ в силу предположения оптимальности, что менеджер всегда выбирает наилучшую опцию $g'$. Давайте теперь попробуем выразить $\manager{Q}^*(s, g)$ через неё же саму, используя $U^*(s', g)$.

\begin{proposition}\label{pr:managerQ*U*}
Q-функция менеджера удовлетворяет следующему рекурсивному уравнению:
\begin{equation}\label{managerQ*U*}
\manager{Q}^*(s, g) = \E_{\pi_g(a \mid s)} \left[ r(s, a) + \gamma \E_{s'} U^*(s', g)\right]
\end{equation}
\begin{proof}
Допустим, менеджер выбрал опцию $g$, и рабочий сделал один шаг в среде (сгенерировалось $a \sim \pi_g(a \mid s)$ и $s'$ в исходном MDP). Дисконтирование случилось только на $\gamma$. После этого с вероятностью $\beta_g(s')$ менеджер сможет выбрать новую подзадачу (это бы соответствовало обычному уравнению оптимальности Беллмана, поскольку прошёл всего один шаг), а с вероятностью $1 - \beta_g(s)$ менеджеру не предоставляется выбора. В такой ситуации можно считать, что менеджер просто <<обязан>> снова выбрать $g$: вероятностная модель со следующего шага выглядит в точности также.
\end{proof}
\end{proposition}

Итак, мы можем для обучения менеджера пользоваться Q-learning-ом для sMDP: просить рабочего $g$ решить подзадачу, дождаться $\manager{s}', \tau$ и делать один шаг обновления. Но <<intra-option>> рекурсивное уравнение \eqref{managerQ*U*} показывает интересную альтернативу: для обучения функции ценности менеджера для перехода $\T \coloneqq (s, g, a, r, s', \done)$ можно рассчитать целевую переменную как
\begin{equation}\label{optimalmanagertarget}
y(\T) \coloneqq r(s, a) + \gamma (1 - \done) U^*(s', g),
\end{equation}
где U-функция вычисляется полностью по формуле \eqref{U*function} (мат.ожидание по Бернуливской $\beta$ можем взять явно), и далее, как обычно, минимизировать MSE:
$$\E_\T \left(y(\T) - \manager{Q}^*(s, g)\right)^2 \to \min_{\manager{Q}^*}$$
Заметим, что переходы мы можем брать любые, в том смысле, что не обязательно, чтобы $g$ было выбрано менеджером именно на данном шаге. Для корректного обучения мат.ожиданий нам требуется лишь, чтобы $a \HM\sim \pi_g(a \HM\mid s)$, $r \HM= r(s, a)$ и $s' \HM\sim p(s' \HM\mid s, a)$. Обратим внимание на первое условие: сейчас мы считаем опции зафиксированным (частью стационарной среды). Если стратегии рабочих меняются (а они будут меняться, так как будут обучаться), условие стационарности нарушается, и off-policy режим обучения мы, естественно, теряем. 

\subsection{Обучение стратегий рабочих}

\begin{proposition}
В предположении оптимальности менеджера оптимальная Q-функция рабочего удовлетворяет следующему уравнению:
$$Q^*(s, a, g) = \left[ r(s, a) + \gamma \E_{s'} U^*(s', g)\right]$$
\begin{proof}
Доказательство в точности повторяет теорему \ref{pr:managerQ*U*} за тем исключением, что действие $a$ уже подано оценочной функции на вход.
\end{proof}
\end{proposition}

Мораль отсюда простая: чтобы учить оценочные функции рабочего, мы можем использовать те же таргеты $y(\T)$ \eqref{optimalmanagertarget}. Только теперь для рабочих оценочная функция дополнительно обусловлена на действие $a$.
$$\E_\T \left(y(\T) - Q^*(s, g, a)\right)^2 \to \min_{Q^*}$$

Важный момент --- здесь можно применять hindsight-приём, который мы встречали в разделе \ref{subsec:hindsight}. Действительно, поскольку единственное, что теперь требуется от переходов --- $s' \HM\sim p(s' \HM\mid s, a)$, мы можем проделать переразметку имеющегося в опыте перехода $\T \coloneqq (s, g, a, r, s', \done)$ на $\hat{\T} \coloneqq (s, \hat{g}, a, r, s', \done)$ для любого $\hat{g}$. То есть: что было бы, если бы в состоянии $s$ мы пользовались бы опцией $\hat{g}$ и выбрали бы действие $a$? Тогда мы попали бы в состояние $s'$ и получили бы награду $r$. Другой информации для получения прецедента для обучения $Q^*(s, \hat{g}, a)$ нам и не нужно. Это значит, что для рабочих мы можем обучать оценочные функции сразу для всех $g \in \G$.

Мы обсудили обучение Q-функций менеджера и рабочего с одношаговых таргетов; конечно же, имея на руках рекурсивные соотношения, мы можем построить полные аналоги всей стандартной теории. Например, для применения Policy Gradient подхода, нам нужны не сами оценочные функции, а их несмещённые оценки; в таких ситуациях достаточно иметь лишь достаточно обучать лишь Q-функцию менеджера. Действительно: если мы знаем $\manager{Q}(s, g)$, то тогда в силу \eqref{QU}:
$$Q(s, g, a) \approx r(s, a) + \gamma U(s', g), \quad s' \sim p(s' \mid s, a),$$
где $U(s', g)$ выражается через Q-функцию менеджера в силу формулы \eqref{Ufunction} --- несмещённая оценка. В качестве бэйзлайна возможно использовать $V(s, g)$, которая совпадает с $\manager{Q}(s, g)$.

\subsection{Обучение функций терминальности}

Перейдём к оставшемся открытому вопросу: как обучать политику терминальности? Рассмотрим такую ситуацию: мы сидим в состоянии $s$ и выполняем опцию $g$. Нужно ли останавливать рабочего, полагая $\beta(s, g) \HM= 1$, или можно продолжать действовать с его помощью? С точки зрения оптимальных оценочных функций, в первом случае мы получим $\manager{V}^*(s)$, а во втором $V^*(s, g)$. Но очевидно, что
$$\manager{V}^*(s) = \max_{\hat{g}} \manager{Q}^*(s, \hat{g}) \ge \manager{Q}^*(s, g) = V^*(s, g),$$
то есть оптимально прерывать рабочего на каждом шаге.

Это, конечно, затыка в нашей теории, поскольку если рабочий прерывается на каждом шаге, никакого temporal abstraction у нас не получится. Попробуем обратится к Policy gradient подходу; мы сможем обучать функции терминальности, пользуясь формулой градиентов по их параметрам, когда сами политики терминальности будут стохастичны. 

Пусть $\beta_{\theta}(s, g) \in [0, 1]$ выдаёт вероятность бернулливской величины и дифференцируемо по параметрам $\theta$. Пусть $\manager{A}(s, g) \coloneqq \manager{Q}(s, g) \HM- \manager{V}(s)$ --- Advantage-функция менеджера. Функционал, который мы оптимизируем, для начального состояния $s_0$, равен по определению $\manager{V}(s_0)$.

\begin{theorem}[Termination Gradient Theorem]
\begin{equation}\label{terminationpg}
\nabla_{\theta} \manager{V}(s) = -\E_{\Traj \mid s_0 = s} \sum_{t \ge 0} \gamma^t \nabla_\theta \beta_{\theta}(s_t, g_t) \manager{A}(s_t, g_t)
\end{equation}
\begin{proof}
Используя уравнения связи \eqref{managerQworkerVworkerQ} и \eqref{QU}:
$$\nabla_{\theta} \manager{V}(s) = \nabla_{\theta} \E_{g} \E_{a} \left[ r(s, a) + \gamma \E_{s'} U(s', g)\right]$$
Мат.ожидания здесь берутся по стратегиям менеджера и рабочего, не зависящих от $\theta$, поэтому мы сразу проносим градиент вплоть до оценочной функции по прибытию, а дальше применяем стандартную для Policy Gradient технику.
\begin{align*}
\nabla_{\theta} U(s', g) &= \nabla_{\theta} \left[\beta_{\theta}(s', g) \manager{V}(s') + (1 - \beta_{\theta}(s', g)) \manager{Q}(s', g) \right] = \\
 &= \nabla_{\theta} \beta_{\theta}(s', g) \left[\manager{V}(s') - \manager{Q}(s', g) \right] + \beta_{\theta}(s', g) \nabla_{\theta} \manager{V}(s') + (1 - \beta_{\theta}(s', g)) \nabla_{\theta} \manager{Q}(s', g)
\end{align*}
Здесь мы хотим вернуться к мат.ожиданиям по траекториям, и в том числе к мат.ожиданиям по $\beta$. Для этого достаточно заметить, что в полученном выражении ровно такое мат.ожидание и стоит. Действительно, убедимся, что
$$\beta_{\theta}(s', g) \nabla_{\theta} \manager{V}(s') + (1 - \beta_{\theta}(s', g)) \nabla_{\theta} \manager{Q}(s', g) = \E_{\beta_{\theta}} \E_{g'} \nabla_{\theta} \manager{Q}(s', g')$$
Если в выражении справа выпало $\beta_{\theta} = 0$, то дальше гарантировано $g' = g$, мат.ожидание по $g'$ вырождается и мы получаем второе слагаемое из выражения слева. Если же в выражении справа выпало $\beta_{\theta} = 1$, то дальше $\E_{g'} \nabla_{\theta} \manager{Q}(s', g') = \nabla_{\theta} \manager{V}(s')$, и мы получаем первое слагаемое из выражения слева.

Собирая всё вместе, получаем рекурсивную формулу:
$$\nabla_{\theta} \manager{V}(s) = \E_{g} \E_{a} \gamma \E_{s'}  \left[ - \nabla_{\theta}\beta_{\theta}(s', g) \manager{A}(s', g) + \E_{\beta} E_{g'} \nabla_{\theta} \manager{Q}(s', g') \right] $$
Собирая рекурсивную формулу в мат.ожидание по всей траектории, получаем доказываемое.
\end{proof}
\end{theorem}

Крайне интуитивная формула \eqref{terminationpg} говорит ровно то, с чего мы начали обсуждение оптимального поведения политики терминальности: если $\manager{A}(s, g) \HM> 0$, то текущая опция лучше той, которую в среднем выбрал бы текущий менеджер, и поэтому стоит продолжать её выполнять; $\beta(s, g)$ уменьшается. Но если $\manager{A}(s, g) < 0$, то менеджер сейчас может выбрать более хорошую опцию, более хорошего рабочего, и поэтому нужно передавать ему управление: $\beta(s, g)$ надо уменьшать.

Как мы знаем, для оптимальных стратегий $\max\limits_{g} \manager{A}(s, g) = 0$, и поэтому такая градиентная оптимизация чревата вырождающими ситуациями. Типично, что менеджер начинает получать управление на каждом шаге. Бороться с этим приходится костылями: например, дополнительными регуляризациями на политику терминальности, чтобы она как можно реже передавала управление менеджеру, и менеджеры с рабочим учились в этих условиях с <<неоптимальной>> политикой терминальности. Популярное решение --- добавить в формуле \eqref{terminationpg} к оценке Advantage небольшое положительное число.

\subsection{Феодальный RL}

В рамках подхода с опциями мы не факт, что выучим какие-то разумные подзадачи; скорее, мы надеемся, что, возможно, рабочие как-то <<распределят>> между собой области среды, за которые будут ответственны. Мы понимаем, что, деля сложную задачу на последовательность более мелких, мы можем переставать думать об исходной задаче на уровне выполнения подзадач. 

\begin{example}
То есть: если было принято решение достать чистую кастрюлю, то истинную цель этой установки (награду за приготовление супа) можно не принимать в расчёты и оптимизировать только награду за доставание кастрюли. Такое мышление открывает возможности для переиспользования оптимальных стратегий доставания кастрюли для других задач (например, приготовления пельменей) --- потенциальный путь к transfer learning.
\end{example}

Соображение лежит в основе принципа \emph{феодализма} (feudalism), который можно сформулировать в виде следующих постулатов:
\begin{itemize}
    \item стратегия, действующая на некотором уровне абстрактности, не должна задумываться о более высокоуровневых задачах (<<единственная задача вассала --- выполнение задач, поставленных феодалом>>).
    \item высокоуровневая стратегия не рассматривает примитивные действия $\A$ (<<феодала не заботит, как вассал будет достигать поставленных ему целей>>).
    \item при наличии нескольких уровней иерархии действует принцип <<вассал моего вассала не мой вассал>>. В дальнейшем повествовании мы ограничимся двумя уровнями иерархии, но теорию легко можно обобщить на большее число уровней.
\end{itemize}

Итак, общая схема иерархического RL с двумя уровнями, к которой мы хотим прийти, выглядит так. Откуда-то берётся набор подзадач; необходимо придумать, откуда его брать. Заводится стратегия-менеджер и стратегии-рабочие для каждой подзадачи. Когда менеджер выбирает подзадачу $g$, запускается стратегия соответствующего рабочего $\pi_g$. Она делает несколько шагов в исходной среде, пока не решит задачу или не <<сдастся>> (кастрюли в шкафу не оказалось и нужно бежать в магазин за новой); критерий остановки процесса выполнения стратегии рабочего остаётся открытым. Далее менеджер снова принимает решение о следующей решаемой подзадаче, оптимизируя исходную награду; рабочие стремятся решить свои подзадачи, для чего им нужно как-то задать функцию награды. 

\begin{example}
В идеале, иерархический RL должен выглядеть как-то примерно так. Менеджер смотрит на текущее состояние и говорит: нужно добраться до банка! Откуда-то берётся функция награды, описывающая задачу $g$ <<добраться до банка>>, и соответствующая стратегия рабочего $\pi(a \mid s, g)$ в течение многих шагов эту задачу решает. В банке происходит определение того, что задача решена, и управление снова передаётся менеджеру. Тот видит, что агент находится в банке, и выбирает новую задачу $g$, например, <<найти кучу денег>>. Вызывается новая стратегия рабочего, он снова решает задачу, менеджер выбирает следующую подзадачу, и так далее.

\begin{center}
    \includegraphics[width=0.7\textwidth]{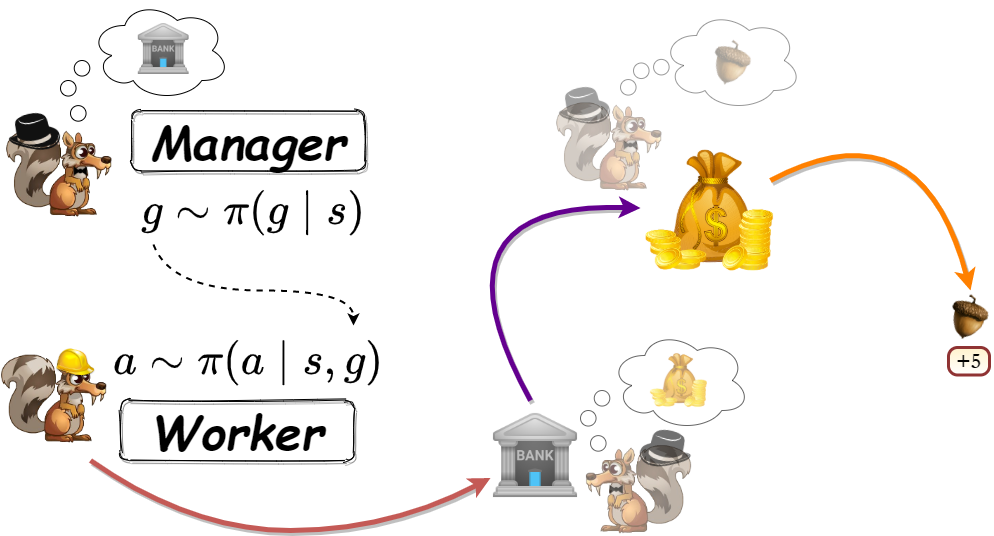}
\end{center}
\end{example}

Иными словами, задачи и подзадачи в концепции RL всегда задаются MDP, и для подобных иерархических RL-алгоритмов нам нужно определиться, что есть MDP для той стратегии, которая принимает решение о подзадачах, и что есть MDP подзадач.

Строить такую идеальную схему, чтобы алгоритм сам смог выучить и набор подзадач, и механизмы определения моментов передачи управления менеджеру, мы на текущий момент не умеем. В теории опций мы абстрагировались от идей подзадач и оптимизировали всю нашу иерархическую конструкцию на максимизацию исходной функции награды, а основная сложность заключалась в обучении политики терминальности, и с ней же были связаны основные причины нестабильности процесса обучения. В рамках двух следующих алгоритмов мы будем следовать заветам феодального RL и выберем какой-нибудь набор подзадач, но упростим схему, убрав политику терминальности и заменив её на эвристики.

\subsection{Feudal Networks (FuN)}

Напрашивается в качестве набора подзадач брать задачу достижения некоторого состояния $g \in \St$. Основная идея \emph{феодальных сетей} (feudal networks) в том, чтобы задавать рабочим не целевое состояние, а направление, в котором нам хотелось бы сдвинуться, в пространстве латентных описаний состояний. Для этого менеджер переводит текущее состояние $s_t$ в латентное пространство $z_t \coloneqq f_\theta(s_t) \in \R^d$, где $\theta$ --- параметры, после чего генерирует вектор $g_t \in \R^d$. Выданный менеджером вектор нормируется, так что $\|g_t\|_2 = 1$: эта хитрость нужна, чтобы запрашиваемое направление не близко к тривиальному нулевому смещению и не достаточно <<большое>>.

Вместо политик терминальности вводится следующее упрощение: менеджер всё-таки будет вызываться на каждом шаге, и для каждого $s_t$ будет генерироваться целевое направление $g_t$; но обучаться он будет, исходя из предположения, что рабочий действительно выполняет поставленную задачу, и через $K$ шагов, где $K$ --- гиперпараметр, агент успешно сдвинется в указанном менеджером направлении. Иначе говоря, для функции переходов менеджера (которая составлена композицией настоящей функции перехода и стратегии рабочего) вводится следующее предположение:
$$\Prob (z_{t + K} = z_t + g_t) = 1$$
Здесь $z_{t+K} \coloneqq f_{\theta}(s_{t+K})$ --- латентное описание состояния через $K$ шагов. В жёсткой форме с подобной функцией переходов работать всё равно не столь удобно, поэтому вместо этого предположим более мягкую форму:
\begin{equation}\label{misesfishermanager}
    p(z_{t + K} \mid z_t, g_t) \propto \exp(d(z_{t+K} - z_t, g_t)),
\end{equation}
где $d$ --- некоторое расстояние между векторами, например, косинусное (с учётом того, что $\|g_t\|_2 = 1$):
$$d(z_{t+K} - z_t, g_t) \coloneqq \frac{(z_{t+K} - z_t)^Tg_t}{\|z_{t + K} - z_t\|_2}$$
В таких предположениях нас волнует лишь направление $z_{t + K} - z_t$, в котором сдвинулось состояние; можно считать, что его мы тоже отнормировали, и тогда предположение \eqref{misesfishermanager} просто задаёт распределение на единичной сфере с центром в $z_k$, у которого мода находится в $g$, и чем менее скоррелированы выбранное направление и <<цель>> $g$, тем меньше вероятность.
\begin{definition}
Распределение 
\begin{equation}\label{misesfisher}
   p(d \mid g) \propto \exp(d^Tg),
\end{equation}
где $\|g\| \HM= 1$ и $\|d\| \HM= 1$, называется \emph{распределением Мизес-Фишера} (von Mises-Fisher distribution).
\end{definition}


\begin{proposition}
Нормировочная константа распределения Мизес-Фишера \eqref{misesfisher} не зависит от $g$
\begin{proof} Рассмотрим нормировочную константу для $g$:
$$\int\limits_{\|d\| = 1} \exp(d^Tg) \diff d$$
Возьмём какое-нибудь другое единичное направление $\hat{g}$. Сделаем замену переменных: пусть $M$ - матрица поворота, переводящая $\hat{g}$ в $g$: $g = M \hat{g}$. Тогда сделаем замену переменных: $\hat{d} \coloneqq M^T d$. Если $d$ пробегает единичную сферу, то так как $M$ --- матрица поворота, $\hat{d}$ тоже пробегает единичную сферу; определитель матрицы поворота $M$ также единичен, поэтому при замене переменной новых множителей не возникает. Получаем
$$\int\limits_{\|d\| = 1} \exp(d^TM\hat{g}) \diff d = \int\limits_{\|\hat{d}\| = 1} \exp(\hat{d}^T\hat{g}) \diff \hat{d},$$
что есть нормировочная константа для $\hat{g}$.
\end{proof}
\end{proposition}

Менеджер может считать, что, выбирая действие $g_t$, он выбирает состояние, в котором окажется через $K$ шагов, причём предлагается считать, что это распределение задано \eqref{misesfishermanager}, то есть:
$$\log \manager{\pi}(g_t \mid s_t) \coloneqq \log p(z_{t + c} \mid z_t, g_t) = d(z_{t+c} - z_t, g_t) + \const(\theta)$$

Подставляя это в формулу градиента для актёра, получается (для одного перехода) следующая формула:
$$\nabla_{\theta}^{\mathrm{manager}} = \nabla_{\theta} d(z_{t+K} - z_t, g_t(\theta)) \manager{A}(s_t, g_t)$$
Здесь $\manager{A}$ - Advantage-оценка критика менеджера (оцениваемая стандартным для Policy Gradient методов через обучение единственной функции $\manager{V}$), а зависимость $z_t, z_{t+K}, \manager{A}$ от параметров $\theta$ игнорируется.


Награда для рабочего в задаче следования тем направлениям, которые указывает менеджер, может быть сформулирована так:
$$r^{g}_t \coloneqq \frac{1}{K}\sum_{i = 0}^{K} d(z_t - z_{t - i}, g_{t - i}),$$
то есть для каждого из последних $K$ шагов проверяется, насколько точно рабочий следует указаниям менеджера, смещается ли в латентном пространстве менеджера описание состояния в указанном им направлении. Эта награда смешивается с наградой из исходного MDP; можно считать, что $r^{g}_t$ для рабочего --- внутренняя мотивация, а истинная награда из среды --- внешняя мотивация.

\subsection{HIRO}

Алгоритм HIRO в целом похож на FuN, хотя общая схема иерархической стратегии выглядит чуть-чуть по-другому. В этом алгоритме авторы рассматривали задачи непрерывного управления и могли для простоты положить $\G \equiv \St$, не переводя состояния в латентное пространства. Менеджер выбирает цель $g_t \in \St$ раз в $K$ шагов, где $K$ --- гиперпараметр. В последующие $K - 1$ моментов времени цели определяются по предыдущей по следующему <<векторному>> правилу:
$$g_{t + 1} - s_{t + 1} = g_t - s_t \qquad \Rightarrow \qquad g_{t + 1} = g_t + s_{t + 1} - s_t$$
Другими словами, если на шаге $t$ менеджер решил, что состояние $s_t$ должно меняться <<в направлении>> $g_t$, то дальше следующие $K$ шагов мы считаем, что хотим этому направлению следовать. Вызов раз в $K$ шагов кажется костылём, но позволяет избежать альтернативы с эвристикой FuN, в котором мы при обучении менеджера предполагали, что рабочий успешно выполняет возложенную на него задачу; тут же менеджера можно обучать лобовым подходом: он живёт даже не в sMDP, а обычном MDP, поскольку всегда гарантируется, что выбранное макро-действие будет выполняться в точности $K$ шагов. Однако, поскольку рабочие будут постепенно обучаться, это MDP нестационарное.

Внутренней мотивацией рабочего, выполняющего цель $g$, полагается расстояние от состояния, в которое попал агент, до целевого состояния $s_t + g_t$:
$$r^{\intr}(s_t, g_t, s_{t+1}) = -\|s_{t + 1} - (s_{t} + g_t)\|_2$$
Как и в FuN, такая внутренняя мотивация смешивается со внешней мотивацией --- наградой из основного MDP. 

Заметим, что рабочего можно спокойно обучать в off-policy режиме (он живёт в самом обычном MDP). Хотелось бы научиться обучать в off-policy режиме и менеджера, но понятно, что пространство действий $\G$ для менеджера с ходом обучения меняет своё семантическое значение. Предлагается лайфхак, очень похожий на идею переразметки траекторий из multi-task RL (см.~раздел \ref{subsec:hindsight}). Сохраним для менеджера в реплей буфера для $s_t, g_t$ всю последующую траекторию $a_t, s_{t+1} \dots a_{t + K - 1}, s_{t + K}$. Допустим, мы засэмплировали этот переход из буфера и хотим использовать для обучения. Проблема в том, что $\pi^g$ для $g$ из перехода уже может быть совершенно другим, и вероятность, что он сгенерировал бы такую траекторию, неприемлемо низкая. Идея: попробуем найти другое $\hat{g} \in \G$, для которого эта вероятность довольно высокая. То есть, мы приблизительно попытаемся решить следующую задачу:
$$\log p(a_t, s_{t+1} \dots a_{t + K - 1}, s_{t + K} \mid s_t, \hat{g}) \to \max_{\hat{g}}$$

Рассмотрим, из чего состоит правдоподобие. Мы предполагаем, что менеджер делает выбор в момент времени $t$, то есть менеджер больше на это правдоподобие не влияет. Есть логарифмы переходов $p(s' \mid s, a)$, которые не зависят от $\hat{g}$, поэтому их можно опустить. Остаются только правдоподобия выборов действий:
$$\sum_{\hat{t} = t}^{t + K - 1} \log \pi_{\hat{g}}(a_{\hat{t}} \mid s_{\hat{t}}) \to \max_{\hat{g}}$$

Почти всегда мы можем для данного $\hat{g}$ посчитать значение выражения. Предлагается взять несколько (штук восемь-десять) $g$, посчитать значение выражения и выбрать наилучшее. 

Какую хорошую стратегию перебора для сэмлпирования кандидатов $g$ выбрать? Поскольку агент в итоге сдвинулся на вектор $s_{t + K} - s_t$, то можно использовать этот вектор в качестве центра гауссианы, из которой будет проводиться сэмплирование. Также предлагается попробовать взять $g$ из буфера и сам центр $s_{t + K} - s_t$.
\section{Частично наблюдаемые среды}\label{sec:PoMDP}

\subsection{Частично наблюдаемые MDP}

Рассматривавшиеся MDP до этого являлись \emph{полностью наблюдаемыми} (fully observable): агенту в качестве наблюдения было доступно всё состояние целиком. Конечно, это довольно-таки существенное упрощение.

\begin{definition}
MDP называется \emph{частично наблюдаемым} (partially observable, принятое сокращение --- PoMDP), если дополнительно задано множество $\mathcal{O}$, называемое \emph{пространством наблюдений} (observation space), и распределение $p(o \mid s)$, определяющая вероятность получить то или иное наблюдение агента $o \in \mathcal{O}$ в момент времени, когда мир находится в состоянии $s \in \St$.
\end{definition}

Если MDP --- <<управляемая>> марковская цепь, то PoMDP можно рассматривать как <<управляемую>> \href{https://en.wikipedia.org/wiki/Hidden_Markov_model}{скрытую марковскую цепь}: действия влияют на то, как будут порождаться следующие состояния, но для наблюдения доступны не сами состояния, а только какие-то другие случайные величины, про которые, однако, известно, что они зависят только от текущего состояния. При этом состояния всё также удовлетворяют свойству марковости, а функция переходов и процесс генерации наблюдений по состоянию --- свойству стационарности.

В PoMDP стратегия должна уметь принимать решение на основании не только текущего наблюдения, но всей истории цепочки наблюдений, так как в них может содержаться информация о текущем состоянии среды. Естественно, эта задача более приближена к реальности, но сильно сложнее. Агенту теперь нужна модель \emph{памяти} (memory), способ хранения и учёта всей истории в течение эпизода.

Заметим, что награда по своему определению наблюдаема. Если раньше можно было считать, что награда --- часть состояния, то теперь награда <<по логике>> модели есть часть наблюдений, а значит, формально может рассматриваться как функция от наблюдений. Далее будем работать в соглашении, что награда --- детерминированная функция от наблюдений.

\subsection{Belief MDP}

Допустим, задано PoMDP, и нам доступны все распределения (вероятности переходов и вероятности наблюдений). Пусть мы знаем вероятность $p(s_0)$, какое состояние генерируется изначально (будем считать, стохастично). Допустим, начался новый эпизод, и мы получили наблюдение $o_0 \in \mathcal{O}$. Что мы можем сказать о том, в каком состоянии $s_0$ мы на самом деле оказались? Ответ на этот вопрос формально даёт формула Байеса:
\begin{equation}\label{initial_belief}
p(s_0 \mid o_0) = \frac{p(o_0 \mid s_0)p(s_0)}{\int_{\mathcal{O}} p(o_0 \mid s_0)p(s_0) \diff o_0}
\end{equation}
Здесь в числителе стоит вероятность первого наблюдения и априорное распределение на состояниях, а в знаменателе стоит нормировочная константа. Пока будем считать, что мы умеем брать любые интегралы (например, если пространства состояний и наблюдений конечны и малы).

Допустим, мы выбрали действие $a_0$, попали в состояние $s_1$ (которое мы не видим) и получили наблюдение $o_1$. Что мы можем сказать о том, в каком состоянии $s_1$ мы находимся? Опять же, пользуясь формулой Байеса и используя вероятностную модель PoMDP, мы можем получить точную формулу наших представлений о том, в каком состоянии на текущий момент пребывает агент. Чтобы проделать вывод, обозначим за $\Traj_{:t}$ наблюдаемую (действия, награды, наблюдения, но не состояния) траекторию до момента времени $t$, включая последнее наблюдение:
$$\Traj_{:t} \coloneqq (o_0, a_0, r_0, o_1, a_1, r_1 \dots o_{t - 1}, a_{t - 1}, r_{t - 1}, o_t)$$

\begin{definition}
Вероятность пребывания в том или ином состоянии при условии наблюдаемой истории $p(s_t \HM\mid \Traj_{:t})$ называется \emph{belief state} и сокращённо обозначается $b_t$.
\end{definition}

По определению $b_0(s_0) \HM= p(s_0 \HM\mid o_0)$, которую мы получили в по формуле Байеса в \eqref{initial_belief}. Как обновить belief state после совершения одного шага в среде (получения одного наблюдения и выбора одного действия)?

\begin{theorem}
С точностью до нормировочной константы:
\begin{equation}\label{beliefupdate}
b_t(s_t) \propto p(o_{t} \mid s_{t})\int\limits_{\St} p(s_t \mid s_{t - 1}, a_{t - 1})b(s_{t - 1}) \diff s_{t - 1}
\end{equation}
\begin{proof}
Нашей целью является посчитать $b_t(s_t) \HM\coloneqq p(s_t \HM\mid \Traj_{:t}) \HM= p(s_t \HM\mid o_t, a_{t-1}, \Traj_{:{t-1}})$. Применим формулу Байеса:
$$b_t(s_t) \propto p(o_t \mid s_t, a_{t-1}, \Traj_{:{t-1}})p(s_t \mid a_{t-1}, \Traj_{:{t-1}}) = (*)$$

Первый множитель здесь по определению вероятностной модели есть $p(o_t \mid s_t)$, а во втором множителе, чтобы воспользоваться функцией переходов, нужно проинтегрировать по неизвестному нам предыдущему состоянию $s_{t - 1}$:
\begin{align*}
(*) 
&= p(o_t \mid s_t) \int\limits_{\St} p(s_t, s_{t - 1} \mid a_{t - 1}, \Traj_{:{t-1}}) \diff s_{t - 1} = \\ 
&= p(o_t \mid s_t) \int\limits_{\St} p(s_t \mid s_{t - 1}, a_{t - 1})p(s_{t - 1} \mid \Traj_{:{t-1}}) \diff s_{t - 1}
\end{align*}

Осталось заметить, что по определению $p(s_{t - 1} \mid \Traj_{:{t-1}}) = b_{t - 1}(s_{t - 1})$ --- информация о текущем состоянии с прошлого шага.
\end{proof}
\end{theorem}

\begin{exampleBox}[righthand ratio=0.6, sidebyside, sidebyside align=center, lower separated=false]{}
Допустим, в PoMDP с 10 состояниями и действиями вправо-влево получаем детерминировано в качестве наблюдения бинарный флаг: есть ли пальма или нет. Начальное состояние определяется случайно равномерно. Посмотрим, как меняется belief state, если агент заспавнился в самом левом состоянии и дальше выбирает действия <<вправо>>.

\tcblower
\animategraphics[controls, width=\linewidth]{1}{Images/BeliefMDP/BeliefMDP-}{1}{5}
\end{exampleBox}

Belief state --- <<достаточная статистика>> для описания всей предыдущей истории. Посчитав его, мы собрали абсолютно всю имеющуюся у нас информацию. Это замечание позволяет формально свести задачу в PoMDP к обычному MDP, но требует знания всех распределения и возможности пересчитывать belief state:

\begin{definition}
Для данного PoMDP \emph{Belief MDP} называется следующее MDP:
\begin{itemize}
    \item Пространством состояний является пространство $\Prob(\St)$ распределений в $\St$;
    \item Пространством действий является $\A$;
    \item Функция переходов $p(b' \mid b, a)$, где $b, b' \in \Prob(\St)$ устроена так: сэмплируются $s \HM\sim b(s)$, $s' \HM\sim p(s' \HM\mid s, a)$, $o' \HM\sim p(o' \mid s')$, после чего новый belief state рассчитывается по формуле:
    $$b'(s') \propto p(o' \mid s')\int\limits_{\St} p(s' \mid s, a) b(s) \diff s $$
    \item Функция награды для текущего состояния $b$ и действия $a$ устроена так: для того же сэмпла $s$, использованного в переходе, выдаётся награда $r(s, a)$.
\end{itemize}
\end{definition}

Теория Belief MDP позволяет предложить алгоритмы динамического программирования наподобие Value Iteration и Policy Iteration для решения задач в PoMDP. Однако, в реальности работать с Belief MDP в средах со сложным пространством состояний довольно неудобно: если все распределения неизвестны, интегралы в формуле обновления \eqref{beliefupdate} не берутся, и даже просто хранить <<распределение над состояниями>> невозможно, то не совсем понятно, что можно извлечь от этой теории в сложных средах. Понятно, что при взаимодействии со средой агент должен помимо прочего стремиться выполнять те действия, которые дают наибольшую информацию о belief state, и позволяют <<локализоваться>> в пространстве состояний.

\subsection{Рекуррентные сети в Policy Gradient}

Видимо, поэтому пока самое распространённое решение для работы в PoMDP --- добавление в архитектуры моделей рекуррентных слоёв (GRU или LSTM). При использовании рекуррентных стратегий скрытое состояние инициализируется нулевым вектором в начале каждого эпизода и <<хранит всю необходимую информацию>> о предыдущих наблюдениях, а все модели на вход получают лишь то, что поступает с сенсоров. Де-факто все наши модели оценочных функций и стратегии становятся функциями от всей истории $o_0, o_1, o_2 \dots o_t$. Обсудим, как это меняет ранее встречавшиеся алгоритмы.

Чистые on-policy алгоритмы, такие как REINFORCE, A2C и мета-эвристики, практически не меняются. В A2C лишь стоит иметь в виду, что градиенты текут только по тому роллауту, который собран на текущей итерации; таким образом, агенту будет тяжело научиться запоминать информацию дольше, чем на $N$ шагов, где $N$ --- длина роллаутов.

В PPO скрытое состояние модели сохраняется для всего собранного датасета. Далее из датасета сэмплируются не мини-батчи, а мини-батчи роллаутов (некоторой длины $N$). Затем скрытое состояние инициализируется сохранённым значением при сборе датасета (поскольку стратегия <<не меняется сильно>>, считается, что это сохранённое состояние более-менее не устарело), и модель прогоняется на засэмплированном роллауте для подсчёта всех градиентов.

\subsection{R2D2}

В off-policy алгоритмах сохранять значение скрытого состояния в реплей буфере на первый взгляд бессмысленно; модель может изменится сколько угодно сильно, и сохранённые значения памяти будут неактуальны. С течением обучения скрытое состояние рекуррентных сетей просто меняет своё семантическое значение: модель со свежими весами просто по-другому будет интерпретировать латентное представление.

Рассмотрим две эвристики для преодоления этой проблемы из алгоритма R2D2. Из реплей буфера генерируются роллауты некоторой длины $N$. Скрытое состояние инициализируется сохранённым в буфере значением (которое потенциально <<протухло>>). Модель прогоняется заново на всём роллауте, но функция потерь и backpropagation считается лишь на последних $M \HM< N$ шагов. Таким образом обучение проводится стандартным для рекуррентных сетей подходом backpropagation through time (BPTT).

Первые же $N \HM- M$ шагов нужны исключительно для <<\emph{прогрева}>> (burn-in) скрытого состояния, чтобы его семантическое значение начало более-менее соответствовать свежим весам модели. Это не так дорого вычислительно, как заново прогонять модель по всему эпизоду; так можно было бы получить <<честные>> актуальные значения памяти, но параллелизовать такие вычисления по батчу, в котором встречаются роллауты из разных моментов эпизодов, неудобно и достаточно дорого. Естественно, аналогичный прогон проводится для таргет-сети.

\begin{center}
    \includegraphics[width=0.9\textwidth]{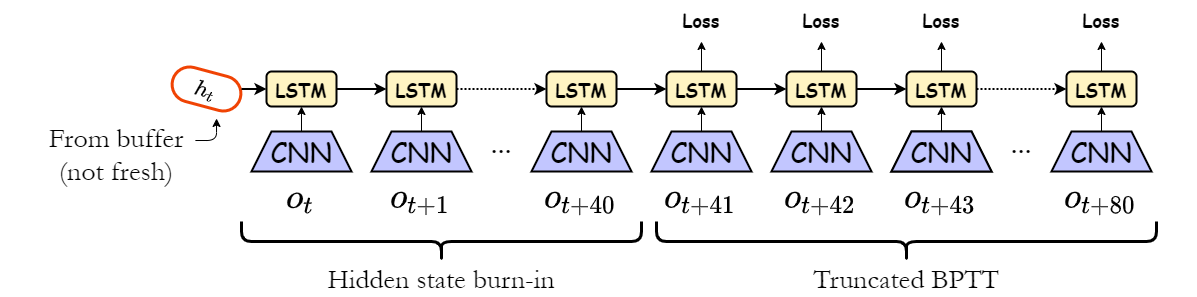}
\end{center}

Вторая эвристика заключается в том, чтобы <<разогретое>> состояние после $N \HM- M$ шагов сохранить в буфере для того роллаута, который начинается с соответствующего перехода, и таким образом сделать сохранённое значение чуть более актуальным.

Как обычно, длина $M$ той части траектории, вдоль которой пропускаются градиенты, и ограничивает то время, <<на которое>> агент может теоретически научиться что-то запоминать. Однако, увеличивать его может быть чревато не только тем, что алгоритм станет вычислительно сложным, но и тем, что в батче оказываются $M$ очень похожих друг на друга переходов, и перебивать эту скоррелированность лоссов придётся размером мини-батча.

\begin{remark}
Может быть удобно для ускорения сэмплирования хранить в реплей буфере эпизоды, уже разбитые на фрагменты удобного размера. Например, авторы R2D2 выбирают $N \HM= 80$, $M \HM= 40$, и при сборе данных собирают 40 последовательных переходов, которые упаковываются в буфер единым комплектом; из-за такой оптимизации промежуточные переходы в этом роллауте никогда не будут засэмплированы для обучения как, например, стартовые (с которых начинается роллаут из $N$ шагов). Похоже, что это не так критично, зато сильно упрощает код и сэмплирование мини-батчей.
\end{remark}

\subsection{Neural Episodic Control (NEC)}

\emph{Эпизодичная память} означает, что вместо (или дополнительно к) рекуррентных связей в качестве памяти используется набор ранее встретившихся состояний или их латентных описаний. Механизм работы с такой памятью очень схож с механизмом внимания. Рассмотрим алгоритм Neural Episodic Control в качестве примера применения такой идеи.

Для каждого действия $a$ будем хранить \emph{словарь} (dictionary), то есть некоторый набор ключ-значение $M_a \coloneqq (k, v)$. \emph{Ключом} (key) $k \in \R^d$ является вектор-эмбеддинг, описывающий некоторое состояние $s$ из опыта агента. \emph{Значением} (value) $v$ является скаляр, приближённо оценивающий $Q^*(s, a)$.

При взаимодействии со средой агент действует следующим образом. Текущее состояние $s$ подаётся на вход сети с параметрами $\theta$, которая выдаёт его описание $q_{\theta}(s) \in \R^d$. Выходом $Q_{\theta}(s, a)$ модели является результат применения \emph{механизма внимания} (attention) к словарю $M_a$ с \emph{запросом} (query) $q_{\theta}(s)$: для каждого ключа $k$ в словаре определяется похожесть запроса на ключ $\rho(q_{\theta}(s), k)$ при помощи некоторой функции близости $\rho$, например:
$$\rho(q, k) \coloneqq \frac{1}{\|q - k\|_2^2 + \delta},$$
где $\delta$ --- небольшая константа для защиты от деления на ноль. Затем для каждой пары ключ-значение определяется его вес:
$$w(k) \propto \rho(q, k),$$
где нормировочная константа вычисляется из соображения $\sum_{k, v \in M_a} w(k) \HM= 1$. Наконец, выход дифференцируемого чтения из словаря определяется как сумма значений с вычисленными весами:
$$Q_{\theta}(s, a) \coloneqq \sum_{k, v \in M_a} w(k)v$$

Поскольку размер словаря может быть достаточно большой, а процедуру нужно проводить для каждого $a \in \A$, предлагается использовать в формуле две аппроксимации. Во-первых, учитываются только топ-50 самых значимых (имеющих наибольший вес $w(k)$) пар ключей-значений. Иными словами, по расстоянию $\rho$ находятся 50 ближайших к $q_{\theta}(s)$ ключей в словаре, и веса с учётом нормировочной константы вычисляются только для них; только они используются для вычисления финальной суммы. Во-вторых, поиск топ-50 ближайших соседей проводится приближённо, для ускорения вычислений.

Видно, что при фиксированном словаре $M_a$ функция $Q_{\theta}(s, a)$, вычисленная таким образом, дифференцируема по параметрам.

При помощи такой модели Q-функции агент выбирает действие $\eps$-жадно и получает для выбранного действия $a$ следующее состояние $s'$ и награду за шаг $r$. Теперь он может посчитать для примера $s, a$ целевую переменную (авторы используют $N$-шаговую оценку, для чего нужно провести $N$ шагов взаимодействия со средой; для простоты положим $N=1$):
$$\hat{Q}(s, a) = r + \gamma \max_{a'} Q_{\theta}(s', a')$$
где значения $Q_{\theta}$ также получены проведением всей вышеописанной процедуры. Полученная пара $(q_{\theta}(s), \hat{Q}(s, a))$ добавляется в словарь $M_a$ и используется при дальнейших проходах через сеть. У словарей, естественно, есть некоторое ограничение по объёму, и при добавлении новой пары в заполненный словарь предлагается выкидывать самую редко используемую пару $(k, v)$: ту, для которой число попаданий в топ-50 ближайших соседей наименьшее (для этого достаточно дополнительно хранить счётчики использований).

Тройка $(s, a, \hat{Q}(s, a))$ добавляется в реплей буфер. На этапе обучения, который проводится после каждого шага взаимодействия со средой, сэмплируется мини-батч троек $(s, a, \hat{Q}(s, a))$ из буфера, и модель с текущими словарями $M_a$ прогоняется на $s, a$ для получения оценки $Q_{\theta}(s, a)$:
\begin{equation*}
\left( Q_{\theta}(s, a) - \hat{Q}(s, a)\right)^2 \to \min_{\theta, k, v}
\end{equation*}
Запись, что минимизация ведётся как по $\theta$, так и по $k, v$, означает, что минимизация ведётся не только по параметрам сети, но и по ключам и значениям, хранящимся на текущий момент в словаре $M_a$. Таким образом, эмбеддинги и значения обновляются в том числе для хранящихся в <<эпизодичной памяти>> состояний.

Заметим, что $\hat{Q}(s, a)$ здесь --- целевая переменная, посчитанная в момент сбора перехода, и она не пересчитывается; иначе говоря, используется по сути on-policy режим обучения. Интерпретация такого хода в том, что отказ пересчитывать значение целевой переменной $\hat{Q}(s, a)$ по сути эквивалентен использованию замороженной таргет-сети; но, чтобы эти значения не устаревали сильно, размер реплей буфера придётся существенно сократить (авторы используют реплей буфер размера $10^5$, что на порядок меньше типично используемого буфера при off-policy обучении).
\section{Мульти-агентное обучение с подкреплением}

\subsection{Связь с теорией игр}

Попробуем представить ситуацию, когда в среде действует два агента. Функция переходов теперь зависит от действий каждого из агентов $p(s' \mid s, a^1, a^2)$, а каждый агент оптимизирует свою функцию награды $r^i(s, a^1, a^2)$, где $i$ --- номер агента.

Рассмотрим сеттинг, аналогичный бандитам: игра с одним состоянием, заканчивающаяся после первого выбора. Допустим также детерминированность функций наград. Мы получим в чистом виде задачу, исследующуюся в \emph{теории игр}: заданы две функции $r^1(a^1, a^2), r^2(a^1, a^2)$, зависящие от действий обоих игроков; игроки оптимизируют каждый свою функцию, при этом производя выбор действий одновременно. Отсюда видно, что любой мульти-агентный RL тесно связан с теорией игр: в частности, при обучении агенты могут застрять не только в локальных оптимумах, но и в \href{https://ru.wikipedia.org/wiki/\%D0\%A0\%D0\%B0\%D0\%B2\%D0\%BD\%D0\%BE\%D0\%B2\%D0\%B5\%D1\%81\%D0\%B8\%D0\%B5_\%D0\%9D\%D1\%8D\%D1\%88\%D0\%B0}{\emph{равновесиях Нэша}} (Nash equilibrium), когда ни один другой агент не может поменять свою стратегию в предположении неизменности стратегий других агентов так, чтобы увеличить свою награду.

В общем случае в среде действует $N$ агентов. Будем обозначать как $a^i$ действие $i$-го агента; $\vec{a}$ --- вектор из всех действий всех агентов; $a^{-i}$ --- действия всех агентов, кроме $i$-го. Тогда функция переходов теперь выглядит как $p(s' \mid s, \vec{a})$, а $i$-ый агент оптимизирует свою функцию награды $r^i(s, \vec{a})$. Природа этих функций наград может быть самой разной.

\begin{definition}
Если функции наград всех агентов совпадают, то задача называется \emph{кооперативной}.
\end{definition}

\begin{definition}
Задача, в которой сумма наград всех агентов за шаг нулевая $\sum\limits_{i} r^i(s, \vec{a}) \HM= 0$, называется \emph{антагонистической} или \emph{игрой с нулевой суммой}.
\end{definition}

Это предельные случаи частных ситуаций: если $r^1$ в целом велико там, где велико $r^2$, то эти два агента, видимо, должны научиться вместе выполнять какую-то задачу. Если же большое $r^1$ влечёт малое $r^2$, то агенты противостоят друг другу. В общем случае у каждого агента может быть какая-то своя задача; этот общий случай иногда называют \emph{смешанной} (mixed) задачей мульти-агентного RL. Зачастую агенты действуют в условиях частичной наблюдаемости, и тогда у каждого агента могут быть свои наблюдения; это означает, что у $i$-го агента своя функция $p^i(o \mid s)$, определяющая наблюдение по текущему состоянию мира. Для простоты далее частичная наблюдаемость опущена. 

\subsection{Централизация обучения}

\emph{Децентрализованный сеттинг} означает, что у агентов нет какой-либо общего <<сервера>> с общими данными, в том числе во время обучения. Простейший подход мульти-агентного RL в рамках такого сеттинга --- завести для каждого агента свою стратегию $\pi^i(a \mid s)$ и оценочную функцию $Q^i(s, a^i)$, которая будет зависеть только от действия $i$-го агента, и оптимизировать их с опыта данного агента любыми обычными RL алгоритмами. Тот факт, что в среде действуют другие агенты, игнорируется. Важный момент: если другие агенты не обучаются (не изменяются с течением времени), то такой подход обучения $i$-го агента корректен.

\begin{proposition}
Если все агенты, кроме $i$-го, стационарны, то $i$-ый агент живёт в MDP с функцией переходов
$$p(s' \mid s, a^i) \coloneqq \int p(s' \mid s, a^i, a^{-i})\pi^{-i}(a^{-i} \mid s) \diff a^{-i},$$
где $\pi^{-i}(a^{-i} \mid s)$ --- стратегия остальных агентов, в предположении их независимого выбора своих действий равная
$$\pi^{-i}(a^{-i} \mid s) \coloneqq \prod_{j \ne i} \pi^j(a^j \mid s)$$
\end{proposition}

Это означает, что если в среде действует другой агент, использующий стационарную стратегию, среда остаётся стационарной и вся теория остаётся справедлива. Однако, если в среде действует другой агент, который обучается (и, значит, меняет свою стратегию с течением времени), то для прочих агентов это означает, что постоянно изменяются <<законы физики>> окружающей среды, и мы под изученную теорию подпадать перестаём. 

В нестационарных средах естественно использовать on-policy обучение. Запуск какого-нибудь on-policy алгоритма с игнорированием факта присутствия других обучающихся агентов обычно оказывается серьёзным бэйзлайном.

\emph{Полностью централизованный сеттинг} возможен для агентов, например, оптимизирующих одну и ту же функцию награды, находящихся в кооперации. Полная централизация означает, что для построения системы из нескольких кооперирующихся агентов все они рассматриваются как единый агент. Наблюдение есть совокупность наблюдений всех агентов, действия --- набор действий для каждого из агентов, и тогда по сути задача сведена к <<одноагентной>> ситуации. В большинстве случаев такой вариант не годится по самой постановке задачи: наличие единого <<центрального сервера>>, который бы принимал решения за всех агентов в среде, может быть просто-напросто очень дорогим. Хотя бы потому что агентов может быть много, и пространство состояний и, главное, действий, взорвётся. В большинстве случаев хочется, чтобы агенты принимали решения на основе только имеющейся у них информации.

Зачастую доступен интересный промежуточный вариант (\emph{частичная централизация}), когда у каждого из кооперирующихся агентов хранится и обучается персональная стратегия, но, например, оценочная функция $Q(s, a^1, a^2 \dots)$ хранится и обучается на общем для всех агентов <<сервере>>. Важно, что эта оценочная функция нужна в on-policy алгоритмах только для обучения, и при использовании полученных стратегий <<сервер>> не потребуется. Это означает, что по окончании обучения в таком сеттинге для использования стратегии никакого центрального сервера не понадобится. Такой подход называется \emph{centralised training with decentralised execution} (CTDE), и многие алгоритмы для мульти-агентного RL строятся именно в рамках этой парадигмы.

В рамках парадигмы CTDE, помимо прочего, возможен \emph{parameter sharing}: если задача симметрична для каких-то агентов, то веса их стратегий считаются общими. Градиенты для их обучения тогда вычисляются на общем сервере; дальше во время использования агенты используют одну и ту же модель для принятия решений. Другими словами, в рамках CTDE подхода при желании обучить колонию из 100 одинаковых муравьёв, вам понадобится лишь одна модель; без частичной централизации же, в децентрализованном сеттинге, подразумевается, что каждый из муравьёв должен как-то сам обучиться, и тогда у каждого из сотни получится в итоге своя модель.

\subsection{Self-play в антагонистических играх}

В симметричных антагонистических играх обучение на опыте игр с самим собой оказалось мощным инструментом. Этот подход называют \emph{self-play}, и заключается он в том, что за игроков играет одна и та же стратегия (такие игры обычно симметричны, и поэтому можно считать, что применяется parameter sharing). Собранный опыт со стороны каждого игрока можно использовать для обучения, и он обладает одним очень ценным свойством: сбалансированностью. На какой бы стадии обучения ни был агент, играет ли он на уровне гранд-мастера или бесконечно сильно тупит, при игре с самим собой в половине случаев он <<выигрывает>>, в половине --- <<проигрывает>>. Разница в сигнале всегда даёт возможность дальнейшего улучшения.

В режиме self-play можно запустить любые алгоритмы обучения: как model-free, так и model-based. В частности, в model-based алгоритмах вроде AlphaGo и MuZero в ходе планирования можно при построении дерева ходы за оппонента проводить при помощи тех же моделей, но не максимизировать, а минимизировать награду. Понятно, что идея применима как в ситуации, когда игроки ходят одновременно, так и когда игроки ходят по очереди (что более типично для игр вроде шахмат и го).

Конечно, self-play не гарантирует, что по итогам обучения стратегия будет способна адаптироваться к произвольному оппоненту. Можно показать, что к обученным self-play методам стратегии можно применить \emph{adversarial-атаку}: зафиксировать её и обучать стратегию оппонента побеждать. Такие стратегии, которые <<взламывают>> стратегию оппонента, зачастую ведут себя очень странно, и могут победить, ничего не делая.

\begin{example}[Adversarial Policy]
По \href{https://adversarialpolicies.github.io/}{данной ссылке} в примерах слева показаны игры стратегии, обученной в режиме self-play. На вид кажется, что полученная стратегия невероятно крута. Далее эта стратегия фиксировалась, и обучалась с нуля новая стратегия оппонента <<побеждать>> фиксированную. Примеры этих игр показаны справа: новая стратегия оппонента просто падает, сбивает с толку <<крутую>> стратегию, и та отправляется в какой-то нокаут.
\end{example}

\subsection{QMix в кооперативных играх}

Рассмотрим кооперативный сеттинг, где все агенты максимизируют одну и ту же функцию награды. Будущую кумулятивную награду после выполнения набора действий $\vec{a}$ в состоянии $s$ для набора политик $\pi^i(a^i \mid s)$ обозначим как $Q^{\mathrm{tot}}(s, \vec{a})$, скаляр. Попробуем построить алгоритм в рамках парадигмы CTDE, то есть обучить <<на центральном сервере>> стратегии $\pi^i(a^i \mid s)$, которые смогут дальше действовать в среде без использования подобного сервера.

Допустим, мы знаем $Q^{\mathrm{tot}}(s, \vec{a})$. Тогда оптимально выбрать действия
$$\argmax_{\vec{a}} Q^{\mathrm{tot}}(s, \vec{a}),$$
однако есть две проблемы. Во-первых, если агентов много, поиск такого аргмаксимума может стать слишком сложной задачей, ровно как и моделирование такой оценочной функции. Во-вторых, в рамках CTDE в частично наблюдаемом сеттинге каждый агент на вход вместо $s$ на самом деле получает своё собственное наблюдение $o^i$ и не знает наблюдения других агентов, когда $Q^{\mathrm{tot}}$ принимает на вход наблюдения всех агентов. С обоими проблемами можно побороться следующим образом.

\begin{definition}
\emph{Смешивающей сетью} (mixing network) назовём моделирование $Q^{\mathrm{tot}}(s, \vec{a})$ в следующем виде:
\begin{equation}\label{vdn}
Q^{\mathrm{tot}}(s, \vec{a}) \approx Q^{\mathrm{tot}}_{\phi}(Q^1(s, a^1), Q^2(s, a^2), \dots Q^N(s, a^N), s),
\end{equation}
где $\phi$ --- параметры, $Q^i(s, a^i)$ --- скаляры, зависящие только от той информации, которая доступна $i$-му агенту.
\end{definition}

В частности, в частично наблюдаемом сеттинге $Q^i(s, a^i)$ вместо $s$ принимает на вход наблюдение $i$-го агента $o^i$ и моделируется рекуррентной сетью. Сразу оговоримся, что здесь $Q^i(s, a^i)$, несмотря на принятое обозначение, не является оценочной функцией и не имеет смысл будущей кумулятивной награды. Это лишь некоторая промежуточная вспомогательная величина, через которую по некоторым правилам выражается $Q^{\mathrm{tot}}(s, \vec{a})$, награда <<всей команды>>.

\begin{example}[Value decomposition network] Рассмотрим самый простой пример смешивающей сети, не использующей параметров $\phi$ вовсе:
$$Q^{\mathrm{tot}}_{\phi}(s, \vec{a}) \coloneqq \sum_i Q^i(s, a^i)$$
У этой декомпозиции есть интересное свойство: 
\begin{equation}\label{consistent_vdn}
\argmax_{\vec{a}} Q^{\mathrm{tot}}(s, \vec{a}) = \left[ 
\begin{matrix}
\argmax\limits_{a^1} Q^1(s, a^1) \\
\argmax\limits_{a^2} Q^2(s, a^2) \\
\vdots \\
\argmax\limits_{a^N} Q^N(s, a^N) \\
\end{matrix}
\right]
\end{equation}
\end{example}

Другими словами, будущая награда команды моделируется через какие-то псевдооценочные функции каждого агента команды в отдельности. Нам крайне интересно свойство \eqref{consistent_vdn}.

\begin{definition}
Скажем, что смешивающая сеть \emph{консистентна} (consistent), если для всех состояний выполняется \eqref{consistent_vdn}.
\end{definition}

Консистентность означает, что каждому агенту вовсе не нужно знать наблюдения других агентов или обращаться к центральному серверу для выбора оптимального действия: он берёт лишь значение свой псевдооценочной функции $Q^i(s, a^i)$ и считает аргмакс по нему. Свойство \eqref{consistent_vdn} гарантирует, что это даст аргмаксимум по всему набору действий $\vec{a}$ всей команды. Поскольку выбор декомпозиции --- наш произвол, нам хочется выбрать такое параметрическое семейство смешивающих сетей, которое гарантирует это свойство. В целом, это не так сложно.

\begin{theorem}
Смешивающая сеть \eqref{vdn} консистентна, если
\begin{equation}\label{qmix_condition}
\frac{\diff Q^{\mathrm{tot}}_{\phi}}{\diff Q^i} \ge 0,
\end{equation}
то есть для всех состояний $s$ и при любых значениях параметров $\phi$ награда команды монотонно зависит от значений псевдооценочных функций агентов $Q^i(s, a^i)$.
\begin{proof}
В силу монотонности, для любых действий $\vec{a}$:
$$Q^{\mathrm{tot}}_{\phi}( \max_{a^1} Q^1(s, a^1), \dots \max_{a^N} Q^N(s, a^N), s) \ge Q^{\mathrm{tot}}_{\phi}( Q^1(s, a^1), \dots Q^N(s, a^N), s),$$
следовательно максимизация всех аргументов монотонной функции максимизирует саму функцию.
\end{proof}
\end{theorem}

Можно ли придумать какое-нибудь богатое семейство смешивающих сетей с параметрами, для которых было бы выполнено \eqref{qmix_condition}? Давайте возьмём нейросеть с параметрами $\phi$, которая принимает на вход скаляры $Q^1(s, a^1), \dots Q^N(s, a^N)$ и выдаёт скаляр $Q^{\mathrm{tot}}_{\phi}$. Тогда:
\begin{proposition}
Пусть нейросеть полносвязная, состоит из чередования линейных слоёв и монотонных функций активаций, и все параметры нейросети $\phi_i$ (кроме, возможно, смещений в линейных слоях) неотрицательны. Тогда выполнено \eqref{qmix_condition}.
\begin{proof}
Композиция монотонных преобразований монотонно; линейная комбинация с положительными весами (и произвольными смещениями) также монотонно не убывает как функция от входов.
\end{proof}
\end{proposition}

Итак, любая нейросеть с положительными весами подходит нам в качестве смешивающей. Сделаем ещё одно наблюдение: эта нейросетка очень маленькая, ведь на вход ей подаётся вектор размерности $N$, где $N$ --- число агентов, а на выходе скаляр. Заметим также, что веса могут зависеть от состояния: мы можем использовать разные $\phi$ для разных состояний $s$, и отсюда возникает идея, как можно повысить гибкость нашей value decomposition network.

\begin{definition}
\emph{Гиперсетью} (hypernetwork) называется нейросеть, выдающая веса для другой нейросети.
\end{definition}

Мы заведём гиперсеть $\phi_{\theta}(s)$, которая для данного состояния $s$ будет с параметрами $\theta$ выдавать на центральном сервере веса для смешивающей сети. В частности, $\phi_{\theta}(s)$ уже не обязано быть монотонным и может моделироваться произвольной обычной нейросетью. Мы получим, что в каждом состоянии у нас есть какое-то своё монотонное преобразование псевдооценочных функций каждого агента в оценочную функцию команды, и это очень удобно.

Итого в алгоритме QMIX оценочная функция моделируется в следующем виде:
$$Q^{\mathrm{tot}}_{\phi_{\theta}(s)}(s, \vec{a}, \psi) \coloneqq Q^{\mathrm{tot}}_{\phi_{\theta}(s)}( Q^1(s, a^1, \psi^1), \dots Q^N(s, a^N, \psi^N)),$$
где $\psi^i$ --- параметры псевдооценочной функции $i$-го агента (возможен parameter sharing, если задача симметрична для некоторых агентов), $\theta$ --- параметры гиперсети, $\phi_{\theta}(s)$ --- получающиеся параметры смешивающей нейросети.

Процесс обучения стандартный и похож на обычный DQN. При взаимодействии со средой $i$-ый агент использует свою псевдооценочную функцию $Q^i(s, a^i)$ и ведёт себя $\eps$-жадно. Собранные переходы $(s, \vec{a}, r, s')$ собираются на центральном сервере. Для перехода $\T \coloneqq (s, \vec{a}, r, s')$ целевая переменная строится как 
$$y(\T) \coloneqq r + \gamma \max_{\vec{a}'} Q^{\mathrm{tot}}_{\phi_{\theta^-}(s)}(s', \vec{a}', \psi^-),$$
где $\theta^-, \psi^-$ --- замороженные параметры таргет-сети. Далее с таким таргетом минимизируется MSE по всем параметрам:
$$\E_\T (y(\T) - Q^{\mathrm{tot}}_{\phi_{\theta}(s)}(s, \vec{a}, \psi))^2 \to \min_{\theta, \psi}$$
После окончания обучения центральный сервер, на котором хранится гиперсеть и смешивающая сеть, не нужны, и агенты используют лишь локальные певдооценочные функции $Q^i(s, a^i)$.

\subsection{Multi-Agent DDPG (MADDPG) в смешанных играх}

Рассмотрим смешанную игру, где функции наград агентов произвольны. На <<центральном сервере>> будем хранить общую оценочную функцию, которая теперь должна выдавать будущие награды сразу для всех агентов. 

\begin{definition}
При данных политиках $\pi^j(a_j \HM\mid s)$ \emph{централизованной оценочной функцией} (centralized state-action value function)  обозначим $\vec{Q}(s, \vec{a})$, возвращающую вектор, $i$-ая компонента которого $Q^i(s, \vec{a})$ равна будущей кумулятивной награде, которую получит $i$-ый агент после выполнения набора действий $\vec{a}$ в состоянии $s$.
\end{definition}

На центральном сервере понятно, как учить такую Q-функцию, причём это можно делать в off-policy режиме. Для произвольного перехода из буфера $\T \coloneqq (s, \vec{a}, \vec{r}, s')$ строим таргет
$$y(\T) \coloneqq \vec{r} + \gamma \vec{Q}(s', \vec{a}'),$$
где $\vec{a}'$ порождены оцениваемыми $\pi^j(a'_j \HM\mid s')$ (для стабилизации процесса --- таргет-сетями для них), и минимизируем стандартное MSE:
$$\E_\T (\vec{Q}(s, \vec{a}) - y(\T))^2 \to \min_{\vec{Q}}$$

Процесс необходимо проводить на центральном сервере, поскольку для вычисления таргета необходимы текущие стратегии всех агентов. Вообще, от этого ограничения можно избавиться, если агентам в их наблюдениях доступны действия всех остальных агентов. Тогда мы сможем перейти к децентрализованному обучению следующим образом: $i$-ый агент может локально хранить и обучать лишь интересующую его компоненту $Q^i(s, \vec{a})$. При построении таргета нужны стратегии опонентов; их $i$-ый агент будет \emph{моделировать}. Для моделирования собираются наблюдения $(s, a^{-i})$, то есть какие действия предприняли остальные агенты в состоянии $s$, и дальше агент учится предсказывать действия других игроков по этой обучающей выборке в supervised-режиме:
$$\E_{s, a^{-i}} \log \mu(a^{-i} \mid s) \to \max_{\mu},$$
обычно с добавлением энтропийного регуляризатора. Имея такое приближение стратегий оппонентов на руках, таргет для обучения критика рассчитывается по формуле
$$y^i(\T) \coloneqq r^i + Q^i(s', a'^i, a'^{-i}),$$
где $a'^i \HM\sim \pi^i(a'^i \mid s')$, а $a'^{-i} \sim \mu(a'^{-i} \mid s')$.

Как использовать централизованную оценочную функцию для обучения актёров? Для обучения $i$-го актёра с параметрами $\theta^i$ можно применить формулу policy gradient:
\begin{equation}\label{MAPG}
\nabla_{\theta^i} = \E_{s} \E_{a^i \sim \pi^i(a^i \mid s)} \nabla_{\theta^i} \log \pi^i(a_i \HM\mid s) Q^i(s, a^i, a^{-i})
\end{equation}
Действительно, для формулы градиентов по $\theta^i$ остальные стратегии можно считать фиксированными; поэтому стандартная формула применима. В ней состояния $s$ приходят из опыта взаимодействия агентов со средой, а действия других агентов $a^{-i}$ должны быть порождены этими самыми стратегиями прочих агентов. 

Авторы MADDPG предлагают перейти здесь от policy gradient-схемы к использованию формулы аля DDPG, и забить на частоты посещения состояний. Тогда в формуле \eqref{MAPG} можно брать $s$, $a^{-i}$ --- из приближения стратегий других агентов. По аналогии с DDPG также можно обучать детерминированную стратегию $\pi^i(s)$ с параметрами $\theta^i$, оптимизируя
$$\E_{s} \E_{a^i} Q^i(s, \pi^i(s), a^{-i}) \to \max_{\theta^i},$$
где $s$ берутся из реплей буфера, $a^{-i}$ моделируются или, в случае CTDE-обучения, берутся из честных стратегий $\pi^j(s), j \HM\ne i$; в последнем случае мат.ожидание по $a^{-i}$ вырождается.

\subsection{Системы коммуникации (DIAL)}

Пожалуй, главной особенностью взаимодействия агентов вроде людей в природе является такое явление, как язык. Давайте попробуем в парадигме RL промоделировать, как агенты могут передавать друг другу сообщения.

\begin{definition}
Скажем, что у агента $i$ есть \emph{канал связи} (communication channel), если на каждом шаге агент помимо действия $a^i$ выбирает некоторое \emph{сообщение} (message) $m^i$, которое не влияет на функцию переходов и функции награды, и которое все агенты получают вместе со своими наблюдениями на следующем шаге.
\end{definition}

Конечно, можно обобщить это понятие и сказать, что канал связи есть между конкретными агентами $i, j$, и этот канал может быть односторонним, двухсторонним, и так далее, но нам это сейчас не принципиально. Мы можем использовать эту идею в произвольных мульти-агентных средах, в том числе обучая агентов децентрализованно, рассматривая $m^i$ как часть пространства действий агентов.

\begin{remark}
В таком подходе удобно выбрать дискретное пространство сообщений, а при моделировании Q-функции вместо того, чтобы выдавать оценку для каждой пары действий $(a^i, m^i)$ отдельно выдаются оценки для $a^i$ и отдельно для $\mu^i$; дальше агент $\eps$-жадно выбирает действие для среды и отдельно $\eps$-жадно выбирает сообщение для отправки. Это небольшая декомпозиция упрощает пространство действий.
\end{remark}

Рассмотрим чуть более хитрое применение канала связи под названием Differentiable Inter-Agent Learning (DIAL), применимое для кооперативных задач (с единой функцией награды) в парадигме CTDE. Здесь естественно выбрать непрерывное пространство сообщений. На центральном сервере во время обучения просто заметим, что любые модели агента $j$, на шаге $t$ минимизирующие свои стандартные функции потерь, дифференцируемы по входным наблюдениям и в том числе по полученным от других агентов сообщениям $\mu^i_{t-1}$. Эту информацию о градиенте можно использовать для обучения $i$-го агента выдавать хорошие сообщения! Концептуально, схема для двух агентов в условиях частичной наблюдаемости выглядит примерно так:

\begin{center}
    \includegraphics[width=0.65\textwidth]{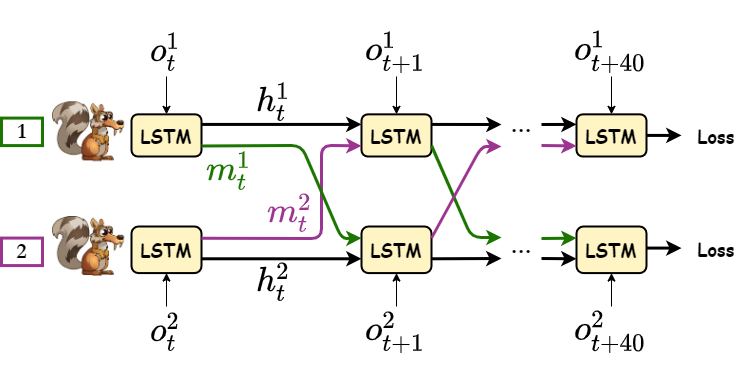}
\end{center}

На рисунке $o_t^i$ --- наблюдение, поступающее на вход $i$-му агенту на шаге $t$. Оно обрабатывается рекуррентной сетью, которое выдаёт, помимо оценочных функций и вероятностей действий, скрытое состояние $h^i_{t}$ для передачи самому себе на следующий шаг и сообщение $m^i_t$ для передачи другим агентам на следующий шаг (как видно на схеме, между этими двумя сущностями практически нет никакой разницы). Далее это сообщение поступает на вход моделям всех других агентов на шаге $t + 1$. В конце вычислений вычисляется некоторая функция потерь в зависимости от используемого алгоритма (например, MSE для оценочных функций или суррогатная функция потерь для обучения стратегии), и этот градиент проходит не только назад по времени, но и по каналам связи в модели других агентов: как видно, вся эта схема end-to-end дифференцируема.


\newpage

\appendix

\chapter{Приложение}

\section{Натуральный градиент}\label{appendix:ng}

\subsection{Проблема параметризации}

Стандартная градиентная оптимизация страдает от зависимости от параметризации. Допустим, мы градиентно оптимизируем
$$f(x) \to \min_x ,$$
и решили сменить параметризацию на $y = y(x)$ (допустим, даже, биективным преобразованием); задача
$$f(y) \to \min_y ,$$
казалось бы, эквивалентна, но траектории градиентного спуска не только будут кардинально отличаться (при эквивалентной инициализации, $x_0$ для первой задачи и $y_0 = y_0(x_0)$ для второй), но и могут сильно различаться по свойствам (в одной параметризации оптимизация проходит намного успешнее другой).

Для нас обычно проблема закопана ещё чуть глубже. Оптимизируемые нами функционалы обычно имеют такой вид:
\begin{equation}\label{NGmotivation}
f(\phi) \coloneqq f(q(x \mid \phi)) \to \min_\phi
\end{equation}
где $q(x \mid \phi)$ --- некоторое параметрически заданное распределение, и функционал $F$ зависит только непосредственно от самого распределения. В качестве такого функционала обычно выступает или логарифм правдоподобия, или функция потерь для задач машинного обучения, также такой вид имеет вспомогательная функция в эволюционных стратегиях \eqref{es} и, самое главное, так выглядит и наш главный оптимизируемый функционал в обучении с подкреплением \eqref{goal}.

Вспомним, как устроен градиентный спуск. Мы рассматриваем некоторую окрестность точки $\phi_0$ и хотим, основываясь на локальных свойствах функции, выбрать направление для изменения текущих параметров. Для этого функция $f(\phi)$ раскладывается в ряд Тейлора до первого члена с центром в точке $\phi_0$, и это разложение используется как приближённая модель поведения функции:
$$\begin{cases}
f(\phi) \approx f(\phi_0) + \langle \left. \nabla_\phi f(\phi) \right|_{\phi = \phi_0}, \phi - \phi_0 \rangle \to \min\limits_{\phi} \\
\|\phi - \phi_0\|_2^2 \le \alpha
\end{cases}$$
где $\alpha$ --- \ENGLISH{learning rate}, а условие $\|\phi - \phi_0\|_2^2 \le \alpha$ задаёт <<\emph{регион доверия}>> (\ENGLISH{trust region}) к построенному приближению. Понятие региона требует понятие расстояния: мы готовы отойти от текущей точки $\phi_0$ не далее чем на $\alpha$ в смысле обычного Евклидова расстояния. Аналитическое решение задачи даёт стандартную формулу градиентного спуска:
$$\phi - \phi_0 \propto - \left. \nabla_\phi f(\phi) \right|_{\phi = \phi_0}$$

Однако, в случае функционала вида \eqref{NGmotivation} параметризации $q(x \mid \phi)$ могут быть таковы, что небольшие изменения параметров $\phi$ могут радикально сильно поменять само распределение $q(x \mid \phi)$, а значит, и значение $f(q(x \mid \phi))$; и наоборот, для внесения каких-то небольших изменений в $q(x \mid \phi)$ необходимо поменять $\phi$ достаточно сильно в смысле Евклидова расстояния.

\begin{example}
$\N(0, 100)$ похож $\N(1, 100)$, когда $\N(0, 0.1)$ совсем не похоже на $\N(1, 0.1)$; при этом для обоих пар евклидово расстояние между значениями параметров равно единице.
\end{example} 

\subsection{Матрица Фишера}\label{appendix:fishermatrix}

Оптимизация при помощи натурального градиента предлагает использовать другую метрику, которая учтёт структуру нашего функционала:
$$\begin{cases}
f(\phi) \approx f(\phi_0) + \langle \left. \nabla_\phi f(\phi) \right|_{\phi = \phi_0}, \phi - \phi_0 \rangle \to \min\limits_{\phi} \\
\KL (q(x \mid \phi_0) \parallel q(x \mid \phi)) \le \alpha
\end{cases}$$

Как решать такую задачу условной оптимизации? Если $\phi \approx \phi_0$, достаточно аппроксимировать дивергенцию $\KL (q(x \mid \phi_0) \parallel q(x \mid \phi))$ при помощи разложения в ряд Тейлора до второго члена. До второго --- потому что первое ноль.

\begin{proposition}\,
\begin{equation*}
\left. \nabla_{\phi} \KL (q(x \mid \phi_0) \parallel q(x \mid \phi)) \right|_{\phi = \phi_0} = 0
\end{equation*}
\begin{proof}
$\KL$-дивергенция в точке $\phi = \phi_0$ равна 0 как дивергенция между одинаковыми распределениями, следовательно как функция от $\phi$ она достигает в этой точке глобального минимума $\Rightarrow$ градиент равен нулю.
\end{proof}
\end{proposition}

\begin{definition}
Для распределения $q(x \mid \phi)$ \emph{матрицей Фишера} (\ENGLISH{Fisher matrix}) называется
$$F_q(\phi) \coloneqq -\mathbb{E}_{q(x \mid \phi)} \nabla^2_\phi \log q(x \mid \phi)$$
\end{definition}

\begin{theorem}
Матрица Фишера есть гессиан $\KL$-дивергенции:
\begin{equation*}
\left. \nabla^2_{\phi} \KL (q(x \mid \phi_0) \parallel q(x \mid \phi)) \right|_{\phi = \phi_0} = F_q(\phi_0) 
\end{equation*}
\beginproof
\begin{equation*}
\left. \nabla^2_{\phi} \KL (q(x \mid \phi_0) \parallel q(x \mid \phi)) \right|_{\phi = \phi_0} = \left. \nabla^2_{\phi} \left[ \const(\phi) - \mathbb{E}_{q(x \mid \phi_0)}  \log q(x \mid \phi) \right] \right|_{\phi = \phi_0} = F_q(\phi_0) \tagqed
\end{equation*}
\end{theorem}

Прежде чем двинуться дальше, остановимся на паре важных для нас свойств матрицы Фишера.

\begin{theorem}[Эквивалентное определение матрицы Фишера]\,
\begin{equation}\label{Fishereqiv}
F_q(\phi) \coloneqq \mathbb{E}_{q(x \mid \phi)} \nabla_\phi \log q(x \mid \phi) (\nabla_\phi \log q(x \mid \phi) )^T
\end{equation}
\beginproof
\begin{align*}
F_q(\phi) &= -\mathbb{E}_{q(x \mid \phi)} \nabla_\phi \nabla_\phi \log q(x \mid \phi) = \\
= \{ \text{градиент логарифма} \} 
&= -\mathbb{E}_{q(x \mid \phi)} \nabla_\phi \frac{\nabla_\phi q(x \mid \phi)}{q(x \mid \phi)} = \\
= \{ \text{градиент отношения} \}
&= -\mathbb{E}_{q(x \mid \phi)} \frac{\nabla^2_\phi q(x \mid \phi)}{q(x \mid \phi)} + \mathbb{E}_{q(x \mid \phi)} \frac{\nabla_\phi q(x \mid \phi)\nabla_\phi q(x \mid \phi)^T}{q(x \mid \phi)^2} = (*)
\end{align*}

Заметим, что первое слагаемое равно нулю. Действительно:
$$\mathbb{E}_{q(x \mid \phi)} \frac{\nabla^2_\phi q(x \mid \phi)}{q(x \mid \phi)} = \int_{x} \nabla^2_\phi q(x \mid \phi) \diff x = \nabla^2_\phi \int_{x} q(x \mid \phi) \diff x = \nabla^2_\phi 1 = 0$$

В оставшемся втором слагаемом перегруппируем множители (заметим, что в знаменатели дроби стоит скаляр):
\begin{align*}
(*) &= \mathbb{E}_{q(x \mid \phi)} \frac{\nabla_\phi q(x \mid \phi)}{q(x \mid \phi)} \left( \frac{\nabla_\phi q(x \mid \phi)}{q(x \mid \phi)} \right)^T = \\
= \{ \text{градиент логарифма} \} 
&= \mathbb{E}_{q(x \mid \phi)} \nabla_\phi \log q(x \mid \phi) (\nabla_\phi \log q(x \mid \phi) )^T \tagqed
\end{align*}
\end{theorem}

\begin{proposition}
Любая матрица Фишера симметрична и положительно определена.
\end{proposition}

\begin{theoremBox}{Репараметризация матрицы Фишера}
Пусть одно распределение параметризовано двумя способами, а то есть $q_\phi(x \mid \phi) \HM\equiv q_\nu(x \mid \nu)$, где $\nu = \nu(\phi)$ --- преобразование с якобианом\footnote[*]{будем считать, что якобиан имеет размер $h_\phi \times h_\nu$, где $h_\nu, h_\phi$ --- размерности $\nu$ и $\phi$ соответственно.} $J$. Тогда:
\begin{equation}\label{Fisherreparam}
F_q(\phi) = JF_q(\nu)J^T
\end{equation}
\beginproof
\begin{align*}
F_q(\phi) &= \E_{q_\phi(x \mid \phi)} (\nabla_\phi \log q(x \mid \phi)) (\nabla_\phi \log q(x \mid \phi) )^T = \\
= \{ q_\phi(x \mid \phi) \equiv q_\nu(x \mid \nu) \} 
&= \E_{q_\nu(x \mid \nu)} (\nabla_\phi \log q(x \mid \nu)) (\nabla_\phi \log q(x \mid \nu) )^T = \\
= \{ \text{chain rule} \}
&= \E_{q_\phi(x \mid \phi)} J \nabla_\nu \log q(x \mid \nu) (\nabla_\nu \log q(x \mid \nu) )^T J^T = \\
&= JF_q(\nu)J^T \tagqed
\end{align*}
\end{theoremBox}

\subsection{Натуральный градиент}

Итак, приближённая задача выглядит следующим образом:
$$\begin{cases}
f(\phi) \approx f(\phi_0) + \langle \left. \nabla_\phi f(\phi) \right|_{\phi = \phi_0}, \phi - \phi_0 \rangle \to \min\limits_{\phi} \\
(\phi - \phi_0)^T F_q(\phi_0) (\phi - \phi_0) \le \alpha
\end{cases}$$
По сути, матрица Фишера задаёт нам приближённо метрику\footnote{А точнее, \emph{Римановскую метрику}. В Евклидовом пространстве общей формой скалярного произведения является $\langle x, y \rangle \HM\coloneqq x^TGy$, где $G$ --- некоторая фиксированная положительно определённая матрица, и индуцированной метрикой соответственно является $d(x, y)^2 \coloneqq  (y - x)^T G (y - x)$. В Римановском пространстве $G$, называемая также \emph{метрическим тензором} (metric tensor), зависит от $x$, и относительное расстояние между точками зависит от положения в пространстве. Римановские пространства используются для описания расстояний между точками на поверхностях, а метрические тензоры для них обладают рядом приятных свойств, которыми, в частности, обладает и матрица Фишера.}, индуцированную $\KL$-дивергенцией. Такая задача также решается аналитично и получается приятный ответ:
\begin{equation}\label{naturalgradient}
\phi - \phi_0 \propto -F_q(\phi_0)^{-1} \left. \nabla_\phi f(\phi) \right|_{\phi = \phi_0}
\end{equation}

\begin{definition}
\emph{Натуральным градиентом} (natural gradient) функции $f(\phi) \coloneqq f(q(x \mid \phi))$ называется
$$\tilde{\nabla}_\phi f(\phi) \coloneqq F_q(\phi)^{-1} \nabla_\phi f(\phi)$$
\end{definition}

Итак, натуральный градиент есть скорректированный матрицей Фишера обычный градиент. Он очень напоминает методы оптимизации второго порядка, так как происходит учёт вида оптимизируемой функции, в том числе масштабирование по осям.

\begin{theorem}[Инвариантность натурального градиента относительно параметризации]
Пусть одно распределение параметризовано двумя способами, а то есть $q_\phi(x \mid \phi) \HM\equiv q_\nu(x \mid \nu)$, где $\nu = \nu(\phi)$ --- преобразование с якобианом $J$.

Тогда натуральные градиенты указывают в одном и том же направлении:
$$\tilde{\nabla}_\phi f(\nu) = J^T \tilde{\nabla}_\phi f(\phi)$$
\beginproof
\begin{align*}
J^T \tilde{\nabla}_\phi f(\phi) &= J^T F_q(\phi)^{-1} \nabla_\phi f(\phi) = \\
= \{ \text{репараметризация матрицы Фишера (\ref{Fisherreparam})} \} 
&= J^T \left( JF_q(\nu)J^T \right) ^{-1} \nabla_\phi f(\phi) \\
= \{ \text{свойства обратной матрицы} \}
&= J^T J^{-T}F_q(\nu)^{-1}J^{-1} \nabla_\phi f(\phi) \\
= \{ \text{chain rule} \}
&= F_q(\nu)^{-1}J^{-1} J \nabla_\nu f(\nu) = \\
&= F_q(\nu)^{-1} \nabla_\nu f(\nu) = \\
&= \tilde{\nabla}_\nu f(\nu) \tagqed
\end{align*}
\end{theorem}



\section{Обоснование формул CMA-ES}\label{appendix:cmaes}

В данном разделе мы вычислим формулу натурального градиента для оптимизации вспомогательного функционала
\begin{equation}\label{appendix_es}
g(\lambda) \coloneqq \E_{\theta \sim q(\theta \mid \lambda)} J(\theta) \to \max_{\lambda},
\end{equation}
для эволюционной стратегии $q(\theta \mid \lambda) \coloneqq \N(\mu, \Sigma)$ с параметрами $\lambda \coloneqq (\mu, \Sigma)$.

\subsection{Вычисление градиента}\label{cmaeslikelihoodgrads}

Сначала нам придётся напрячься и продифференцировать логарифм правдоподобия по параметрам $\mu$ и $\Sigma$:
$$\log q(\theta \mid \mu, \Sigma) = \const(\mu, \Sigma) - \frac{1}{2} \log \det \Sigma - \frac{1}{2}(\theta - \mu)^T\Sigma^{-1}(\theta - \mu)$$

Поскольку нам придётся дифференцировать функционал по матрице, вычислять градиенты удобно в терминах дифференциала. Будем использовать следующую нотацию: будем обозначать за $D_x f(x)[h]$ дифференциал функции $f(x)$ по переменной $x$ с приращением $h$. Дифференциал имеет ту же размерность, что и выход функции $f$, что и делает его использование таким удобным. В итоге мы должны получить линейную часть приращения $f(x)$; так, в итоге вычислений мы получим представление дифференциала в виде $D_x f(x)[h] = \langle \nabla_x f(x), h \rangle$ и сможем <<вытащить>> градиент.

\begin{proposition}\,
$$\nabla_\mu \log q(\theta \mid \mu, \Sigma) = \Sigma^{-1}(\theta - \mu)$$
\begin{proof}
$$D_\mu \log q(\theta \mid \mu, \Sigma)[h] = \frac{1}{2}h^T\Sigma^{-1}(\theta - \mu) + \frac{1}{2}(\theta - \mu)^T\Sigma^{-1}h = \langle \Sigma^{-1}(\theta - \mu), h \rangle$$

Отсюда получаем градиент как вектор, скалярно перемножаемый на приращение $h$: $\Sigma^{-1}(\theta - \mu)$.
\end{proof}
\end{proposition}

Для $\Sigma$ нам понадобится пара табличных формул векторного дифференцирования (здесь $\Tr$ --- оператор взятия \href{https://ru.wikipedia.org/wiki/\%D0\%A1\%D0\%BB\%D0\%B5\%D0\%B4_\%D0\%BC\%D0\%B0\%D1\%82\%D1\%80\%D0\%B8\%D1\%86\%D1\%8B}{следа матрицы}):
\begin{equation}\label{logdetdiff}
D_\Sigma \log \det \Sigma [H] = \Tr(\Sigma^{-1}H)
\end{equation}
\begin{equation}\label{inversediff}
D_\Sigma \Sigma^{-1} [H] = -\Sigma^{-1}H\Sigma^{-1}
\end{equation}

\begin{proposition}\, 
$$\nabla_\Sigma \log q(\theta \mid \mu, \Sigma) = -\frac{1}{2}\Sigma^{-1} + \frac{1}{2}\Sigma^{-1}(\theta - \mu)(\theta - \mu)^T\Sigma^{-1}$$
\begin{proof}
\begin{align*}
D_\Sigma \log q(\theta \mid \mu, \Sigma)[H] &= - \frac{1}{2} D_\Sigma \log \det \Sigma [H] - \frac{1}{2}(\theta - \mu)^T D_\Sigma \Sigma^{-1} [H] (\theta - \mu)\\
= \{ \text{подставляем формулы \eqref{logdetdiff} и \eqref{inversediff}} \}
&= - \frac{1}{2} \Tr(\Sigma^{-1}H) + \frac{1}{2}(\theta - \mu)^T \Sigma^{-1}H\Sigma^{-1} (\theta - \mu)\\
= \{ \text{фокус: если $v$ --- скаляр, $v = \Tr(v)$} \}
&= - \frac{1}{2} \Tr(\Sigma^{-1}H) + \frac{1}{2}\Tr \left( (\theta - \mu)^T \Sigma^{-1}H\Sigma^{-1} (\theta - \mu) \right) = \\
= \{ \text{тождество $\Tr(ABC) = \Tr(BCA)$} \}
&= - \frac{1}{2} \Tr(\Sigma^{-1}H) + \frac{1}{2}\Tr \left( \Sigma^{-1}(\theta - \mu)(\theta - \mu)^T\Sigma^{-1}H \right)
\end{align*}

Дифференциал --- это скалярное произведение градиента на приращение $H$, которое в пространстве матриц задано оператором $\Tr$.
\end{proof}
\end{proposition}

Из этих формул можно увидеть, почему градиентный подъём для оптимизации \eqref{appendix_es} по $\mu, \Sigma$ не так хорош. Допустим, матрица ковариации адаптировалась так, что вдоль оси $\theta_0$ разброс большой. Пусть где-то справа нашлись особи с огромным $J(\theta)$; тогда взвешенное среднее
$$\frac{1}{N}\sum_{\theta \in \Pop} J(\theta)\theta$$
будет указывать примерно в центр этого скопления, будет говорить сильно увеличить компоненту $\theta_0$. Однако, формула градиента говорит сделать поправку на матрицу ковариации: мол, мы генерировали вдоль $\theta_0$ с большим разбросом, и поэтому вдоль этой оси градиент нужно пропорционально сбить. В итоге, к центру скопления делается куда меньший шаг, чем по идее должен был бы по итогам генерации целого поколения особей. Примерно тоже самое наблюдается с матрицей ковариации: в формуле также делается поправка на текущее значение матрицы ковариации (домножение на $\Sigma^{-1}$ с двух сторон) вместо движения к эмпирической (взвешанной на $\hat{J}(\theta)$) матрице.

\subsection{Произведение Кронекера}

Нам понадобится работать с матрицей Фишера, которая имеет размерность <<количество параметров на количество параметров>>. Иными словами, нам понадобится сравнивать попарно между собой все элементы матрицы $\Sigma$, а это значит, нам придётся чуть-чуть залезть в \href{https://ru.wikipedia.org/wiki/\%D0\%9F\%D1\%80\%D0\%BE\%D0\%B8\%D0\%B7\%D0\%B2\%D0\%B5\%D0\%B4\%D0\%B5\%D0\%BD\%D0\%B8\%D0\%B5_\%D0\%9A\%D1\%80\%D0\%BE\%D0\%BD\%D0\%B5\%D0\%BA\%D0\%B5\%D1\%80\%D0\%B0}{алгебру Кронекера}. Для этого мы введём операцию \emph{векторизации матрицы} $\vect$, вытягивающей все элементы матрицы в вектор. 

\begin{proposition}
Векторизация --- линейная операция; поэтому, в частности:
\begin{equation}\label{exp_in_vec}
    \E_\theta \vect (f(\theta)) = \vect(\E_\theta f(\theta))
\end{equation}
\begin{equation}\label{nab_in_vec}
    \nabla_{\vect(\Sigma)} f(\Sigma) = \vect(\nabla_\Sigma f(\Sigma))
\end{equation}
\begin{proof}
Можно проверить непосредственно: мы просто переписываем все элементы матрицы в другой форме, сложению и умножению на скаляры всё равно.
\end{proof}
\end{proposition}

\begin{definition}
Произведением Кронекера двух матриц $A \in \R^{n \times m}, B \in \R^{p \times q}$ называется
$$A \otimes B \coloneqq 
\begin{pmatrix}
a_{11}B & \dots & a_{1m}B \\
\vdots & \ddots & \vdots \\
a_{n1}B & \dots & a_{nm}B
\end{pmatrix}
\in \R^{np \times mq},$$
где $a_{ij}$ --- элементы матрицы $A$.
\end{definition}

\begin{example}
\begin{align*} 
\begin{pmatrix}
3 & 0 \\
1 & -2
\end{pmatrix}
\otimes
\begin{pmatrix}
4 & 0 & -1 \\
-4 & 2 & 5 
\end{pmatrix} 
&=
\begin{pmatrix}
3 \begin{pmatrix}
4 & 0 & -1 \\
-4 & 2 & 5 
\end{pmatrix} & 0 \begin{pmatrix}
4 & 0 & -1 \\
-4 & 2 & 5 
\end{pmatrix} \\
1 \begin{pmatrix}
4 & 0 & -1 \\
-4 & 2 & 5 
\end{pmatrix} & -2 \begin{pmatrix}
4 & 0 & -1 \\
-4 & 2 & 5 
\end{pmatrix}
\end{pmatrix} = \\
&=
\begin{pmatrix}
12 & 0 & -3 & 0 & 0 & 0 \\
-12 & 6 & 15 & 0 & 0 & 0 \\
4 & 0 & -1 & -8 & 0 & 2 \\
-4 & 2 & 5 & 8 & -4 & -10
\end{pmatrix}
\end{align*}
\end{example}

\begin{theorem}Для матриц $A, B, C, D$ с размерностями, для которых существует произведения $AC$ и $BD$, верно:
\begin{equation}\label{kroneckermixproducts}
(A \otimes B)(C \otimes D) = AC \otimes BD
\end{equation}
\begin{proof}
Проверяется непосредственно: пусть $a_{ij}$ --- элементы матрицы $A \in \R^{n \times m}$, $c_{ij}$ --- элементы матрицы $C \in \R^{m \times q}$, тогда:
\begin{align*}(A \otimes B)(C \otimes D) &=
\begin{pmatrix}
a_{11}B & \dots & a_{1m}B \\
\vdots & \ddots & \vdots \\
a_{n1}B & \dots & a_{nm}B
\end{pmatrix}
\begin{pmatrix}
c_{11}D & \dots & c_{1q}D \\
\vdots & \ddots & \vdots \\
c_{m1}D & \dots & c_{mq}D
\end{pmatrix} \\
&=
\begin{pmatrix}
\sum_{k = 1}^m a_{1k}Bc_{k1}D & \dots & \sum_{k = 1}^m a_{1k}Bc_{kq}D \\
\vdots & \ddots & \vdots \\
\sum_{k = 1}^m a_{nk}Bc_{k1}D & \dots & \sum_{k = 1}^m a_{nk}Bc_{kq}D
\end{pmatrix}
\end{align*}

Вводя обозначение $e_{ij} \coloneqq \sum_{k = 0}^m a_{ik}c_{kj}$, получаем Кронекерово произведение между матрицей $E$, состоящей из этих элементов, и матрицей $BD$. Осталось заметить, что матрица $E = AC$ по определению.
\end{proof}
\end{theorem}

\begin{proposition} Для обратимых матриц $A$, $B$:
\begin{equation}\label{kroneker_inverse}
(A \otimes B)^{-1} = A^{-1} \otimes B^{-1}
\end{equation}
\beginproof[Пояснение]
\begin{align*}
(A^{-1} \otimes B^{-1})(A \otimes B) = A^{-1}A \otimes B^{-1}B = I \otimes I = I \tagqed 
\end{align*}
\end{proposition}

\begin{theorem}Для матриц $A, B, C$, для которых существует произведение $ABC$, верно:
\begin{equation}\label{kronecker_vect}
\vect (ABC) = (C^T \otimes A)\vect(B)
\end{equation}
\begin{proof} Проверяется непосредственно. Пусть $c_{ij}$ --- элементы матрицы $C$, $b_i$ --- строки матрицы $B$. Тогда $\vect(B) = (b_1, b_2 \dots b_m)^T$, и:
\begin{align*}(C^T \otimes A)\vect(B) &=
\begin{pmatrix}
c_{11}A & \dots & c_{m1}A \\
\vdots & \ddots & \vdots \\
c_{1n}A & \dots & c_{mn}A
\end{pmatrix}
\begin{pmatrix}
b_1^T \\
\vdots\\
b_m^T
\end{pmatrix}
=
\begin{pmatrix}
\sum_{i=1}^{m} c_{i1}Ab_i^T \\
\vdots \\
\sum_{i=1}^{m} c_{in}Ab_i^T \\
\end{pmatrix}
\end{align*}

Осталось заметить, что $\sum_{i=1}^{m} c_{ij}Ab_i^T$ есть транспонированная $j$-ая строчка матрицы $ABC$.
\end{proof}
\end{theorem}

\subsection{Вычисление матрицы Фишера}

Получение матрицы Фишера для нормального распределения представляет собой напряжное техническое упражнение. Будем далее считать, что наш набор параметров есть $\lambda \coloneqq (\mu, \vect(\Sigma))$; тогда матрица Фишера для нас будет матрицей, размера $(h + h^2) \times (h + h^2)$. Представим её в блочно-диагональном виде:
$$F_q(\lambda) =
\begin{pmatrix}
F_{\mu \mu} & F_{\mu \vect{\Sigma}} \\
F_{\mu \vect{\Sigma}}^T & F_{\vect{\Sigma} \vect{\Sigma}}
\end{pmatrix}
$$
где $F_{\mu \mu} \in \R^{h \times h}$, $F_{\vect{\Sigma} \vect{\Sigma}} \in \R^{h^2 \times h^2}$. Мы воспользовались симметричностью матрицы Фишера (что было видно из её эквивалентного определения \eqref{Fishereqiv}).

Будем искать эти блоки по одному. Для удобства введём обозначение
$$S \coloneqq (\theta - \mu)(\theta - \mu)^T$$
и продублируем формулы градиентов логарифма правдоподобия, полученные в разделе \ref{cmaeslikelihoodgrads}:
\begin{equation}\label{cmaes_mu_grad}
\nabla_\mu \log q(\theta \mid \mu, \Sigma) = \Sigma^{-1}(\theta - \mu)
\end{equation}
\begin{equation}\label{cmaes_sigma_grad}
\nabla_\Sigma \log q(\theta \mid \mu, \Sigma) = -\frac{1}{2}\Sigma^{-1} + \frac{1}{2}\Sigma^{-1}S\Sigma^{-1}\end{equation}

Заметим, что по свойствам нормального распределения:
$$\E_{\theta} \theta = \mu \qquad \E_{\theta} S = \Sigma$$

\begin{theorem}
Матрица Фишера для нормального распределения имеет вид:
$$
F_q(\lambda) =
\begin{pmatrix}
\Sigma^{-1} & 0 \\
0 & \frac{1}{2}\left(\Sigma^{-1} \otimes \Sigma^{-1} \right)
\end{pmatrix}
$$
\beginproof[Доказательство для $F_{\mu\mu}$]
\begin{equation*}
F_{\mu\mu} = -\E_{\theta} \nabla^2_\mu \log q(\theta \mid \mu, \Sigma) = \{ \text{\eqref{cmaes_mu_grad}} \} = -\E_{\theta} \nabla_\mu \Sigma^{-1}(\theta - \mu) = \E_{\theta} \Sigma^{-1} = \Sigma^{-1} \tagqed
\end{equation*}

\begin{proof}[Доказательство для $F_{\mu \vect{\Sigma}}$]
Сначала посчитаем вторую производную.
$$D_\Sigma \nabla_\mu \log q(\theta \mid \mu, \Sigma) [H]= \{ \text{\eqref{cmaes_mu_grad}} \} = -D_\Sigma \Sigma^{-1} (\theta - \mu) [H] = \{ \text{\eqref{inversediff}} \} = \Sigma^{-1}H\Sigma^{-1}(\theta - \mu)$$

Тут дальше есть проблема: нас, вообще говоря, интересует градиент. Однако производная вектора $\nabla_\mu \log q(\theta \mid \mu, \Sigma)$ по матрице это трёхмерный тензор, и надо танцевать с векторизацией. Мы схитрим: давайте посмотрим на мат.ожидание дифференциала по $\theta$:
$$-\E_\theta \Sigma^{-1}H\Sigma^{-1}(\theta - \mu) = -\Sigma^{-1}H\Sigma^{-1} \E_\theta(\theta - \mu) = 0$$
Как следствие, это нулевой тензор, и матрица Фишера тоже ноль.
\end{proof}

\beginproof[Доказательство для $F_{\vect{\Sigma} \mu}$ (Sanity check)]
Проверим, что и симметричный блок тоже ноль (формально это уже доказано в силу симметрии). Для этого нужно дифференцировать по $\mu$ первую производную по $\Sigma$  \eqref{cmaes_sigma_grad}. Сразу же будем смотреть на мат.ожидание этого выражения по $\theta$:
$$-\E_\theta D_\mu \nabla_\Sigma \log q(\theta \mid \mu, \Sigma) [h] = \frac{1}{2}\Sigma^{-1} \E_\theta D_\mu S [h] \Sigma^{-1}$$

При этом $\E_\theta D_\mu S [h] = 0$, поскольку:
\begin{equation*}
\E_\theta D_\mu S [h] = -\E_\theta h(\theta - \mu)^T - \E_\theta (\theta - \mu)h^T = 0 \tagqed
\end{equation*}

\begin{proof}[Доказательство для $F_{\vect{\Sigma} \vect{\Sigma}}$]
\begin{align*}
&-\E_\theta D_{\Sigma} \nabla_{\vect{\Sigma}} \log q(\theta \mid \mu, \Sigma) [H] = \\
= &-\E_\theta D_{\Sigma} \vect \left( \nabla_\Sigma \log q(\theta \mid \mu, \Sigma) \right) [H] = \\
= & \{ \text{подставляем \eqref{cmaes_sigma_grad}} \} = \\
= &-\E_\theta D_{\Sigma} \vect \left( -\frac{1}{2}\Sigma^{-1} + \frac{1}{2}\Sigma^{-1}S\Sigma^{-1} \right) [H] = \\
= & \{ \text{вносим минус, $D_{\Sigma}$ и $\E_{\theta}$} \} = \\
= &\vect \left( \frac{1}{2}D_{\Sigma}\Sigma^{-1}[H] - \frac{1}{2} \E_\theta D_{\Sigma} \left( \Sigma^{-1}S\Sigma^{-1} \right) [H] \right) = \\
= & \{ \text{дифф. билинейной функции} \} = \\
= &\vect \left( \frac{1}{2}D_{\Sigma}\Sigma^{-1}[H] - \frac{1}{2}D_\Sigma \Sigma^{-1} [H] \E_\theta S \Sigma^{-1} - \frac{1}{2}\Sigma^{-1}\E_\theta S D_\Sigma \Sigma^{-1}[H] \right) = \\
= & \{ \text{дифф. обратной матрицы \eqref{inversediff}} \} = \\
= &\vect \left( -\frac{1}{2}\Sigma^{-1}H\Sigma^{-1} + \frac{1}{2}\Sigma^{-1}H\Sigma^{-1}\E_\theta S \Sigma^{-1} + \frac{1}{2}\Sigma^{-1}\E_\theta S \Sigma^{-1}H\Sigma^{-1} \right) = \\
= \{ \E_\theta S = \Sigma \}
= & \vect \left( -\frac{1}{2}\Sigma^{-1}H\Sigma^{-1} + \frac{1}{2}\Sigma^{-1}H\Sigma^{-1} + \frac{1}{2} \Sigma^{-1}H\Sigma^{-1} \right) = \\
= & \frac{1}{2} \vect \left( \Sigma^{-1}H\Sigma^{-1} \right) = \\
= \{ \text{свойство \eqref{kronecker_vect}} \}
= & \frac{1}{2} (\Sigma^{-1} \otimes \Sigma^{-1}) \vect (H)
\end{align*}

Приращение вектора, полученное в виде умножения матрицы на векторизацию приращения, эквивалентно тому, что функция под дифференциалом рассматривалась бы как функция от $\vect{\Sigma}$:
$$\E_\theta D_{\vect{\Sigma}} \nabla_{\vect{\Sigma}} \log q(\theta \mid \mu, \Sigma) [\vect{H}] = \frac{1}{2} (\Sigma^{-1} \otimes \Sigma^{-1}) \vect (H)$$

Отсюда $F_{\vect{\Sigma}\vect{\Sigma}} = \frac{1}{2}\left(\Sigma^{-1} \otimes \Sigma^{-1} \right).$
\end{proof}
\end{theorem}

\begin{proposition}
Обратная матрица Фишера для нормального распределения имеет вид:
\begin{equation}\label{inversenormalFisher}
F_q^{-1}(\lambda) =
\begin{pmatrix}
\Sigma & 0 \\
0 & 2\left(\Sigma \otimes \Sigma \right)
\end{pmatrix}
\end{equation}
\begin{proof}[Пояснение]
Для обращения нижнего блока применить формулу \eqref{kroneker_inverse}.
\end{proof}
\end{proposition}

\subsection{Covariance Matrix Adaptation Evolution Strategy (CMA-ES)}

Вспомним общую формулу градиента для эволюционных стратегий:
\begin{equation}\label{appendix_ESgradient}
    \nabla_\lambda g(\lambda) = \E_{\theta \sim q(\theta \mid \lambda)} \nabla_\lambda \log q(\theta \mid \lambda) J(\theta)
\end{equation}

Достаточно домножить её на обратную матрицу Фишера, чтобы получить формулу натурального градиента для обновления $\lambda = (\mu, \Sigma)$.

\begin{theorem}
Натуральный градиент для $\mu$ в \eqref{appendix_es} выглядит так:
$$\tilde{\nabla}_\mu g(\mu, \Sigma) = \frac{1}{N}\sum_{\theta \in \Pop} \hat{J}(\theta) (\theta - \mu)$$
\beginproof
\begin{align*}
\tilde{\nabla}_\mu g(\mu, \Sigma) &= F^{-1}_{\mu \mu} \nabla_\mu g(\mu, \Sigma) = \\
= \{ \text{подставляем \eqref{appendix_ESgradient}} \}
&= \frac{1}{N}\sum_{\theta \in \Pop} \hat{J}(\theta) F^{-1}_{\mu \mu} \nabla_\mu \log q(\theta \mid \mu, \Sigma) = \\
&= \{ \text{подставляем $F^{-1}_{\mu \mu}$ из \eqref{inversenormalFisher} и градиент из \eqref{cmaes_mu_grad}} \} = \\
&= \frac{1}{N}\sum_{\theta \in \Pop} \hat{J}(\theta) \Sigma \Sigma^{-1} (\theta - \mu) = \\ 
&= \frac{1}{N}\sum_{\theta \in \Pop} \hat{J}(\theta) (\theta - \mu)  \tagqed
\end{align*}
\end{theorem}

Заметим, что для OpenAI-ES \ref{openai_es} при константной матрице ковариации натуральный градиент и градиент отличаются на константу (она <<сокращается с learning rate>>); поэтому можно считать, что OpenAI-ES есть тоже алгоритм натуральной эволюционной стратегии.

\begin{theorem}
Натуральный градиент для $\Sigma$ в \eqref{appendix_es} выглядит так:
$$\tilde{\nabla}_{\Sigma} g(\mu, \Sigma) = \frac{1}{N}\sum_{\theta \in \Pop} \hat{J}(\theta) (S - \Sigma)$$
\begin{proof}
\begin{align*}
\tilde{\nabla}_{\vect(\Sigma)} g(\mu, \Sigma) &= F^{-1}_{\vect(\Sigma) \vect(\Sigma)} \nabla_{\vect(\Sigma)} g(\mu, \Sigma) = \\
= \{ \text{подставляем \eqref{appendix_ESgradient}} \}
&= \frac{1}{N}\sum_{\theta \in \Pop} \hat{J}(\theta) F^{-1}_{\vect(\Sigma) \vect(\Sigma)} \vect \left( \nabla_\Sigma \log q(\theta \mid \mu, \Sigma) \right) = \\
&= \{ \text{подставляем $F^{-1}_{\vect(\Sigma) \vect(\Sigma)}$ из \eqref{inversenormalFisher} и градиент из \eqref{cmaes_sigma_grad}} \} = \\
&= \frac{1}{N}\sum_{\theta \in \Pop} \hat{J}(\theta) \left(\Sigma \otimes \Sigma \right) \vect \left( \Sigma^{-1}S\Sigma^{-1} - \Sigma^{-1} \right) = \\
= \{ \text{свойство \eqref{kroneckermixproducts} } \}
&= \frac{1}{N}\sum_{\theta \in \Pop} \hat{J}(\theta) \vect \left( \Sigma\Sigma^{-1}S\Sigma^{-1}\Sigma - \Sigma\Sigma^{-1}\Sigma \right) = \\ 
&= \vect \left( \frac{1}{N}\sum_{\theta \in \Pop} \hat{J}(\theta) (S - \Sigma) \right)
\end{align*}

Отсюда, убирая векторизацию, получаем формулу.
\end{proof}
\end{theorem}
\section{Сходимость Q-learning}\label{appendix:qlearning}

В данном разделе приведено доказательство теоремы \ref{th:TDconvergencetheorem}. Пусть дано MDP с конечным пространством состояний $\St$ и действий $\A$. $Q_0(s, a)$ --- начальное приближение, на $k$-ом шаге $Q_k(s, a)$ строится по правилу
\begin{equation}\label{Qlearningupdateproof}
Q_{k+1}(s, a) = (1 - \alpha_k(s, a))Q_k(s, a) + \alpha_k(s, a) \left( r(s, a) + \gamma \max_{a'} Q_k(s', a')\right)
\end{equation}
где $s' \sim p(s' \mid s, a)$, а $\alpha_k(s, a)$ --- случайные величины, на которые накладывается единственное требование: для всех $s, a$ с вероятностью 1 выполнено:
\begin{equation}\label{TDconvergenceconditionsproof}
\sum_{k \ge 0} \alpha_k(s, a) = +\infty \qquad \sum_{k \ge 0} \alpha_k(s, a)^2 < +\infty
\end{equation}

Докажем, что $Q_n(s, a) \xrightarrow{ n \to +\infty } Q^*(s, a)$ с вероятностью 1.

\subsection{Action Replay Process}

Для доказательства рассмотрим конструкцию под названием Action Replay Process. Давайте запустим алгоритм Q-learning (проведём, чисто теоретически, бесконечное число итераций) и запишем для каждого реализации $\alpha_k(s, a)$ и $s'_k(s, a)$ --- те следующие состояния, которые использовались на $k$-ом шаге для обновления $Q_k(s, a)$. Будем проводить такую аналогию --- запишем эту историю <<на карточках>>: на $k$-ой карточке записано для каждой пары $s, a$ по одному сэмплу $s'_k(s, a) \HM\sim p(s' \HM\mid s, a)$, а также степень $\alpha_k(s, a)$, с которой этот сэмпл был использован для обновления Q-функции; можно считать, что для них в случившейся реализации выполнено \eqref{TDconvergenceconditionsproof}, поскольку это происходит с вероятностью 1. Карточки, считаем, <<сложены в стопку>>, начиная с нулевой карточки, на которой, условно, напишем наше исходное приближение Q-функции $Q_0(s, a)$ для всех $s, a$. Мы сейчас будем эту стопку карт <<просматривать>> от конца к началу.

\newcommand{\ARP}{\mathrm{ARP}}
\begin{definition}
Для данной реализации алгоритма Q-learning Action Replay Process (ARP) будем называть следующее MDP:
\begin{itemize}
    \item Пространством состояний будем считать пару $s, n$, где $s \in \St$, $n$ --- номер карточки.
    \item Пространством действий будем считать $\A$.
    \item Процесс генерации следующего состояния по данному состоянию $s, n$ и действию $a$, который мы будем обозначать как $p_{\ARP}(s', n' \mid s, n, a)$, задаётся следующим образом. Если номер карточки $n$ равен нулю, то <<стопка карт закончилась>>, и следующее состояние --- терминальное. Если $n > 0$, то бросаем нечестную монетку, которая с вероятностью $\alpha_{n - 1}(s, a)$ выдаёт результат <<остановиться>>, а с вероятностью $1 \HM- \alpha_{n - 1}(s, a)$ выдаёт результат <<пойти дальше>>. <<Остановиться>> означает, что мы полагаем итоговым следующим состоянием пару $s'_{n - 1}(s, a), n - 1$. <<Пойти дальше>> означает, что верхняя карточка колоды удаляется --- $n$ всё равно уменьшается на единицу, --- и процедура повторяется уже для тройки $s, n - 1, a$: мы снова подбрасываем монетку и так далее, пока не выпадет <<остановиться>> или колода карт не закончится.
    \item Награда для тройки $s, n, a$ есть $r(s, a)$, если $n > 0$, и $Q_0(s, a)$ иначе.
\end{itemize}
\end{definition}

\needspace{7\baselineskip}
\begin{wrapfigure}{r}{0.35\textwidth}
\vspace{-0.3cm}
\centering
\includegraphics[width=0.35\textwidth]{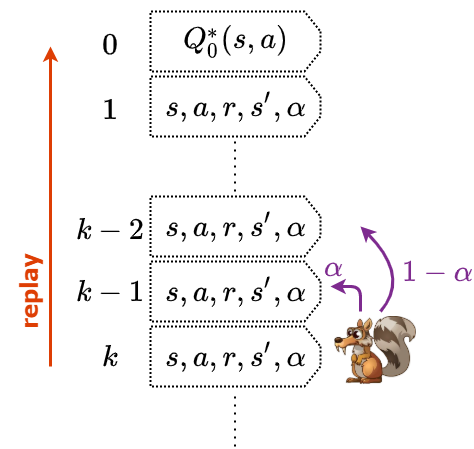}
\end{wrapfigure}

Данное определение внезапно содержит в себе всю основную идею доказательства. Что тут происходит? Давайте посмотрим на первые $n$ шагов результаты работы Q-learning-а. Попробуем построить MDP, которое было бы в некотором смысле <<похожее>> на исходное MDP, но использующее только собранную историю (т.к. доступа к $p(s' \mid s, a)$ у нас нет). Если $n \HM= 0$, и истории нет, то мы считаем, что мы бы получили $Q_0(s, a)$ в качестве награды за всю оставшуюся игру --- такого наше <<исходное>> приближение MDP. Для больших $n$ для данной пары состояние-действие мы в качестве следующего состояния хотели бы засэмплировать $s' \HM\sim p(s' \HM\mid s, a)$, но вместо этого у нас есть лишь коллекция $s'_k(s, a)$. Давайте с вероятностью $\alpha_{n-1}(s, a)$ возьмём в качестве сэмпла $s'_n(s, a)$, с вероятностью $(1 \HM- \alpha_{n-1}(s, a))\alpha_{n-2}(s, a)$ возьмём в качестве сэмпла $s'_{n-1}(s, a)$, и так далее. Мы знаем, что при стремлении $n$ к бесконечности альфы, во-первых, уходят к нулю (это гарантирует сходимость ряда из квадратов альф), а во-вторых, сэмплов будет бесконечно много (бесконечно много альф обязано быть ненулевыми из-за расходимости ряда из альф). Однако после каждого шага в ARP, $n$ уменьшается (как минимум на единицу), и через некоторое число шагов такая игра гарантированно завершится --- вся история из первых $n$ шагов будет <<проиграна>> как на повторе от $n$-го шага до первого. Hence the name.

\subsection{Ключевые свойства ARP}

Принципиально по построению выполнен такой фокус:
\begin{theorem}
В любом ARP оптимальная Q-функция $Q_{\ARP}^*(s, n, a)$ в точности равна
\begin{equation}\label{ARPQ}
Q_{\ARP}^*(s, n, a) = Q_n(s, a)
\end{equation}
\begin{proof}
По индукции. Для $n = 0$ по определению следующее состояние всегда будет терминальным, а наградой для $s, n, a$ является $Q_0(s, a)$. Значит, $Q_{\ARP}^*(s, 0, a) = Q_0(s, a)$.

Пусть выполнено $Q_{\ARP}^*(s, n, a) \HM= Q_n(s, a)$ для любых $s, a$. Тогда для любых $s, a$ величина $Q_{\ARP}^*(s, n \HM+ 1, a)$ равна следующему: с вероятностью $\alpha_n(s, a)$ после выполнения действия $a$ в состоянии $s, n \HM+ 1$ будет получена награда $r(s, a)$, а следующим состоянием будет $s'_n(s, a), n$; тогда дальнейшее оптимальное поведение даст награду $\max\limits_{a'} Q_{\ARP}^*(s'_n(s, a), n, a') \HM= \max\limits_{a'} Q_n(s'_n(s, a), a')$ по предположению индукции. А с вероятностью $1 \HM- \alpha_n(s, a)$ мы не остановимся на $n$ и повторим бросок монетки для $s, n, a$, после которого при дальнейшем оптимальном поведении мы получаем награду $Q^*_{\ARP}(s, n, a) \HM= Q_n(s, a)$. Собирая это вместе, получаем:
$$Q_{\ARP}^*(s, n + 1, a) = \alpha_n(s, a) \left( r(s, a) + \gamma \max_{a'} Q_n(s'_n(s, a), a') \right) + (1 - \alpha_n(s, a)) Q_n(s, a)$$
Справа в точности стоит $Q_{n+1}(s, a)$! Иначе говоря, функция переходов в ARP специально построена так, что оценочные функции удовлетворяют формулам обновления Q-learning-а.
\end{proof}
\end{theorem}

Доказательство сходимости Q-learning-а идейно сводится к тому, что при стремлении $n$ к бесконечности ARP с начальным состоянием $s_0, n$ (и коэф. дисконтирования $\gamma$) становится всё больше похож на исходное, настоящее MDP.

Покажем, что в ARP неявно содержится информация о $p(s' \mid s, a)$. Рассмотрим следующую величину:
$$p_{\ARP}(s' \mid s, n, a) = \sum_{n' = 1}^{n-1} p_{\ARP}(s', n' \mid s, n, a),$$
то есть вероятность после выбора действия $a$ из состояния $s, n$ оказаться в $s'$, если неважно, сколько карточек $n'$ у нас останется после одного шага.

\begin{theorem}\,
\begin{equation}\label{arptranslimit}
p_{\ARP}(s' \mid s, n, a) \xrightarrow{ n \to +\infty } p(s' \mid s, a)
\end{equation}
\begin{proof}
Рассмотрим $p_{\ARP}(s' \HM\mid s, n \HM+ 1, a)$ --- вероятность оказаться в $s'$ после выполнения $a$ из состояния $s, n \HM+ 1$ вне зависимости от $n'$. С вероятностью $\alpha_{n}(s, a)$ исходом будет $s'_n(s, a)$: если оно равно рассматриваемому $s'$, то это даёт $\alpha_n(s, a)$ вероятность для $p_{\ARP}(s' \mid s, n \HM+ 1, a)$, иначе 0; с вероятностью $1 \HM- \alpha_{n}(s, a)$ процесс генерации следующего состояния будет повторён из $s, n, a$, и тогда вероятность оказаться в состоянии $s'$ равна $p_{\ARP}(s' \HM\mid s, n, a)$ по определению. Итого получаем:
$$p_{\ARP}(s' \mid s, n + 1, a) = (1 - \alpha_{n}(s, a))p_{\ARP}(s' \mid s, n, a) + \alpha_{n}(s, a) \mathbb{I}[s'_n(s, a) = s']$$

Мы получили формулу экспоненциального сглаживания \eqref{expsmoothingtheoremexpr} для величин $\mathbb{I}[s'_n(s, a) = s']$. При этом альфы удовлетворяют условиям сходимости \eqref{RobbinsMonro}! Значит, по теореме о сходимости экспоненциального сглаживания \ref{th:expsmoothingconvergence} эти величины в пределе сходятся к мат.ожиданию случайной величины $\mathbb{I}[s'_n(s, a) = s']$. Поскольку $s'_n(s, a)$ для любого $n$ генерировался из $p(s' \HM\mid s, a)$, то 
$$\E \mathbb{I}[s'_n(s, a) = s'] \HM= p(s' \HM\mid s, a),$$
и, следовательно, именно к нему стремится $p_{\ARP}(s' \HM\mid s, n, a)$.
\end{proof}
\end{theorem}

Мы показали, что процесс генерации $s'$ в ARP <<корректно>> имитирует реальную $p(s' \HM\mid s, a)$ при большом числе карточек. Значит, наше ARP с большим числом карточек всё больше похоже на настоящее MDP. Коли так, то и наверняка и оптимальная Q-функция для ARP при стремлении $n$ к бесконечности всё больше похожа на $Q^*(s, a)$ исходного MDP:
$$\lim_{n \to +\infty} Q_{\ARP}^*(s, n, a) = Q^*(s, a)$$
Если это так, то в совокупности с \eqref{ARPQ} мы получаем доказываемое: для любой реализации Q-learning-а 
$$\lim_{n \to +\infty} Q_n(s, a) = \lim_{n \to +\infty} Q_{\ARP}^*(s, n, a) = Q^*(s, a)$$

\subsection{Схожесть ARP и настоящего MDP}

Нам осталось формально показать, почему <<похожесть MDP>> влечёт похожесть оптимальных Q-функций. Техническим препятствием для этого является то, что ARP на каждом шаге <<тратит карточки>> --- мы, идя с конца колоды к началу, теряем какое-то случайное число карточек, а, оставшись с маленьким числом карточек, уже не умеем <<хорошо имитировать>> настоящую функцию переходов.

Следующее утверждение является вспомогательным для основной теоремы: оно говорит, что можно запастись достаточным количеством карточек, чтобы можно было сделать шаг и остаться всё равно со сколь угодно большим числом карточек.

\begin{theorem}
Для любого ARP и любого целого числа карточек $m$ можно выбрать число карточек $n$ так, что для всех $s, a, s'$ вероятность $p_{\ARP}(n' \HM < m \mid s, n, a, s')$ бесконечна мала. 
\begin{proof}
Вероятность оказаться с числом карточек меньше $m$, стартуя из уровня $n$, не меньше чем $\prod_{i=m}^n (1 - \alpha_i(s, a))$ ---  это вероятность в принципе прокрутить историю от $n$-й карточки до $m$-й. Воспользуемся следующим фактом: при любых $\alpha \HM\in [0, 1]$ верно
$$1 - \alpha \le \exp (-\alpha).$$
Подставляя это неравенство в произведение, получаем:
$$\prod_{i=m}^n (1 - \alpha_i(s, a)) \le \prod_{i=m}^n \exp (-\alpha_i(s, a)) = \exp \left( -\sum_{i=m}^n \alpha_i(s, a) \right) \xrightarrow{ n \to +\infty } \exp(-\infty) = 0$$
В последнем переходе мы воспользовались тем, что ряд альф расходится, а значит и любой его хвост, начинающийся с любого конечного $m$, тоже расходится.
\end{proof}
\end{theorem}

Теперь обсудим идею основного доказательства о том, что $Q^*_{ARP}(s, n, a) \xrightarrow{ n \to +\infty } Q^*(s, a)$. Если мы в ARP сидим в состоянии с большим $n$, то у нас есть три причины, по которым $Q^*_{ARP}(s, n, a)$ отличается от $Q^*(s, a)$:
\begin{itemize}
    \item с некоторой маленькой вероятностью мы после нескольких первых шагов окажемся в ARP в состоянии с маленьким значением $n$, где все наши приближения уже не работают. Мы выберем $n$ достаточно большим, чтобы эта вероятность была очень маленькой.
    \item наше приближение функции переходов $p_{\ARP}(s' \mid s, n, a)$ при больших $n$ близка, но не точно совпадает с истинной $p(s' \mid s, a)$. Мы будем пользоваться тем, что как истинная оценочная функция, так и наша оценочная функция ограничены: $Q_{\ARP}^*(s, n, a) \HM< C$, $Q^*(s, a) \HM< C$ для некоторой константы $C$ (это следует из наших стандартных требований регулярности к MDP, которое справедливы и для ARP), поэтому достаточно выбрать $n$ достаточно большим, чтобы сделать эту ошибку маленькой.
    \item наконец, основная ошибка заключается в том, что мы принимаем решения последовательно, а значит, ошибка из-за погрешности функции переходов будет накапливаться. Очень условно, это <<ошибка внутри нашей аппроксимации Q-функции>>. С ней мы будем бороться наиболее хитрым образом: мы рассмотрим ошибку в награде, собираемой на протяжении первых $k$ шагов. Остальная часть этой ошибки будет домножаться на $\gamma^k$, следовательно, можно будет выбрать $k$ достаточно большим, чтобы ошибка была меньше наперёд заданного малого числа $\epsilon > 0$.
\end{itemize}

Введём следующее обозначение: <<максимальная ошибка, если у нас на руках $n$ карточек>>:
$$\nu(n) \coloneqq \max_{s, a} |Q_{\ARP}^*(s, n, a) - Q^*(s, a)|$$
Надо доказать, что она стремится к нулю. Можно считать, что и $\nu(n)$ ограничено в силу ограниченности оценочных функций, чем мы будем пользоваться, когда карточек остаётся мало.

\begin{theorem}
Пусть $n$ и $m$ таковы, что для любых $s, a, s'$ выполнено
$$|p_{\ARP}(s' \mid s, n, a) - p(s' \mid s, a)| < \epsilon$$
$$p_{\ARP}(n' < m \mid s, n, a, s') < \epsilon$$
для данного $\epsilon$. Тогда для некоторой константы $C$ справедливо следующее рекуррентное соотношение:
$$\nu(n) \le \gamma \max_{n' \ge m} \nu(n') + C\epsilon$$
\beginproof
\begin{align*}
\nu(n) &= \max_{s, a} |Q_{\ARP}^*(s, n, a) - Q^*(s, a)| = \\
&= \{ \text{уравнение оптимальности Беллмана \eqref{Q*Q*}; слагаемые с наградой сокращаются} \} = \\
&= \gamma \max_{s, a} |\sum_{s', n'} p_{\ARP}(s', n' \mid s, n, a) \max_{a'} Q_{\ARP}^*(s', n', a') - \sum_{s'} p(s' \mid s, a) \max_{a'} Q^*(s', a')| = \\
&= \{ \text{добавляем и вычитаем $\sum\limits_{s', n'} p_{\ARP}(s', n' \mid s, n, a) \max\limits_{a'} Q^*(s', a')$} \} = \\
&= \gamma \max_{s, a} |\sum\limits_{s', n'} p_{\ARP}(s', n' \mid s, n, a) (\max_{a'} Q_{\ARP}^*(s', n', a') -  \max_{a'} Q^*(s', a')) - \\ &\qquad + \sum_{s'} (p_{\ARP}(s' \mid s, n, a) - p(s' \mid s, a)) \max_{a'} Q^*(s', a') | \le \\
&\le \{ \text{используем свойство максимумов \eqref{diffmax} $\max_x f(x) - \max_x g(x) \le \max_x \left| f(x) - g(x) \right|$} \} \le \\
&\le \gamma \max_{s, a} \bigl[ \sum_{s', n'} p_{\ARP}(s', n' \mid s, n, a) \max_{a'} |Q_{\ARP}^*(s', n', a') -  Q^*(s', a')| + \\ &\qquad + \sum_{s'} |p_{\ARP}(s' \mid s, n, a) - p(s' \mid s, a)| \max_{a'} Q^*(s', a') \bigr] = \\
&= \{ \text{правило произведения} \} = \\
&= \gamma \max_{s, a} \bigl[ \sum_{s'} p_{\ARP}(s' \mid s, n, a) \sum_{n'} p_{\ARP}(n' \mid s, n, a, s') \max_{a'} |Q_{\ARP}^*(s', n', a') -  Q^*(s', a')| + \\ &\qquad + \sum_{s'} |p_{\ARP}(s' \mid s, n, a) - p(s' \mid s, a)| \max_{a'} Q^*(s', a') \bigr] \le \\
&\le \{ \text{определение $\nu(n)$ и свойство $\E f(x) \le \max\limits_x f(x)$} \} \le \\
&\le \gamma \max_{s, a} \bigl[ \sum_{n'} p_{\ARP}(n' \mid s, n, a, s') \nu(n') + \\ &\qquad + \sum_{s'} |p_{\ARP}(s' \mid s, n, a) - p(s' \mid s, a)| \max_{a'} Q^*(s', a') \bigr] \le \\
&\le \{ \text{ошибка аппроксимации функции переходов и ограниченность $Q^*(s, a)$} \} \le \\
&\le \gamma \max_{s, a} \left[ \sum_{n'} p_{\ARP}(n' \mid s, n, a, s') \nu(n') + C\epsilon \right] = \\
&= \{ \text{рассматриваем два случая: $n' < m$ и $n' \ge m$} \} = \\
&\le \gamma \max_{s, a} \left[ \sum_{n' \ge m} p_{\ARP}(n' \mid s, n, a, s') \nu(n') + \sum_{n' < m} p_{\ARP}(n' \mid s, n, a, s') v(n') + C\epsilon \right] \le \\
&\le \{ \text{пользуемся условием теоремы и тем, что $\nu(n)$ ограничено} \} \le \\
&\le \gamma \max_{s, a} \left[ \sum_{n' \ge m} p_{\ARP}(n' \mid s, n, a, s') \nu(n') + C'\epsilon \right] \le \\
&\le \{ \text{свойство $\E f(x) \le \max\limits_x f(x)$} \} \le \\
&\le \gamma \max_{n' \ge m} \nu(n') + C'\epsilon \tagqed
\end{align*}
\end{theorem}

\begin{theorem}
$$\nu(n) \xrightarrow{ n \to +\infty } 0$$
\begin{proof}
Пусть дано $\epsilon > 0$. Покажем, что начиная с некоторого номера, $\nu(n) < 2C\epsilon$. Для этого выберем целое $k$ так, чтобы $\gamma^k < \epsilon$, и применим предыдущую теорему $k$ раз следующим образом. Выберем какое-нибудь $m_0$ и подберём $m_1$ так, чтобы для всех $s, a, s'$ 
$$p_{\ARP}(n' < m_0 \mid s, m_1, a, s') < \frac{\epsilon}{k}.$$

Убедимся, что $m_1$ достаточно большое, что $|p_{\ARP}(s' \mid s, m_1, a) - p(s' \mid s, a)| < \frac{\epsilon}{k}$ (если нет, то заменим на достаточно большое). Затем подберём $m_2$ так, что для всех $s, a, s'$ $$p_{\ARP}(n' < m_1 \mid s, m_2, a, s') < \frac{\epsilon}{k}$$
и так далее вплоть до $m_k$. 

Тогда для всех $n > m_k$:
$$\nu(n) \le \gamma \max_{n' \ge m_{k - 1}} \nu(n') + C\frac{\epsilon}{k}$$
Аналогично, для всех $n > m_{k - 1}$:
$$\nu(n) \le \gamma \max_{n' \ge m_{k - 2}} \nu(n') + C\frac{\epsilon}{k}$$

Последовательно раскручивая эту цепочку $k$ раз получим, что для всех $n > m_k$:
\begin{align*}
\nu(n) &\le \gamma \max_{n' \ge m_{k - 1}} \nu(n') + C\frac{\epsilon}{k} \le \gamma^2 \max_{n' \ge m_{k - 2}} \nu(n') + 2C\frac{\epsilon}{k} \le \dots \le \\
&\le \gamma^k \max_{n' \ge m_0} \nu(n') + kC\frac{\epsilon}{k} \le C\epsilon + C\epsilon = 2C \epsilon
\end{align*}
Вот такие дела.
\end{proof}
\end{theorem}


\newcommand*{\sectionsources}[1]{\textbf{\textcolor{ChadBlue}{\underline{#1}}}}   

\newpage
\begin{center}
\huge \textbf{\textcolor{ChadBlue}{\underline{Материалы}}}

\normalsize
\vspace{0.3cm}
Большая часть материалов взята из основных курсов по обучению с подкреплением:

\vspace{0.3cm}
\href{https://www.youtube.com/playlist?list=PL_iWQOsE6TfURIIhCrlt-wj9ByIVpbfGc}{Курс Сергея Левина}; 

\vspace{0.3cm}
\href{https://github.com/yandexdataschool/Practical_RL}{Курс Practical RL};

\vspace{0.3cm}
\href{https://deeppavlov.ai/rl_course_2020}{Цикл докладов Advanced RL};

\vspace{0.3cm}
\href{https://www.davidsilver.uk/teaching/}{Курс Дэвида Сильвера};

\vspace{0.3cm}
\href{https://www.youtube.com/watch?v=ISk80iLhdfU&list=PLqYmG7hTraZBKeNJ-JE_eyJHZ7XgBoAyb&index=1}{Курс DeepMind};

\vspace{0.3cm}
\href{https://www.udacity.com/course/reinforcement-learning--ud600}{Курс GeorgiaTech};
\end{center}

В \underline{\textbf{главе 2}} большинство материала взято из \href{https://cs.gmu.edu/~sean/book/metaheuristics/Essentials.pdf}{книги} \cite{luke2013essentials}. Хороший обзор эволюционных стратегий можно найти в \href{https://lilianweng.github.io/lil-log/2019/09/05/evolution-strategies.html}{блоге Lil'log} \cite{weng2019ES}. Алгоритм NEAT предложен в \cite{stanley2002evolving}. Алгоритм WANN развил его идею в \cite{gaier2019weight}. Кросс-энтропийный метод как метод оптимизации и метод вычисления вероятности маловероятных событий предложен в \cite{botev2013cross}; его применение к задаче RL обычно связывают с \cite{szita2006learning}, где его применили к тетрису. OpenAI-ES описана в \cite{salimans2017evolution}; упомянутый алгоритм ARS, действующий примерно также, предложен в \cite{mania2018simple}. Идея адаптировать ковариационную матрицу в эволюционных стратегиях восходит корнями к \cite{hansen1996adapting}; полный технический обзор всего набора эвристик, использующихся как алгоритм CMA-ES, можно прочитать \href{https://arxiv.org/pdf/1604.00772.pdf}{здесь} \cite{hansen2016cma}. Доказательство того, что адаптация матрицы ковариации по сути является натуральным градиентным спуском, было независимо получено в \cite{akimoto2010bidirectional} и \cite{glasmachers2010exponential}.

Больше информации по \underline{\textbf{главе 3}} и более подробную библиографию можно получить \href{https://drive.google.com/file/d/1Z4W_-0IaMNpZnhnMkqcDVM_EA79GFJo-/view}{в классической книге Саттона-Барто} \cite{sutton2018reinforcement}; отмечу только некоторые дополнительные ссылки. Лемма RPI была представлена в \cite{kakade2002approximately}. Алгоритм Q-learning, изначально придуманный в \cite{watkins1989learning}, был придуман как эвристика, но позже авторам удалось доказать сходимость в \cite{watkins1992q}; это доказательство через ARP и приведено в приложении. Связь с теорией стохастической аппроксимации, начатой ещё в далёком 1951-ом году статьёй \cite{robbins1951stochastic}, была обнаружена после этого в \cite{tsitsiklis1994asynchronous}, что позволило доказать более сильные утверждения вроде сходимости TD($\lambda$). Наиболее общую форму алгоритмов off-policy оценивания стратегии и оценку Retrace представили в \cite{munos2016safe} уже в 2016-ом году.

\underline{\textbf{Глава 4}} основана на алгоритме DQN \cite{mnih2013playing}, продемонстрировавшем потенциал совмещения глубокого обучения с классической теорией. Идея борьбы с переоценкой при помощи двух аппроксимаций Q-функций была предложена в \cite{hasselt2010double} для табличного алгоритма. Twin (<<Clipped Double>>) оценка предложена была позже (в рамках алгоритма TD3) в \cite{fujimoto2018addressing}. Double DQN предложен в \cite{van2016deep}; Dueling DQN в \cite{wang2015dueling}; Noisy Nets в \cite{fortunato2017noisy}. Приоритизированный реплей был использован в DQN в \cite{schaul2015prioritized}; идею высчитывать приоритеты онлайн и добавлять в буфер уже <<с правильным>> приоритетом реализовали в алгоритме R2D2 \cite{horgan2018distributed}. Эвристика многошагового DQN была описана в составе Rainbow \cite{hessel2018rainbow}. Distributional подход и алгоритм c51 был инициирован в \cite{bellemare2017distributional}; идея перехода к квантильной аппроксимации и алгоритм QR-DQN был описан в \cite{dabney2018distributional}; алгоритм IQN предложен в \cite{dabney2018implicit}. Эквивалентность distributional-алгоритмов с обычным подходом в табличном сеттинге была показана в \cite{lyle2019comparative}.

В \underline{\textbf{главе 5}} метод пробрасывания градиентов через стохастические узлы вычислительного графа REINFORCE и его применение к задаче RL были придуманы в \cite{williams1992simple}. Actor-Critic методы, в которых учится как стратегия, так и оценочная функция, позволяющая обучаться с неполных эпизодов, предложены в \cite{sutton2000policy}. Применение Policy Gradient подхода с нейросетевой аппроксимацией и алгоритм A2C предложен в \cite{mnih2016asynchronous}. Список разных вариаций Policy Gradient алгоритмов можно найти в \href{https://lilianweng.github.io/lil-log/2018/04/08/policy-gradient-algorithms.html}{блоге Lil'log} \cite{weng2018PG}. TRPO был предложен в \cite{schulman2015trust}; обучение с длинных роллаутов привело к использованию GAE оценок, что было предложено в \cite{schulman2015high}. Алгоритм PPO описан в \cite{schulman2017proximal}, однако стандартные реализации, добившиеся высоких результатов на бенчмарках и способствовавшие распространению алгоритма, использовали дополнительные инженерные эвристики; на их существенное влияние на результаты обращено внимание в \cite{engstrom2019implementation}. Позже в \cite{achiam2017constrained} была представлена более точная нижняя оценка на погрешность суррогатной функции, давшая теоретическое обоснование использованию усреднённой по состояниям KL-дивергенции, которая и была представлена в тексте.

Применение идей policy gradient и value-based подхода для обучения нейросетей в задаче непрерывного управления, описанное в \underline{\textbf{главе 6}}, предложено в \cite{lillicrap2015continuous} в алгоритме DDPG. Более стабильная версия этого алгоритма TD3 описана в \cite{fujimoto2018addressing}. Алгоритм Soft Q-learning, использующий теорию Maximum Entropy RL для обучения нейросетей, описан в \cite{haarnoja2017reinforcement}; алгоритм Soft Actor-Critic, ставший практическим алгоритмом для работы с непрерывными действиями в рамках этого сеттинга, предложен в \cite{haarnoja2018soft}. 

За полным погружением в математику, стоящую за многорукими бандитами (\underline{\textbf{глава 7.1}}), можно обратиться к \href{https://tor-lattimore.com/downloads/book/book.pdf}{книге} \cite{lattimore2020bandit}. Нижняя оценка регрета (теорема Лаи-Роббинса) получена в \cite{lai1985asymptotically}. Асимптотическая оптимальность алгоритма UCB показана в \cite{auer2002finite}. Сама задача многоруких бандитов была впервые рассмотрена Томпсоном в \cite{thompson1933likelihood}; он же эвристически предложил сэмплирование Томпсона, которое позже тоже оказалось асимптотически оптимальным \cite{kaufmann2012thompson}. Пример \ref{ex:onlineshortestpath} c нахождением кратчайшего маршрута в графе взят из туториала по сэмплированию Томпсона \cite{russo2017tutorial}. Алгоритм, обучающий байесовские модели динамики и награды для табличных MDP и использующих сэмплирование Томпсона для разрешения дилеммы exploration-exploitation в MDP взят из \cite{osband2013more}.

Несмотря на то, что описанный в \underline{\textbf{главе 7}} model-based подход исследовался в RL всегда, применение моделей мира и концепция сновидений популяризовалась благодаря статье \cite{ha2018world}. Победа алгоритма AlphaGo в го на основе совмещения MCTS и нейросетей в итоге была обобщена сначала в алгоритм AlphaZero \cite{silver2018general}, а затем и на случай неизвестной динамики в алгоритм $\mu$-Zero \cite{schrittwieser2019mastering}. Теория линейно-квадратичных регуляторов восходит ещё к оптимальному управлению; по LQR обычно ссылаются на \cite{bemporad2002explicit}, а по расширению iLQR --- на \cite{li2004iterative}.

Фреймворк Maximum Entropy Inverse RL, рассмотренный в \underline{\textbf{главе 8.1}}, был предложен в \cite{ziebart2008maximum}. Обобщение алгоритма из этой статьи с табличного случая на произвольный в виде процедуры Guided Cost Learning описана в \cite{finn2016guided}. Связь задачи с минимизацией расстояния между occupancy measure была исследована в \cite{ho2016generative}, где был предложен алгоритм GAIL. Расширение GAIL на случай, когда в записях эксперта доступны только наблюдения (алгоритм GAIfO) представлен в \cite{torabi2018generative}. Пример \ref{ex:quadrocopter} со сведением к классификации задачи обучения квадрокоптера полётам по лесу описан в статье \cite{giusti2015machine}. 

Моделирование любопытства и скуки, описанное в \underline{\textbf{главе 8.2}} у агентов было описано ещё Шмидхубером в далёком 1991 году \cite{schmidhuber1991possibility} вместе с проблемой шумных телевизоров. Эвристика RND придумана в \cite{burda2018exploration}; фильтрующие свойства модели обратной динамики, понятие контролируемого состояния и алгоритм ICM описаны в \cite{pathak2017curiosity}. Пример минимизации хаоса \ref{ex:chaosminimization} основан на \cite{berseth2019smirl}.

В \underline{главе \textbf{8.3}} переразметка траекторий произвольными целями, также называемая алгоритмом Intentional-Unintentional, описана в \cite{cabi2017intentional}. Идея HER придумана в \cite{andrychowicz2017hindsight}. Обобщение идеи для переразметки произвольных траекторий и связь этой задачи с обратным обучением с подкреплением одновременно замечена в \cite{eysenbach2020rewriting} и \cite{li2020generalized}. Мета-контроллеры для автоматического подбора гиперпараметров использовались в алгоритме Agent57, обошедшем человека сразу во всех 57 играх Atari \cite{badia2020agent57}.

В иерархическом RL, описанном в \underline{главе \textbf{8.4}}, алгоритм Option-Critic и формулы градиентов для обучения политики терминальности представлены в \cite{bacon2017option}. Концепция феодализма и феодальные сети FuN предложены в \cite{vezhnevets2017feudal}. Обучение похожей иерархической схемы в off-policy при помощи переразметки в виде алгоритма HIRO описано в \cite{nachum2018data}.

Задача обучения в условиях частичной наблюдаемости \underline{главы \textbf{8.5}}  поставлена в \cite{smallwood1973optimal}. Обучение рекуррентных сетей в RL в рамках алгоритма DRQN описано в \cite{hausknecht2015deep}; эвристики разогрева и хранения скрытого состояния предложены в алгоритме R2D2 \cite{horgan2018distributed}. Эпизодичная память NEC предложена в \cite{pritzel2017neural}.

В \underline{главе \textbf{8.6}} Adversarial-атаки на алгоритмы, обученные в режиме self-play, продемонстрированы в \cite{gleave2019adversarial}. Алгоритм QMix для кооперативных игр и идея моделировать смешивающую сеть в классе монотонных функций предложена в \cite{rashid2018qmix}. Общий алгоритм MADDPG и идея моделирования других агентов предложены в \cite{lowe2017multi}. Идея моделировать и учить протоколы коммуникации между агентами описаны в \cite{foerster2016learning}.

\needspace{7\baselineskip}
\begin{wrapfigure}{r}{0.15\textwidth}
\vspace{-0.5cm}
\centering
\includegraphics[width=0.15\textwidth]{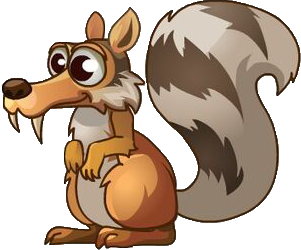}
\vspace{-0.8cm}
\end{wrapfigure}

Изображения взяты из рассмотренных статей, книги \cite{sutton2018reinforcement} и из распространённых сред (OpenAI Gym \cite{brockman2016openai}, Mario \cite{gym-super-mario-bros}, Unity ML Agents \cite{juliani2018unity}). Кастомные изображения для примеров и схем были нарисованы в \href{https://www.draw.io/}{draw.io}; изображение белки на них взято из \href{https://twitter.com/racefornuts/status/690043558208913408}{вот этого твита}; судя по всему, это незаюзанный концепт-арт для некой игры <<Трагедия белок>>... а вот, пригодился!

\newpage
\huge \textbf{\textcolor{ChadBlue}{Литература}}
\normalsize
\bibliography{DRL}

\newpage
\thispagestyle{empty}
\begin{center}
\vspace*{5cm}
\includegraphics[width=10cm]{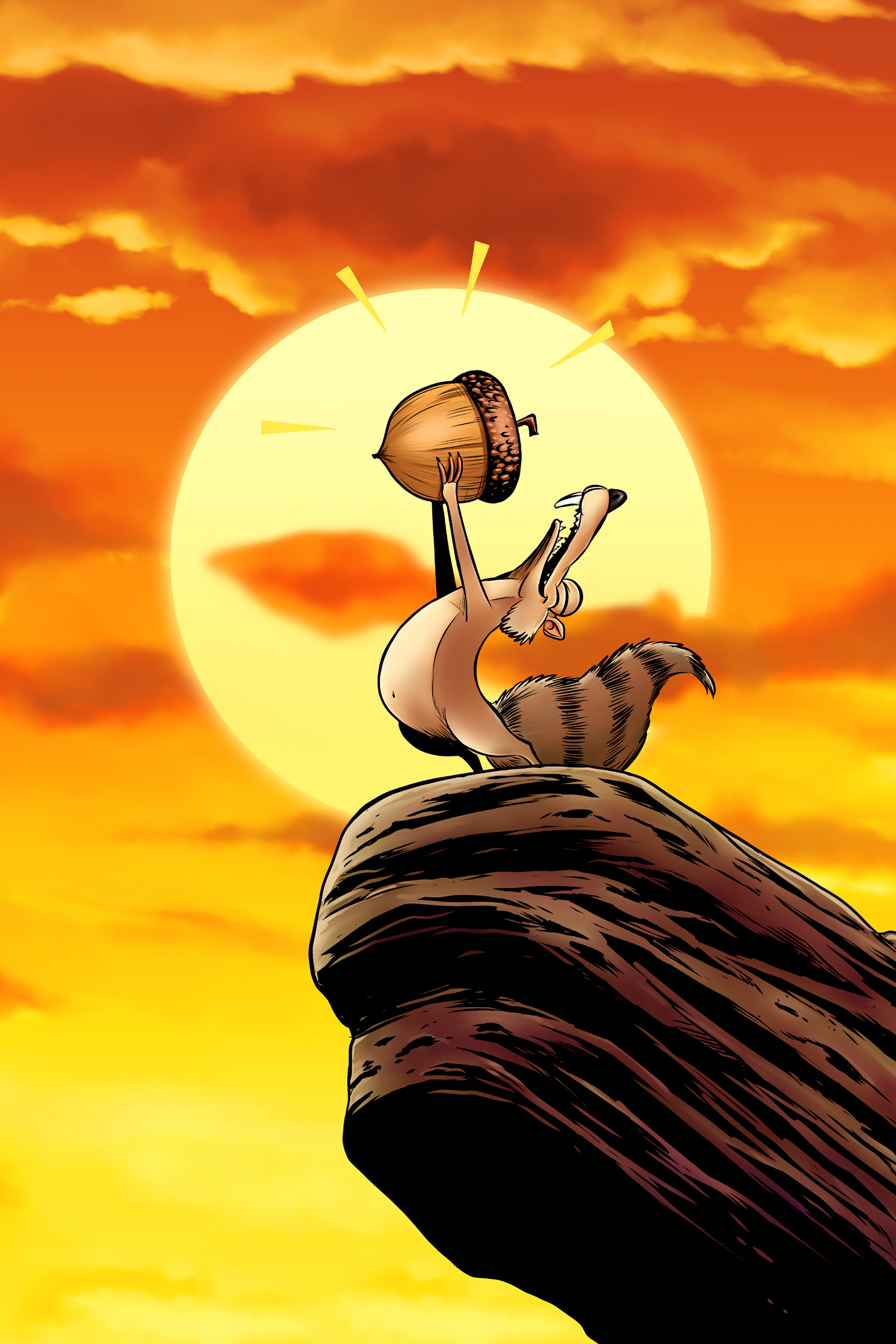}
\end{center}
\end{document}